\documentclass{article}
\usepackage{multirow}
\usepackage{microtype}
\usepackage{graphicx}
\usepackage[table,xcdraw]{xcolor}
\usepackage{subcaption}
\usepackage{booktabs} 
\usepackage{lipsum}
\usepackage{hyperref}

\usepackage{lmodern}

\usepackage[preprint]{icml2026}
\usepackage[utf8]{inputenc}
\usepackage[T1]{fontenc}

\usepackage{amsmath}
\usepackage{amssymb}
\usepackage{mathtools}
\usepackage{amsthm}

\usepackage{cuted}    
\usepackage{capt-of}
\usepackage{caption}

\usepackage[capitalize,noabbrev]{cleveref}

\theoremstyle{plain}

\theoremstyle{definition}

\theoremstyle{remark}

\def\our{DIAMOND}
\def\ourfine{\our{}-Fine}

\usepackage{microtype}      
\usepackage{xcolor}         
\usepackage[textsize=tiny]{todonotes}
\usepackage{soul}
\usepackage{url}

\icmltitlerunning{DIAMOND: Directed Inference for Artifact Mitigation in Flow Matching Models}

\usepackage{array}
\usepackage{graphicx}

\begin{document}


\twocolumn[
  \icmltitle{
  \our{}: Directed Inference for Artifact Mitigation in Flow Matching  Models
  }



  \icmlsetsymbol{equal}{*}

  \begin{icmlauthorlist}
   \icmlauthor{Alicja Polowczyk}{equal,yyy}
    \icmlauthor{Agnieszka Polowczyk}{equal,yyy}
    \icmlauthor{Piotr Borycki}{comp}
    \icmlauthor{Joanna Waczyńska}{comp}
    \icmlauthor{Jacek Tabor}{comp}
    \icmlauthor{Przemys\l{}aw Spurek}{comp,sch}
  \end{icmlauthorlist}

  \icmlaffiliation{yyy}{Silesian University of Technology}
  \icmlaffiliation{comp}{Jagiellonian University}
  \icmlaffiliation{sch}{IDEAS Research Institute}

  \icmlcorrespondingauthor{}{\{agnieszkapolowczyk11 alicjapolowczyk47\}@gmail.com}

  \icmlkeywords{Machine Learning, ICML}

  \vskip 0.3in
]

\begin{strip}
\centering
\includegraphics[width=0.9\linewidth]{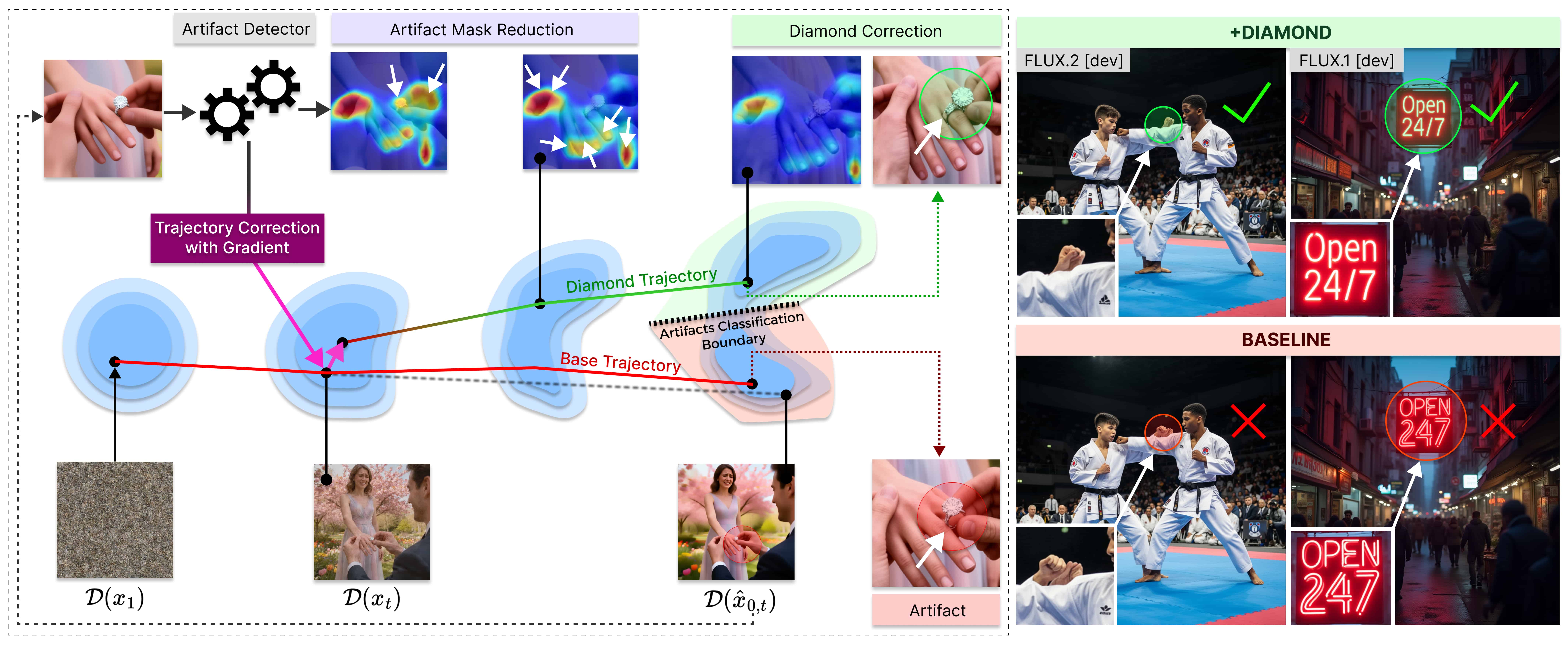}\\
    \captionof{figure}{We propose \textbf{\our{}}, an inference-time trajectory correction mechanism to mitigate artifacts in Rectified Flow and Diffusion Models. First, we generate an image $\mathcal{D}({x}_{1})$ from the initial probability distribution sample, which then undergoes the generative process. The Base Trajectory (red) leads to an image containing artifacts, whereas our corrected \textbf{\our{}} Trajectory (green) results in the artifact-free image. At timestep $t$, we estimate the final image $\hat{x}_{0,t}$ (gray dashed line) and use an Artifact Detector to apply a gradient-based trajectory correction (purple arrows), shifting it away from artifact region.}
    \label{fig:trajectorid}
\end{strip}



\printAffiliationsAndNotice{\icmlEqualContribution}  

\begin{abstract}

Despite impressive results from recent text-to-image models like FLUX, visual and anatomical artifacts remain a significant hurdle for practical and professional use.
Existing methods for artifact reduction, typically work in a post-hoc manner, consequently failing to intervene effectively during the core image formation process. Notably, current techniques require problematic and invasive modifications to the model weights, or depend on a computationally expensive and time-consuming process of regional refinement. To address these limitations, we propose \our{}, a training-free method that applies trajectory correction to mitigate artifacts during inference. By reconstructing an estimate of the clean sample at every step of the generative trajectory, \our{} actively steers the generation process away from latent states that lead to artifacts. Furthermore, we extend the proposed method to standard Diffusion Models, demonstrating that \our{} provides a robust, zero-shot path to high-fidelity, artifact-free image synthesis without the need for additional training or weight modifications in modern generative architectures. Code is available
at \url{https://gmum.github.io/DIAMOND/} 
\end{abstract}

\section{Introduction}


Generative text-to-image models~\cite{chang2023muse, ding2022cogview2, lu2023tf, malarz2025classifier} have transformed visual content creation by enabling photorealistic synthesis from natural language. Recent advancements in Rectified Flow architectures, such as FLUX.2, have established new benchmarks in synthesis quality and detail precision through deterministic trajectory modeling. Despite these achievements, generated images frequently exhibit structural and anatomical artifacts, posing a significant challenge for professional and practical applications. For example, in e-commerce or marketing visuals, a generated hand with six fingers instead of five, might make the image unsuited for product promotion.

Research has introduced various strategies to mitigate these issues. However, current state-of-the-art methods face critical bottlenecks. One research direction is to fine-tune the model by enhancing suspicious regions during or after image generation \cite{wang2025diffdoctor, xing2025focus}, which improves image fidelity but is computationally very expensive and therefore harder to apply in real-time or large-scale scenarios. Another line of work focuses on real-time artifact detection by analyzing irregularities during inference \cite{cao2025temporal}. Although more efficient, these solutions rely on heuristic thresholds, which may lead to inaccurate artifact localization. Moreover, the correction performed by randomly replacing the latent in the identified regions, rather than applying non-artifact correction.



\begin{figure}[t]
\centering
\setlength{\tabcolsep}{1pt}
\renewcommand{\arraystretch}{0.8}
\begin{tabular}{ccc}
 & Baseline &  +\our{}  \\
\rotatebox{90}{\scriptsize\hspace{8pt}{FLUX.2 [dev]}} &
\includegraphics[width=0.235\textwidth]{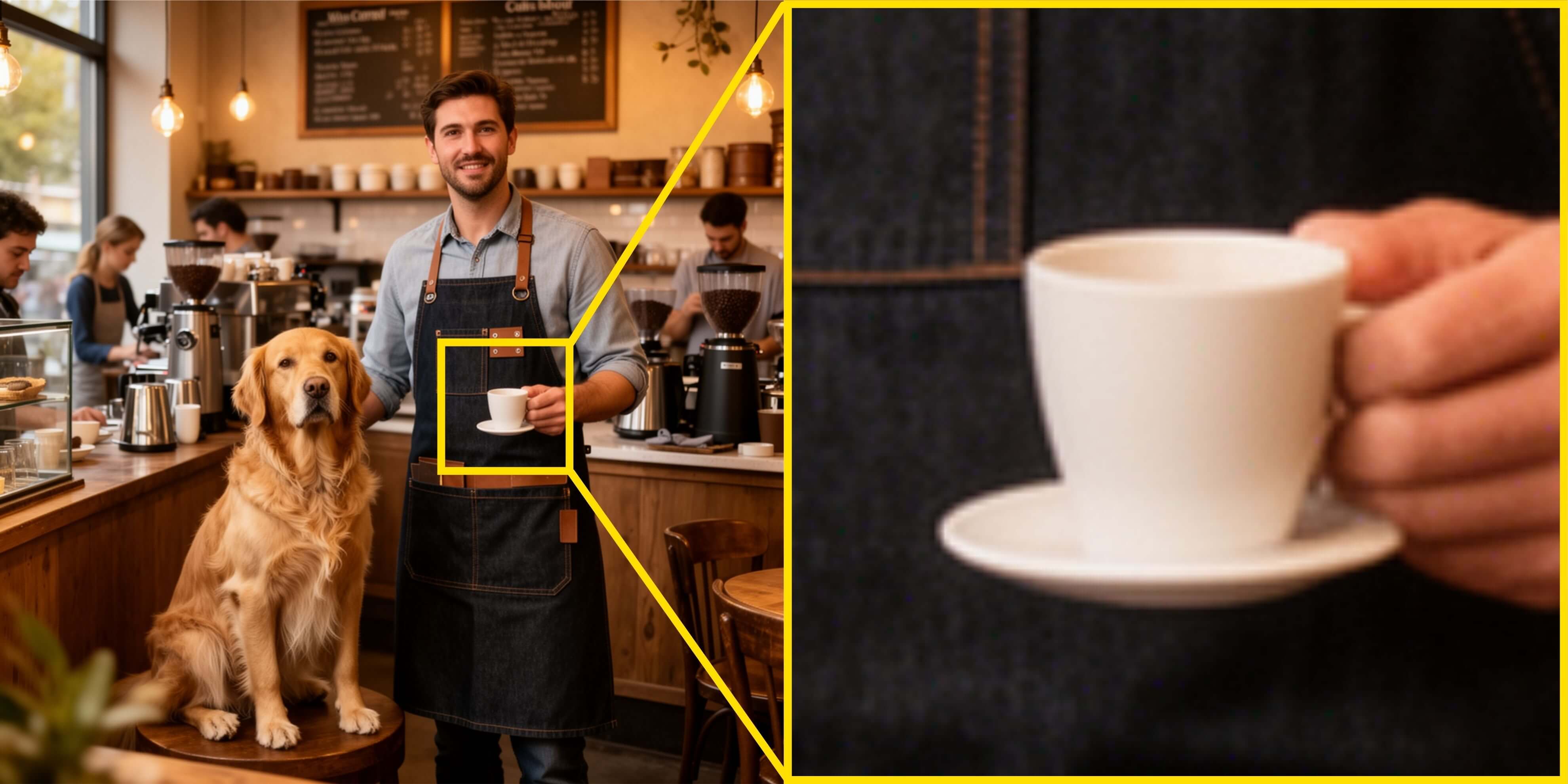} &
\includegraphics[width=0.235\textwidth]{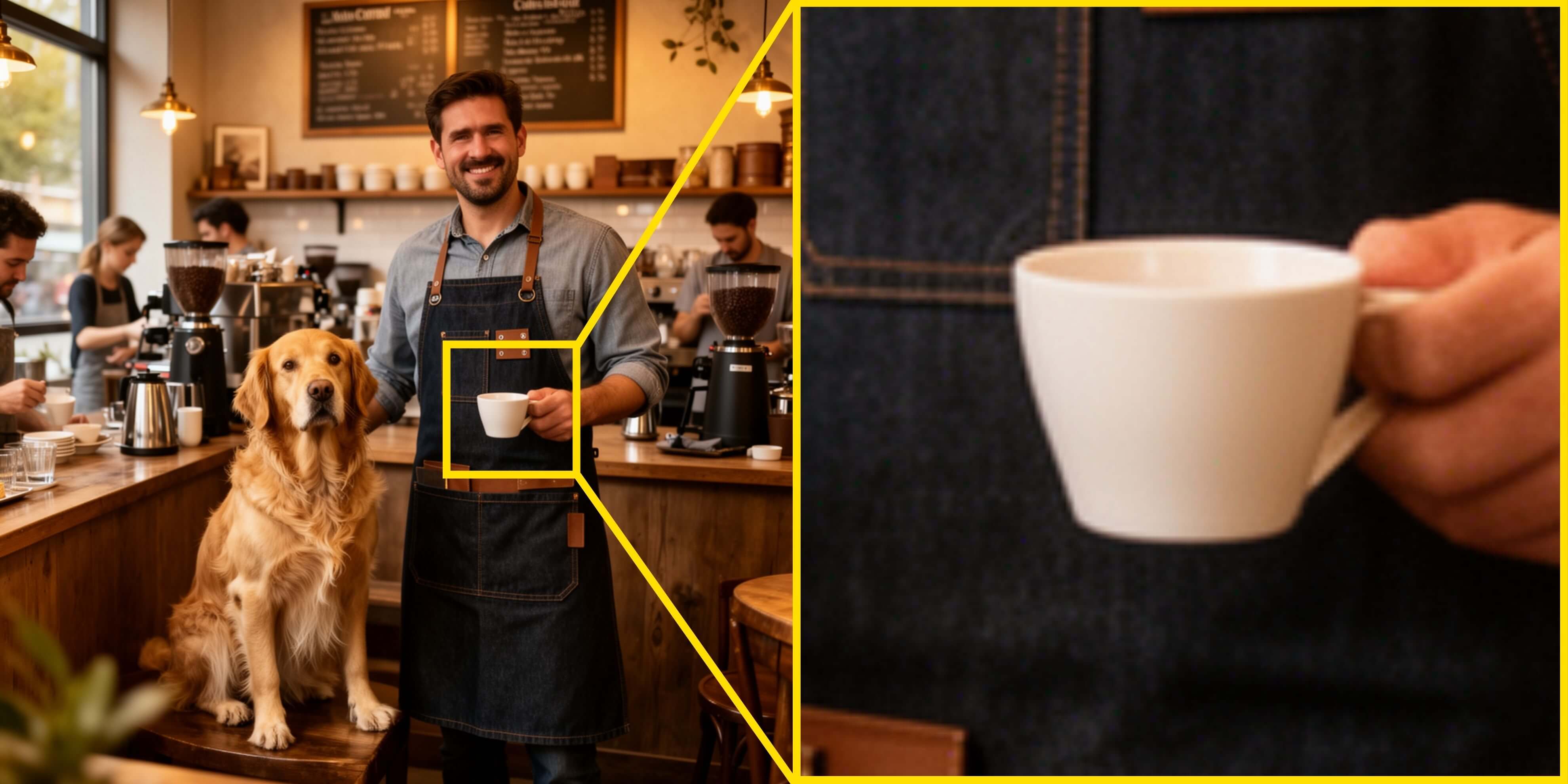} \\
\rotatebox{90}{\scriptsize\hspace{6pt}{ FLUX.1 [dev]}} &
\includegraphics[width=0.235\textwidth]{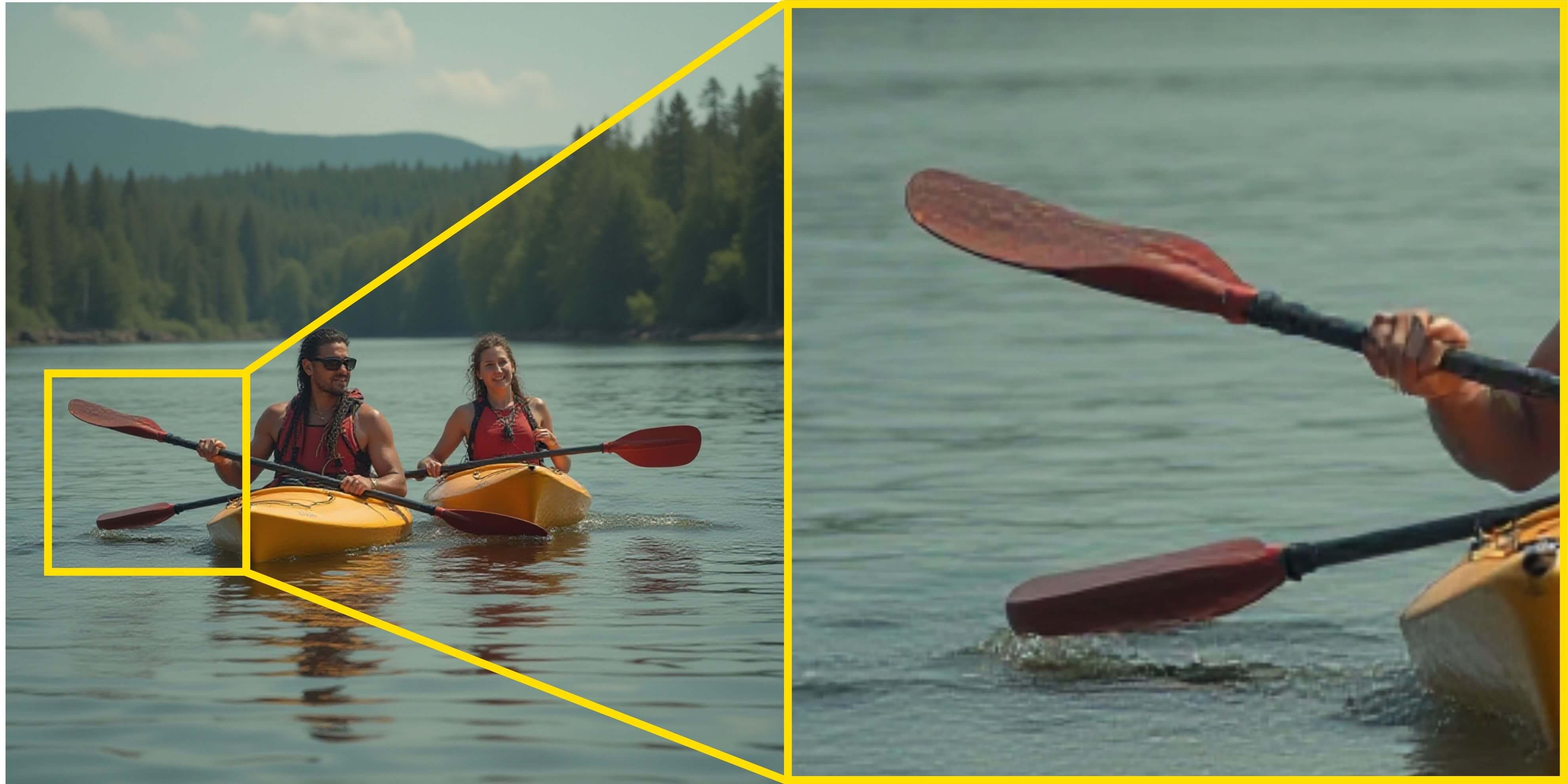} &
\includegraphics[width=0.235\textwidth]{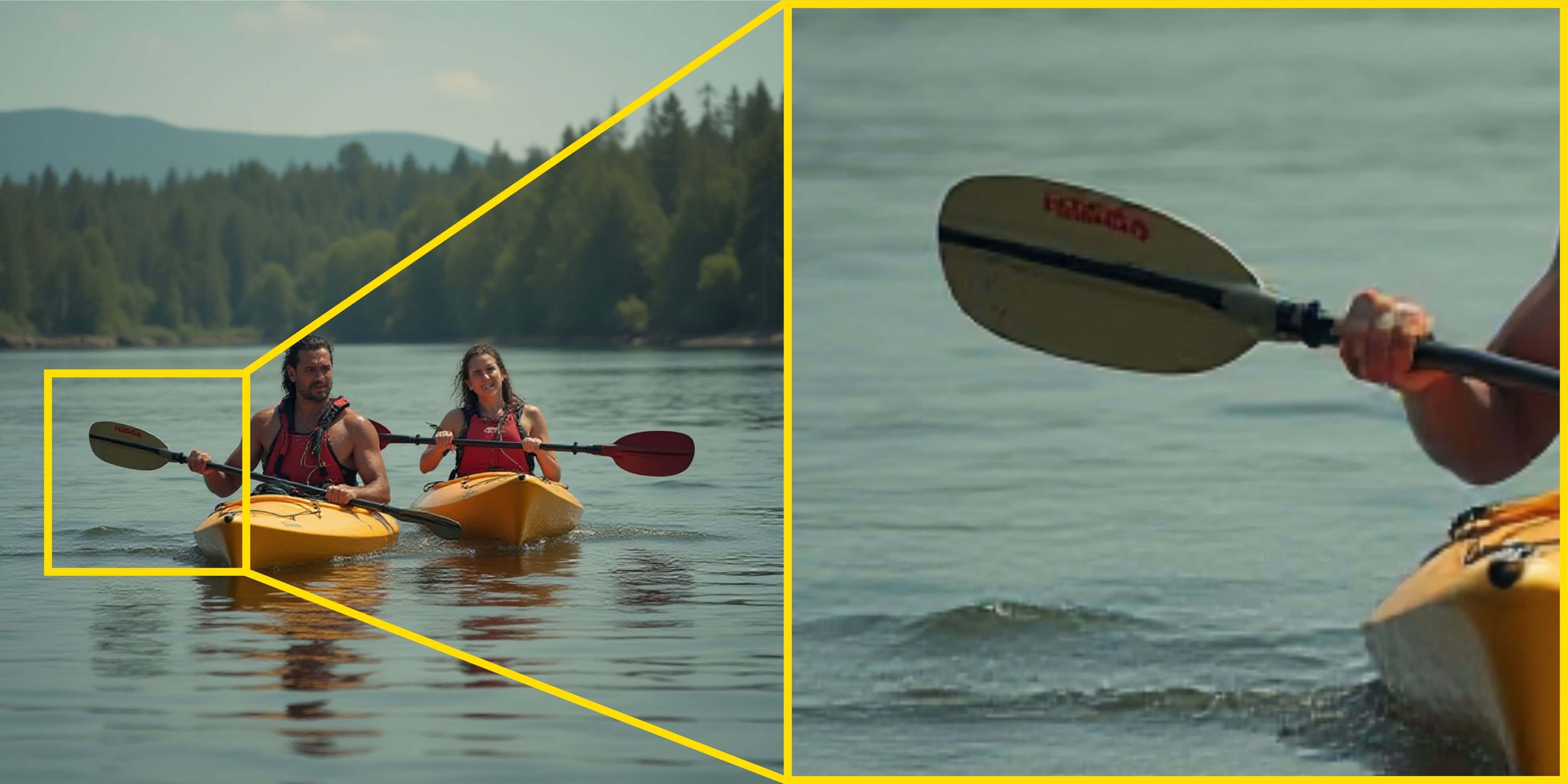} \\
\rotatebox{90}{\scriptsize\hspace{3pt}{FLUX.1 [schnell]}} &
\includegraphics[width=0.235\textwidth]{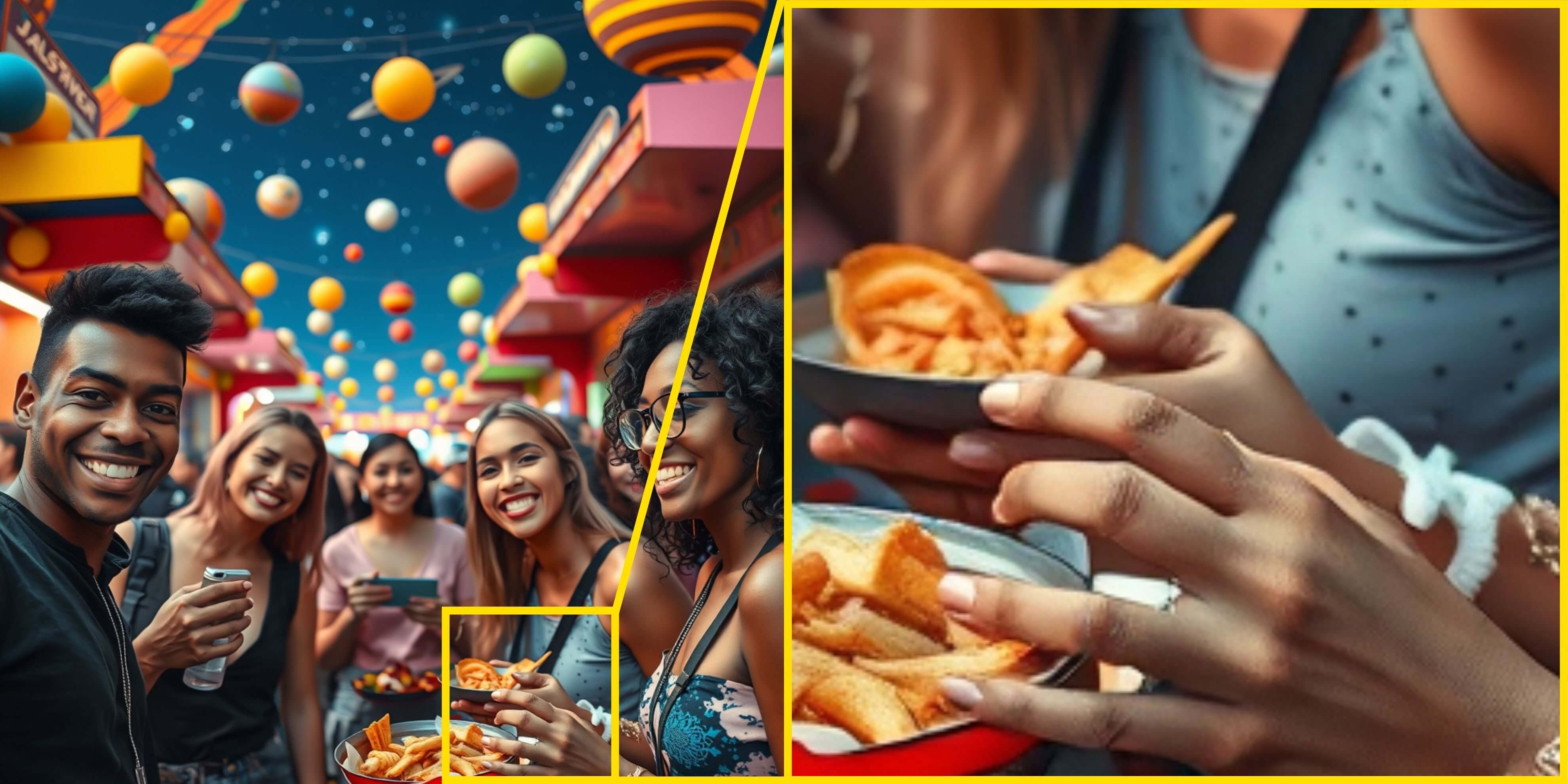} &
\includegraphics[width=0.235\textwidth]{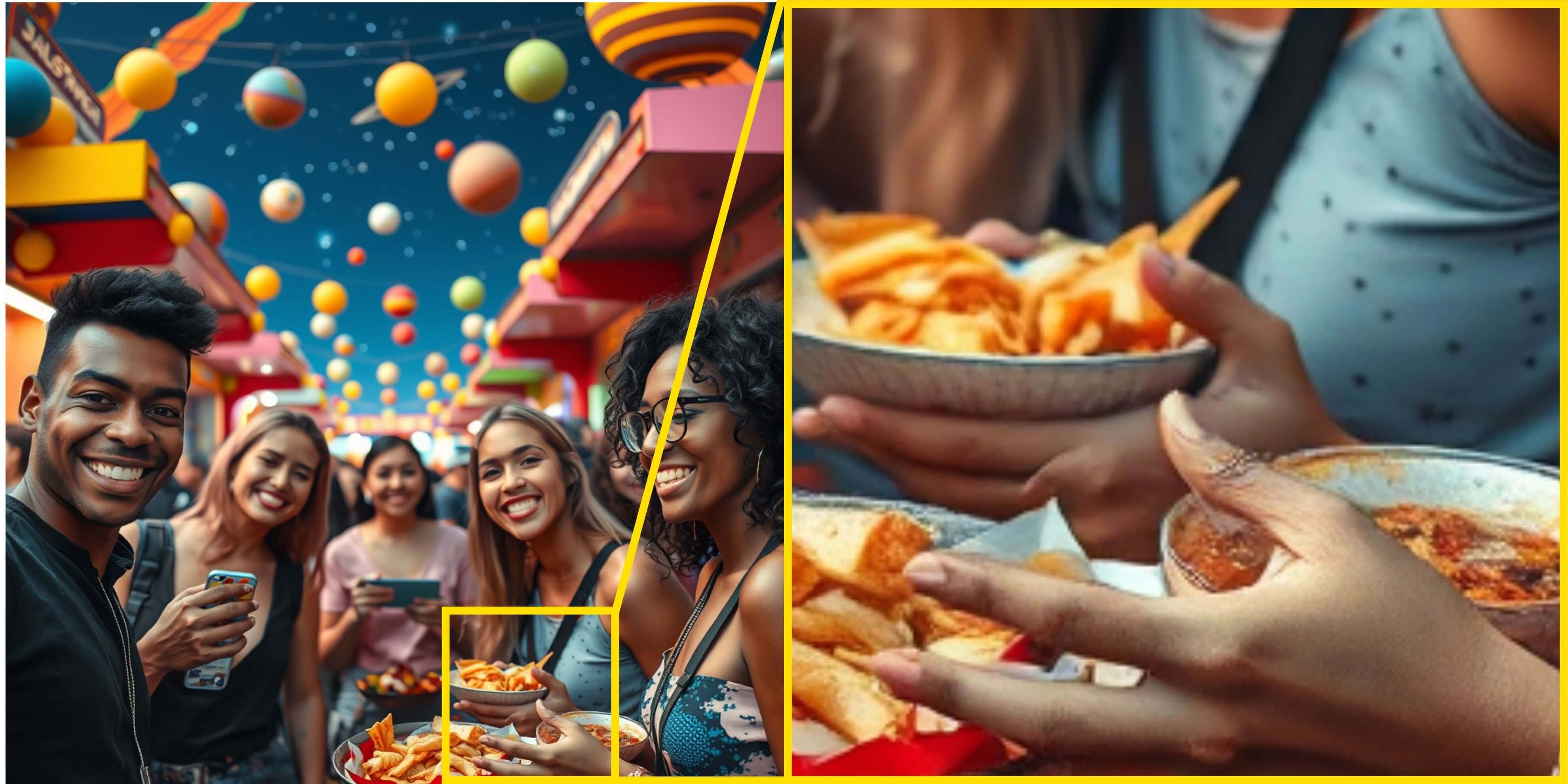} \\
\rotatebox{90}{\scriptsize\hspace{10pt}{ \quad SDXL}} &
\includegraphics[width=0.235\textwidth]{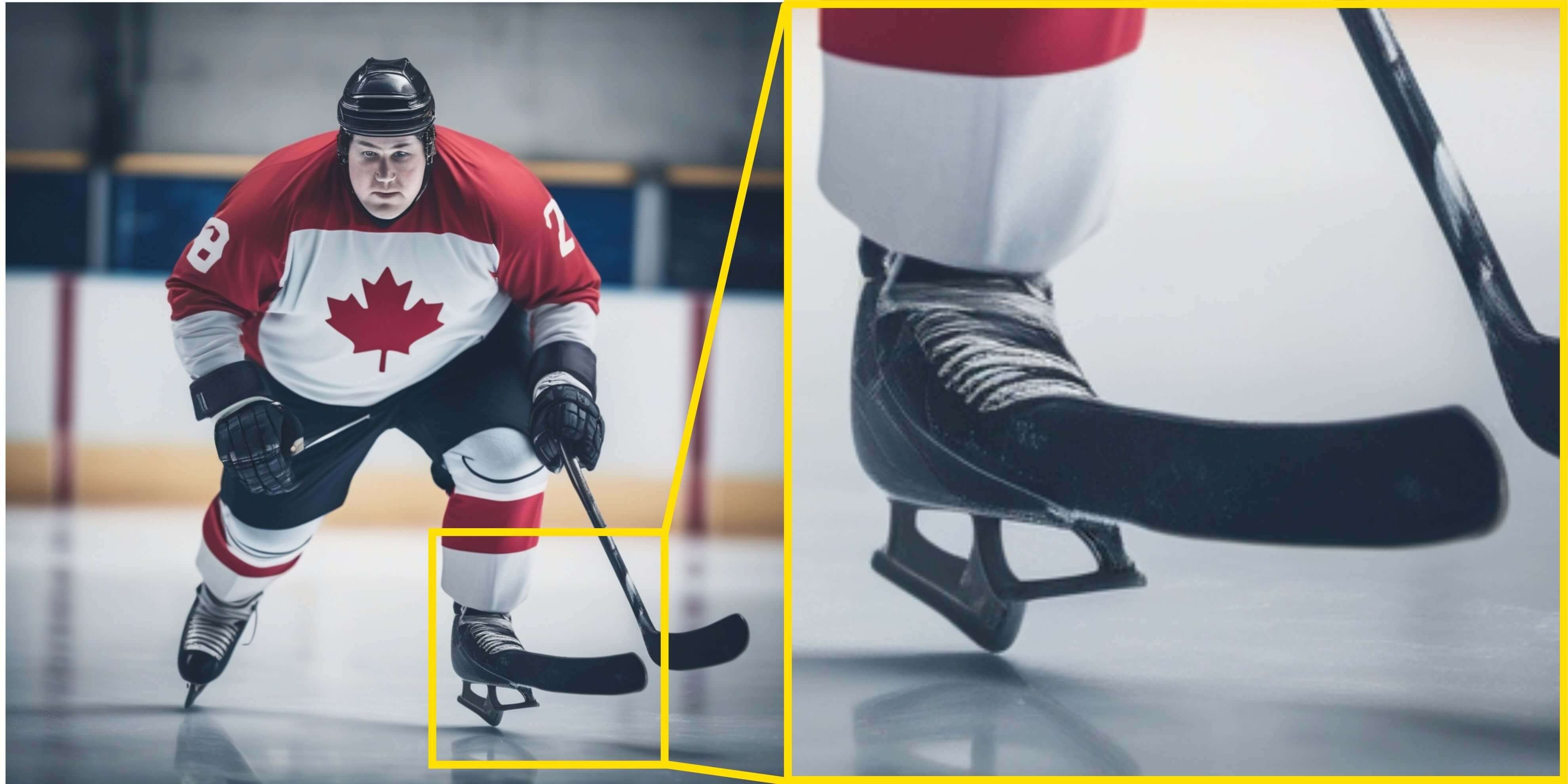} &
\includegraphics[width=0.235\textwidth]{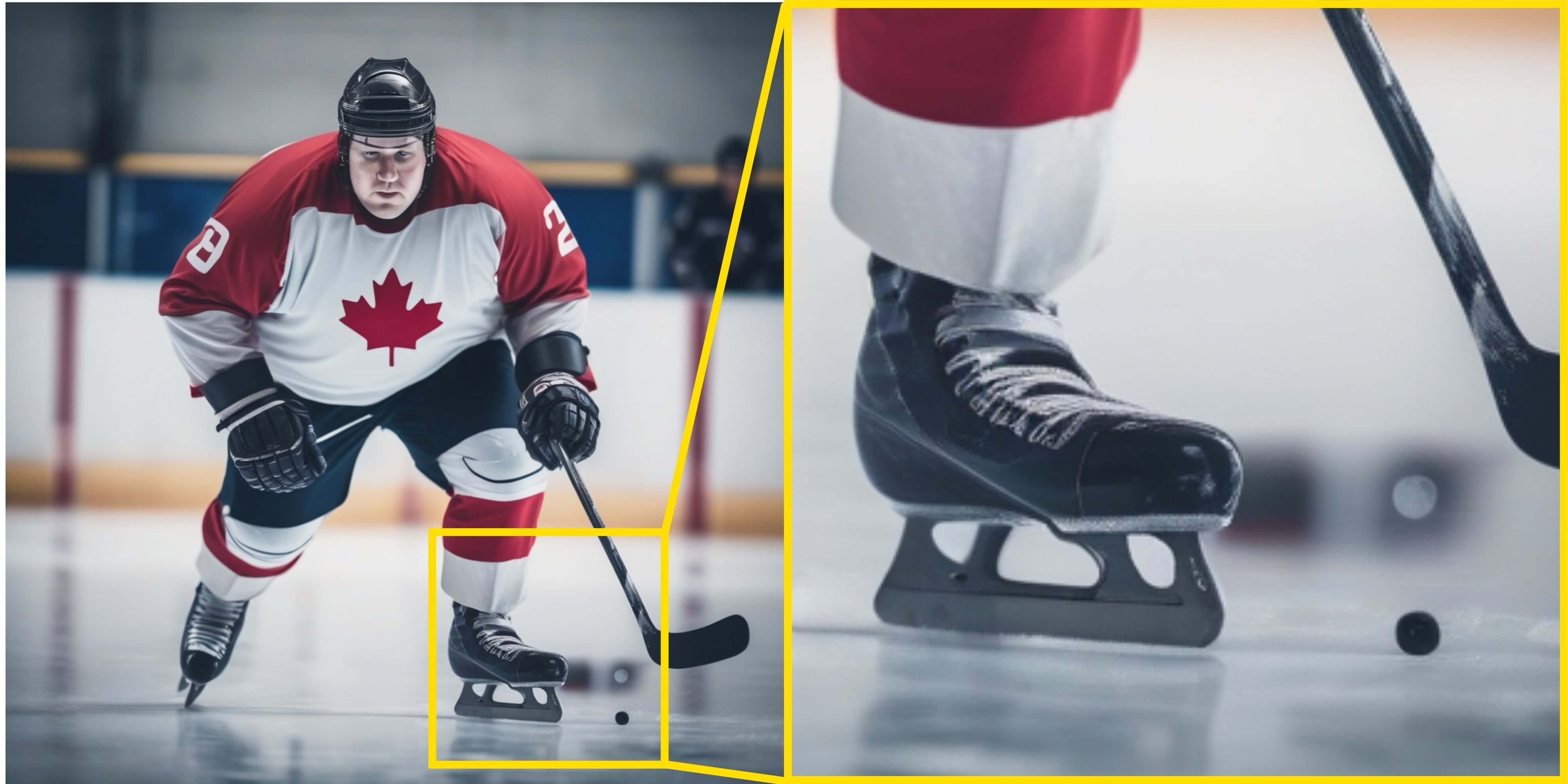} 
\end{tabular}
\caption{\textbf{Artifact generation in FLUX.2, FLUX.1 and SDXL, with corrections applied using \our{}.} While these models produce visible artifacts, \our{} detects and corrects them automatically without additional training.
}
\label{fig:comparison}
\vspace{-3mm}
\end{figure}

\begin{figure*}[t]
    \centering
    \includegraphics[width=\linewidth]{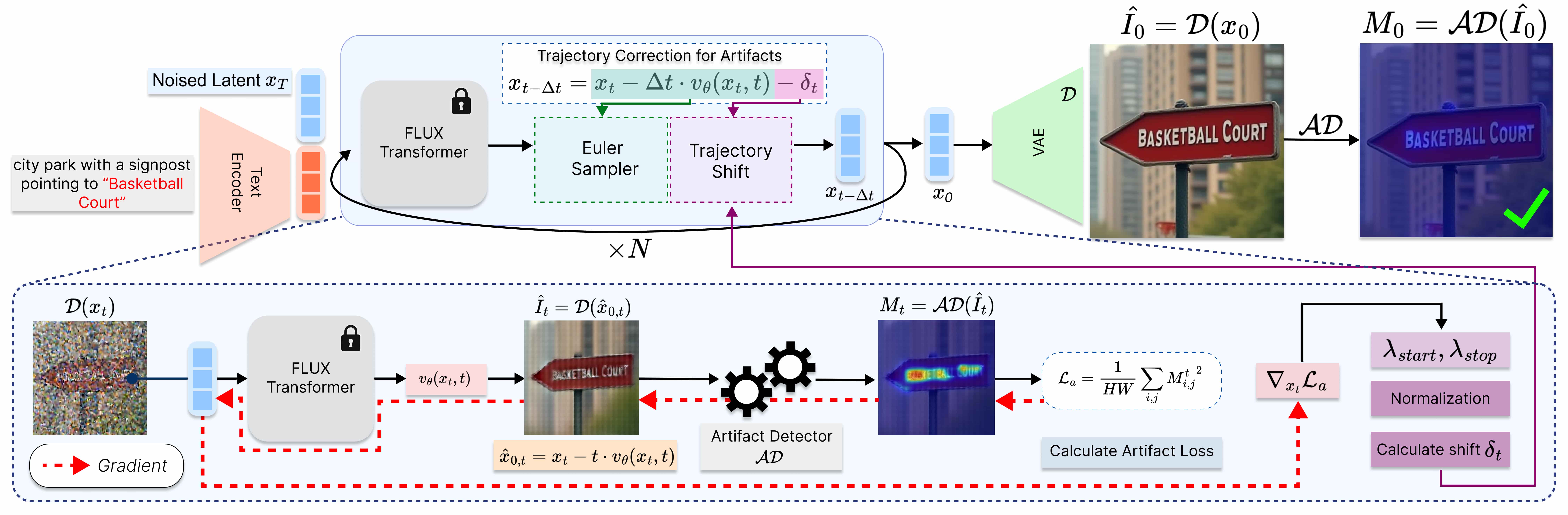}
    \caption{\textbf{Overview of \our{}}. Our technique employs an inference-time pipeline specifically designed to mitigate artifacts without requiring additional training. Artifact suppression is achieved by correcting the trajectory using gradients derived from a pixel-wise segmentation loss. During inference, the latent representation is iteratively updated using controlled shifts along the trajectory, enabling progressive correction of artifact-prone regions.}
    \label{fig:pipeline}
    \vspace{-1mm}
\end{figure*}

Overall, current research either focuses on improving image generation using fine-tuning or chooses lightweight but less precise modifications applied during the generation process. To address these constraints, we introduce \our{} Directed Inference for Artifact Mitigation in Flow Matching Models) which integrates geometric trajectory guidance directly into the inference process, see Fig. \ref{fig:trajectorid}.
Our method reconstructs an estimate of the clean image at each timestep of the generation trajectory, based on the predicted velocity field and noise levels for Rectified Flow and Diffusion models, respectively. The image corresponding to the clean estimate enables the calculation of gradients using a differentiable Artifact Detector, which can be used to steer the trajectory away from latent states associated with artifacts. We employ segmentation-based guidance by incorporating an auxiliary pixel-wise loss in each time interval in the generation process, enabling fine-grained spatial supervision and precise correction of localized artifacts within selected image regions. Notably, this process works without requiring additional training or LoRA fine-tuning. To ensure optimal performance across different models, we utilize a dynamic scheduling mechanism to scale the guidance intensity. 

Quantitative results on FLUX.1 [dev] show that our approach drastically reduces the mean artifact frequency from a 100\% baseline to less than 10\% on \textit{words} dataset. Furthermore, \our{} is also applicable to other Flow Matching and Diffusion models, achieving a significantly lower artifact pixel ratio compared to competing baselines, providing a robust, zero-shot path to high-fidelity synthesis, see Fig.~\ref{fig:comparison}.


Our main contributions are as follows:
\begin{itemize}
\vspace{-0.3cm}
\item We introduce \our{}, a zero-shot, training-free framework for removing visual artifacts in Rectified Flow models without modifications to model weights.
\vspace{-0.3cm}
\item We propose a trajectory guidance mechanism that leverages on-the-fly clean image 
estimation  to steer the velocity field during the inference process.
\vspace{-0.3cm}
\item  We demonstrate that \our{} reduces the mean artifact frequency on DiffDoctor benchmark and applies to both diffusion and flow-matching models.
\vspace{-0.3cm}
\end{itemize}

 \section{Related Works}
The evaluation and refinement of generative models have led to various frameworks for measuring and improving synthesis quality. Preference-based scorers such as ImageReward \cite{xu2023imagereward} provides a general-purpose reward model to align text to image outputs with human aesthetic and semantic preferences. Similarly, PickScore \cite{kirstain2023pick} leverages community-driven preference data to score and select images that best reflect user intent. While they can be used to filter out generated images, these methods lack the ability to repair the model's outputs.

To address specific anatomical failures, specialized tools like HandRefiner \cite{lu2024handrefiner} and HandCraft \cite{qin2025handcraft} employ structural priors to correct hand distortions. Diagnostic approaches such as PAL \cite{zhang2023perceptual} focus on identifying perceptual flaws, while SAD \cite{menn2025synthesizing} synthesizes artefacts to enhance the training of error detection classifiers. Statistical and temporal consistency are further analyzed by BayesDiff \cite{kou2024bayesdiff}, which uses a Bayesian framework to estimate denoising uncertainty, and ASCED \cite{cao2025temporal}, which monitors stability throughout the generation process. 

Automated refinement loops, including SARGD \cite{zheng2024self}, Self-Refining \cite{lee2025refining}, and SLD~\cite{wu2024self},~utilize iterative feedback to improve visual fidelity. Among existing repair strategies, \mbox{DiffDoctor~\cite{wang2025diffdoctor}~diagnoses} and corrects artifacts during synthesis. However, its practical application to modern rectified flow models is limited by the absence of publicly available LoRA weights. Finally, Focus-N-Fix \cite{xing2025focus} addresses local artifacts through regional fine-tuning. This approach remains computationally expensive and sensitive to configuration discrepancies, compared to our proposed inference-time guidance.



\section{\our{}}
This section introduces our \our{} method for inference-time artifact removal in Rectified Flow models. The goal of \our{} is the mitigation of occasional visual anomalies produced by generative models. An overview of the method is presented in Fig. \ref{fig:pipeline}. The method is based on a differentiable, pretrained Artifact Detector ($\mathcal{AD}$) \cite{wang2025diffdoctor,lu2024handrefiner,qin2025handcraft} that identifies artifact regions during generation. This enables immediate intervention during sampling, enabling on-the-fly gradient-based corrections at each timestep \cite{Poleski_2025_WACV,dhariwal2021diffusion}. First we recall some preliminary information on Flow Matching. This is followed by the definition of an artifact and methodology related to their detection. Finally, the section ends with the full pipeline of \our{}.

\textbf{Flow Matching Models}  
Rectified Flow Transformers such as FLUX~\cite{flux2024} are based on a Flow Matching principle, which learns vector field $v_\theta:\mathbb{R}^D\times [0,1]\to \mathbb{R}^D$ that transports a probability distribution from a Gaussian noise $\pi_1=\mathcal{N}\left(0,\mathbb{I}\right)$ to the unknown data distribution~$\pi_0$. 

This process is defined by an Ordinary Differential Equation (ODE) that describes the trajectory of a latent variable $x_t$ as time flows from $t=1$ to~$t=0$: \(
     \dfrac{\mathrm{d}}{\mathrm{d}t} x_t =  v_\theta(x_t, t)
\).

\begin{figure*}[t]
\centering
\setlength{\tabcolsep}{1.2pt}
\renewcommand{\arraystretch}{0.9}
\begin{tabular}{ccccc}
FLUX.1 [dev] &  +DiffDoctor & +HPSv2 & +HandsXL & + \our{} \\
\includegraphics[width=0.19\textwidth]{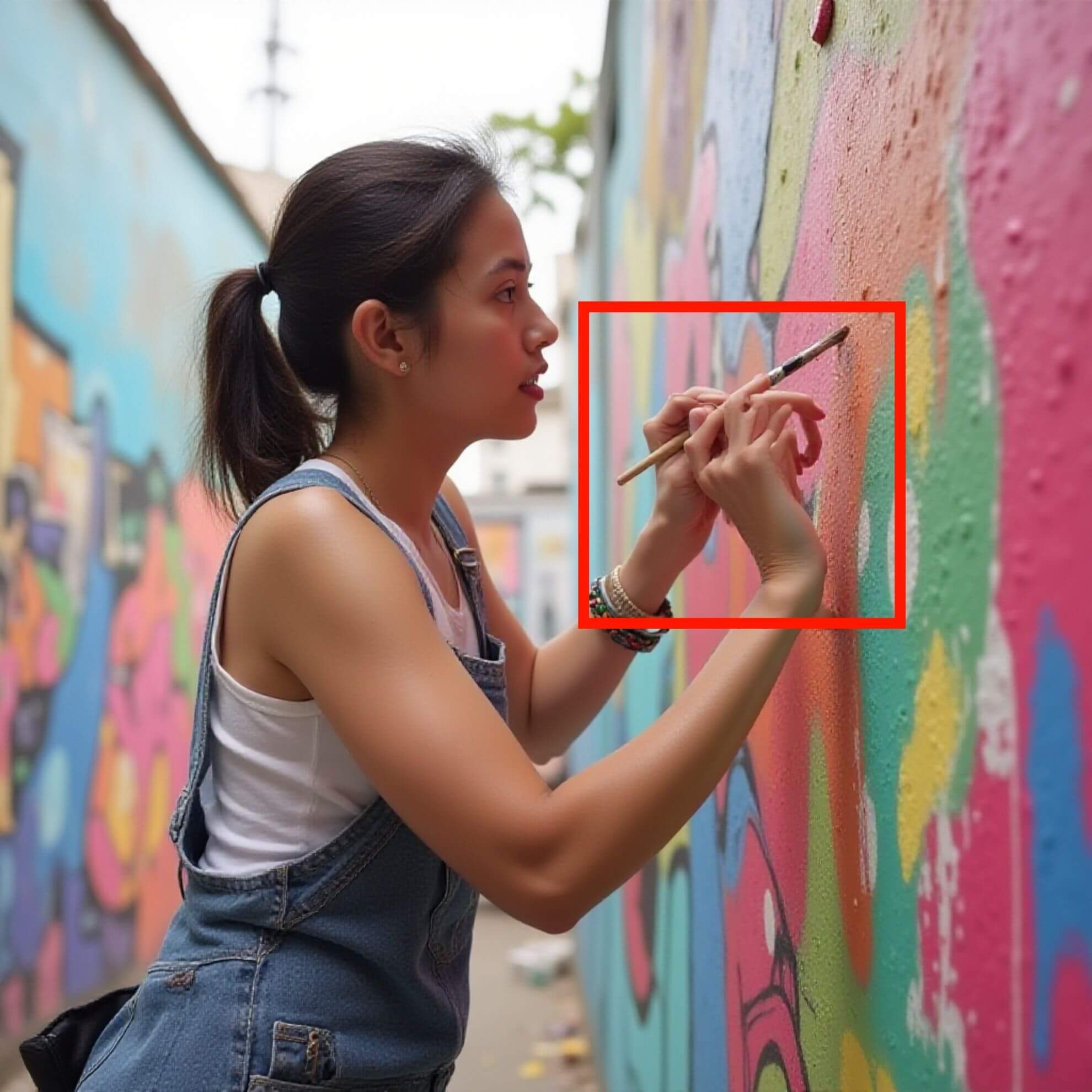} &
\includegraphics[width=0.19\textwidth]{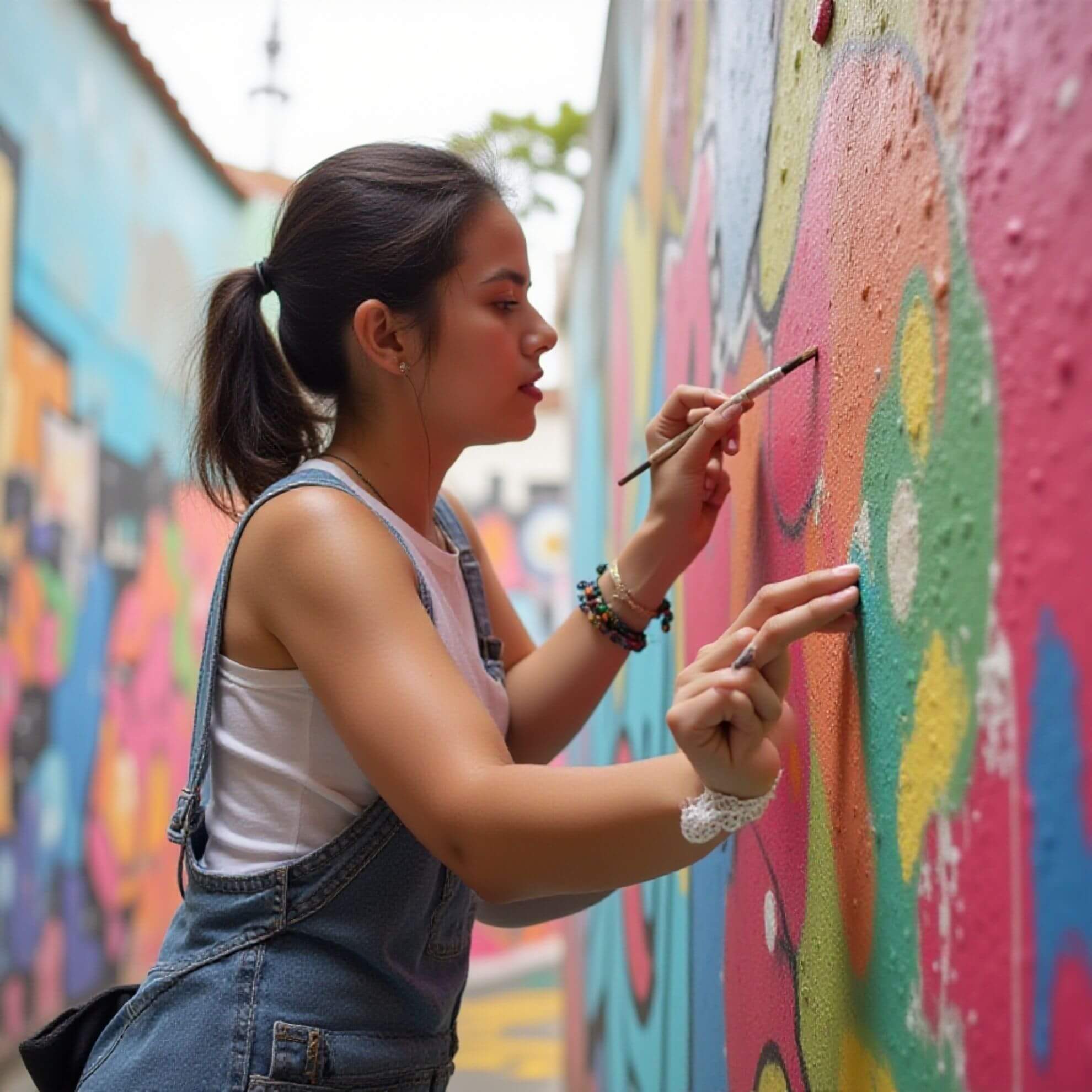} &
\includegraphics[width=0.19\textwidth]{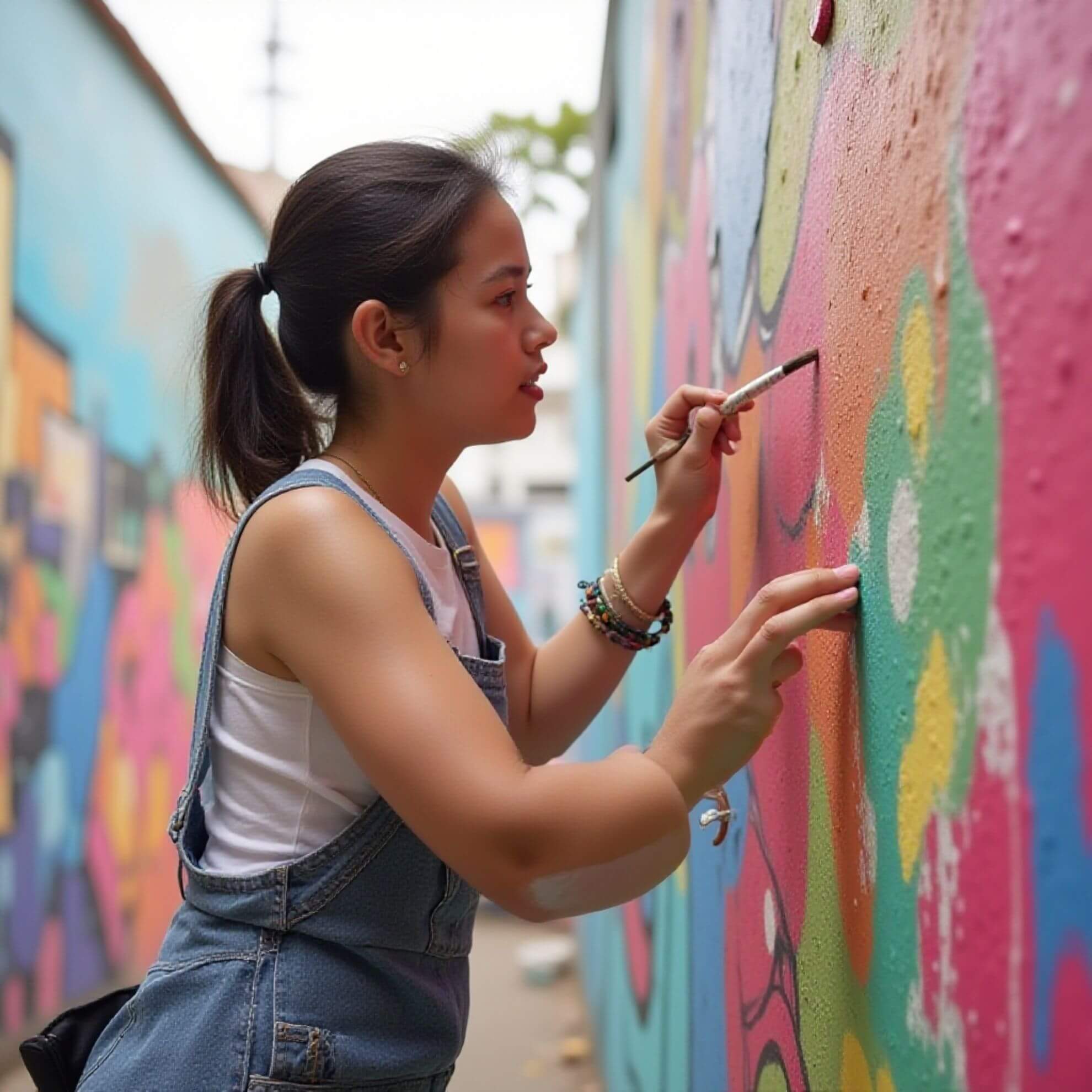} &
\includegraphics[width=0.19\textwidth]{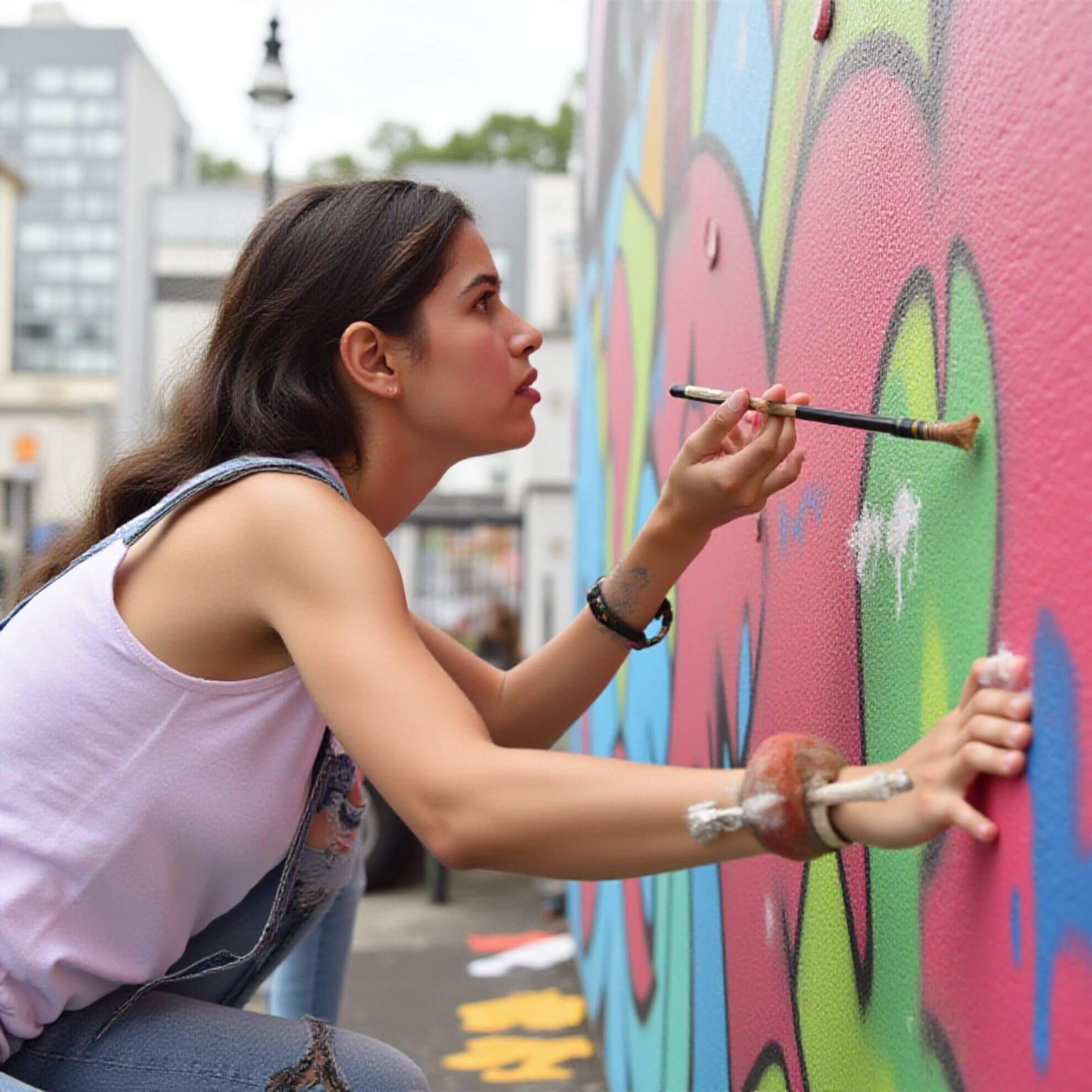} &
\includegraphics[width=0.19\textwidth]{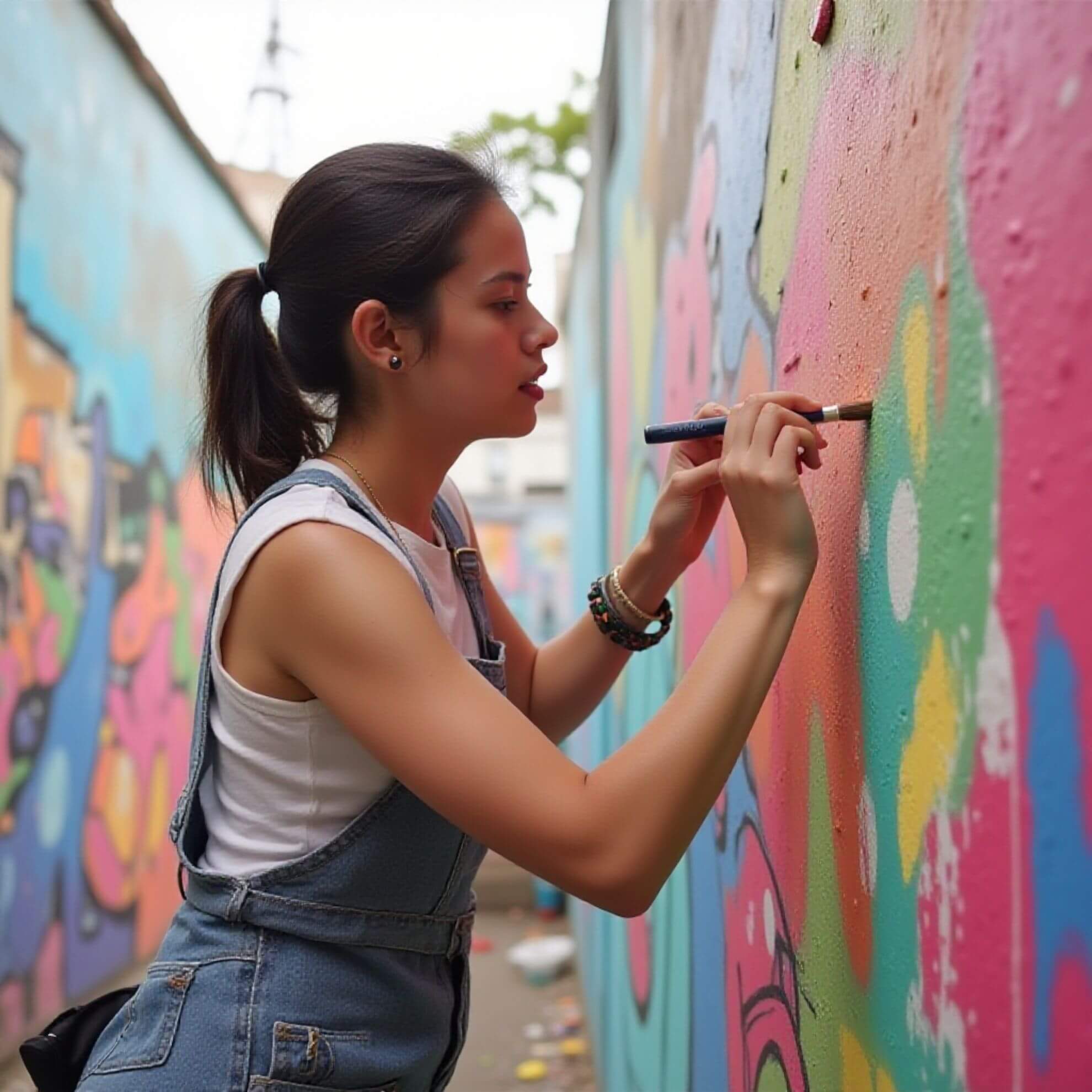} \\
\includegraphics[width=0.19\textwidth]{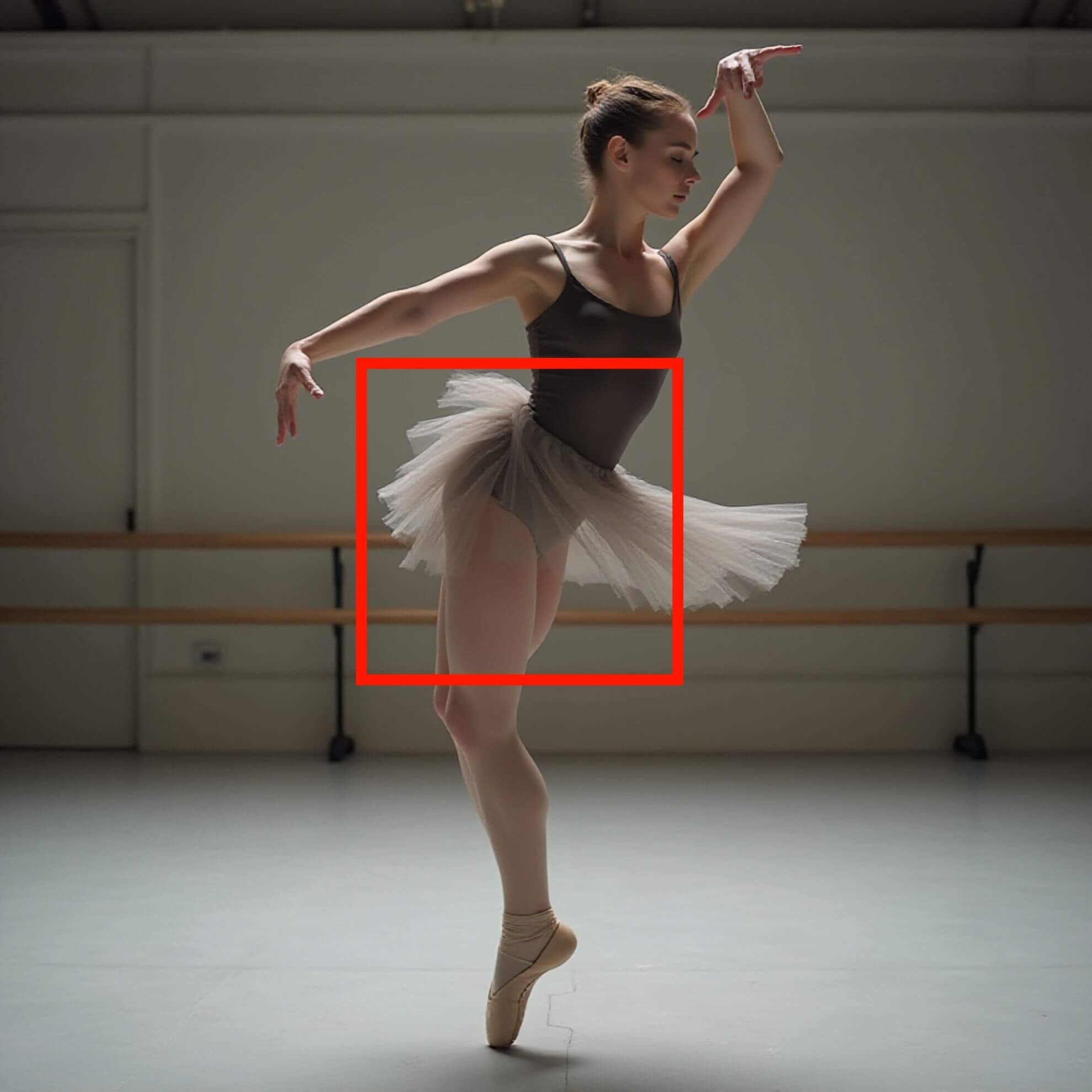} &
\includegraphics[width=0.19\textwidth]{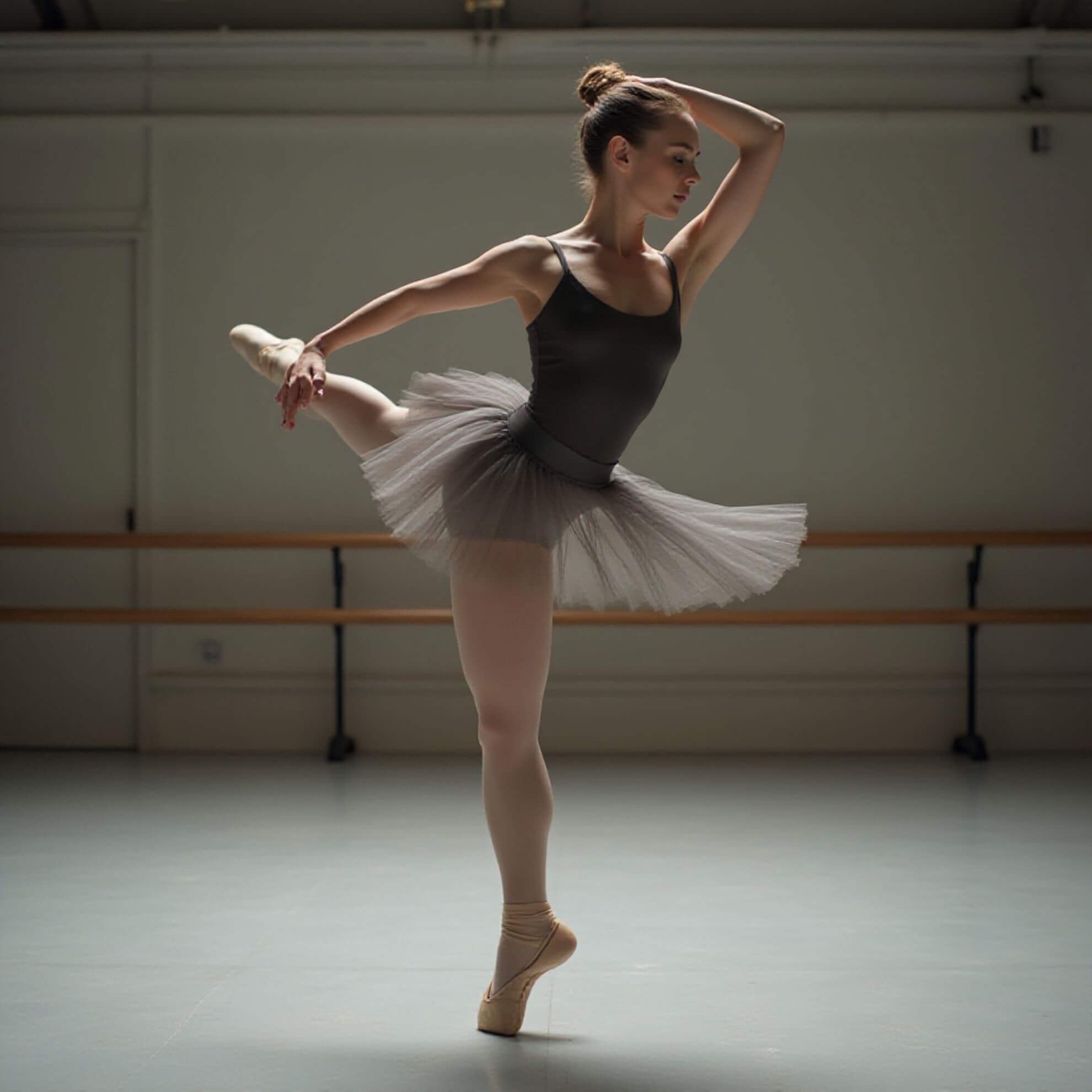} &
\includegraphics[width=0.19\textwidth]{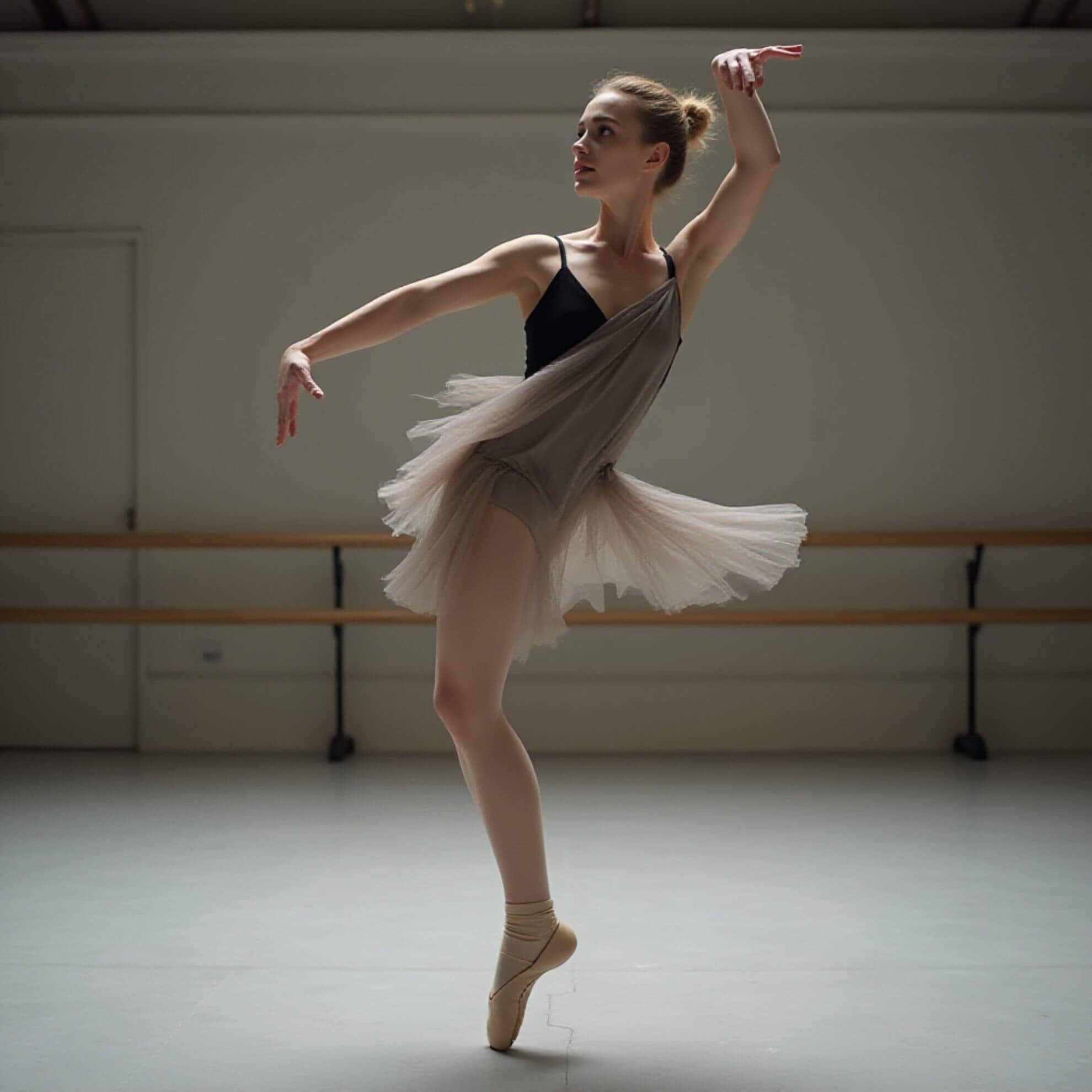}  &
\includegraphics[width=0.19\textwidth]{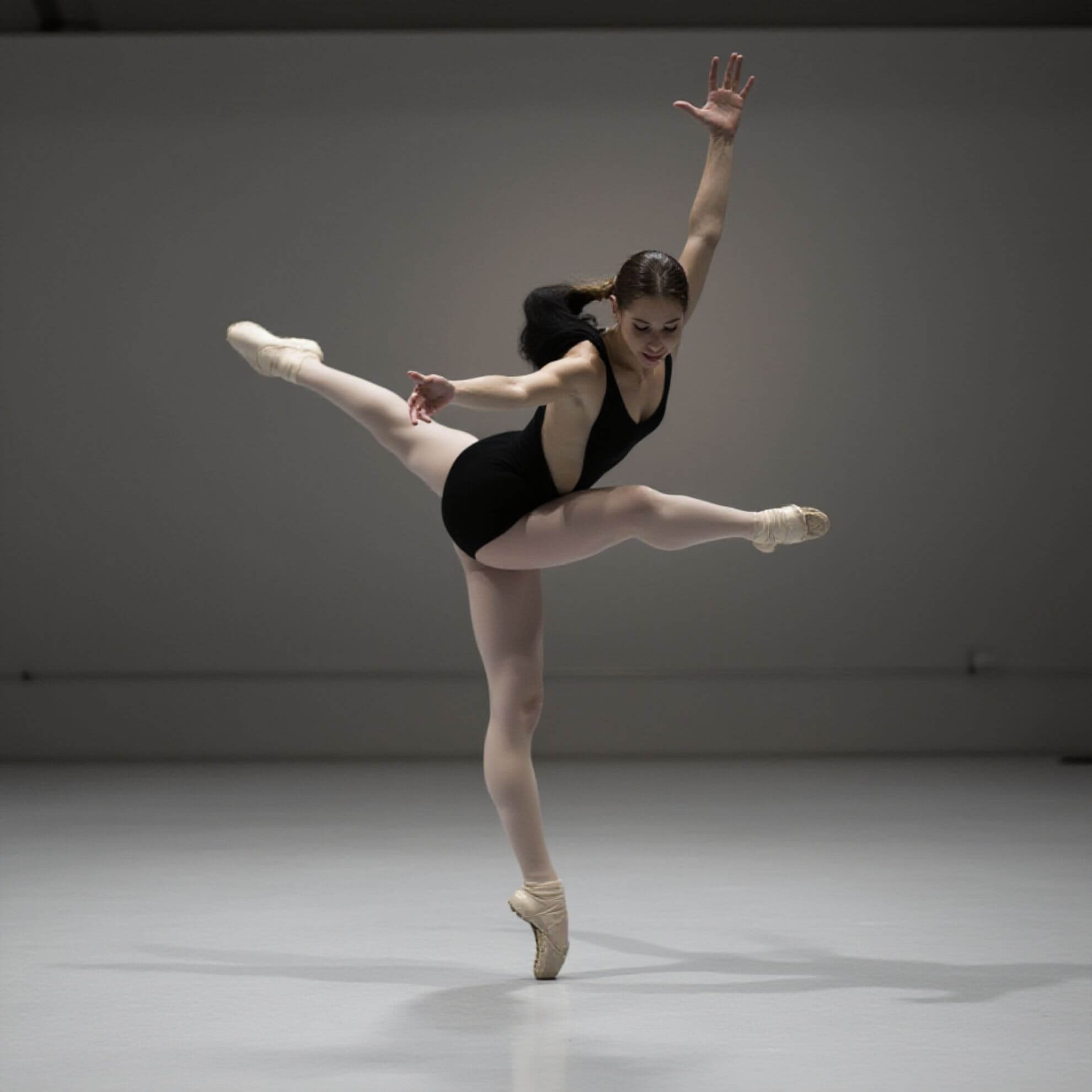} &
\includegraphics[width=0.19\textwidth]{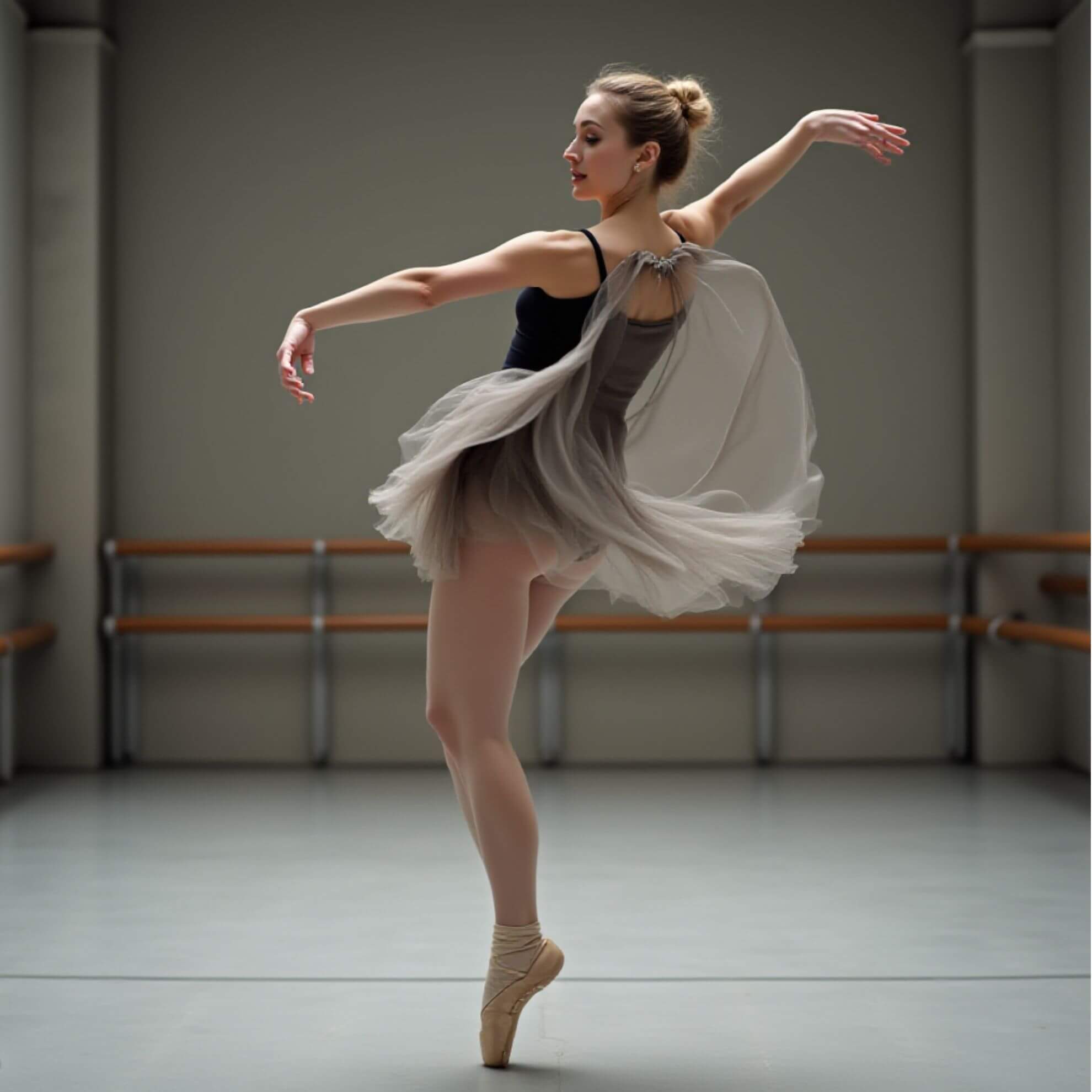} \\
\includegraphics[width=0.19\textwidth]{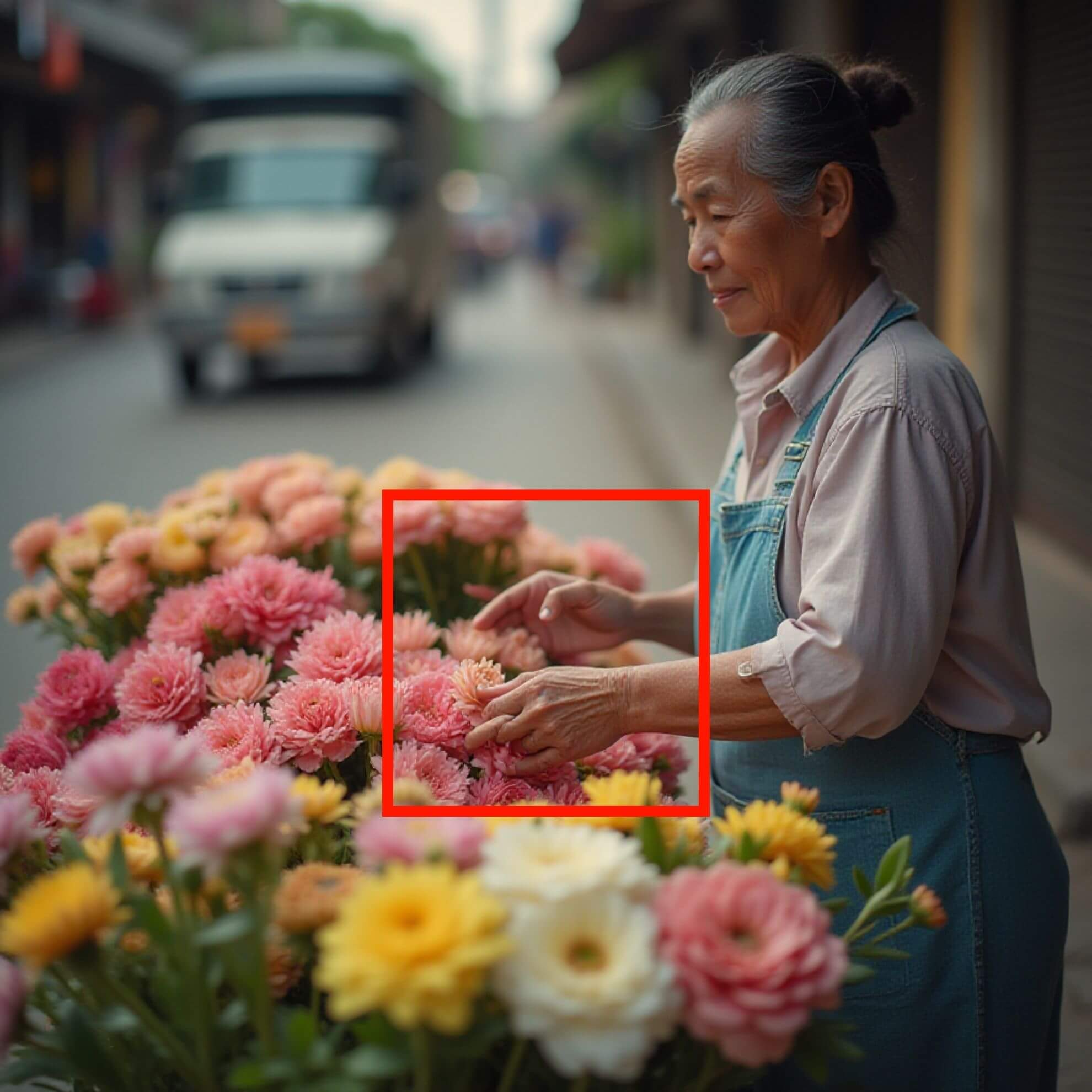} &
\includegraphics[width=0.19\textwidth]{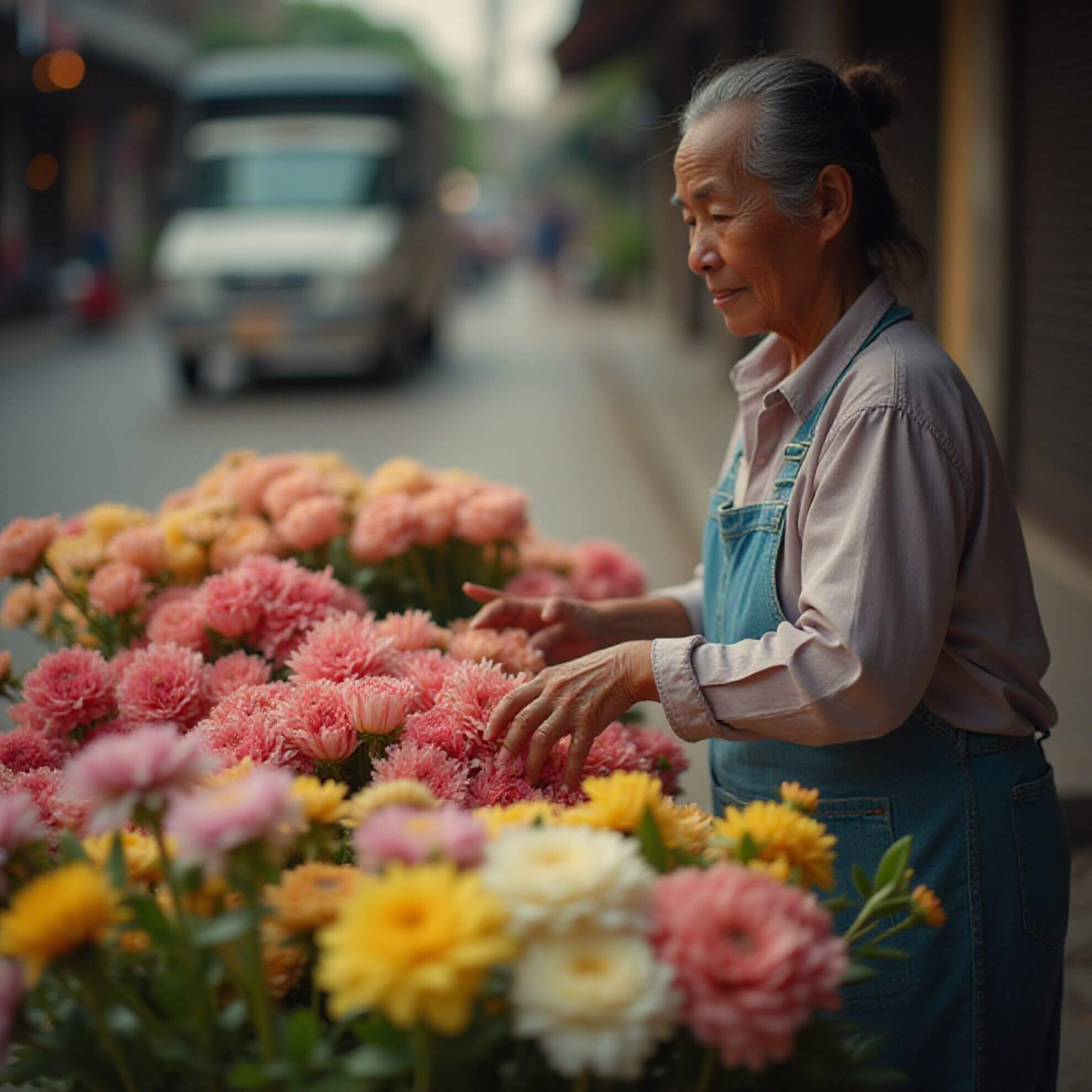}  &
\includegraphics[width=0.19\textwidth]{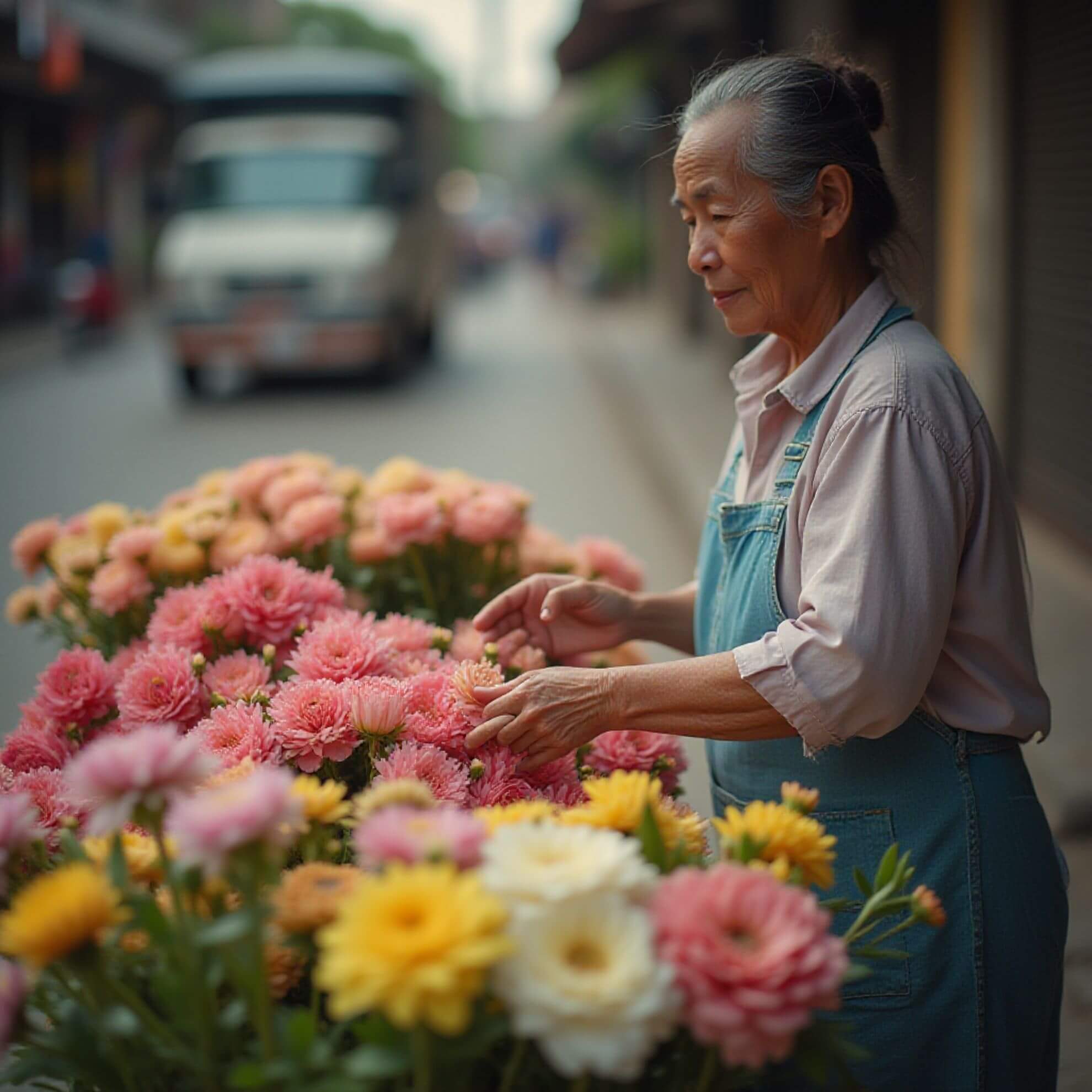} &
\includegraphics[width=0.19\textwidth]{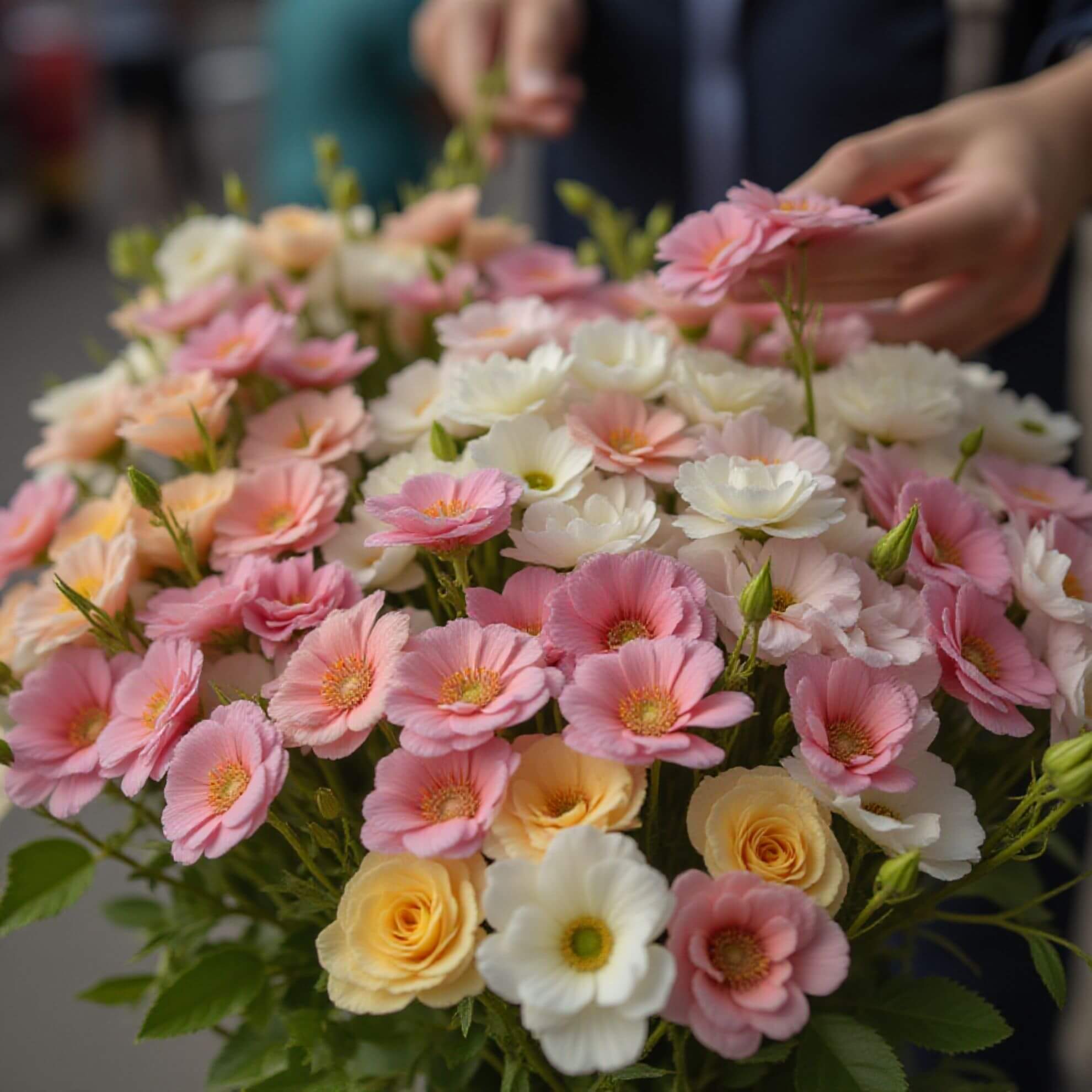} &
\includegraphics[width=0.19\textwidth]{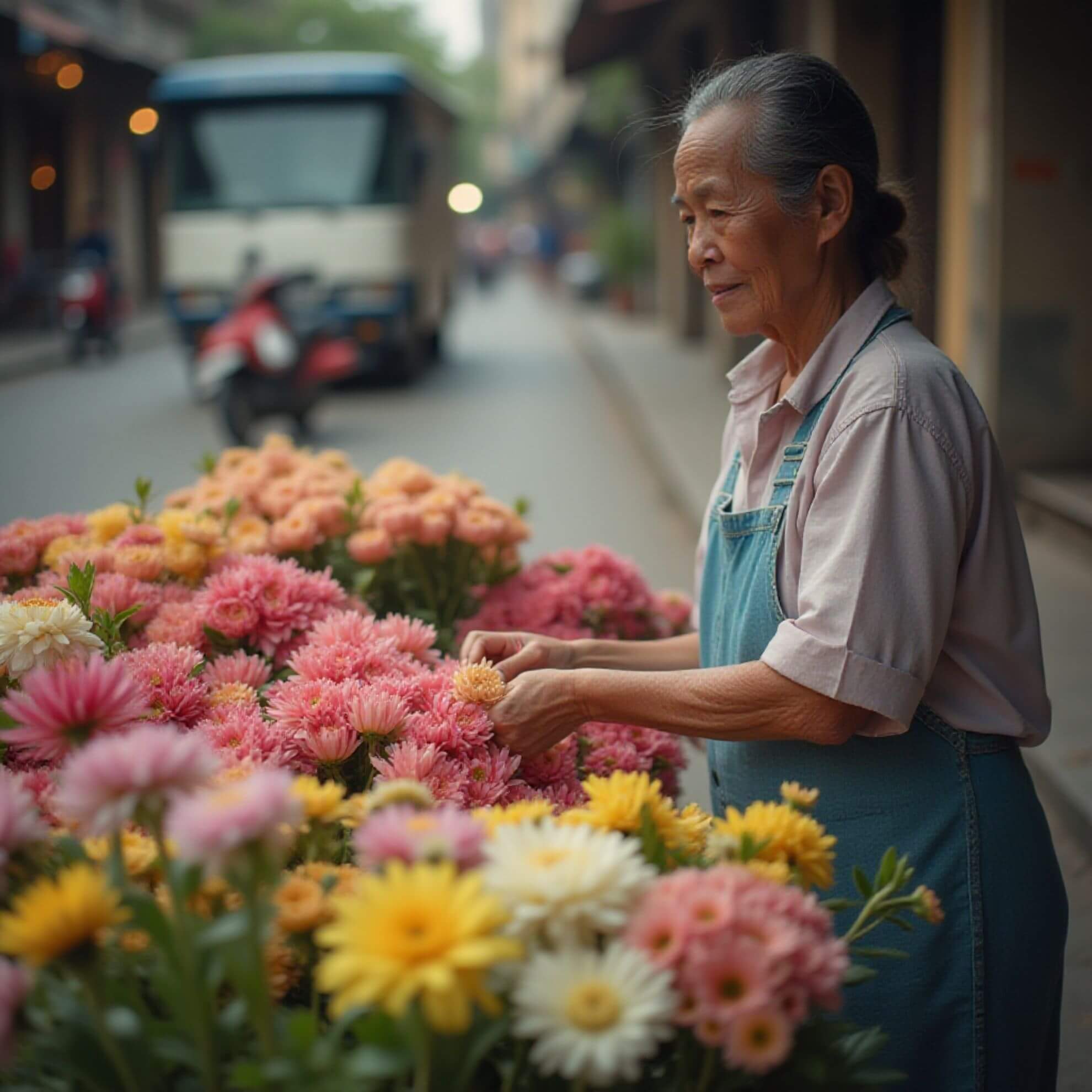} \\
\end{tabular}
\caption{\textbf{Qualitative comparisons on images from the \textit{people} dataset for FLUX.1 [dev].} Our model uses 10 inference steps and here does not apply base-model identity preservation. Even without regularization $\mathcal{L}_{\text{rec}}$, the model diminishes artifacts without significantly altering the generated image, unlike HandsXL.}
\label{fig:fluxdev1peopledataset}
\vspace{-4mm}
\end{figure*}

In order to sample from this distribution, the ODE is numerically integrated from $t=1$ to $t=0$. Typically using a standard Euler solver, resulting in the following update rule:
\[
    x_{t-\Delta t} = x_t - \Delta t \cdot v_\theta(x_t, t)
\]

A key property of Rectified Flow is the objective to learn transport paths as straight as possible. Ideally, the trajectory connects the initial noise $x_1$ and the final data $x_0$ linearly:
\[
    x_t = (1-t)x_0 + tx_1
\]

Differentiating this trajectory with respect to time yields the velocity \(v = \dfrac{\mathrm{d}}{\mathrm{d}t} x_t = x_1 - x_0\). By substituting $x_1 = v + x_0$ back into the interpolation equation, we derive the relationship between the current state $x_t$ and the clean data $x_0$:
\[
    x_t = (1-t)x_0 + t(v + x_0) = x_0 + t v
\]

Rearranging this term allows us to predict the clean data latent $\hat{x}_{0,t}$ from any intermediate timestep $t$ by subtracting the noise component:
\begin{equation}
    \label{eq:rec_img}
    \hat{x}_{0,t} = x_t - t \cdot v_\theta(x_t, t)
\end{equation}

Our method relies on this derivation to perform on-the-fly trajectory correction, estimating the final image structure $\hat{x}_{0,t}$ to look for appearing artifacts before the generation is complete, see Fig.~\ref{fig:fluxdev1peopledataset}.

\begin{figure}[t]
\centering
\setlength{\tabcolsep}{1.2pt}
\renewcommand{\arraystretch}{0.9}
\begin{tabular}{cccccccc}
 & $t_9$ & $t_8$ & $t_7$ & $t_4$ & $t_2$ & $t_0$\\
\rotatebox{90}{\tiny\hspace{3pt}{$\mathcal{D}(\hat{x}_{0,t})$}} & \includegraphics[width=0.07\textwidth, height=0.07\textwidth]{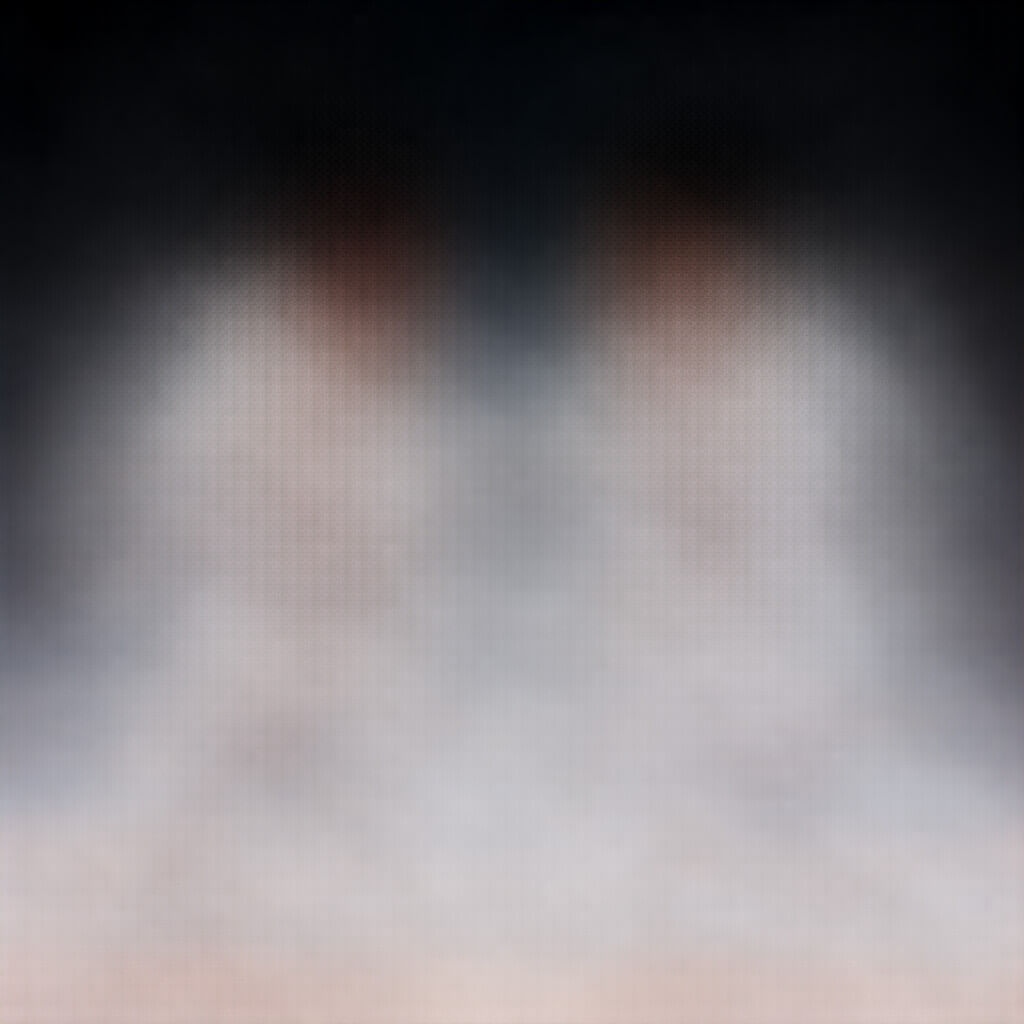} &
\includegraphics[width=0.07\textwidth, height=0.07\textwidth]{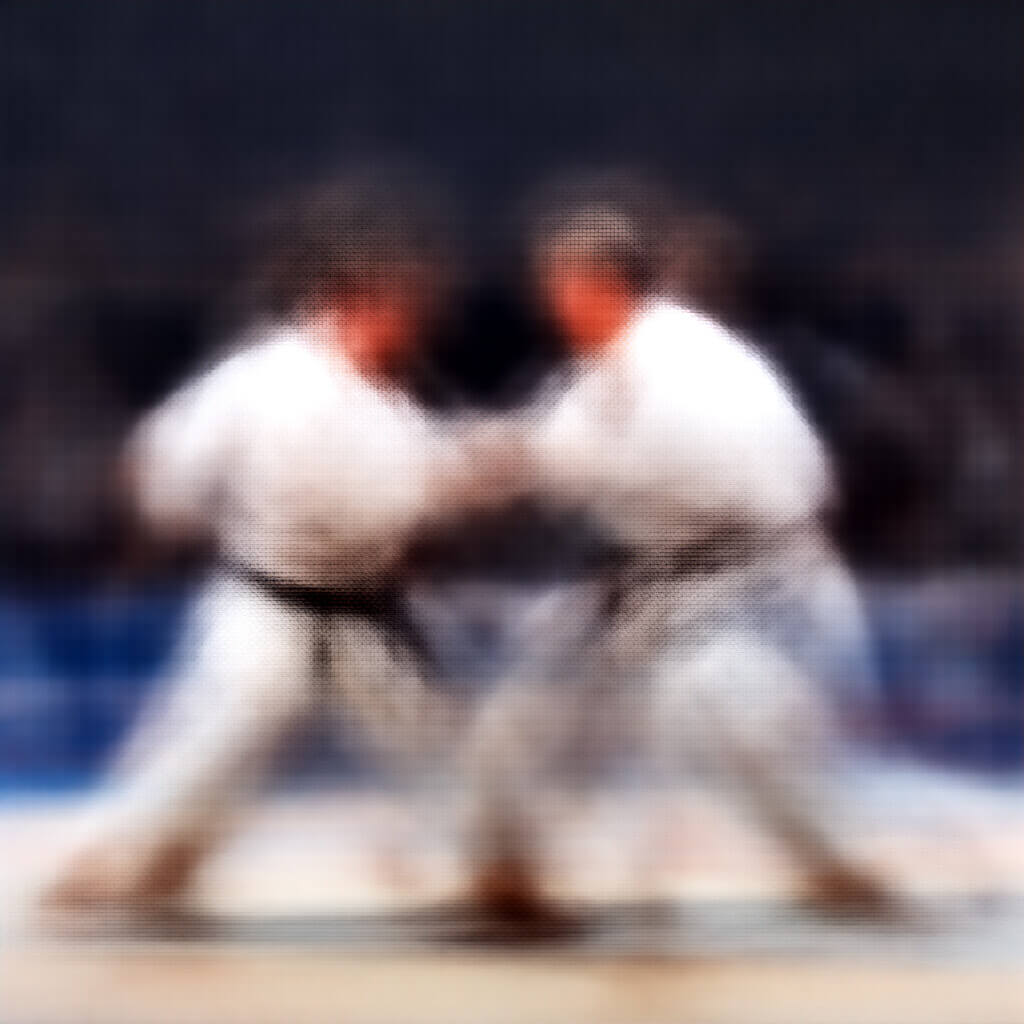} &
\includegraphics[width=0.07\textwidth, height=0.07\textwidth]{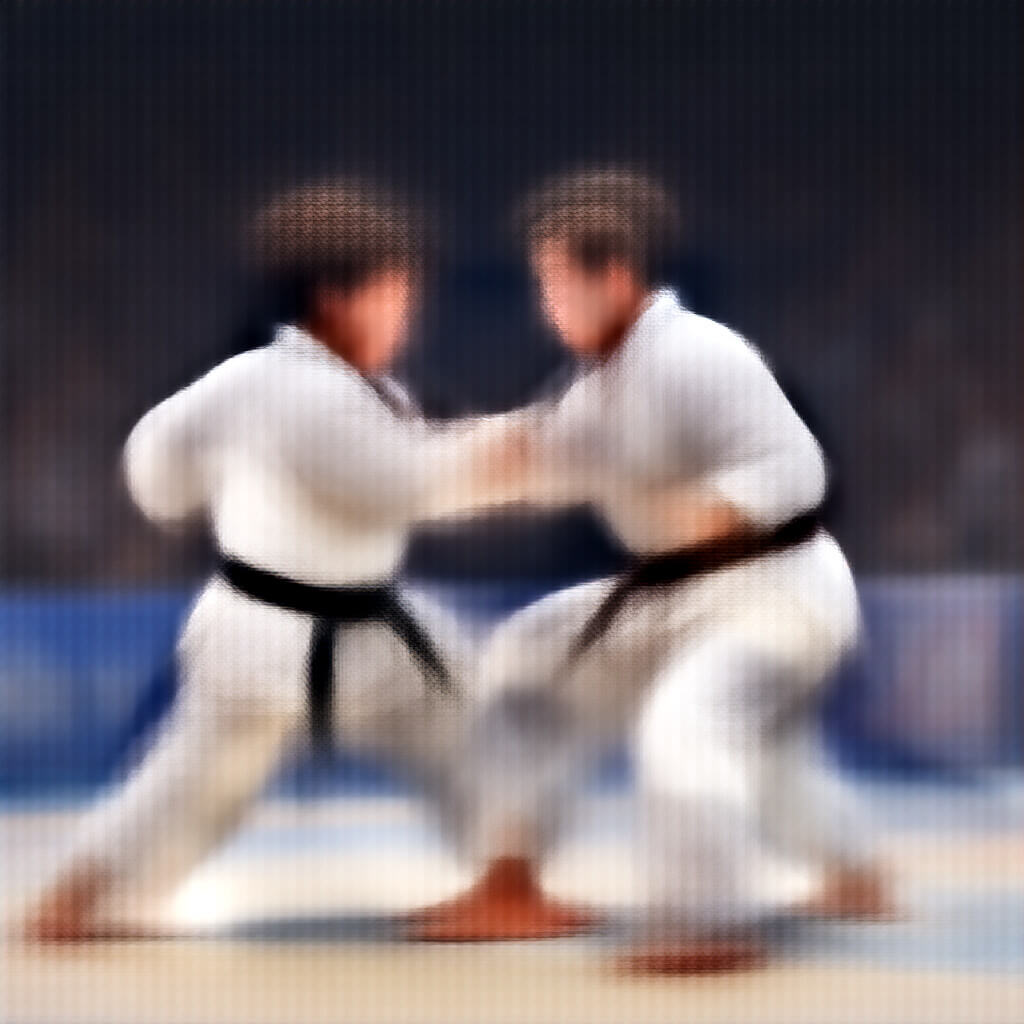} &
\includegraphics[width=0.07\textwidth, height=0.07\textwidth]{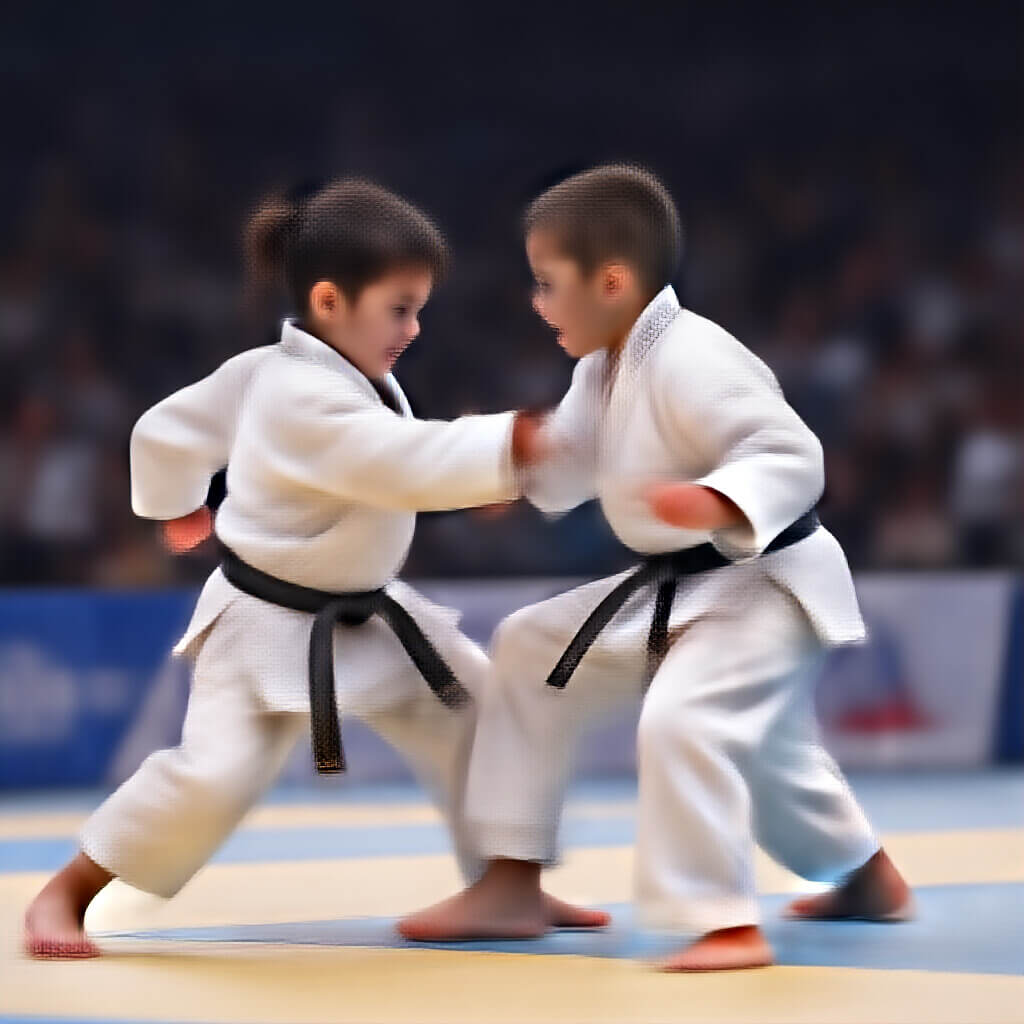} &
\includegraphics[width=0.07\textwidth, height=0.07\textwidth]{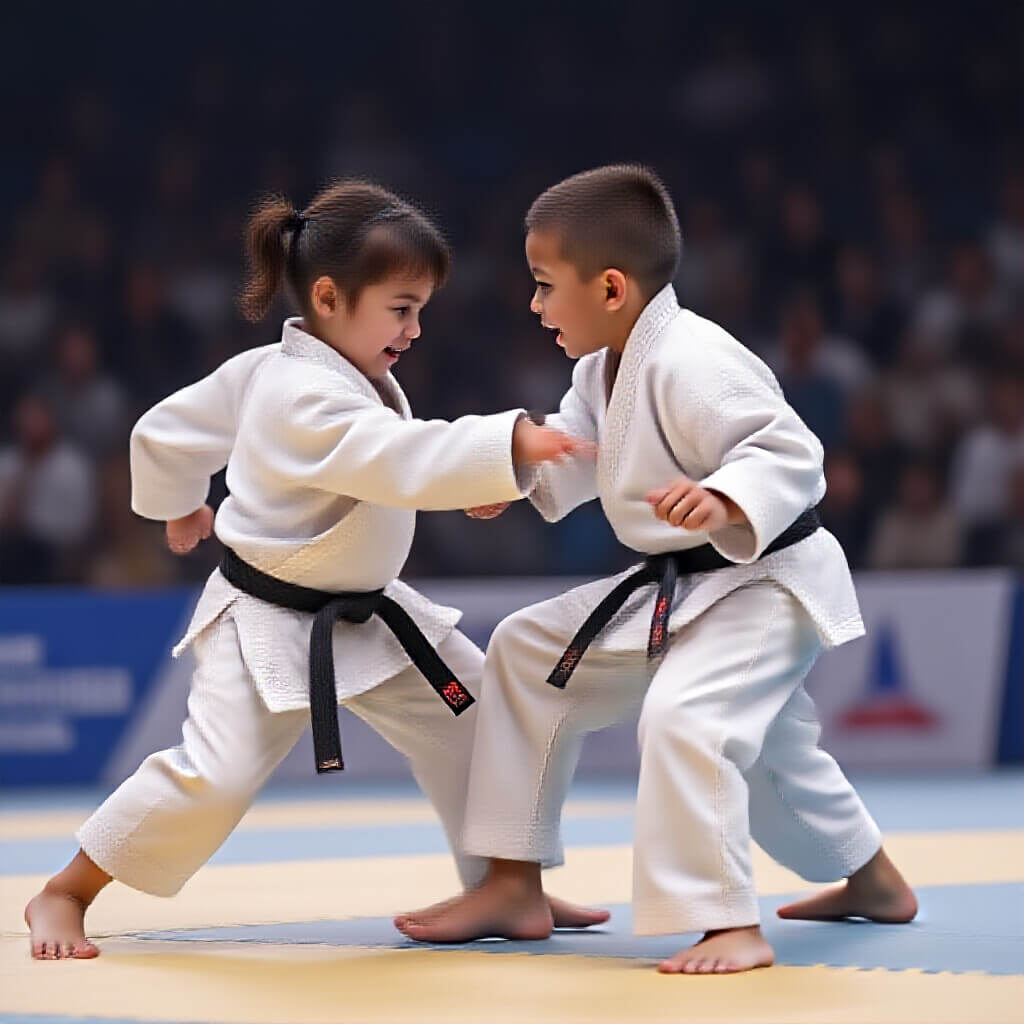}
& \includegraphics[width=0.07\textwidth, height=0.07\textwidth]{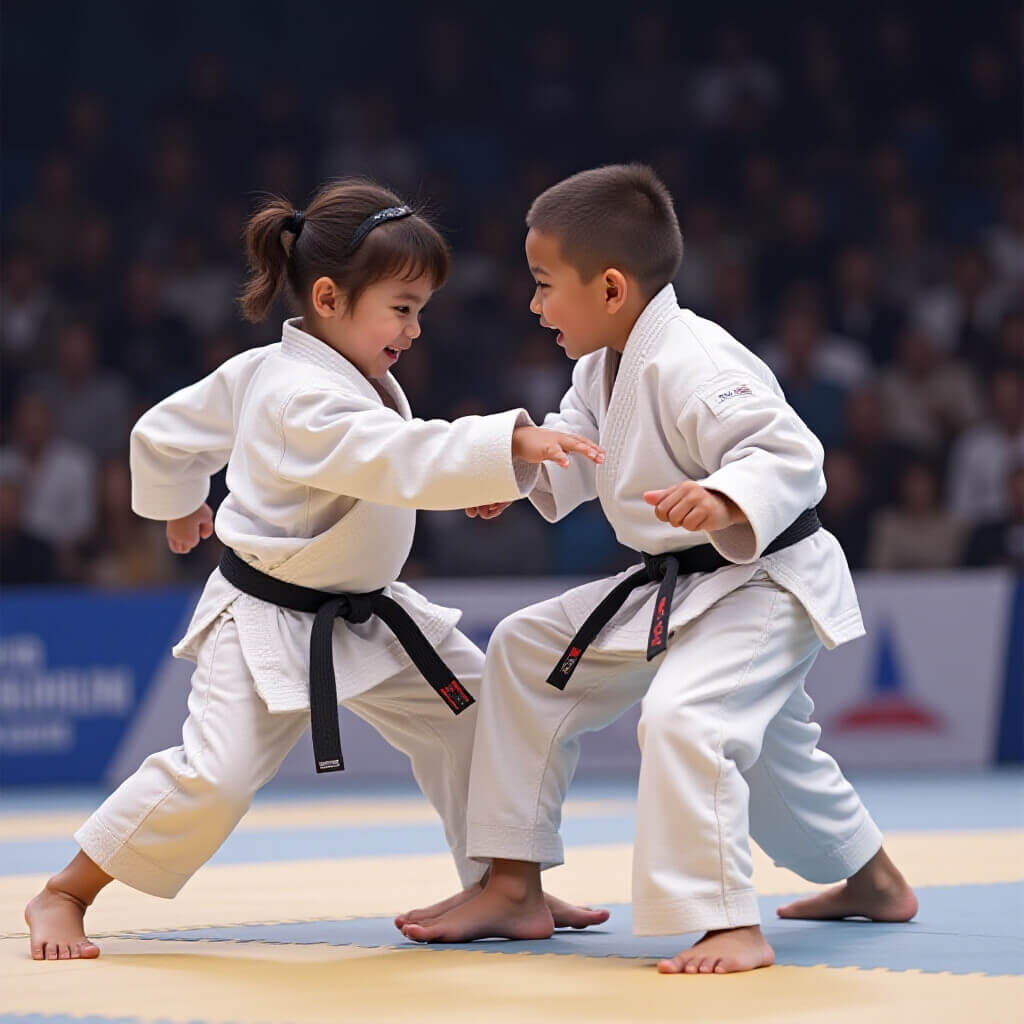} \\
\rotatebox{90}{\tiny\hspace{6pt}{Artifact}} \rotatebox{90}{\tiny{\hspace{8pt} Mask}}& \includegraphics[width=0.07\textwidth, height=0.07\textwidth]{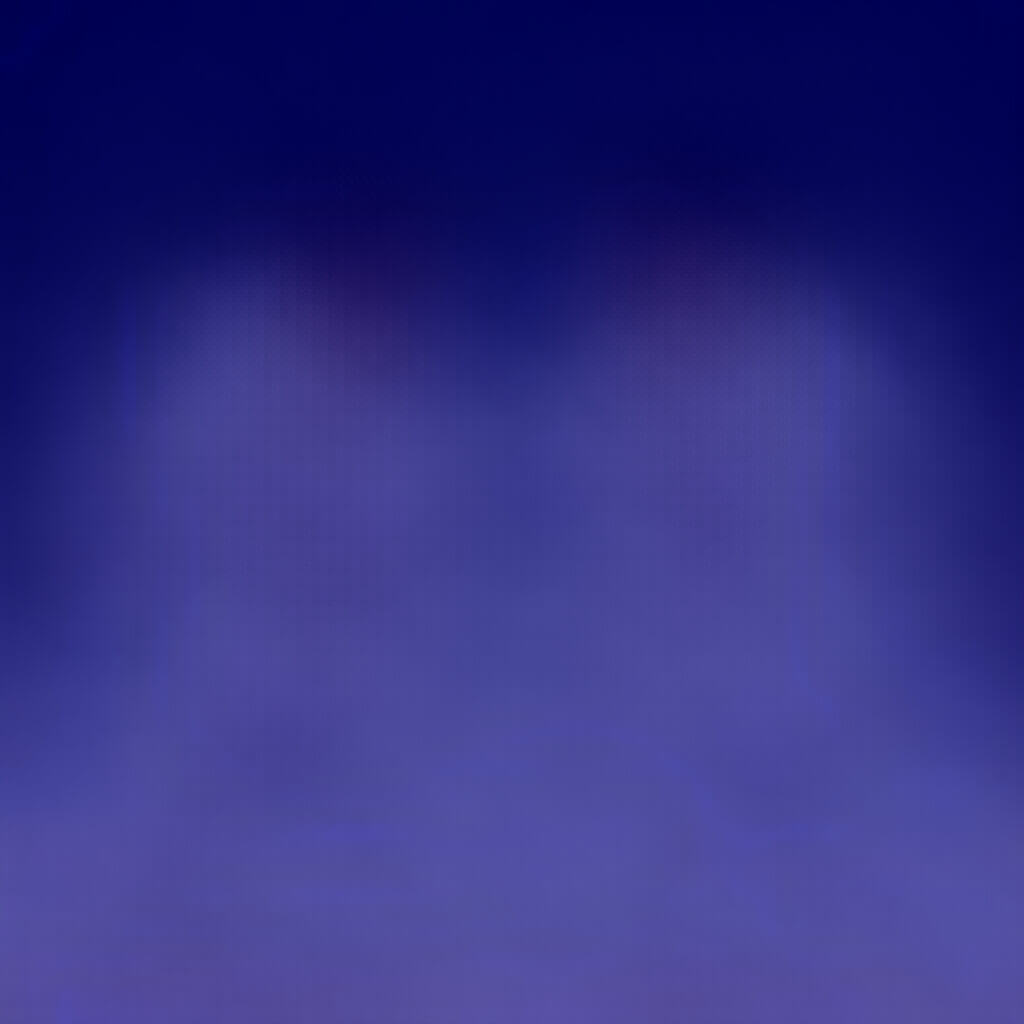} &
\includegraphics[width=0.07\textwidth, height=0.07\textwidth]{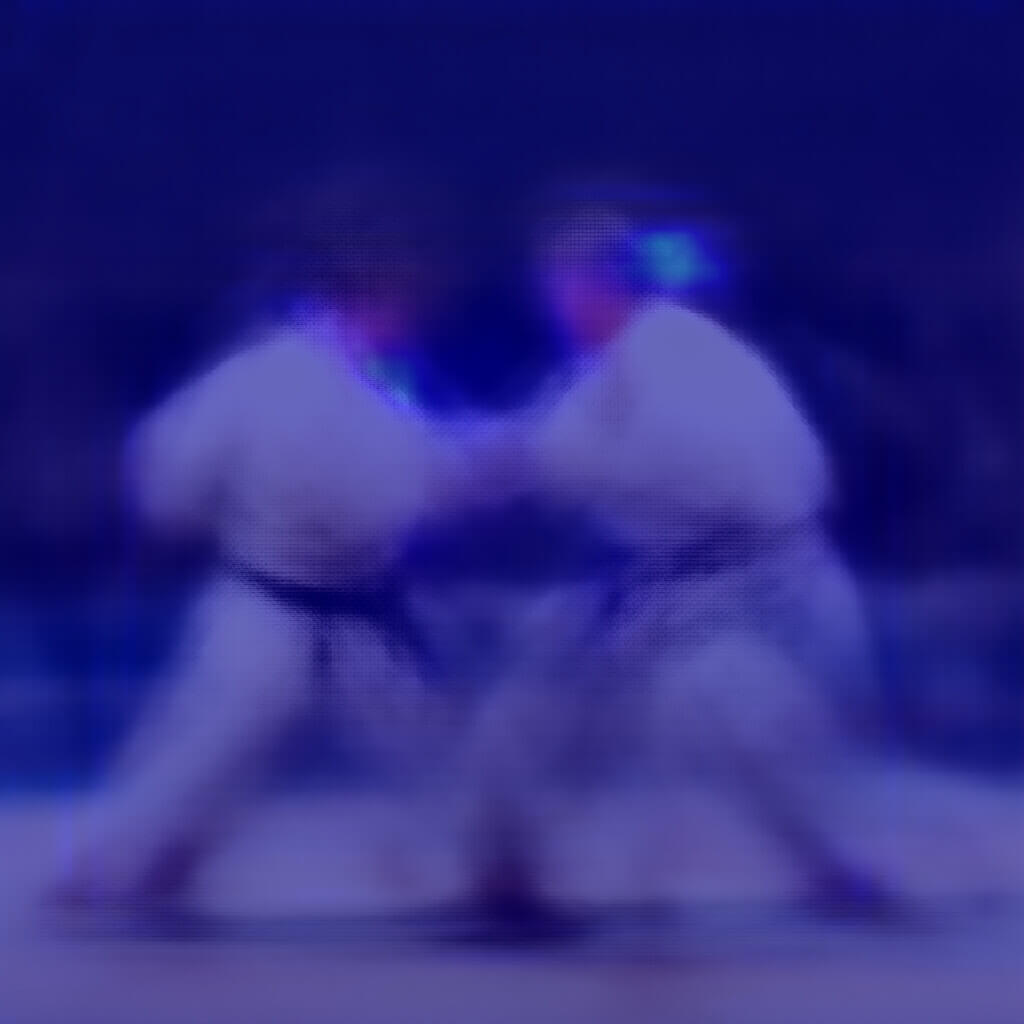} &
\includegraphics[width=0.07\textwidth, height=0.07\textwidth]{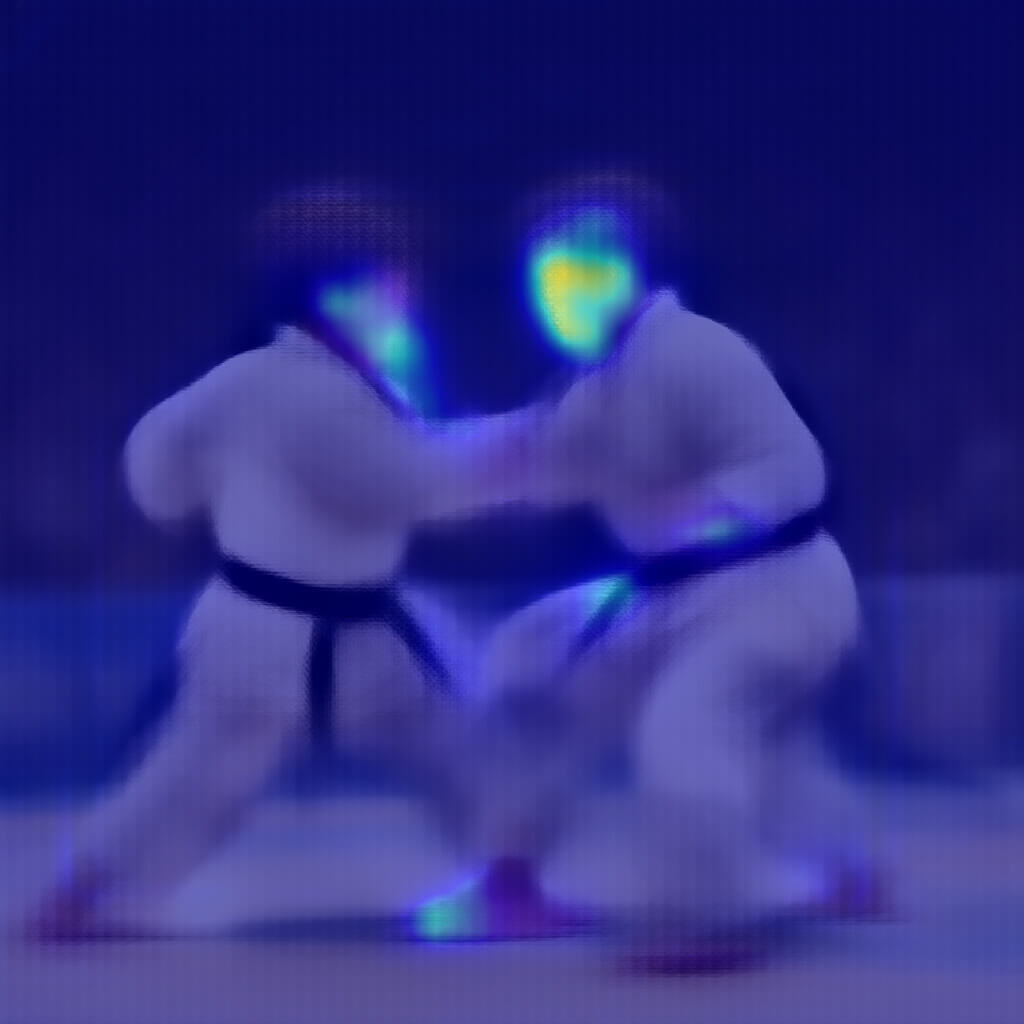} &
\includegraphics[width=0.07\textwidth, height=0.07\textwidth]{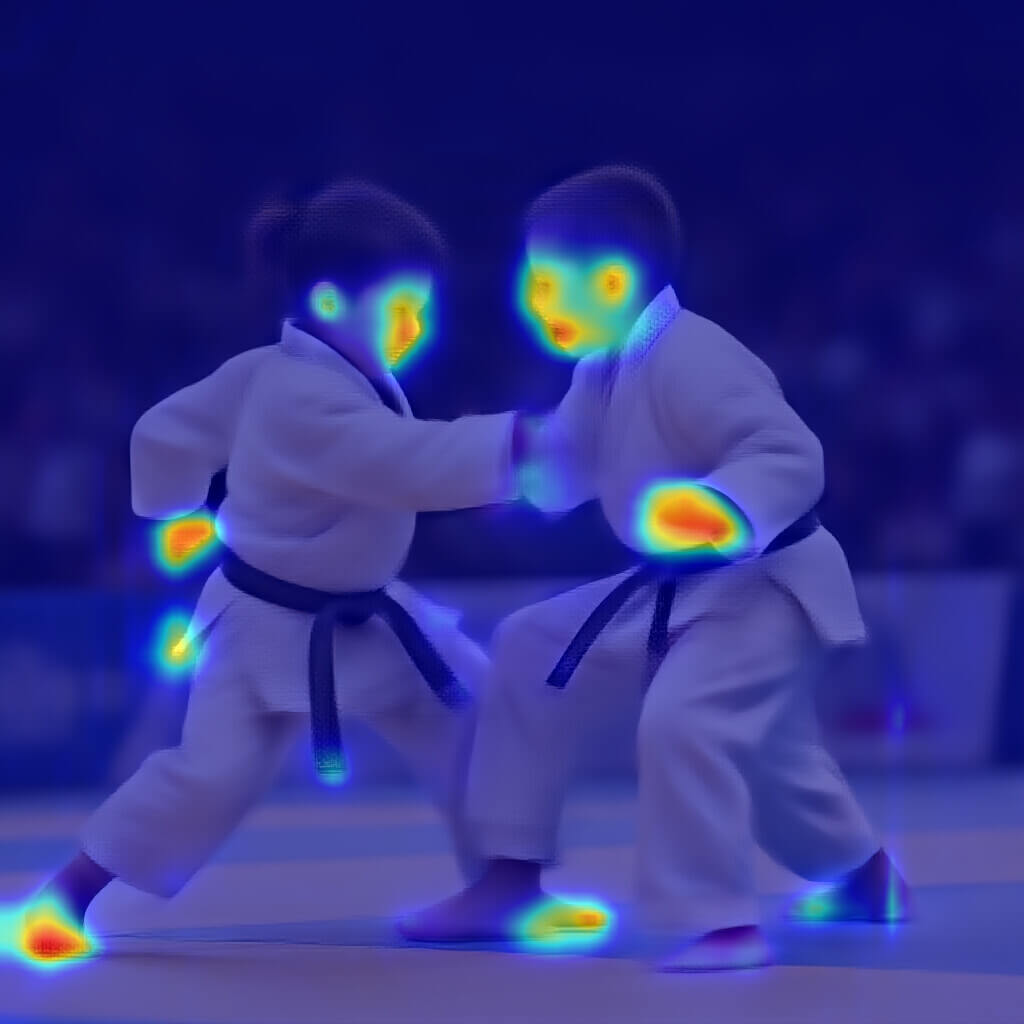} &
\includegraphics[width=0.07\textwidth, height=0.07\textwidth]{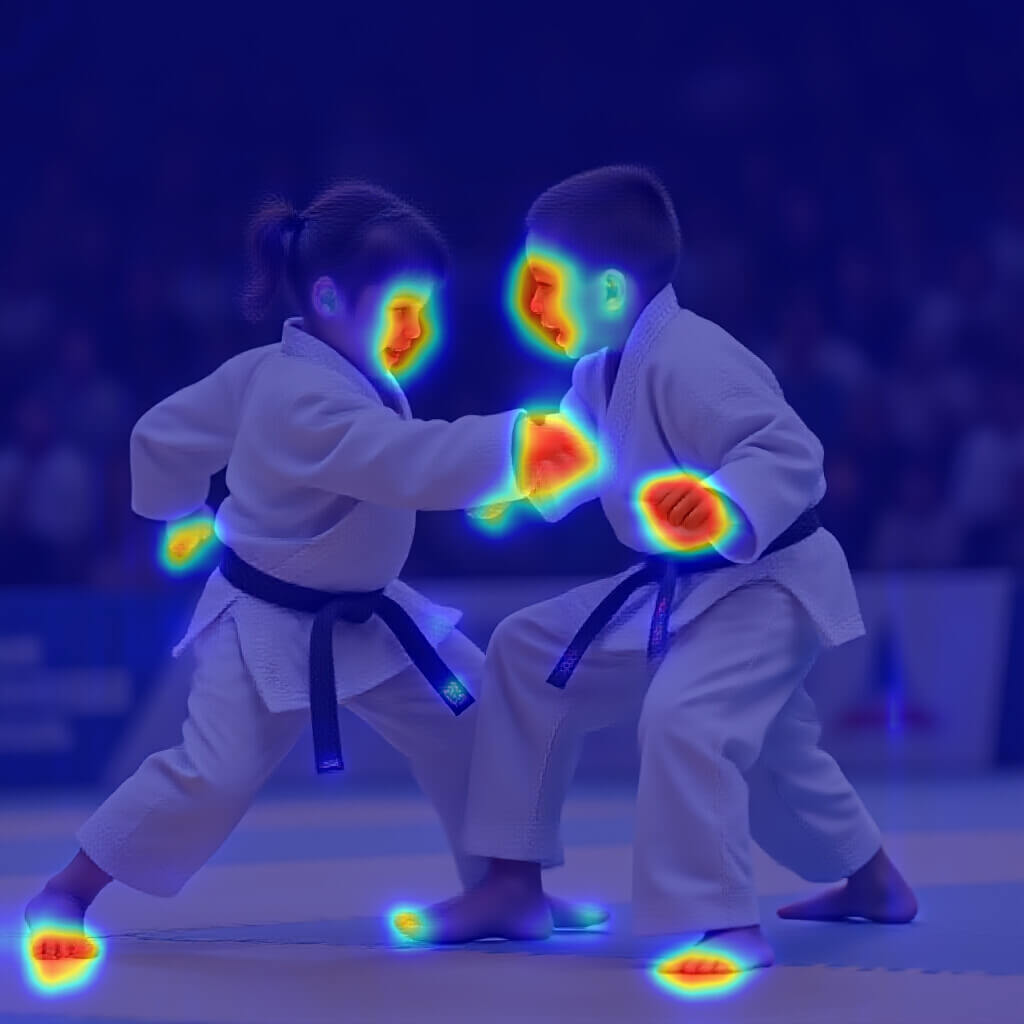}
& \includegraphics[width=0.07\textwidth, height=0.07\textwidth]{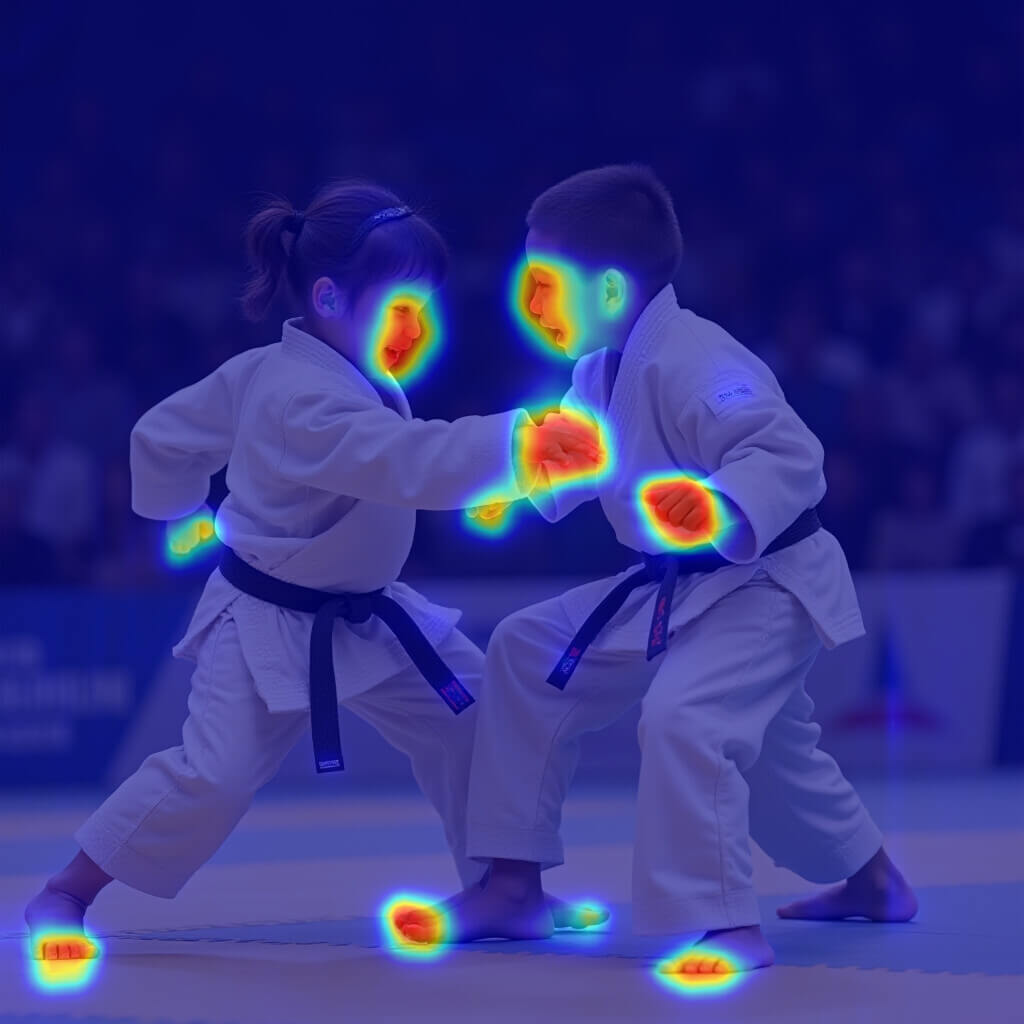} \\
\rotatebox{90}{\tiny\hspace{6pt}{$\mathcal{D}(x_t)$}} & \includegraphics[width=0.07\textwidth, height=0.07\textwidth]{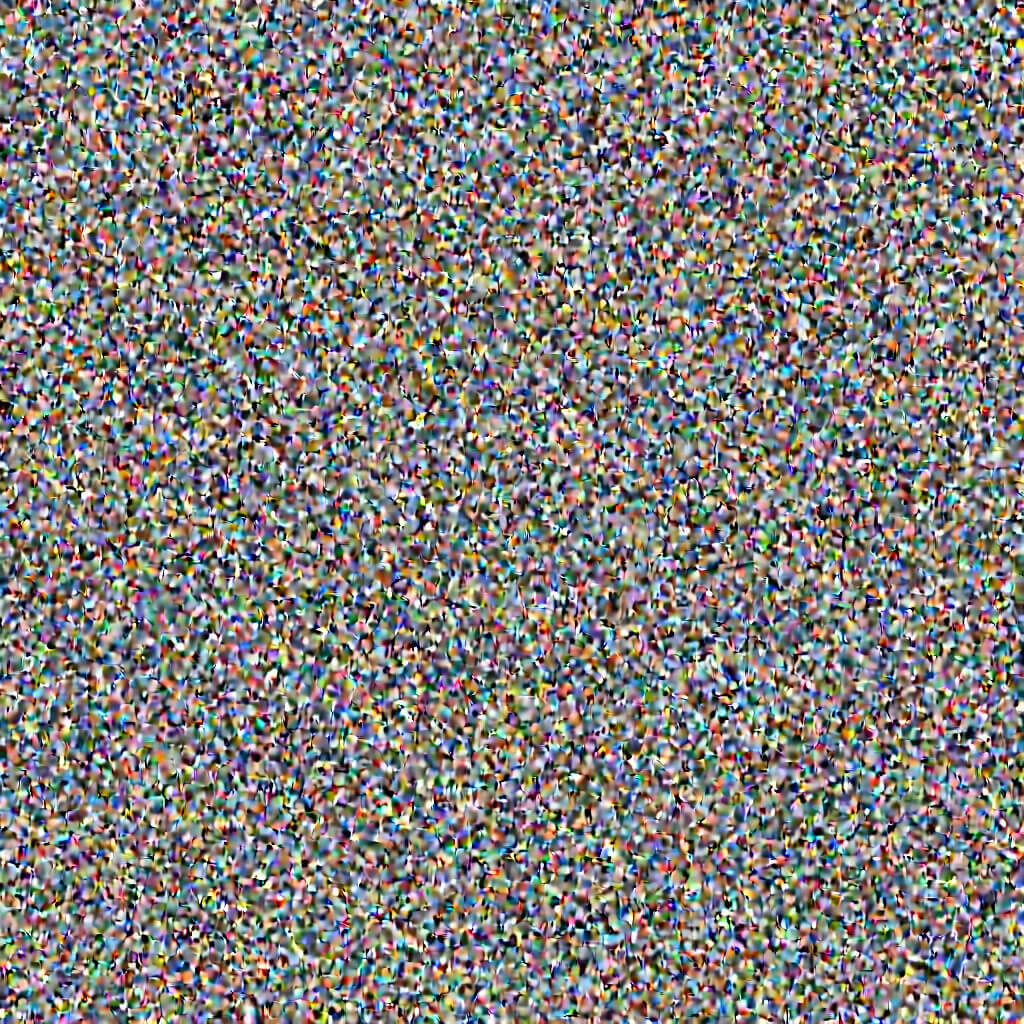} &
\includegraphics[width=0.07\textwidth, height=0.07\textwidth]{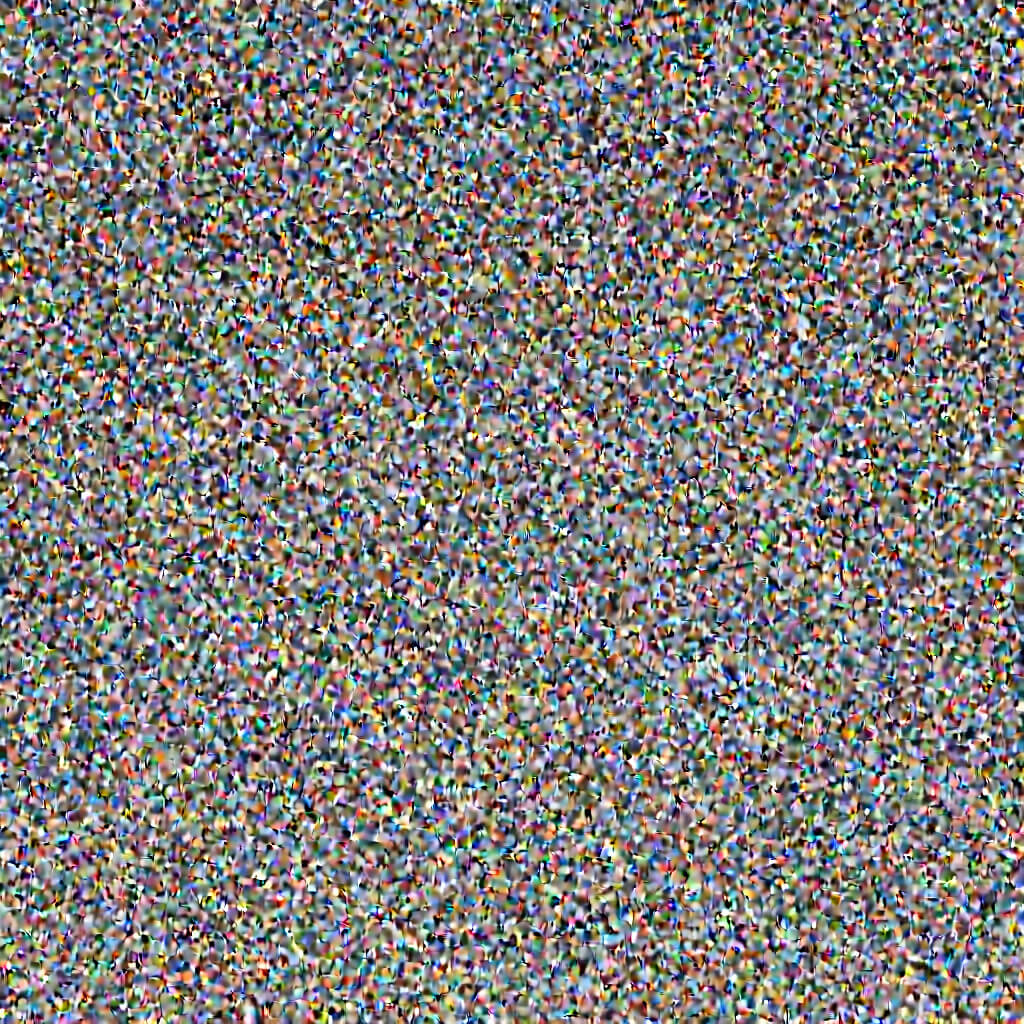} &
\includegraphics[width=0.07\textwidth, height=0.07\textwidth]{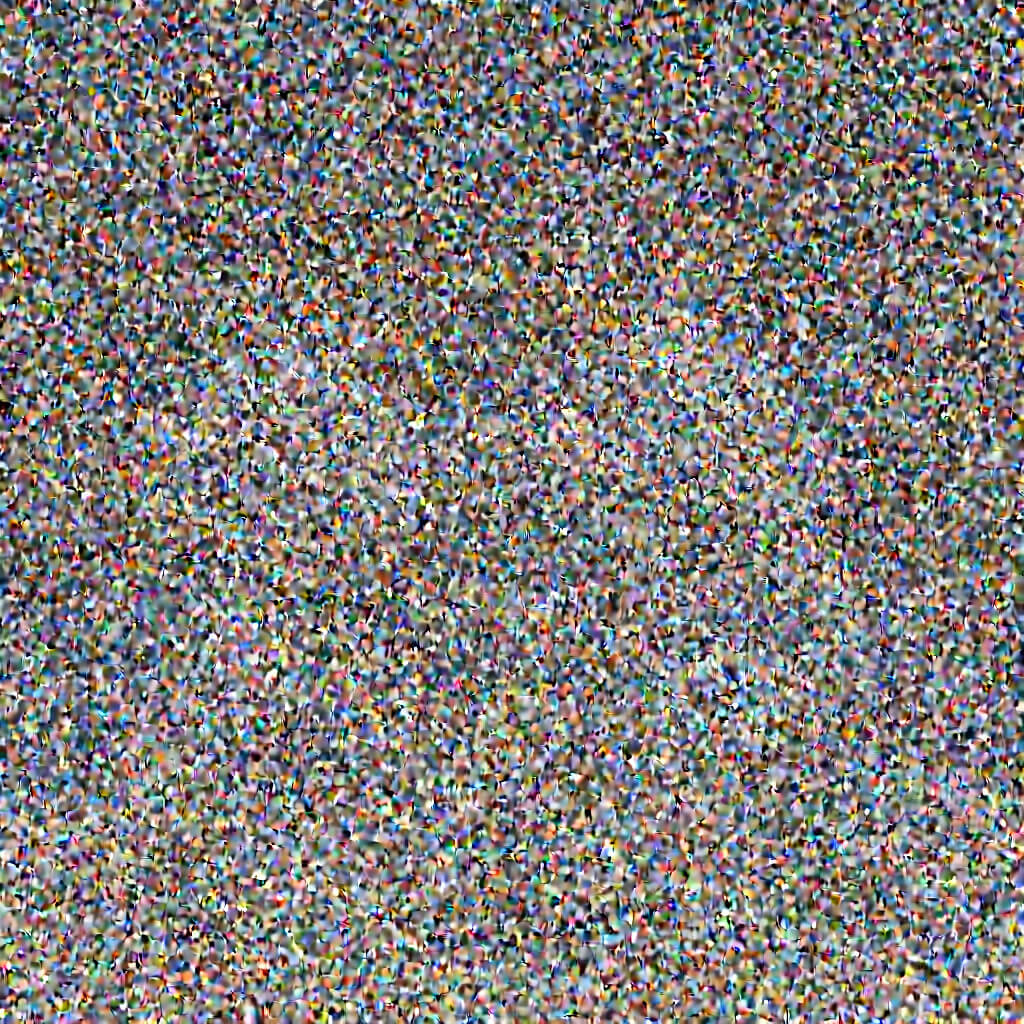} &
\includegraphics[width=0.07\textwidth, height=0.07\textwidth]{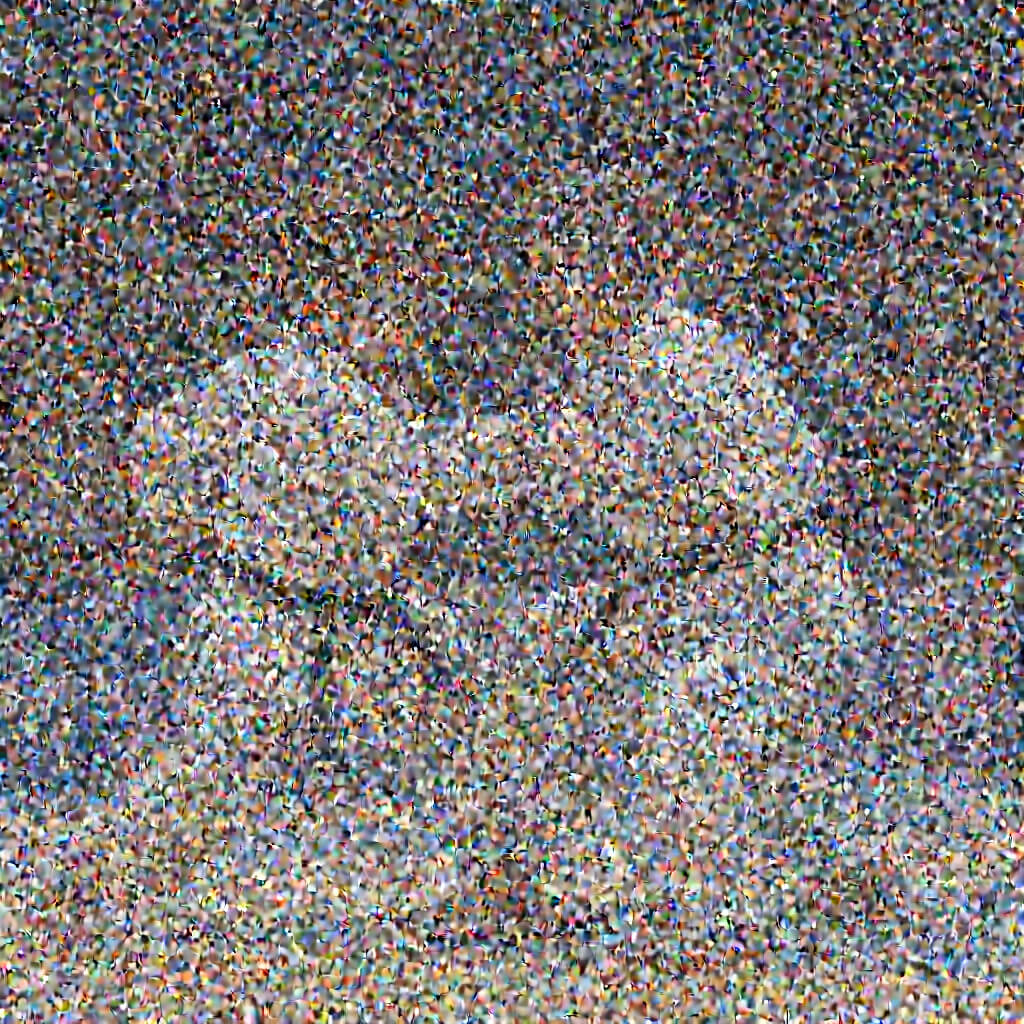} &
\includegraphics[width=0.07\textwidth, height=0.07\textwidth]{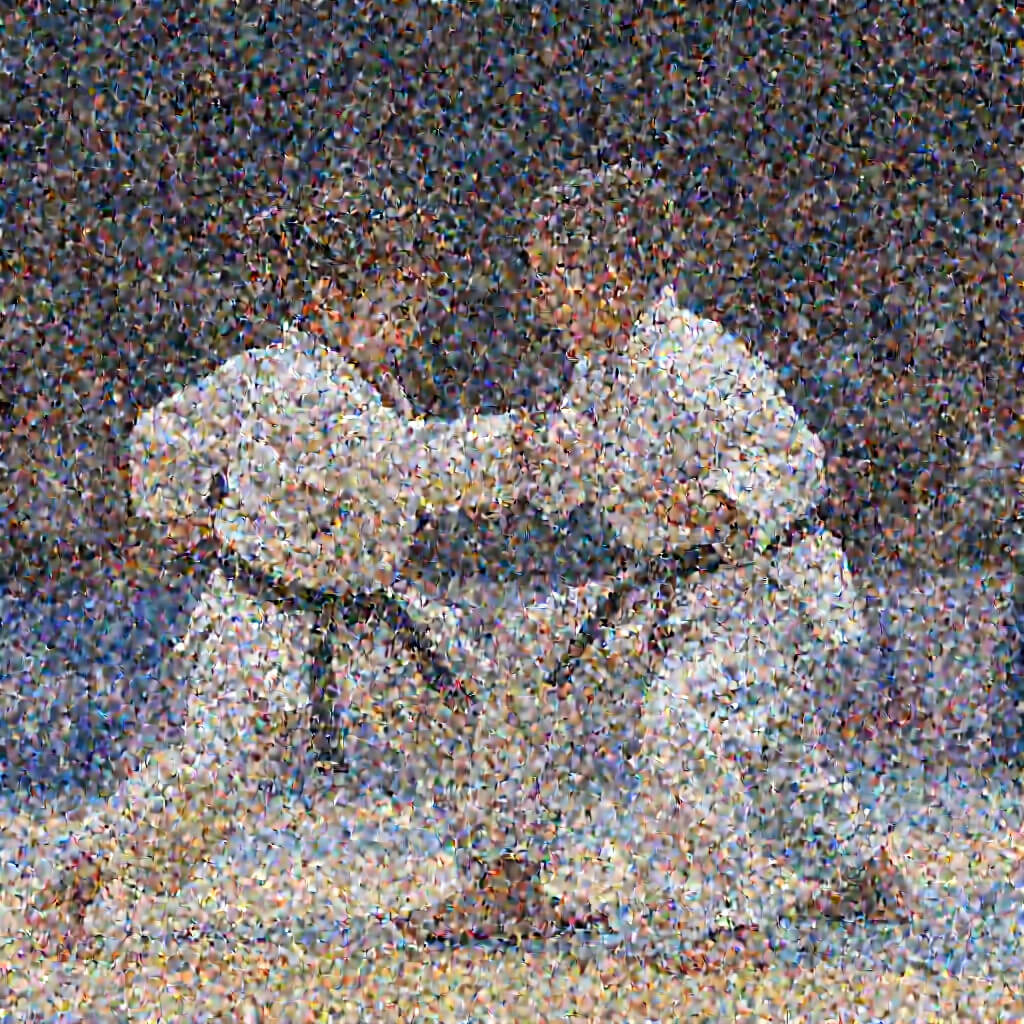}
& \includegraphics[width=0.07\textwidth, height=0.07\textwidth]{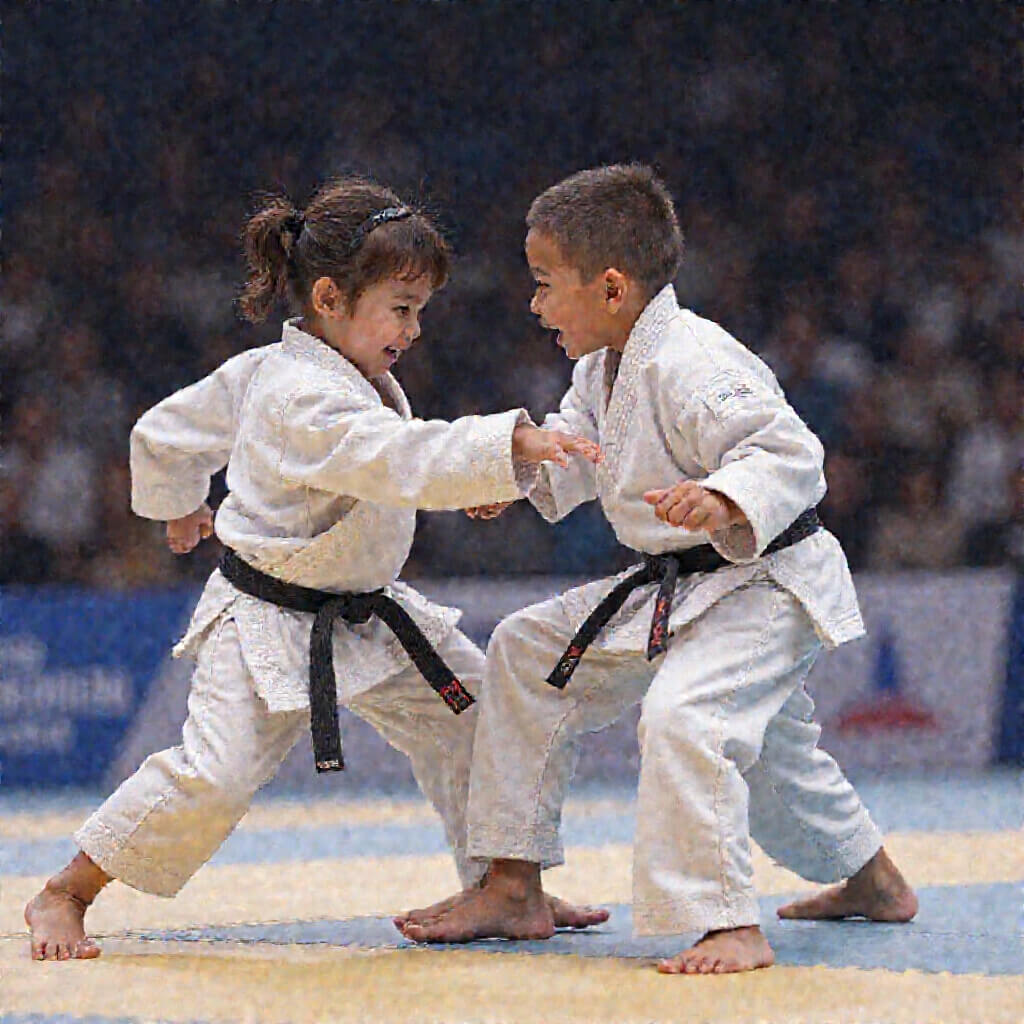} \\
\rotatebox{90}{\tiny\hspace{6pt}{Artifact}} \rotatebox{90}{\tiny{\hspace{8pt} Mask}} & \includegraphics[width=0.07\textwidth, height=0.07\textwidth]{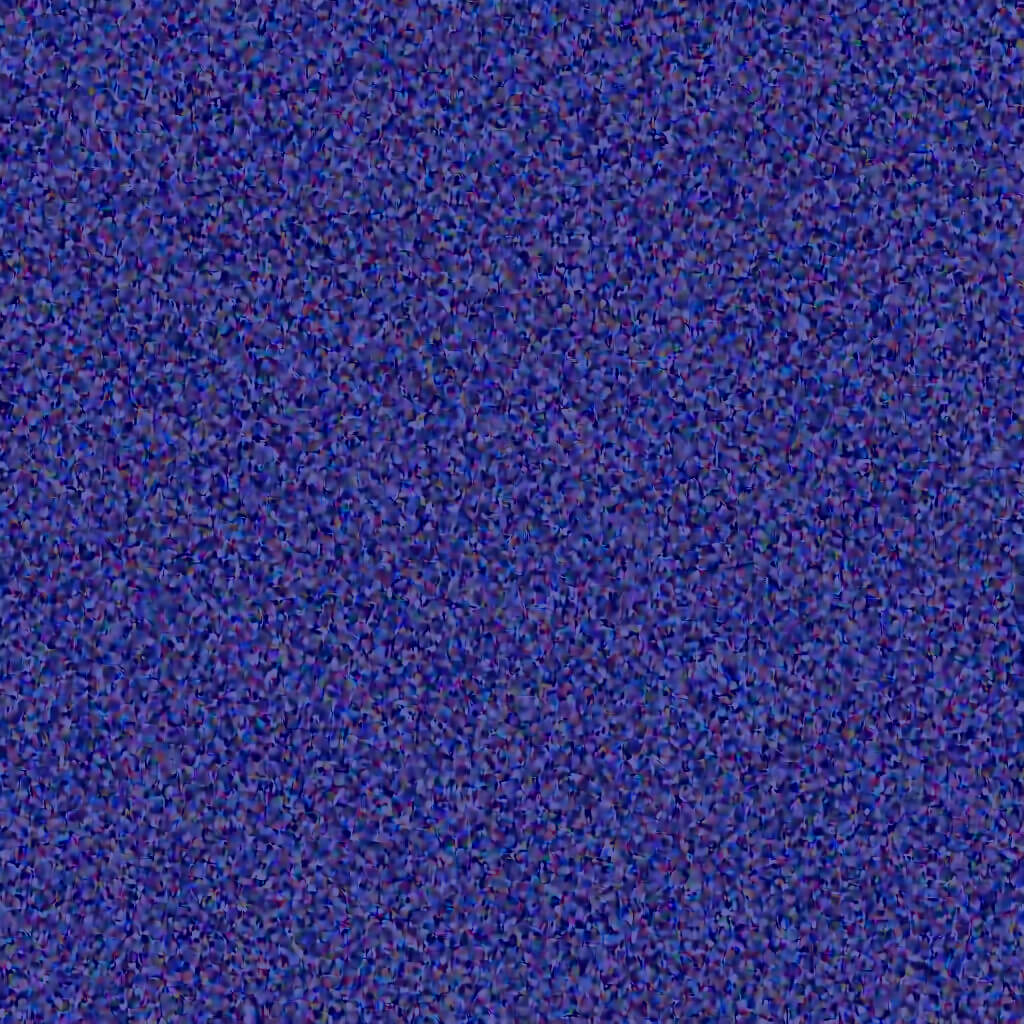} &
\includegraphics[width=0.07\textwidth, height=0.07\textwidth]{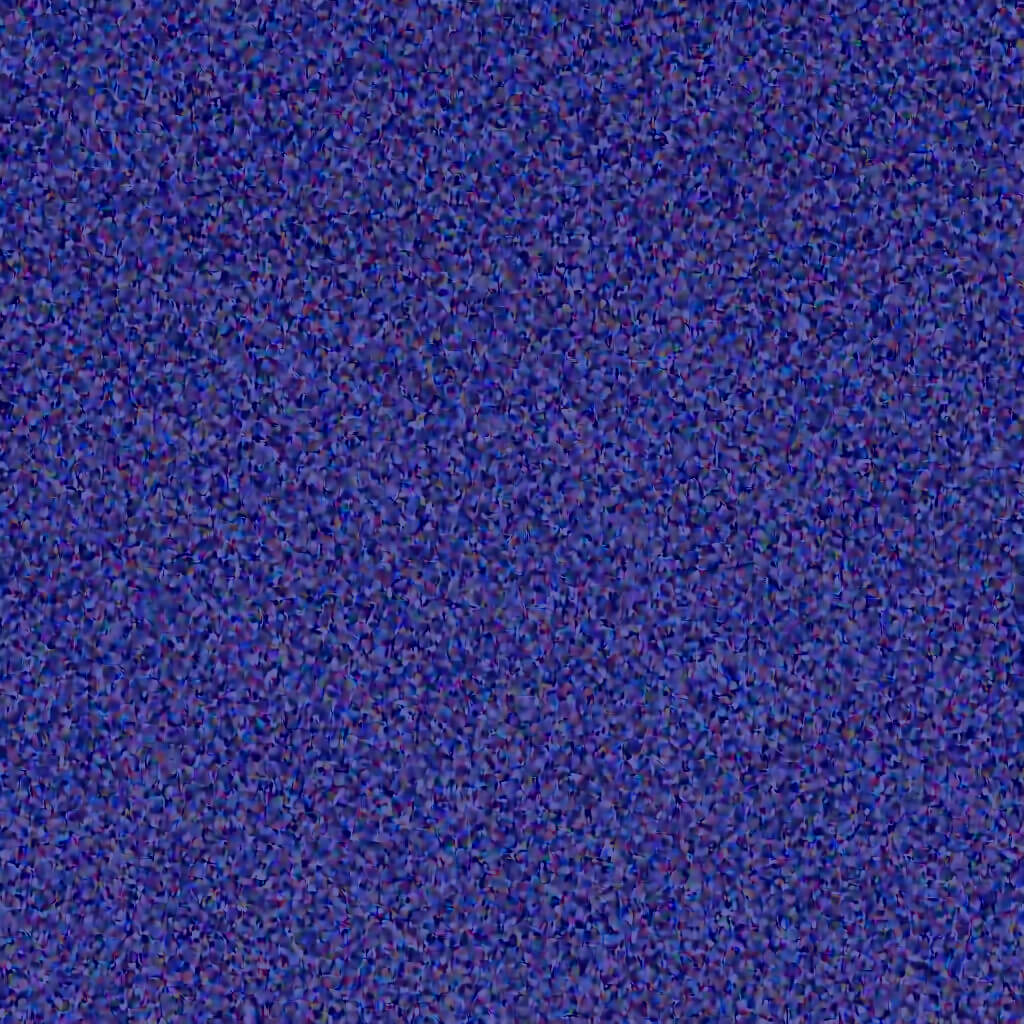} &
\includegraphics[width=0.07\textwidth, height=0.07\textwidth]{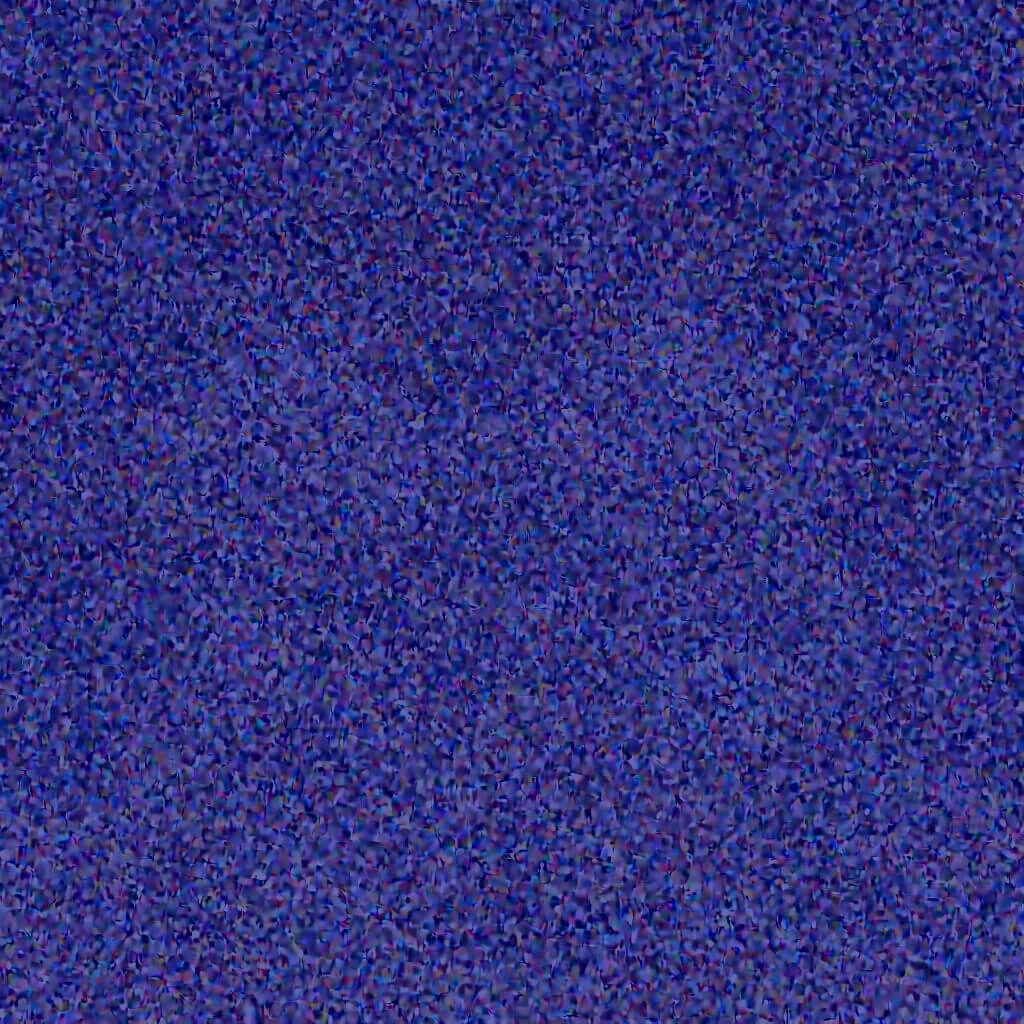} &
\includegraphics[width=0.07\textwidth, height=0.07\textwidth]{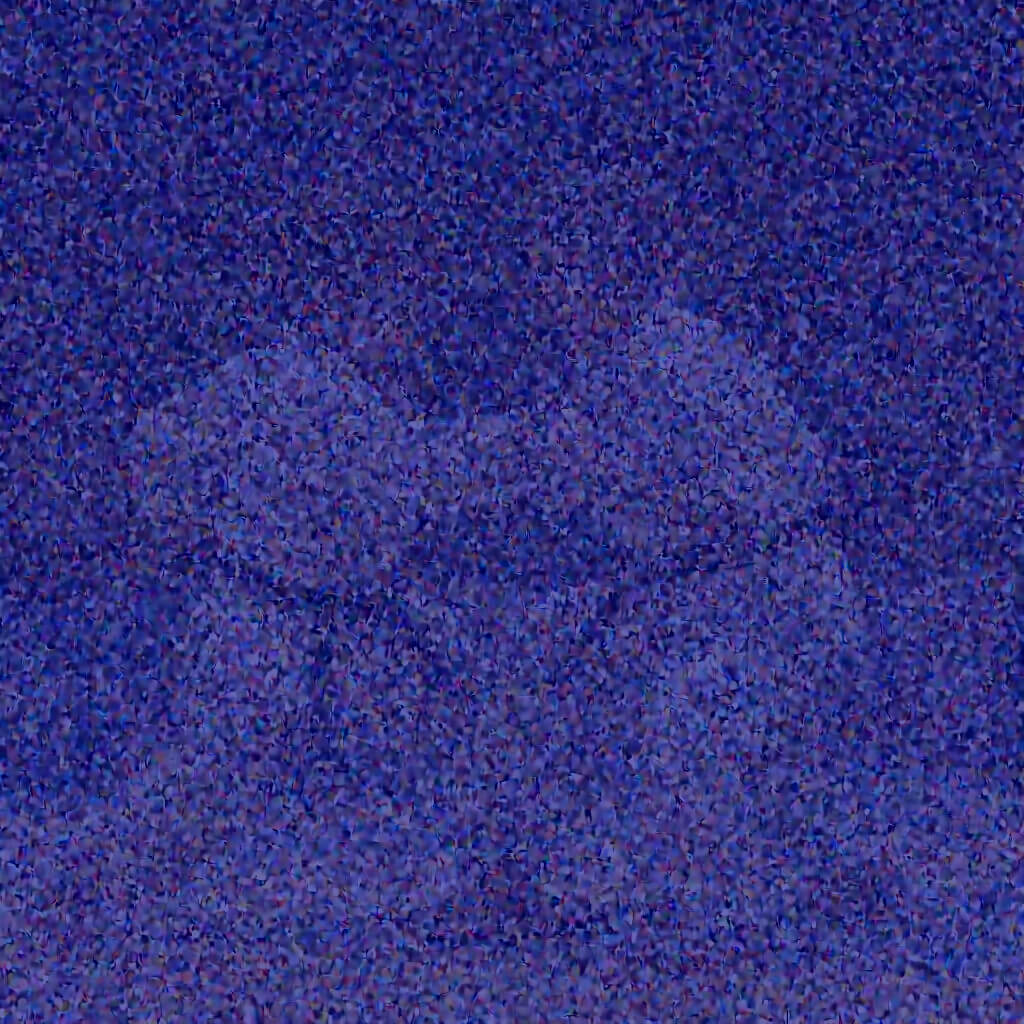} &
\includegraphics[width=0.07\textwidth, height=0.07\textwidth]{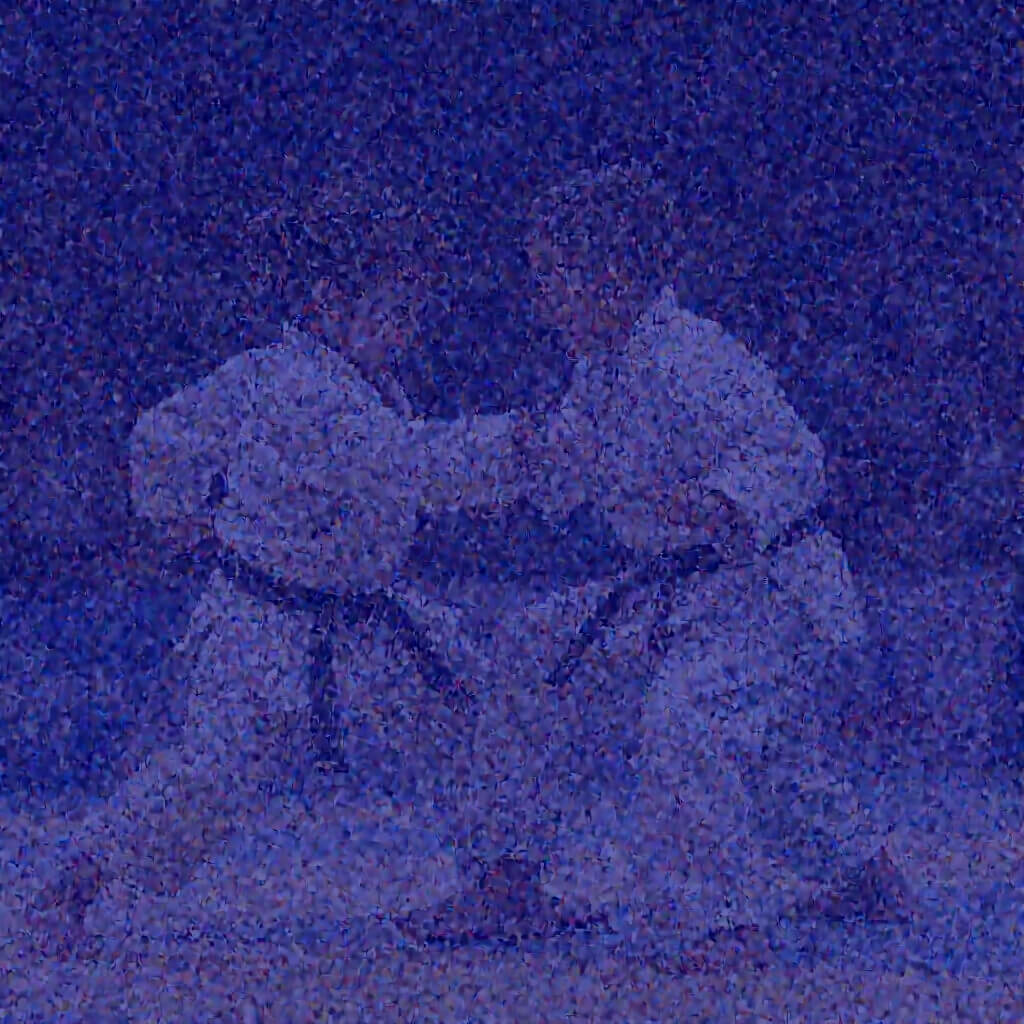}
& \includegraphics[width=0.07\textwidth, height=0.07\textwidth]{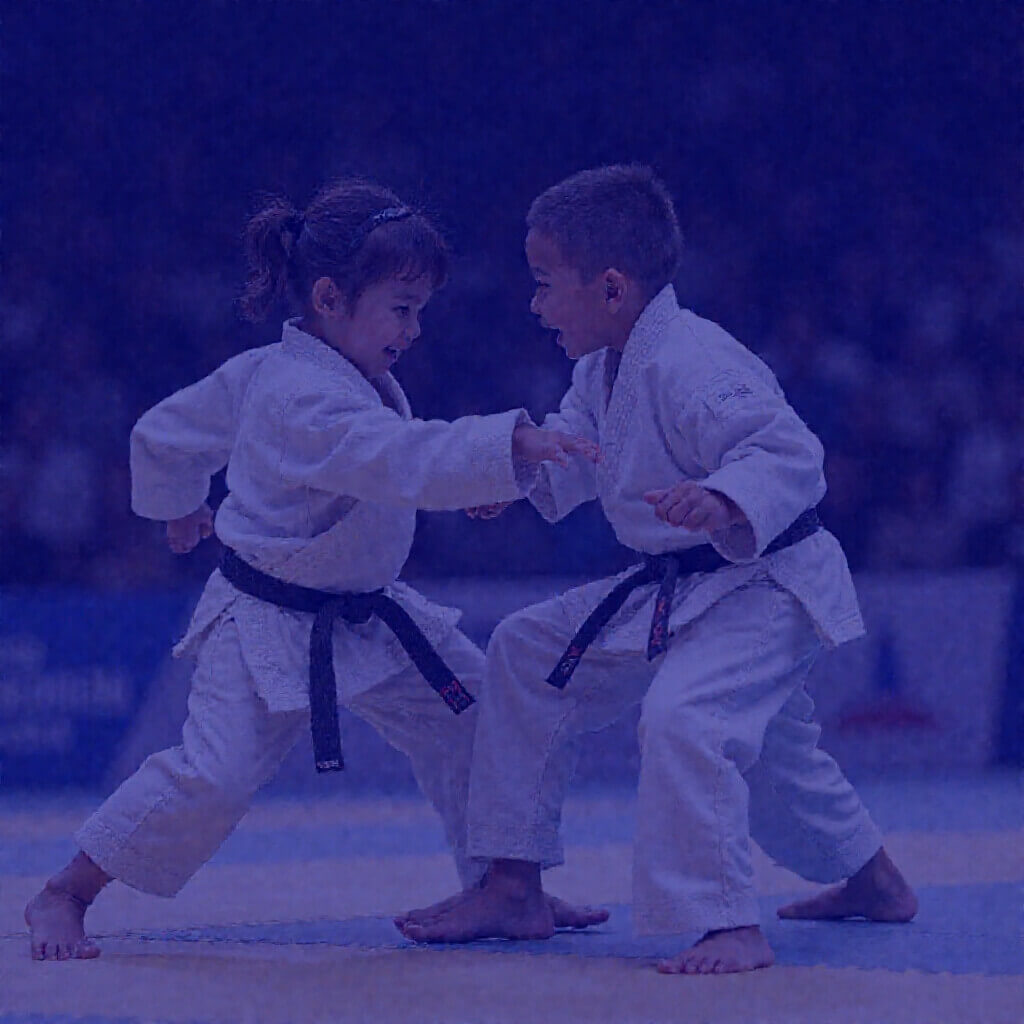} \\
\end{tabular}
\caption{\textbf{Comparison of images generated using estimated clean data latents $\hat{x}_{0,t}$ and noisy latents $x_t$ during intermediate inference steps for FLUX.1 [dev].} Artifact masks are shown to highlight the effectiveness of using the estimated latents. Contrary to using $x_t$, the detector is able to find artifact regions even during early stages when using $\hat{x}_{0,t}$ estimation.}
\label{fig:x0andxt}
\vspace{-2mm}
\end{figure}

\textbf{Artifact Detection}
In order to detect artifacts a definition of an artifact has to be provided. We follow the definition of an artifact from DiffDoctor~\cite{wang2025diffdoctor}. In this formulation the artifacts belong to one of the three types: shape distortions (e.g., distorted hands, faces, words), unreasonable content (e.g., extra limbs, more-or-fewer fingers), and watermarks. Notably, artifacts which require complex reasoning such as floating objects without physical support are not included in this definition. 

To identify regions containing artifacts, unless specified otherwise we use the Artifact Detector introduced in DiffDoctor, which follows the above definition. The detector maps each pixel of an input image to the probability that the pixel is part of an artifact.


\textbf{Artifact-aware Sampling}
The detection module is integrated directly into the sampling loop of the flow matching model. At timestep $t$, we seek to update the latent $x_t$ such that the resulting image contains fewer artifacts while maintaining faithfulness to the prompt. 

A crucial distinction of \our{} from existing methods, which typically directly operate on the noisy latent $x_t$. However, artifact detectors trained on clean data latent might struggle with generalization to intermediate noisy latent, resulting in sub-optimal guidance. 

To address this, we introduce a clean latent estimation step. Instead of working with the noisy latent, we directly calculate the clean data latent estimation $\hat{x}_{0,t}$ is calculated using Eq.~\ref{eq:rec_img}. This estimated latent is then mapped to the pixel space using the VAE decoder $\mathcal{D}$: $
    \hat{I}_t = \mathcal{D}(\hat{x}_{0,t})
$.

By working directly on the estimated clean data latent $\hat{x}_{0,t}$ rather than the noisy state $x_t$, we ensure the detector operates in its optimal domain. To the best of our knowledge, we are the first to apply this specific $\hat{x}_{0,t}$ prediction guidance scheme within the FLUX architecture, see Fig.~\ref{fig:x0andxt}.

The estimated image $\hat{I}_t$ is passed to the detector to obtain an artifact mask $M^t = \mathcal{AD}(\hat{I}_t)$ containing probabilities. The mask is then used to calculate the artifact loss
\begin{equation}
    \mathcal{L}_{a}=\dfrac{1}{HW} \sum_{i,j} {M^{t}_{i,j}}^2,
\end{equation}
which is equal to the mean squared error between the mask and artifact-free image mask (e.g. mask containing only zeroes).

To discourage the creation of an artifact, the trajectory is altered using the shift term based on the value of $\mathcal{L}_a$ with respect to $x_t$. Since the detector, the decoder, and the linear extrapolation step are all differentiable operations, we compute the gradient with regards to $x_t$ and use it to alter the trajectory.

We compute a displacement vector $\delta_t$ by normalizing the gradient and scaling it by a time-dependent scalar $\lambda_t$:
\[
    \delta_t = \lambda_t \frac{\nabla_{x_t}\mathcal{L}_{a}}{\|\nabla_{x_t}\mathcal{L}_{a}\|_2 + \epsilon}
\]
where $\epsilon$ is a small constant for numerical stability.  

Unlike standard classification guidance, where the gradients are typically unnormalized~\cite{dhariwal2021diffusion}. In this setting, the magnitude of the log-probability implicitly contains the confidence of the classifier. However, in our pipeline, the magnitude of the raw gradient $\|\nabla_{x_t}\mathcal{L}_{a}\|_2$ is tied to the probability of all pixels containing artifacts. Consequently, a large artifact region results in a much larger gradient norm, compared to a small artifact, despite both requiring equal attention.

\begin{figure}[!t]
\centering
\setlength{\tabcolsep}{1.2pt}
\renewcommand{\arraystretch}{0.9}
\begin{tabular}{cccccccc}
 & $t_9$ & $t_8$ & $t_7$ & $t_5$ & $t_3$ &$ t_0$ & Final\\
\rotatebox{90}{} & \includegraphics[width=0.06\textwidth, height=0.06\textwidth]{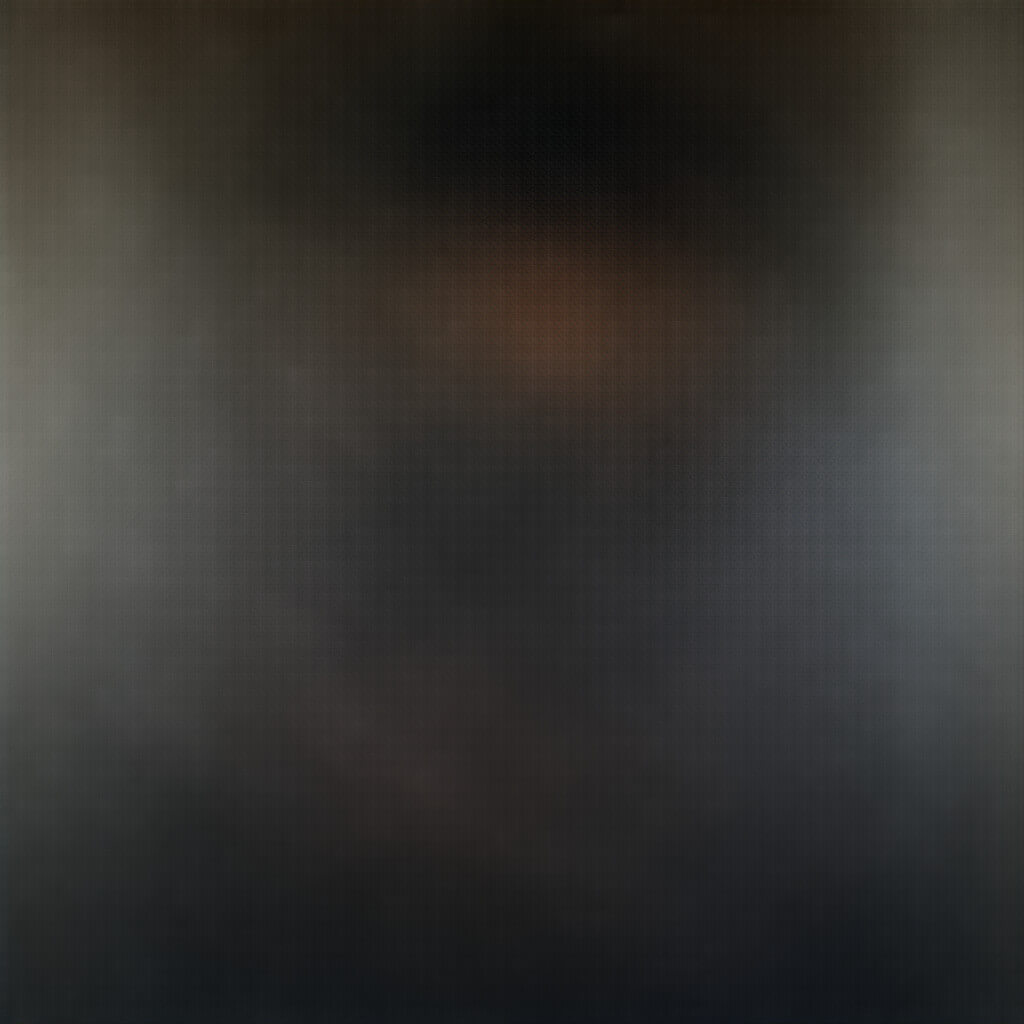} &
\includegraphics[width=0.06\textwidth, height=0.06\textwidth]{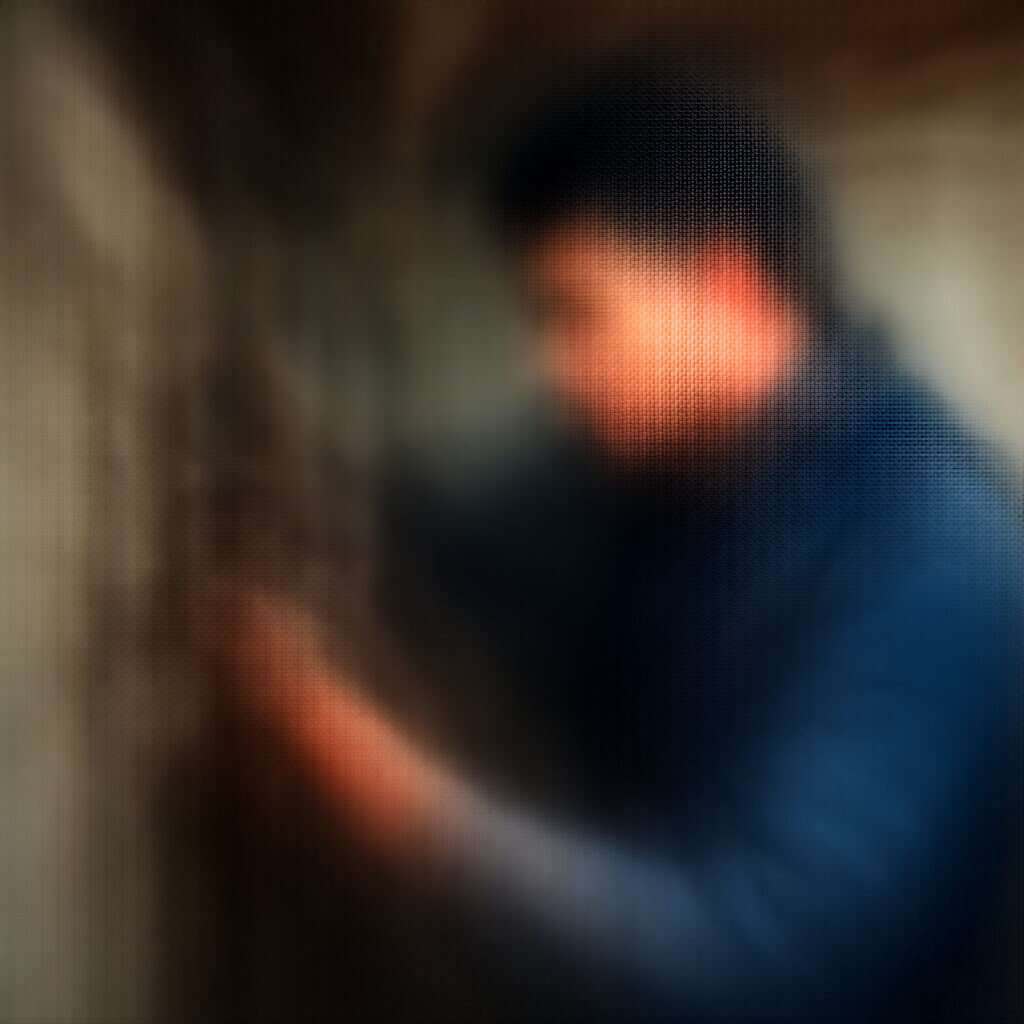} &
\includegraphics[width=0.06\textwidth, height=0.06\textwidth]{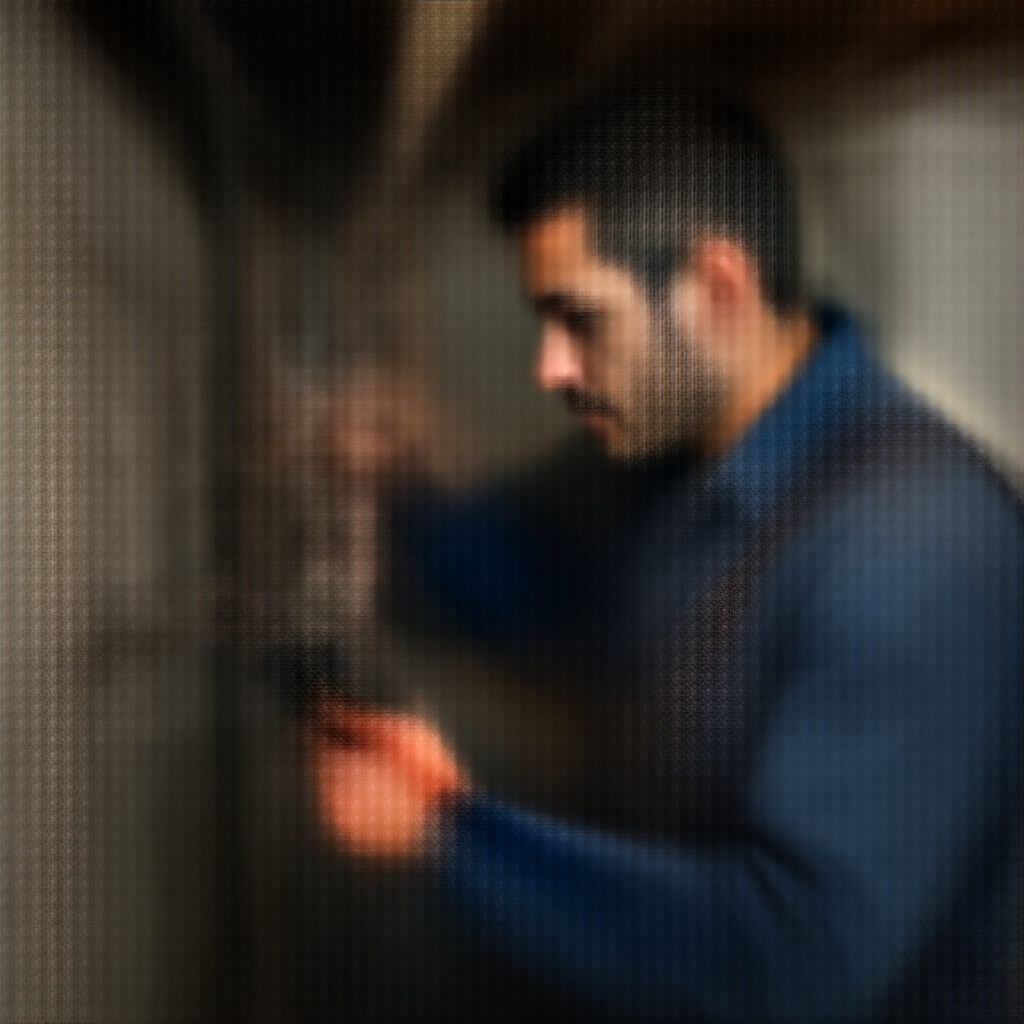} &
\includegraphics[width=0.06\textwidth, height=0.06\textwidth]{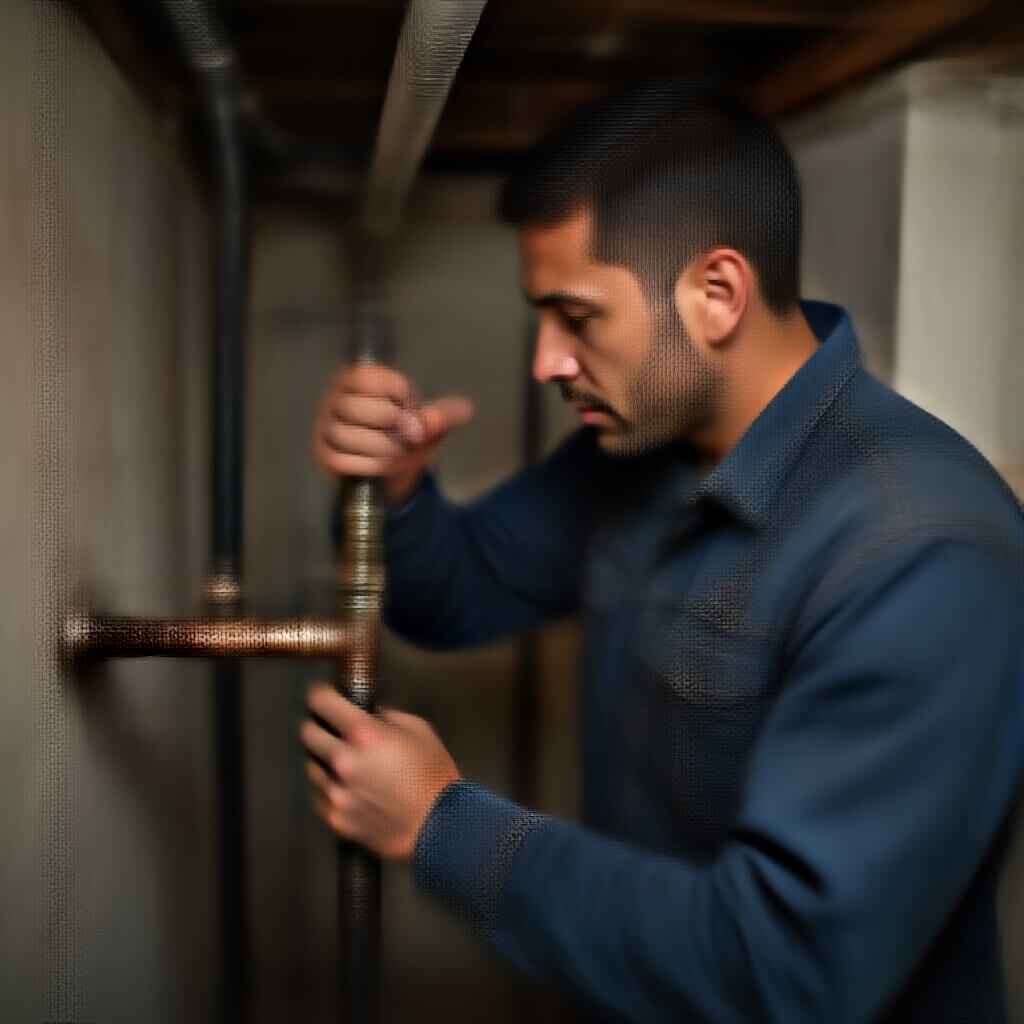} &
\includegraphics[width=0.06\textwidth, height=0.06\textwidth]{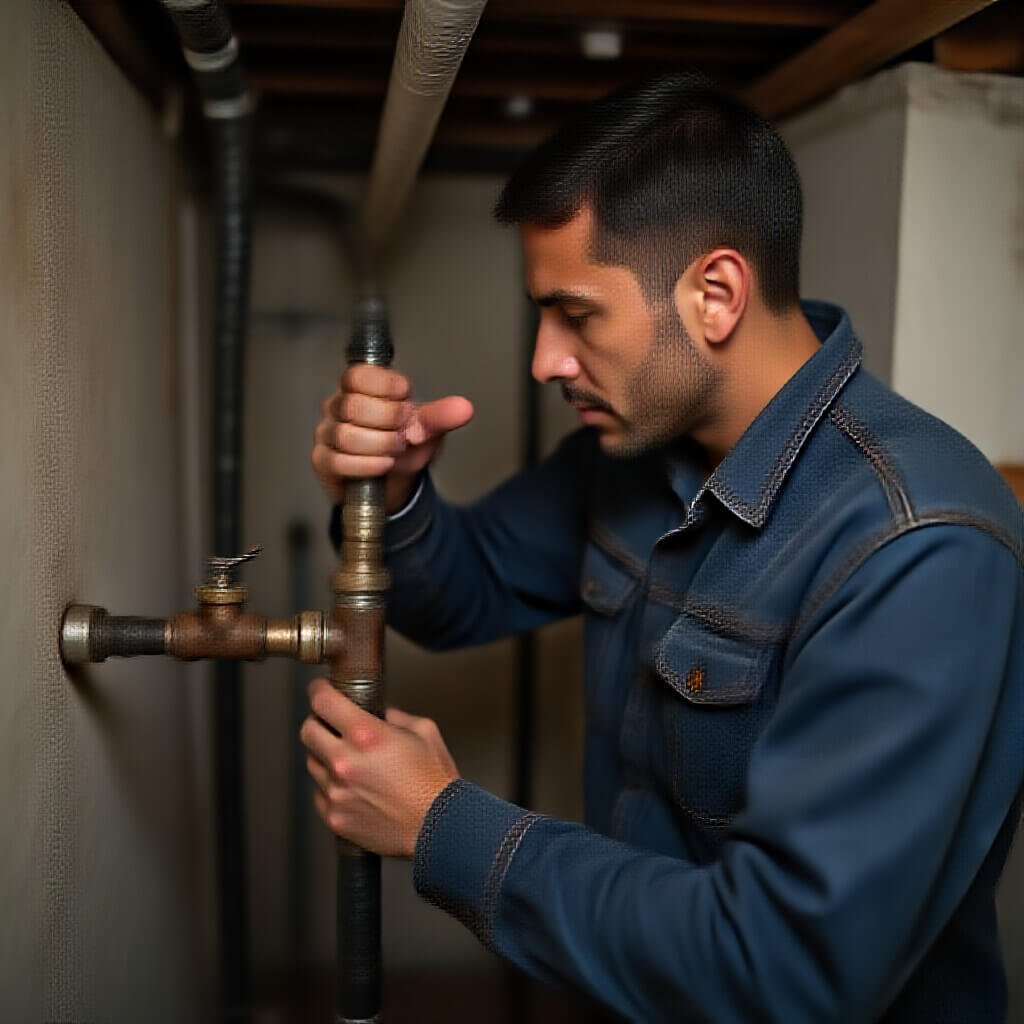}
& \includegraphics[width=0.06\textwidth, height=0.06\textwidth]{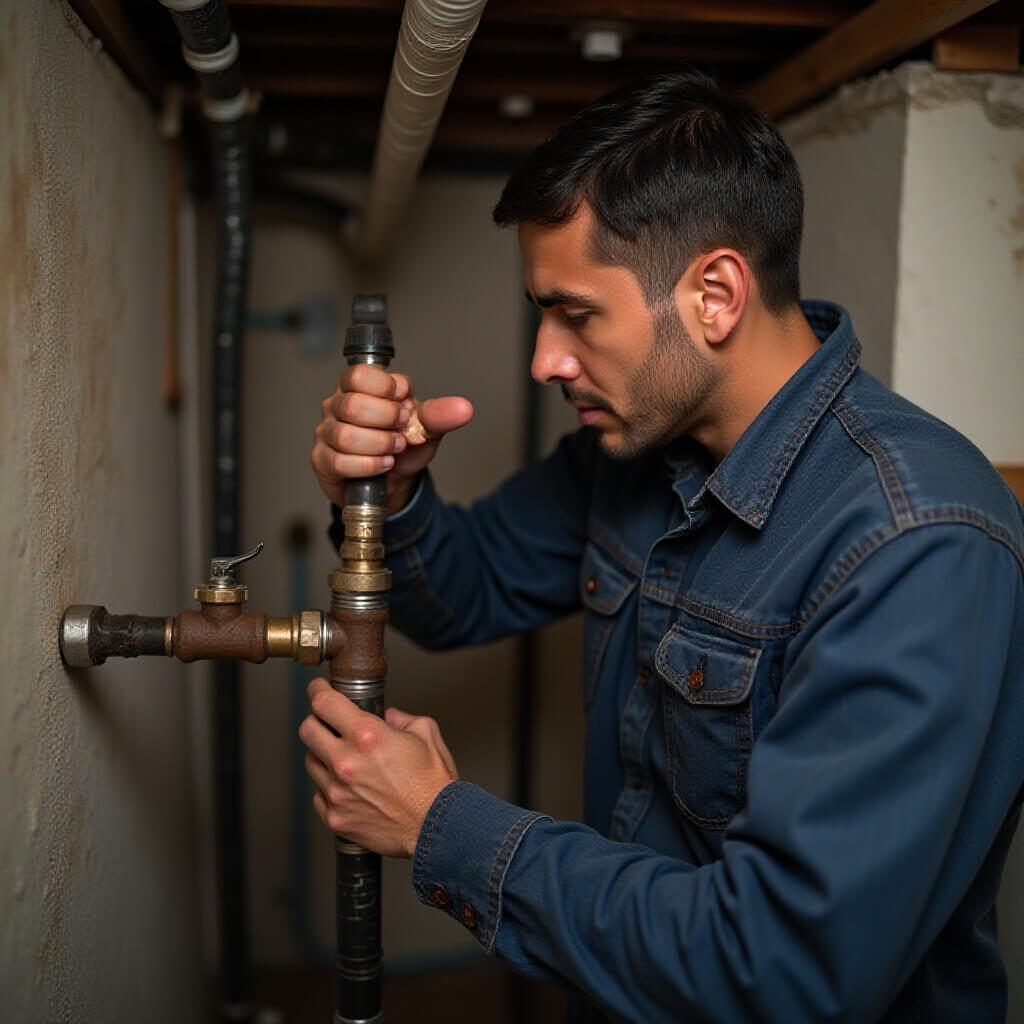} &
\includegraphics[width=0.06\textwidth, height=0.06\textwidth, trim={200, 130, 410, 210}, clip]{images/55.jpg} \\
\rotatebox{90}{\small{\quad Baseline} \hspace{-1cm}} & \includegraphics[width=0.06\textwidth, height=0.06\textwidth]{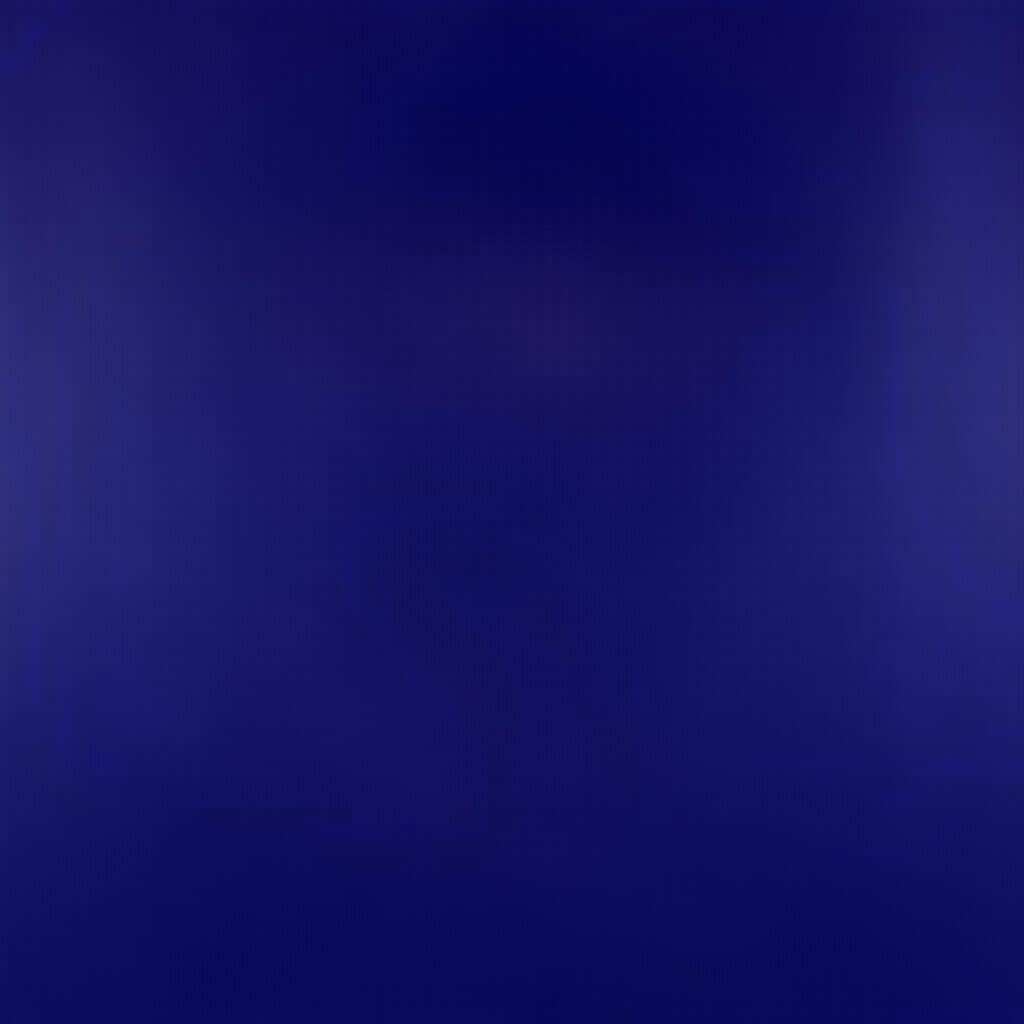} &
\includegraphics[width=0.06\textwidth, height=0.06\textwidth]{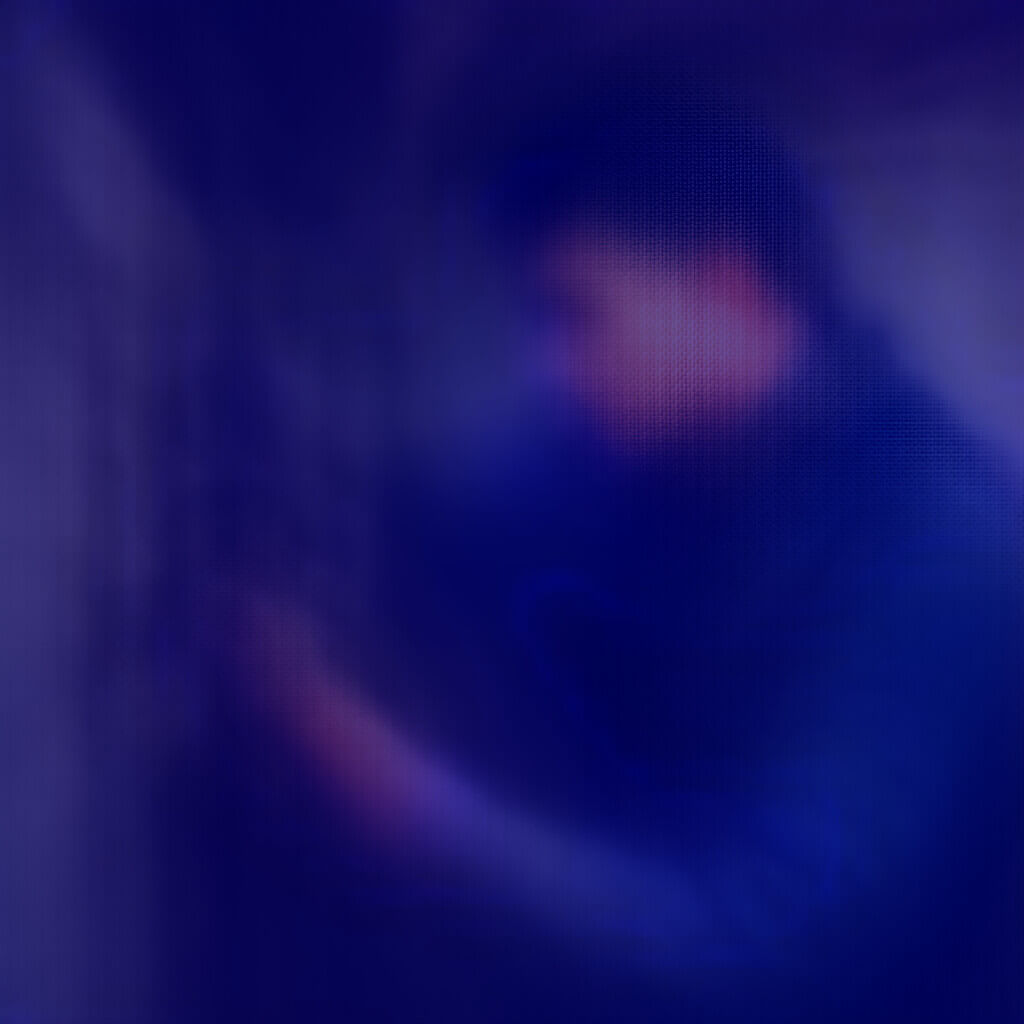} &
\includegraphics[width=0.06\textwidth, height=0.06\textwidth]{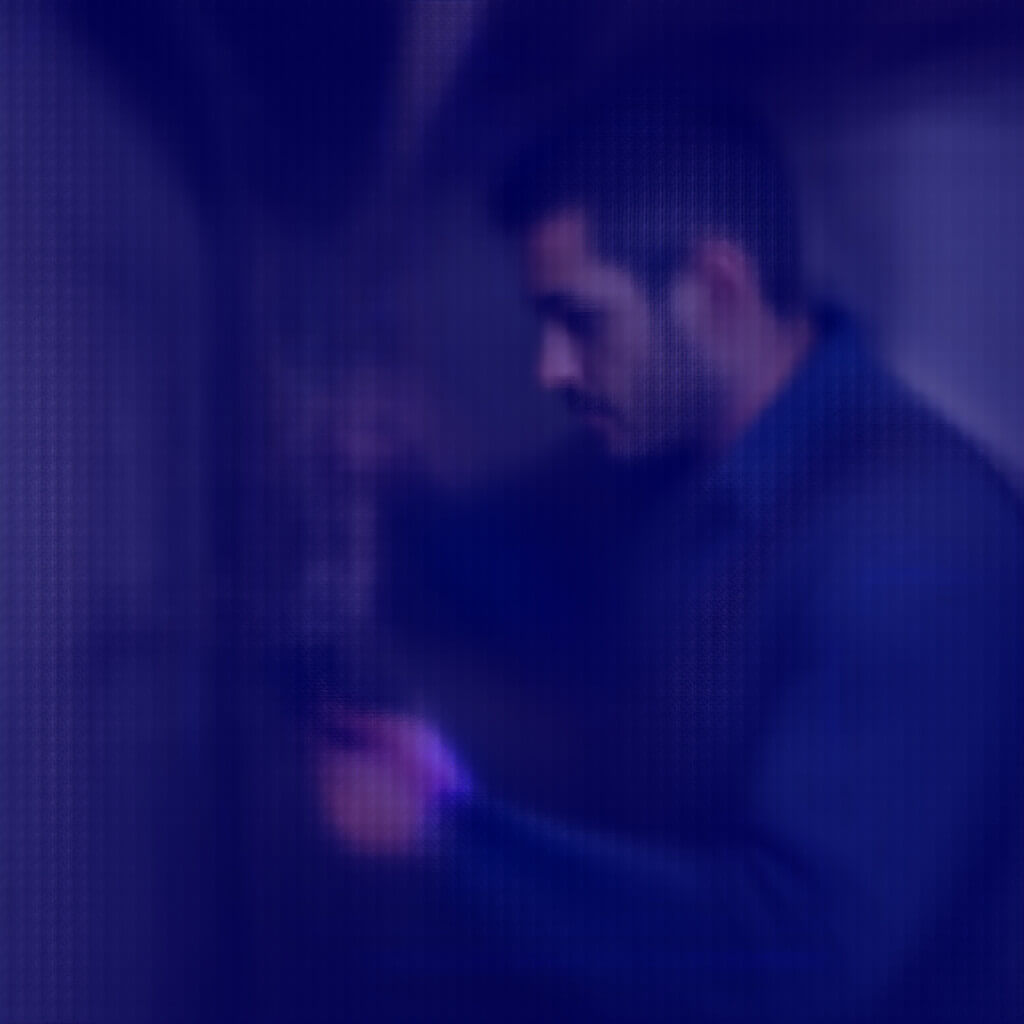} &
\includegraphics[width=0.06\textwidth, height=0.06\textwidth]{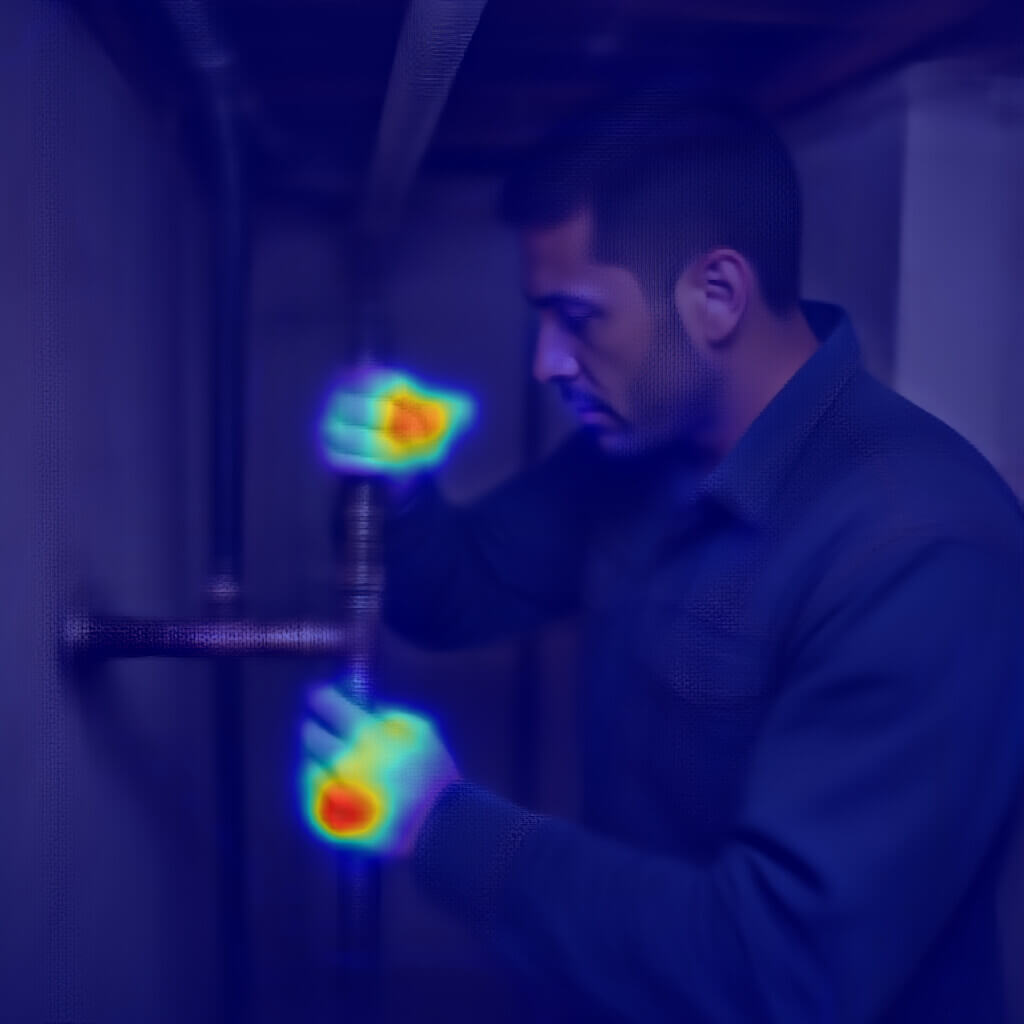} &
\includegraphics[width=0.06\textwidth, height=0.06\textwidth]{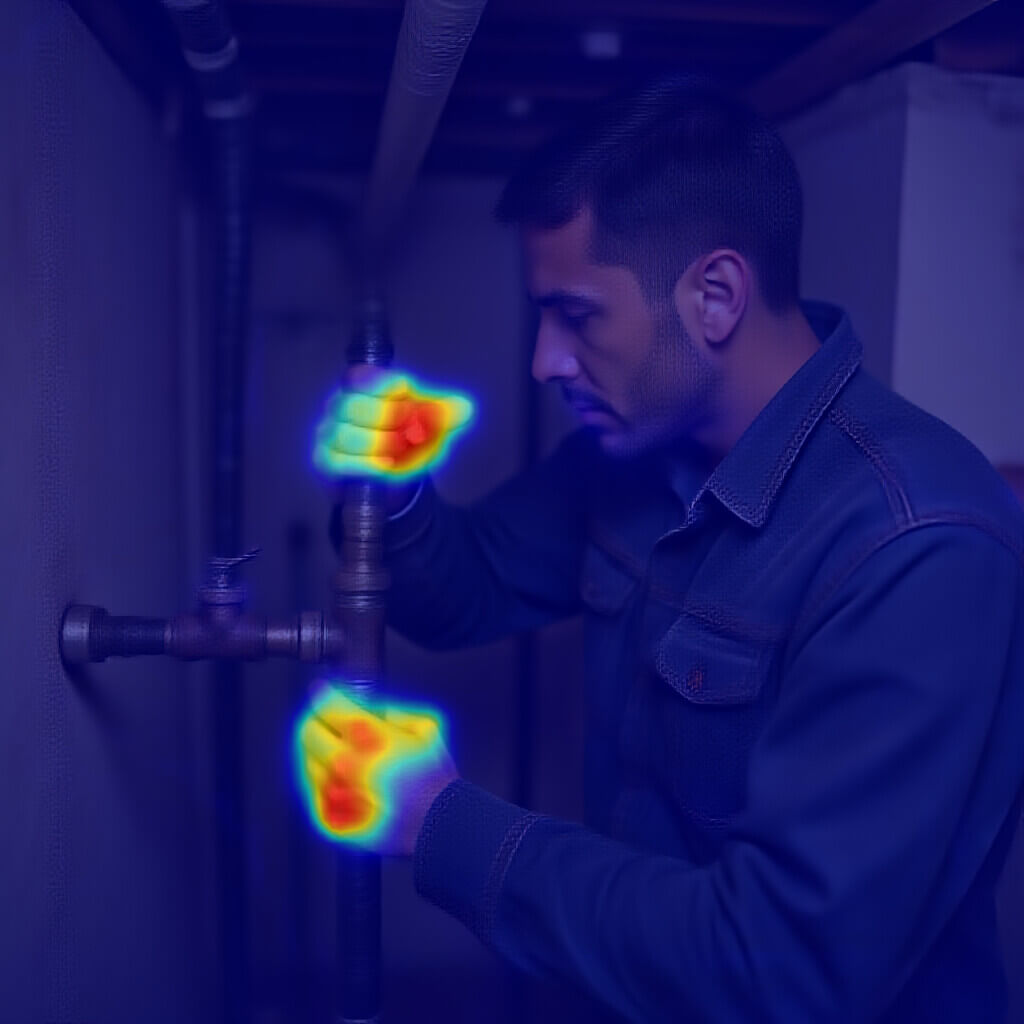}
& \includegraphics[width=0.06\textwidth, height=0.06\textwidth]{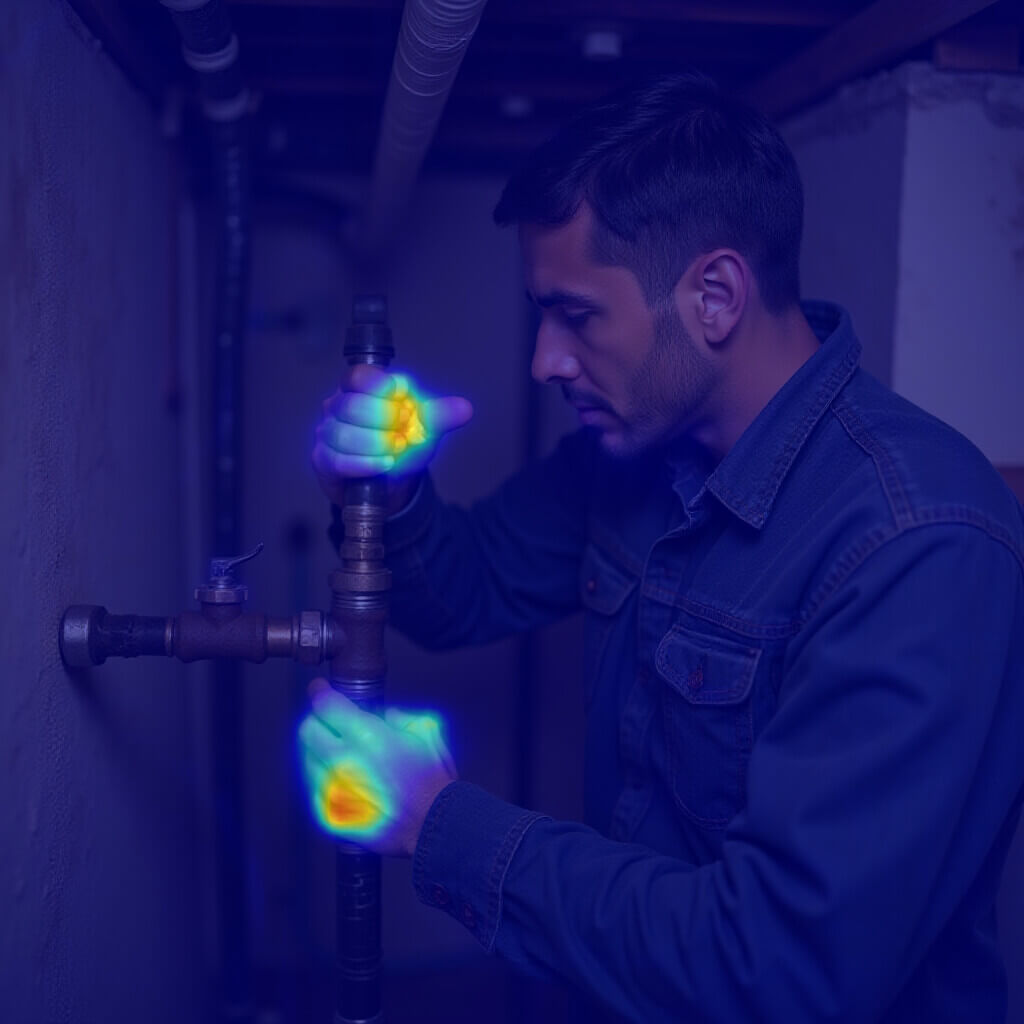} &
\includegraphics[width=0.06\textwidth, height=0.06\textwidth, trim={200, 130, 410, 210}, clip]{images/61.jpg} \\
 &  \includegraphics[width=0.06\textwidth, height=0.06\textwidth]{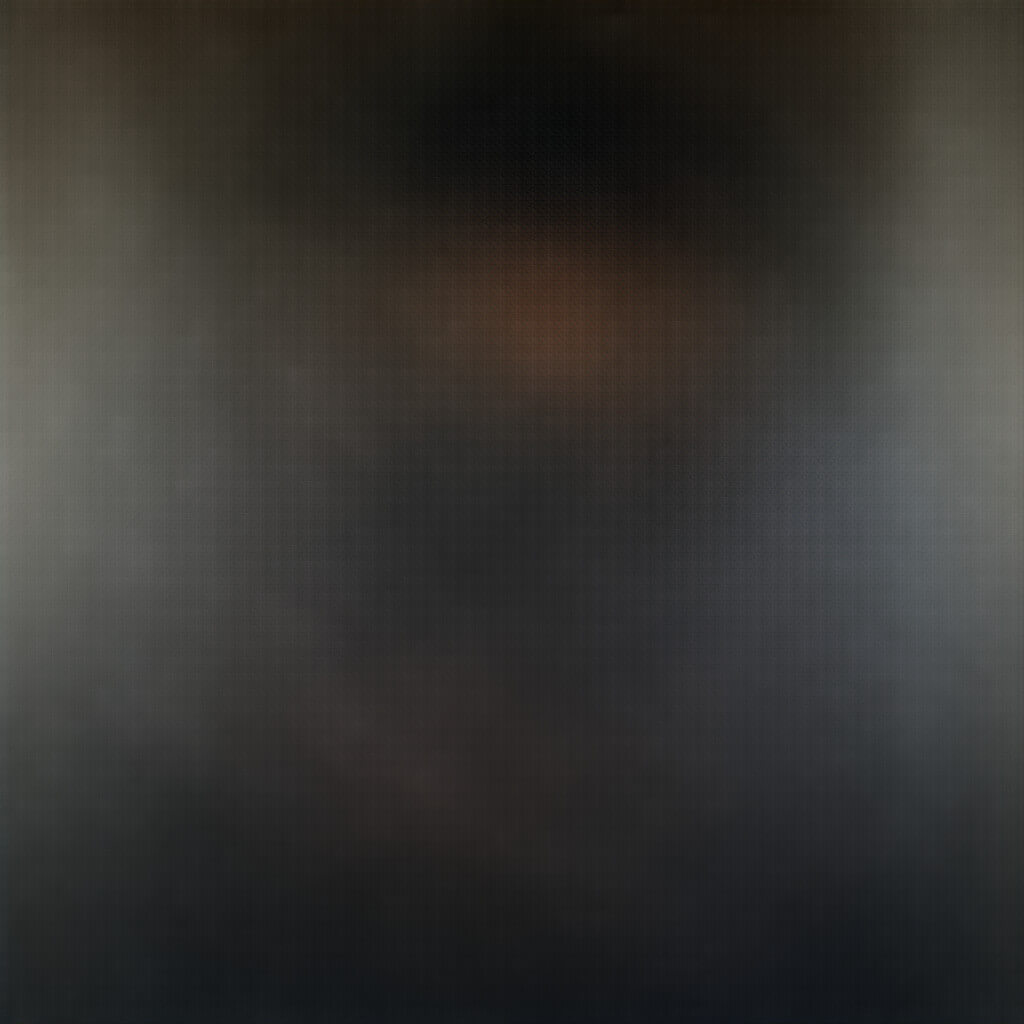} &
\includegraphics[width=0.06\textwidth, height=0.06\textwidth]{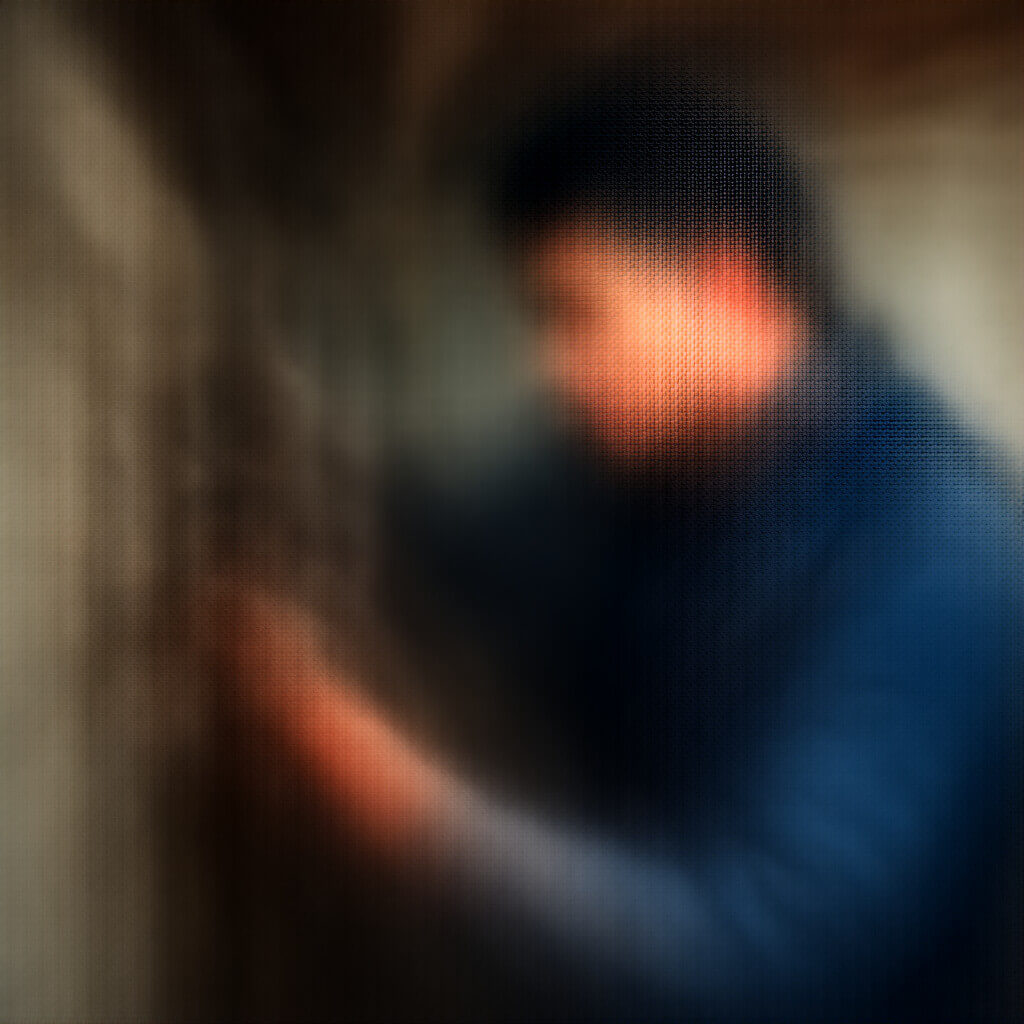} &
\includegraphics[width=0.06\textwidth, height=0.06\textwidth]{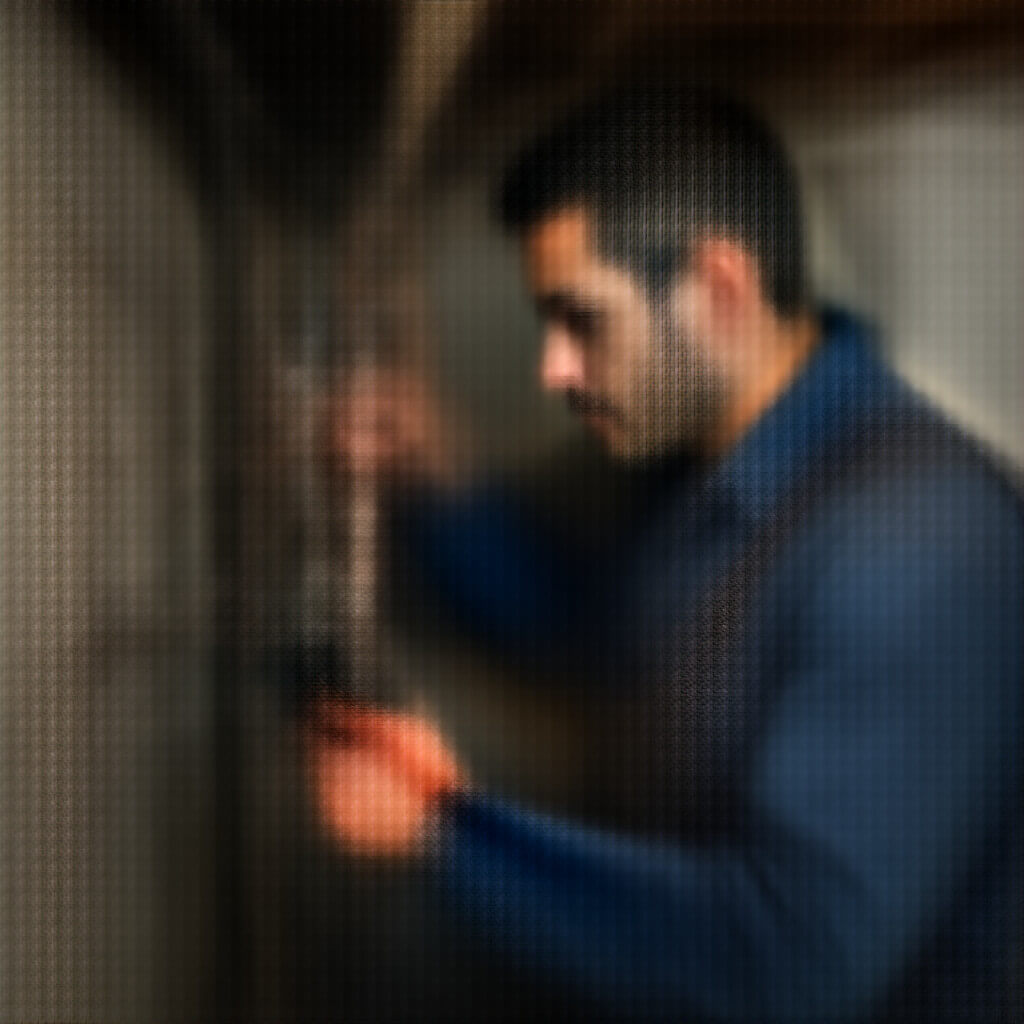} &
\includegraphics[width=0.06\textwidth, height=0.06\textwidth]{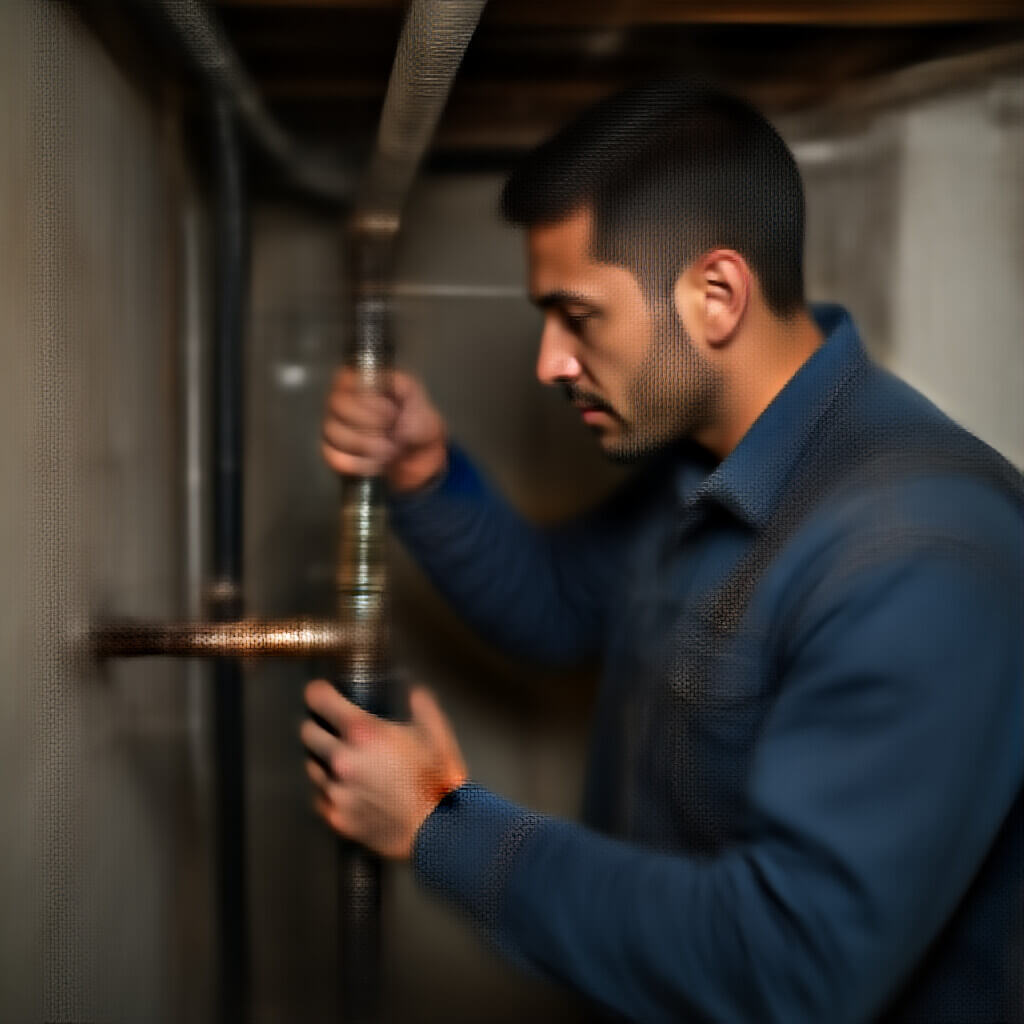} &
\includegraphics[width=0.06\textwidth, height=0.06\textwidth]{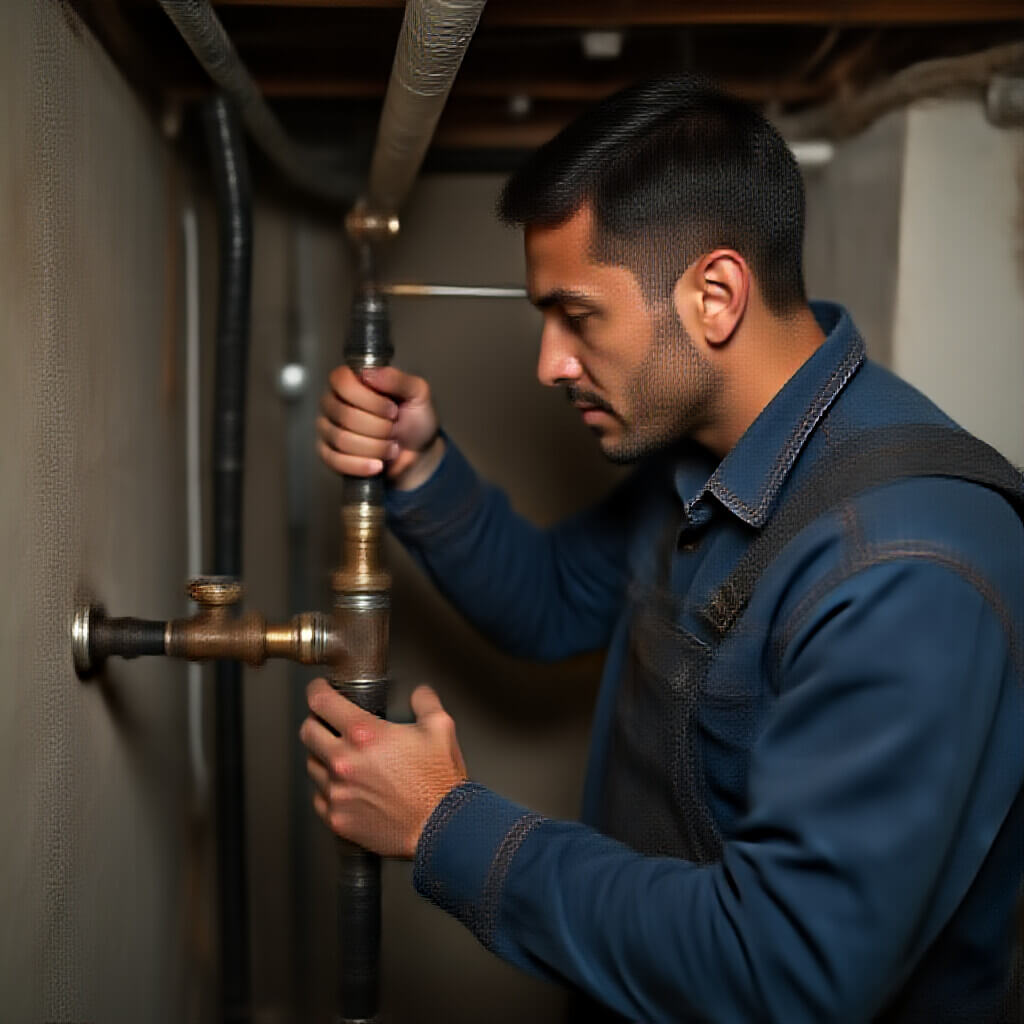}
& \includegraphics[width=0.06\textwidth, height=0.06\textwidth]{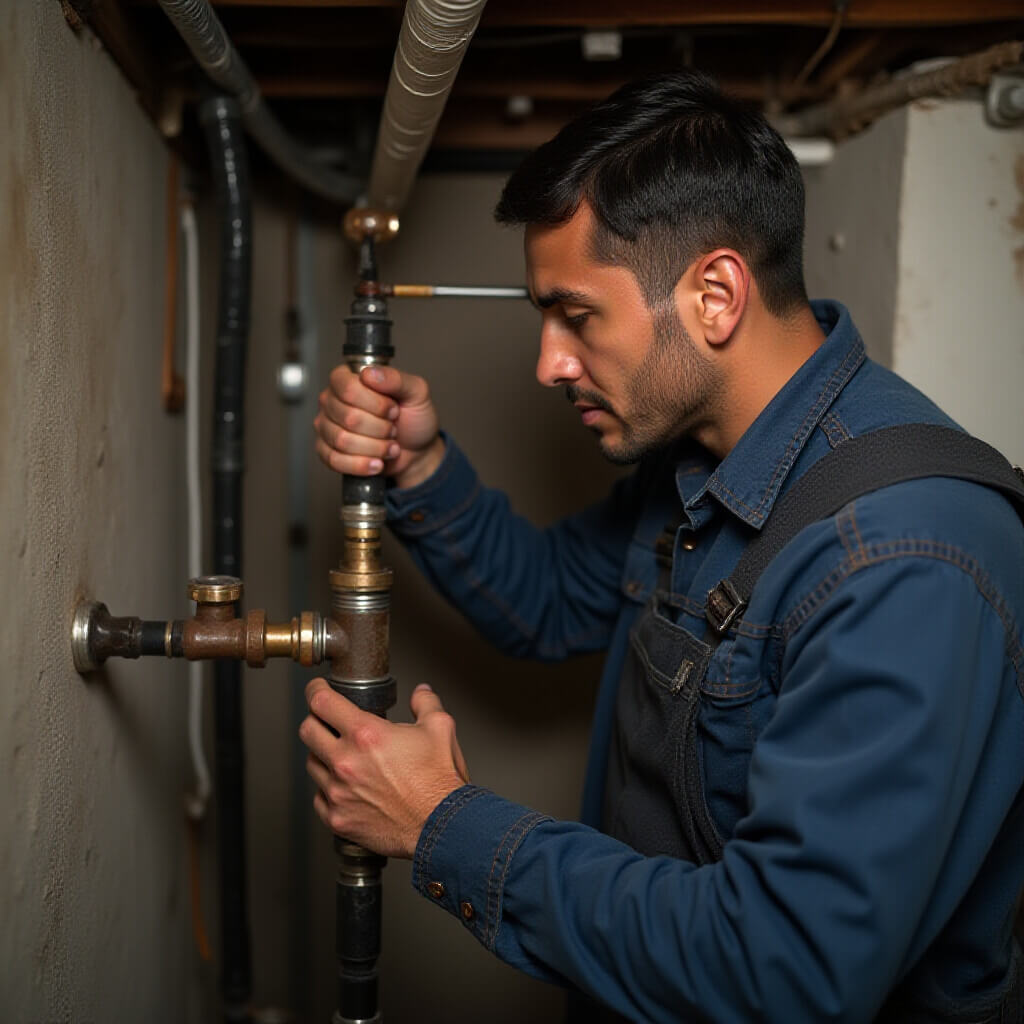} &
\includegraphics[width=0.06\textwidth, height=0.06\textwidth, trim={200, 130, 410, 210}, clip]{images/67.jpg} \\
\rotatebox{90}{\small{\quad Our w norm} \hspace{-1cm}} & \includegraphics[width=0.06\textwidth, height=0.06\textwidth]{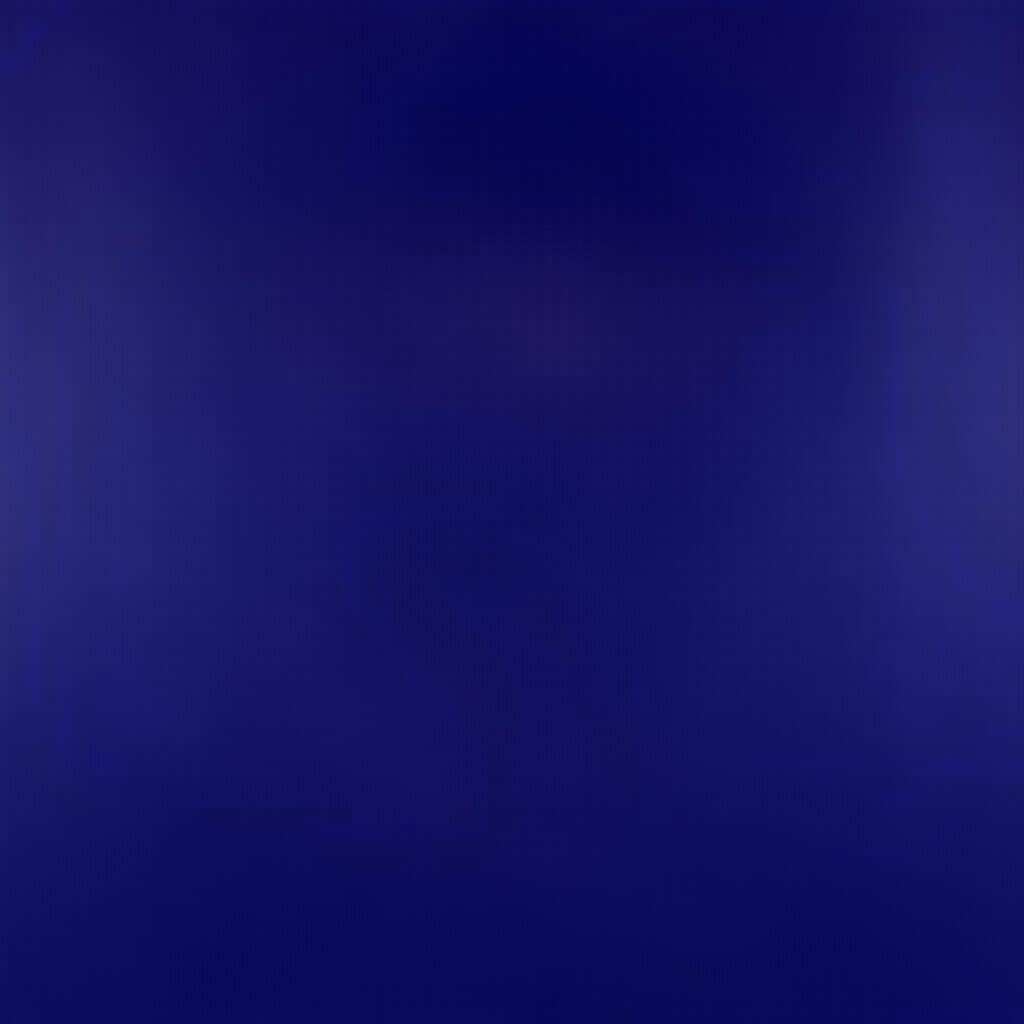} &
\includegraphics[width=0.06\textwidth, height=0.06\textwidth]{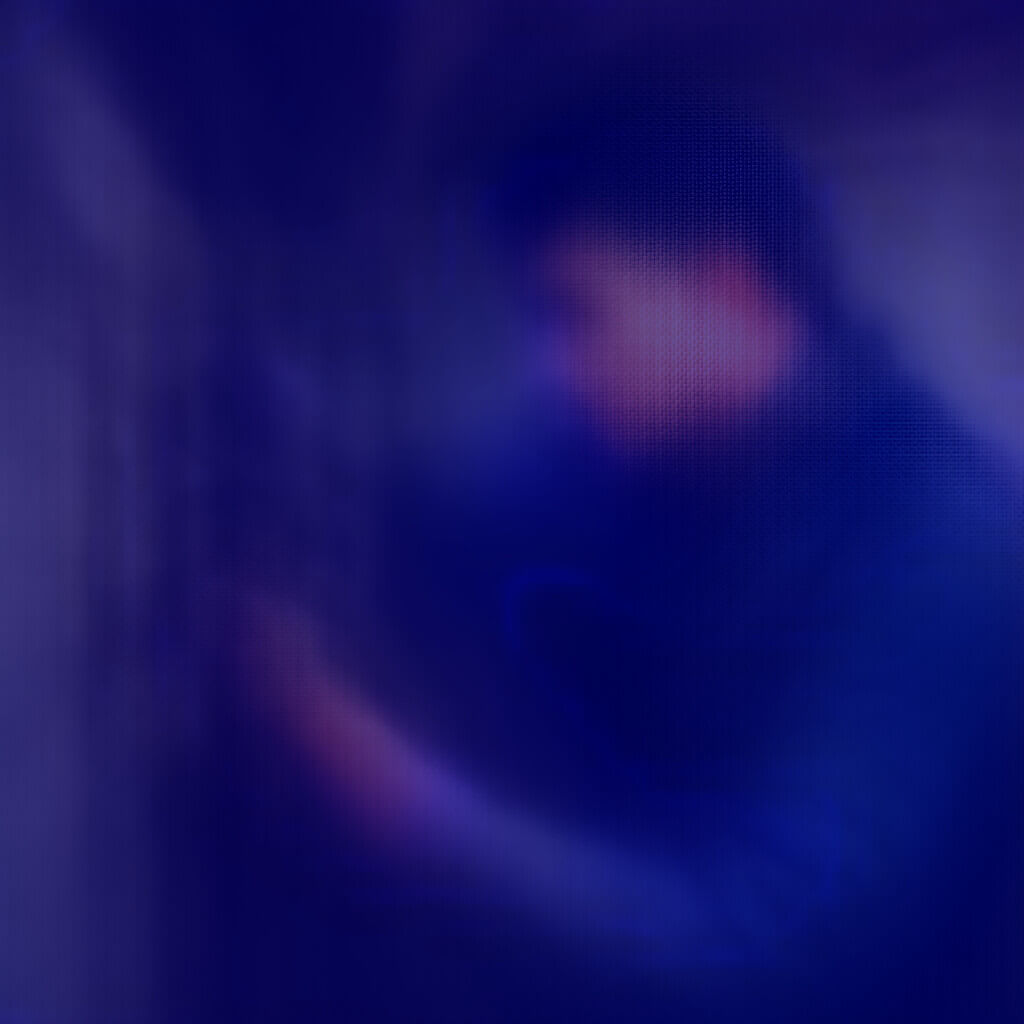} &
\includegraphics[width=0.06\textwidth, height=0.06\textwidth]{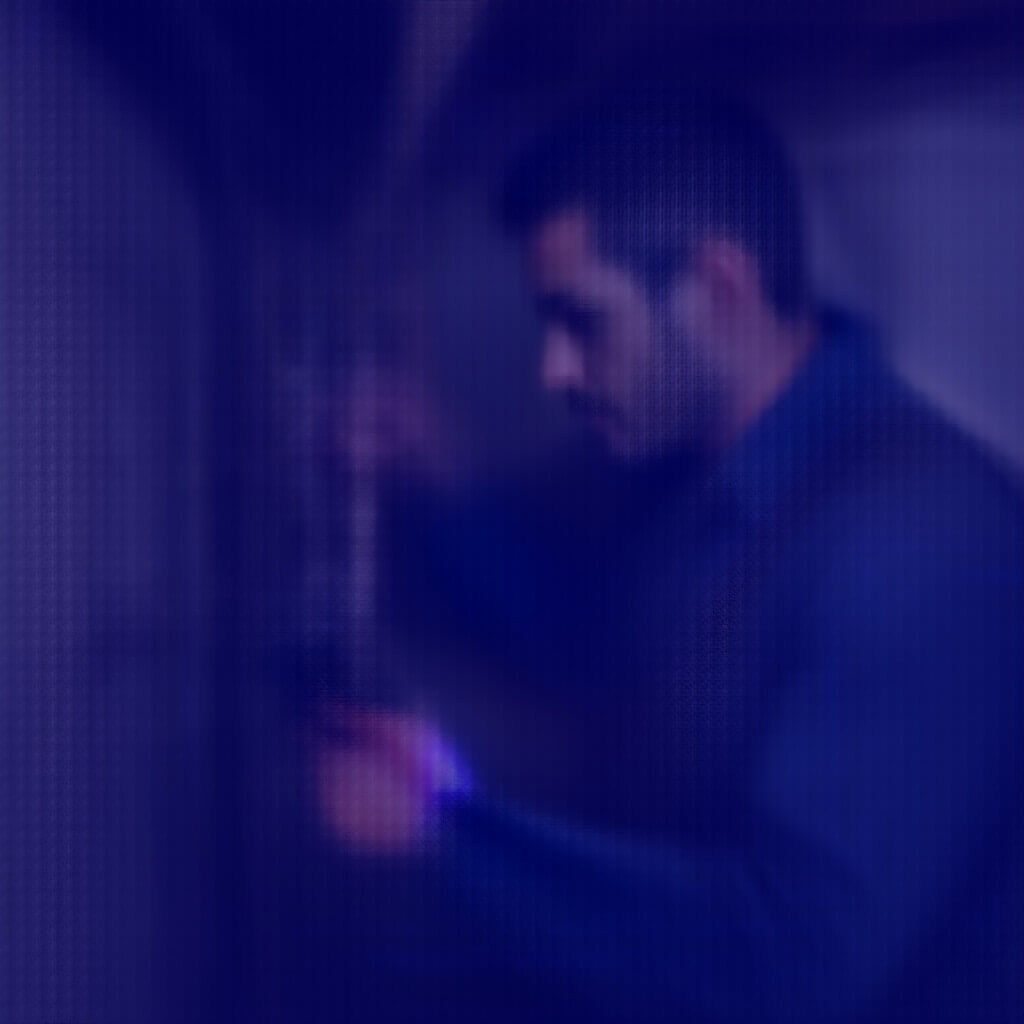} &
\includegraphics[width=0.06\textwidth, height=0.06\textwidth]{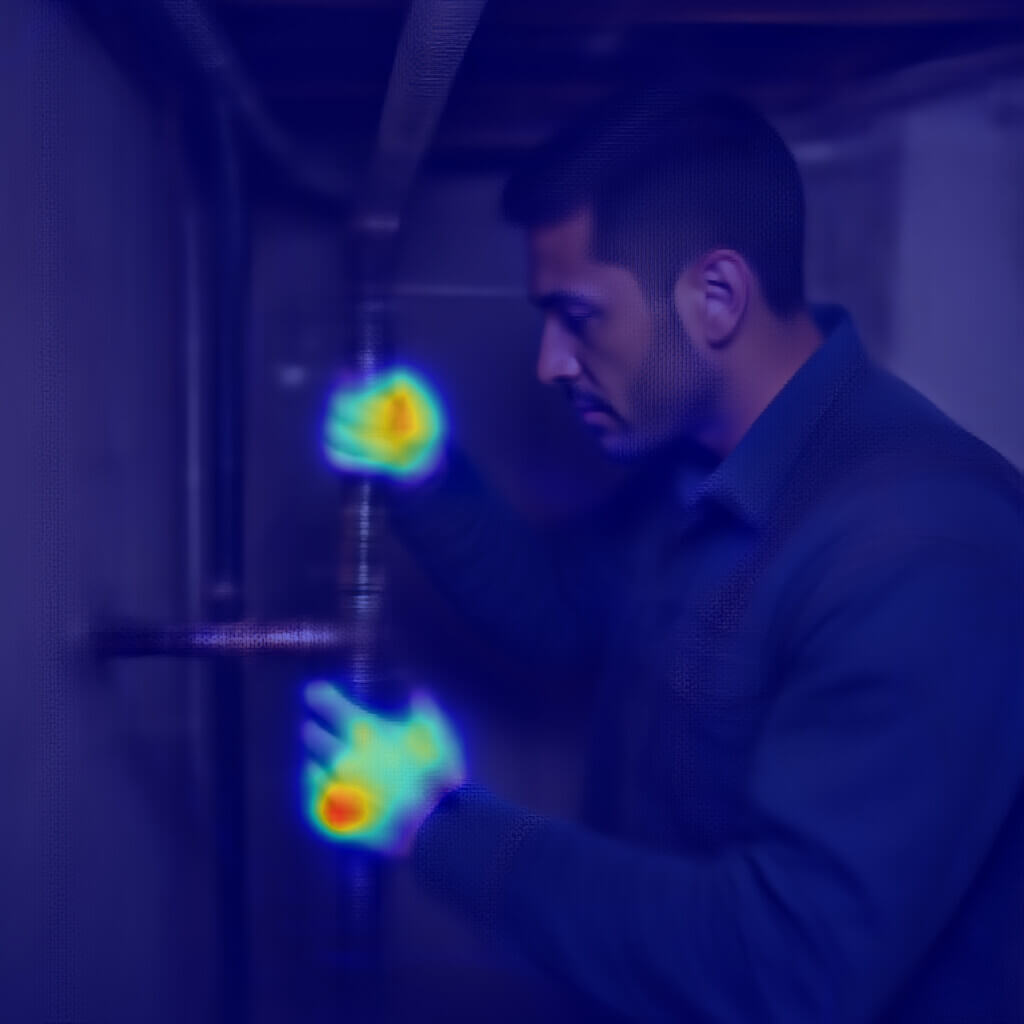} &
\includegraphics[width=0.06\textwidth, height=0.06\textwidth]{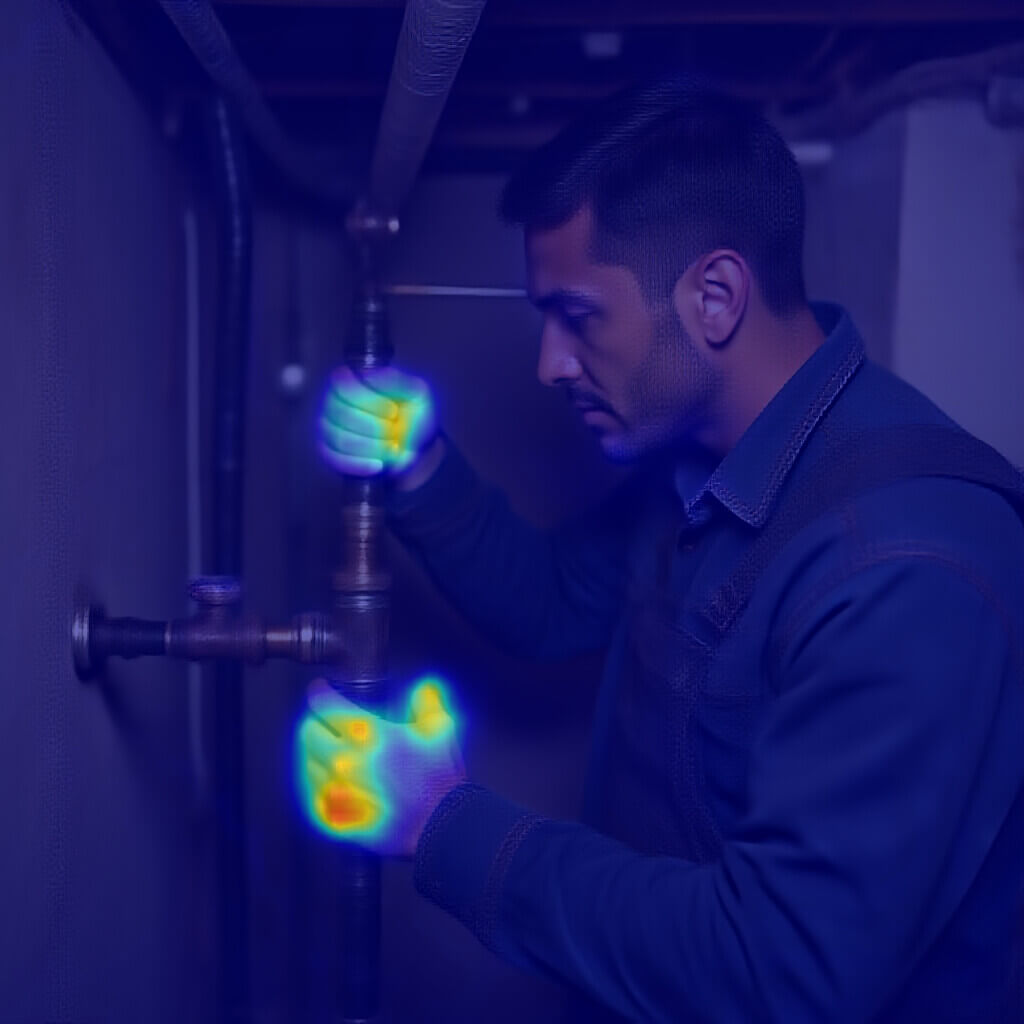}
& \includegraphics[width=0.06\textwidth, height=0.06\textwidth]{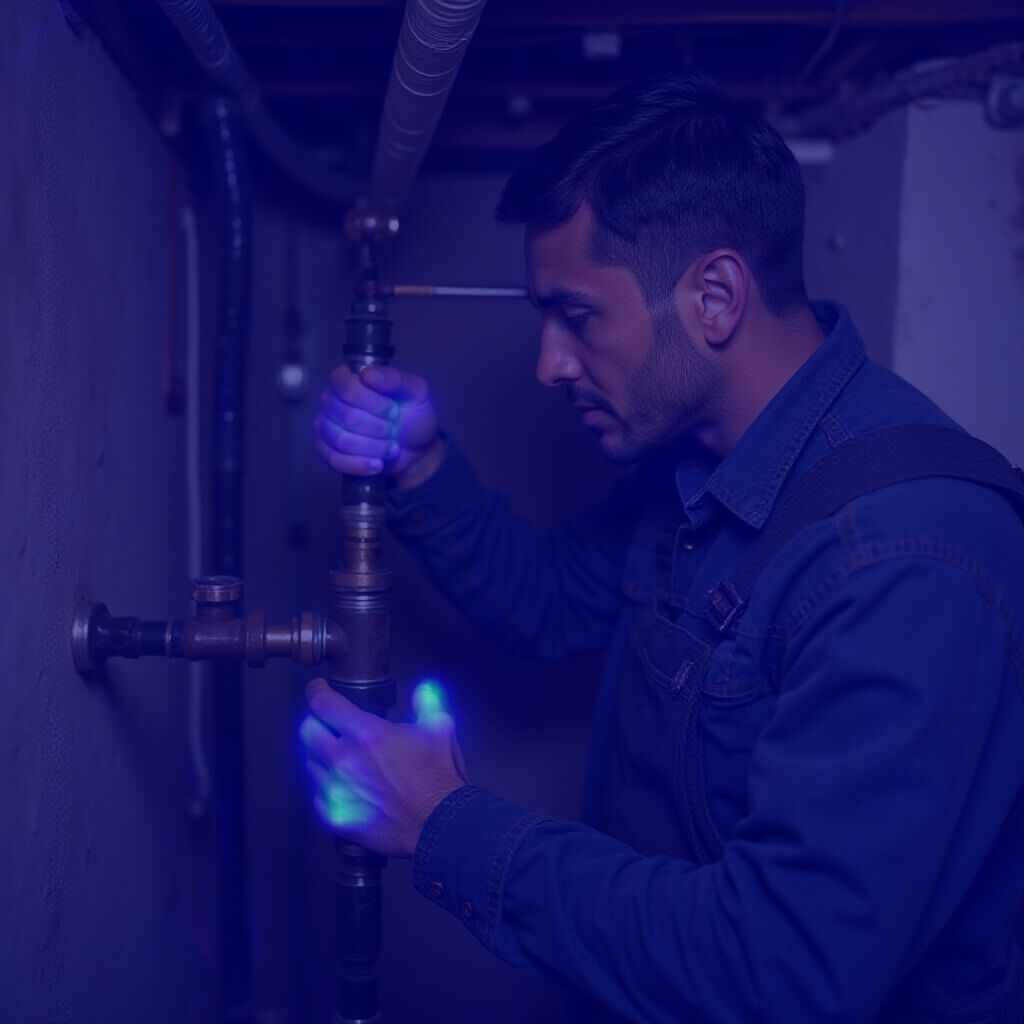} &
\includegraphics[width=0.06\textwidth, height=0.06\textwidth, trim={200, 130, 410, 210}, clip]{images/73.jpg} \\ 
 & \includegraphics[width=0.06\textwidth, height=0.06\textwidth]{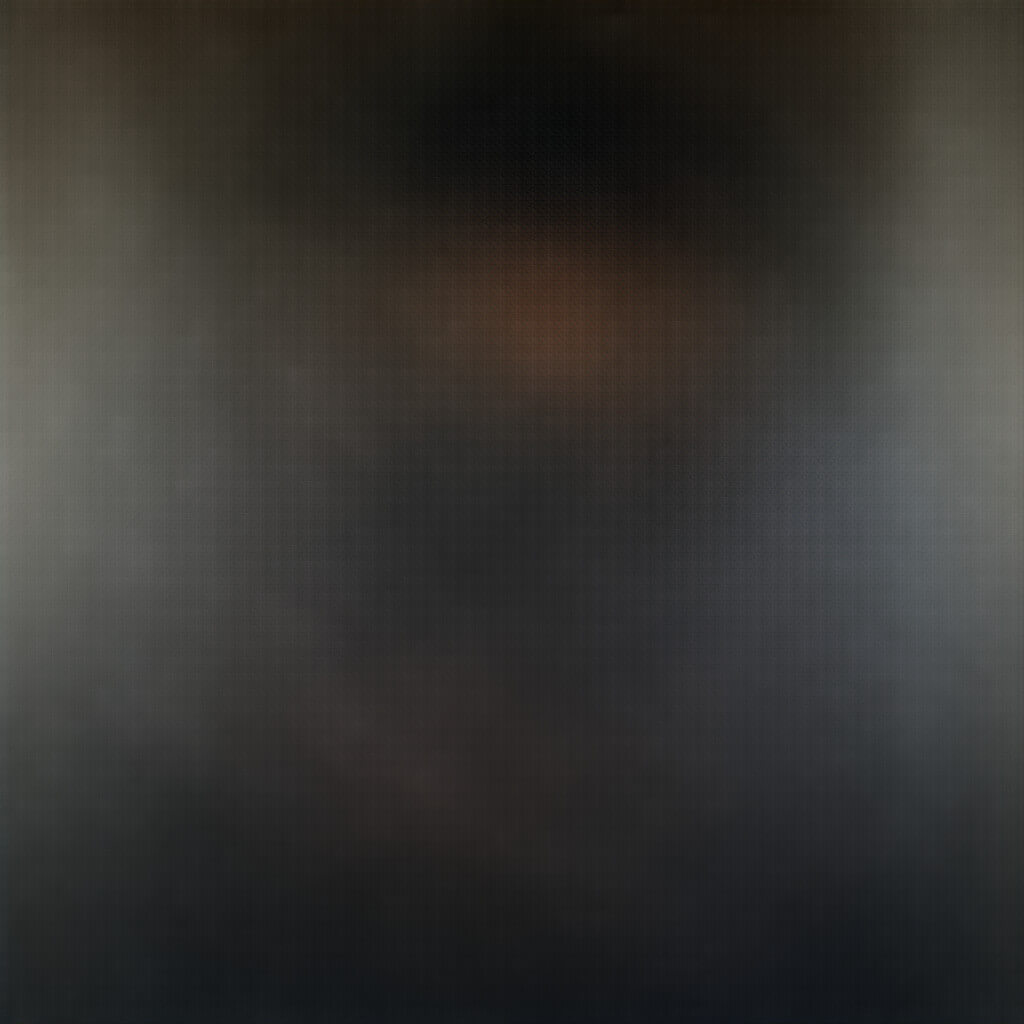} &
\includegraphics[width=0.06\textwidth, height=0.06\textwidth]{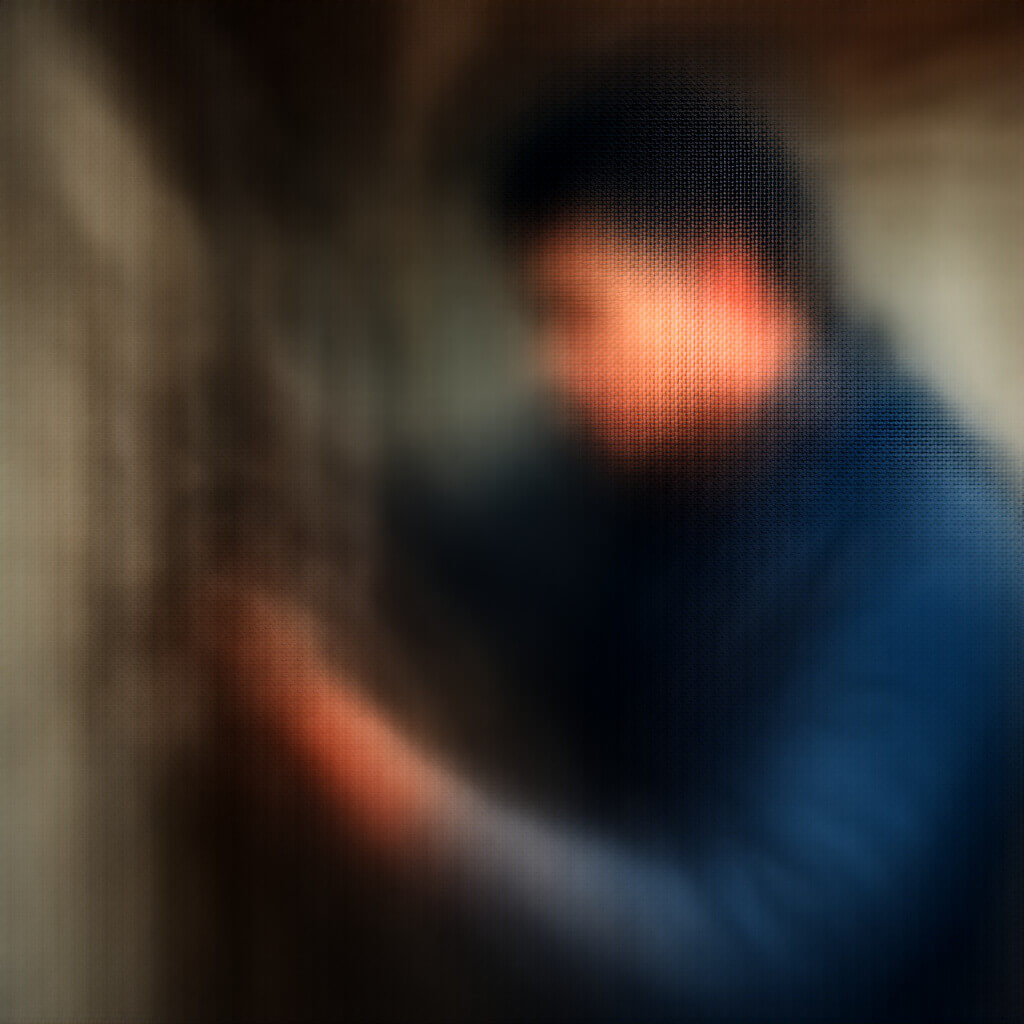} &
\includegraphics[width=0.06\textwidth, height=0.06\textwidth]{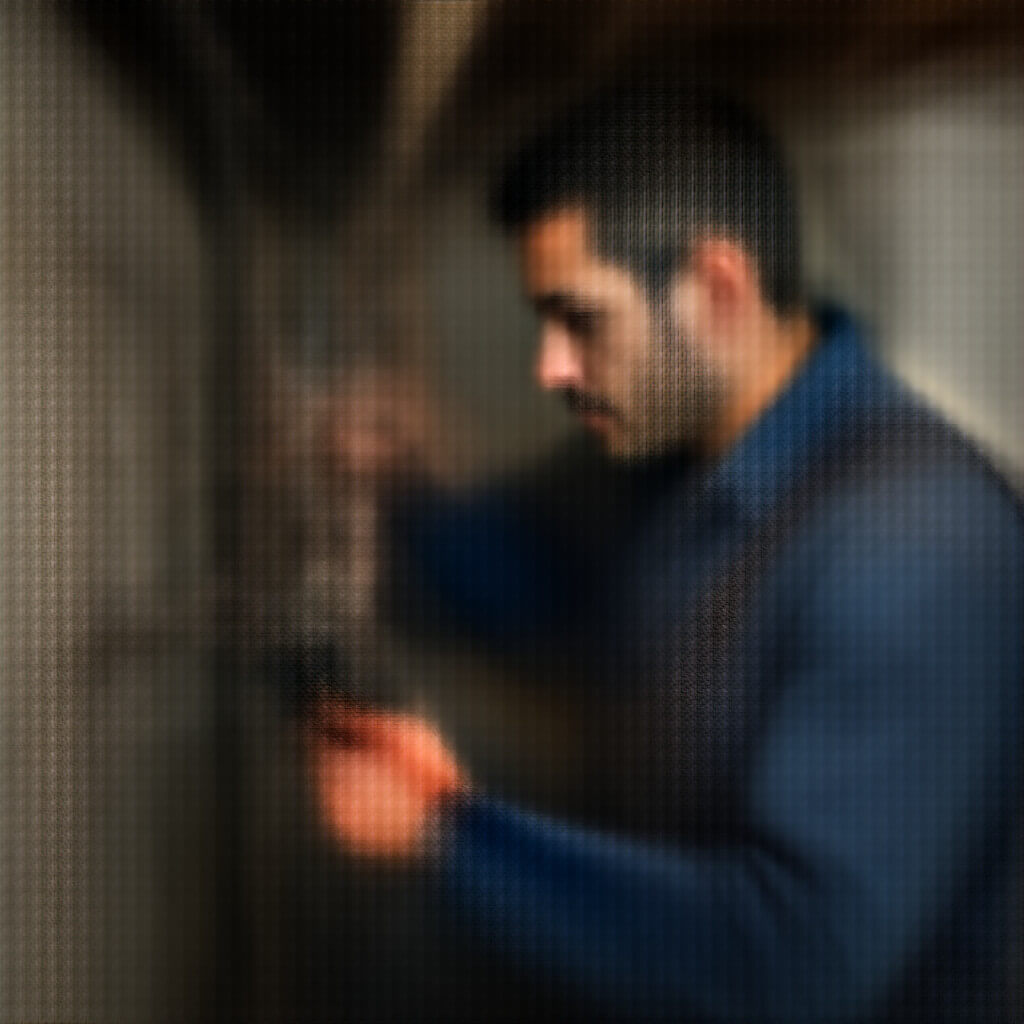} &
\includegraphics[width=0.06\textwidth, height=0.06\textwidth]{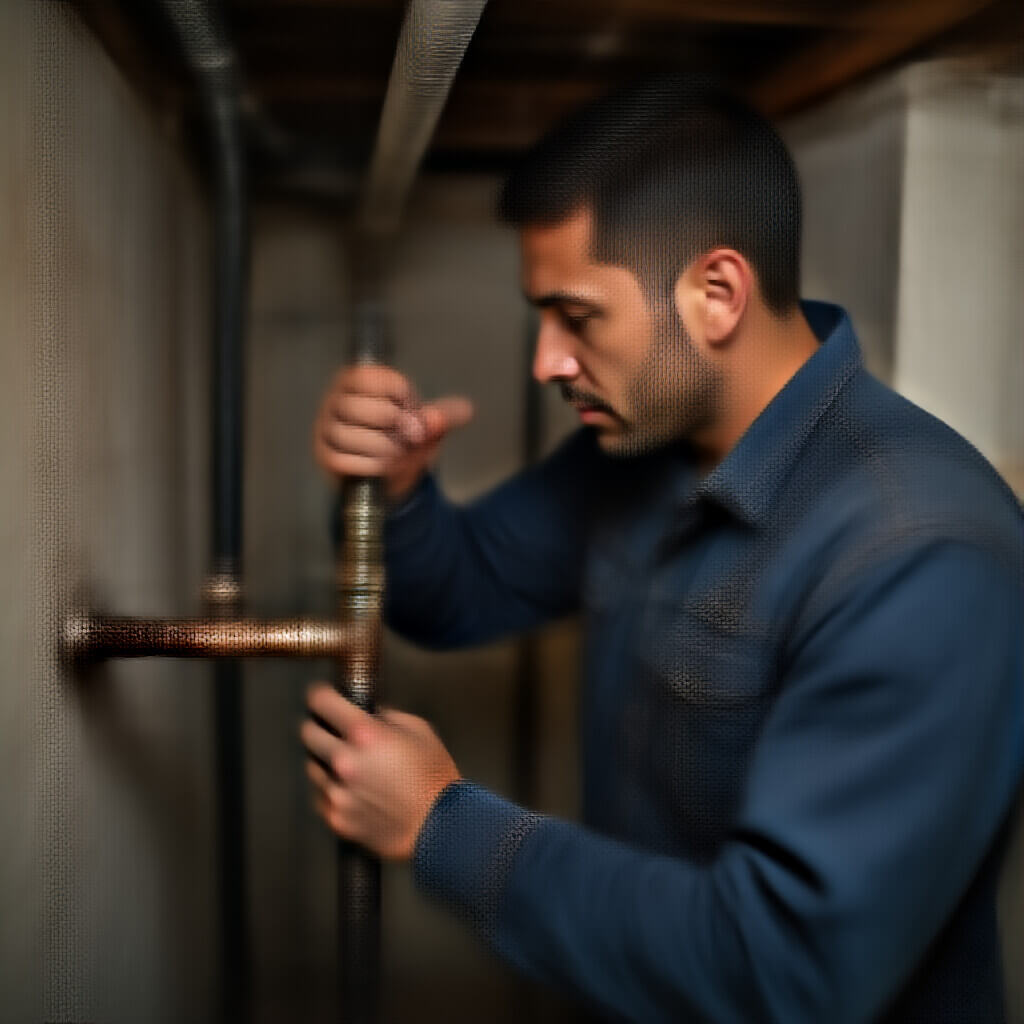} &
\includegraphics[width=0.06\textwidth, height=0.06\textwidth]{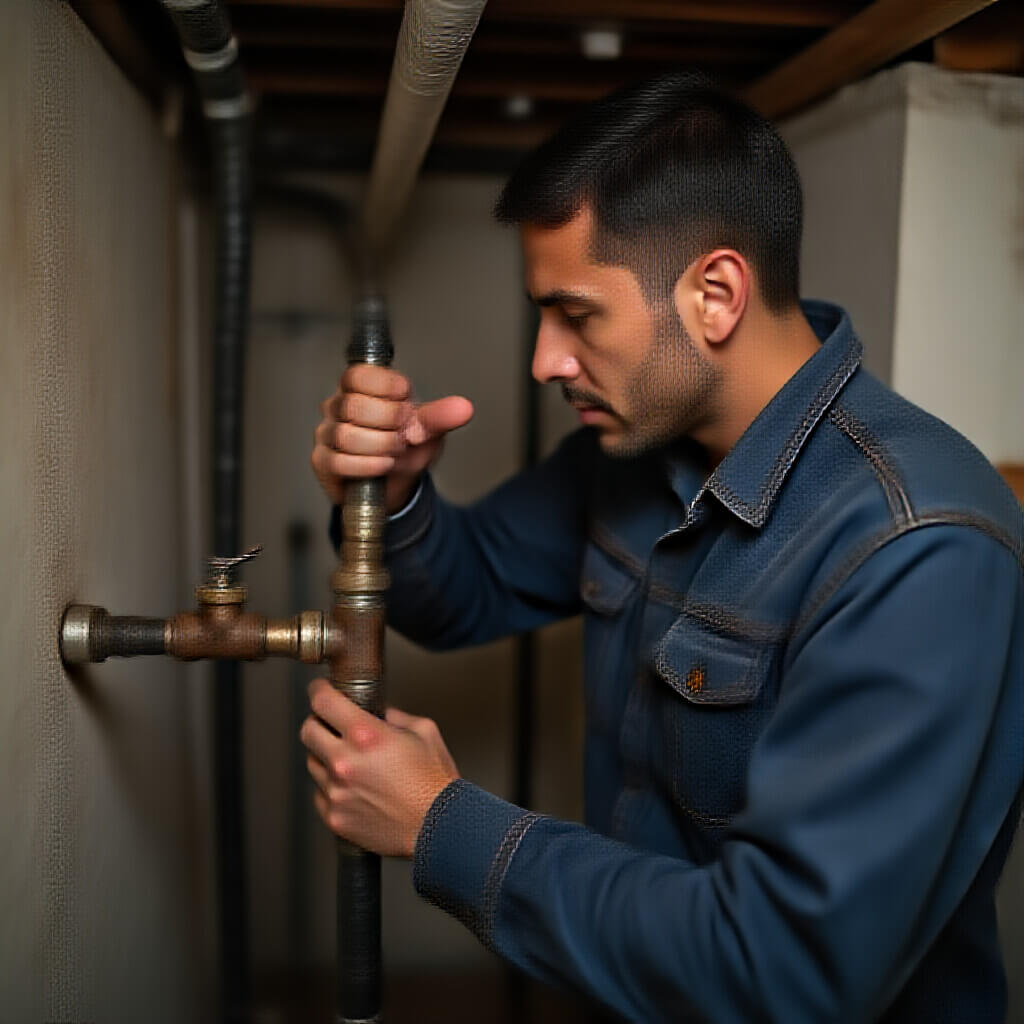}
& \includegraphics[width=0.06\textwidth, height=0.06\textwidth]{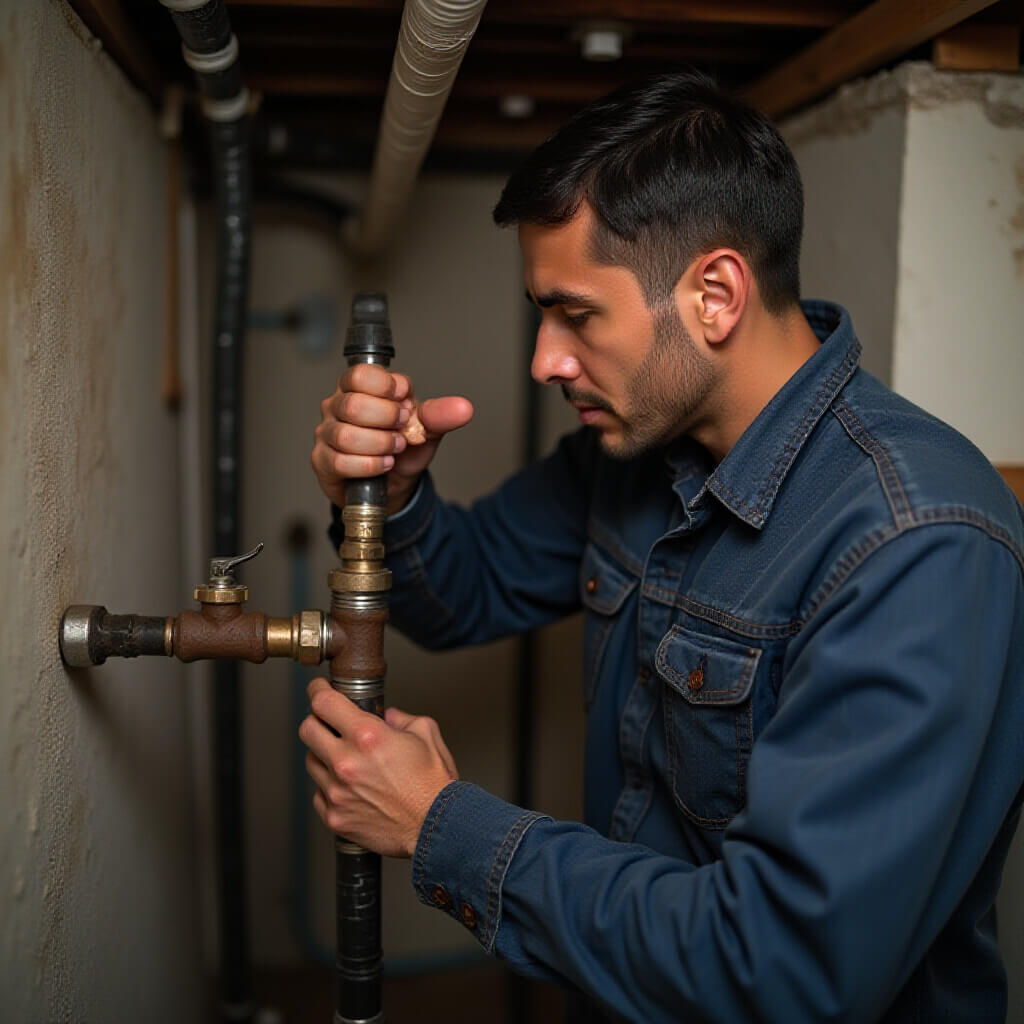} &
\includegraphics[width=0.06\textwidth, height=0.06\textwidth, trim={200, 130, 410, 210}, clip]{images/79.jpg}\\
\rotatebox{90}{\small{\quad Our w/o norm} \hspace{-1cm}} & \includegraphics[width=0.06\textwidth, height=0.06\textwidth]{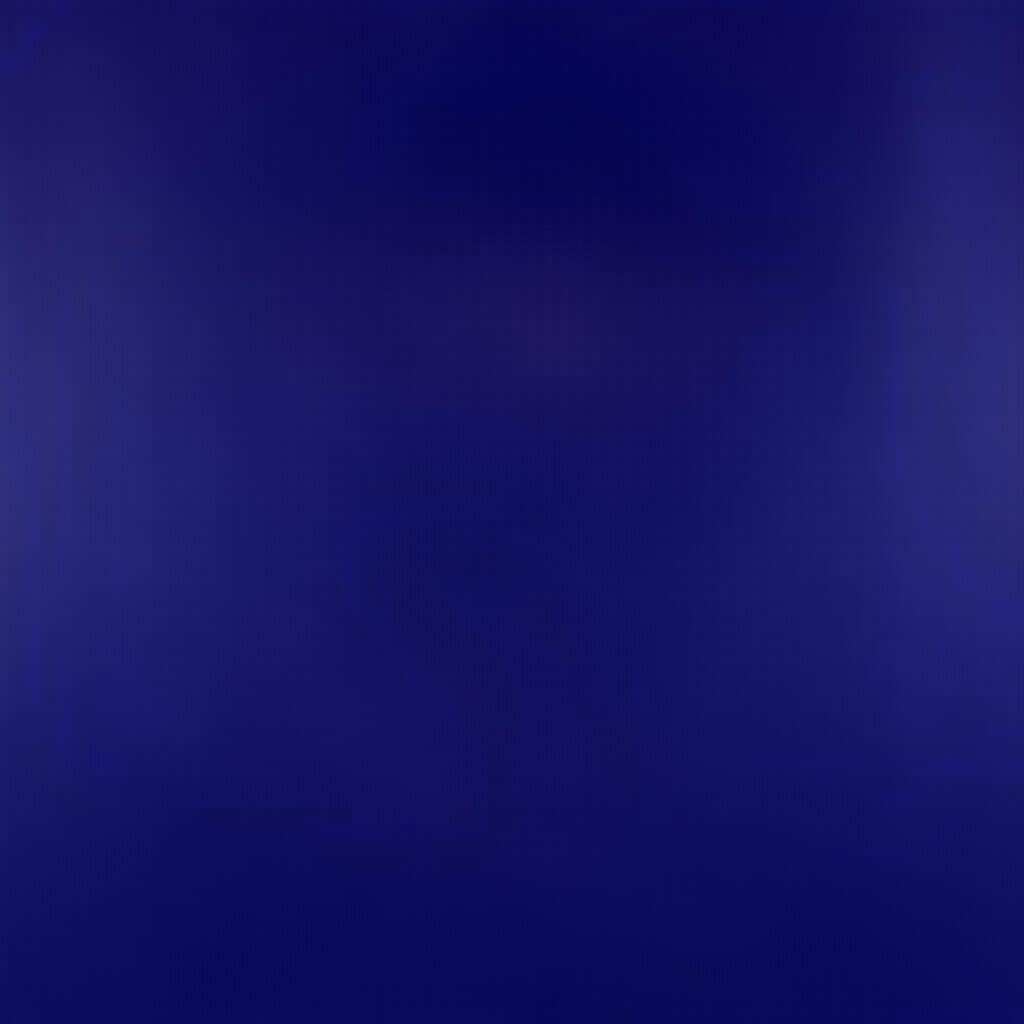} &
\includegraphics[width=0.06\textwidth, height=0.06\textwidth]{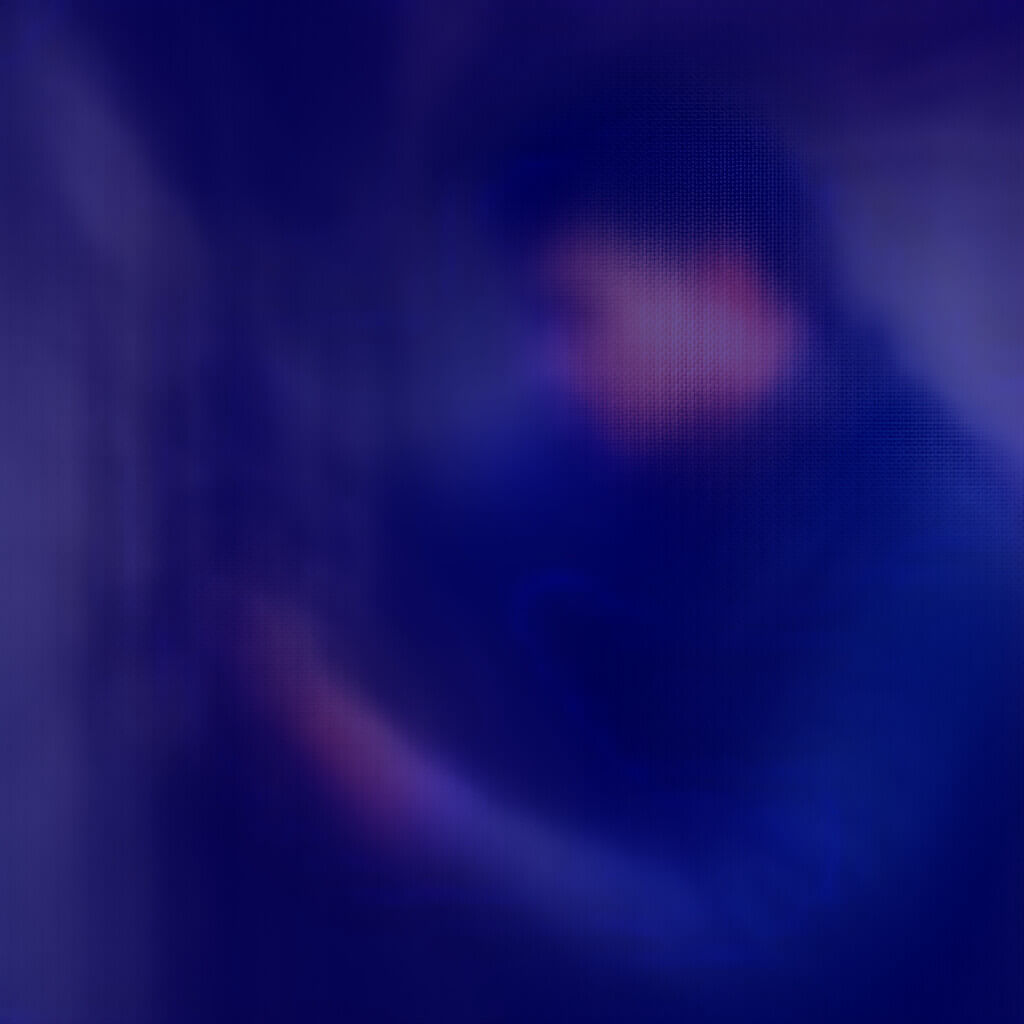} &
\includegraphics[width=0.06\textwidth, height=0.06\textwidth]{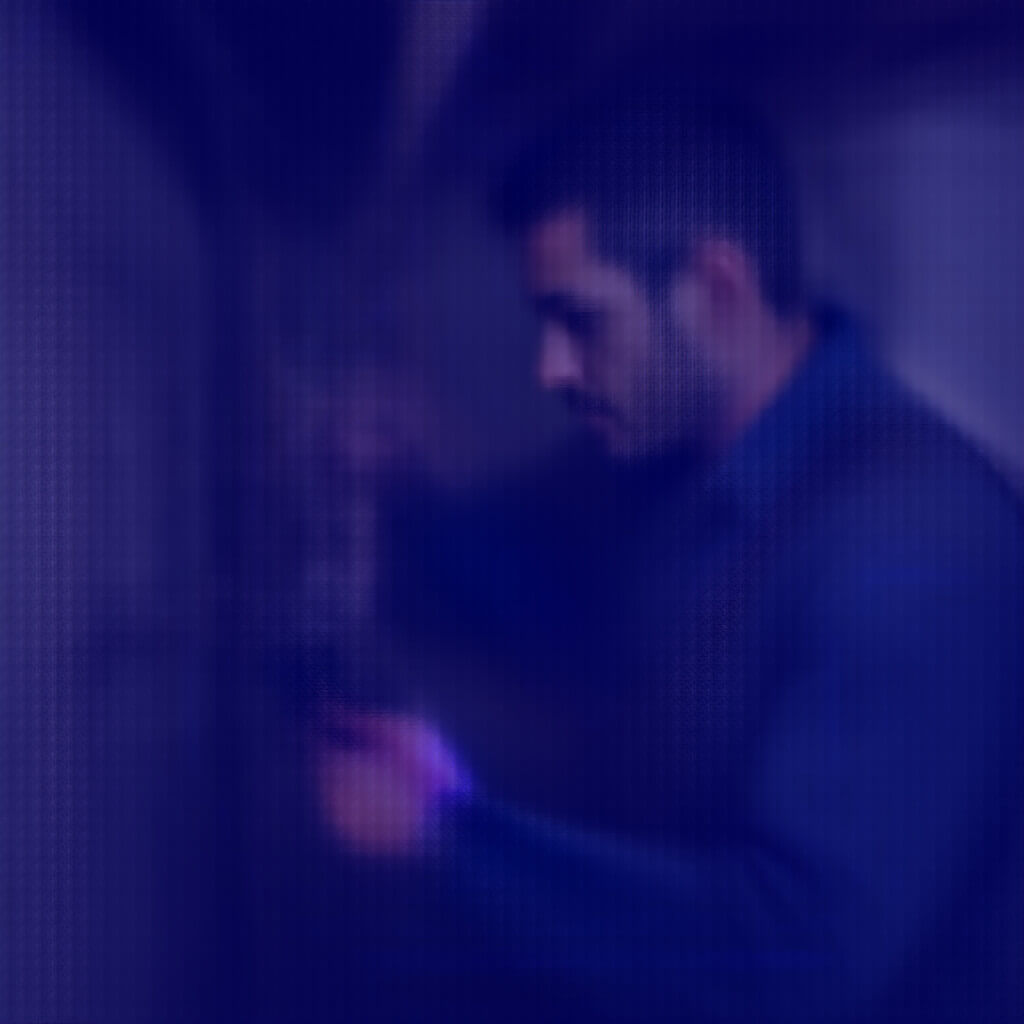} &
\includegraphics[width=0.06\textwidth, height=0.06\textwidth]{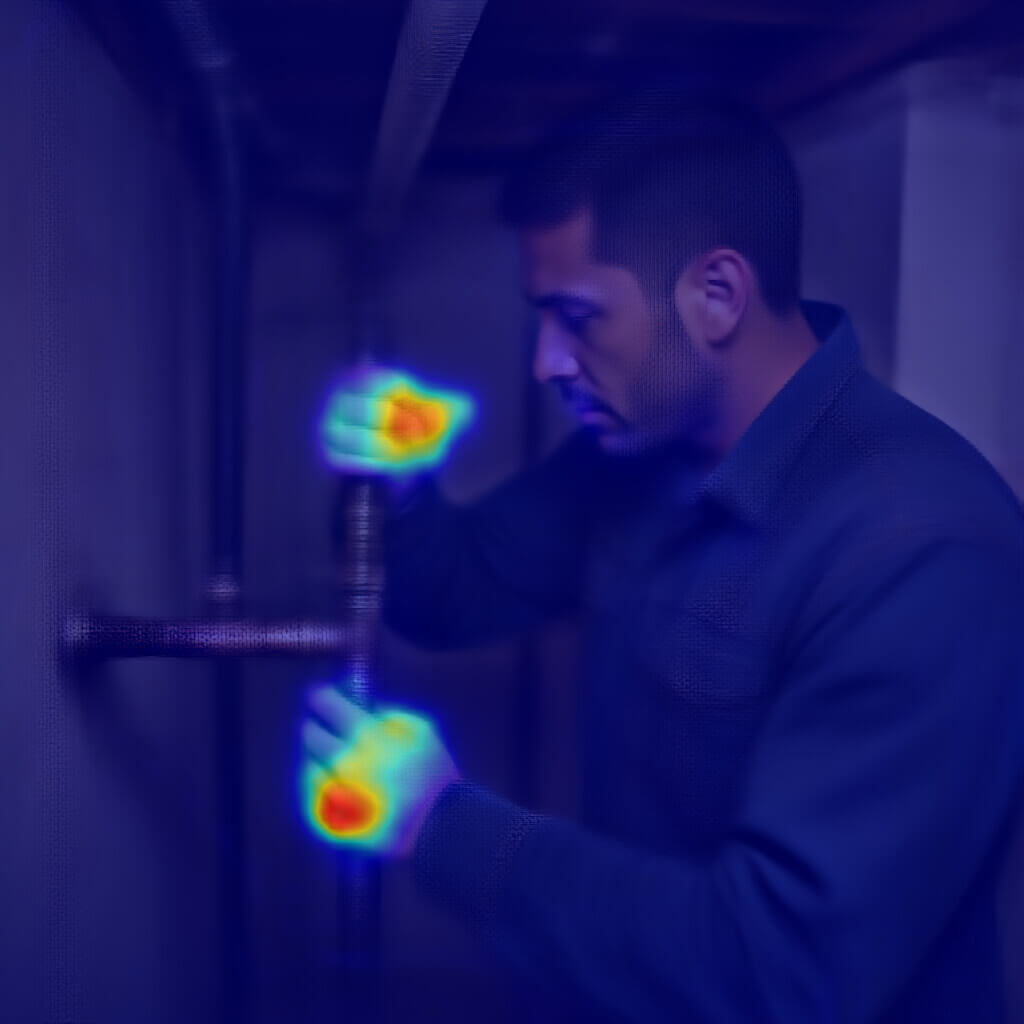} &
\includegraphics[width=0.06\textwidth, height=0.06\textwidth]{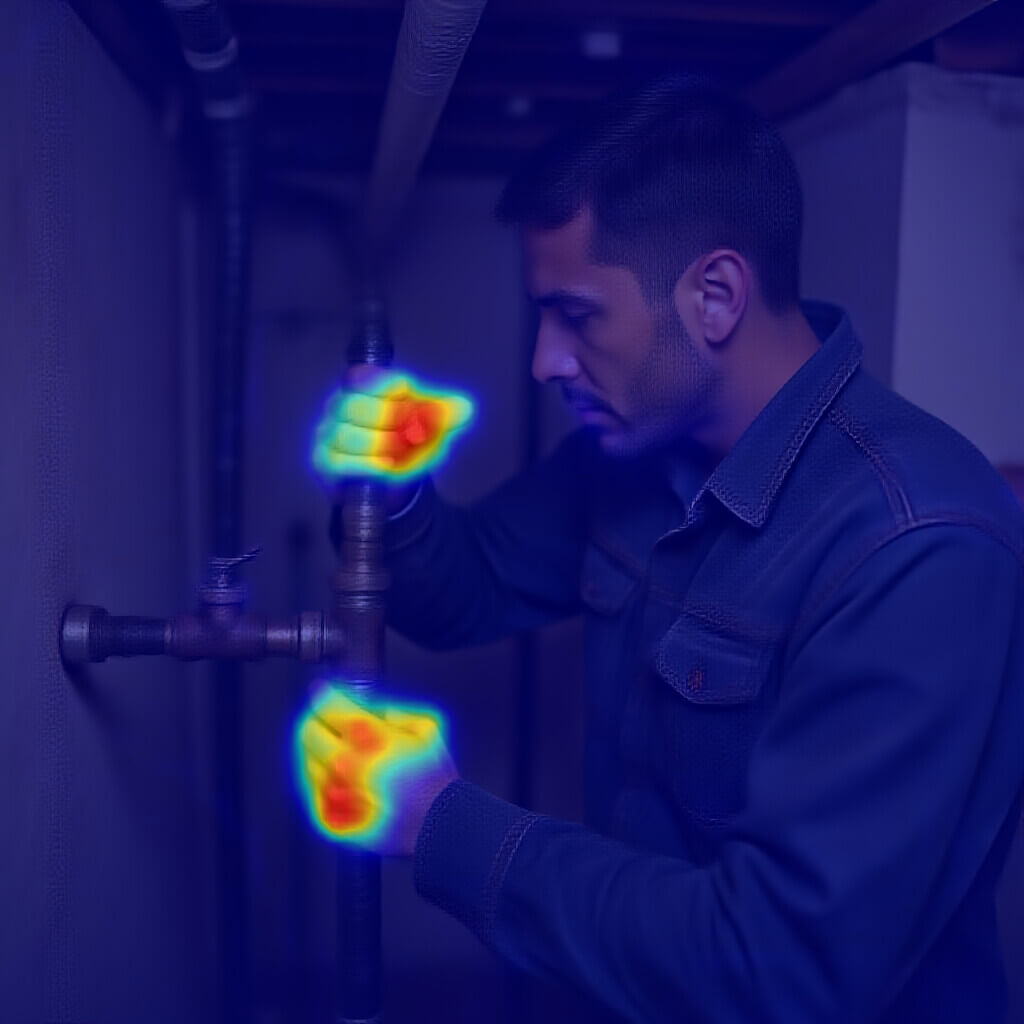}
& \includegraphics[width=0.06\textwidth, height=0.06\textwidth]{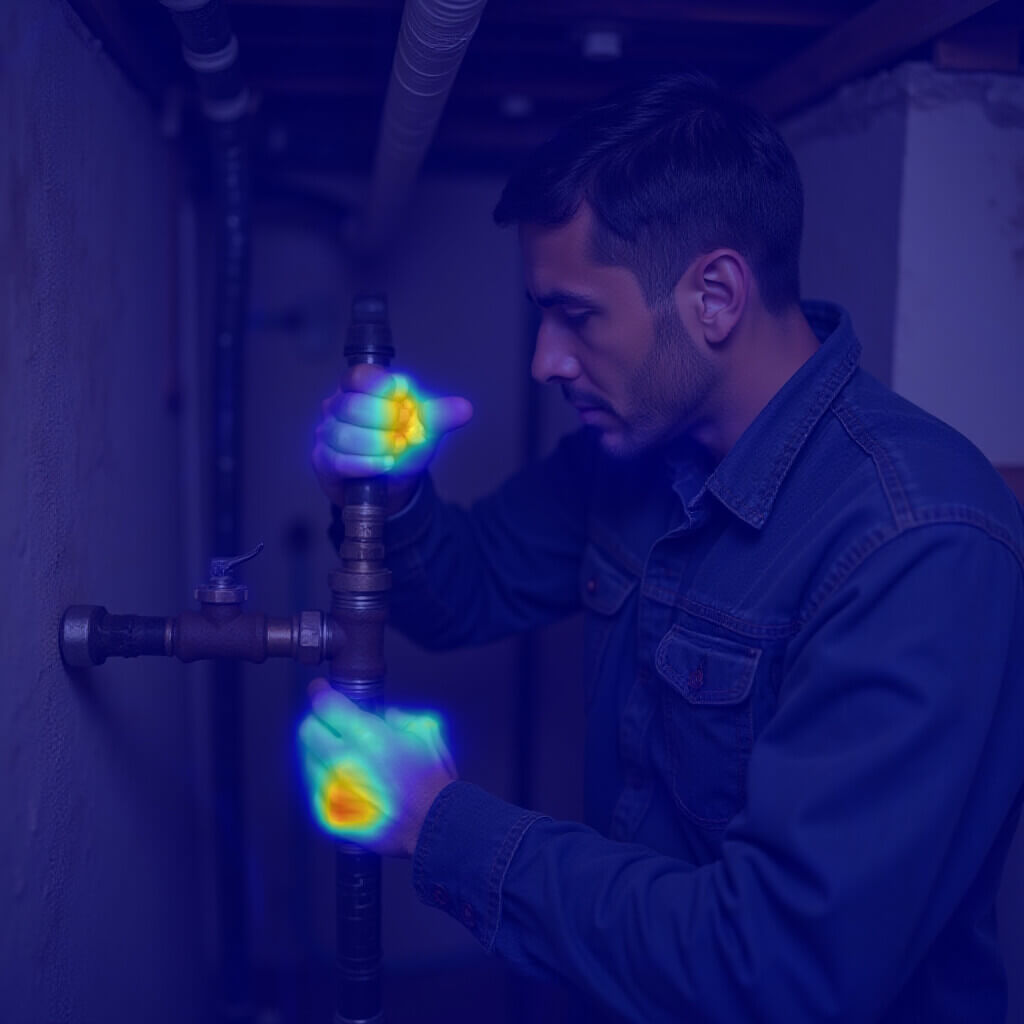} &
\includegraphics[width=0.06\textwidth, height=0.06\textwidth, trim={200, 130, 410, 210}, clip]{images/85.jpg} 
\end{tabular}
\caption{\textbf{Comparison of generation trajectories w/o and w gradient normalization.} Gradient normalization leads to a more stable trajectory and a noticeable mitigation of visual artifacts.
}
\label{fig:nonorm}
\end{figure}

By normalizing $\nabla_{x_t}\mathcal{L}_{a}$, we preserve only the direction, decoupling it from its magnitude. Hence, ensuring that the update strength is controlled solely by $\lambda_t$, and is not influenced by the artifact's size. The importance of this normalization step can be seen in Fig.~\ref{fig:nonorm}.

\begin{figure}[!t]
\centering
\setlength{\tabcolsep}{1.2pt}
\renewcommand{\arraystretch}{0.9}
\begin{tabular}{cccccc}
 & $t_{26}$ & $t_{23}$ & $t_{10}$ & $t_0$ \\
\rotatebox{90}{\footnotesize\hspace{6pt}{\quad SDXL}} & \includegraphics[width=0.11\textwidth, height=0.11\textwidth]{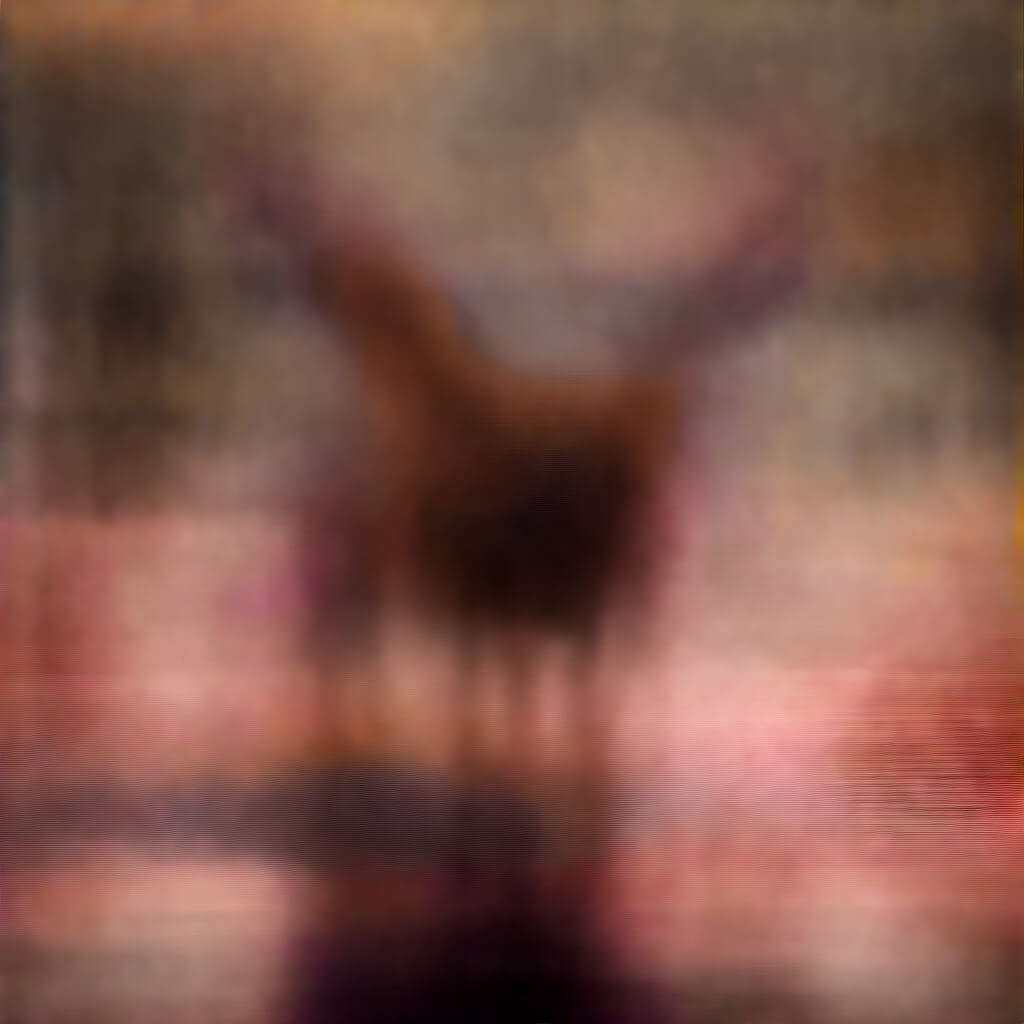} &
\includegraphics[width=0.11\textwidth, height=0.11\textwidth]{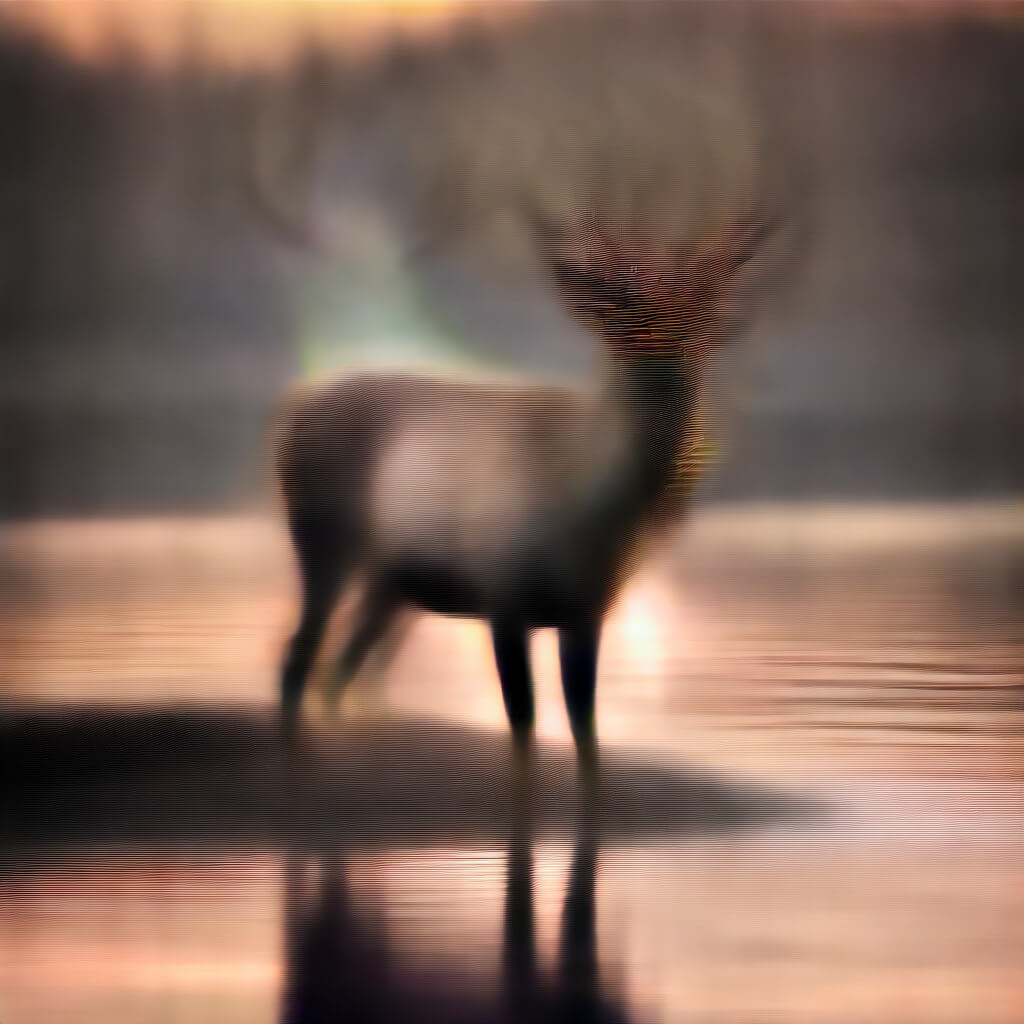} &
\includegraphics[width=0.11\textwidth, height=0.11\textwidth]{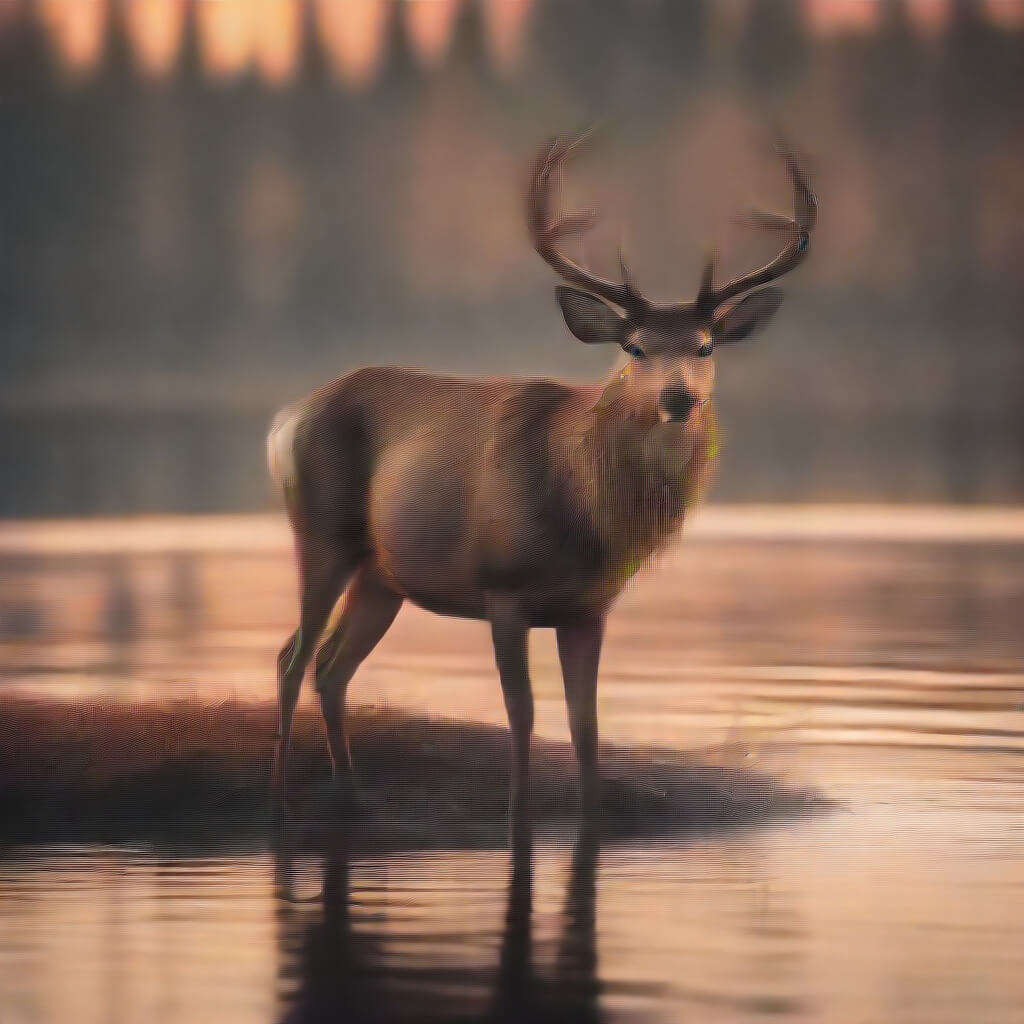} &
\includegraphics[width=0.11\textwidth, height=0.11\textwidth]{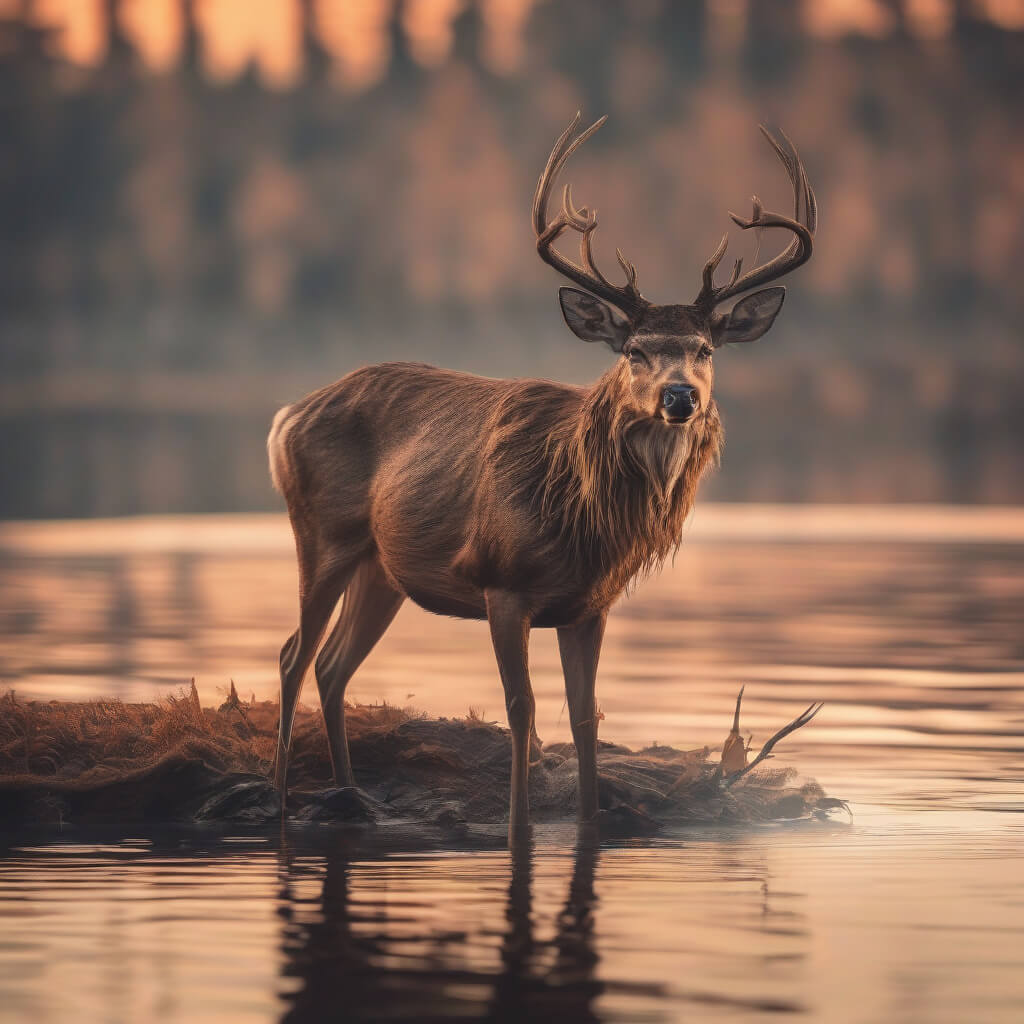} \\

\rotatebox{90}{\footnotesize\hspace{1pt}{FLUX.2 [dev]}} & \includegraphics[width=0.11\textwidth, height=0.11\textwidth]{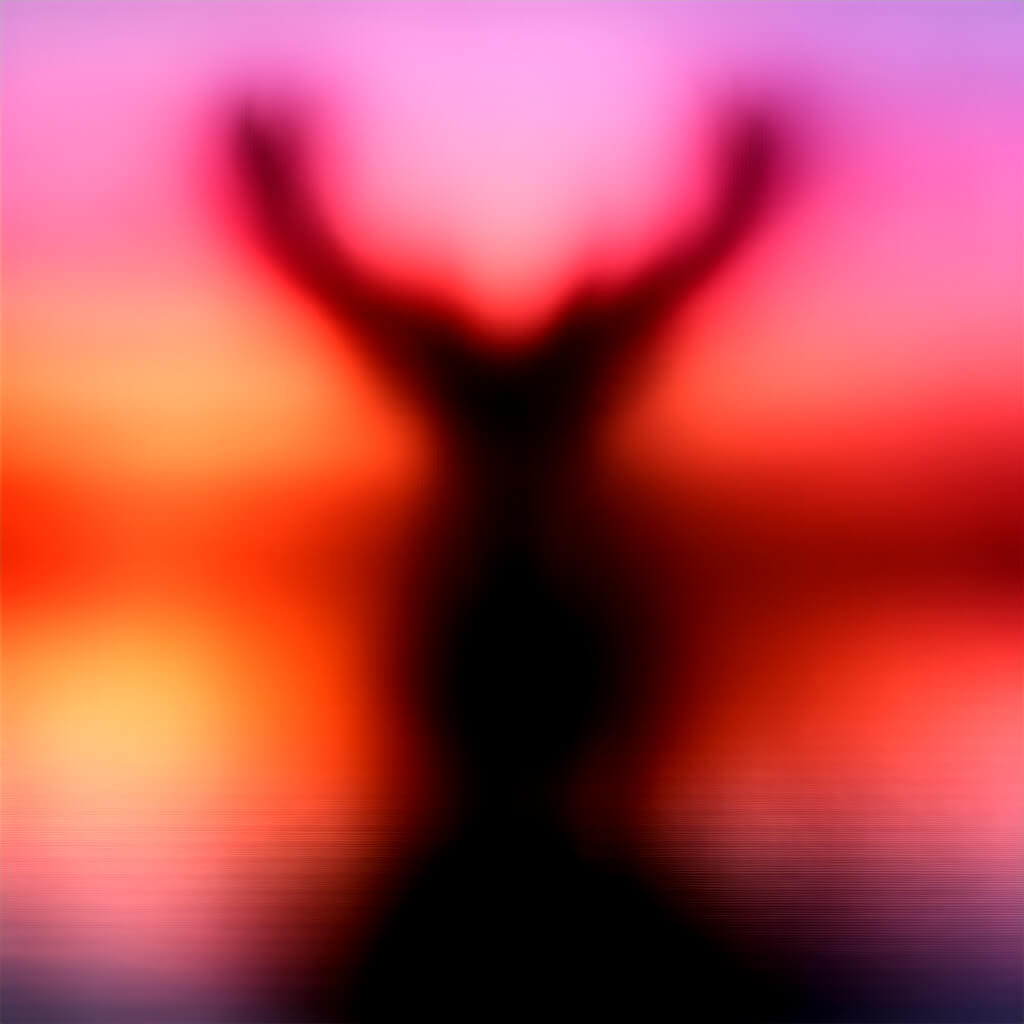} &
\includegraphics[width=0.11\textwidth, height=0.11\textwidth]{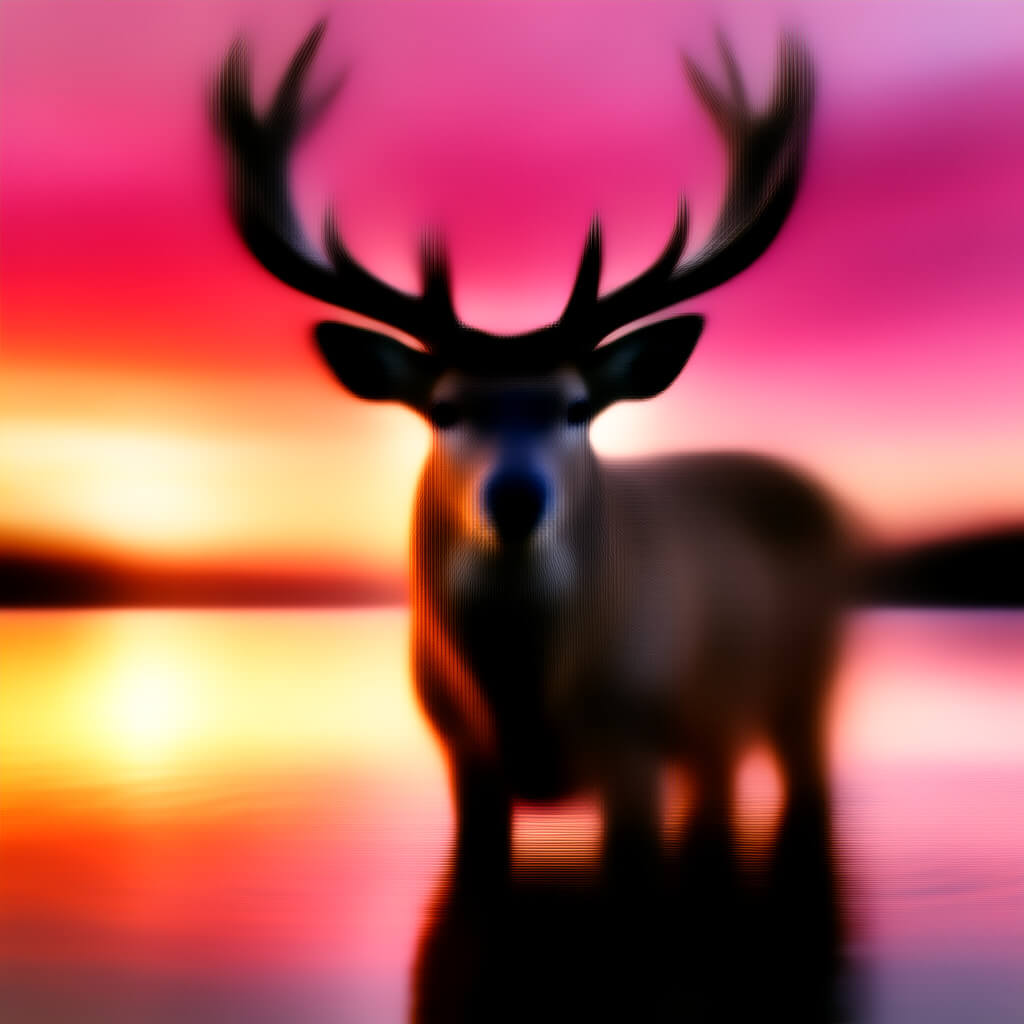} &
\includegraphics[width=0.11\textwidth, height=0.11\textwidth]{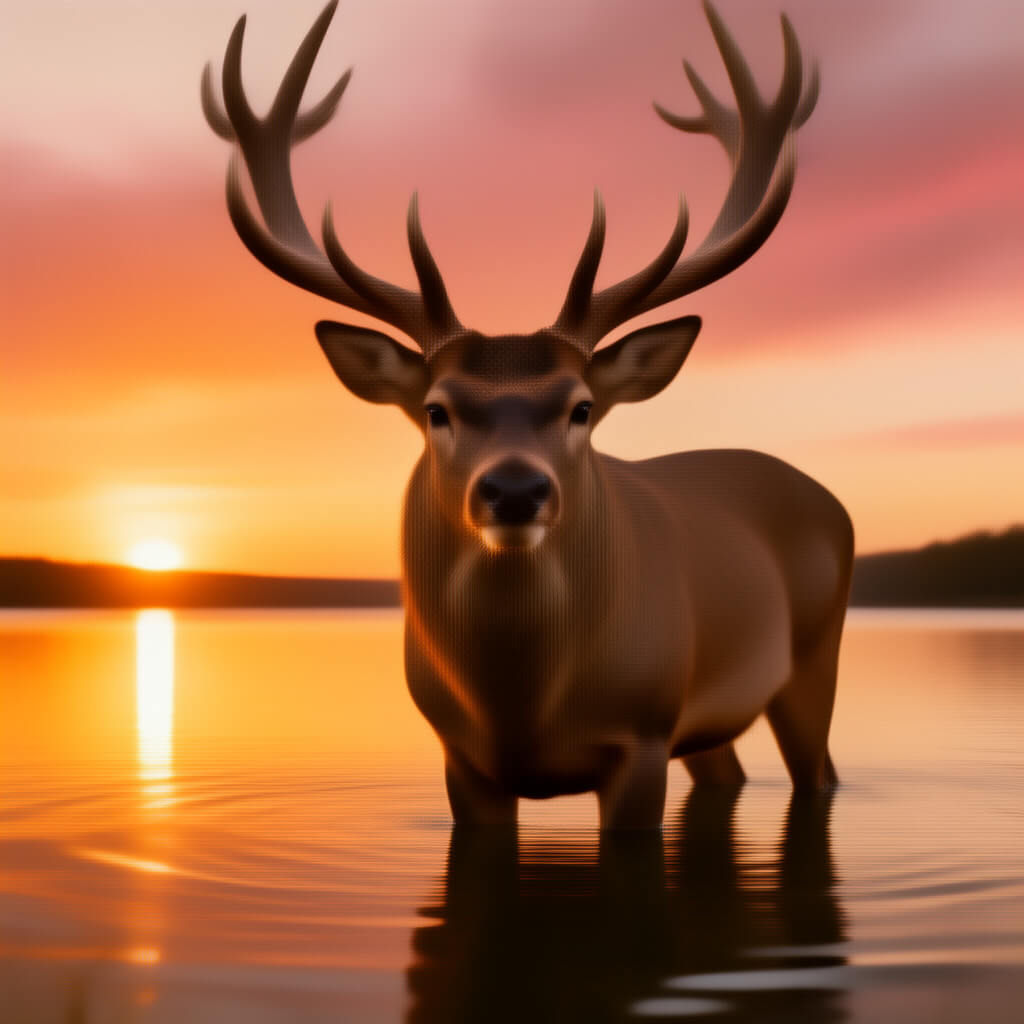} &
\includegraphics[width=0.11\textwidth, height=0.11\textwidth]{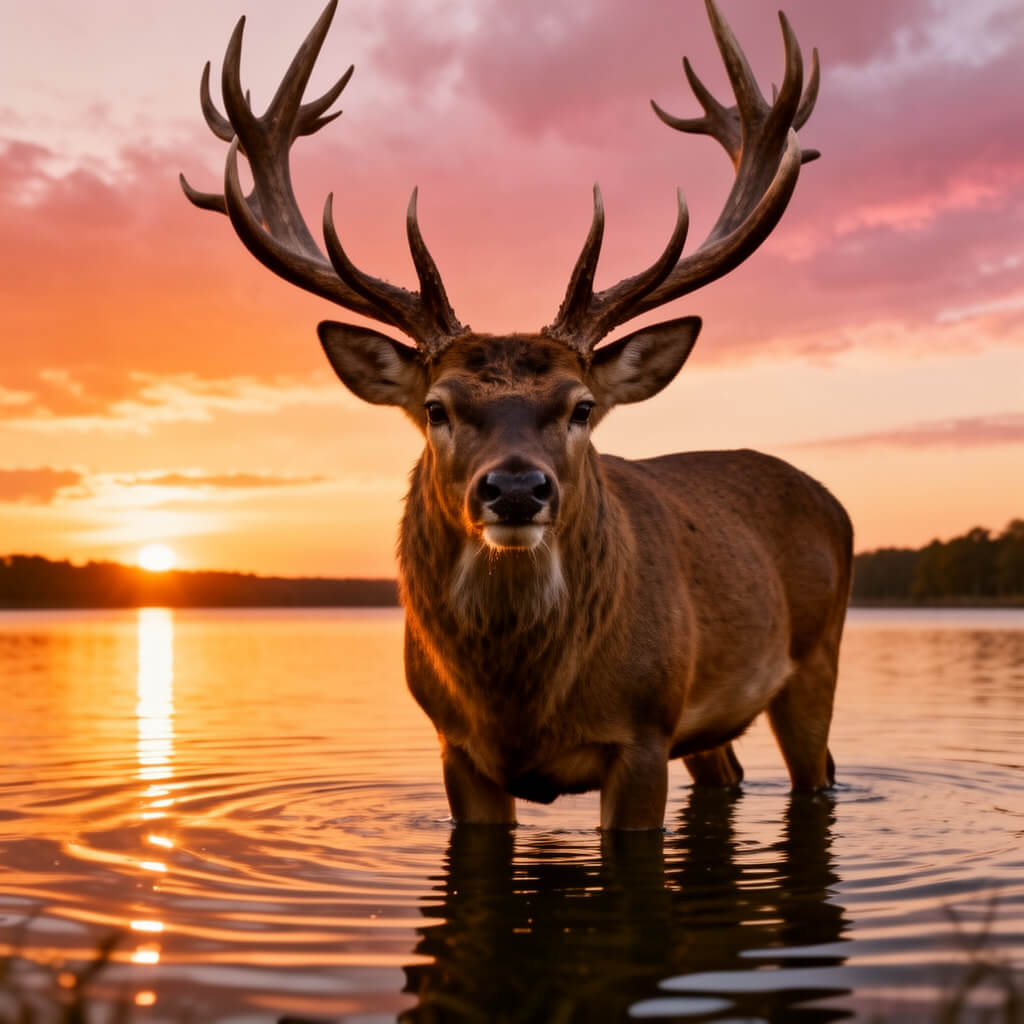} \\
\end{tabular}
\caption{\textbf{Visualization of trajectories in diffusion and rectified flow models.} Even though SDXL and FLUX.2 [dev] exhibit different speed of convergence, they share a similar image formation process starting with general structure and later sharpening the images. A total of 30 iterations are used during inference.}
\label{fig:fluxvsdiff}
\vspace{-4mm}
\end{figure}

\begin{table*}[t]
 \centering
\caption{\textbf{Evaluation of the \our{} on Three Datasets Across Four Models.} Mean $\pm$ standard deviation over different random seeds for each dataset. \our{} achieves the minimal level  of artifacts (the lowest Mean Artifact Freq and Artifact Pixel Ratio) without significantly changing the entire image (MAE) while still maintaining very good quality scores (ImageReward) and prompt-image alignment (CLIP-T).
}
\setlength{\tabcolsep}{4.45pt}
{\fontsize{6.8pt}{10.4pt}\selectfont
\begin{tabular}{lccccccc}
\hline
Model              & CLIP-T $\uparrow$ & Mean Artifact Freq (\%) $\downarrow$ & ImageReward $\uparrow$ & Artfiact Pixel Ratio (\%) $\downarrow$ & MAE $\downarrow$   & MAE (A) $\downarrow$                & MAE (NA) $\downarrow$\\ \hline
\multicolumn{8}{c}{\textit{animals dataset}} \\ \hline
FLUX.2 {[}dev{]}   & 39.532 $\pm$ 0.157  & 100.000 $\pm$ 0.000 & \textbf{1.354 $\pm$ 0.082} & 0.325 $\pm$ 0.048 & - & -  & -   \\
\textbf{+ \our{}}        & \textbf{39.598} $\pm$ 0.282   & \textbf{30.772 $\pm$ 4.772} & 1.343 $\pm$ 0.051 & \textbf{0.074 $\pm$ 0.021}  & \textbf{9.402 $\pm$ 0.459} & \textbf{31.241 $\pm$ 4.250 } & \textbf{9.319 $\pm$ 0.540}   \\ \hline
\multicolumn{8}{c}{\textit{words dataset}} \\ \hline
FLUX.1 {[}dev{]}   & \textbf{36.443 $\pm$ 0.095}  & 100.000 $\pm$ 0.000               & 0.790 $\pm$ 0.035                     & 0.152 $\pm$ 0.011                   & -  & -  &  -        \\
+ DiffDoctor       & 36.418 $\pm$ 0.256  & 46.970 $\pm$ 6.313                & 0.744 $\pm$  0.038                     & 0.065 $\pm$ 0.014                   &  11.469 $\pm$ 0.695  &\textbf{ 35.787 $\pm$ 1.093 }& 11.435 $\pm$ 0.691    \\
+ HPSv2            & 35.689 $\pm$ 0.399  & 56.823 $\pm$ 6.192                & 0.698 $\pm$ 0.081                      & 0.093 $\pm$ 0.017                   & 23.751 $\pm$ 0.289  & 45.050 $\pm$ 4.156  & 	23.721 $\pm$ 0.284   \\
\textbf{+ \our{}} & 36.151 $\pm$ 0.197  & \textbf{9.848 $\pm$ 2.901} & \textbf{0.819 $\pm$  0.065} & \textbf{0.009 $\pm$ 0.002} &\textbf{ 8.744 $\pm$ 0.097}  & 36.064 $\pm$ 2.001 & \textbf{8.707 $\pm$ 0.099}         \\ \hline
\multicolumn{8}{c}{\textit{people dataset}} \\ \hline
FLUX.1 {[}schnell{]}   & 37.377 $\pm$ 0.237 & 100.000 $\pm$ 0.000 & \textbf{1.175 $\pm$ 0.026} & 0.812 $\pm$ 0.070 &-  & - &  -   \\
+ DiffDoctor        & 	37.448 $\pm$ 0.240 & 65.250 $\pm$ 4.500 & 1.158 $\pm$ 0.027 & 0.271 $\pm$ 0.051  & 12.856 $\pm$ 0.119 & 27.014 $\pm$ 1.243 & 12.755 $\pm$ 0.116   \\
+ HPSv2             & \textbf{37.552 $\pm$ 0.204} & 91.750 $\pm$ 0.957 & 1.170 $\pm$ 0.022 & 0.765 $\pm$ 0.066 & 15.605 $\pm$ 0.042 & 31.162 $\pm$ 1.365 & 15.498 $\pm$ 0.041   \\
\textbf{+ \our{}}  & 37.324 $\pm$ 0.124  & \textbf{32.250 $\pm$ 0.957} & 1.162 $\pm$ 0.026 & \textbf{0.103 $\pm$ 0.042} &\textbf{ 5.135 $\pm$ 0.049} & \textbf{21.217 $\pm$ 0.945} & \textbf{5.026 $\pm$ 0.051}   \\ \hline
\multicolumn{8}{c}{\textit{people dataset}} \\ \hline
Stable Diffusion XL & 38.454 $\pm$ 0.071 & 100.000 $\pm$ 0.000 & 1.153 $\pm$ 0.047 & 0.621 $\pm$ 0.058 & - & -  & -   \\
+ HandsXL & 38.061 $\pm$ 0.128  & 79.845 $\pm$ 2.262 & 1.092 $\pm$ 0.043 & 0.618 $\pm$ 0.050 & 26.168 $\pm$ 0.788 & 37.102 $\pm$ 1.160 & 26.119 $\pm$ 0.781   \\
\textbf{+ \our{}}  & \textbf{38.503 $\pm 0.139$}& \textbf{66.328 $\pm$ 3.634}  & \textbf{1.158 $\pm$ 0.045} & \textbf{0.303 $\pm$ 0.050} & \textbf{7.951 $\pm$ 0.241} & \textbf{18.151 $\pm$ 0.409}  & \textbf{7.906 $\pm$ 0.236}   \\ \hline
\end{tabular}
}
\label{tab:comparison_fluxdev_schnell_sdxl}
\vspace{-2mm}
\end{table*}

\textbf{$\mathbf{\lambda_t}$ Schedule} 
The scalar $\lambda_t$ determines the update strength at the current timestep. The choice of schedule is motivated by the generative behavior of diffusion models. It is established that these models generate low-frequency data, corresponding to coarse features (e.g., global structure, color scheme), in the early stages, while high-frequency details and image sharpening occur in the later stages of the generation process~\cite{perception_diffusion}.

While flow matching and diffusion models differ in the formulation, they share a similar structural image formation, starting with coarse features and refining higher-level details at each timestep, see Fig. ~\ref{fig:fluxvsdiff}. 
Since artifacts typically correspond to structural changes, they appear in the early stages. To address this, we employ a \textit{Power} schedule that applies strong correction during the early noise-dominated stages and decay it during later steps. Given a start value $\lambda_{start}$, an end value $\lambda_{end}$, number of steps $N$, current step index $i$, and a power factor $p$, the schedule is defined as:
\[
    \lambda_t = \lambda_{end} + (\lambda_{start} - \lambda_{end}) \left(1 - \frac{i}{N-1}\right)^p
\]

This ensures that the artifact correction actively steers the image generation during image formation, but allows the model to more freely update details. Ablation study of the effect of different $\lambda_t$ hyperparameters is shown in Appendix.

Finally, this correction is integrated into the solver's update step. The final update rule for the transition to $x_{t-\Delta t}$ is defined as:
\[
    x_{t-\Delta t} = x_t - \Delta t \cdot v_\theta(x_t, t) - \delta_t
\]

The first term $\Delta t \cdot v_\theta(x_t, t)$ ensures the trajectory follows the trajectory learned by the Flow model, while the second term $\delta_t$ applies the necessary shift to steer the latent state towards the artifact-free image. 





\textbf{Generalization to Diffusion Models}
While our primary focus is on rectified flow models, \our{} naturally generalizes to diffusion models, as shown in Fig. \ref{fig:handsxl_comparison}. To demonstrate this, we describe a generalized version of \our{} for modern diffusion models, such as Stable Diffusion XL (SDXL)~\cite{ICLR2024_sdxl}. To derive the formulas we will use the formulation of diffusion models from~\cite{Karras2022edm}.

Unlike flow based models, which predict velocity vector, diffusion models predict the noise component. However, the estimation of the clean latent $\hat{x}_{0,t}$ can still be derived from the noise equation. Based on it the clean data latent is estimated as:
 $   \hat{x}_{0,t} = x_t - \sigma_t \epsilon_\theta(x_t, t)
$ 
where $x_t$ is the current latent and $\sigma_t$ is the scheduler's sigma value for the current timestep, and $\epsilon_\theta(x_t, t)$ is the noise prediction, and $t$ is from $t=T$ to $t=0$.

The estimated $\hat{x}_{0,t}$ is decoded to pixel space to compute the artifact loss $\mathcal{L}_a$. The resulting displacement vector $\delta_t$ is calculated by normalizing the gradient $\nabla_{x_t}\mathcal{L}_{a}$.

Similarly to flow matching models, we integrate this correction directly into the standard Euler discretization step used by the diffusion solver. Let $\sigma_t$ and $\sigma_{t-1}$ represent the noise levels at the current and next timestep, respectively. The update rule combines the denoising step with our artifact correction term:
\[
    x_{t-1} = x_t - (\sigma_t - \sigma_{t-1})\, \epsilon_\theta(x_t, t) - \delta_t
\]


This formula highly resembles the one used for flow matching models. The only difference is the first part, representing the standard trajectory update moving from noise level $\sigma_t$ to $\sigma_{t-1}$ defining the diffusion process.

\textbf{Restricted time interval} In practice, both diffusion and flow-based models operate through a sequence of inference steps $N$. For each model, the number of such steps may vary. In flow-based models, the time range from $t=1$, to $t=0$ is discretized into $N$ intervals. For each step $i = N-1, \ldots, 1, 0.
$
Equivalently, time range can be expressed from $t=t_{N-1}$ to $t=t_0$,
where $t_0 = 0$. 
It is also worth noting that for $t = 0$, Eq.~(\ref{eq:rec_img}) implies
$
\hat{x}_{0,0} = x_0.
$

In our model, we show that applying corrections within a restricted time interval from $
t \in [\tau_{\text{end}},\; \tau_{\text{start}} \,]
$ leads to improved result quality  \cite{ab867a56d3224cad9b2963f04a47095e}.  When it comes down to implementation for both diffusion and flow-based models $ N-1-i_{\tau_{\text{start}}}$ is the initial iteration of the correction change, and $i_{\tau_{\text{end}}}$ is the final iteration of the trajectory correction.


\section{Experiments}
In this section, we present numerical and visual comparison with other methods and configurations. We discuss the construction of the datasets used, and detail the metrics used for evaluation. The hyperparameters for each model and \our{} are described comprehensively in the Appendix. 

\begin{figure}[t]
\centering
\setlength{\tabcolsep}{1pt}
\begin{tabular}{cccc}
 & Baseline & +HandsXL & +\our{} \\
\rotatebox{90}{\footnotesize\hspace{8pt}{FLUX.1 [dev]}} &
\includegraphics[width=0.15\textwidth]{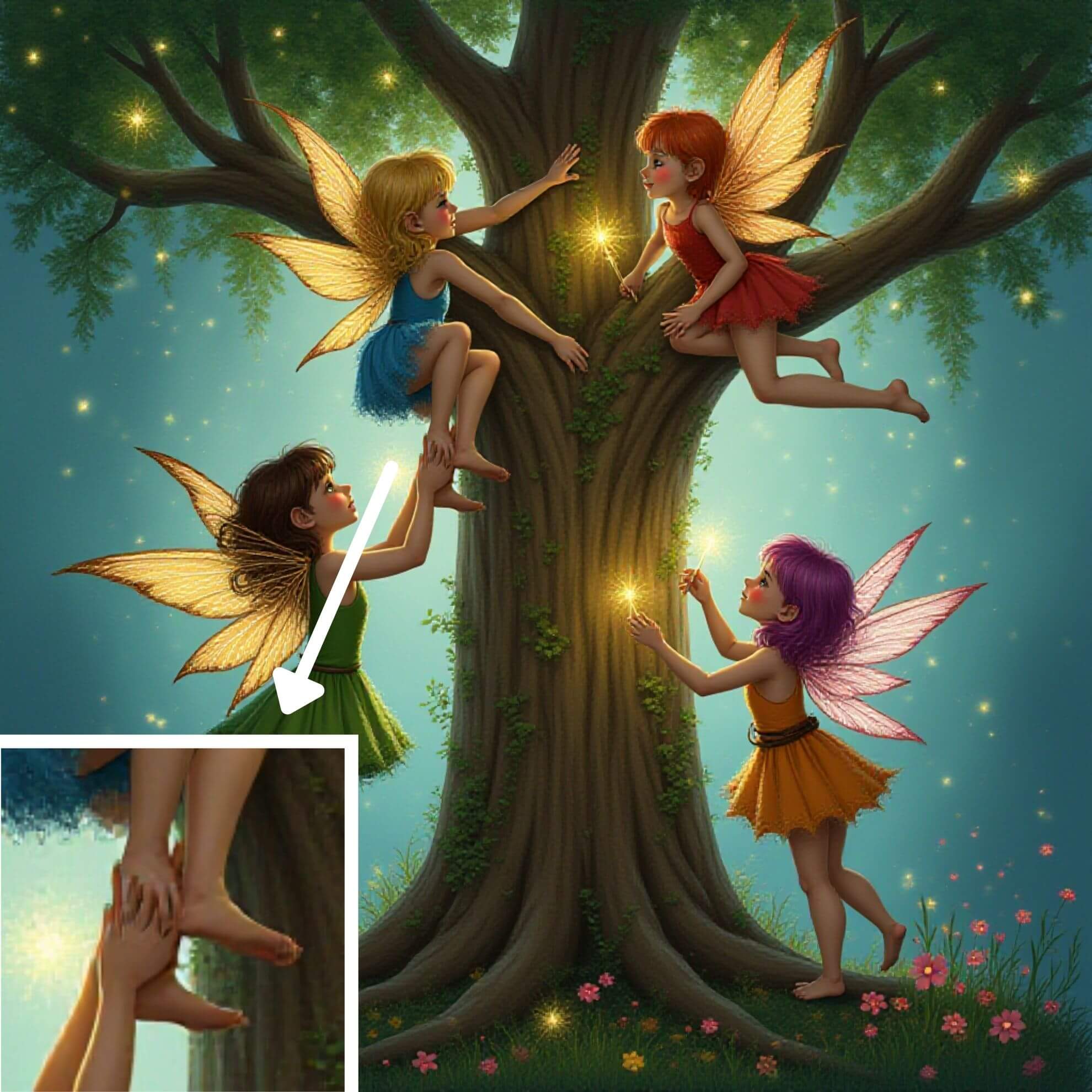} &
\includegraphics[width=0.15\textwidth]{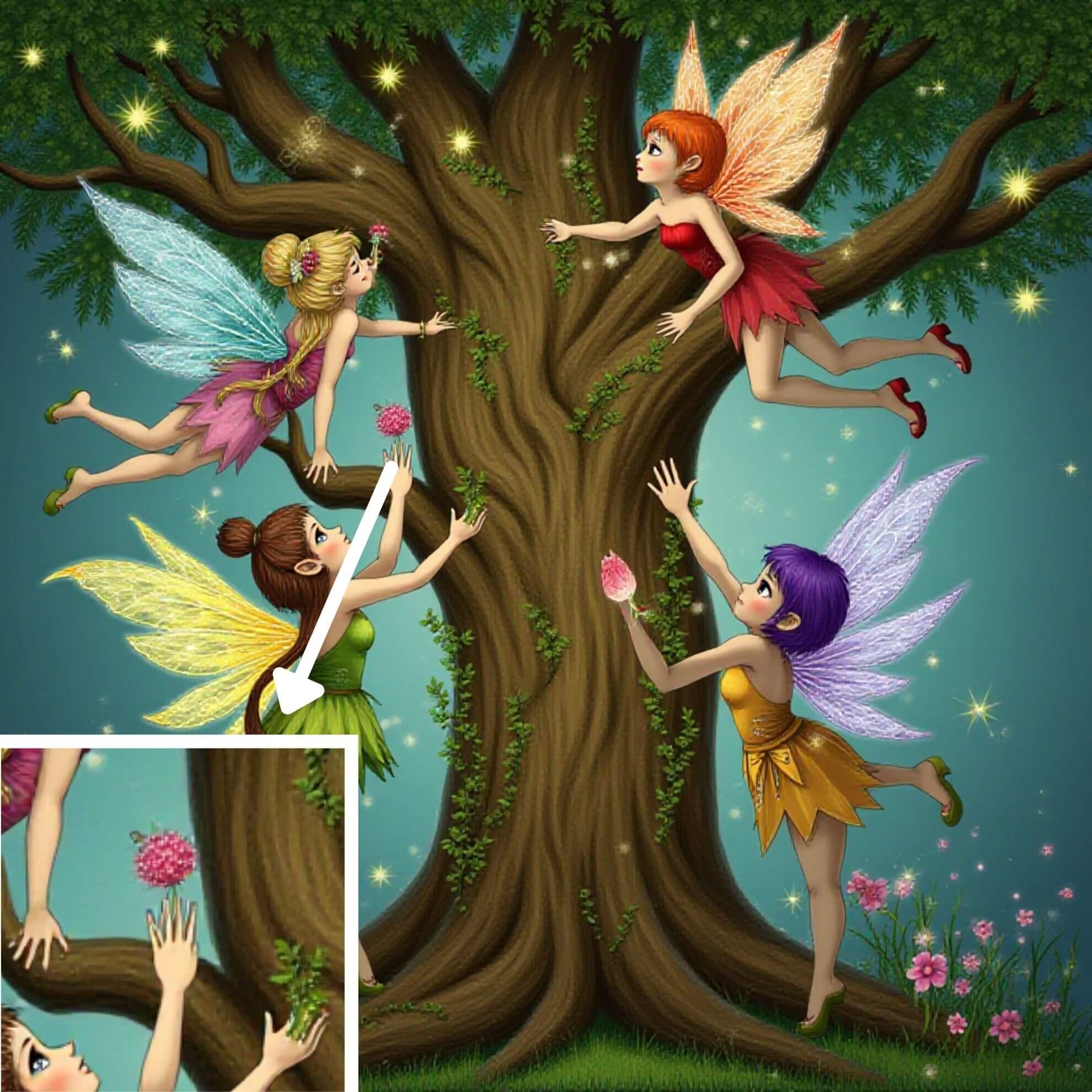} &
\includegraphics[width=0.15\textwidth]{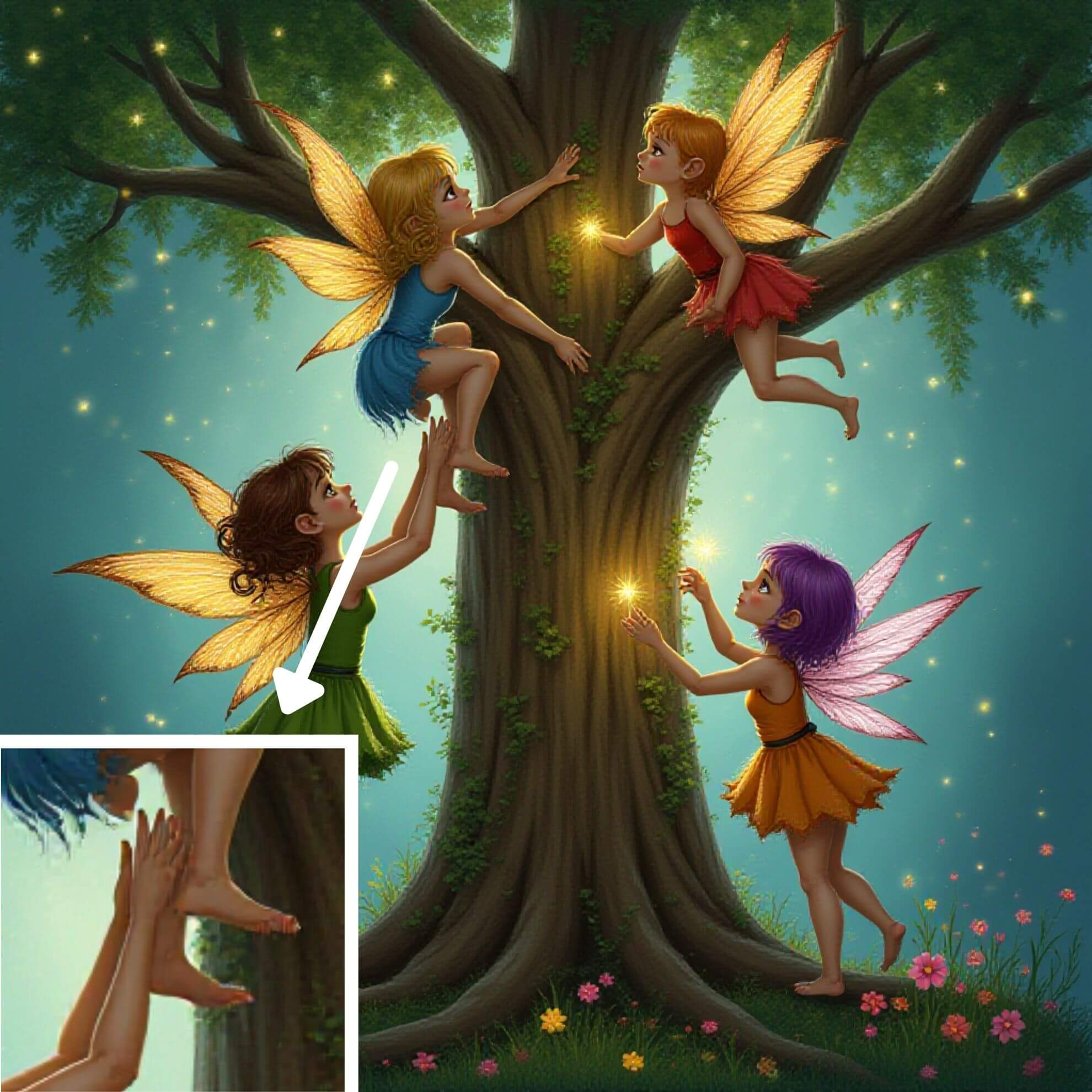} \\

\rotatebox{90}{\footnotesize\hspace{16pt}{\quad SDXL}} &
\includegraphics[width=0.15\textwidth]{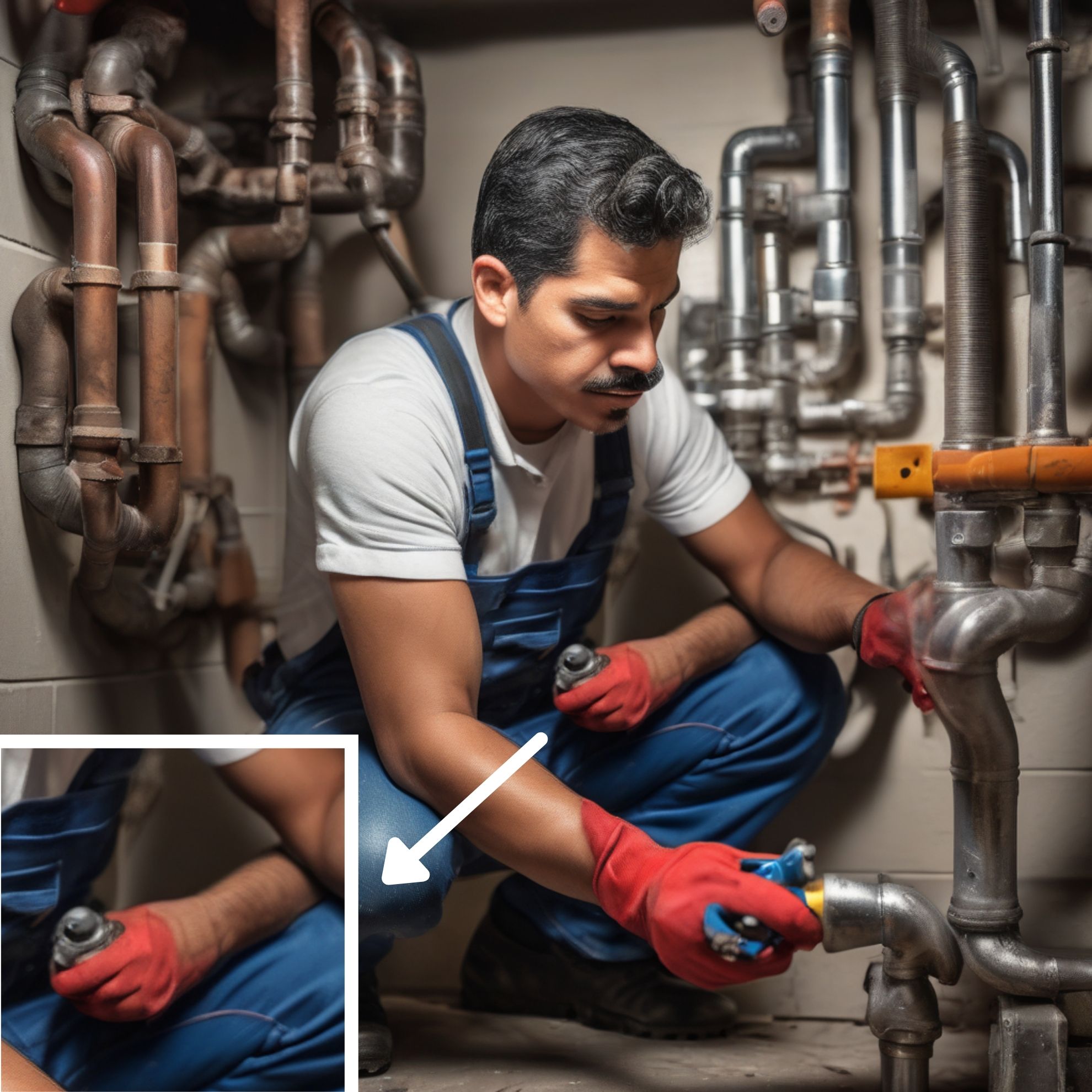} &
\includegraphics[width=0.15\textwidth]{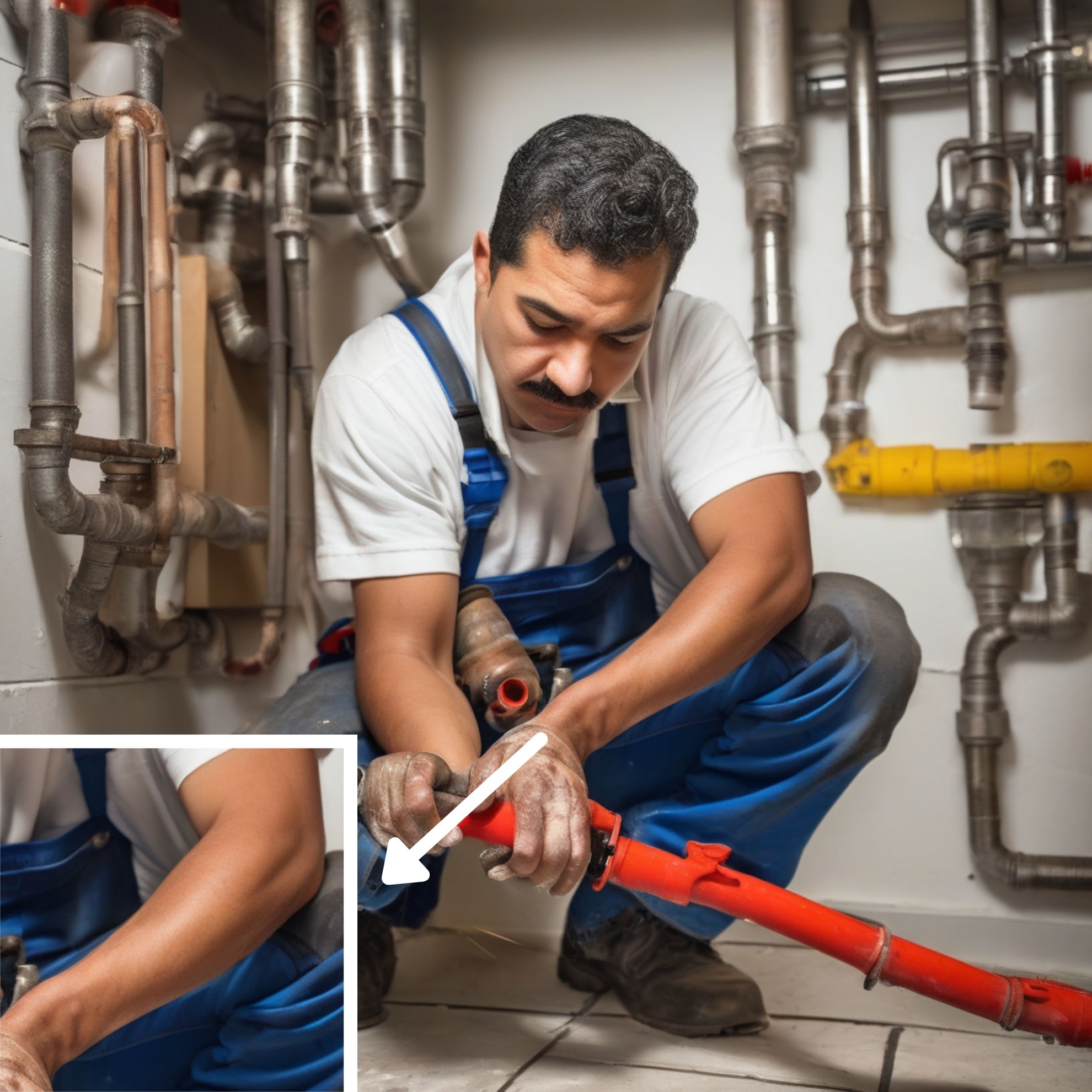} &
\includegraphics[width=0.15\textwidth]{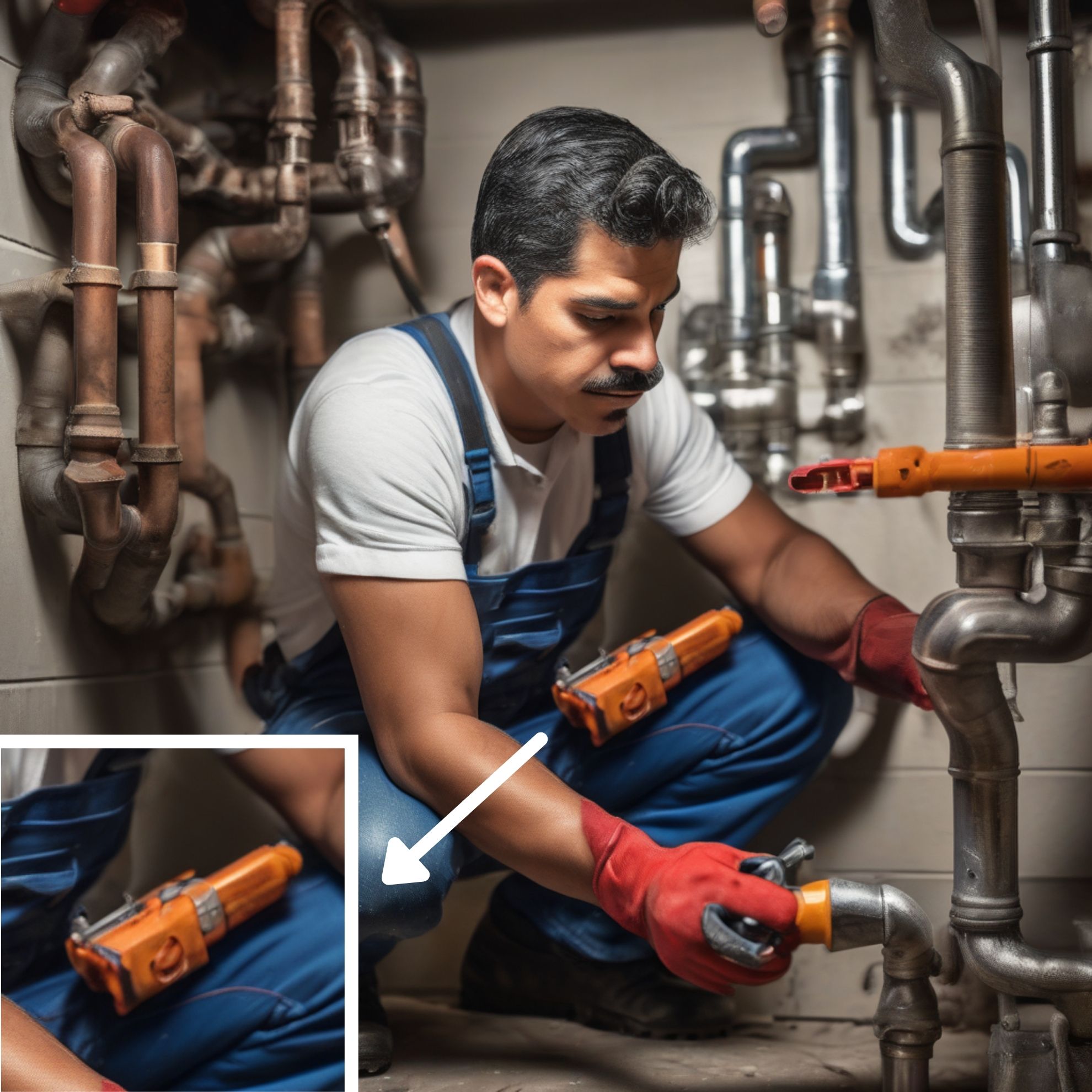} \\
\end{tabular}
\caption{\textbf{Artifact correction comparison with HandsXL.}
Results for FLUX.1 [dev] (top) and SDXL (bottom). \our{} allows to reduce artifacts so much better than HandsXL.  }
\label{fig:handsxl_comparison}
\vspace{-6mm}
\end{figure}

\begin{figure}[t]
\centering
\setlength{\tabcolsep}{1.2pt}
\renewcommand{\arraystretch}{0.9}
\begin{tabular}{lccc}
 & Baseline  & +\our{}  &  +\ourfine{} \\
\rotatebox{90}{\hspace{3pt} FLUX.1 [dev]} & \includegraphics[width=0.14\textwidth, height=0.14\textwidth]{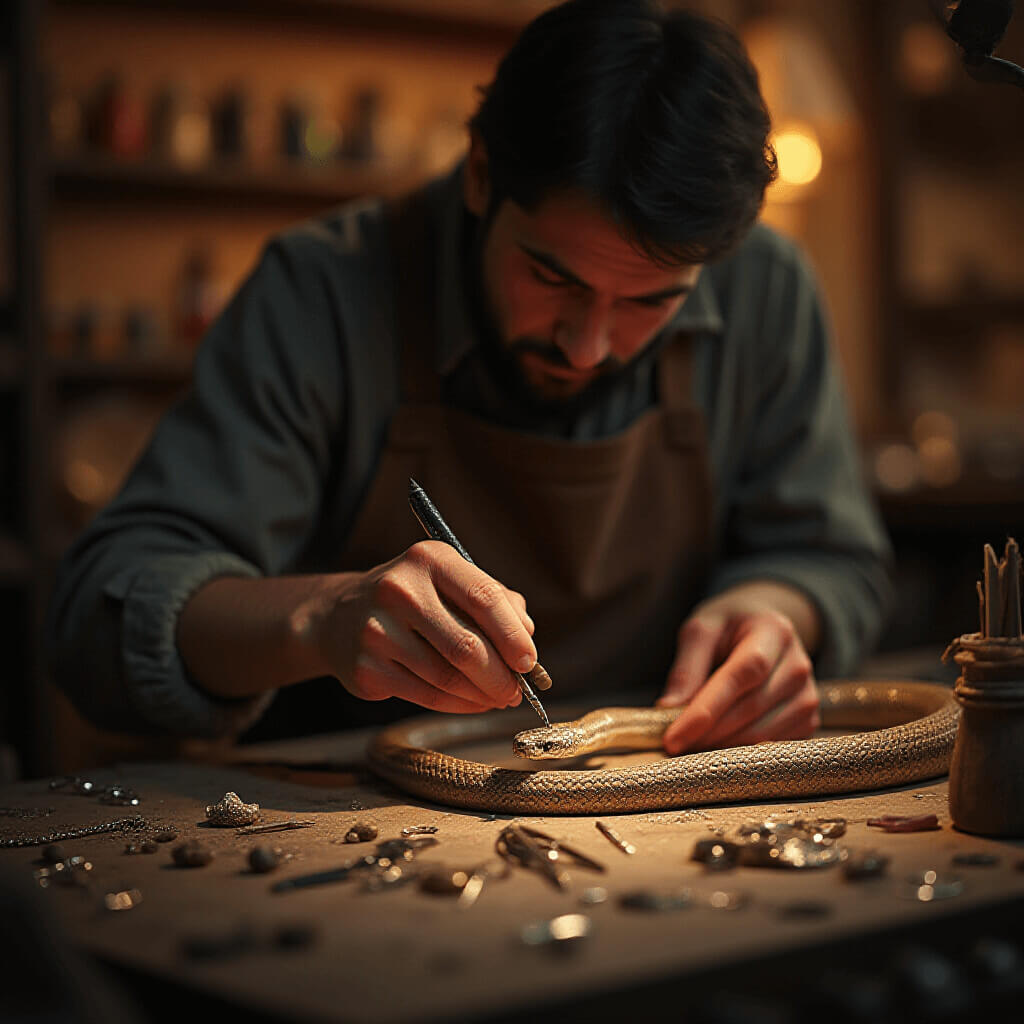} &
\includegraphics[width=0.14\textwidth, height=0.14\textwidth]{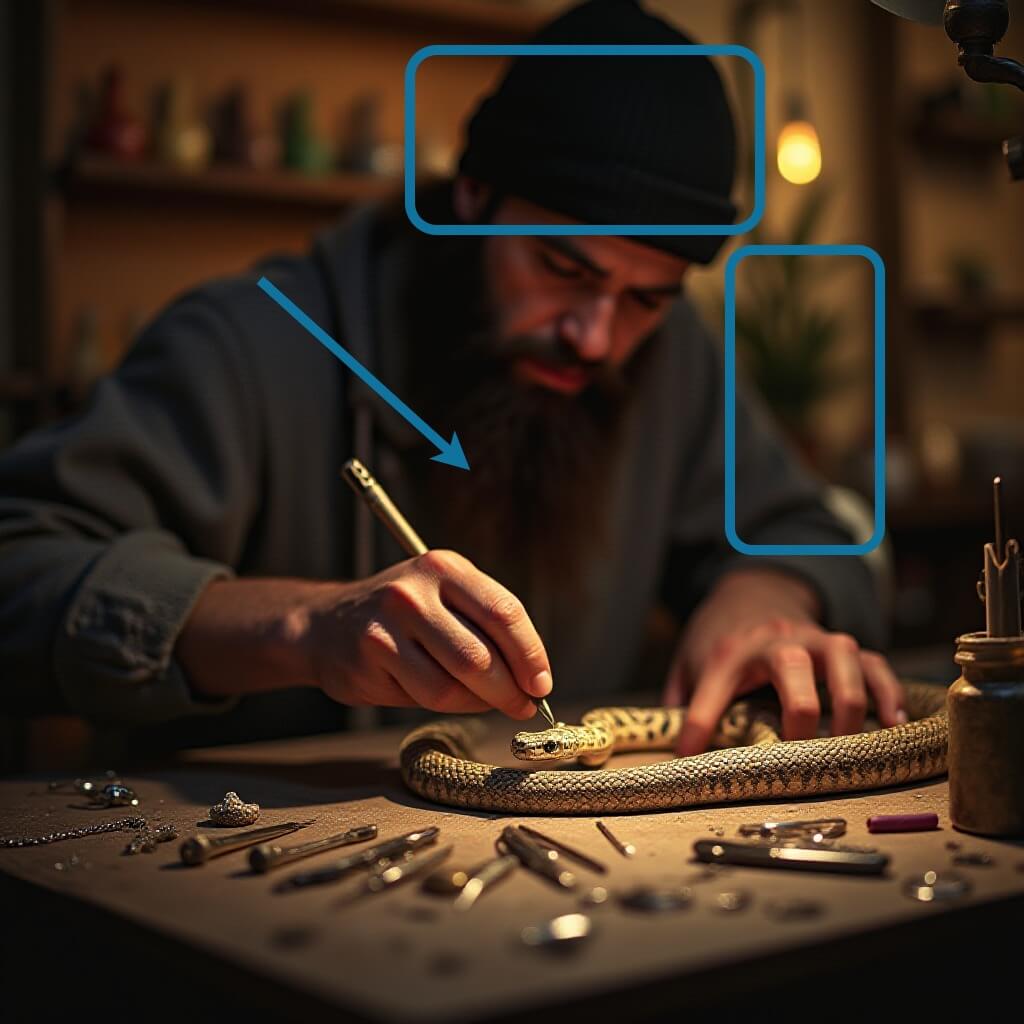} &
\includegraphics[width=0.14\textwidth, height=0.14\textwidth]{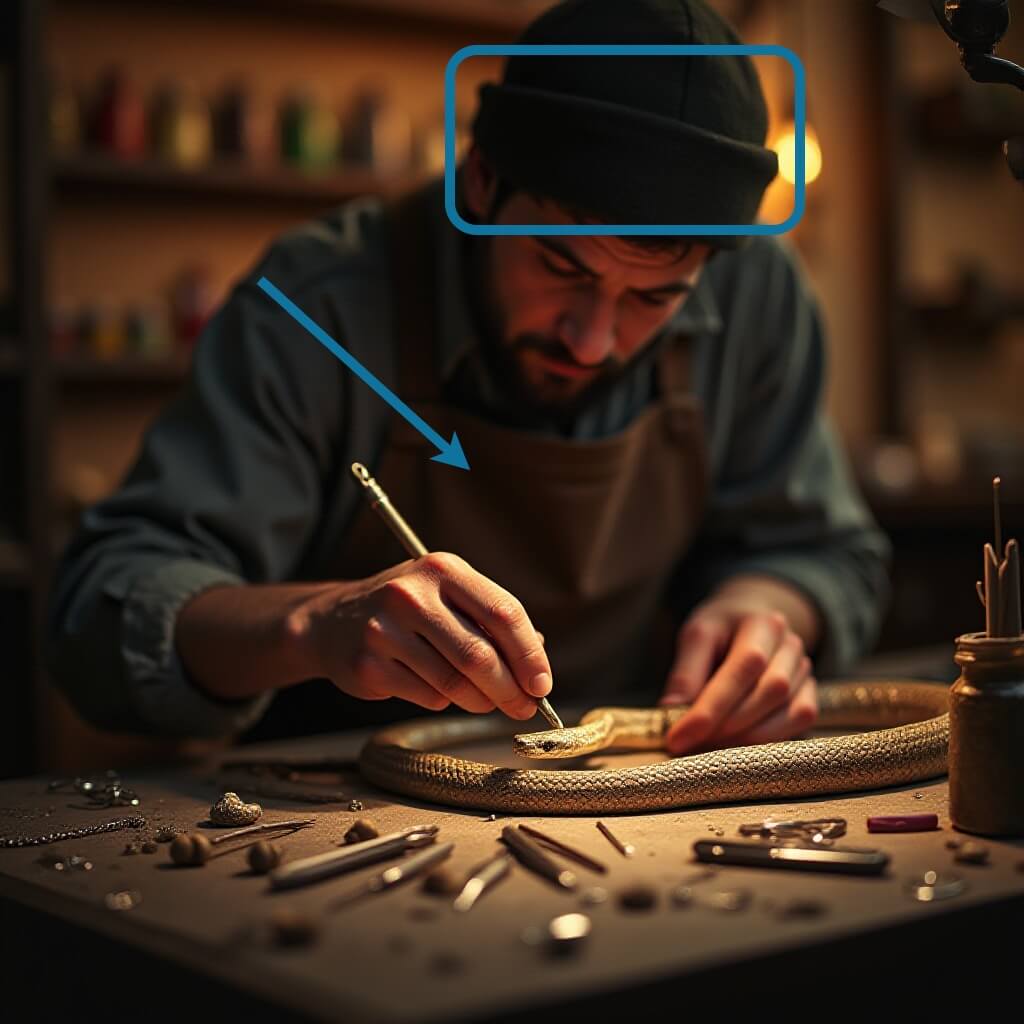} \\
\rotatebox{90}{Artifact detection} & 
\includegraphics[width=0.14\textwidth, height=0.14\textwidth]{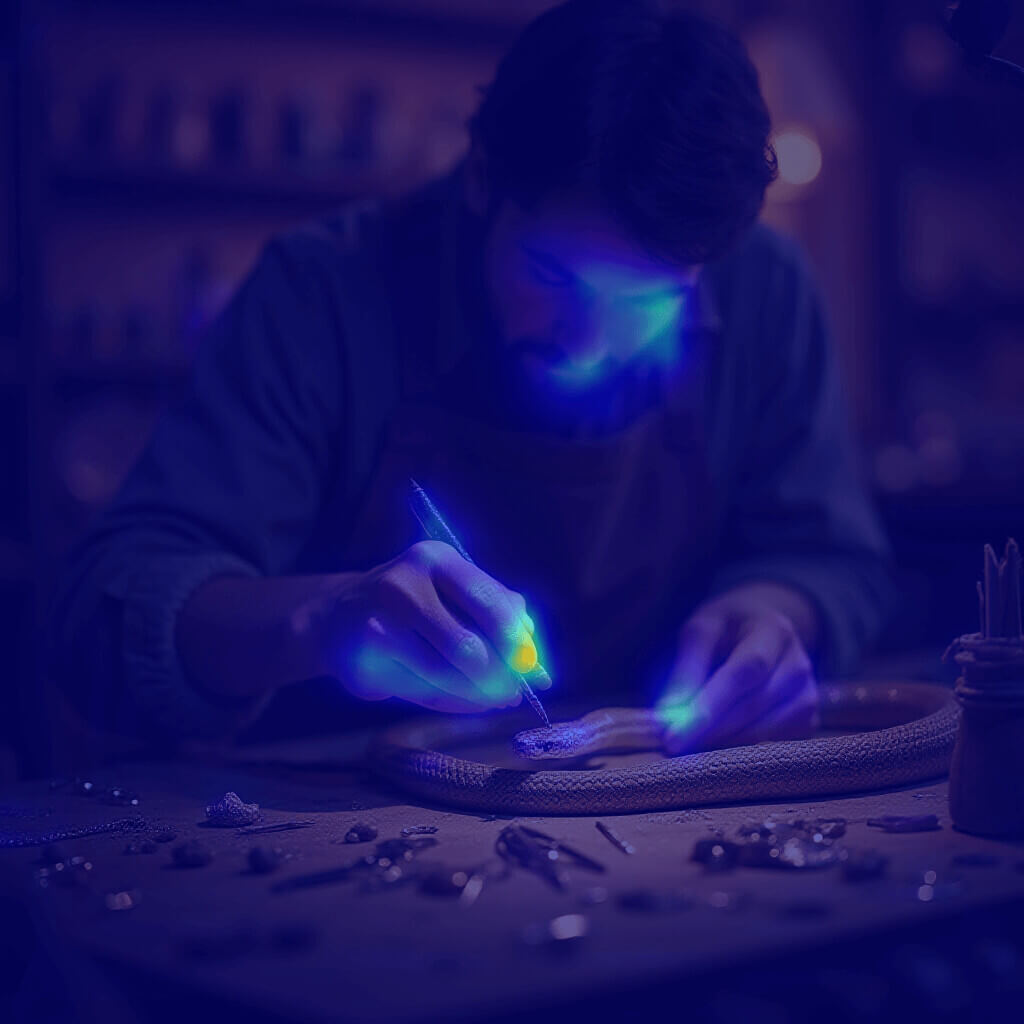} &
\includegraphics[width=0.14\textwidth, height=0.14\textwidth]{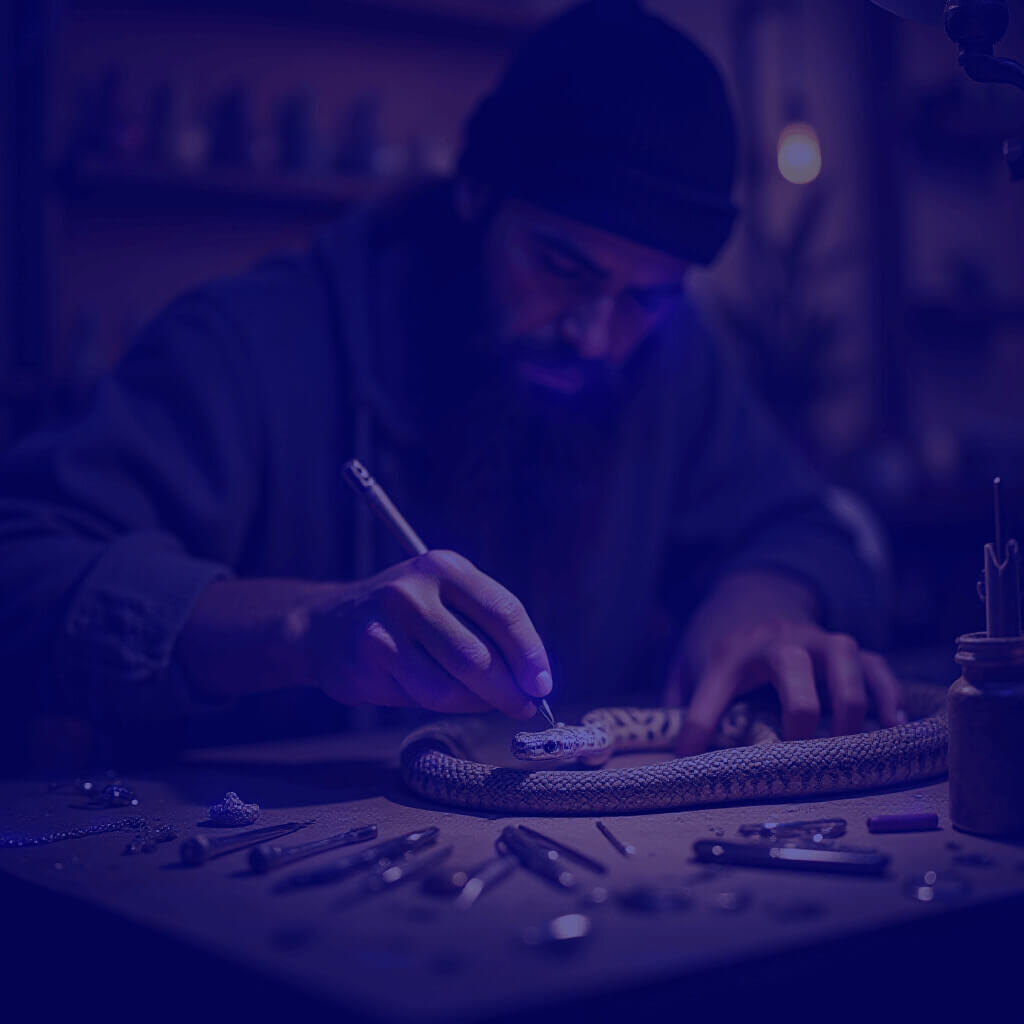} &
\includegraphics[width=0.14\textwidth, height=0.14\textwidth]{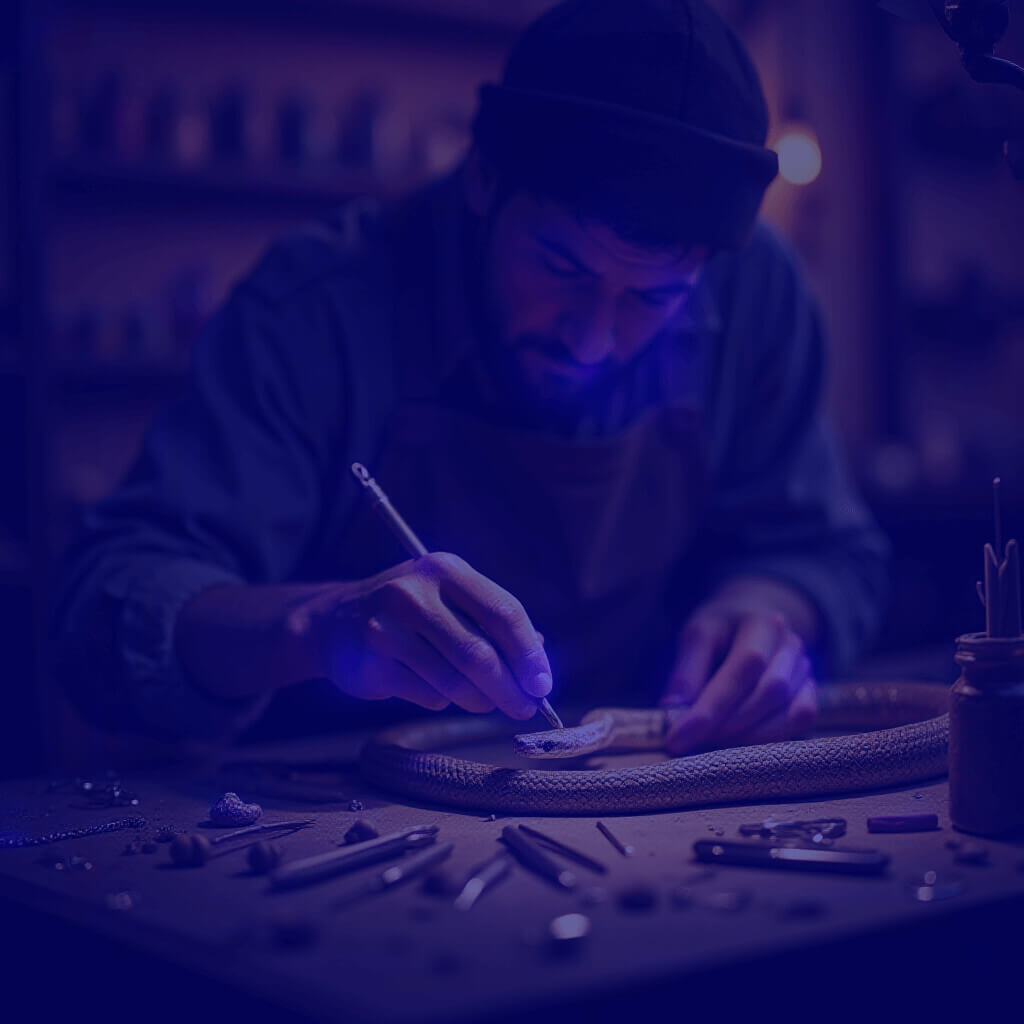} \\
\rotatebox{90}{\hspace{3pt} FLUX.1 [dev]} &\includegraphics[width=0.14\textwidth, height=0.14\textwidth]{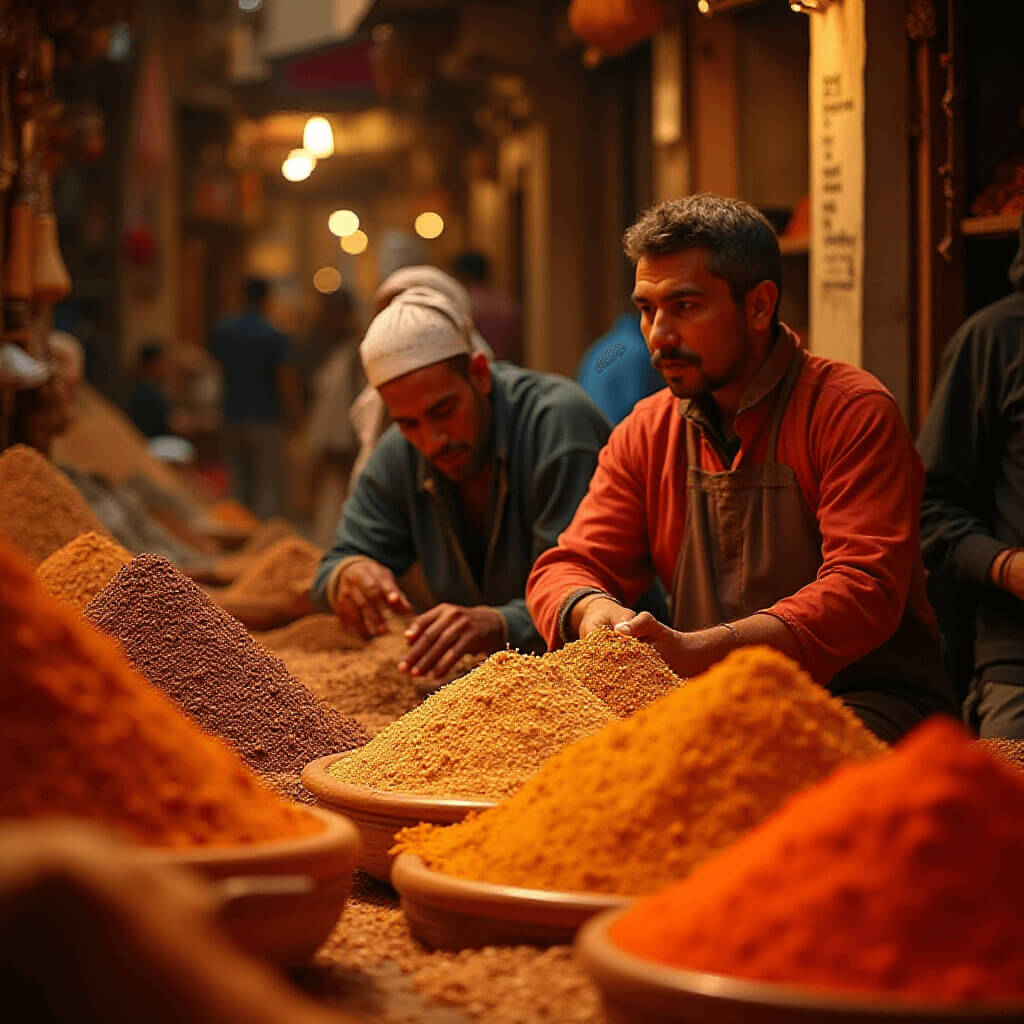} &
\includegraphics[width=0.14\textwidth, height=0.14\textwidth]{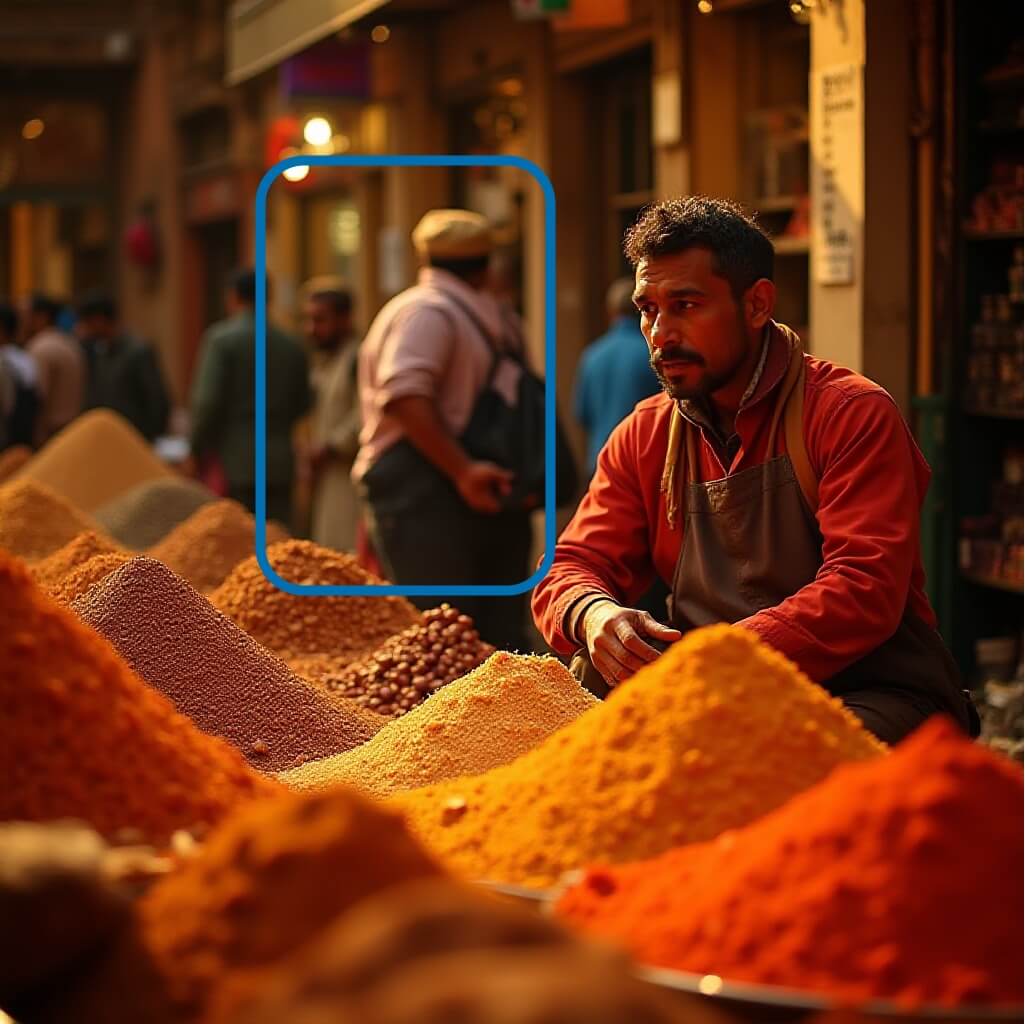} &
\includegraphics[width=0.14\textwidth, height=0.14\textwidth]{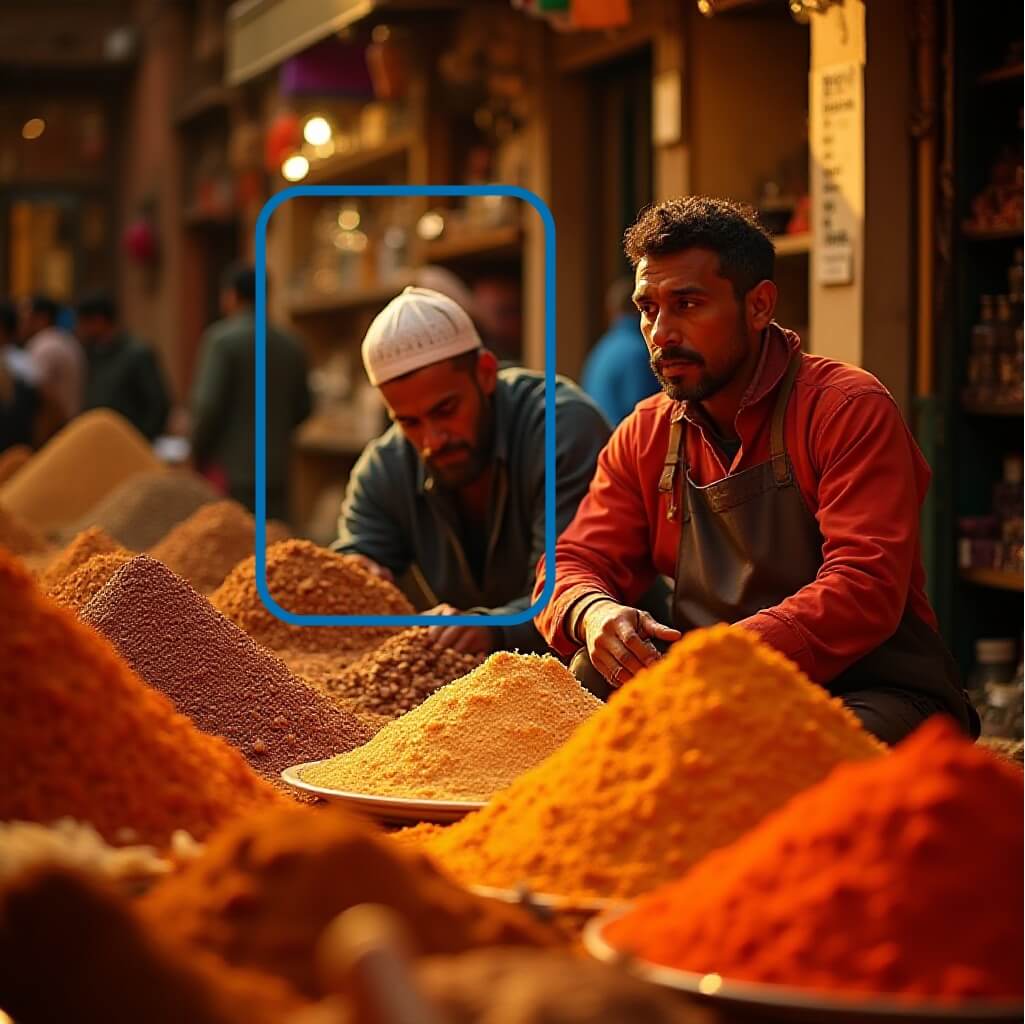} \\
\rotatebox{90}{Artifact detection} & 
\includegraphics[width=0.14\textwidth, height=0.14\textwidth]{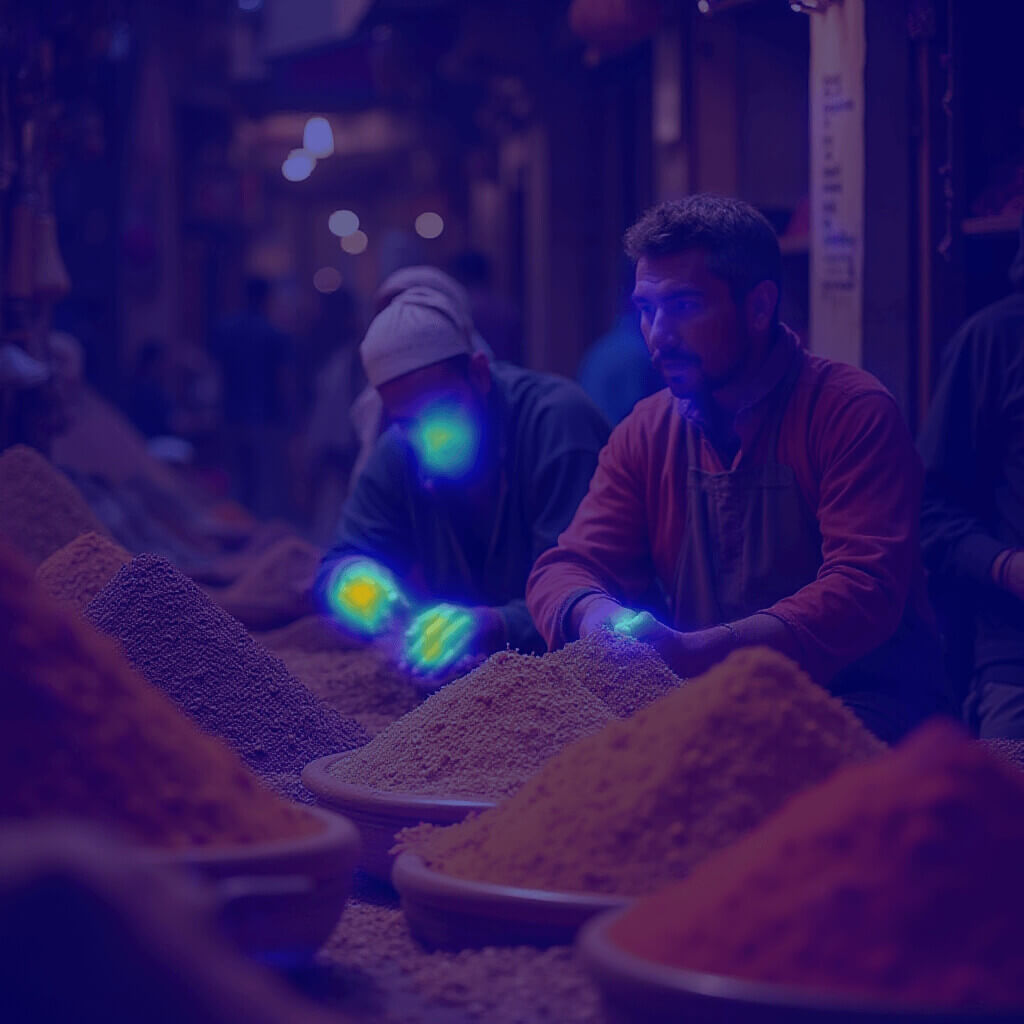} &
\includegraphics[width=0.14\textwidth, height=0.14\textwidth]{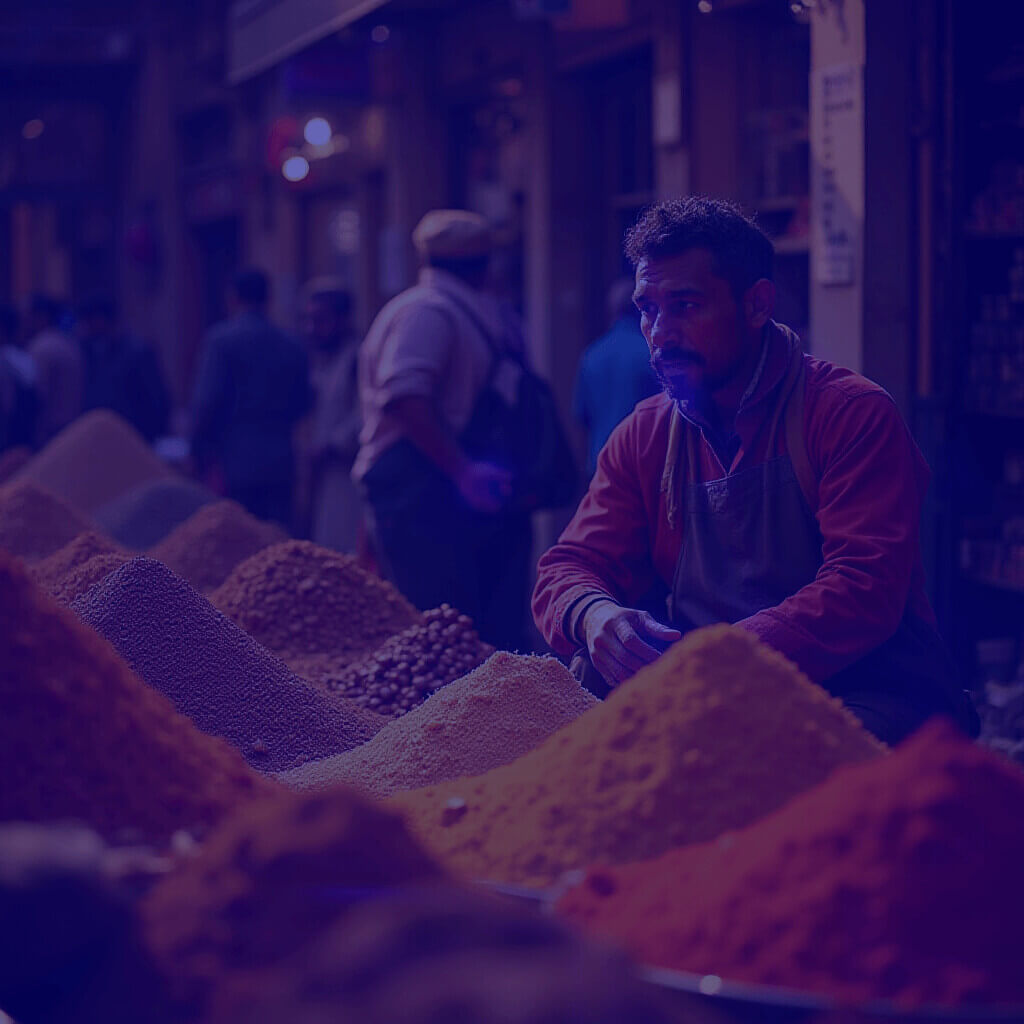}  &
\includegraphics[width=0.14\textwidth, height=0.14\textwidth]{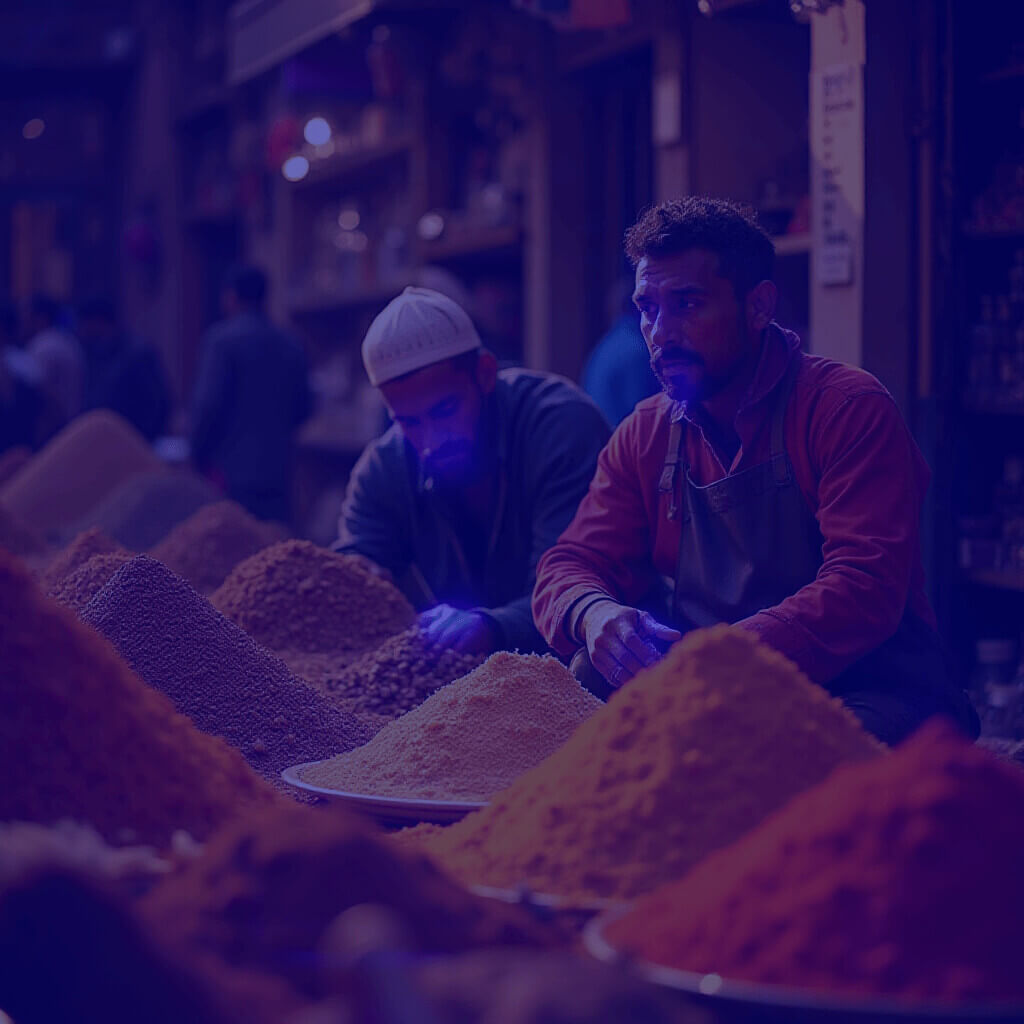}  \\
\end{tabular}
\caption{\textbf{Visualization of final images with $\mathcal{L}_\text{rec}$ normalization (\our{}-Fine) applied during the trajectory for $\alpha=0.1$.} The $\mathcal{L}_\text{rec}$ term enforces similarity between the edited and original images in regions unaffected by artifacts. 
}
\label{fig:artifacts}
\vspace{-1cm}
\end{figure}

\begin{figure*}[!t]
\centering
\setlength{\tabcolsep}{1.5pt}
\renewcommand{\arraystretch}{0.9}
\begin{tabular}{c c c c c c c c c c c}
\footnotesize{$I_{base}$} & \footnotesize{$\mathcal{AD}^D(I_{base})$} & \footnotesize{$\mathcal{AD}^R(I_{base})$} & \footnotesize{$I_D$} & \footnotesize{$\mathcal{AD}^D(I_{D})$} & \footnotesize{$I_R$} & \footnotesize{$\mathcal{AD}^R(I_{D})$} & \footnotesize{$I_{D+R}$} & \footnotesize{$\mathcal{AD}^{D+R}(I_{D+R})$} \\
\includegraphics[width=0.104\textwidth]{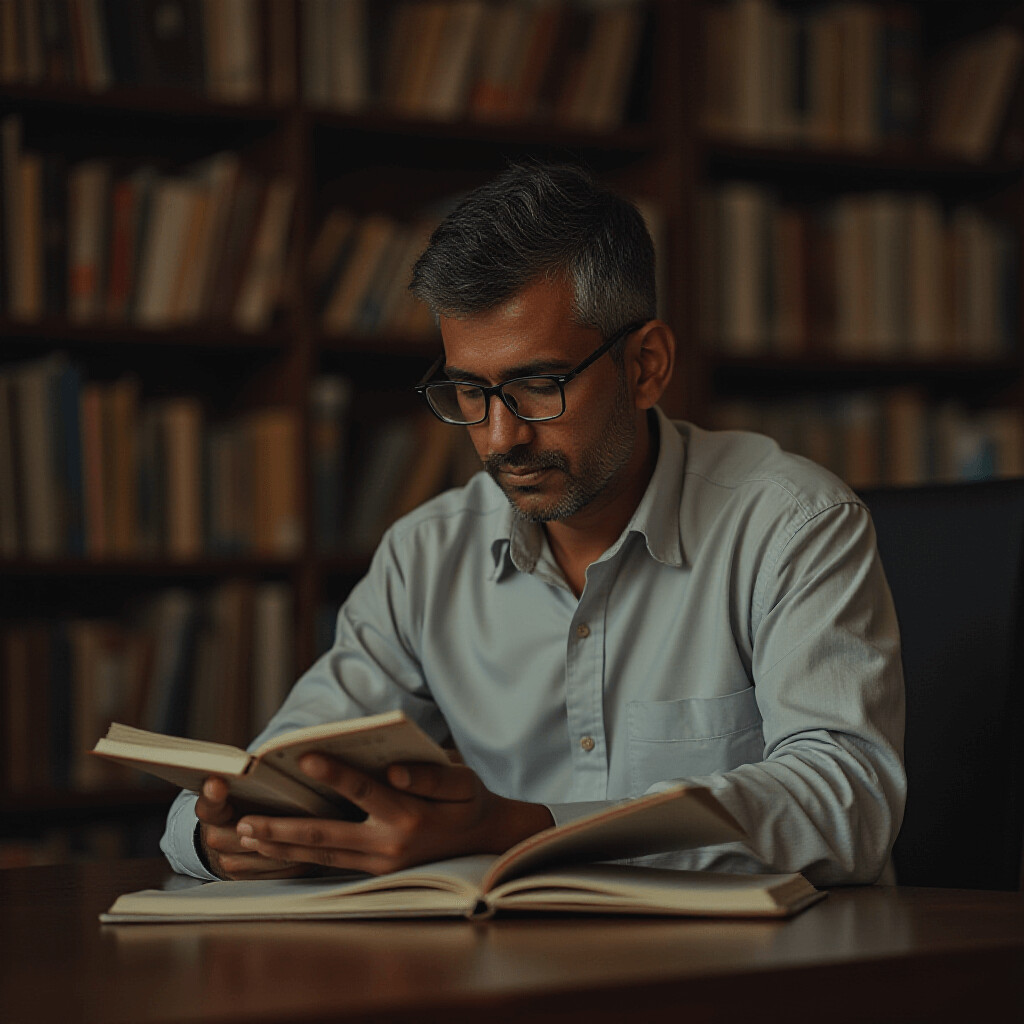} &
\includegraphics[width=0.104\textwidth]{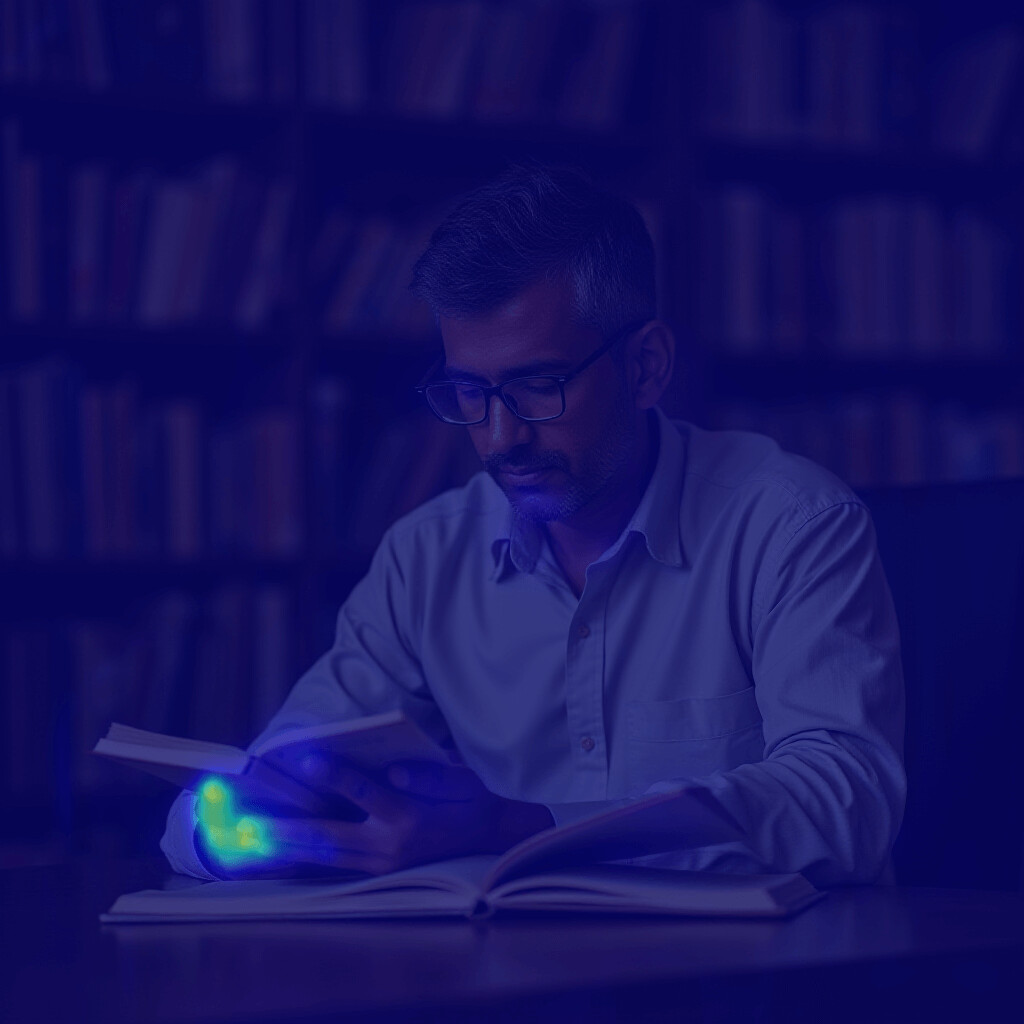} &
\includegraphics[width=0.104\textwidth]{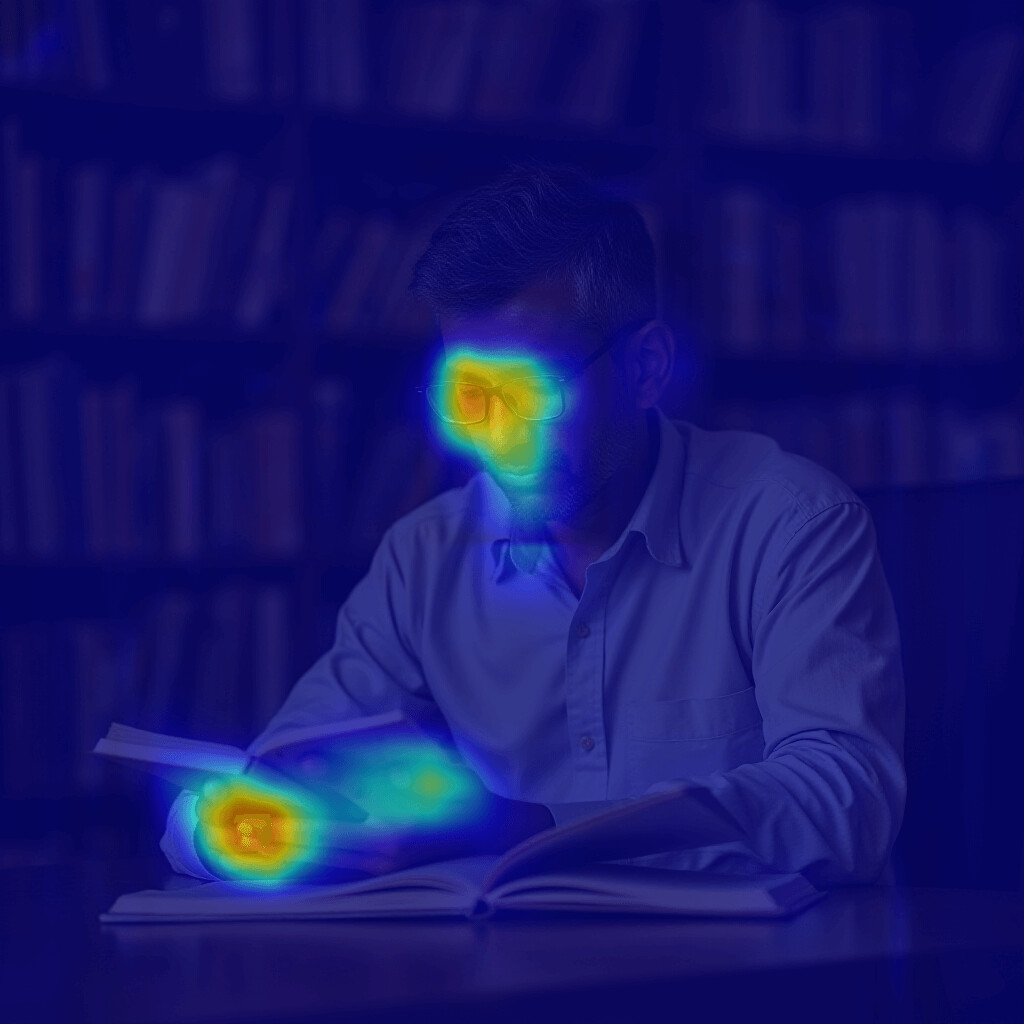} &
\includegraphics[width=0.104\textwidth]{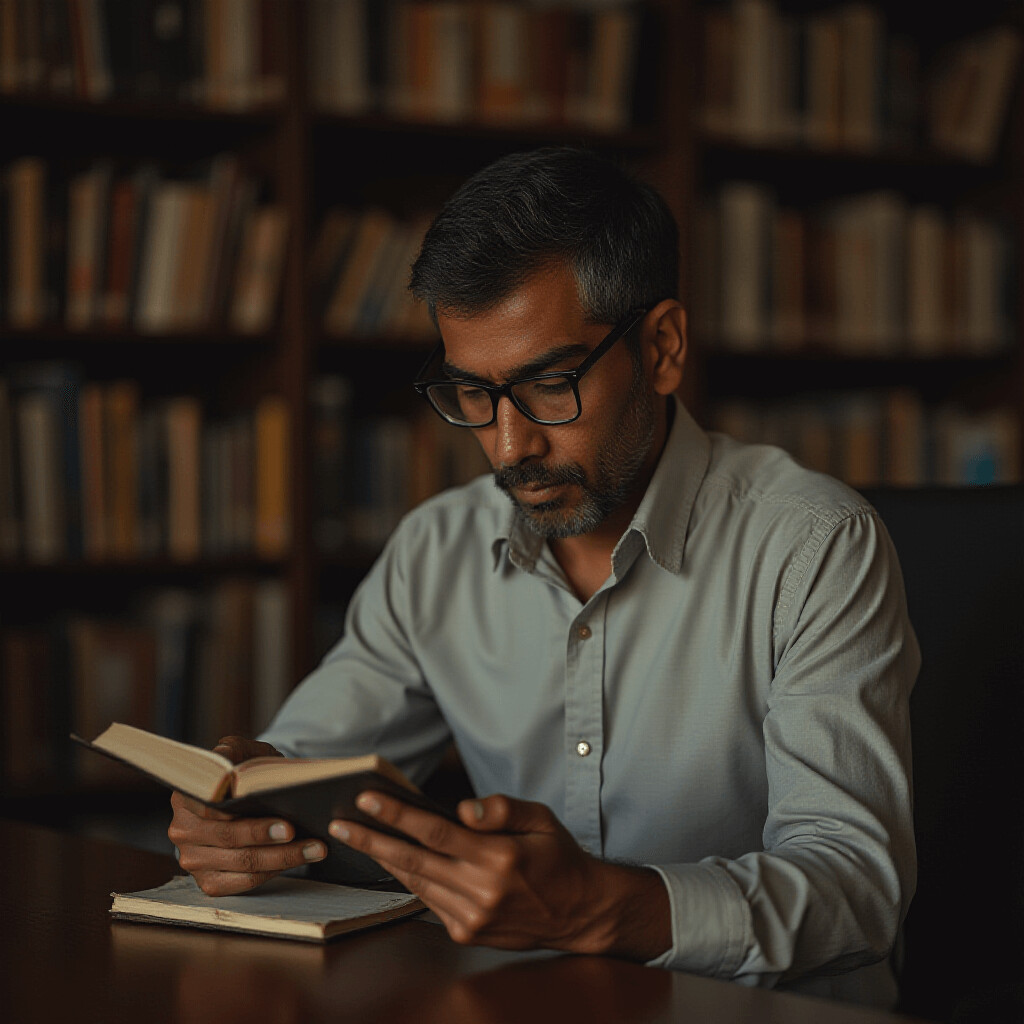} &
\includegraphics[width=0.104\textwidth]{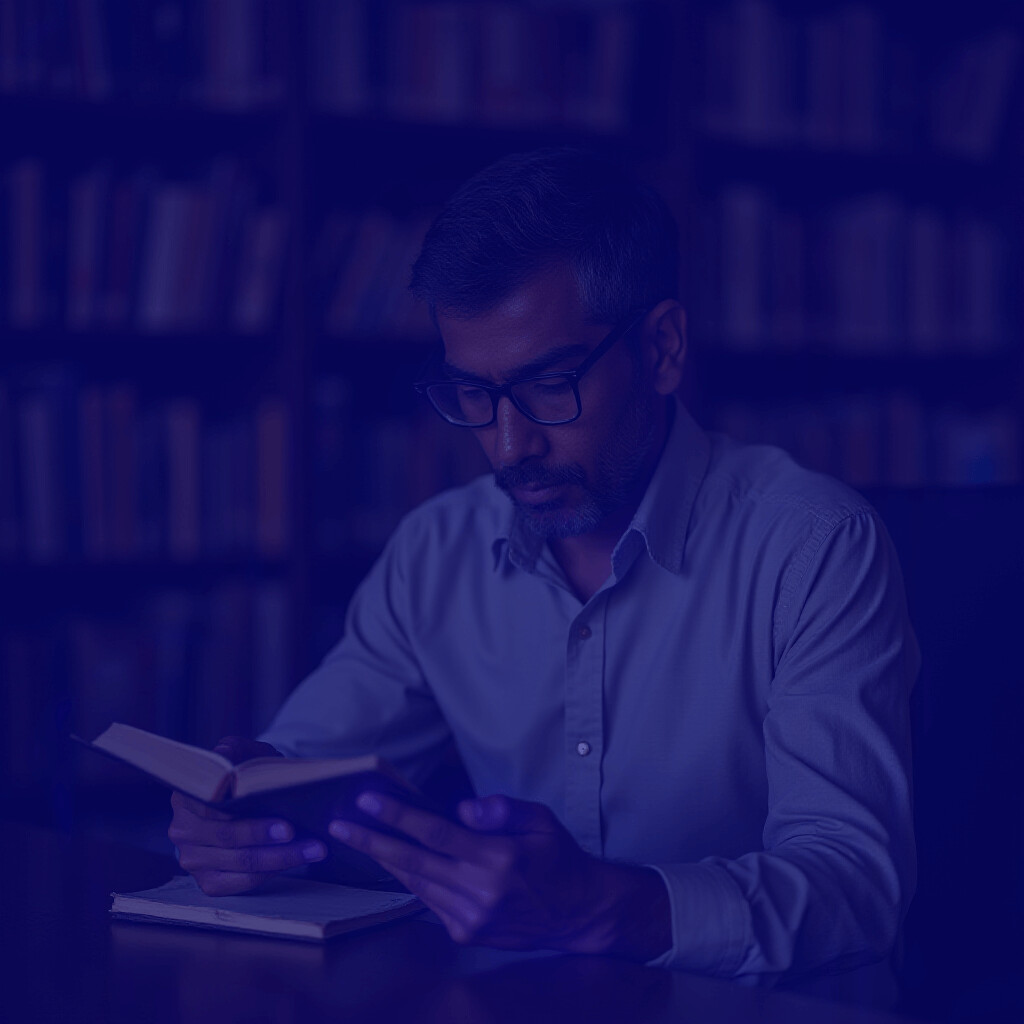} &
\includegraphics[width=0.104\textwidth]{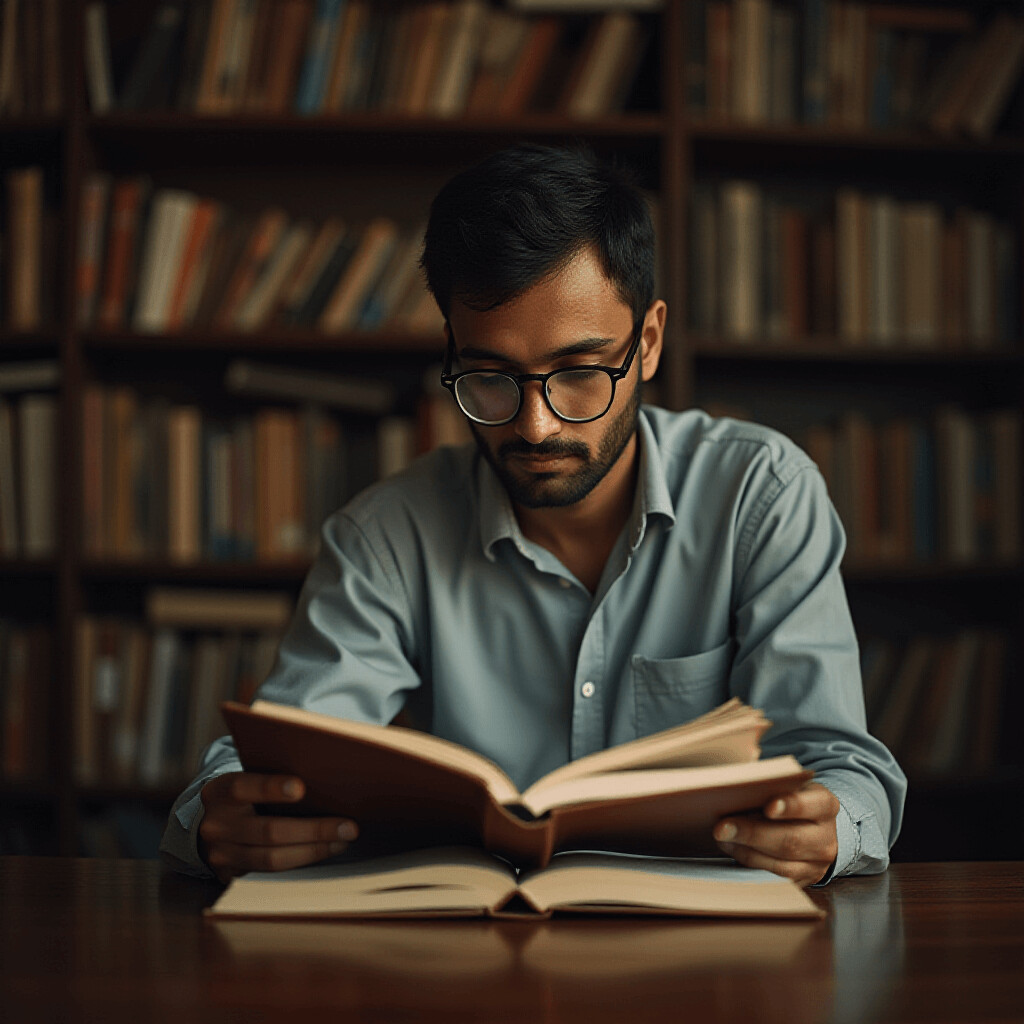} &
\includegraphics[width=0.104\textwidth]{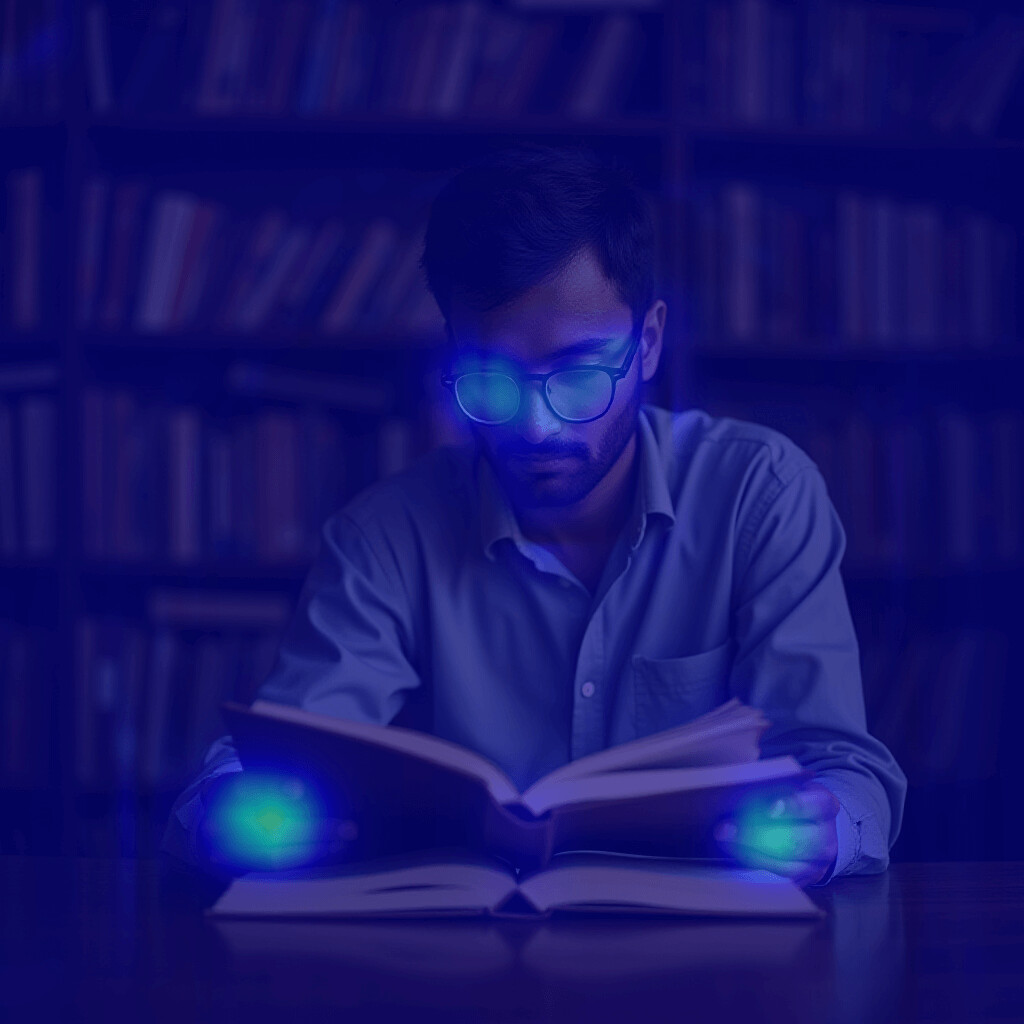} &
\includegraphics[width=0.104\textwidth]{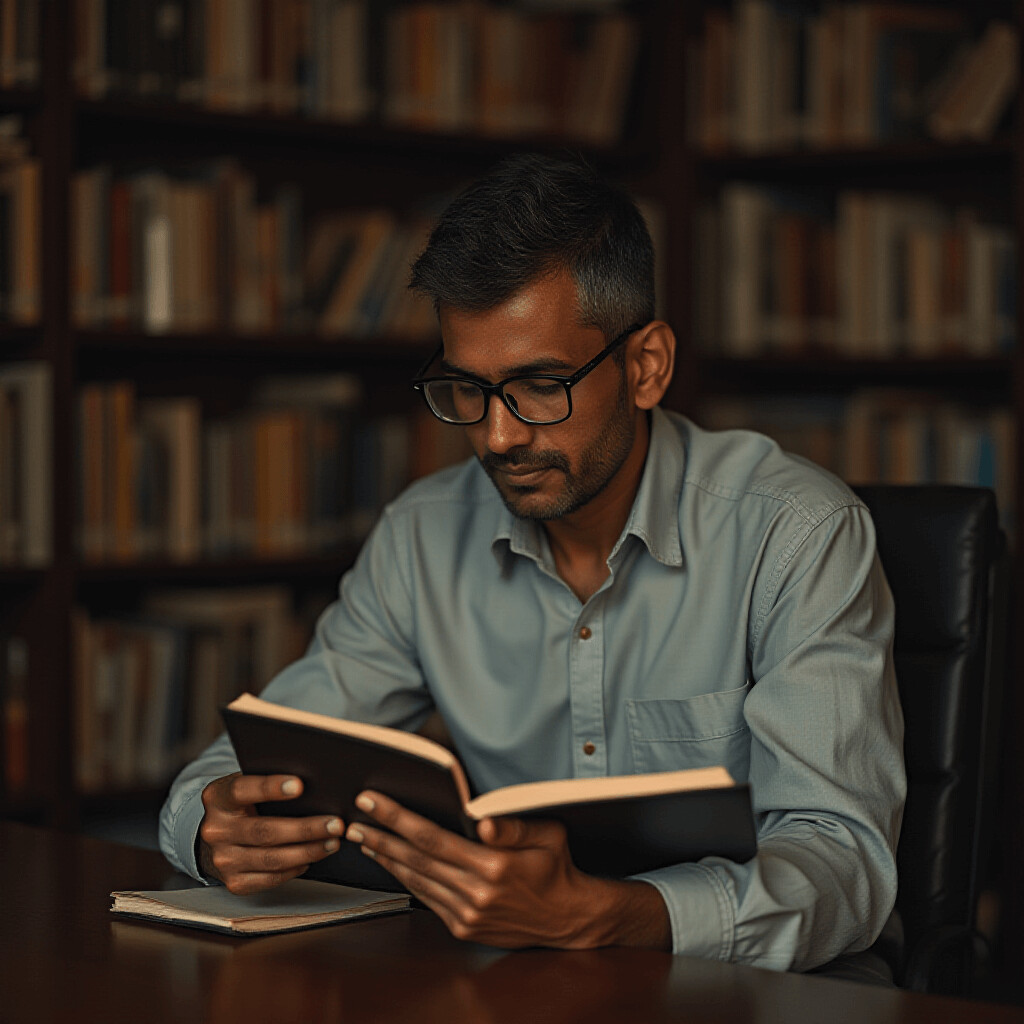} &
\includegraphics[width=0.104\textwidth]{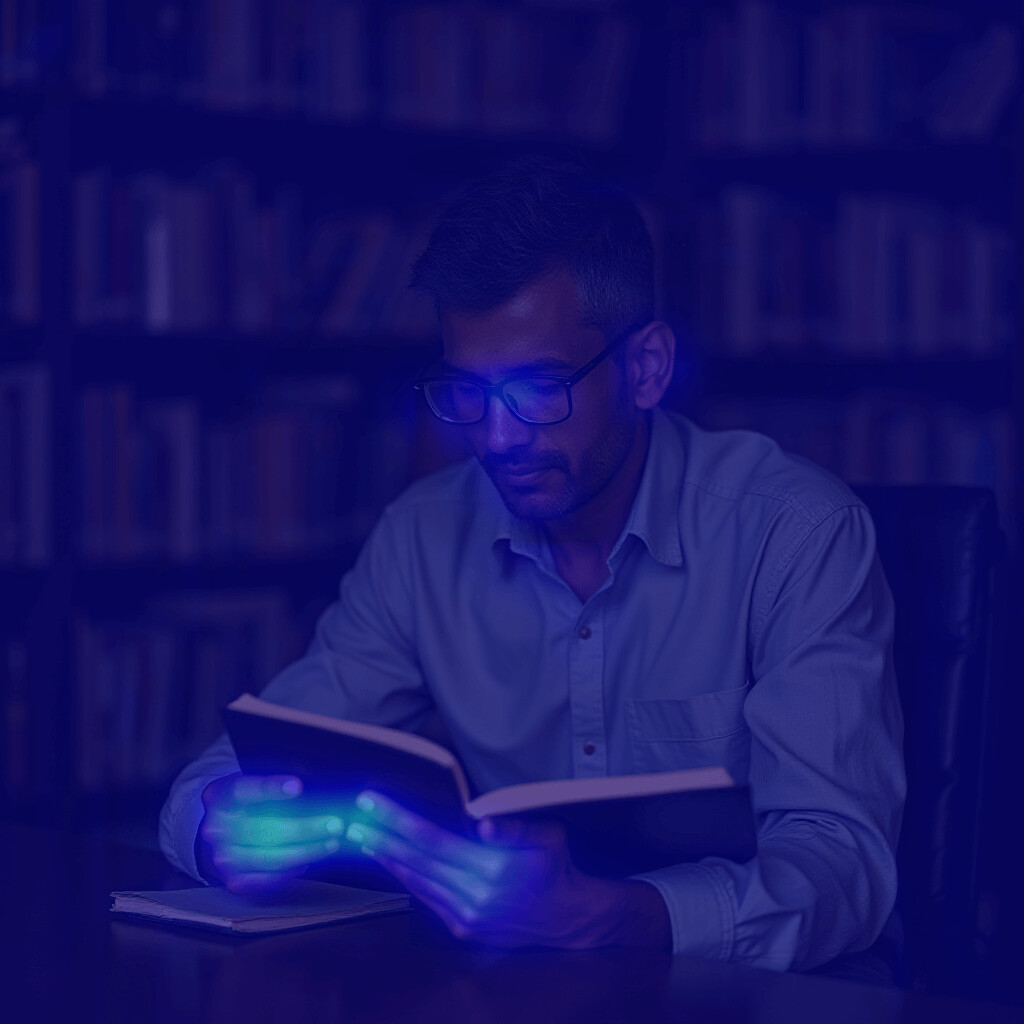} 
\end{tabular}
\caption{\textbf{Qualitative comparison} of DIAMOND using different artifact detectors $\mathcal{AD}$ during trajectory refinement. The detector used is denoted by R - RichHF*, D - DiffDoctor and R + D, a average combination of the obtained masks. The overlay is calculated using the artifact detector shown in superscript, while the subscript indicates the detector used during trajectory correction, with the exception of base indicating the image generated using baseline method.}
\label{fig:diffdetectors2}
\vspace{-1mm}
\end{figure*}

\textbf{Datasets}
To create dataset containing artifacts for the evaluation, we utilized textual prompts from DiffDoctor, which are available on their github repository\footnote{\url{https://github.com/ali-vilab/DiffDoctor}}. The prompts belong to one of the following three categories \textit{humans} (100 prompts), \textit{words} (100 prompts), and \textit{animals} (26 prompts).

\begin{figure}[!t]
\vspace{-3mm}
\centering
\setlength{\tabcolsep}{1pt}
\renewcommand{\arraystretch}{1.0}
\begin{tabular}{ccccc}
 & Baseline & +DiffDoctor & +HPSv2 & +\our{} \\


\rotatebox{90}{\scriptsize\hspace{5pt} {FLUX.1 [dev]}} &
\includegraphics[width=0.11\textwidth]{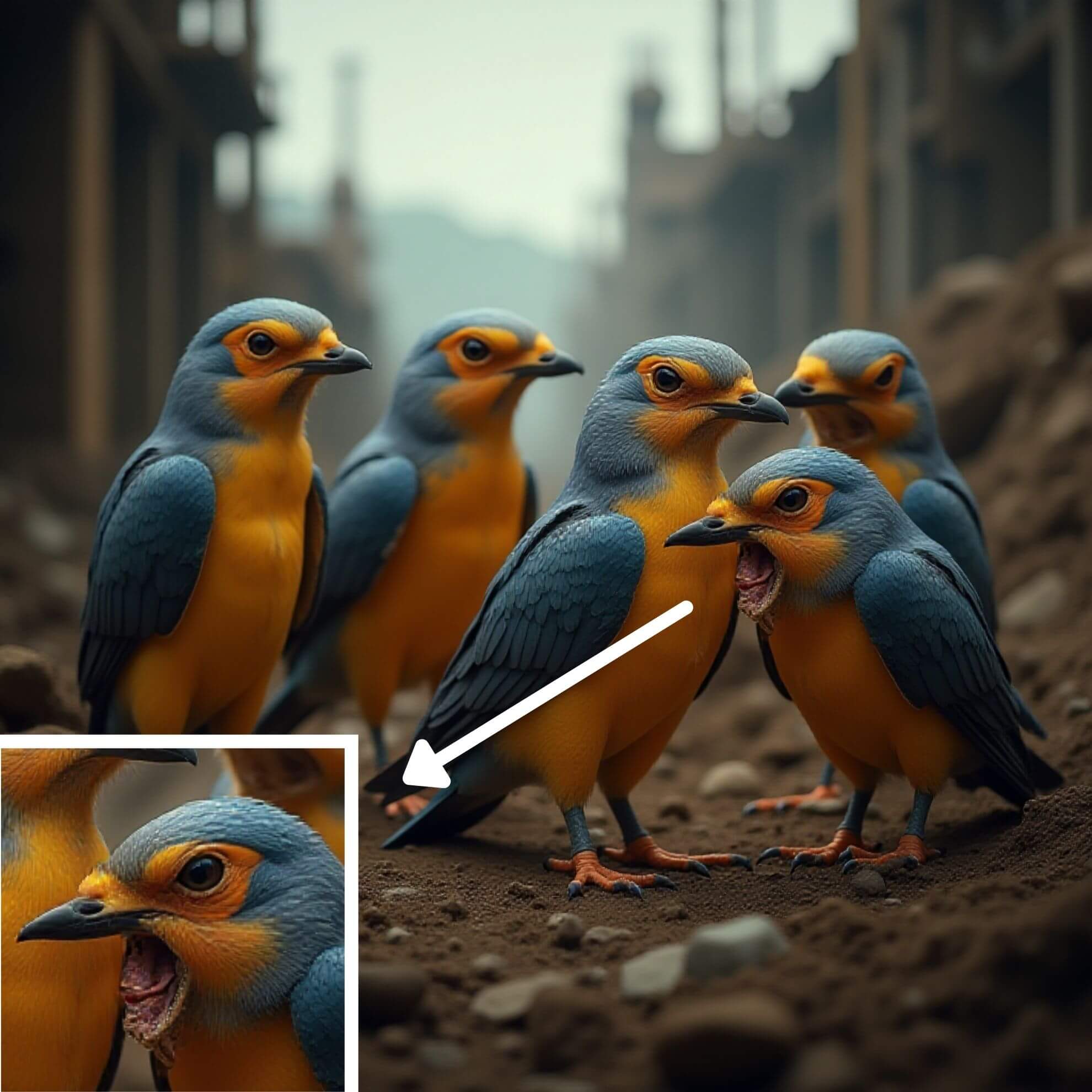} &
\includegraphics[width=0.11\textwidth]{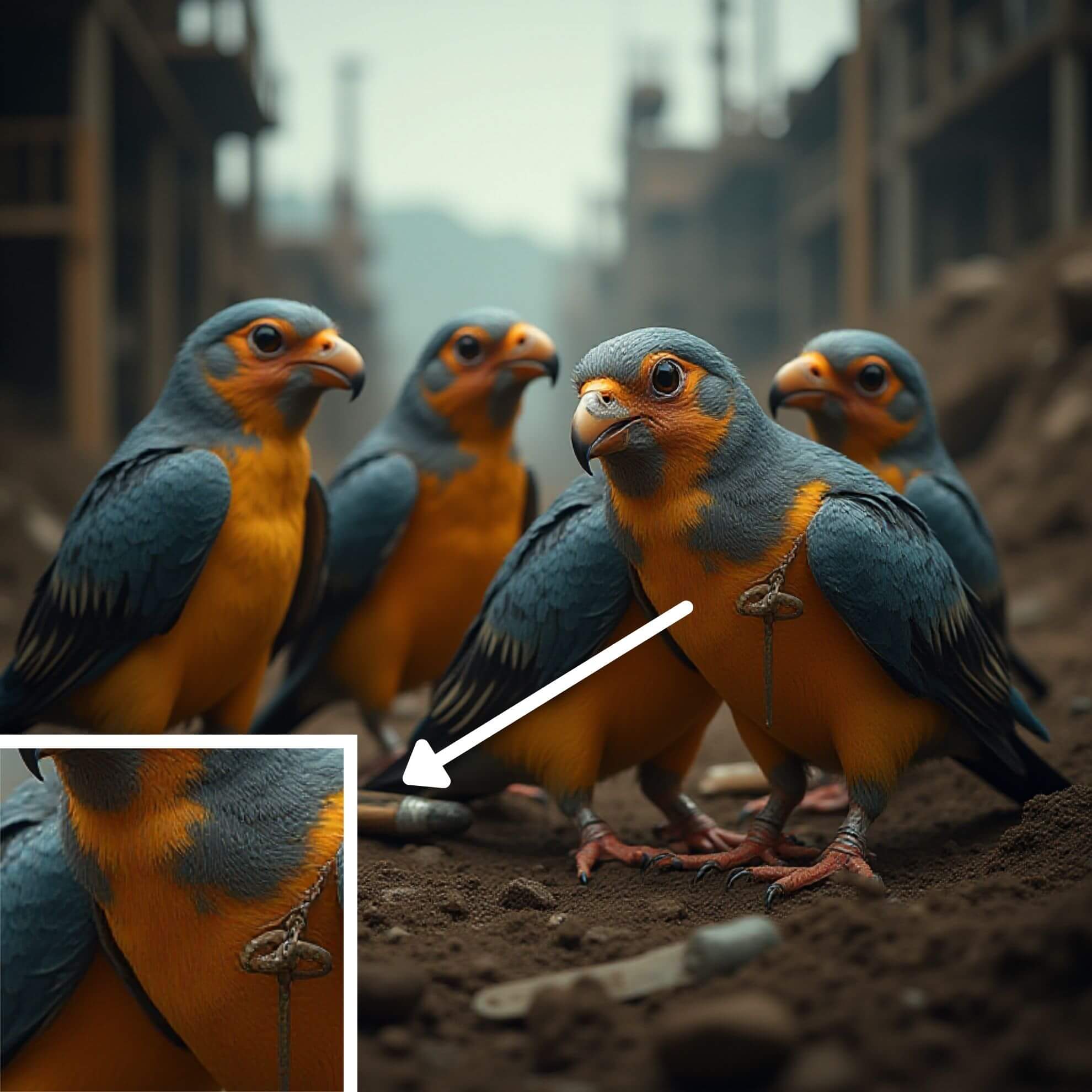} &
\includegraphics[width=0.11\textwidth]{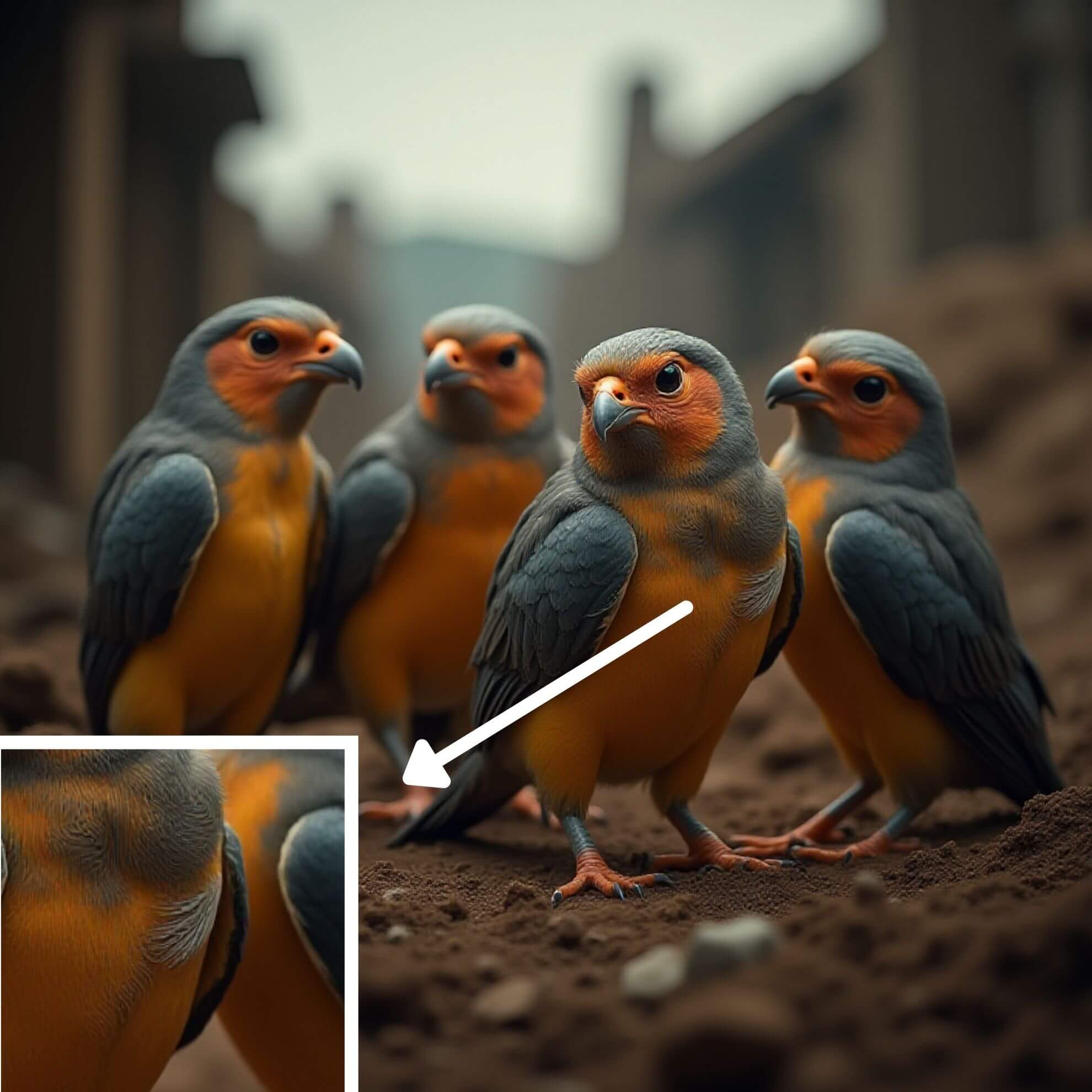}&
\includegraphics[width=0.11\textwidth]{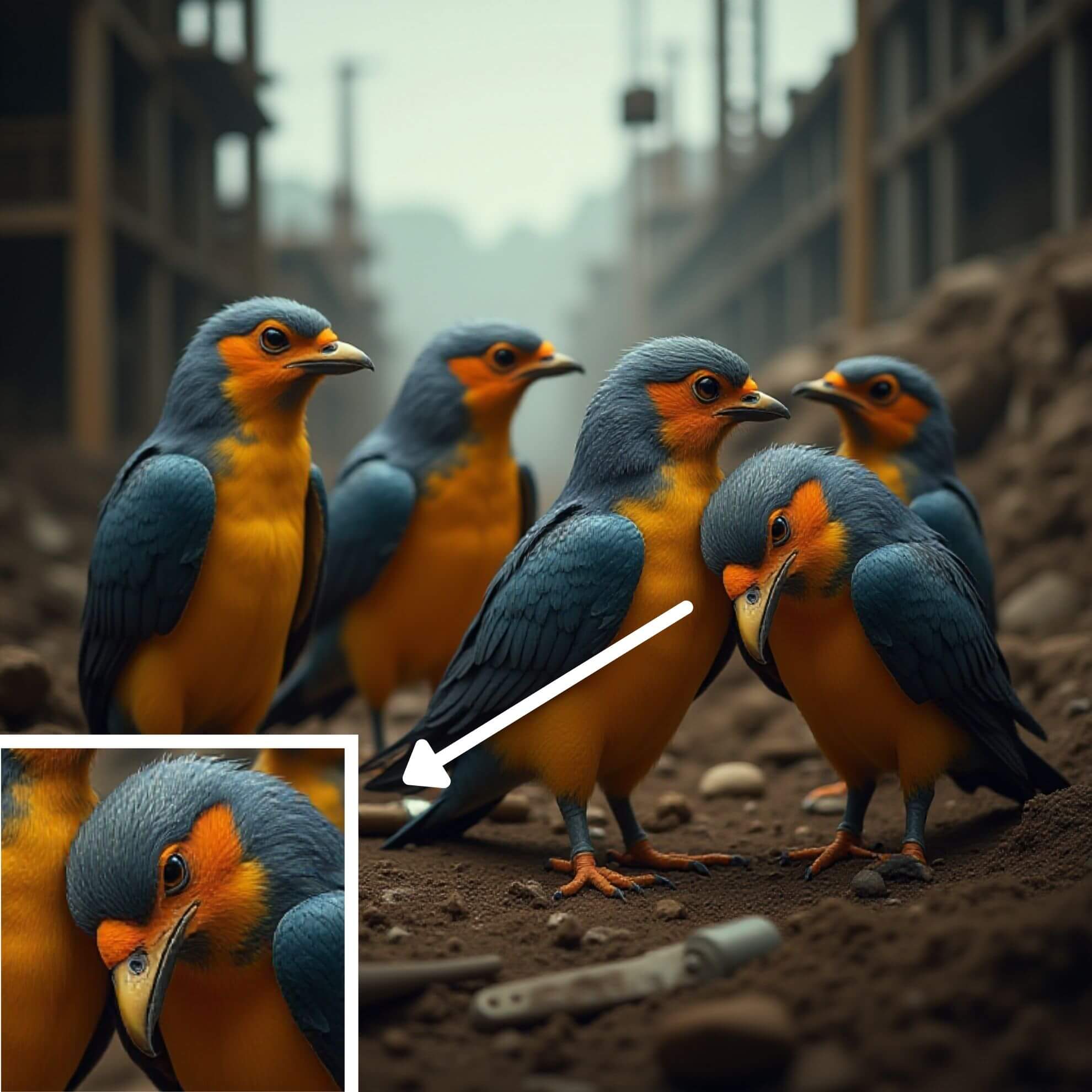} \\

\rotatebox{90}{\scriptsize\hspace{0pt} {FLUX.1 [schnell]}} &
\includegraphics[width=0.11\textwidth]{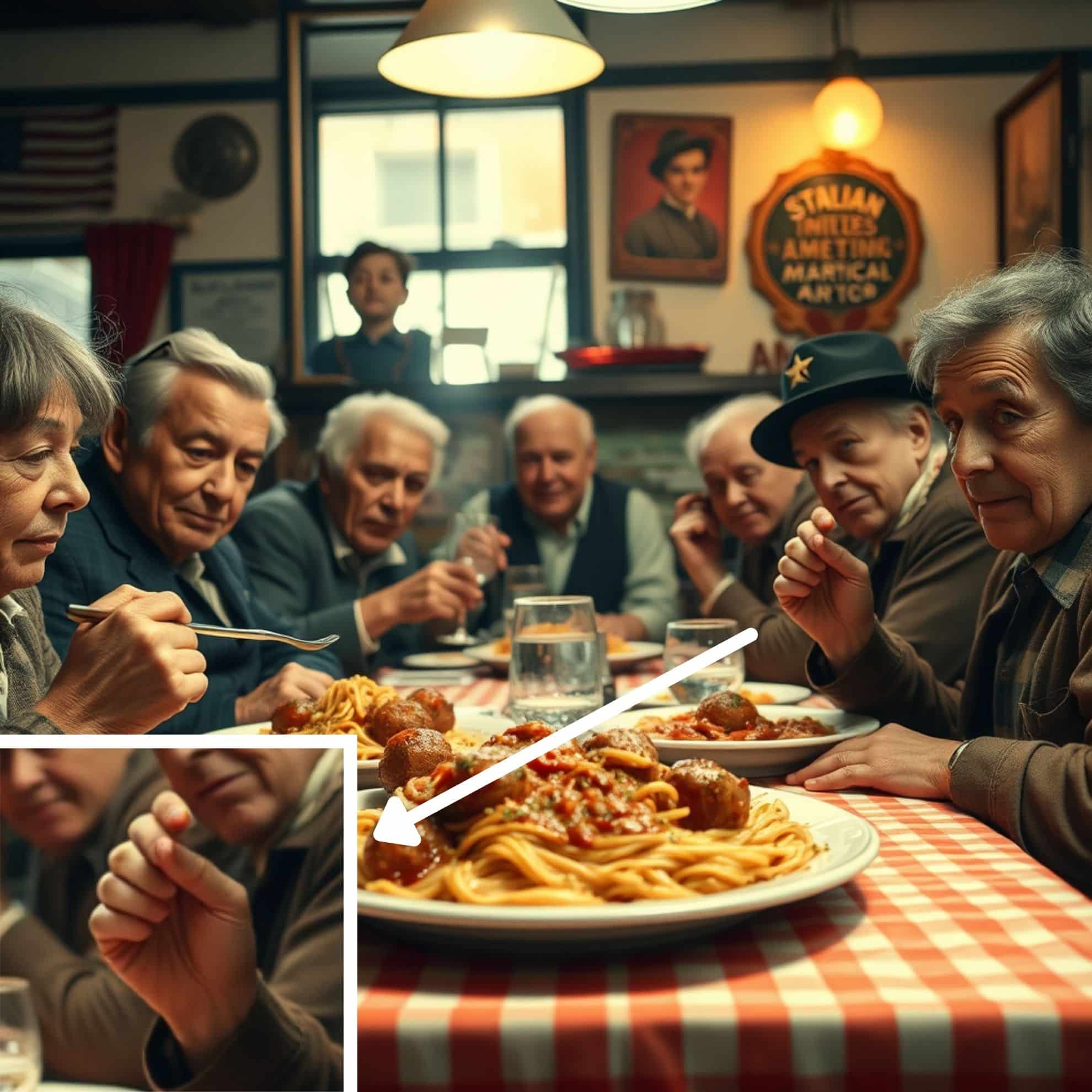} &
\includegraphics[width=0.11\textwidth]{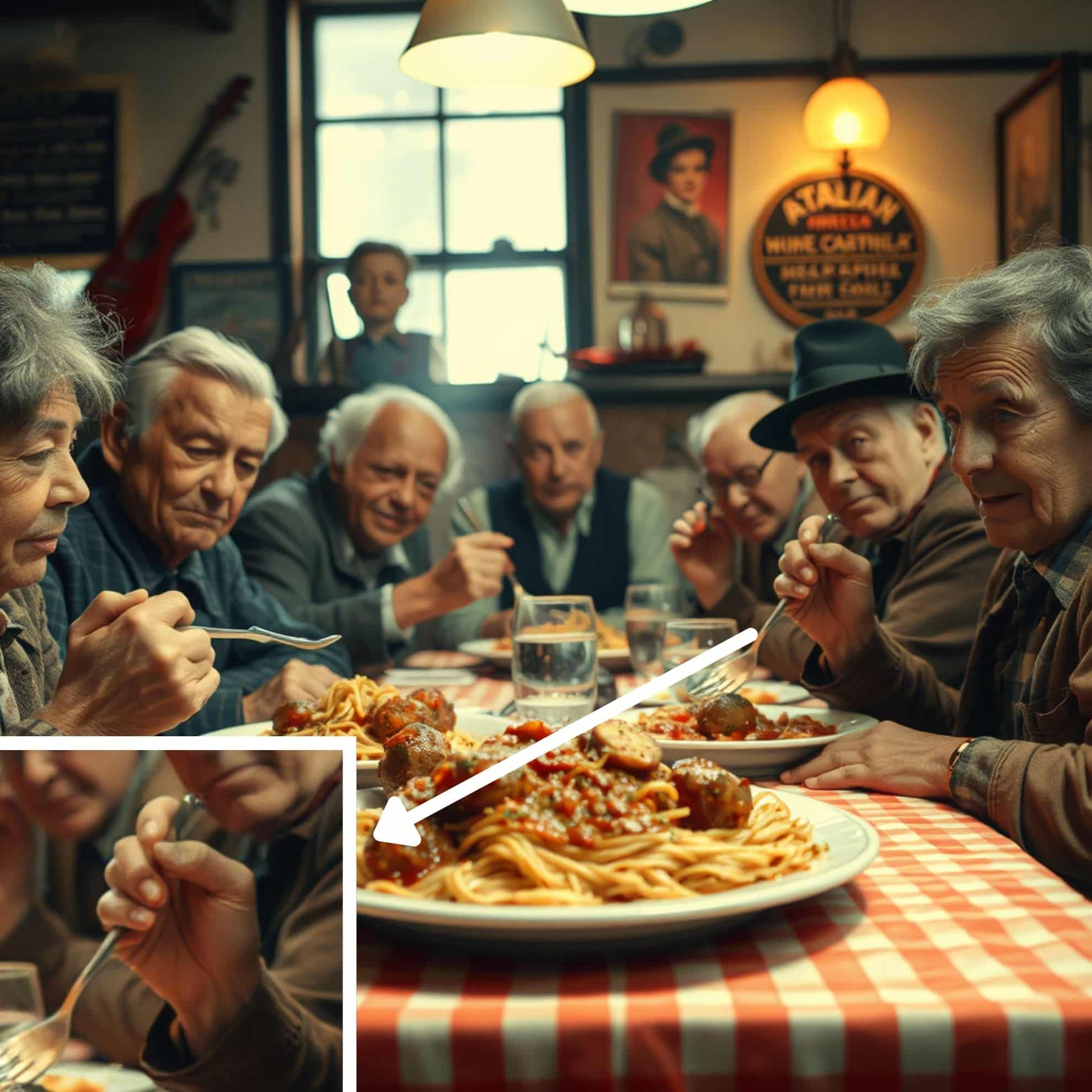} &
\includegraphics[width=0.11\textwidth]{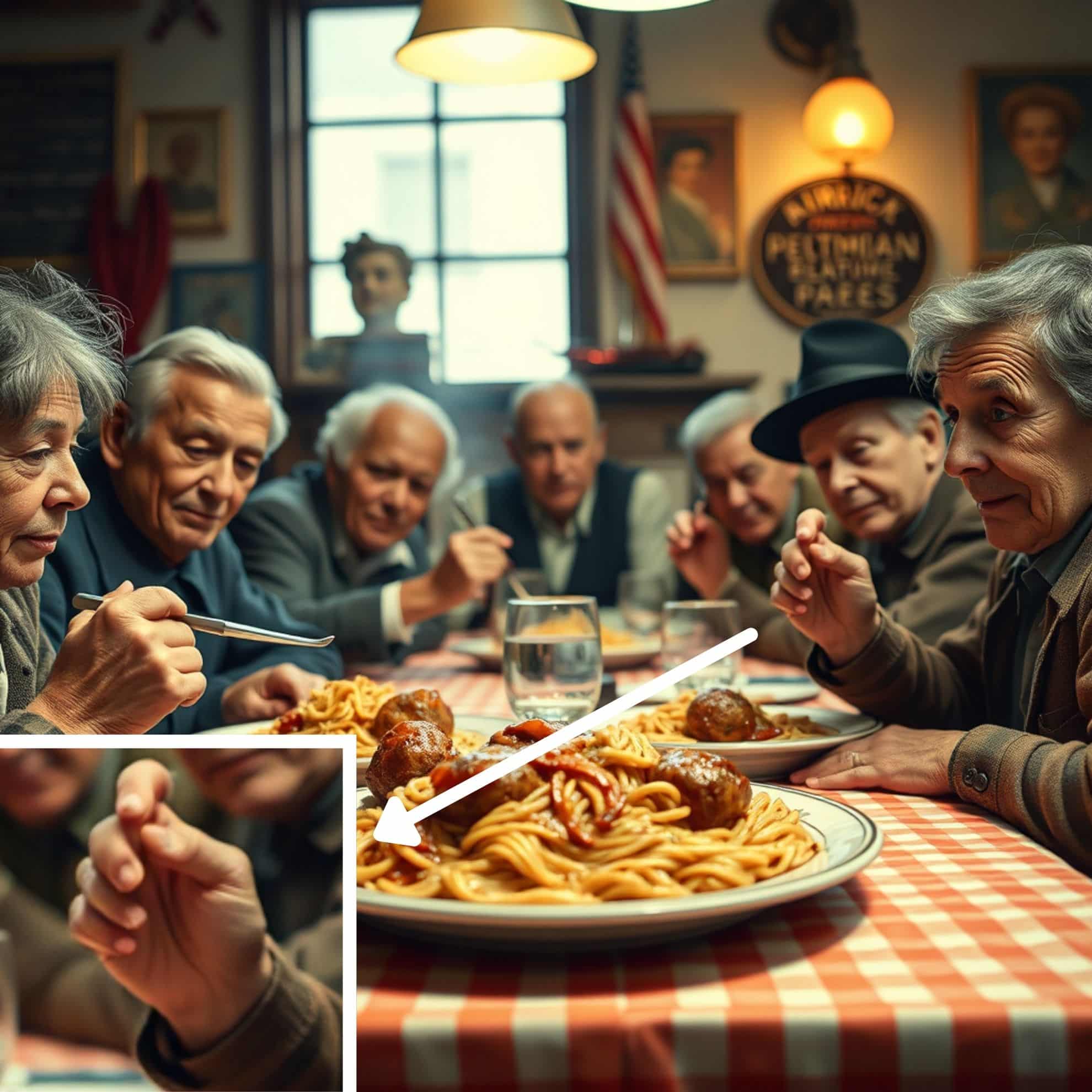} &
\includegraphics[width=0.11\textwidth]{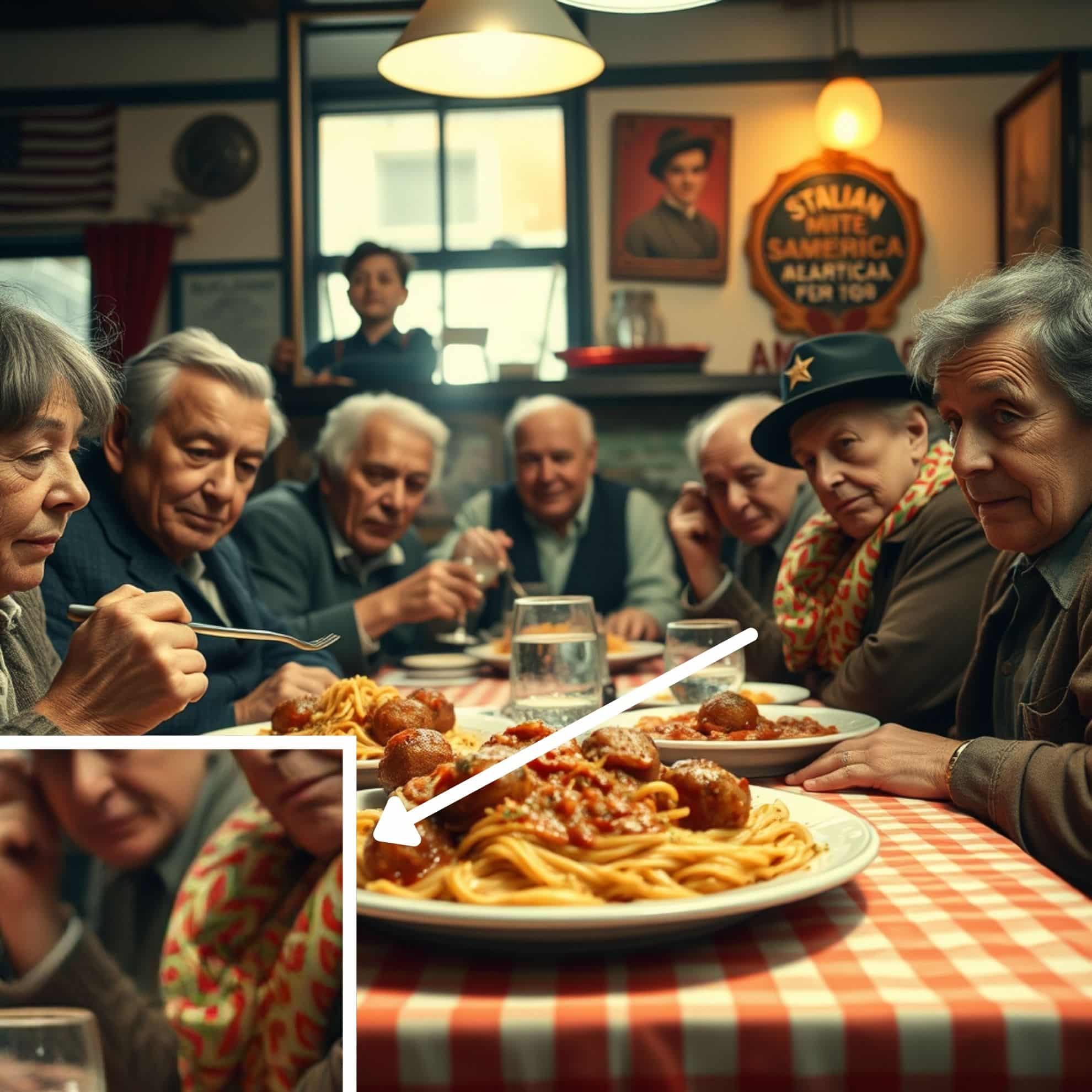} \\
\end{tabular}
\caption{\textbf{Comparison of artifact correction methods.}
Results for FLUX.1 [dev] (top two rows) and FLUX.1 [schnell] (bottom row)
across different correction methods. \our{} was the only one to correctly remove the hand artifact.}
\label{fig:flux1_comparison}
\vspace{-0.5cm}
\end{figure}

The original datasets provided only text prompts without random seeds, so we applied filtering to select images with detectable artifacts. 
The acceptable threshold for accepting a photo containing artifacts was if the generated image contained at least one pixel with a probability greater than 0.5 in the artifact mask. Evaluation prompts have been created for FLUX.2 [dev] (Fig. \ref{fig:comparison}), FLUX.1 [dev] (Figs. \ref{fig:comparison}, \ref{fig:fluxdev1peopledataset}, \ref{fig:handsxl_comparison}, \ref{fig:artifacts}, \ref{fig:diffdetectors2} and \ref{fig:flux1_comparison}, Tab. \ref{tab:comparison_fluxdev_schnell_sdxl}), FLUX.1 [schnell] (Figs. \ref{fig:comparison} and \ref{fig:artifacts}) and SDXL~(Figs.~\ref{fig:comparison} and~\ref{fig:handsxl_comparison}). Detailed instructions for generating datasets are described in the Appendix.

\textbf{Quality Evaluation Metrics}
To evaluate the quality and semantic consistency of generated images, we used the following two metrics, as shown in Tab. \ref{tab:comparison_fluxdev_schnell_sdxl}. The results demonstrate that our \our{} produces high-quality images while exhibiting the lowest number of artifacts compared to state-of-the-art methods. The first of them is CLIP-T, which calculates the cosine similarity between CLIP~\cite{radford2021learning} image embeddings and their text embeddings. We utilized the \texttt{ViT-L-14} model pre-trained on the \texttt{laion2b\_s32b\_b82k}\footnote{\url{https://huggingface.co/laion/CLIP-ViT-L-14-laion2B-s32B-b82K}} dataset. To evaluate the human preferences we used the ImageReward~\cite{xu2023imagereward} model, which predicts the human preference scores for image-text pairs. Consequently, the induced metric measures whether the visual quality was maintained after the artifact-aware sampling process.

\textbf{Artifact Reduction Metrics}
To assess the effectiveness of our method in artifact removal, we employ specific metrics based on the output of the Artifact Detector. The first metric 
Mean Artifact Frequency (MAF) measures the fraction of generated images containing artifacts. An image is classified as containing an artifact if it contains at least one pixel with a artifact-confidence score greater or equal to~0.5. Additionally, the Artifact Pixel Ratio (APR) evaluates the proportion of pixels classified as artifacts relative to the image resolution. Finally, to check whether the correction process preserved the image identity we calculate Mean Absolute Error (MAE) between the original generation process of the baseline model, and the artifact-aware generation using \our{}. This metric is further split into two disjoint regions: non-artifact (NA) and artifact (A) to distinguish between background preservation and correction intensity. As shown in Tab.\ref{tab:comparison_fluxdev_schnell_sdxl} \our{} removes the most artifacts for different base models compared to other techniques. Qualitative results for FLUX.1 [dev], FLUX.1 [schnell] and SDXL are presented in Figs. \ref{fig:fluxdev1peopledataset}, \ref{fig:handsxl_comparison} and~\ref{fig:flux1_comparison}. More numerical and visual results are provided in the Appendix.


\textbf{Base model identity preservation}
A visual comparison of different methods shows that some approaches, such as HandsXL \footnote{\url{https://civitai.com/models/200255}}, rarely alter the global context of the image, see Figs.\ref{fig:fluxdev1peopledataset} and \ref{fig:handsxl_comparison}. We observed that although the CLIP-T metric indicates a high similarity, the resulting images still differ noticeably from the baseline. We introduce a $\mathcal{L}_\text{rec}$ term. Its objective is to more closely approximate the image output produced by the baseline model, regularization term that encourages the output to remain close to that of the baseline:
$
    \mathcal{L}_\text{rec} = \alpha\,\lambda_t\,\text{MAE}( M_{NA}\mathcal{D}(\hat{x}_{0,t}), M_{NA}I_\text{base} ),
$
where $\alpha$ is a hyperparameter controlling the strength of this regularization, $I_\text{base}$ is the image generated using the baseline model, and $M_{NA}$ is the corresponding binary mask for non-artifact regions.  $\mathcal{L}_{\text{rec}}$ is applied from the second step, since after the first step, the model has not yet obtained a reliable gradient direction. A qualitative and numerical comparison between our method without regularization ($\alpha=0$) and with correction (our-Fine, $\alpha=0.1$) is presented in Fig.~\ref{fig:artifacts} and Tab.~\ref{tab:gradient_norm}. Setting of $\alpha=0.1$ allows for the best trade-off between artifacts and base model identity preservation. More results are in the Appendix.










\textbf{Detectors} Artifact detection can be performed using pre-trained models. Tab.\ref{tab:detectors} presents the impact of different detectors on the quality of the generated images. Notably, the metrics  CLIP-T and ImageReward, which are independent of the artifact detector demonstrate that \our{} paired with the artifact detector from DiffDoctor achieves the best overall generative performance. A qualitative comparison is provided in Fig. \ref{fig:diffdetectors2} and more visualizations are in the Appendix.
The results indicate a strong dependency between detector quality and overall model performance. 
DiffDoctor produces accurate, subtle modifications, whereas RichHF* introduces excessive artifacts that significantly alter the image.

\textbf{Gradient normalization}
After computing the artifact mask, we evaluate the corresponding artifact loss $\mathcal{L}_{a}$. The resulting gradient $\nabla_{x_t}\mathcal{L}_{a}$ is then used during inference to estimate the shift required for trajectory correction. To stabilize and improve the magnitude of this shift, we apply gradient normalization. Fig. \ref{fig:nonorm} compares our method with and without gradient normalization, where Tab. \ref{tab:gradient_norm} presents a numerical comparison. Both show that normalization amplifies the effect of artifact mitigation.



\begin{table}[!t]
\centering
\caption{\textbf{Ablation Study of gradient and $\mathcal{L}_\text{rec}$ normalization on \textit{people} dataset}, where: \textbf{w/o n:} no normalization;\textbf{ w/ n:} normalization; \textbf{w/ n; $\alpha=0.1$:} normalization with $\mathcal{L}_\text{rec}$ normalization. W/n: allows for a significant improvement in the trajectory toward artifact removal. $\mathcal{L}_\text{rec}$ preserves the semantics of the base model while slightly restoring artifacts. }

\setlength{\tabcolsep}{2pt}
{
\fontsize{5.3pt}{8.9pt}\selectfont
\begin{tabular}{lccccccc}
\hline
Model              & CLIP-T $\uparrow$ & Mean Artifact Freq $\downarrow$ & MAE (A) $\downarrow$ & MAE (NA) $\downarrow$ \\ \hline
Baseline FLUX.1 {[}dev{]}   & 35.820 $\pm$ 0.175  & 100.000 $\pm$ 0.000       & -       & -          \\
+ \textbf{\our{} w/o n;}  & 35.850 $\pm$ 0.213  & 95.000 $\pm$ 3.464        & 4.741 $\pm$ 0.508       &  1.291 $\pm$ 0.069          \\
+ \textbf{\our{} w n;}           & 35.762 $\pm$ 0.101  & \textbf{15.500 $\pm$ 2.380 }      & 25.764 $\pm$ 0.604       & 9.545 $\pm$ 0.240          \\
+ \textbf{\our{} w n; $\alpha=0.1$} & \textbf{35.771 $\pm$ 0.181}  & 20.000 $\pm$ 2.449 & \textbf{23.690 $\pm$ 0.944 }& \textbf{8.187 $\pm$ 0.168} \\ \hline
\end{tabular}
}
\label{tab:gradient_norm}
\vspace{-3mm}
\end{table}

\begin{table}[!t]
\centering
\caption{\textbf{Ablation Study of Artifact Detector Selection for \our{} with FLUX.1 [dev] as a base model on \textit{people} dataset.} The highest CLIP-T and ImageReward scores are achieved by DiffDoctor.
}
\setlength{\tabcolsep}{9pt}
{
\fontsize{8.5pt}{9.8pt}\selectfont
\begin{tabular}{lcc}
\hline
Detector &  CLIP-T $\uparrow$  & ImageReward $\uparrow$ \\ \hline
DiffDoctor & 35.762 $\pm$ 0.101  &0.968 $\pm$ 0.034   \\
RichHF* & 35.624 $\pm$ 0.225 & 0.922 $\pm$ 0.037  \\
DiffDoctor+RichHF* & 35.682 $\pm$ 0.181 & 0.952 $\pm$ 0.048   \\ \hline
\end{tabular}
}
\label{tab:detectors}
\vspace{-0.5cm}
\end{table}


\section{Conclusion}
We presented \our{}, a training-free inference time method for mitigating visual artifacts in rectified flow and diffusion-based generative models. Unlike prior approaches that rely on post hoc refinement or model fine-tuning, \our{} intervenes directly during sampling by correcting the generative trajectory when artifact-prone states are detected.
The core idea of DIAMOND is to reconstruct an estimate of the clean sample at intermediate timesteps, enabling reliable artifact detection and gradient-based trajectory correction in a suitable domain. This allows the method to apply localized and targeted updates that suppress artifacts while preserving global structure and prompt alignment.
\textbf{Limitation} Our method relies on the performance of both the underlying generative model and the artifact detector. When an artifact is not correctly identified by the detector, the corresponding correction may be suboptimal.

\bibliography{flux}
\bibliographystyle{icml2026}

\newpage
\clearpage
\appendix
\onecolumn
\section*{Appendix}
In the supplementary materials, we provide additional experiments. Section \ref{sec:exp_setup} details additional configurations for implementing the Flow and Diffusion models and describes the dataset generation. Appendix \ref{sec:add_results} presents more visual results of our method that reduce artifacts and also describes ablation on adding an MAE regularizer during the trajectory and selecting the correct gradient scaling value from the artifact detector.

\section{Experimental setup}
\label{sec:exp_setup}

While the implementation of our method will be released publicly after publication, we focus here on documenting the exact configuration required to reproduce the benchmark results. In particular, we describe how the evaluation benchmark was constructed, including the set of prompts and fixed random seeds that consistently induce visual artifacts in the baseline models. We note that such a prompt–seed benchmark is especially valuable in the context of FLUX.2 [dev], where evaluation setup are currently limited.

\subsection{Implementation of existing methods}
\label{subsec:impl_existing_methods}

A public repository was used during the experiments, providing a ready-made script for fine-tuning flow matching models, including DiffDoctor and HPSv2 \cite{wang2025diffdoctor}. These methods require expensive resources, so an NVIDIA H200 Tensor Core GPU was used as the hardware. The FLUX.1 [dev] and FLUX.1 [schnell] models were fine-tuned according to the hyperparameters from the paper, with a rank of LoRA = 64. The third method - HandsXL, is a publicly available version of LoRA weights, applicable to the FLUX.1 [dev] and Stable Diffusion XL. For the flux model, version 1.0 was used, while version 5.5 was used for the stable diffusion model. They were loaded according to the instructions, which emphasized that it requires a valid lora\_{scale}. In our experiments, due to the large impact of the LoRA adapter on the final image, the value lora\_scale = 0.1 was used, which is also recommended as the default value. In contrast, experiments were conducted using the official FLUX.2 [dev] open-weight checkpoint, without any form of model quantization.


\subsection{Implementation of \our{}}
\label{subsec:implement_diamond}
The hyperparameter values used to scale the gradient values are presented in Tab. \ref{tab:hyperparameters}, including information on the number of inference steps and guidance scale. For SDXL, this guidance scale is for CFG, and for Flow Matching Models, guidance is distilled and built into the model.

\begin{table*}[b]
\caption{\textbf{Hyperparameters Settings for \our{}.} Four models were tested on three datasets: \textit{animals}, \textit{people}, and \textit{words}. $\lambda_{start}$, $\lambda_{end}$ - initial and final values of the $\lambda_t$ parameter. $p$ - exponent controlling the rate of decay of the $\lambda_t$. $\tau_{start}$, $\tau_{end}$ define the range of steps in which the gradient from the classifier is added during the trajectory. The following configuration is recommended and accepted as default.}
\centering
\setlength{\tabcolsep}{4pt}
{
\fontsize{9pt}{13pt}\selectfont
\begin{tabular}{lcccccccc}
\hline
Base model + \our{}           & \multicolumn{1}{l}{Dataset} & $\lambda_{start}$ & $\lambda_{end}$ & $\tau_{\text{start}}$ & $\tau_{\text{end}}$ & $p$ & inference steps & guidance scale \\ \hline
\multirow{3}{*}{Flux.1 [dev]} & Animals                     & 25                & 1               & 0                     & 0                   & 2   & 10              & 3.5            \\
                              & People                      & 25                & 1               & 0                     & 0                   & 2   & 10              & 3.5            \\
                              & Words                       & 25                & 1               & 0                     & 0                   & 3   & 10              & 3.5            \\ \hline
Flux.1 [schnell]             & People                      & 40                & 1               & 0                     & 0                   & 4   & 4               & 3.5            \\ \hline
\multirow{3}{*}{Flux.2 [dev]} & Animals                     & 20                & 1               & 5                     & 5                   & 4   & 30              & 4.0            \\
                              & People                      & 20                & 1               & 5                     & 5                   & 4   & 30              & 4.0            \\
                              & Words                       & 20                & 1               & 5                     & 5                   & 4   & 30              & 4.0            \\ \hline
SDXL                          & People                      & 25                & 1               & 0                     & 5                   & 4   & 30              & 7.5            \\ \hline
\end{tabular}
}
\label{tab:hyperparameters}
\end{table*}

\subsection{Datasets}
\label{subsec:datasets}
In this section, we describe the detailed process of dataset creation, specifying which datasets were used for each model due to its properties. Four distinct starting seeds were defined to create four independent image sets for each dataset. For the \textit{people} and \textit{words} datasets, the following seeds were used: 4000, 40000, 400000, and 4000000. However, for the \textit{animals} dataset, we chose different initial seeds for better-generated photos with more artifacts: 4000, 40000, 400000, and 3000000.

To create each set, we started by generating an image for the first prompt and the first initial seed. This image was then processed by the Artifact Detector. If at least one pixel was found with a probability greater than 0.5 in the artifact mask, the seed for that prompt was saved. Otherwise, the seed for that prompt was discarded, and the image was generated again, but with a seed incremented by 1. The condition was then checked again to see if it contained at least one artifact. If not, the process was repeated. A maximum of 1000 iterations was allowed for each prompt to find a valid image. This process resulted in four separate sets of prompt-seed pairs for each dataset. The generation was repeated for all the models separately.

Our method was evaluated using the latest model from the FLUX.2 series, namely the \textbf{FLUX.2 [dev]} variant.\footnote{\url{https://github.com/black-forest-labs/flux2}}. It generates the highest quality images, so the probability of obtaining any artifacts is rare. It is characterized by very precise placement of the given text in images and attention to detail. Generating a massive words dataset is a big challenge because despite several hundred attempts, some prompts do not generate any artifacts at all. Nevertheless, we included this model in our experiments using prompts from the {\textit{animals} dataset}, which we managed to create, examining the effect of changing the trajectory on improving image quality.  It is noteworthy that the inference time is longer, reaching up to 50 steps compared to standard flow matching models, which often use only a few. In our experiments, we assumed inference\_steps=30 as a good choice for obtaining the best trade-off.

The following three prompts from  the teaser are presented below:
\begin{itemize}
    \item \textbf{Prompt 1:} "First-person POV proposal scene. The viewer's hands are placing a large diamond ring onto the finger of a standing woman in front of them. The woman's face is visible, smiling and emotional, wearing an elegant pastel dress. Low-angle perspective shows the man kneeling. Colorful spring garden with pink flowering trees in the background. Bright natural daylight, rich but realistic colors. Focus on realistic hands, accurate finger anatomy, detailed diamond reflections. Photorealistic, no cinematic effects, no text or symbols."
    \item \textbf{Prompt 2:} "Two athletes from different backgrounds competing in a karate tournament, mats, focused expressions, dynamic, energetic, vivid details."
    \item \textbf{Prompt 3:} "A busy shopping district with a neon sign that says "Open 24/7," urban, matte, sharp focus, vibrant colors."
\end{itemize}

The \textbf{FLUX.1 [dev]} model was configured according to the work \cite{wang2025diffdoctor}, where inference steps=10 was assumed. This development version produces more frequent visual errors in photos, resulting from the poorer model architecture. The model tends to duplicate, omit, or blur letters, see Figs. \ref{fig:dev_text} and \ref{fig:dev_text2}. Furthermore, considering only 10 steps, the probability of artifact occurrence increases. Therefore, all {three datasets} were used in the experiments, examining both the correct creation of \textit{humans}, \textit{animals}, and grammatically correct \textit{texts}.

\begin{figure}[h]
\centering
\setlength{\tabcolsep}{1.2pt}
\renewcommand{\arraystretch}{0.9}
\begin{tabular}{lcccc}
 & $t_{3}$  & $t_{2}$  & $t_{1}$ & $t_{0}$ \\
\rotatebox{90}{\hspace{6mm}FLUX.1 [schnell]} & \includegraphics[width=0.24\textwidth, height=0.24\textwidth]{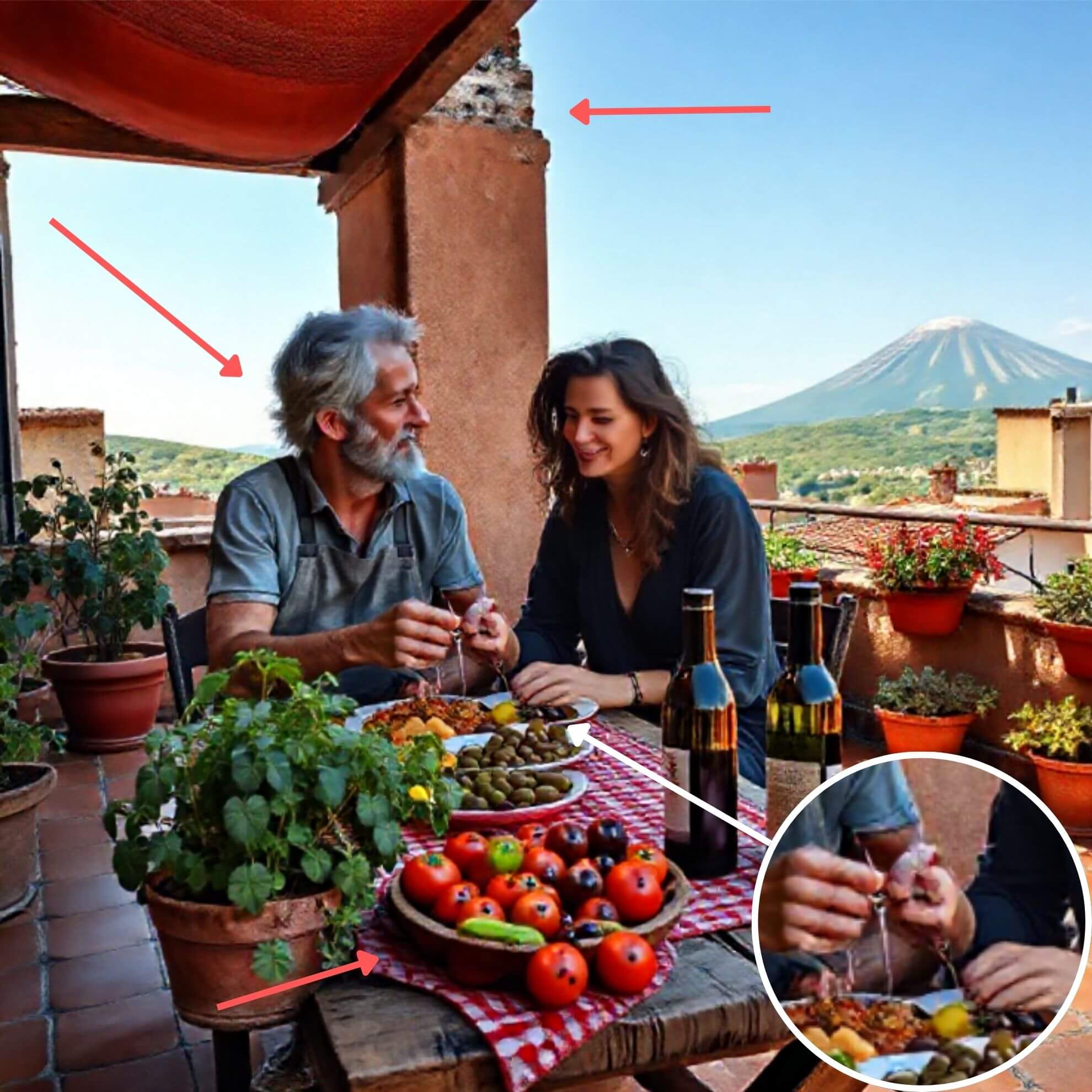} &
\includegraphics[width=0.24\textwidth, height=0.24\textwidth]{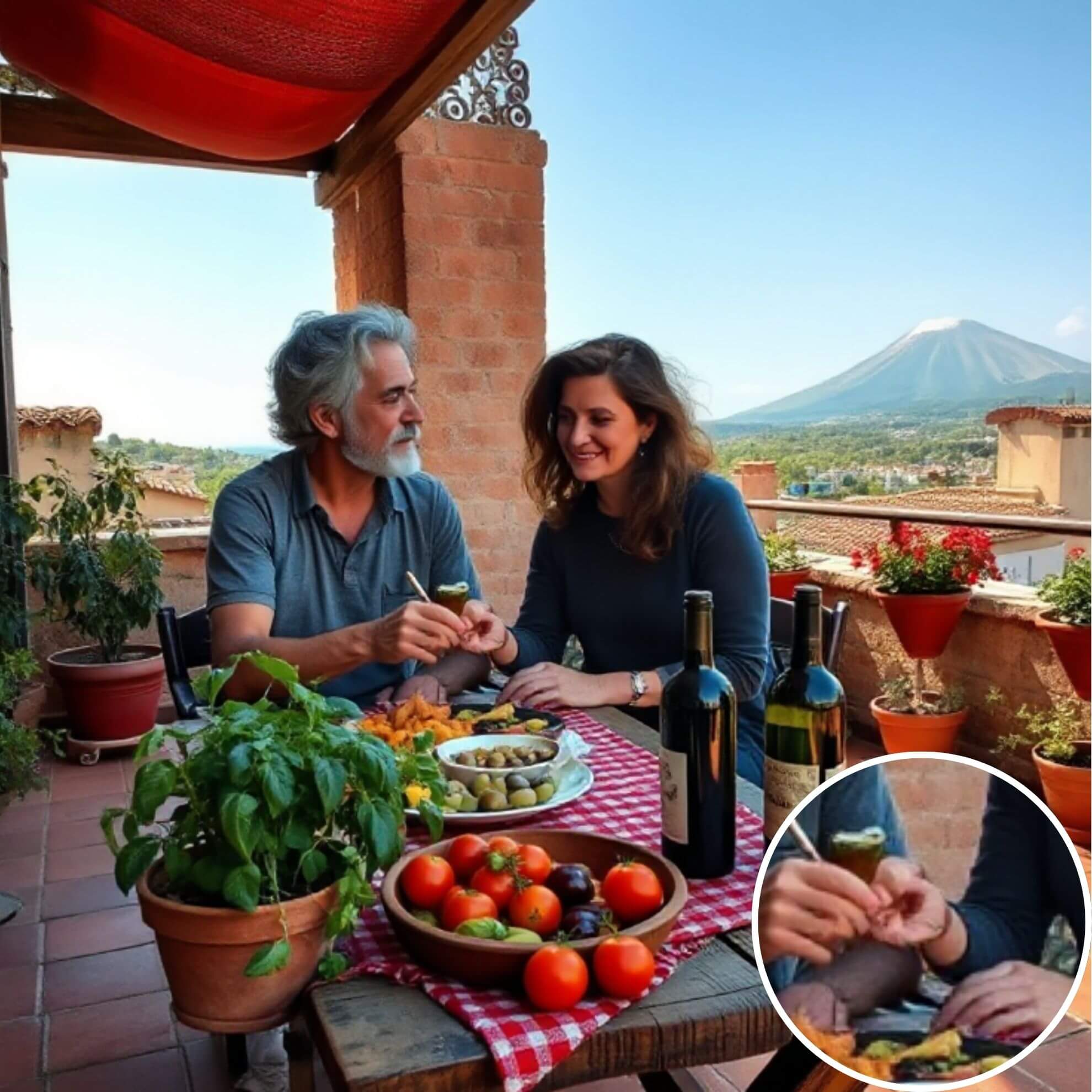} &
\includegraphics[width=0.24\textwidth, height=0.24\textwidth]{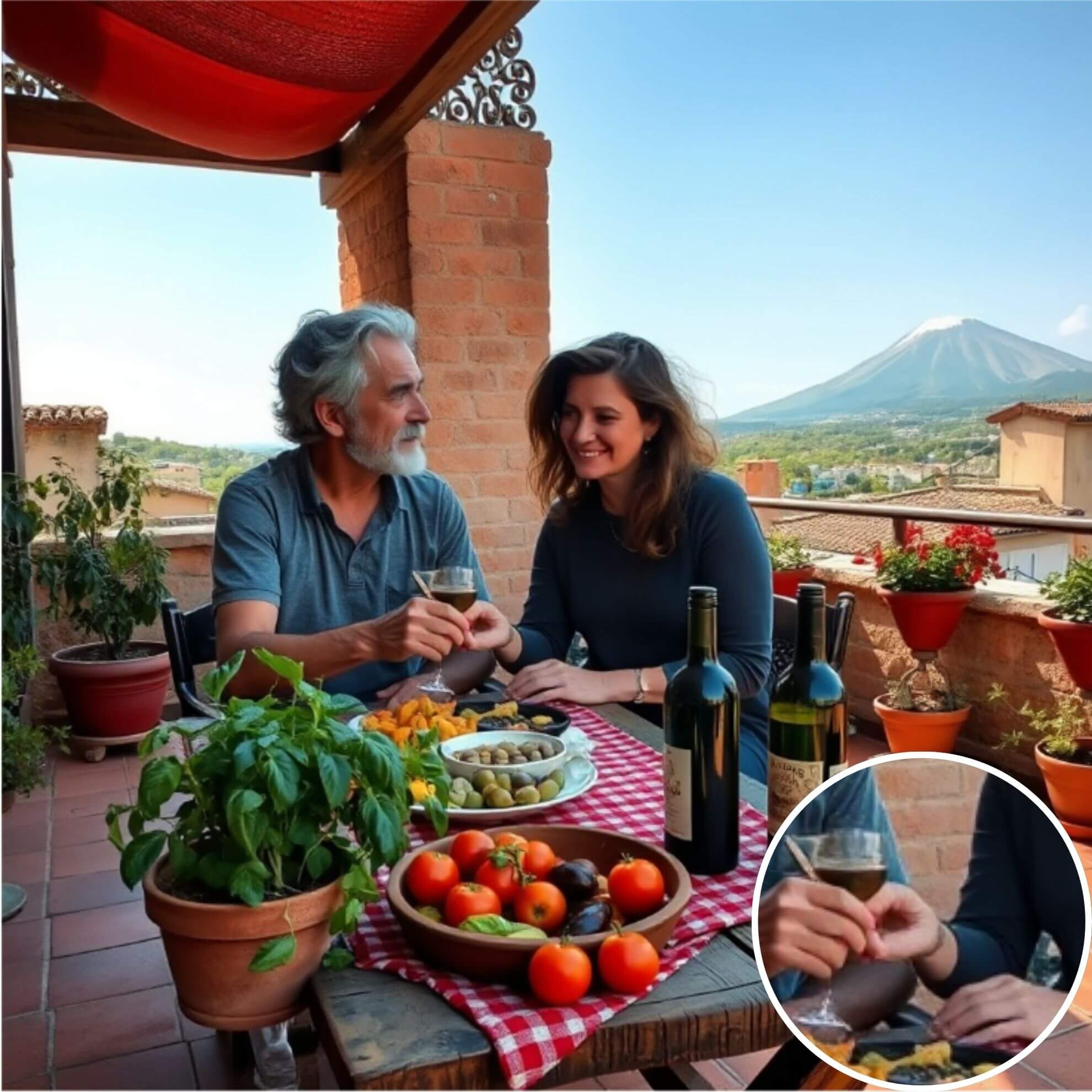} &
\includegraphics[width=0.24\textwidth, height=0.24\textwidth]{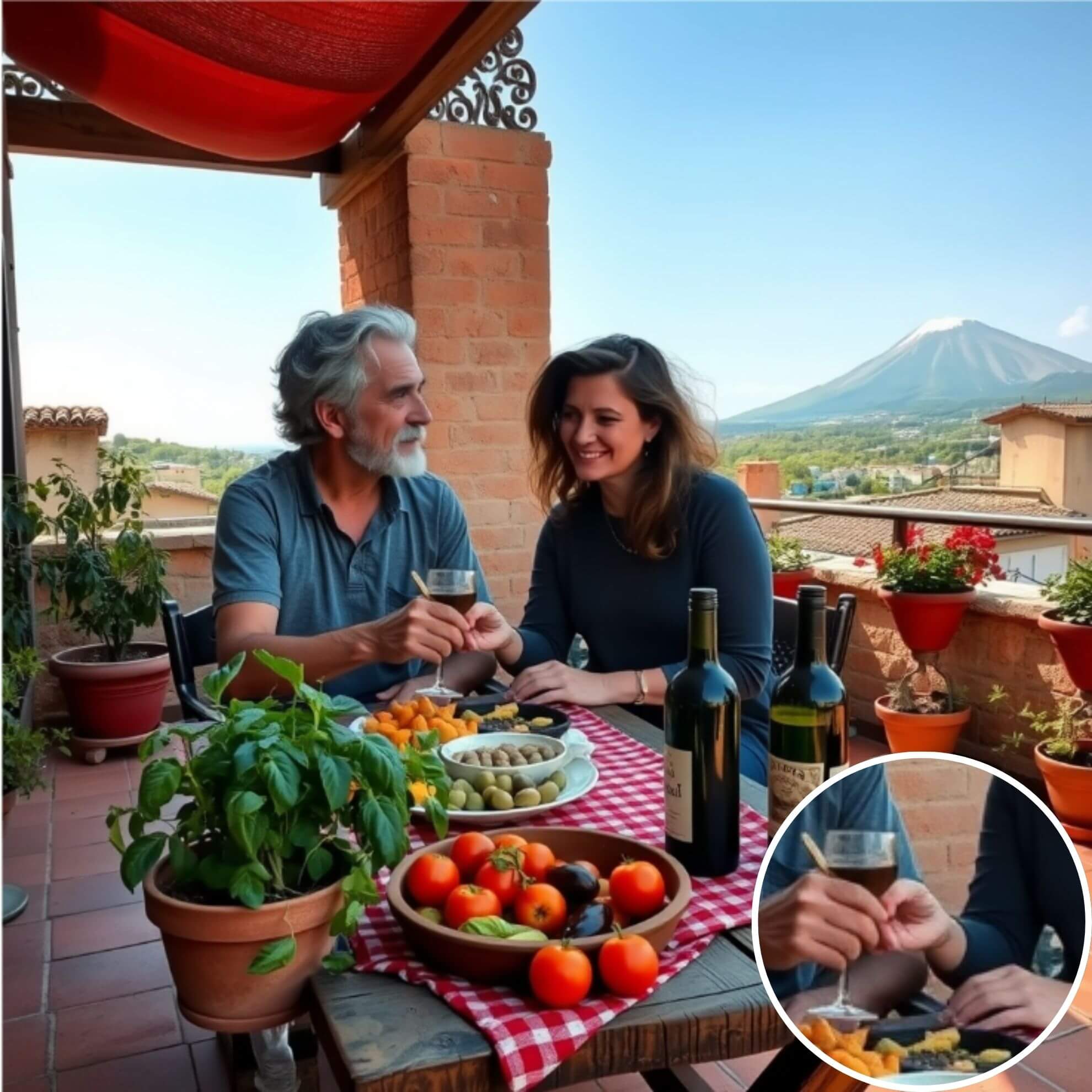} \\
\rotatebox{90}{\hspace{9mm}+\our{}} & 
\includegraphics[width=0.24\textwidth, height=0.24\textwidth]{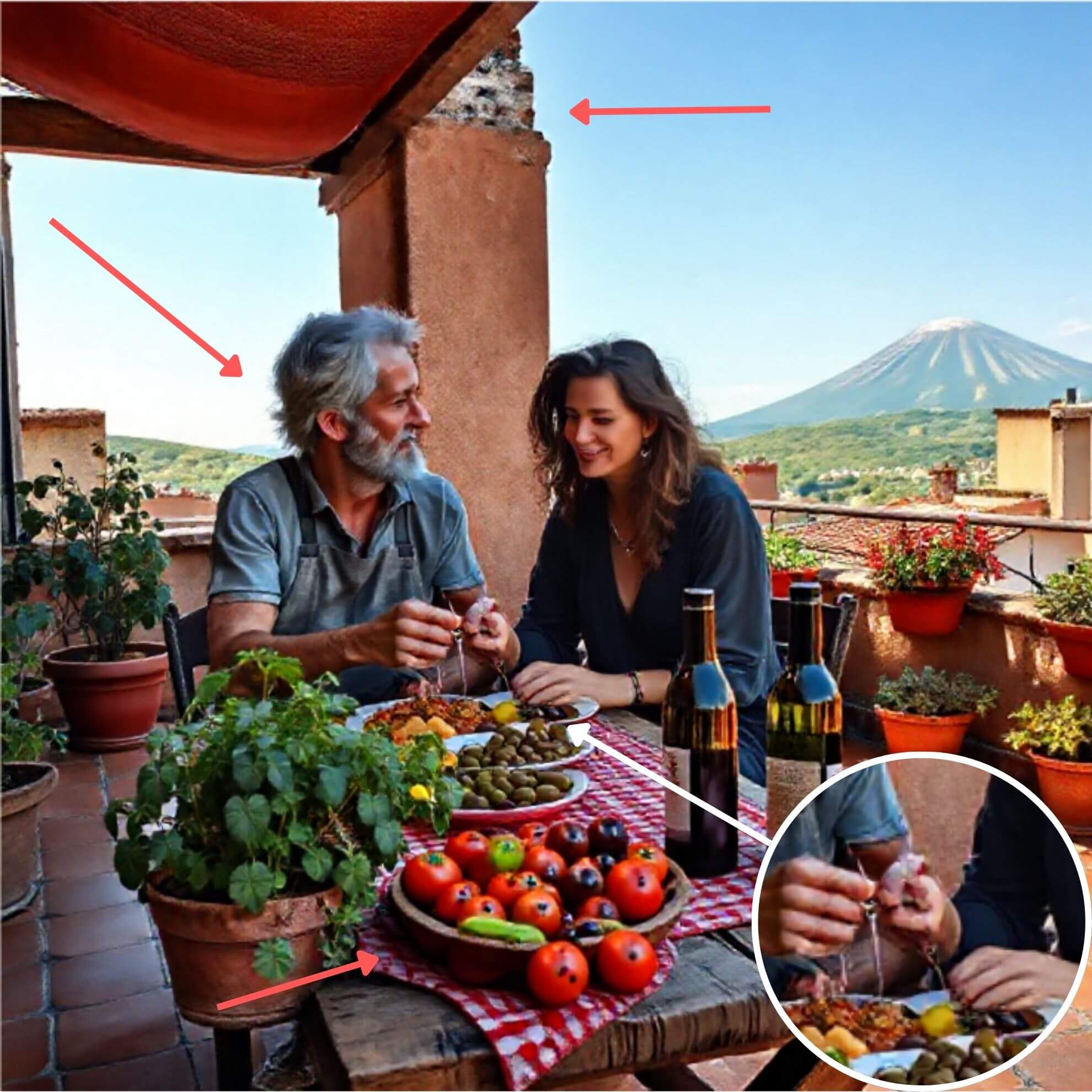} &
\includegraphics[width=0.24\textwidth, height=0.24\textwidth]{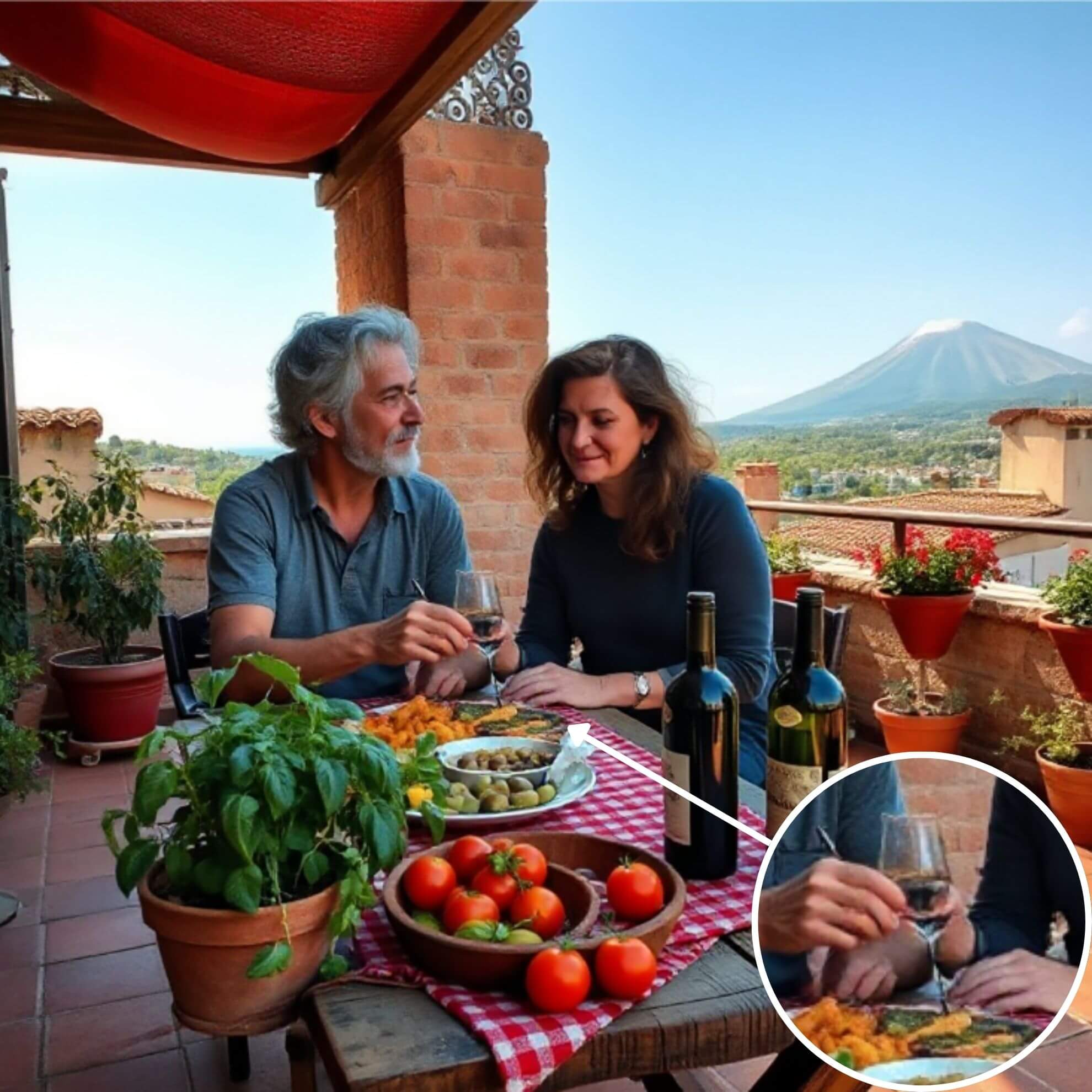} &
\includegraphics[width=0.24\textwidth, height=0.24\textwidth]{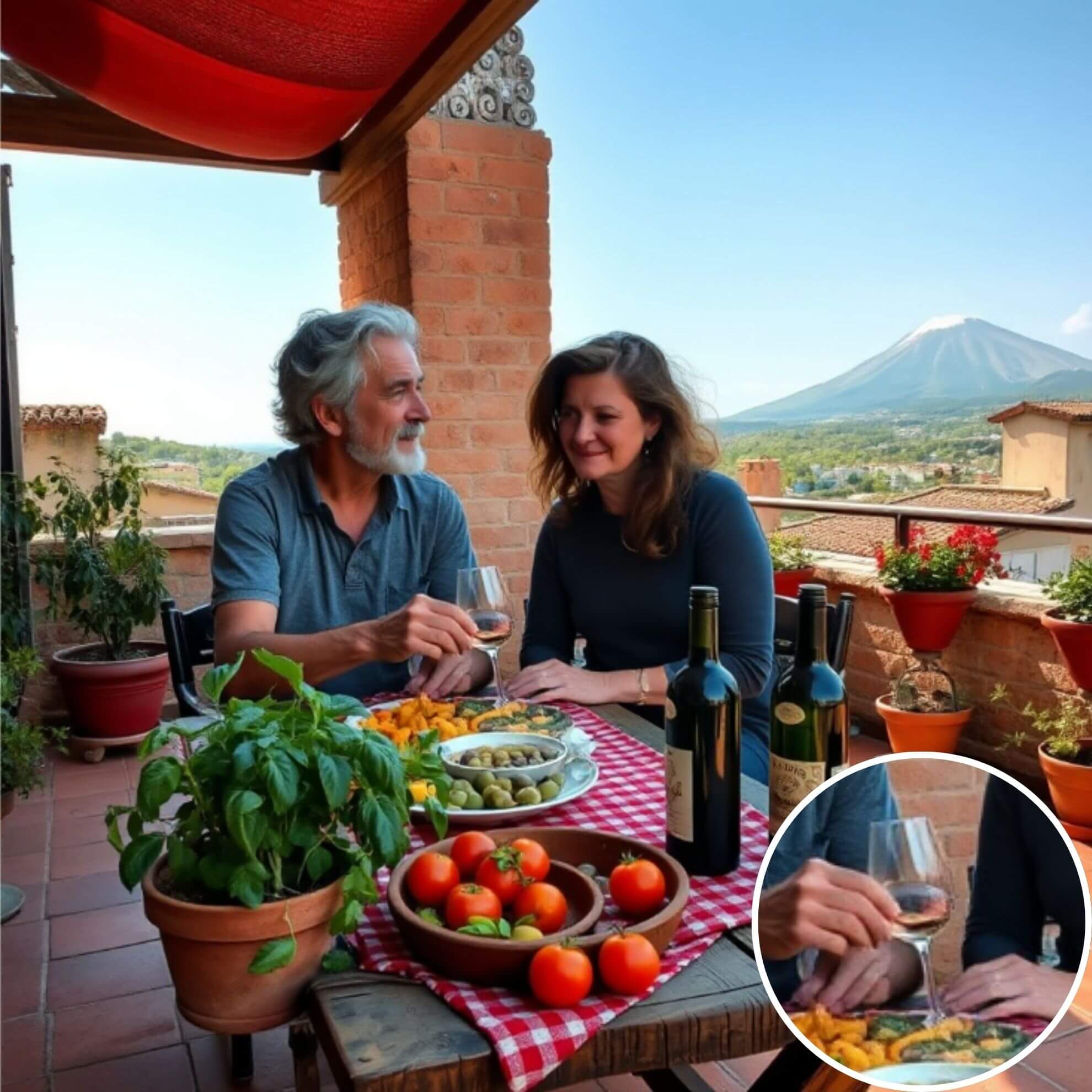} &
\includegraphics[width=0.24\textwidth, height=0.24\textwidth]{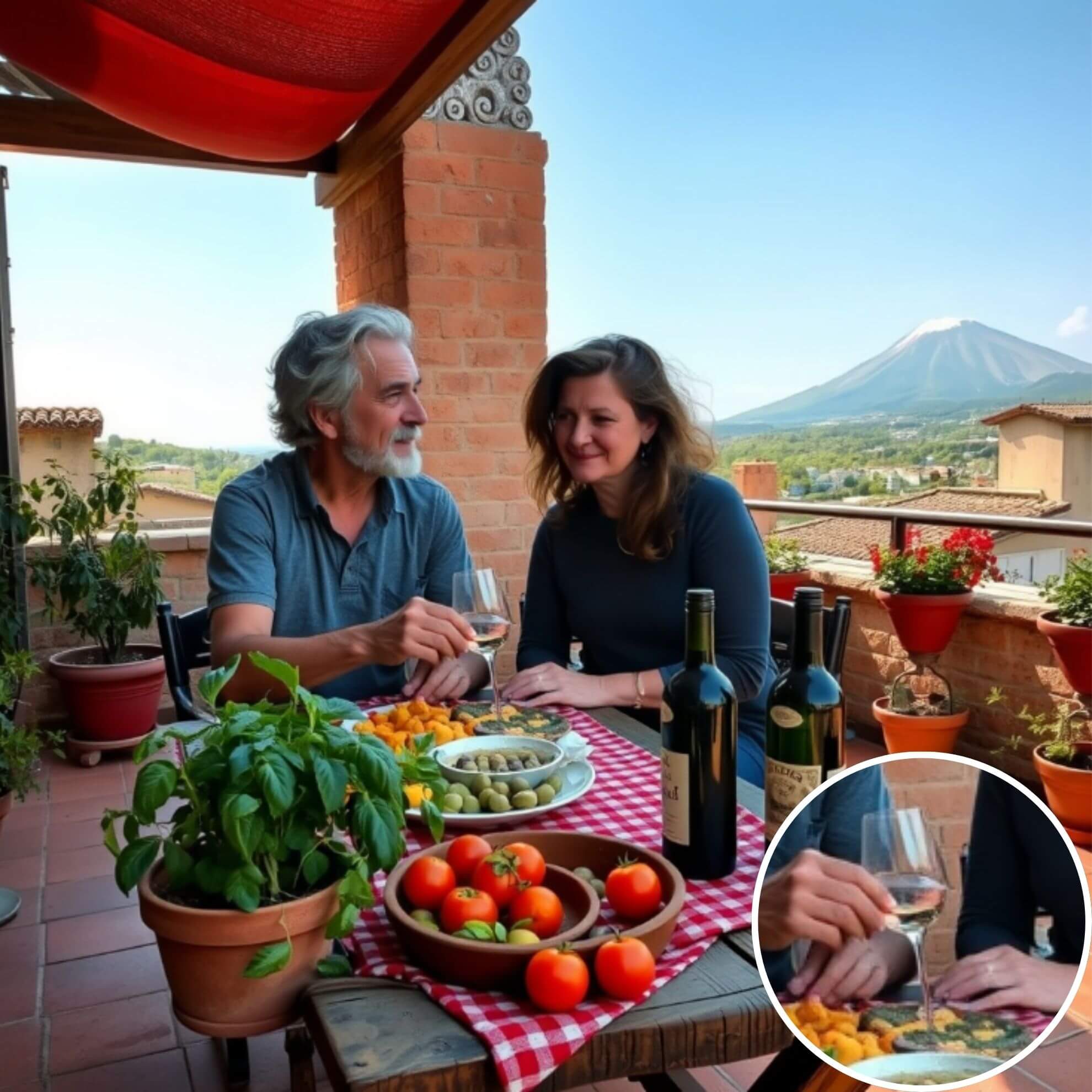} \\ 
\end{tabular}
\caption{\textbf{Trajectory evolution in FLUX.1 [schnell] with \our{}.} The generation direction is already strongly directed after the first step, so major corrections are not possible. We can see that the baseline trajectory in subsequent steps focuses on the texture of the wall, the man, and distorted objects (e.g., tomatoes/vegetables), without changing the position of the people or the background. \our{} allows for the upgrade of smaller parts of the image, in this case, for hands.}
\label{fig:schnell_iteration}
\end{figure}

We also used a fast version of the flow matching model: \textbf{FLUX.1 [schnell]}, which is optimized for often single-step image generation. This leads to lower-quality images with more visual artifacts. Due to the ultra-fast inference time, we perform additional analyses on the general {\textit{people }dataset} to analyze trajectory improvement in the forward pass. We assumed inference steps = 4 \cite{wang2025diffdoctor}. \our{} enables only limited image corrections (due to the shorter trajectory), substantially smaller than those performed by multi-step models (e.g, dev), see Fig. \ref{fig:schnell_iteration}. The hyperparameters used to generate the images in Figs. \ref{fig:dev_text}, \ref{fig:dev_text2}, \ref{fig:schnell_iteration} are provided in Tab. \ref{tab:hyperparameters}.

\textbf{Stable Diffusion XL (SDXL)} was included in the evaluation, exclusively for the \textit{people} dataset. This model exhibits greater difficulty in generating complex scenes and texts than previous models. Ultimately, the observed artifacts arise primarily from inherent limitations of the model, rather than from the diffusion trajectory. Therefore, we focus only on evaluating the diffusion model for the {\textit{people} dataset}, as texts and animals pose a greater challenge for this model. The inference steps were assumed to be 30.

\newpage

\section{Additional results}\label{sec:add_results}
This section describes the additional results for the datasets \textit{people}, \textit{words}, and \textit{animals}, respectively.

\begin{figure*}[h]
\centering
\setlength{\tabcolsep}{1.5pt}
\renewcommand{\arraystretch}{0.9}
\begin{tabular}{cc}
FLUX.1 [dev]\hspace{80pt}+\our{} & FLUX.1 [dev]\hspace{80pt}+\our{}  \\
\includegraphics[width=0.49\textwidth]{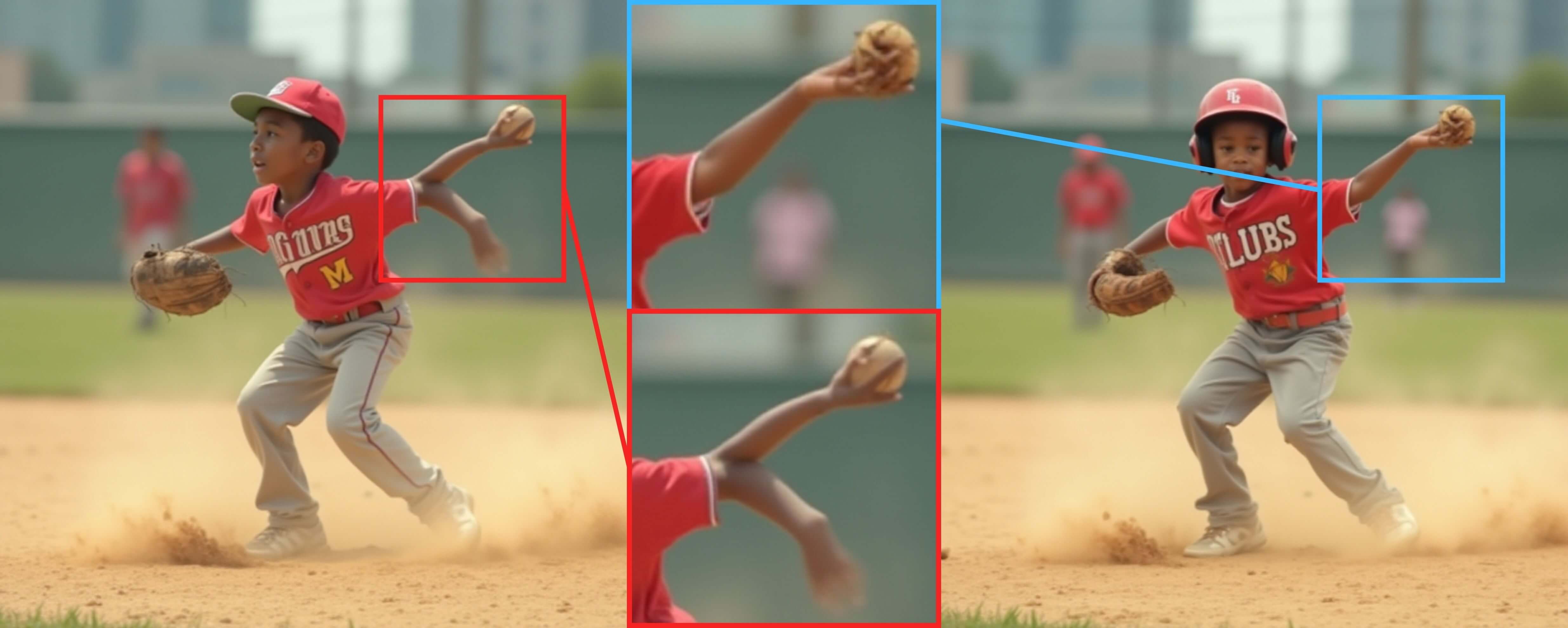} &
\includegraphics[width=0.49\textwidth]{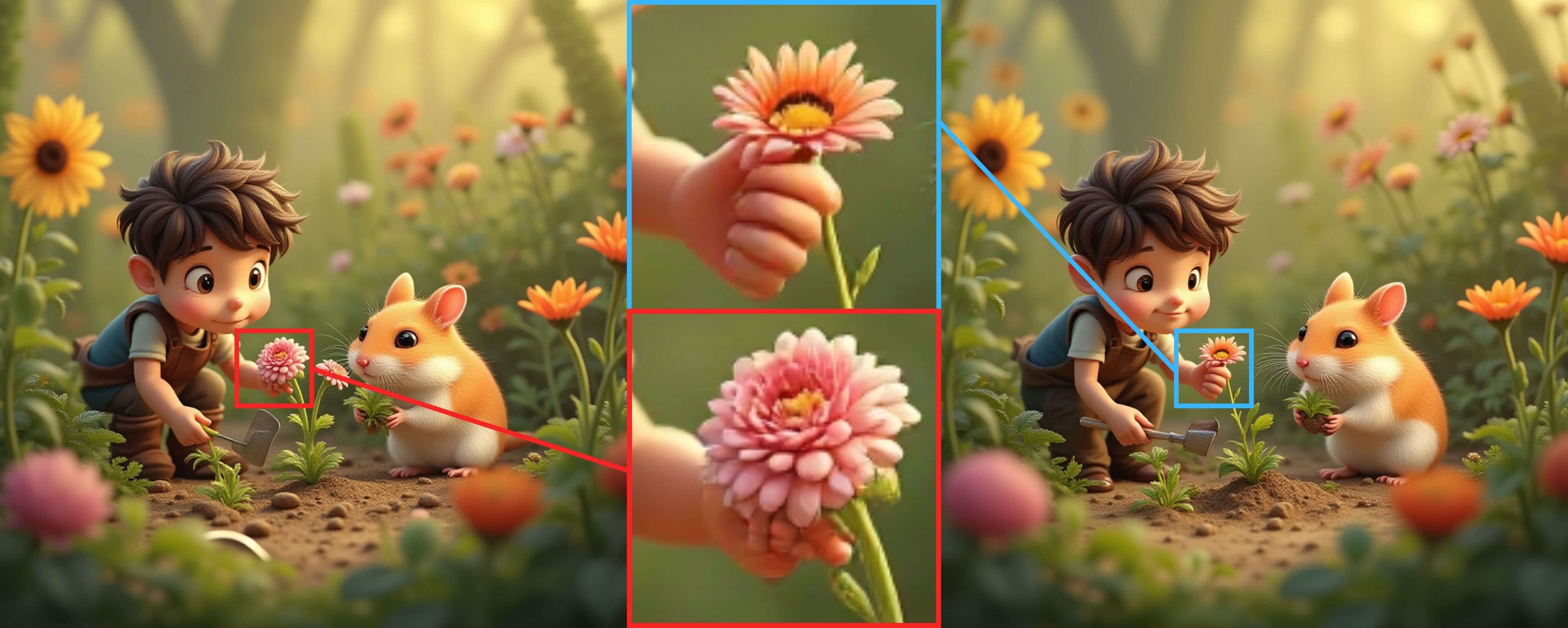} \\

\includegraphics[width=0.49\textwidth]{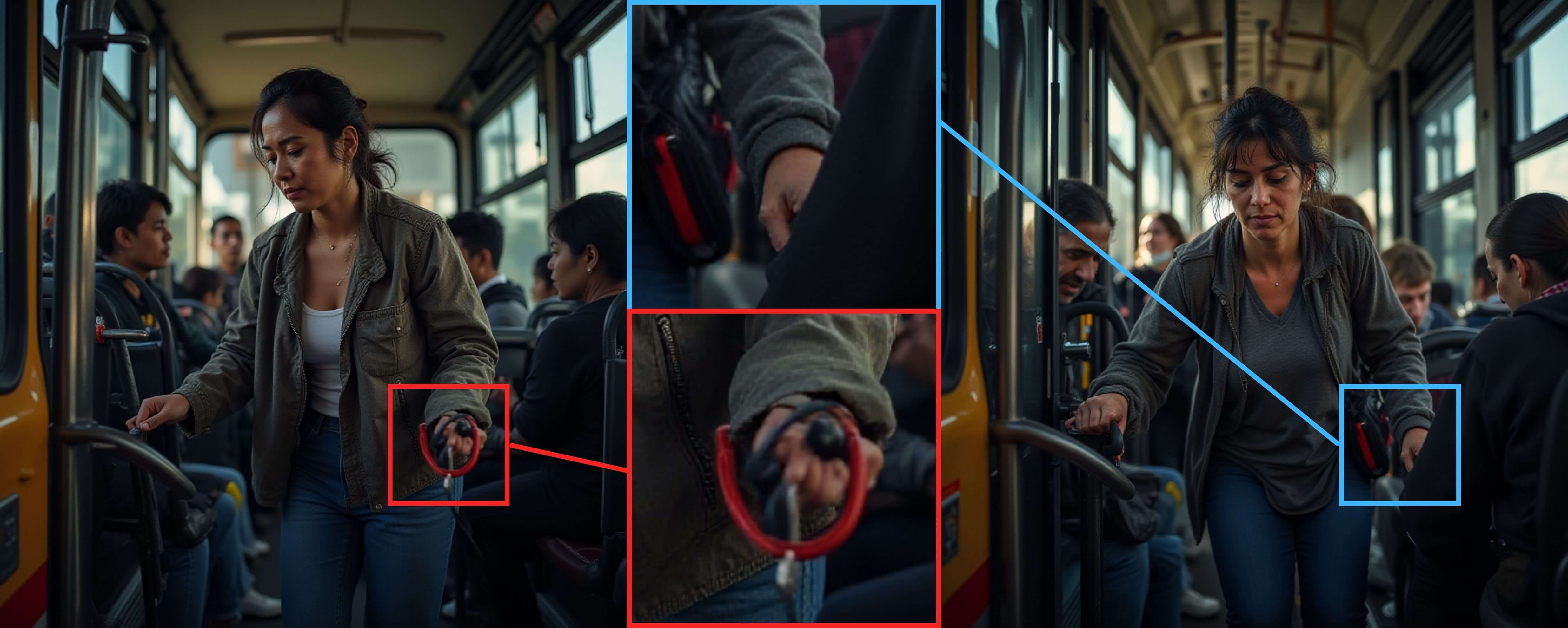} &
\includegraphics[width=0.49\textwidth]{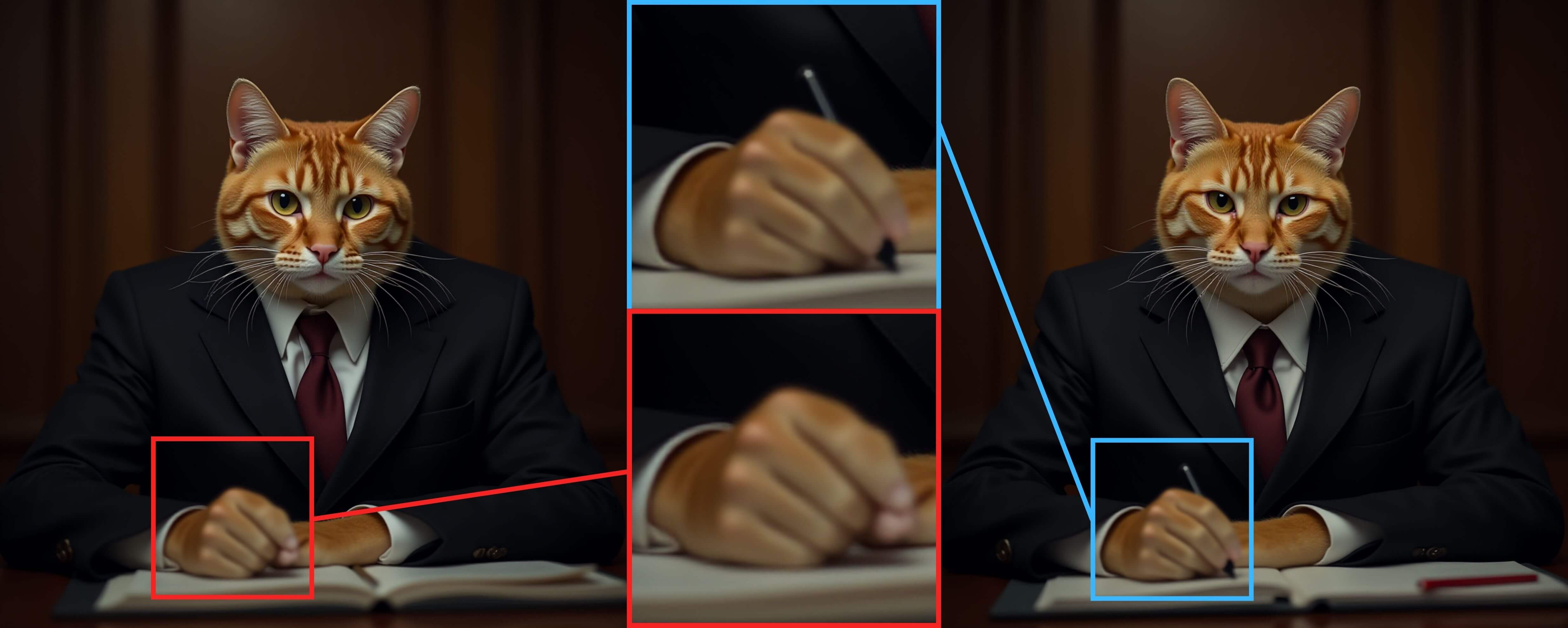} \\

\includegraphics[width=0.49\textwidth]{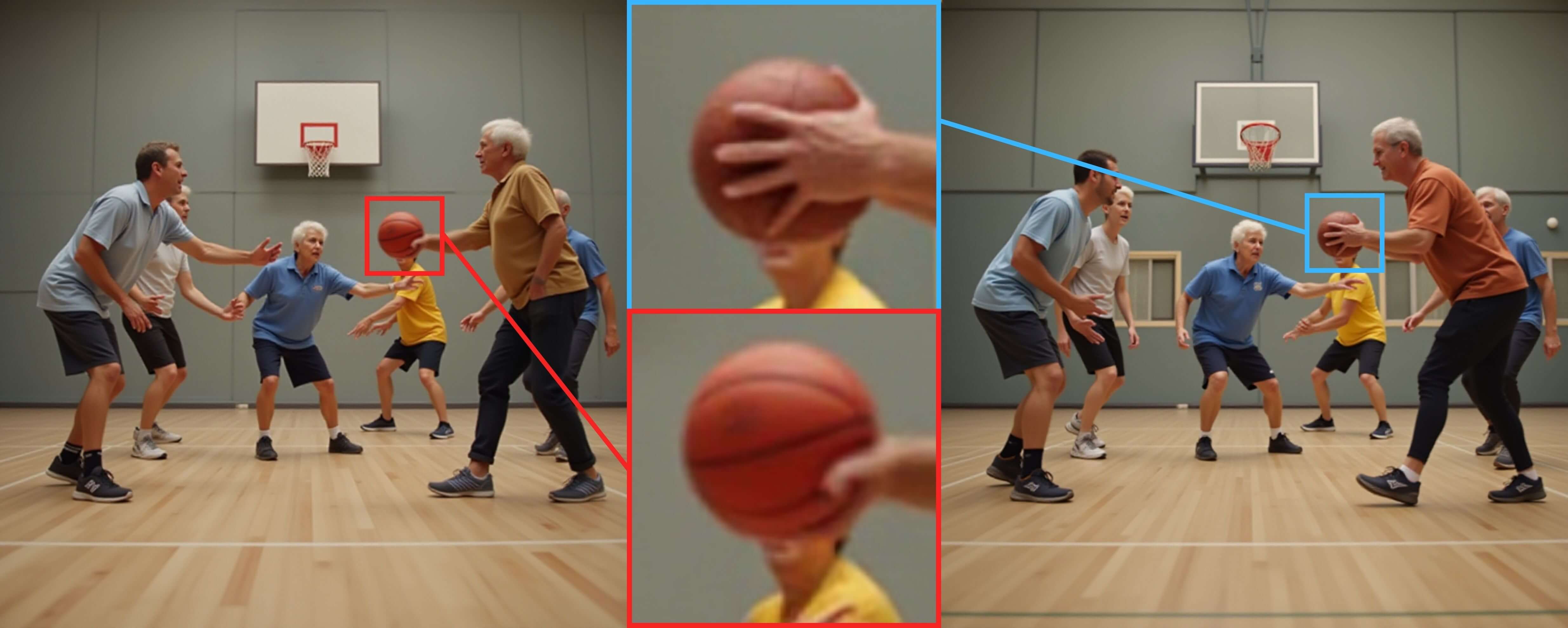} &
\includegraphics[width=0.49\textwidth]{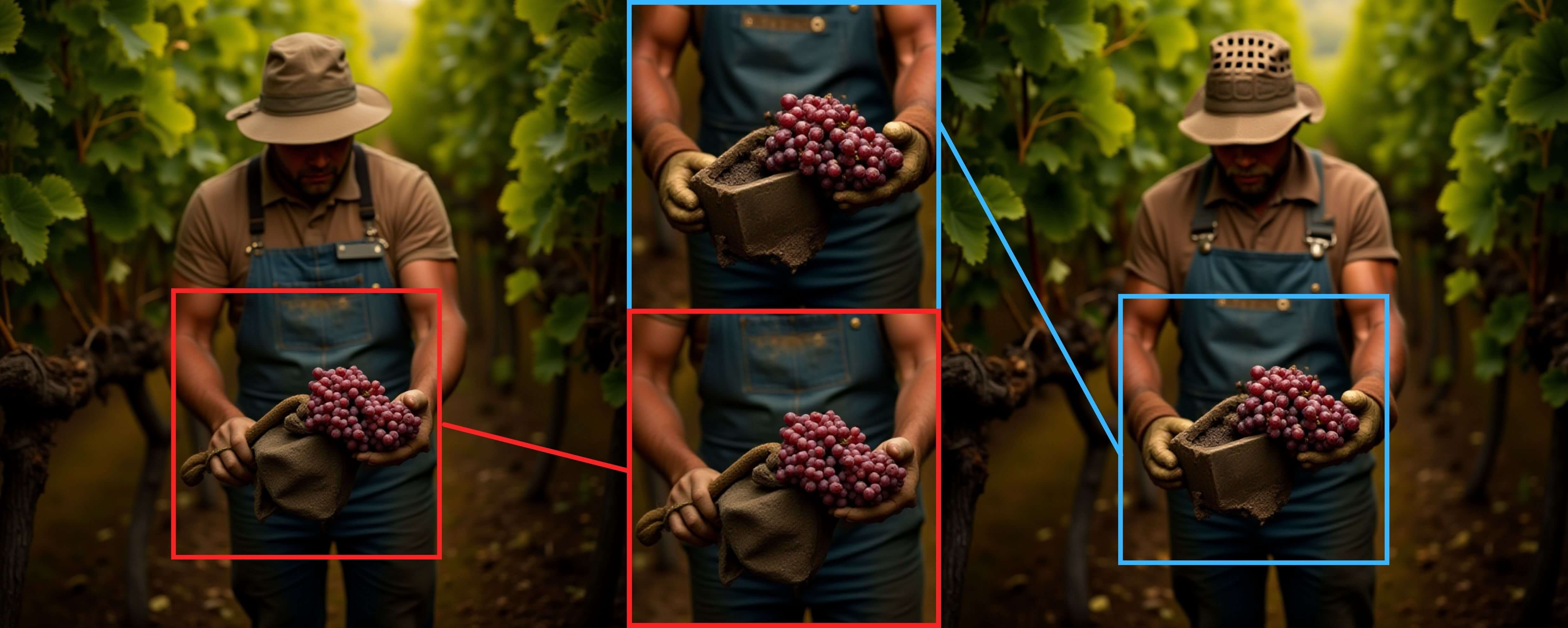} \\

\includegraphics[width=0.49\textwidth]{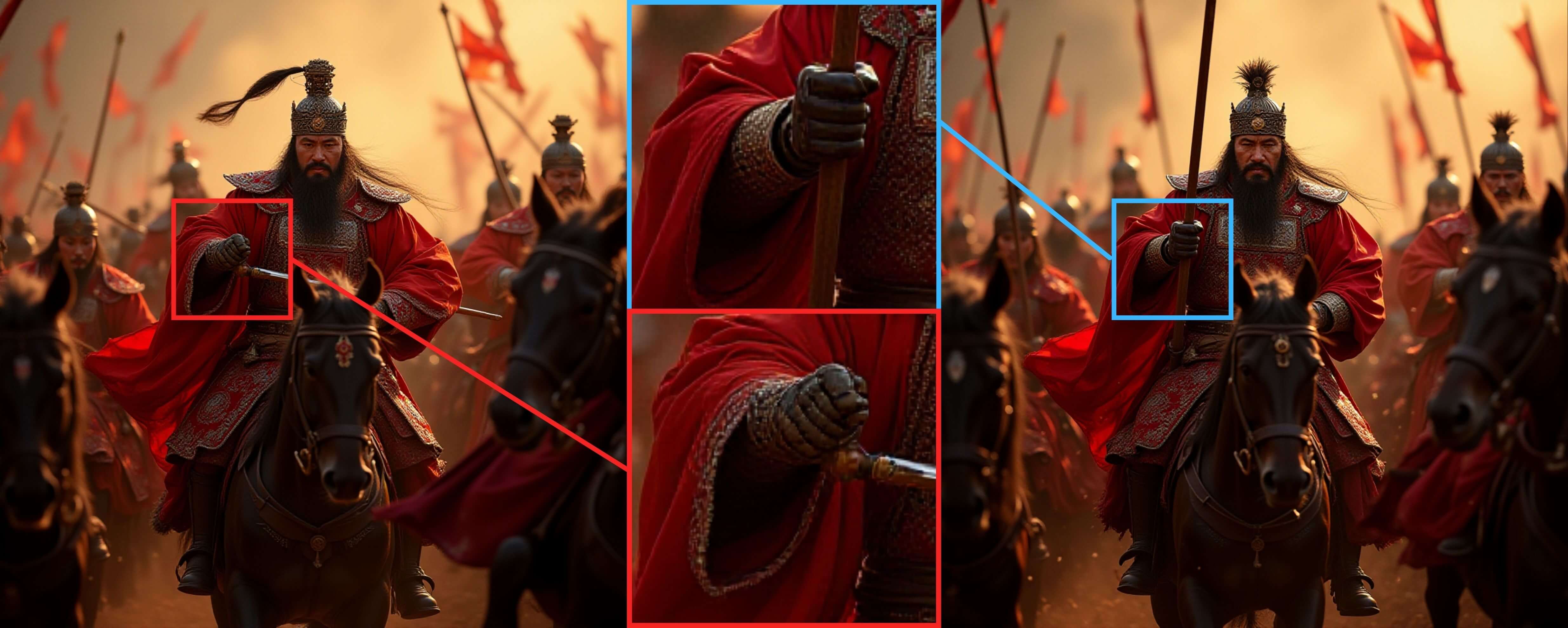} &
\includegraphics[width=0.49\textwidth]{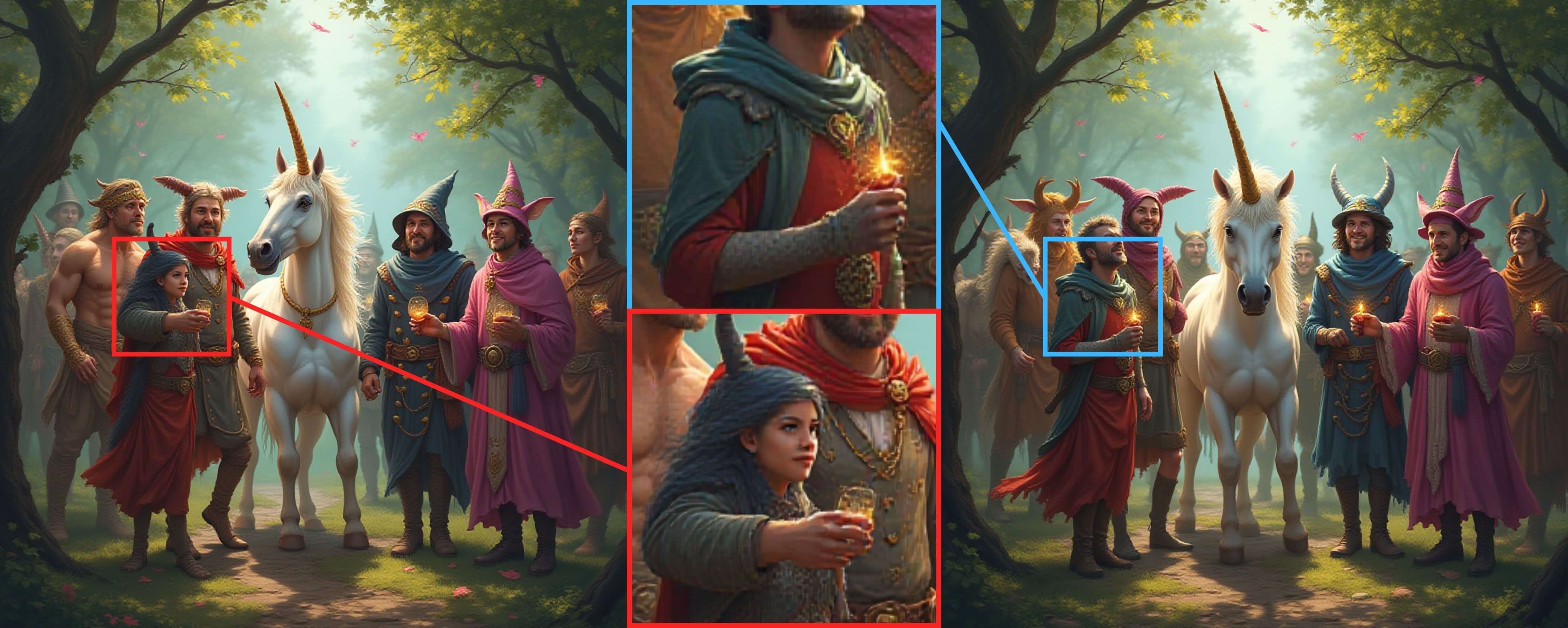}

\end{tabular}
\caption{\textbf{Images for the FLUX.1 [dev] on the \textit{people} dataset using \our{} for 10 inference steps.} In this type of flow matching model, we can direct the trajectory by correcting large artifacts, such as improving the body shape (see the image with the wizards), and smaller ones (correcting the hand holding the pen or the basketball).}
\label{fig_flux_dev_people}
\end{figure*}

\begin{figure*}[!t]
\centering
\setlength{\tabcolsep}{1.2pt}
\renewcommand{\arraystretch}{0.9}
\begin{tabular}{ccc}
SDXL & +HandsXL & +\our{} \\
 \includegraphics[width=0.19\textwidth]{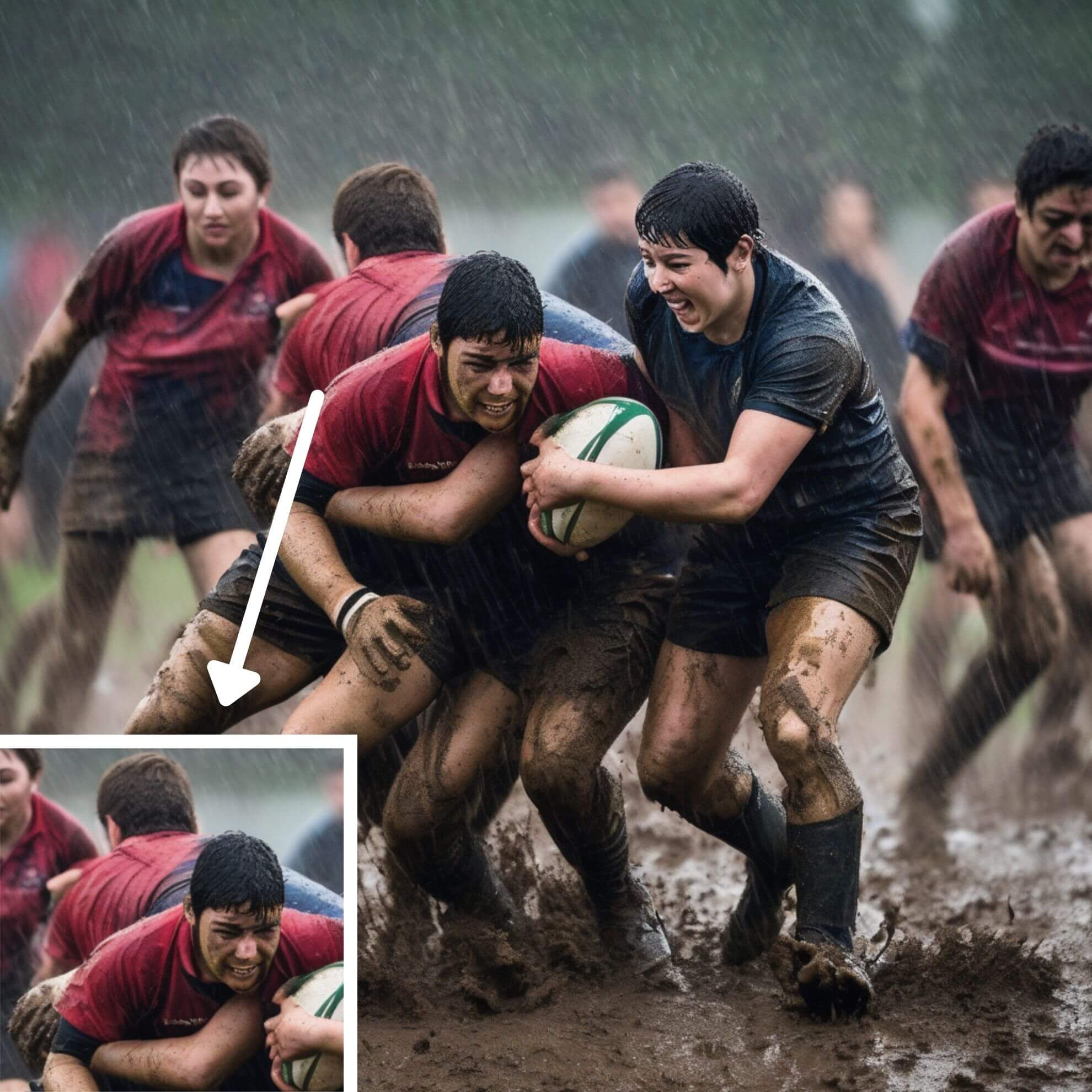} &
\includegraphics[width=0.19\textwidth]{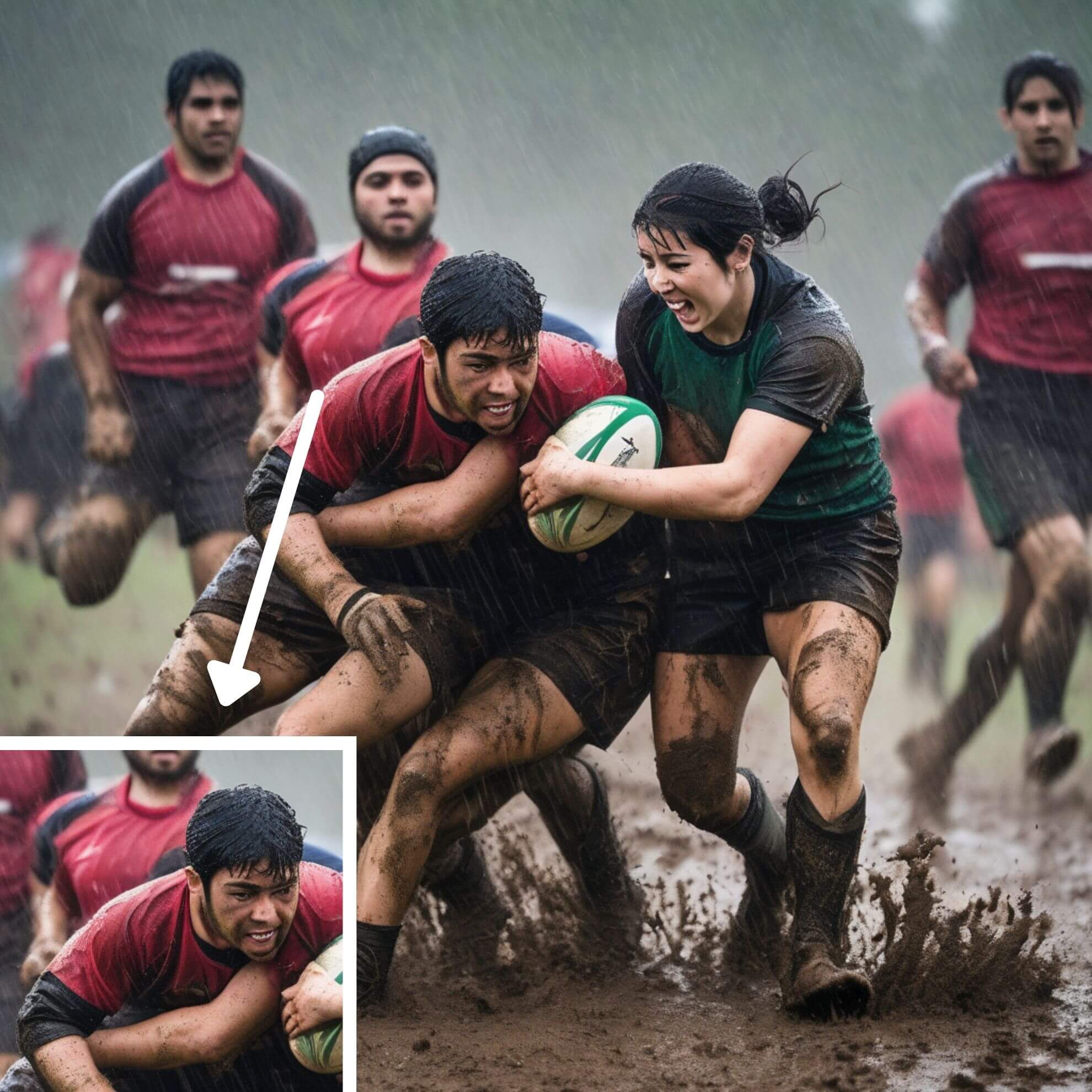} &
\includegraphics[width=0.19\textwidth]{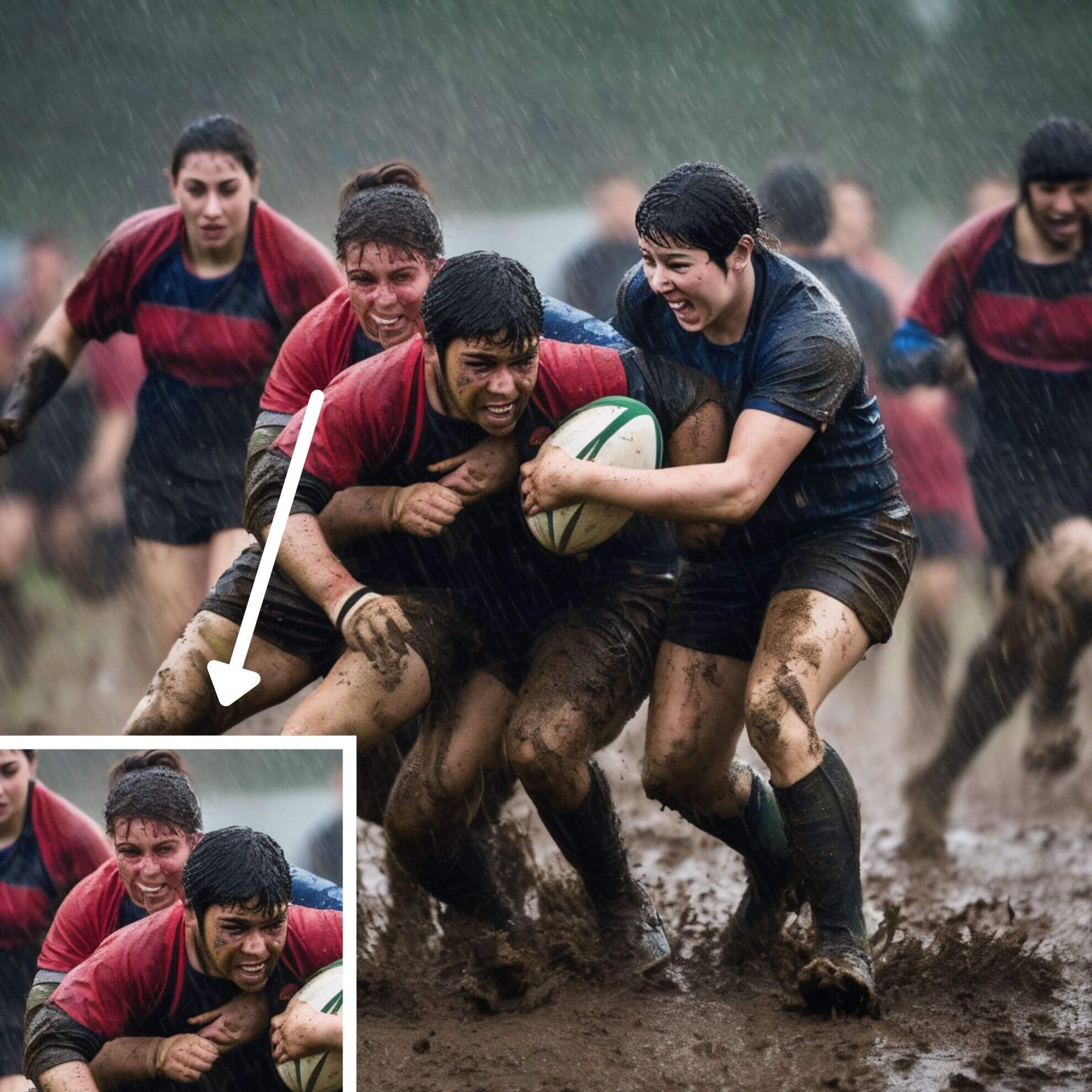} \\
 \includegraphics[width=0.19\textwidth]{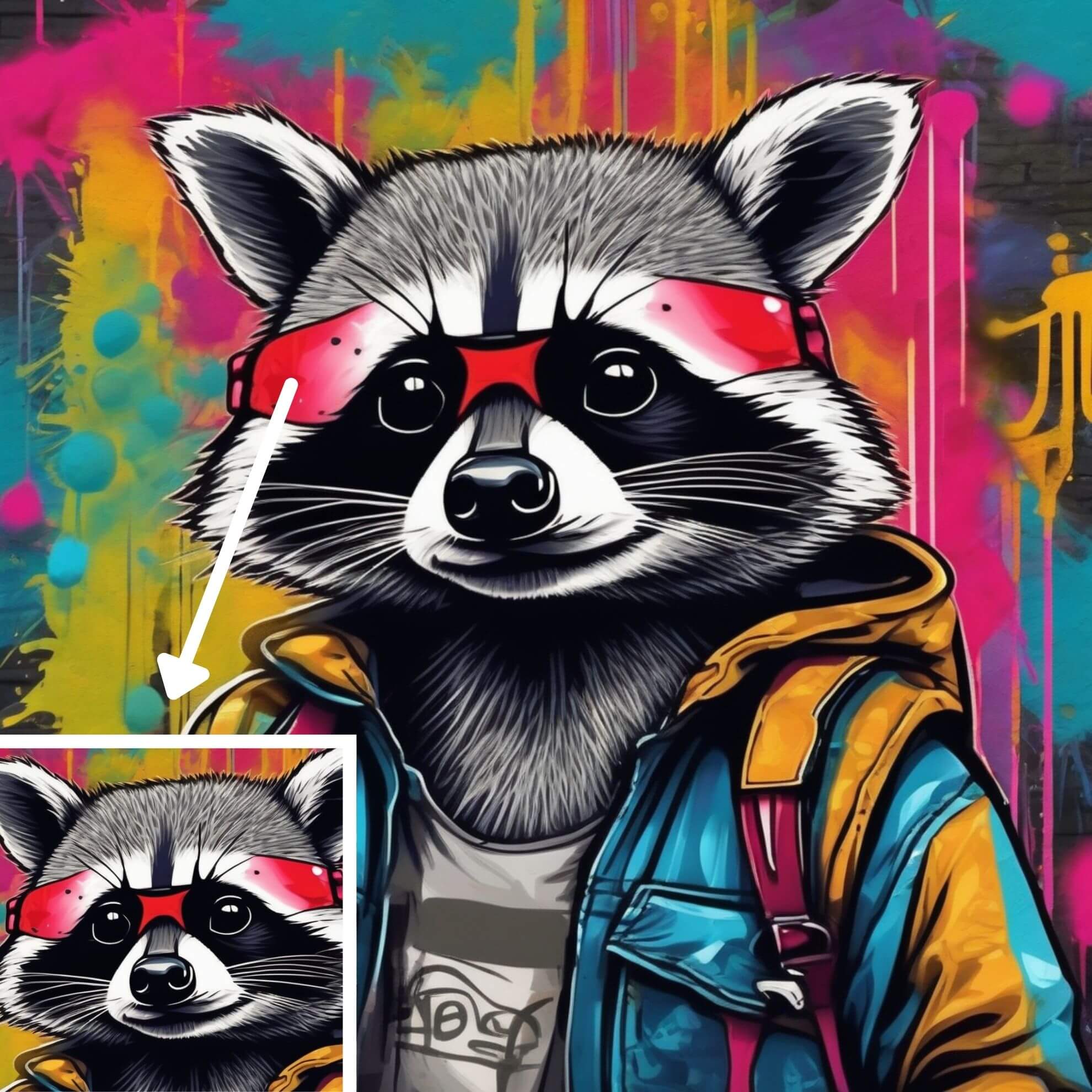} &
\includegraphics[width=0.19\textwidth]{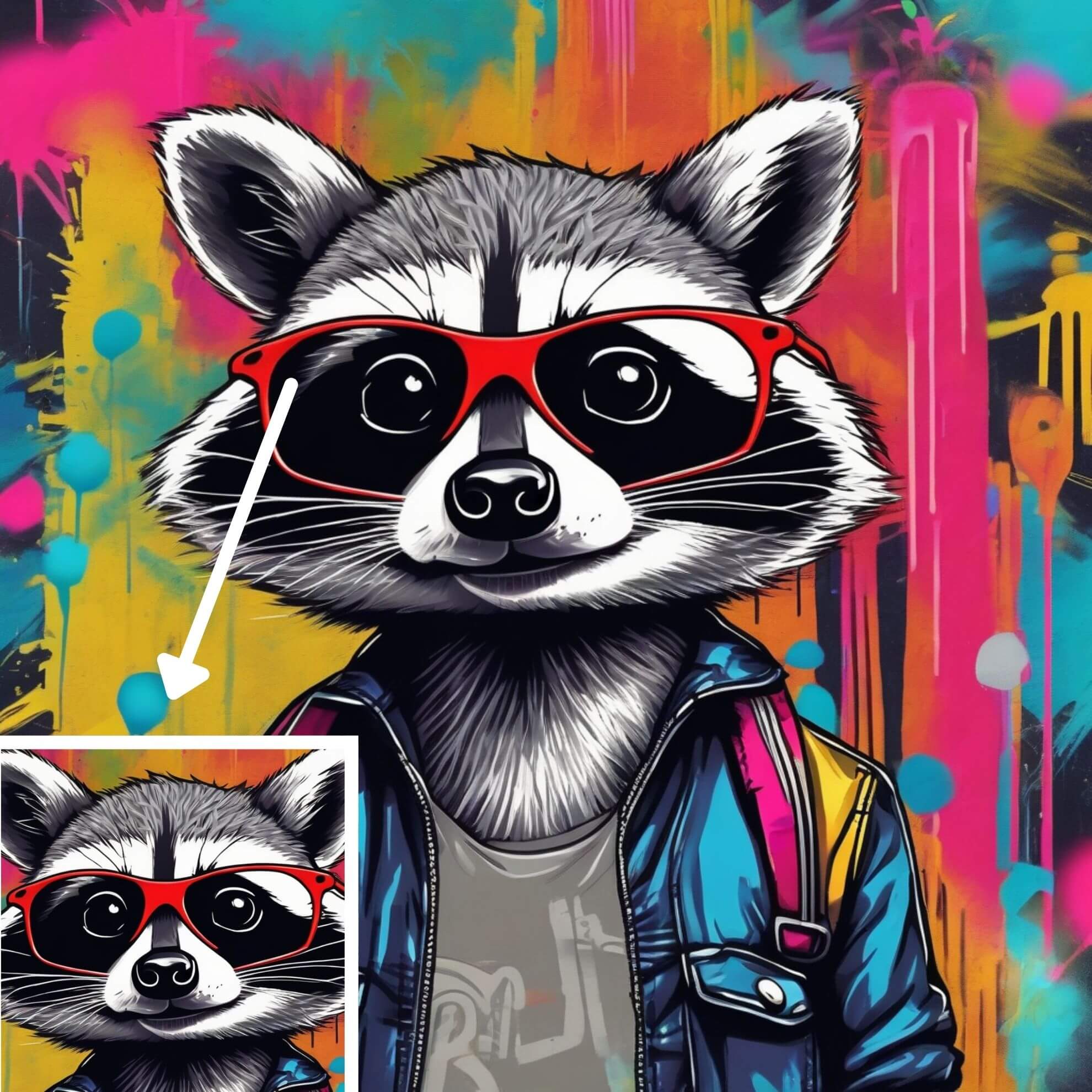} &
\includegraphics[width=0.19\textwidth]{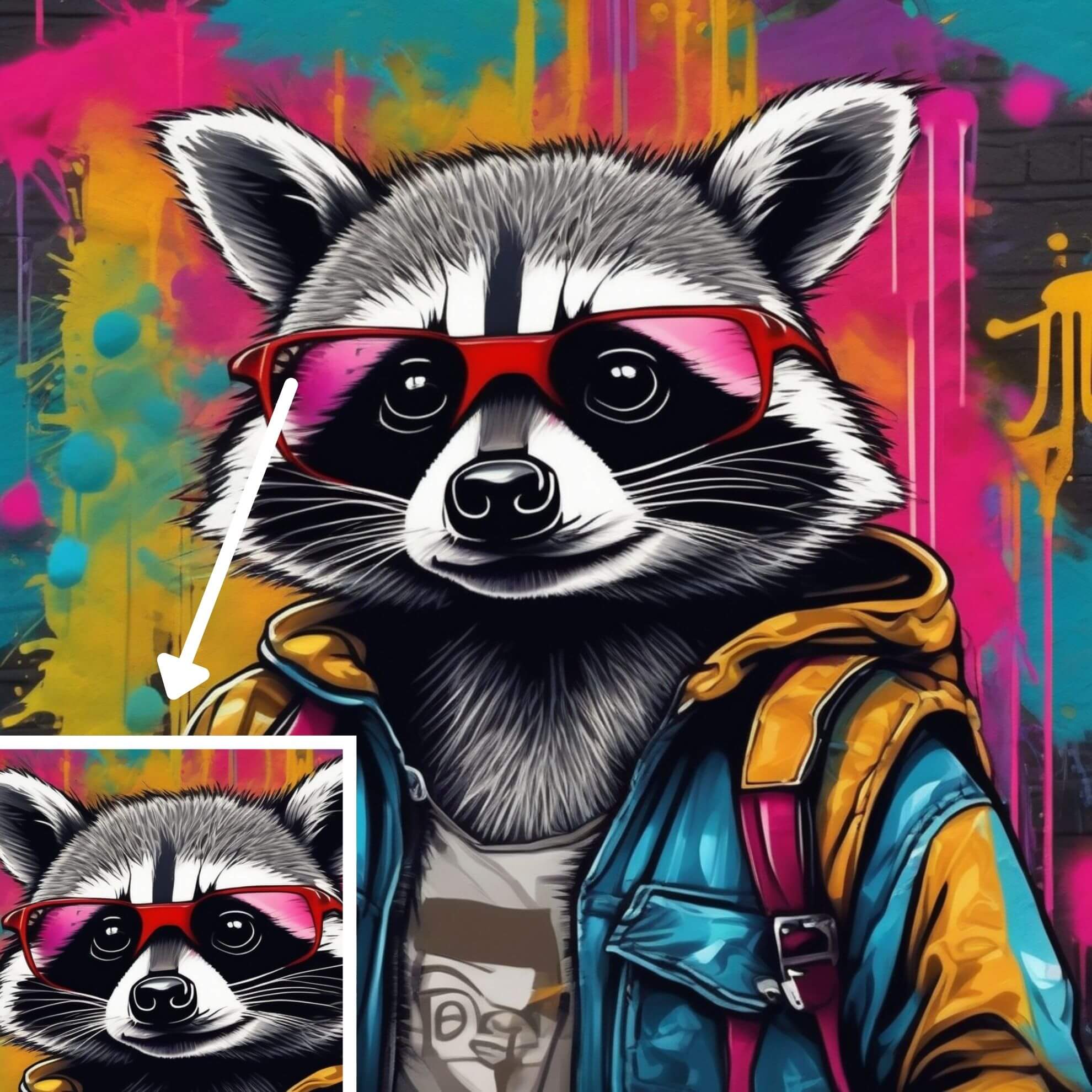} \\
 \includegraphics[width=0.19\textwidth]{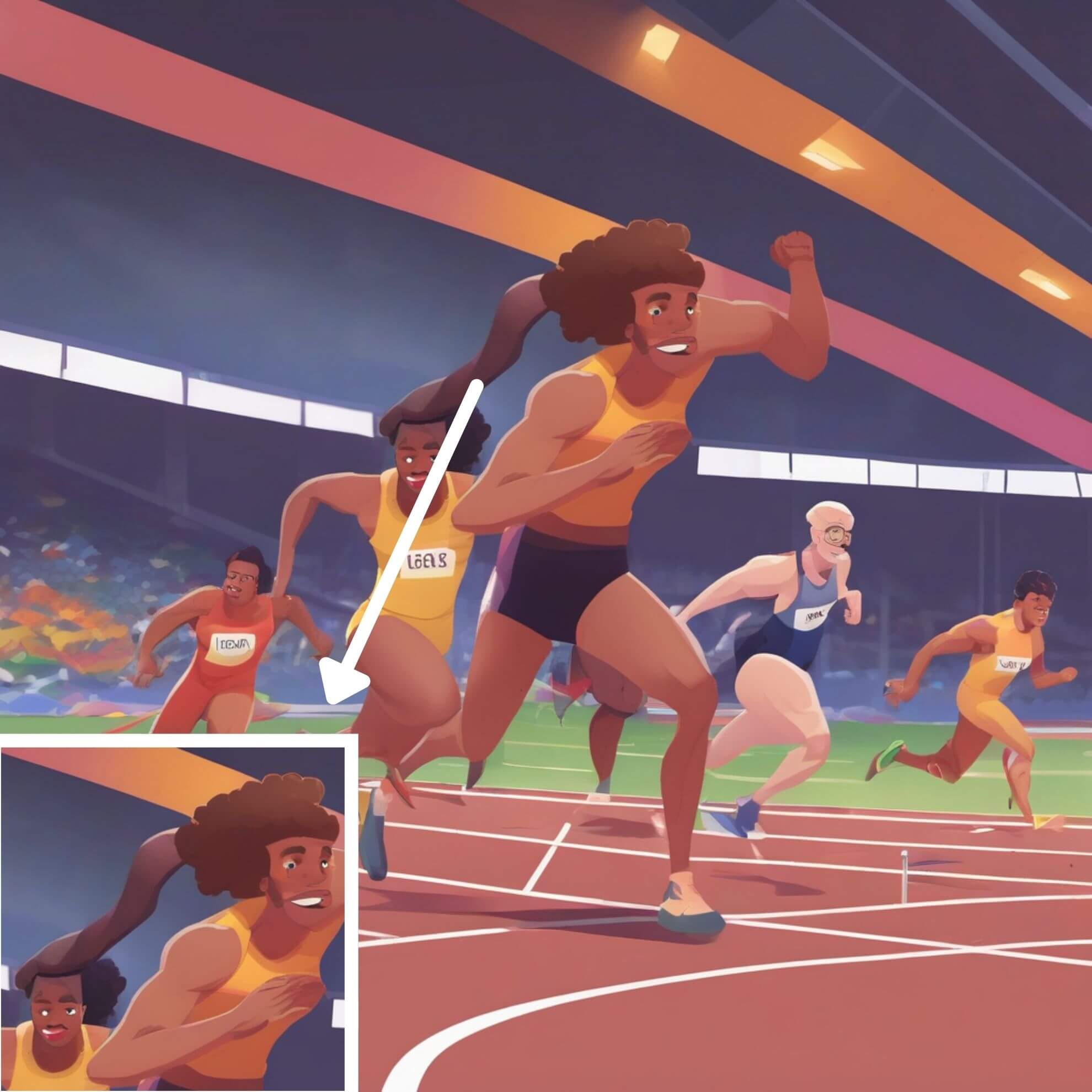} &
\includegraphics[width=0.19\textwidth]{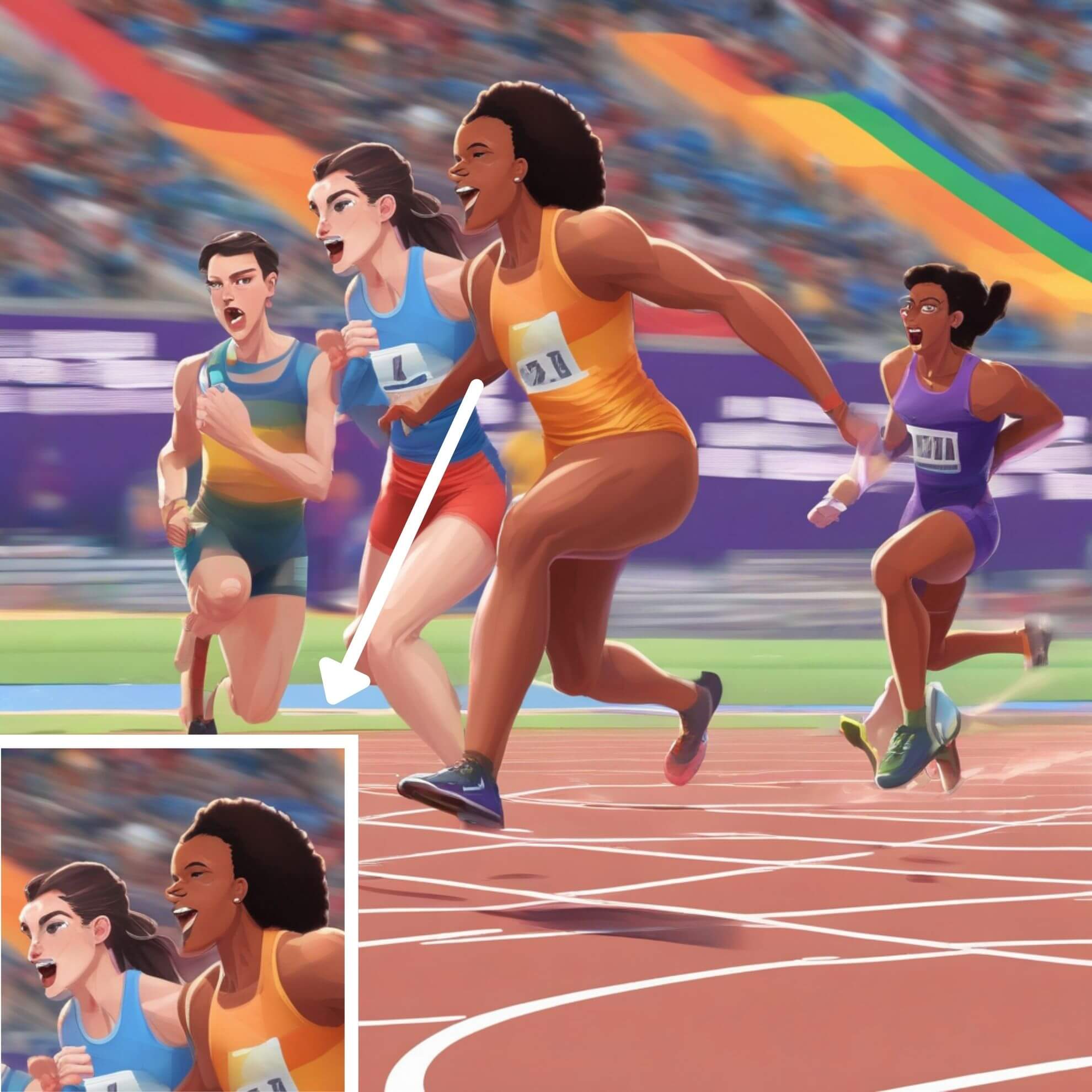} &
\includegraphics[width=0.19\textwidth]{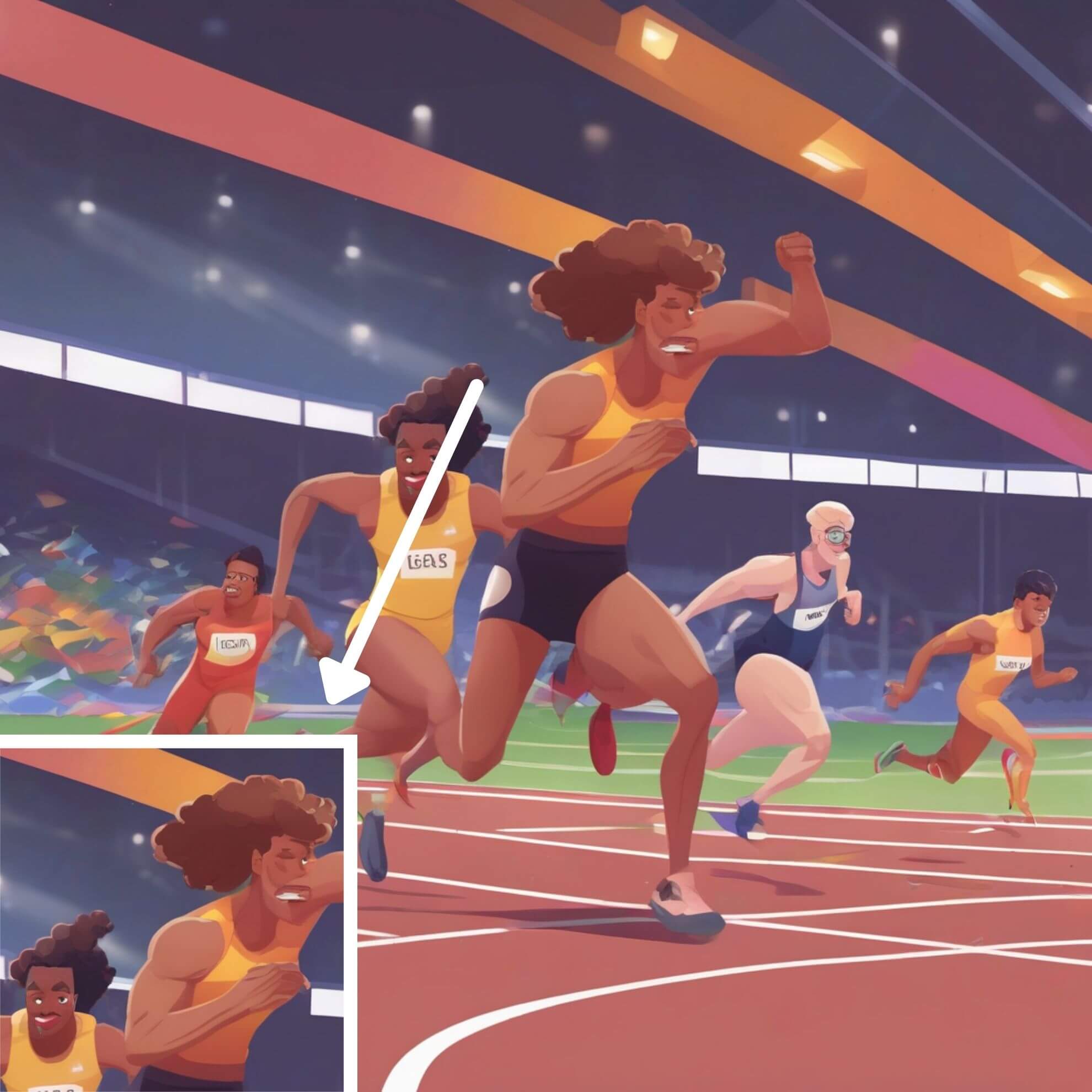} \\
 \includegraphics[width=0.19\textwidth]{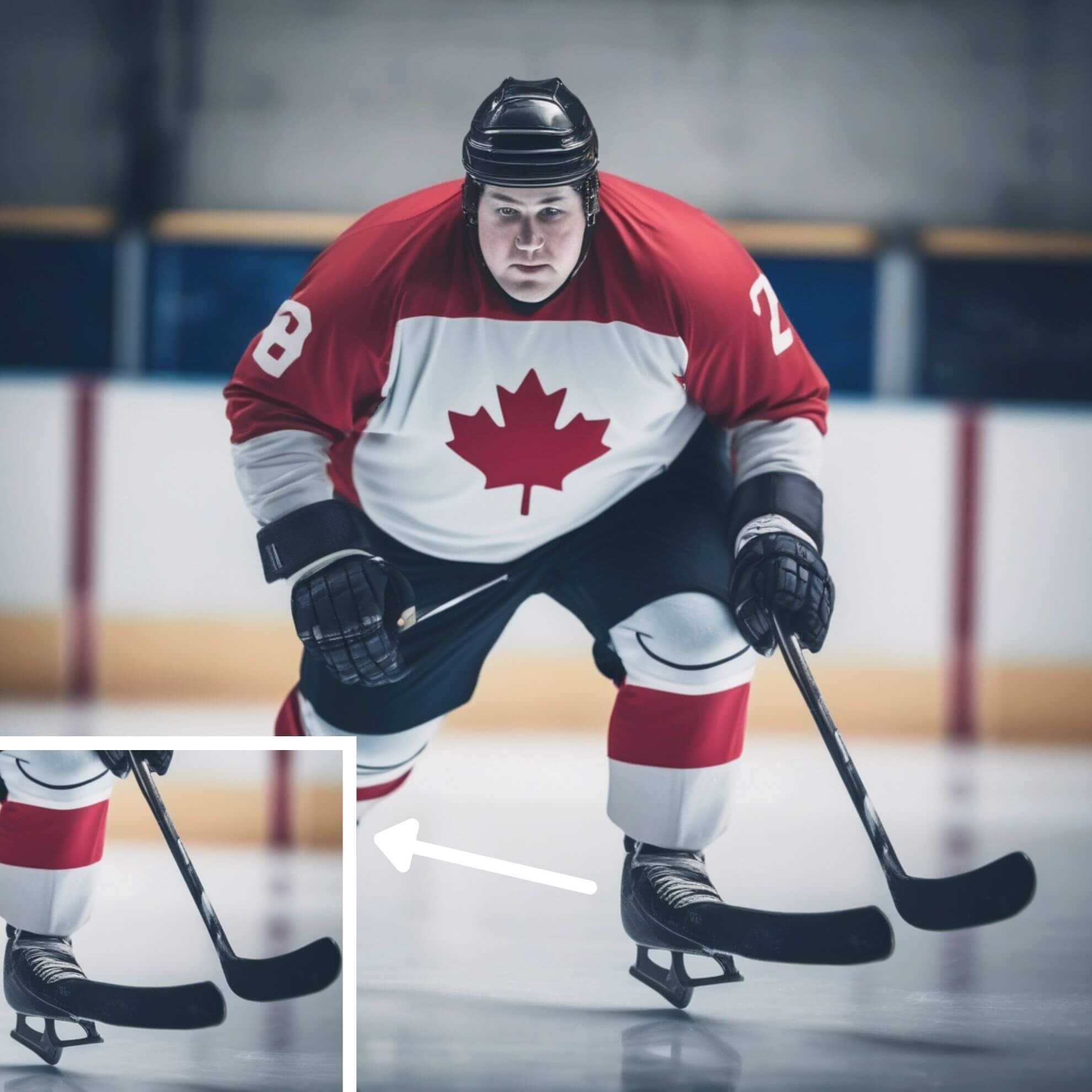} &
\includegraphics[width=0.19\textwidth]{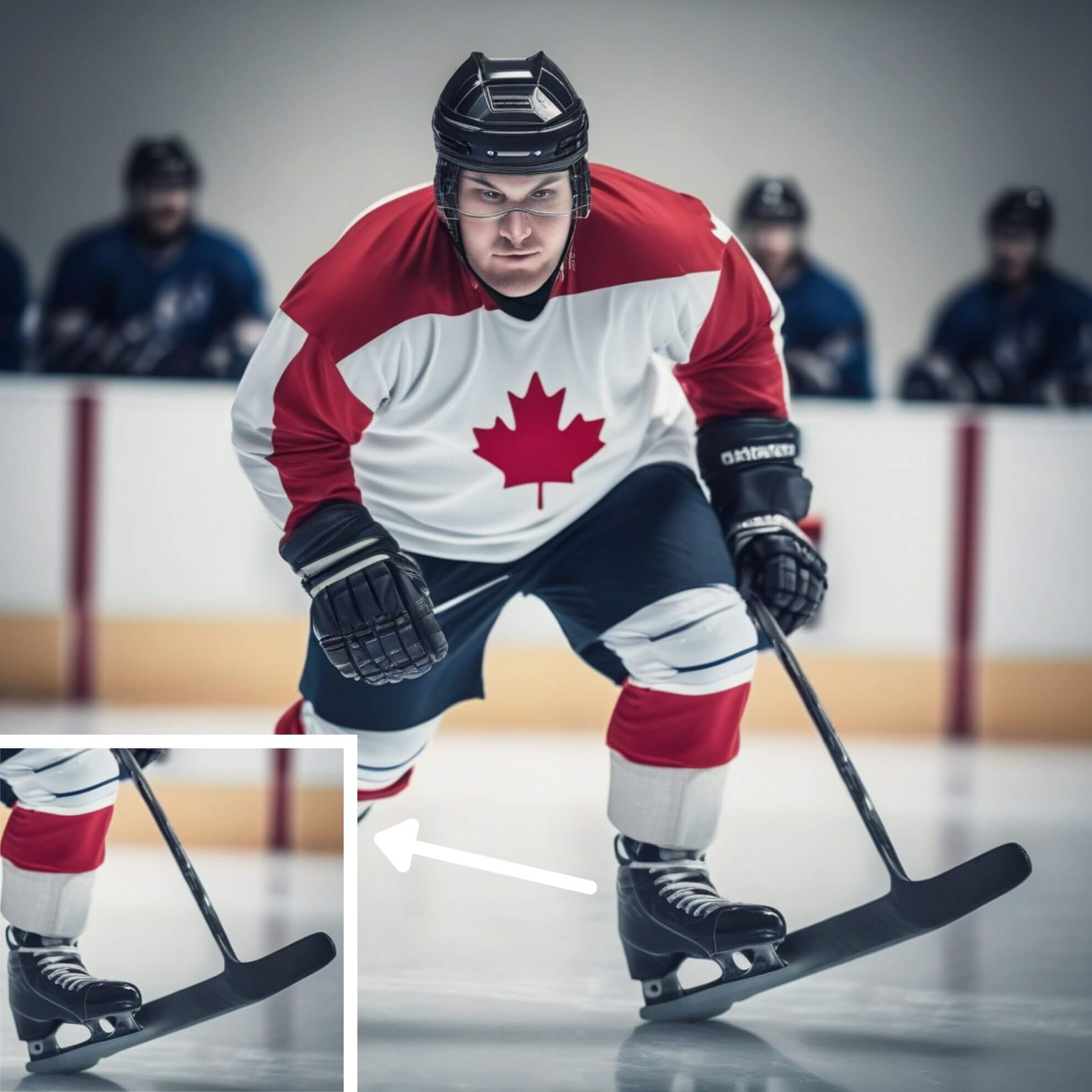} &
\includegraphics[width=0.19\textwidth]{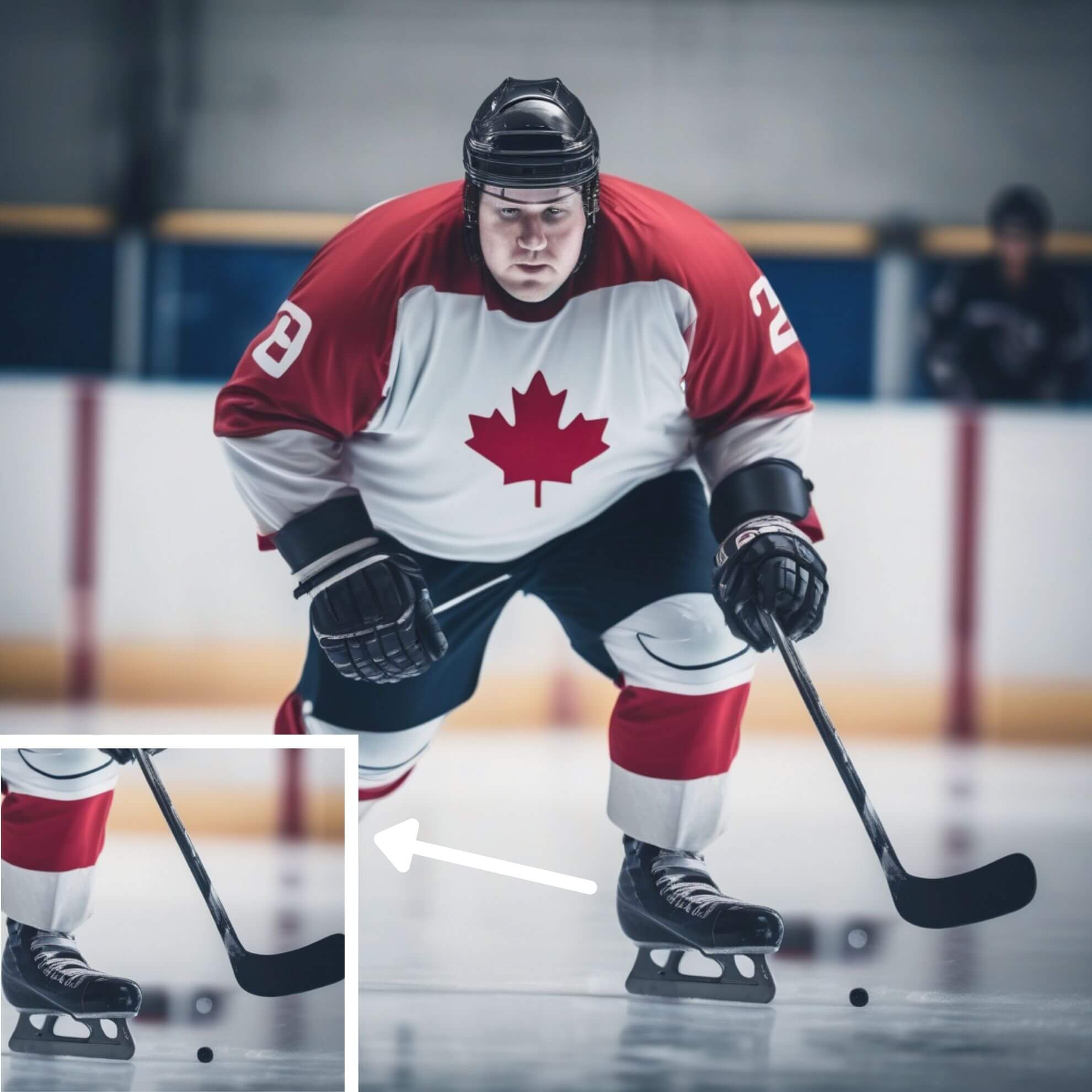}  \\
 \includegraphics[width=0.19\textwidth]{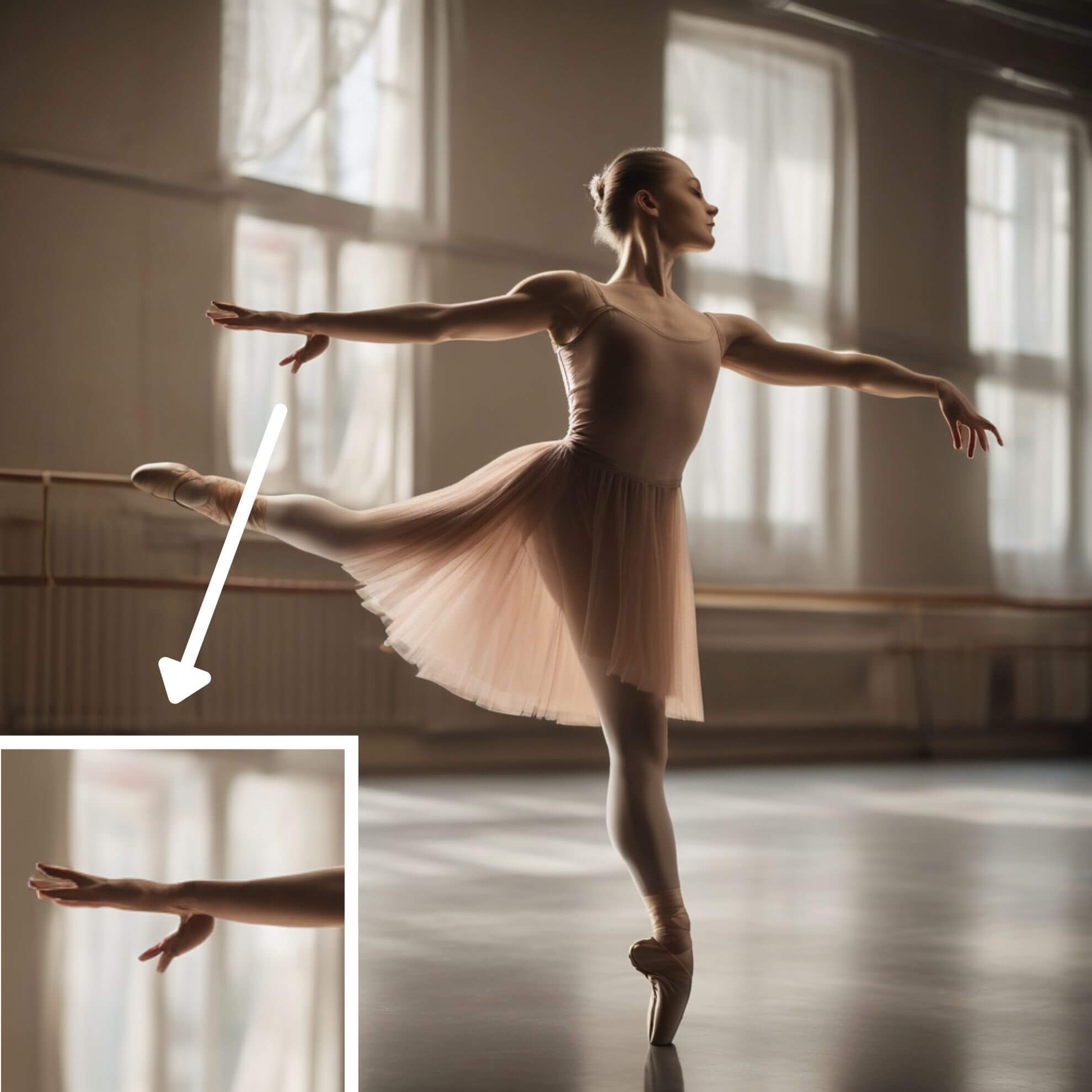} &
\includegraphics[width=0.19\textwidth]{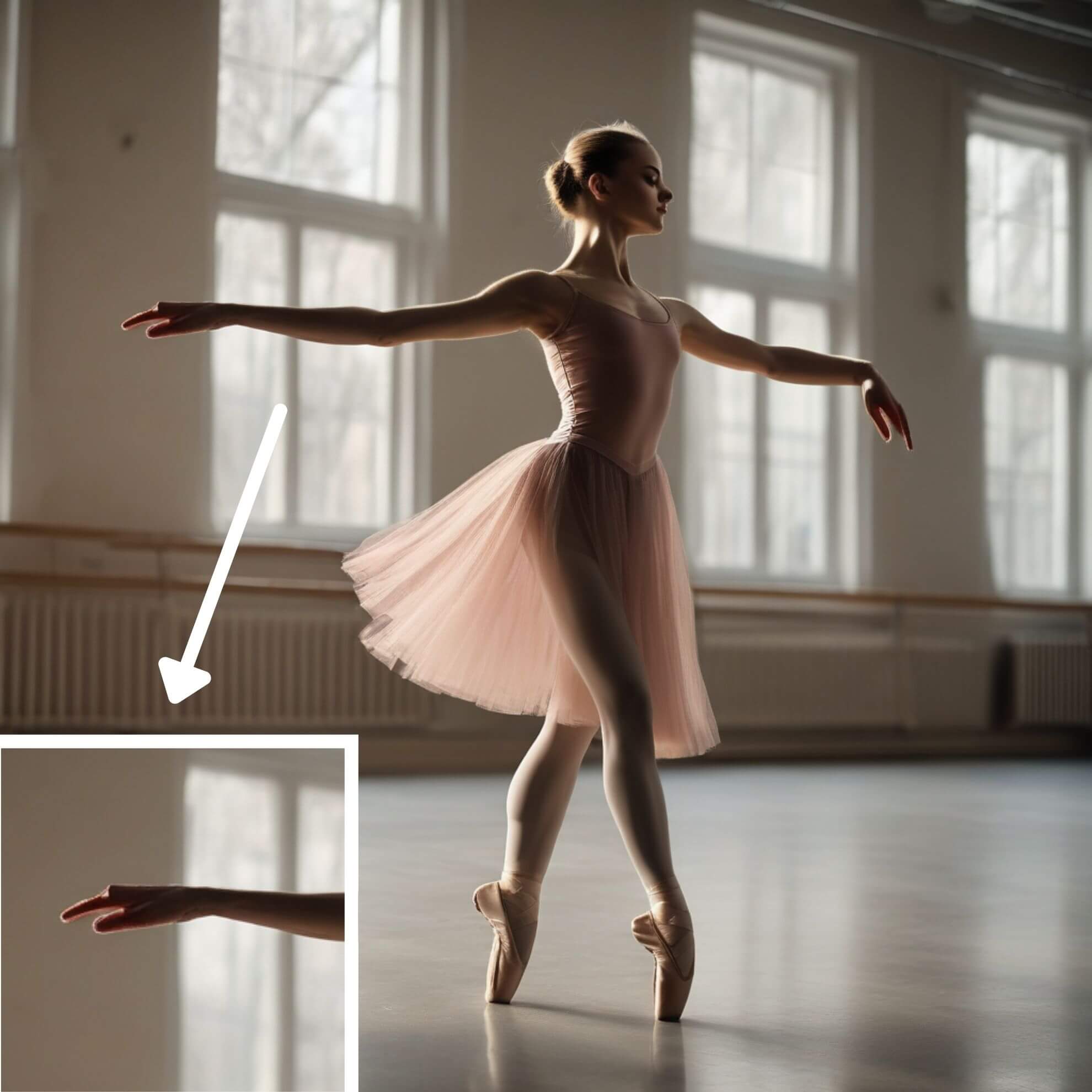} &
\includegraphics[width=0.19\textwidth]{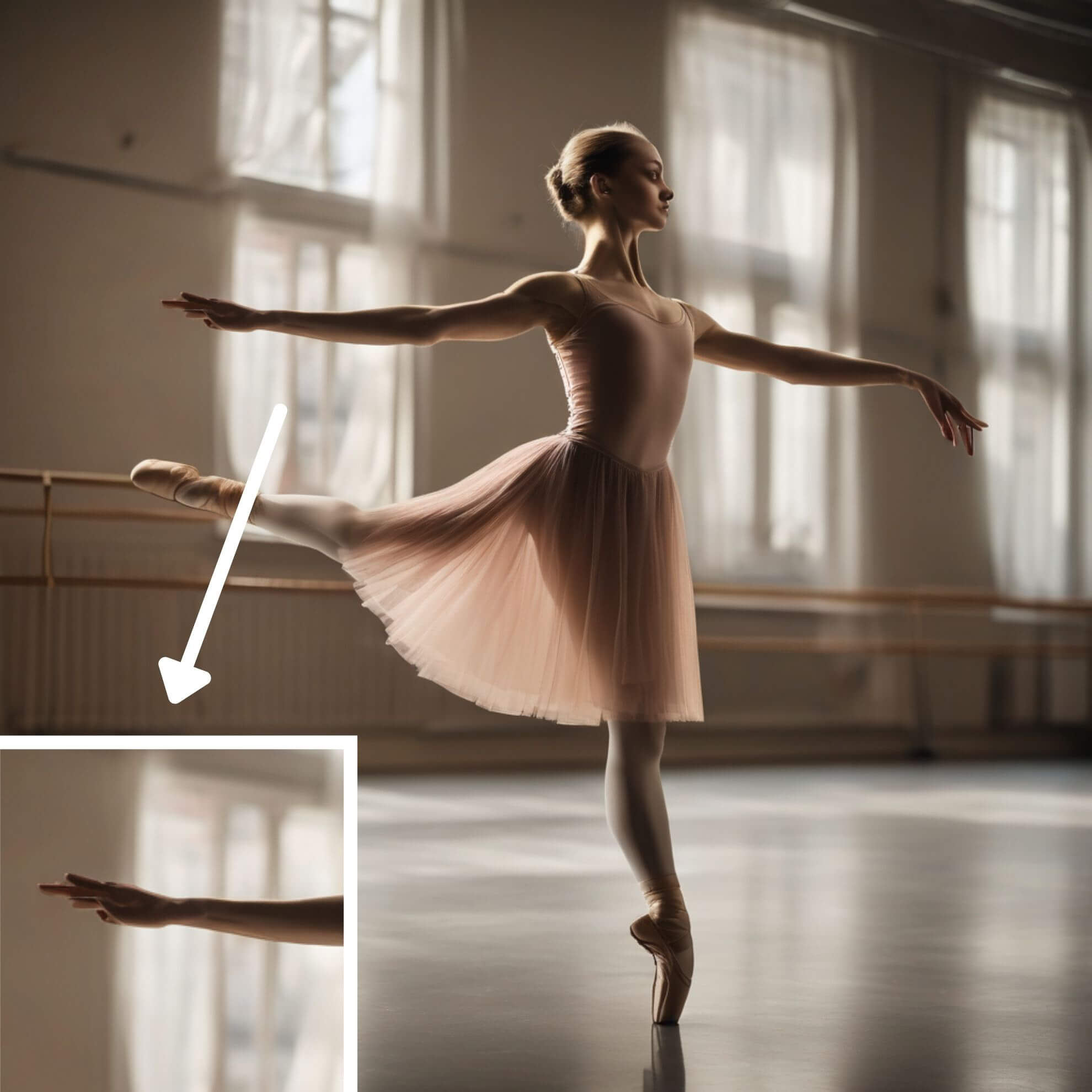} \\
 \includegraphics[width=0.19\textwidth]{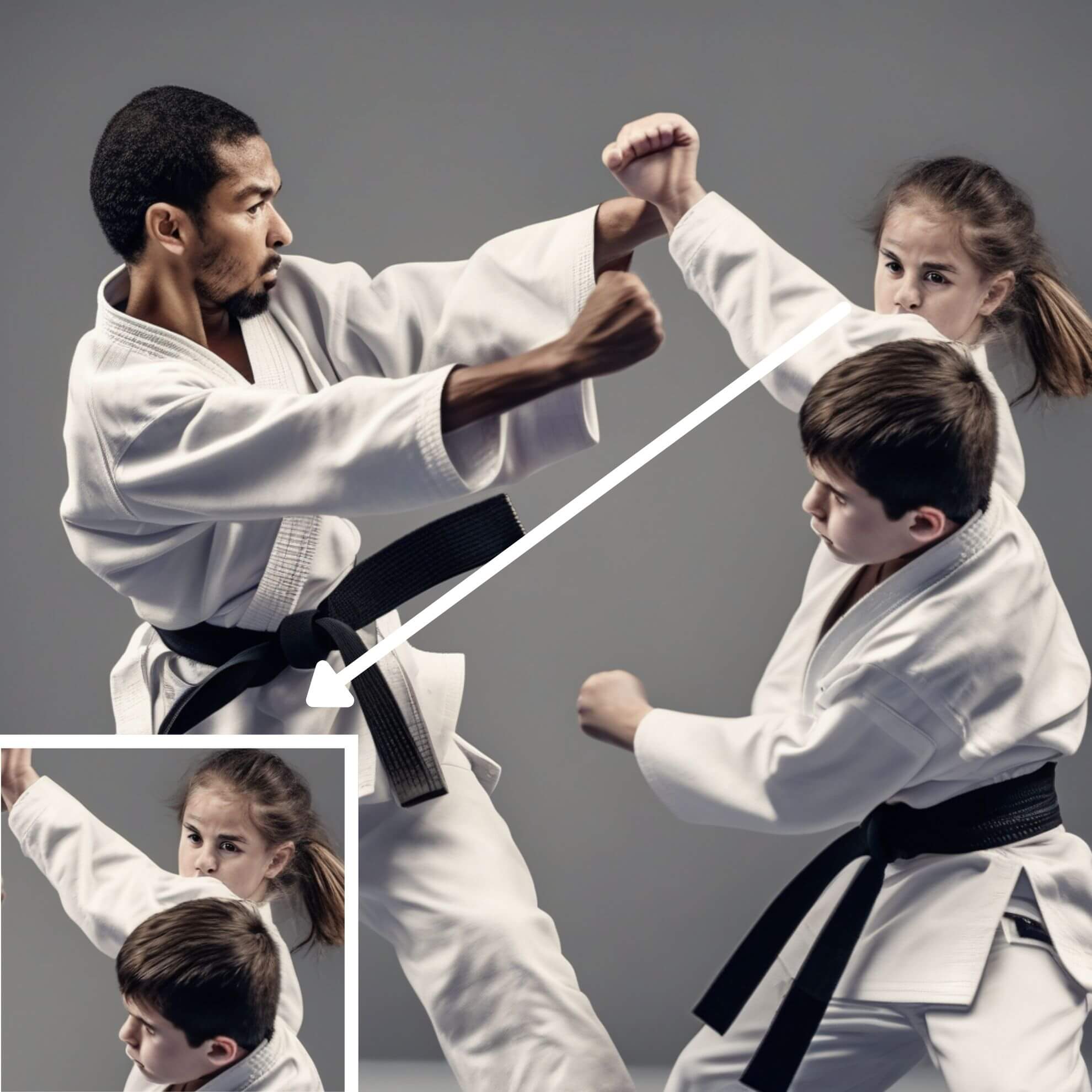} &
\includegraphics[width=0.19\textwidth]{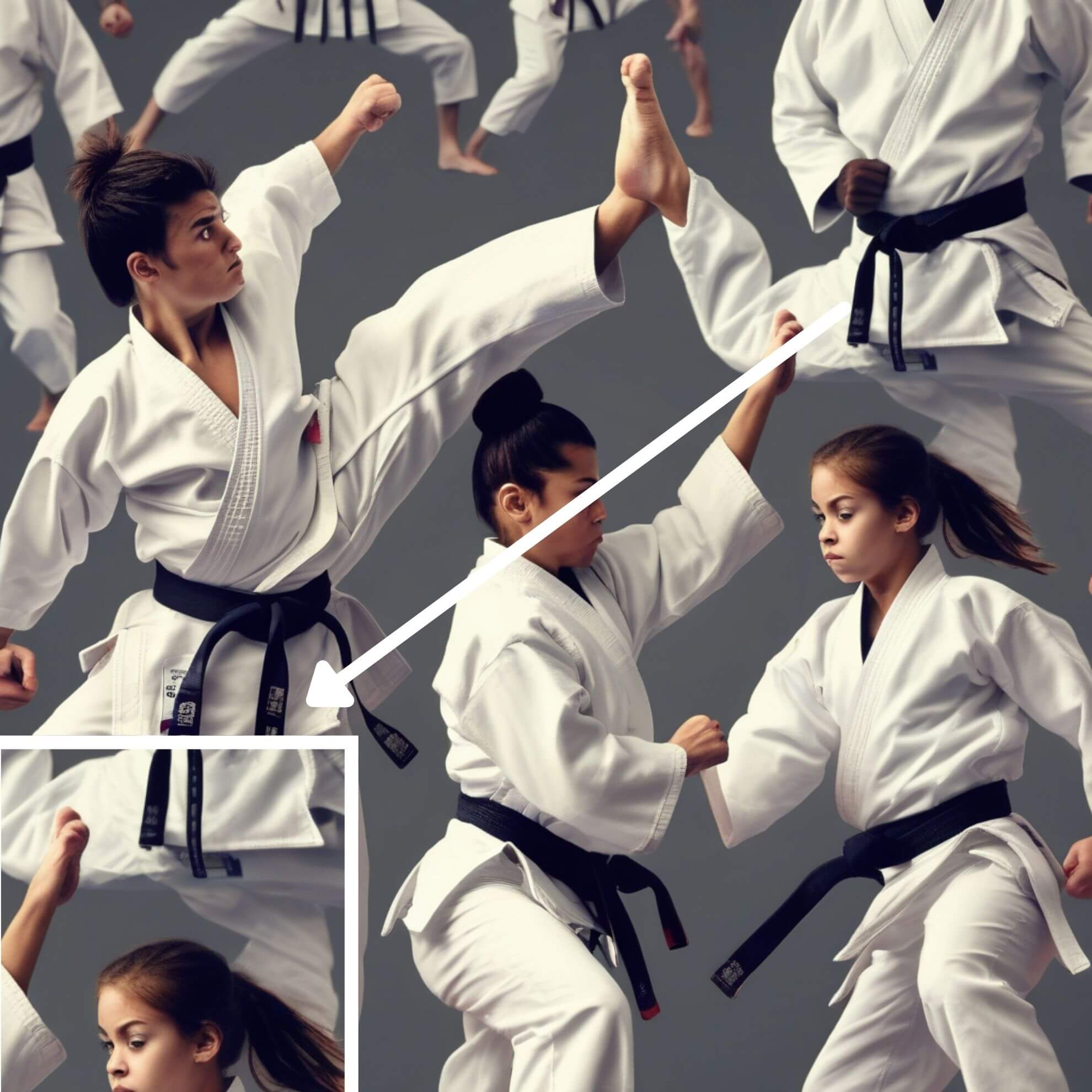} &
\includegraphics[width=0.19\textwidth]{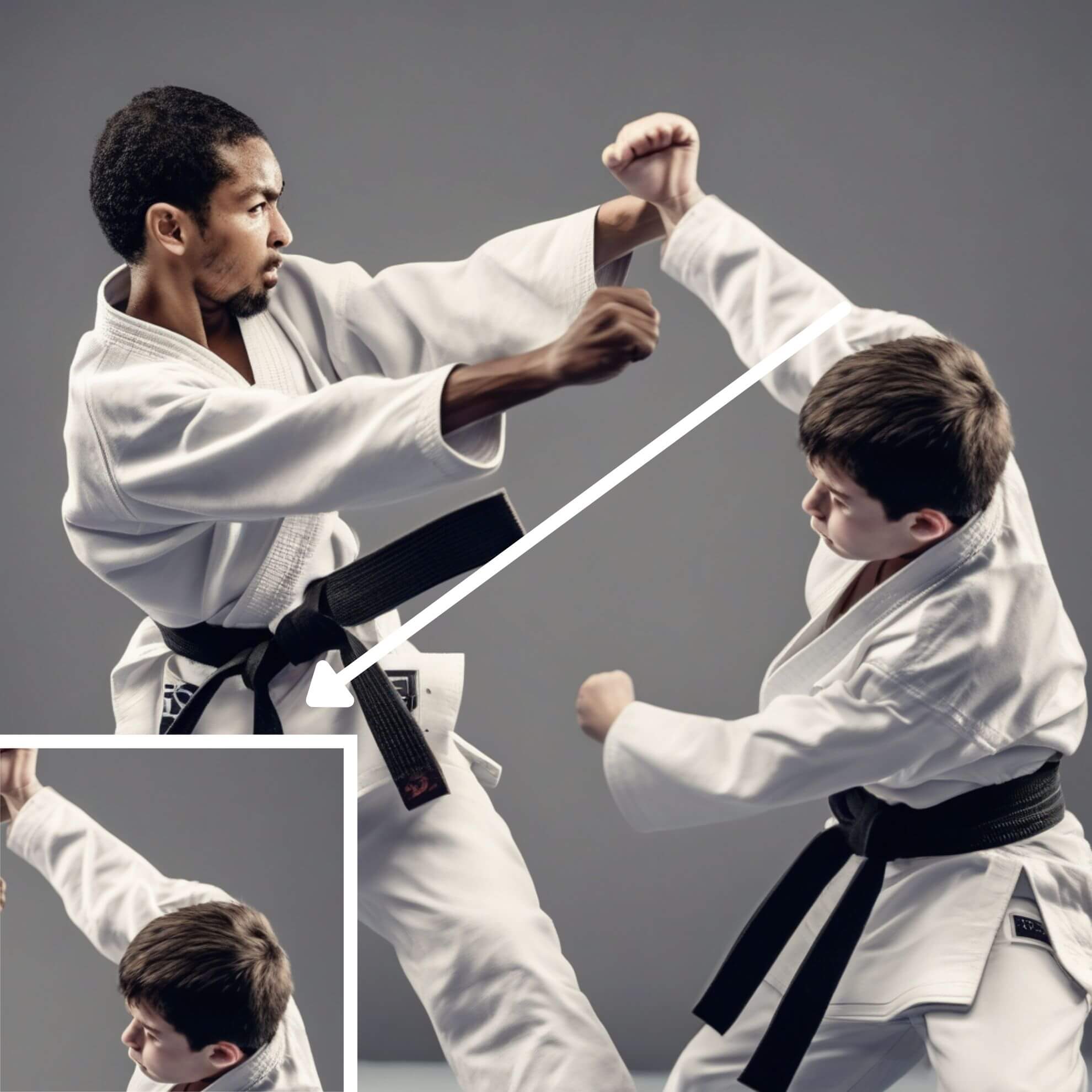} 
\end{tabular}
\caption{\textbf{Images for the SDXL using \our{} for 30 inference steps.} The HandsXL technique introduces more pronounced modifications to images, while our approach focuses on more precise removal limited to artifact areas.}
\label{fig:sota_sdxl}
\end{figure*}


\subsection{Human artifacts}
\label{subsec:human_art}
Qualitative results for the \textit{people} dataset for FLUX.1 [dev] are presented on Fig. \ref{fig_flux_dev_people}. \our{} correctly changes image trajectories to minimize artifacts. Common model errors include deformed hands, multiple legs, and underdeveloped image elements, such as missing a pen for a cat. \our{} allows the model to focus on artifacts during image generation and dynamically steer the trajectory, allowing the model to correctly generate an image, removing duplicate hands, kayaks, etc. Tab. \ref{tab:diffalphareg} presents numerical results for this dataset. \our{} can remove distortions while maintaining high CLIP-T and ImageReward. Furthermore, Fig. \ref{fig:diffdetectors} presents a comparison of generated images using different detectors. \our{} allows the use of multiple responses from classifiers for artifact removal.

Additionally, visualizations for the SDXL and FLUX.2 [dev] are presented in Figs. \ref{fig:sota_sdxl} and \ref{fig:flux2_images}. \our{} performs better because it can remove artifacts without significantly affecting the rest of the image. This is clearly visible for the marathon runners or the karate fight scene, where the compositions are completely different from the reference image for HandsXL.

\begin{table*}[h]
\centering
\caption{ \textbf{Evaluation of FLUX.1 [dev] on the \textit{people} dataset.} Results for our technique compared to other state-of-the-art models. We compare our \our{} with gradient normalization against the version without gradient normalization (\our{} w/o norm), showing that normalization leads to a greater reduction of artifacts.}
\setlength{\tabcolsep}{3.7pt}
{
\fontsize{6.8pt}{10pt}\selectfont
\begin{tabular}{lccccccc}
\hline
Model              & CLIP-T $\uparrow$ & Mean Artifact Freq (\%) $\downarrow$ & ImageReward $\uparrow$ & Artfiact Pixel Ratio (\%) $\downarrow$ & MAE $\downarrow$   & MAE (A) $\downarrow$  & MAE (NA) $\downarrow$ \\ \hline
Baseline FLUX.1 {[}dev{]}   & 35.820 $\pm$ 0.175  & 100.000 $\pm$ 0.000               & 0.961 $\pm$ 0.035       & 0.525 $\pm$ 0.039            & -  & -       & -          \\
+ DiffDoctor       & 35.879 $\pm$ 0.338  & 44.000 $\pm$ 2.944                & 0.917 $\pm$ 0.061       & 0.190 $\pm$ 0.016            & 12.445 $\pm$ 0.131  & 26.029 $\pm$ 1.092       & 12.377 $\pm$ 0.134        \\
+ HPSv2            & 35.938 $\pm$ 0.079  & 76.000 $\pm$ 2.708                & 0.953 $\pm$ 0.036       & 0.428 $\pm$ 0.049            & 6.632 $\pm$ 0.333  & 16.998 $\pm$ 0.714       & 6.582 $\pm$ 0.114        \\
+ HandsXL          & 35.642 $\pm$ 0.204  & 47.750 $\pm$ 0.957                & 0.899 $\pm$ 0.062       & 0.270 $\pm$ 0.038            & 24.470 $\pm$ 0.269  & 38.002 $\pm$ 0.813       & 24.404 $\pm$ 0.262          \\
+ \textbf{\our{} w/o norm} & 35.850 $\pm$ 0.213  & 95.000 $\pm$ 3.464 & 	0.959 $\pm$ 0.037 & 0.526 $\pm$ 0.040 & 1.306 $\pm$ 0.073 & 4.741 $\pm$ 0.508 &  1.291 $\pm$ 0.069          \\
+ \textbf{\our{}} & 35.762 $\pm$ 0.101  & 15.500 $\pm$ 2.380                & 0.968 $\pm$ 0.034        & 0.068 $\pm$ 0.022            & 9.617 $\pm$ 0.240  & 25.764 $\pm$ 0.604       & 9.545 $\pm$ 0.240          \\ \hline
\end{tabular}
}

\label{tab:diffalphareg}
\end{table*}

\begin{figure*}[h]
\centering
\setlength{\tabcolsep}{1.5pt}
\renewcommand{\arraystretch}{0.9}

\begin{tabular}{c c c c c c c c c c c}
\footnotesize{$I_{base}$} & \footnotesize{$\mathcal{AD}^D(I_{base})$} & \footnotesize{$\mathcal{AD}^R(I_{base})$} & \footnotesize{$I_D$} & \footnotesize{$\mathcal{AD}^D(I_{D})$} & \footnotesize{$I_R$} & \footnotesize{$\mathcal{AD}^R(I_{D})$} & \footnotesize{$I_{D+R}$} & \footnotesize{$\mathcal{AD}^{D+R}(I_{D+R})$} \\

\includegraphics[width=0.10\textwidth]{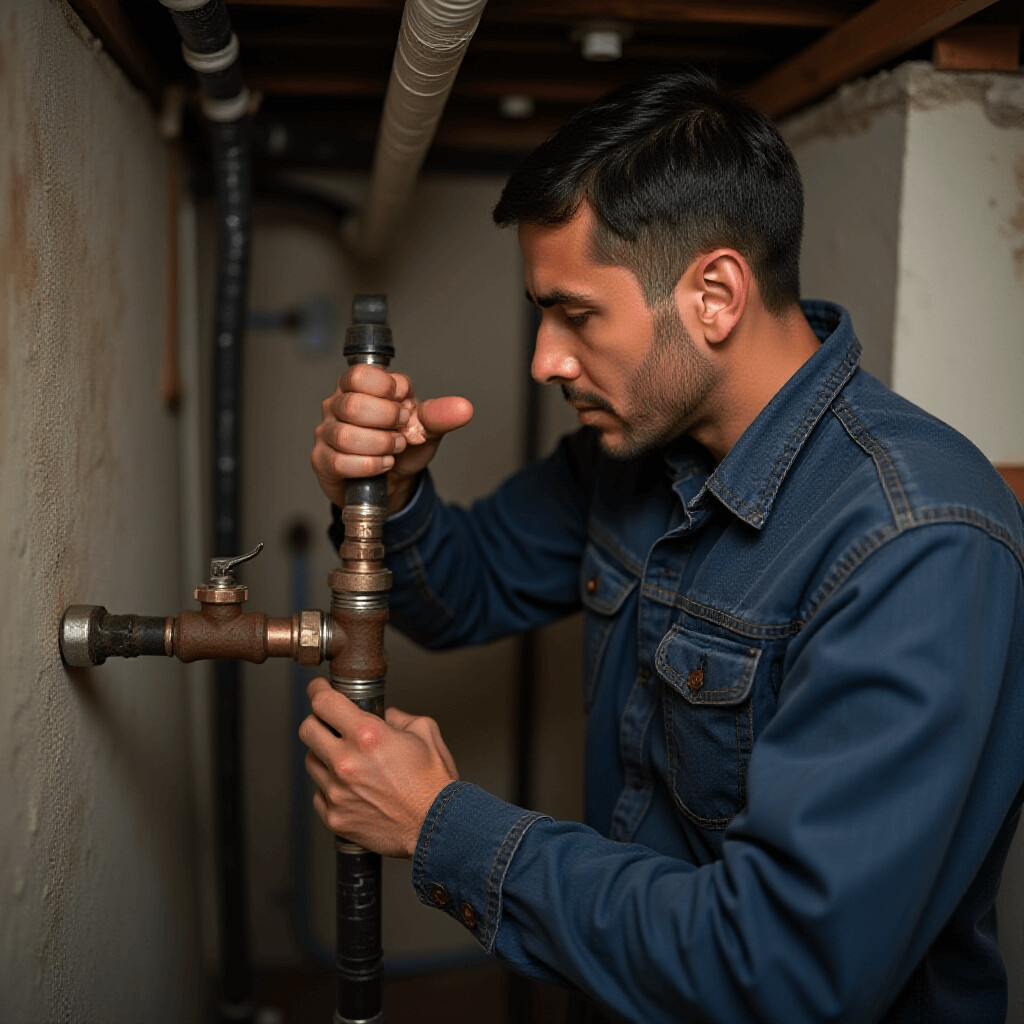} &
\includegraphics[width=0.10\textwidth]{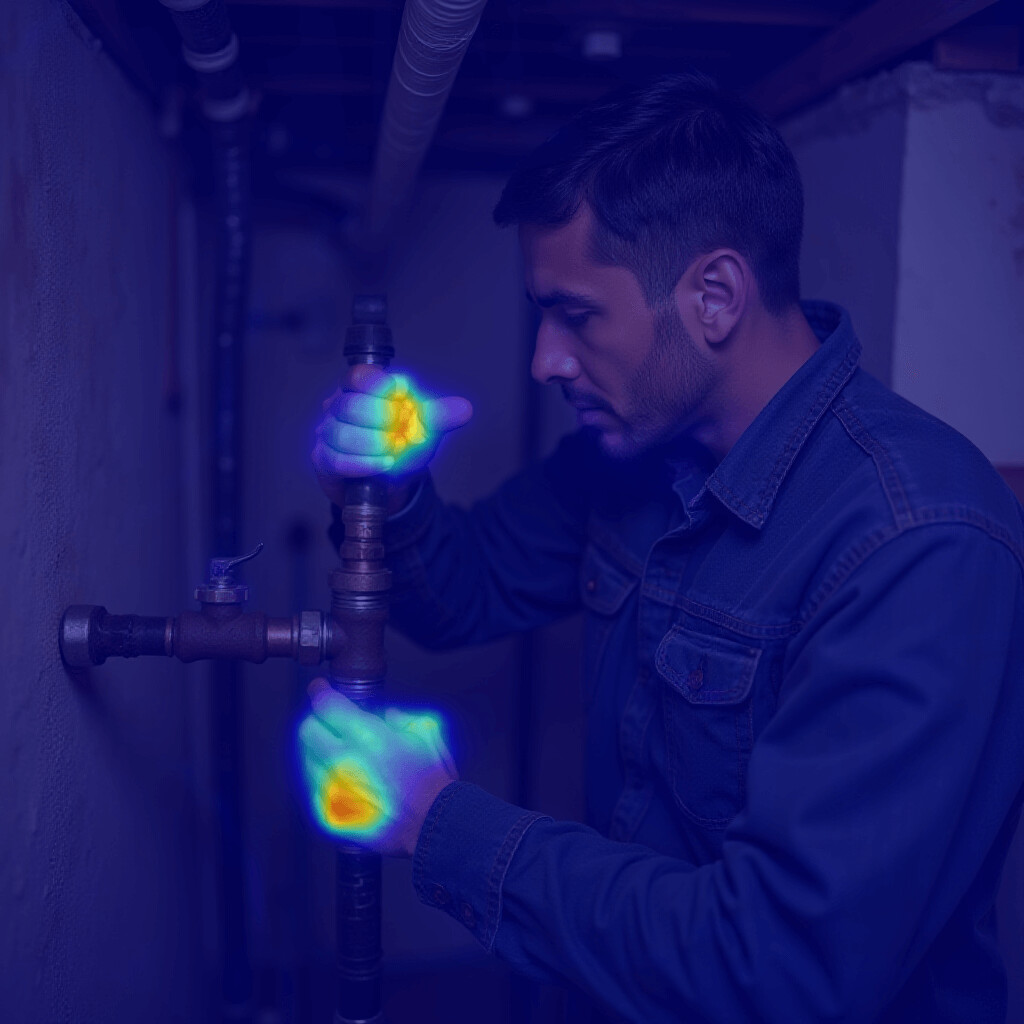} &
\includegraphics[width=0.10\textwidth]{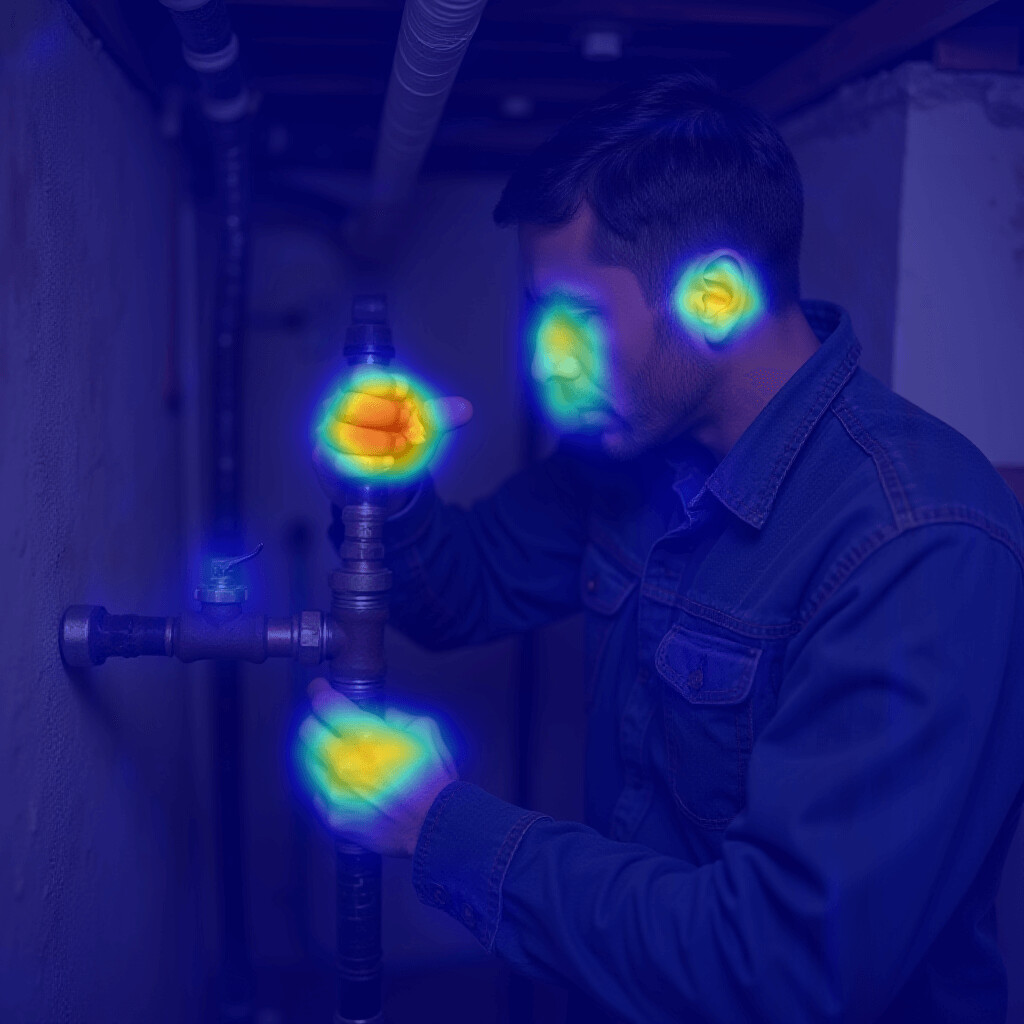} &
\includegraphics[width=0.10\textwidth]{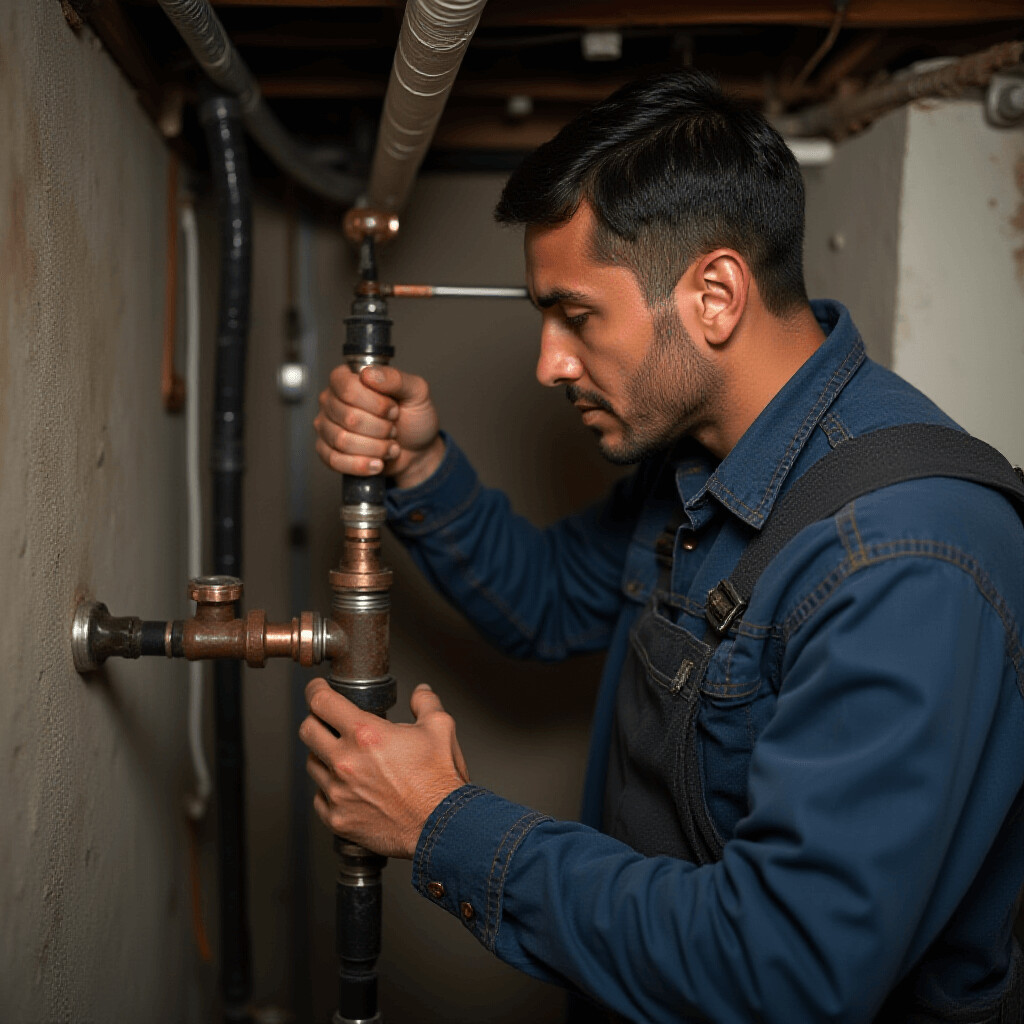} &
\includegraphics[width=0.10\textwidth]{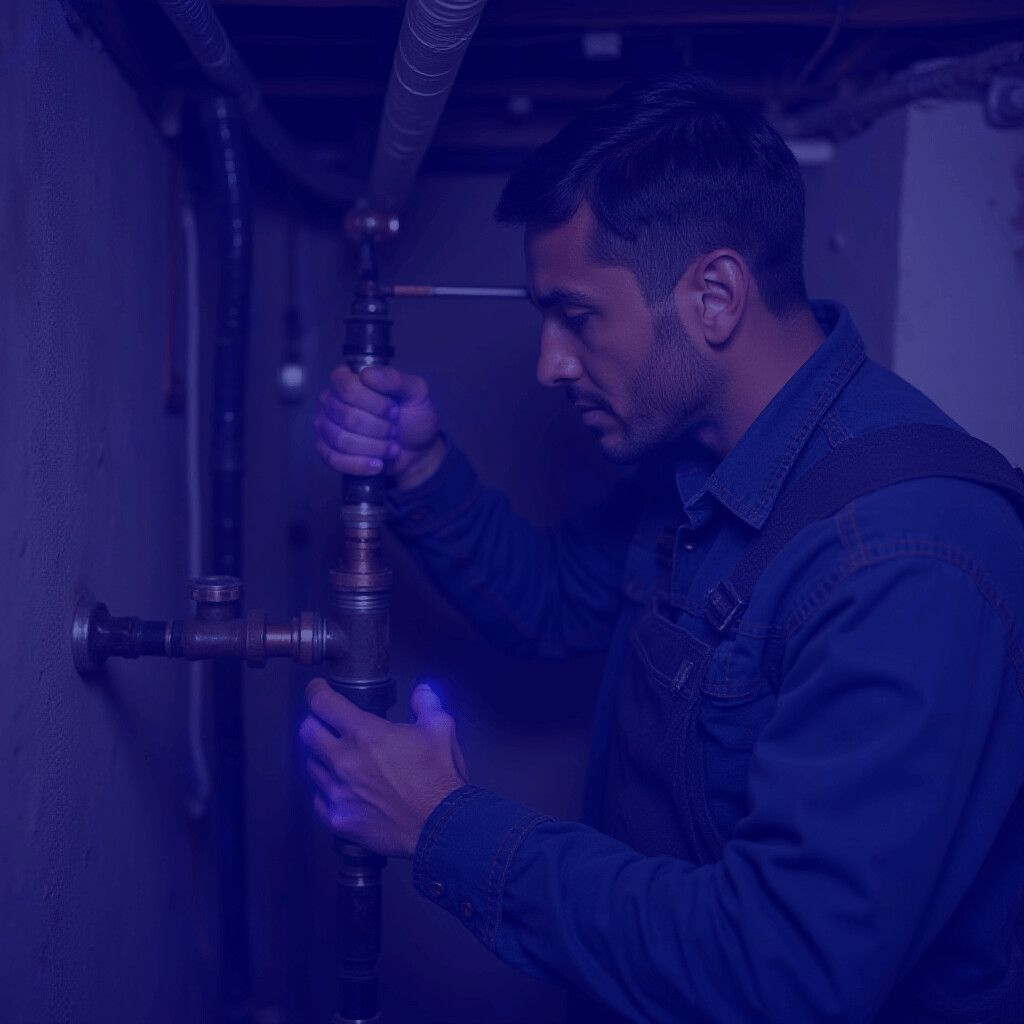} &
\includegraphics[width=0.10\textwidth]{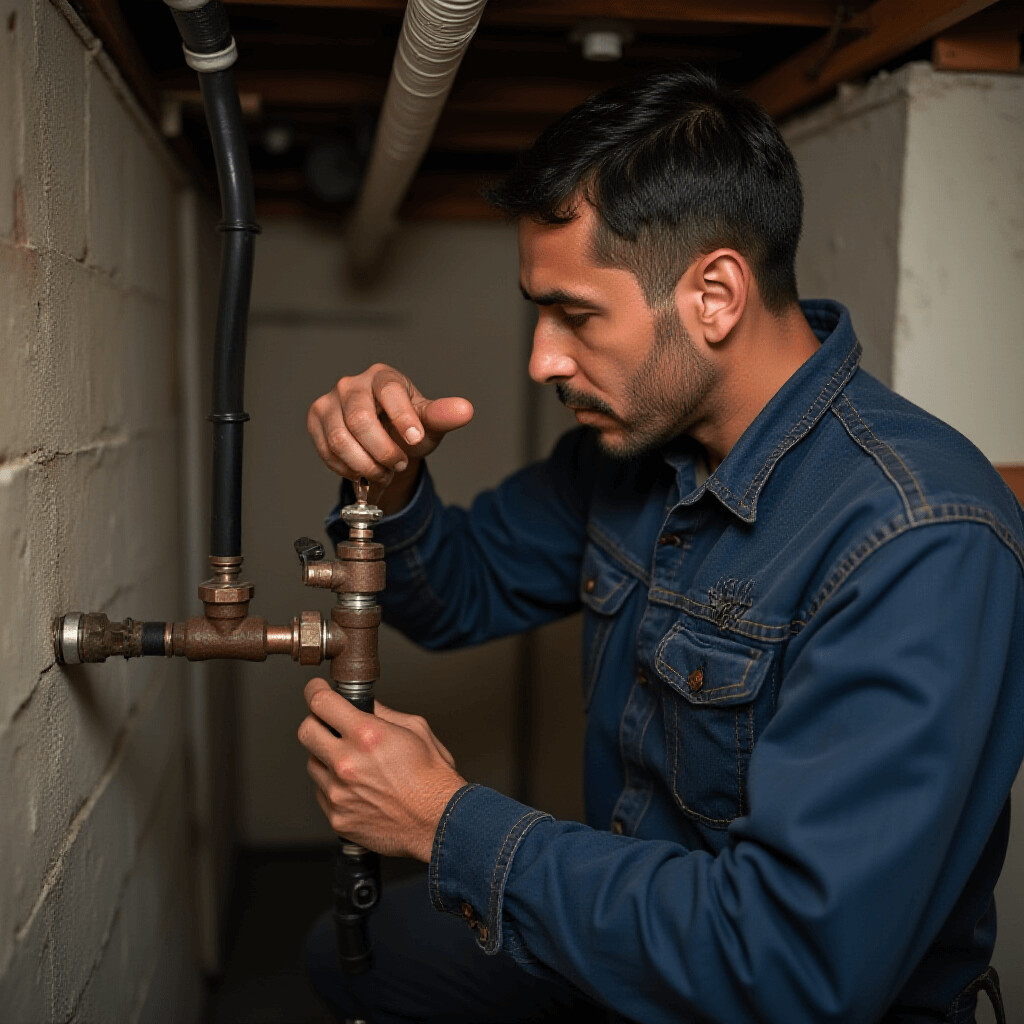} &
\includegraphics[width=0.10\textwidth]{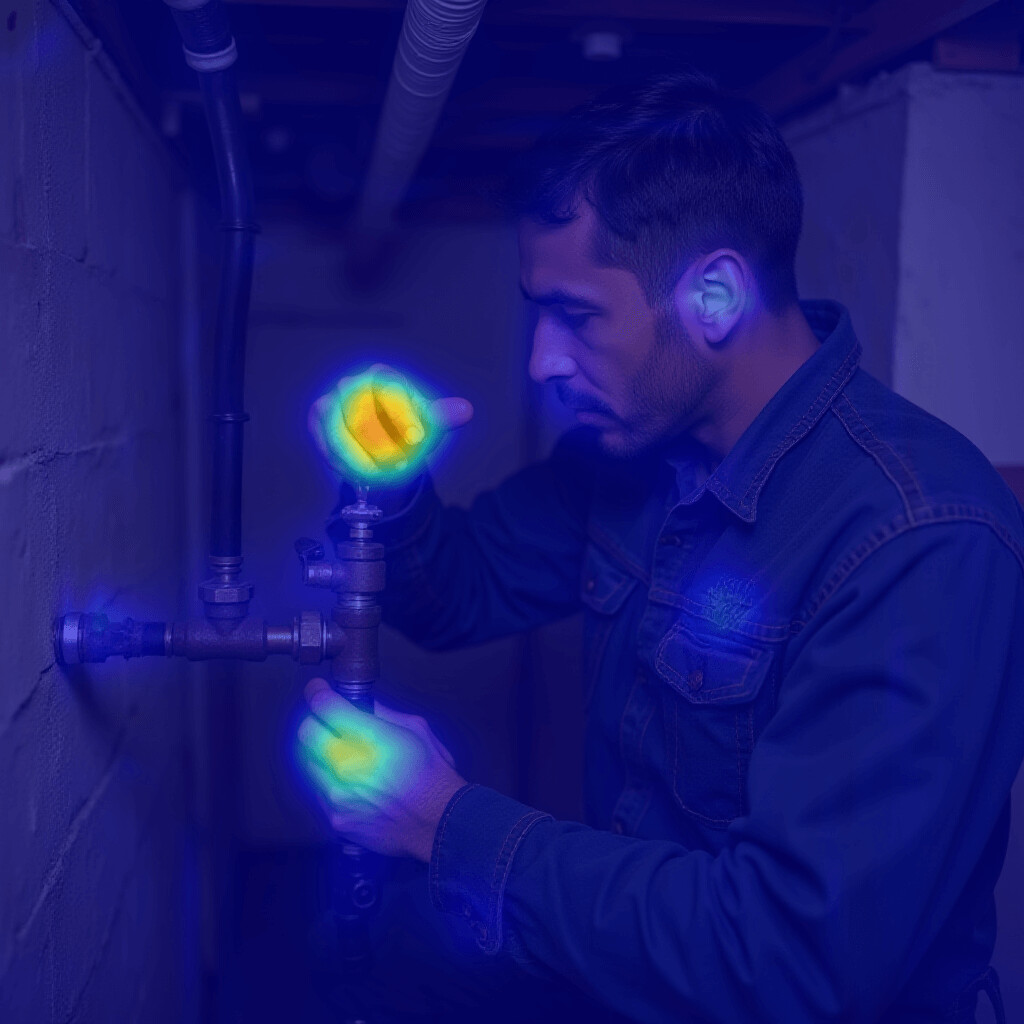} &
\includegraphics[width=0.10\textwidth]{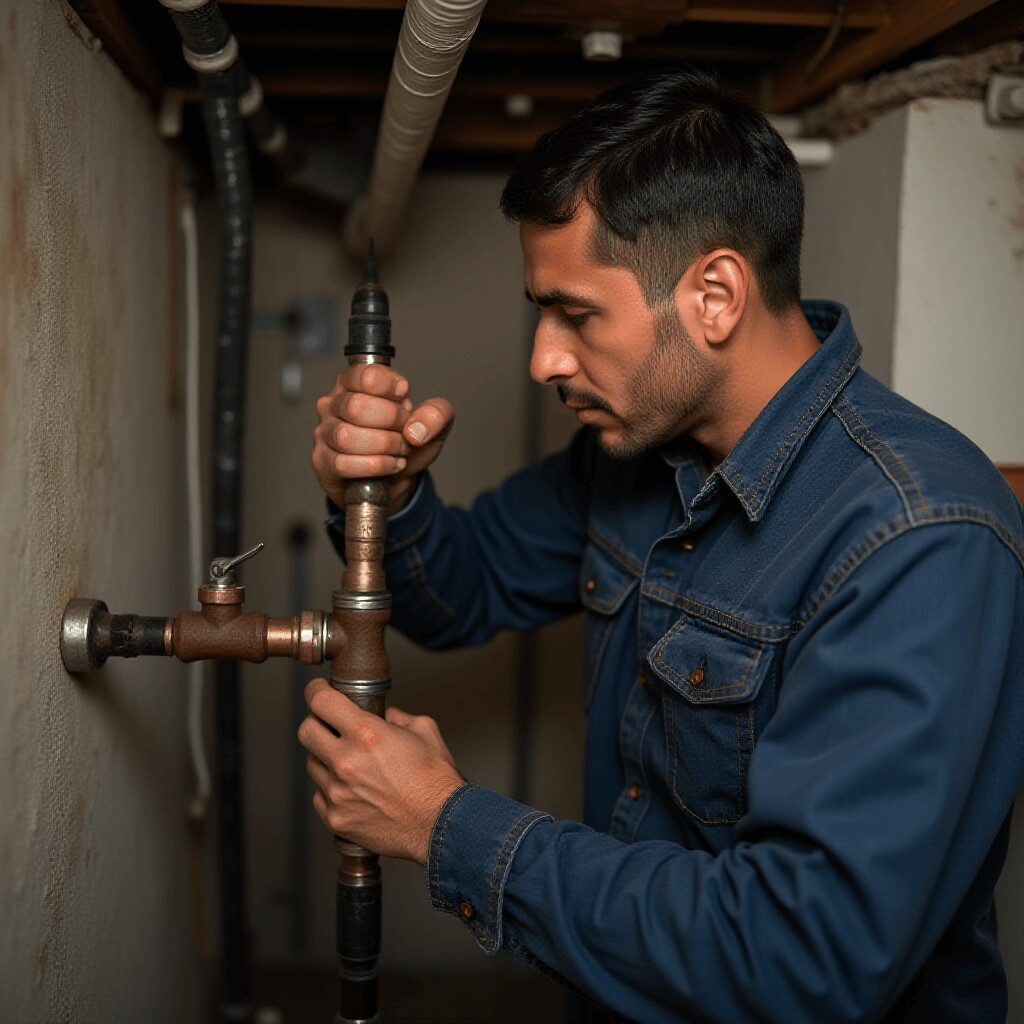} &
\includegraphics[width=0.10\textwidth]{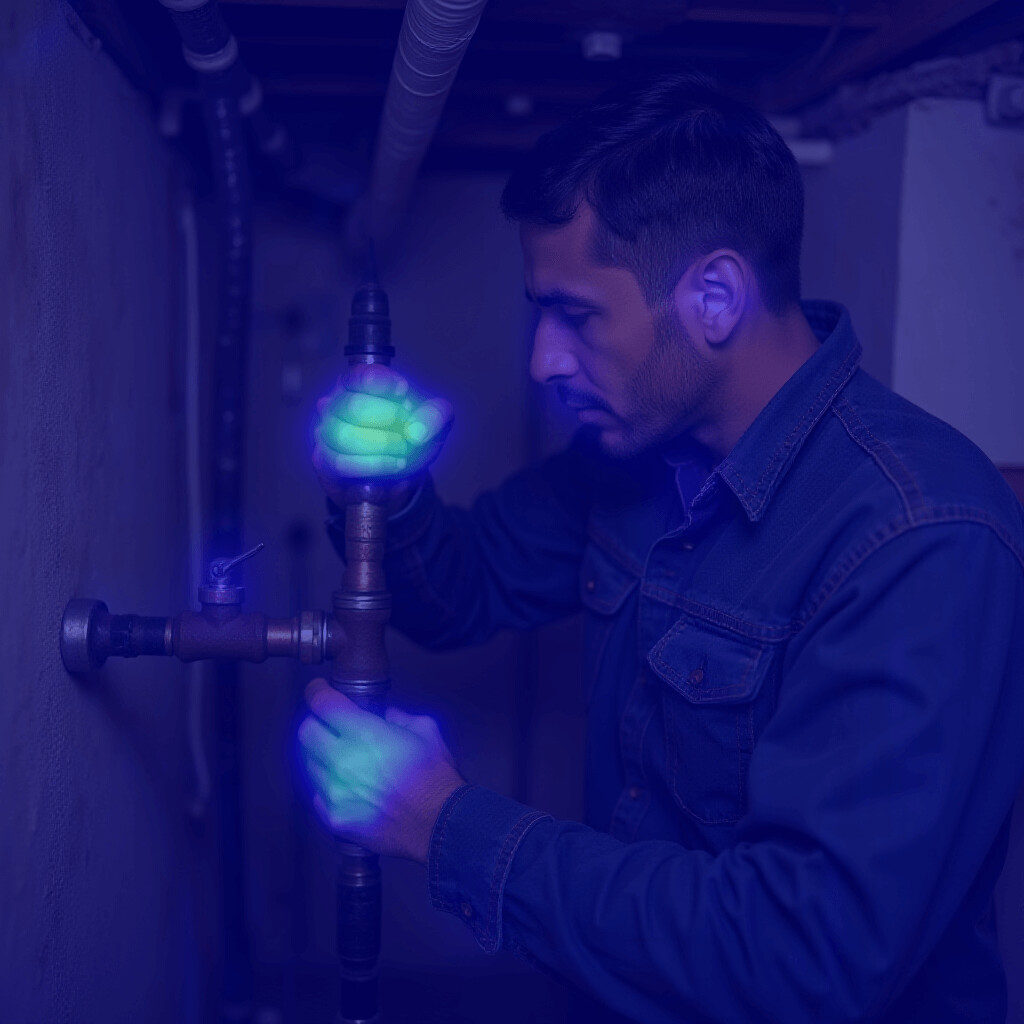} \\

\includegraphics[width=0.10\textwidth]{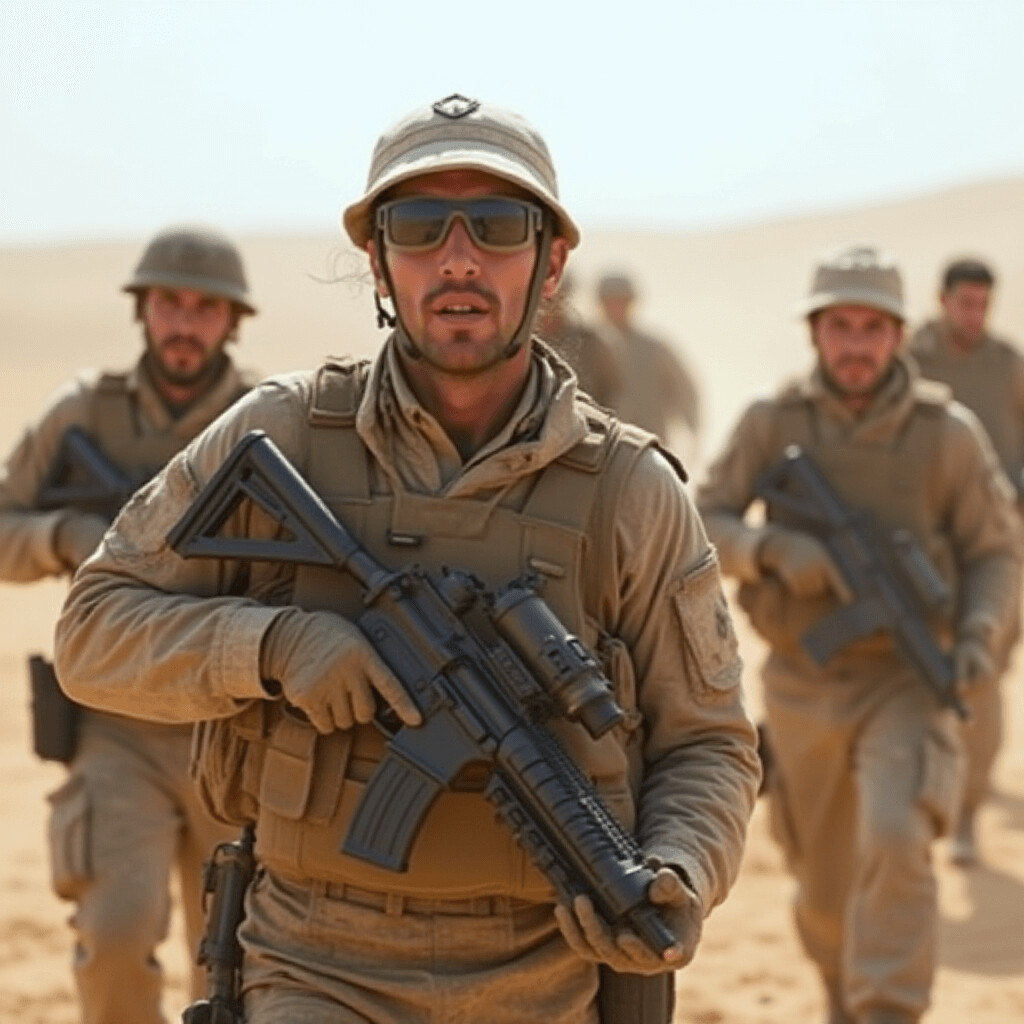} &
\includegraphics[width=0.10\textwidth]{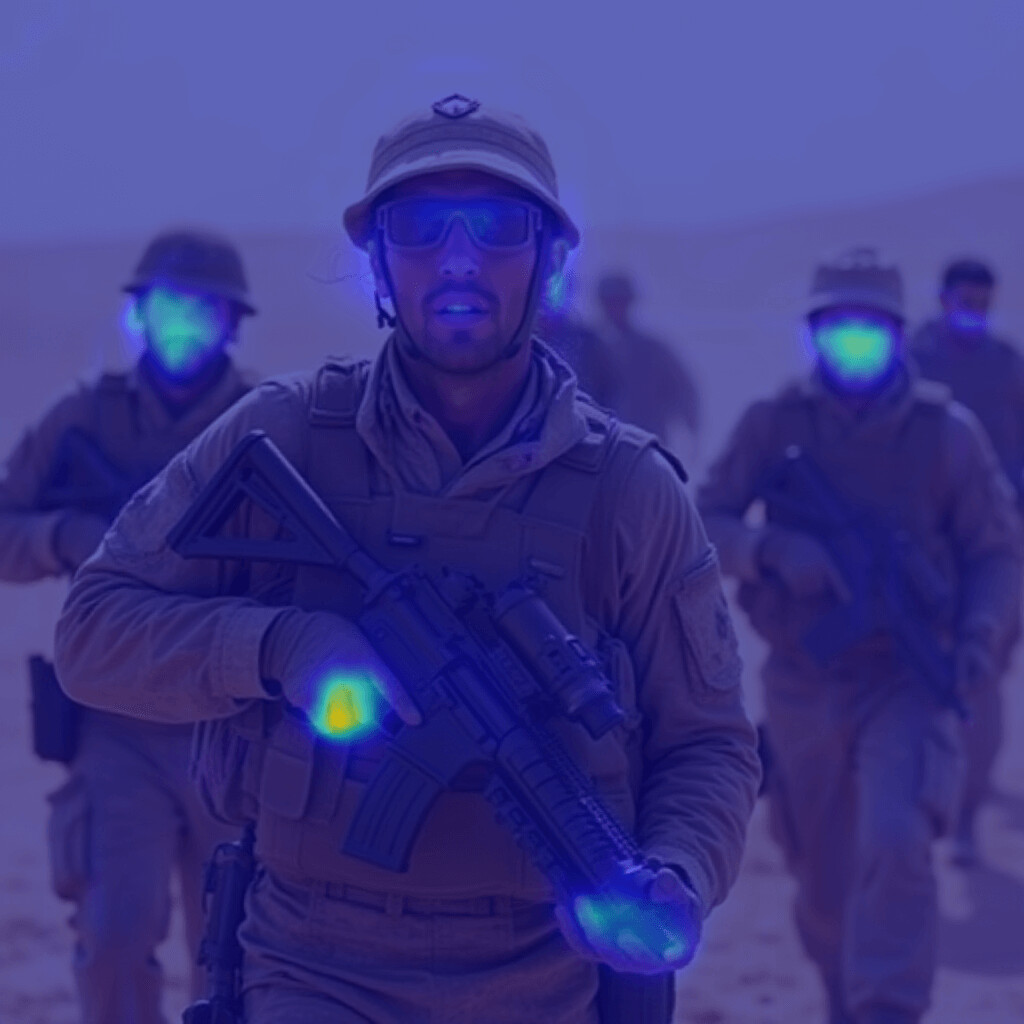} &
\includegraphics[width=0.10\textwidth]{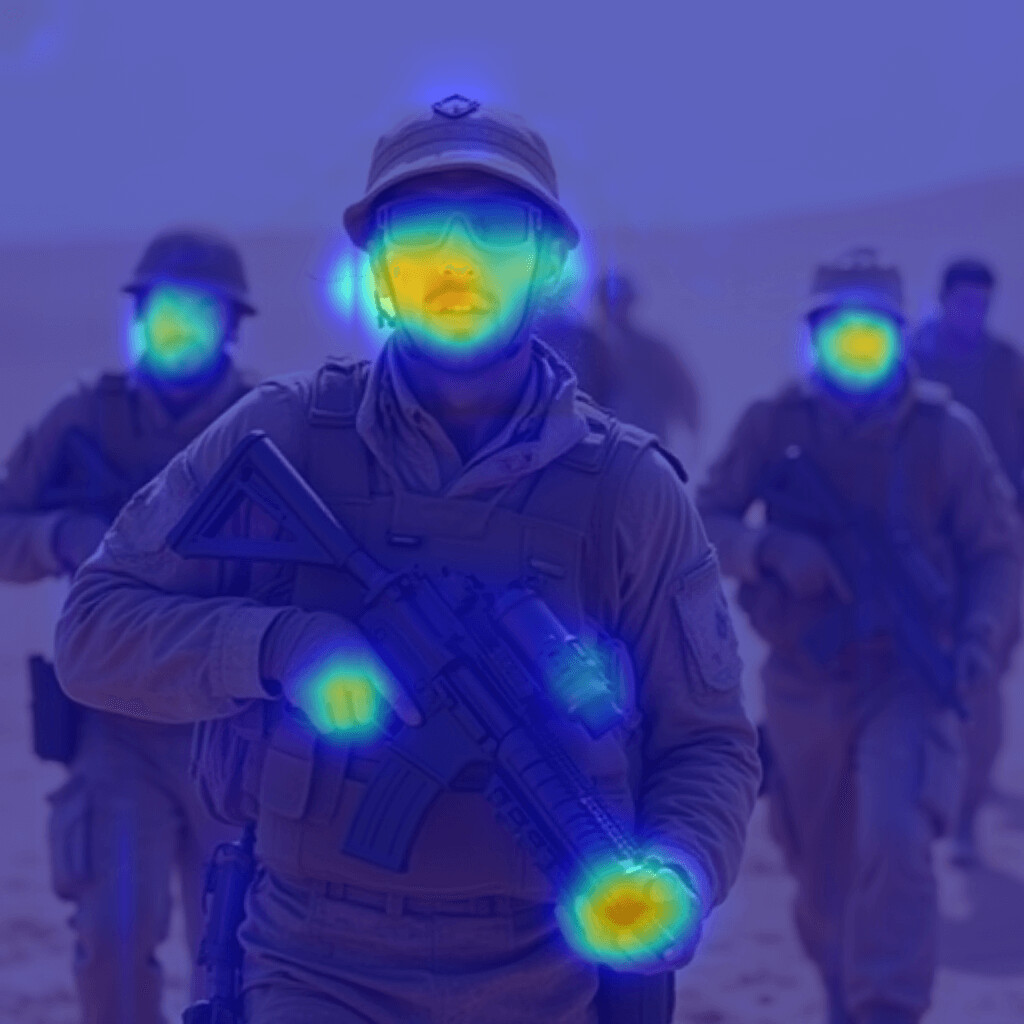} &
\includegraphics[width=0.10\textwidth]{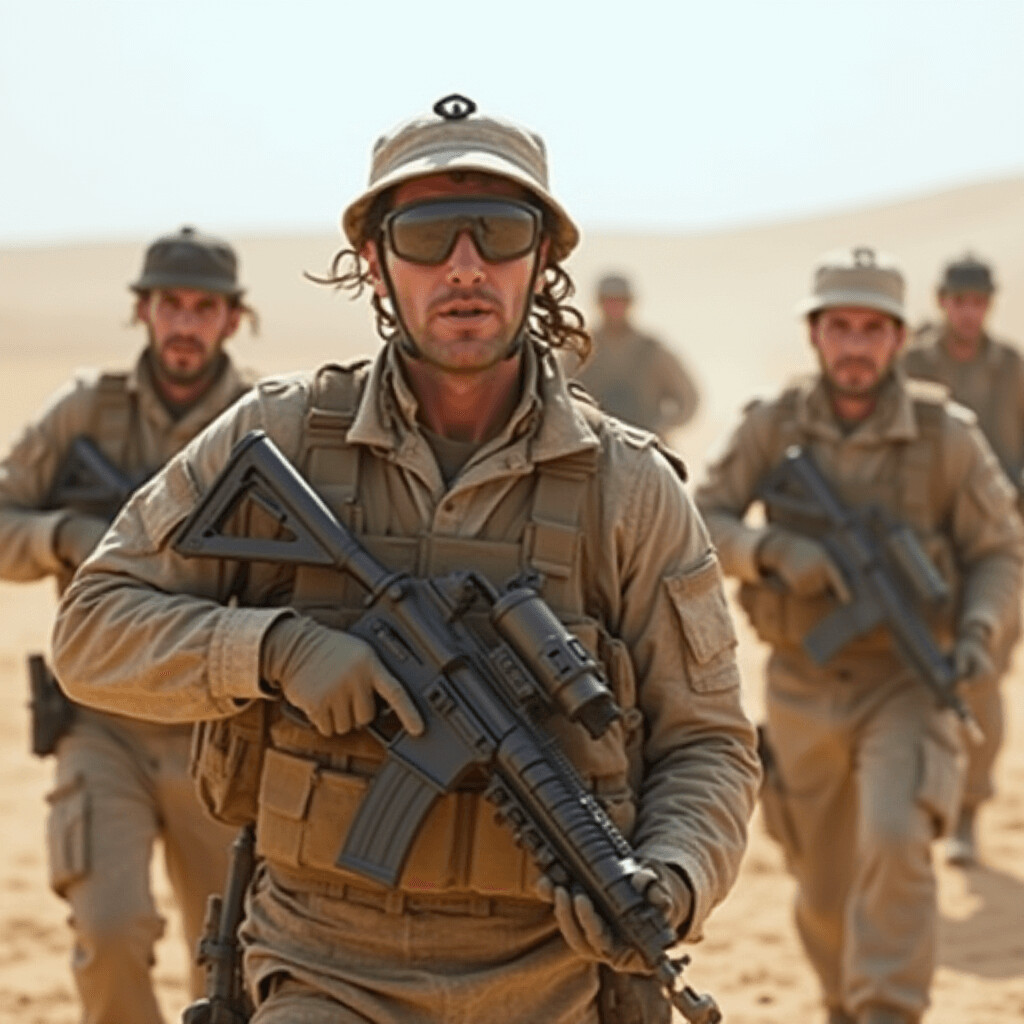} &
\includegraphics[width=0.10\textwidth]{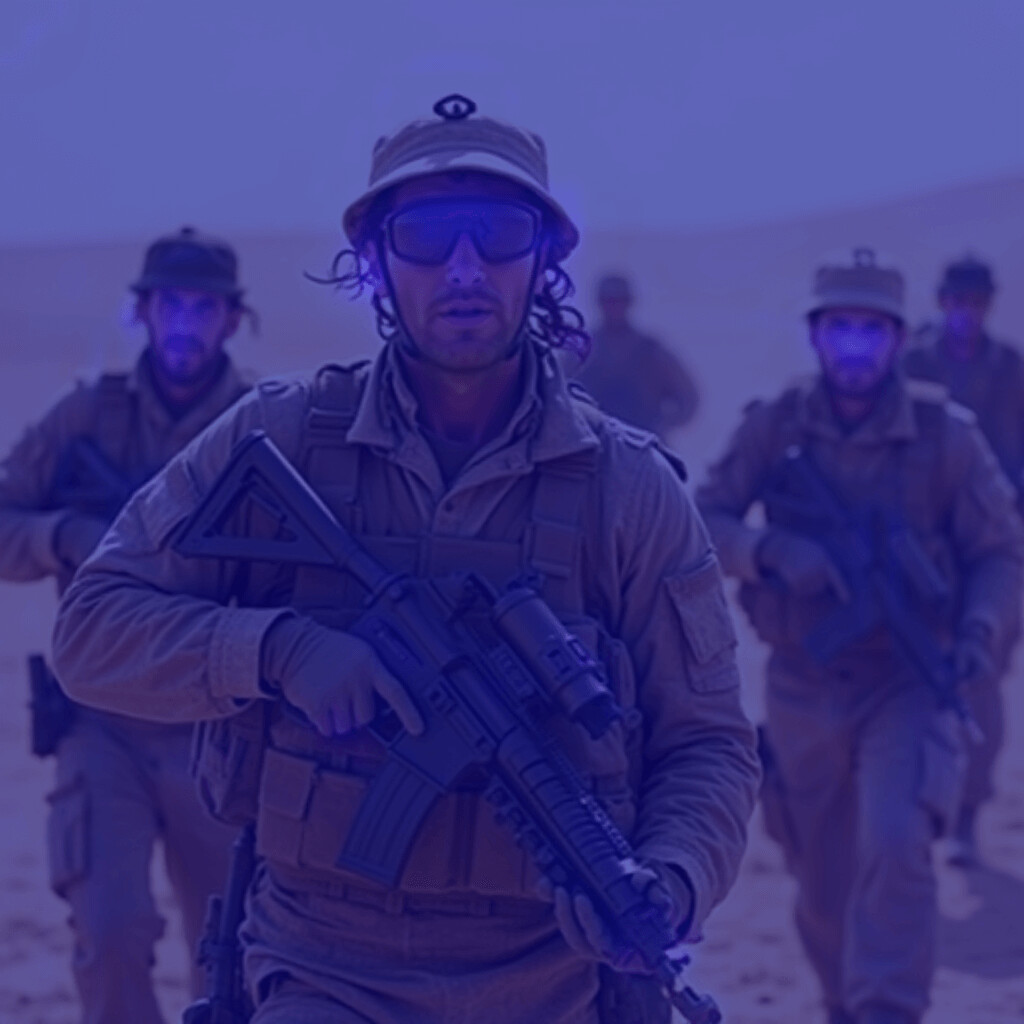} &
\includegraphics[width=0.10\textwidth]{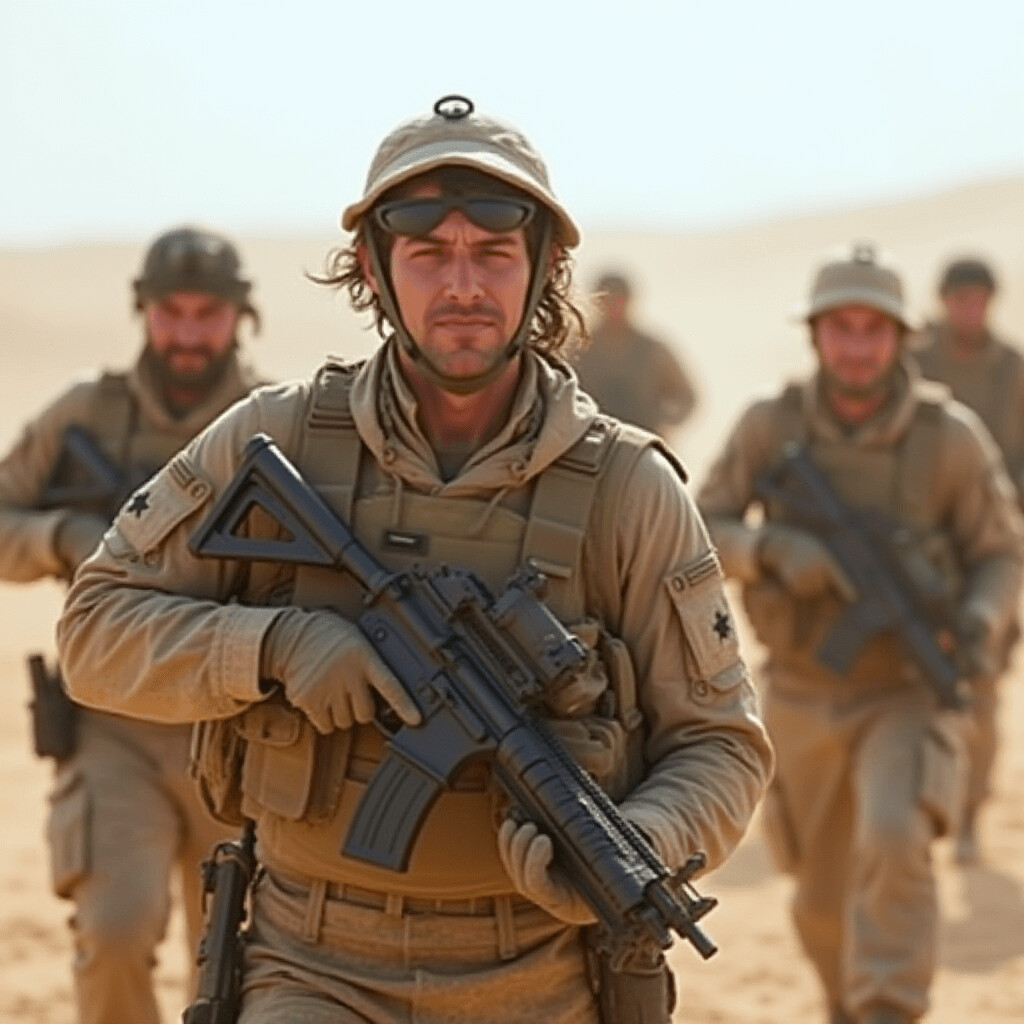} &
\includegraphics[width=0.10\textwidth]{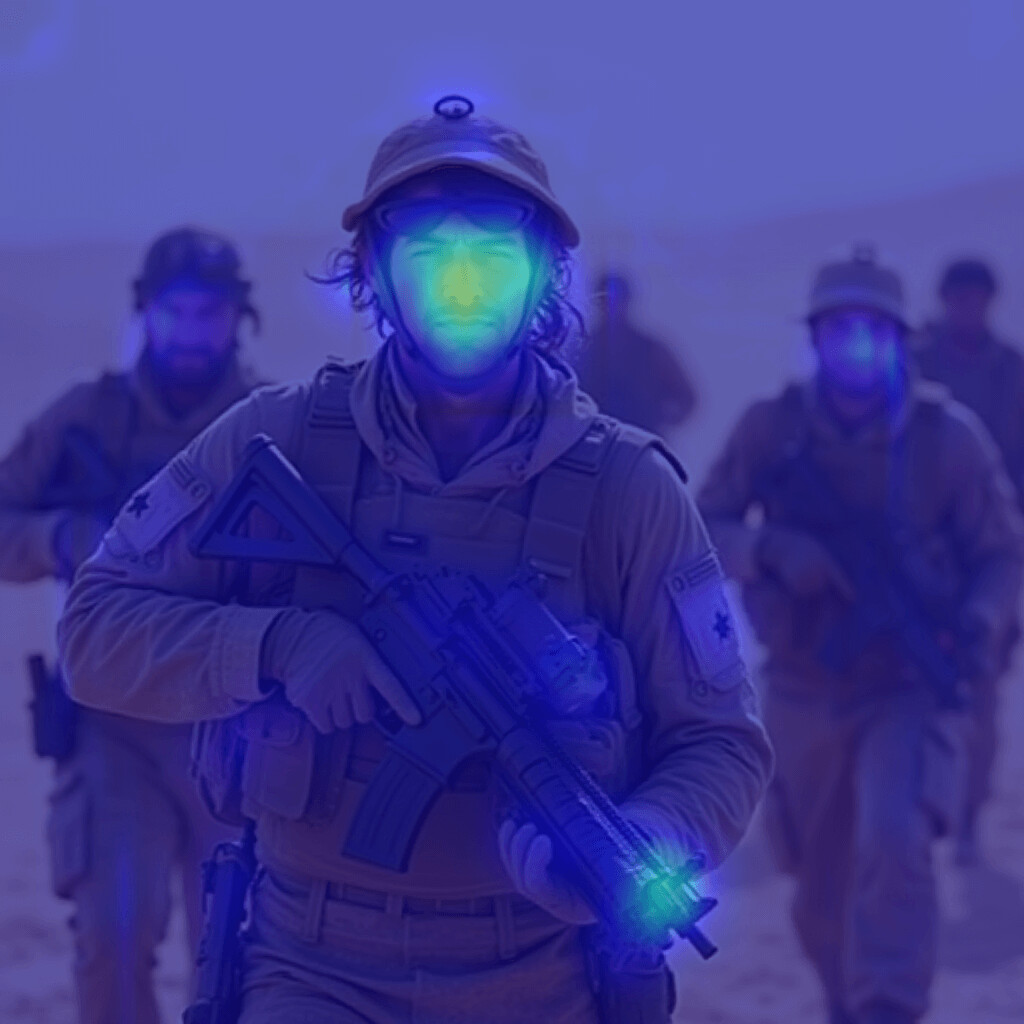} &
\includegraphics[width=0.10\textwidth]{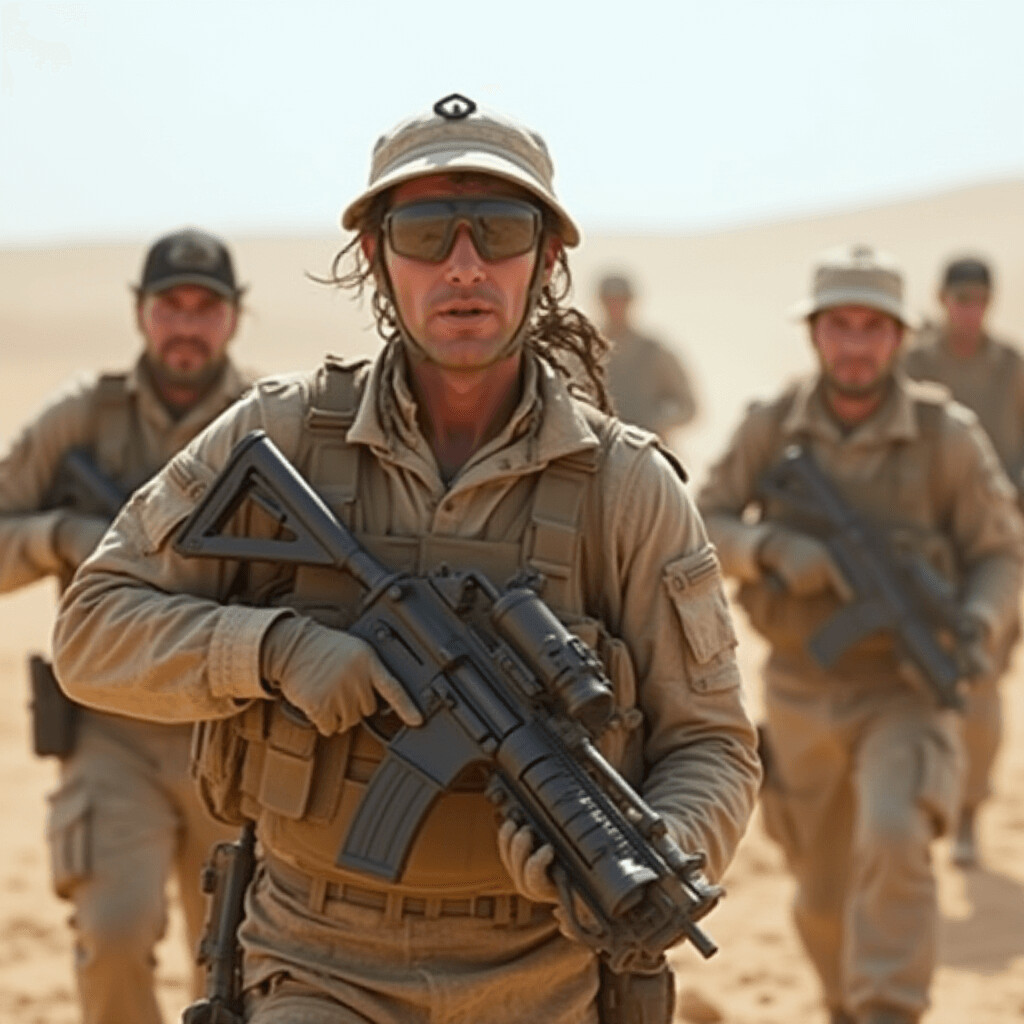} &
\includegraphics[width=0.10\textwidth]{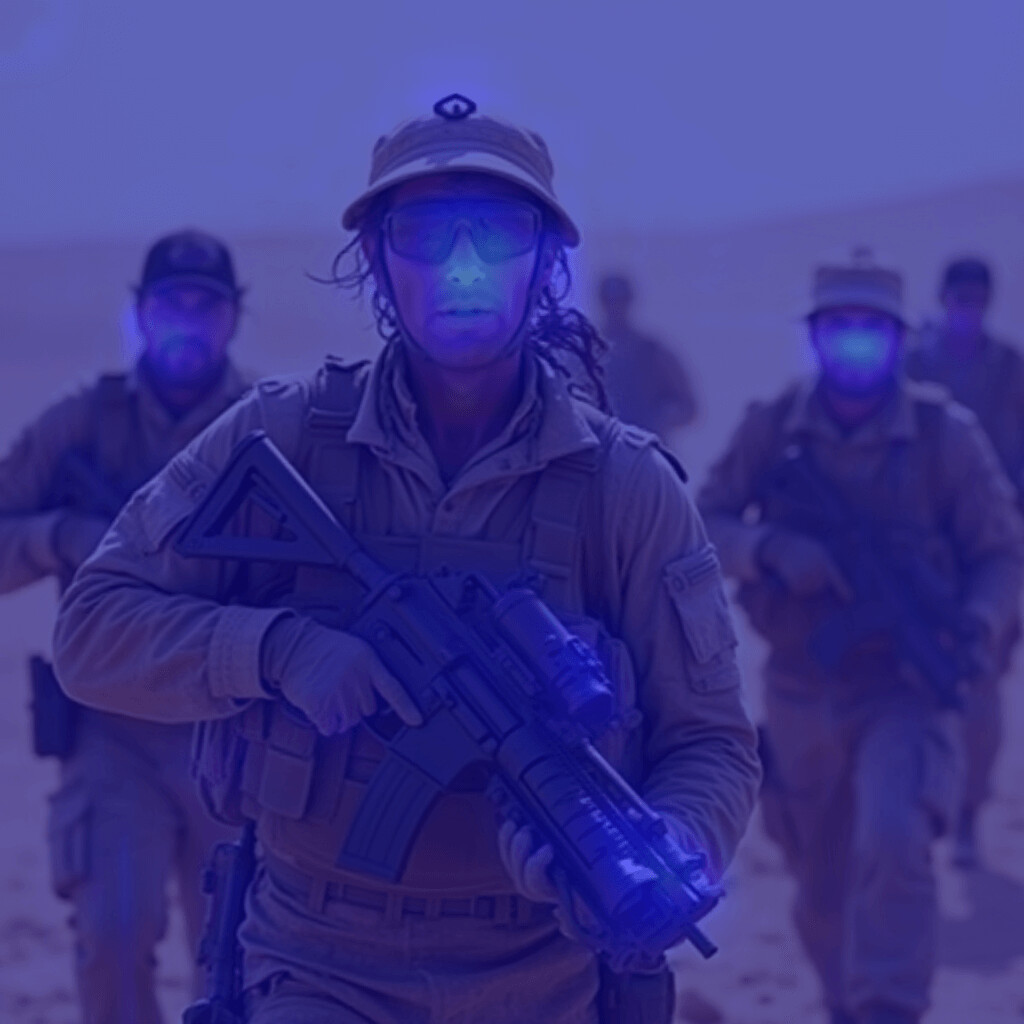} \\
\end{tabular}

\caption{\textbf{Additional examples qualitative comparison} of \our{} using different artifact detectors $\mathcal{AD}$ during trajectory refinement. The detector used is denoted by R - RichHF*, D - DiffDoctor and R + D, a average combination of the obtained masks. The overlay is calculated using the artifact detector shown in superscript, while the subscript indicates the detector used during trajectory correction, with the exception of base indicating the image generated using baseline method.}
\label{fig:diffdetectors}
\end{figure*}

\begin{table*}[h]
\setlength{\tabcolsep}{5.5pt}
\caption{\textbf{Ablation Study of Step-Wise MAE-Based Trajectory Regularization for \textit{people} dataset.} Results on FLUX.1 [dev] for different values of the parameter $\alpha$ for additional gradient scaled by the coefficient $\mathcal{L}_\text{rec}$  = $\alpha \lambda_{t}$. The best configuration is achieved at $\alpha=0.1$, as it keeps artifacts low while minimally affecting non-artifact regions, providing a good quality-preservation trade-off. 
}
{\fontsize{7.5pt}{11pt}\selectfont
\begin{tabular}{lccccccc}
\hline
$\alpha$ & CLIP-T $\uparrow$  & Mean Artifact Freq (\%)$\downarrow$ & ImageReward $\uparrow$ & Artfiact Pixel Ratio (\%) $\downarrow$ & MAE $\downarrow$  & MAE (A) $\downarrow$           & MAE (NA) $\downarrow$          \\ \hline
0.0     & 35.762 $\pm$ 0.101  & 15.500 $\pm$ 2.380                & 0.968 $\pm$ 0.034        & 0.068 $\pm$ 0.022            & 9.617 $\pm$ 0.240  & 25.764 $\pm$ 0.604       & 9.545 $\pm$ 0.240             \\
0.1     & 35.771 $\pm$ 0.181  & 20.000 $\pm$ 2.449 & 0.966 $\pm$ 0.024 & 0.078 $\pm$ 0.023 & 8.254 $\pm$ 0.169 & 23.690 $\pm$ 0.944 & 8.187 $\pm$ 0.168\\
0.3     & 35.751 $\pm$ 0.053  & 30.750 $\pm$ 1.708 & 0.969 $\pm$ 0.026 & 0.119 $\pm$ 0.026 & 8.653 $\pm$ 0.057 & 23.755 $\pm$ 0.859 & 8.585 $\pm$ 0.061 \\
0.5     & 35.800 $\pm$ 0.093  & 40.000 $\pm$ 5.598 & 0.959 $\pm$ 0.027 & 0.155 $\pm$ 0.035 & 9.325 $\pm$ 0.034 & 24.891 $\pm$ 0.848 & 9.256 $\pm$ 0.040\\
0.7     & 35.872 $\pm$ 0.077  & 47.250 $\pm$ 3.403 & 0.961 $\pm$ 0.030 & 0.196 $\pm$ 0.026 & 10.058 $\pm$ 0.126 & 26.429 $\pm$ 1.289 & 9.986 $\pm$ 0.332\\
1.0     & 35.873 $\pm$ 0.072 & 49.500 $\pm$ 3.697  & 0.947 $\pm$ 0.034 & 0.240 $\pm$ 0.021 & 11.063 $\pm$ 0.184 & 27.862 $\pm$ 0.670 & 10.986 $\pm$ 0.194\\ \hline
\end{tabular}
\label{tab:mae_people}
}
\end{table*}

\begin{figure*}[h!]
\centering
\setlength{\tabcolsep}{3pt}
\renewcommand{\arraystretch}{1.0}
\begin{tabular}{ccccccc}
FLUX.1 [dev] & $\alpha = 0$ & $\alpha = 0.1$ & $\alpha = 0.3$ & $\alpha = 0.5$ & $\alpha = 0.7$ & $\alpha = 1$\\
\includegraphics[width=0.13\textwidth, height=0.13\textwidth]{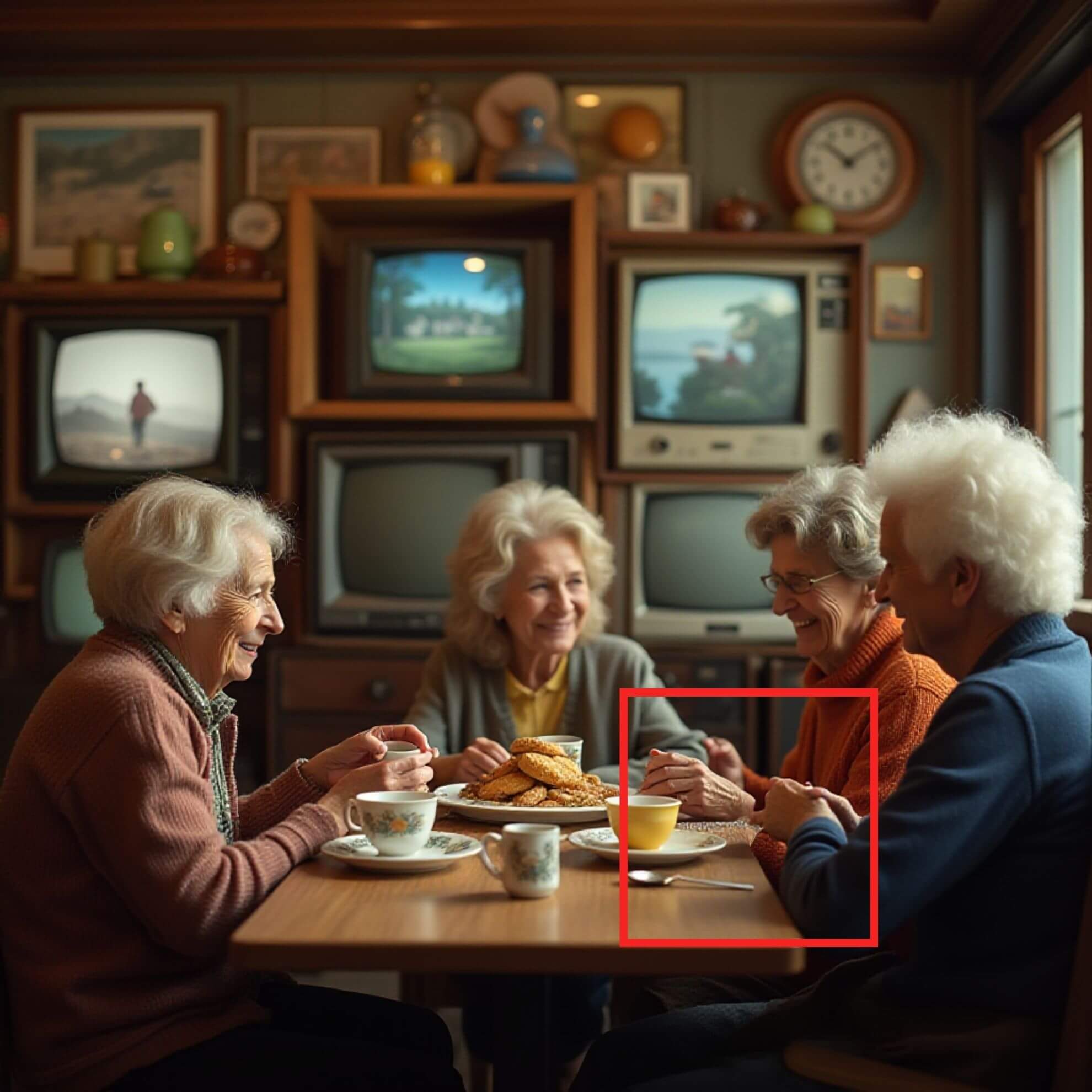} & 
\includegraphics[width=0.13\textwidth, height=0.13\textwidth]{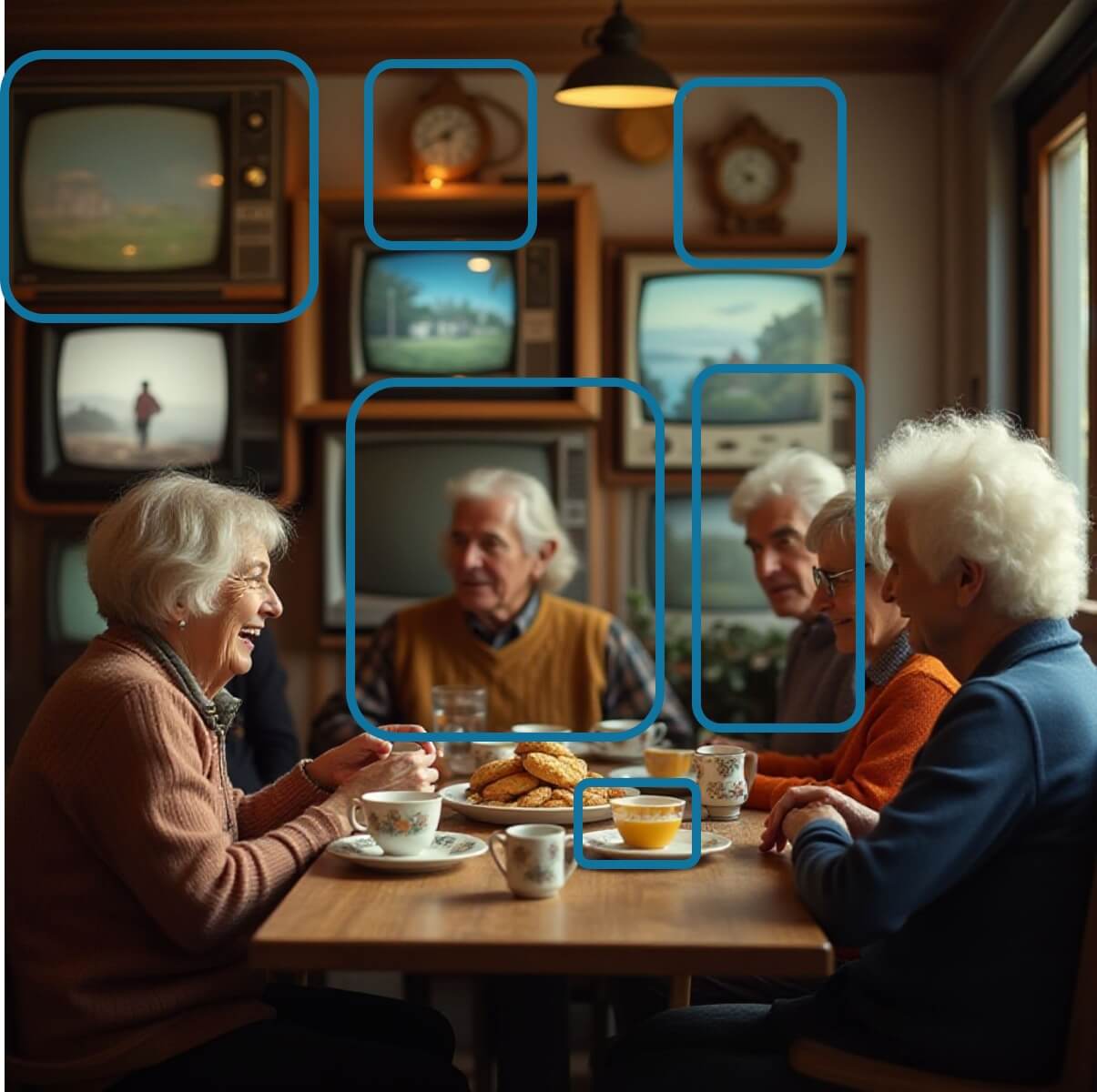} &
\includegraphics[width=0.13\textwidth, height=0.13\textwidth]{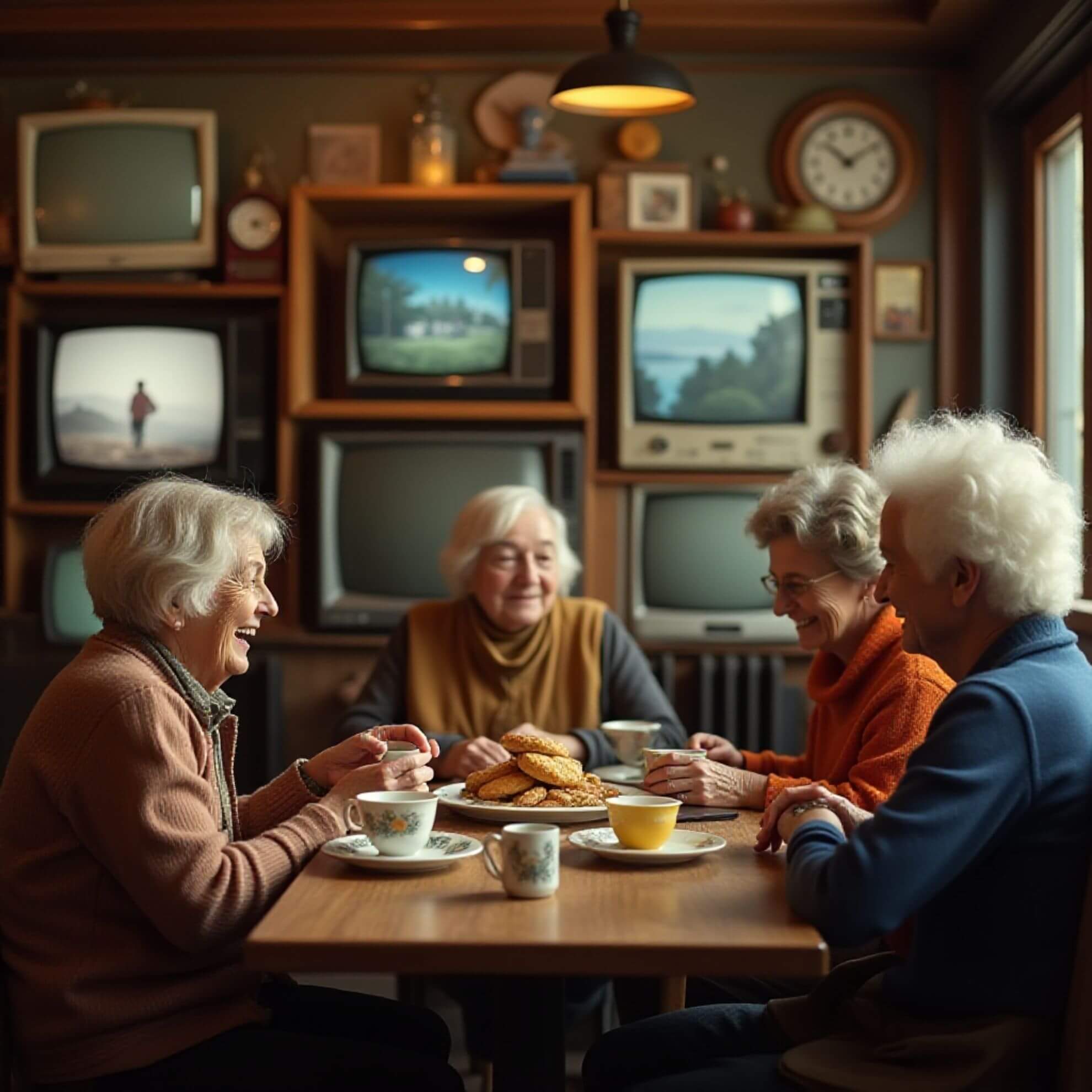} &
\includegraphics[width=0.13\textwidth, height=0.13\textwidth]{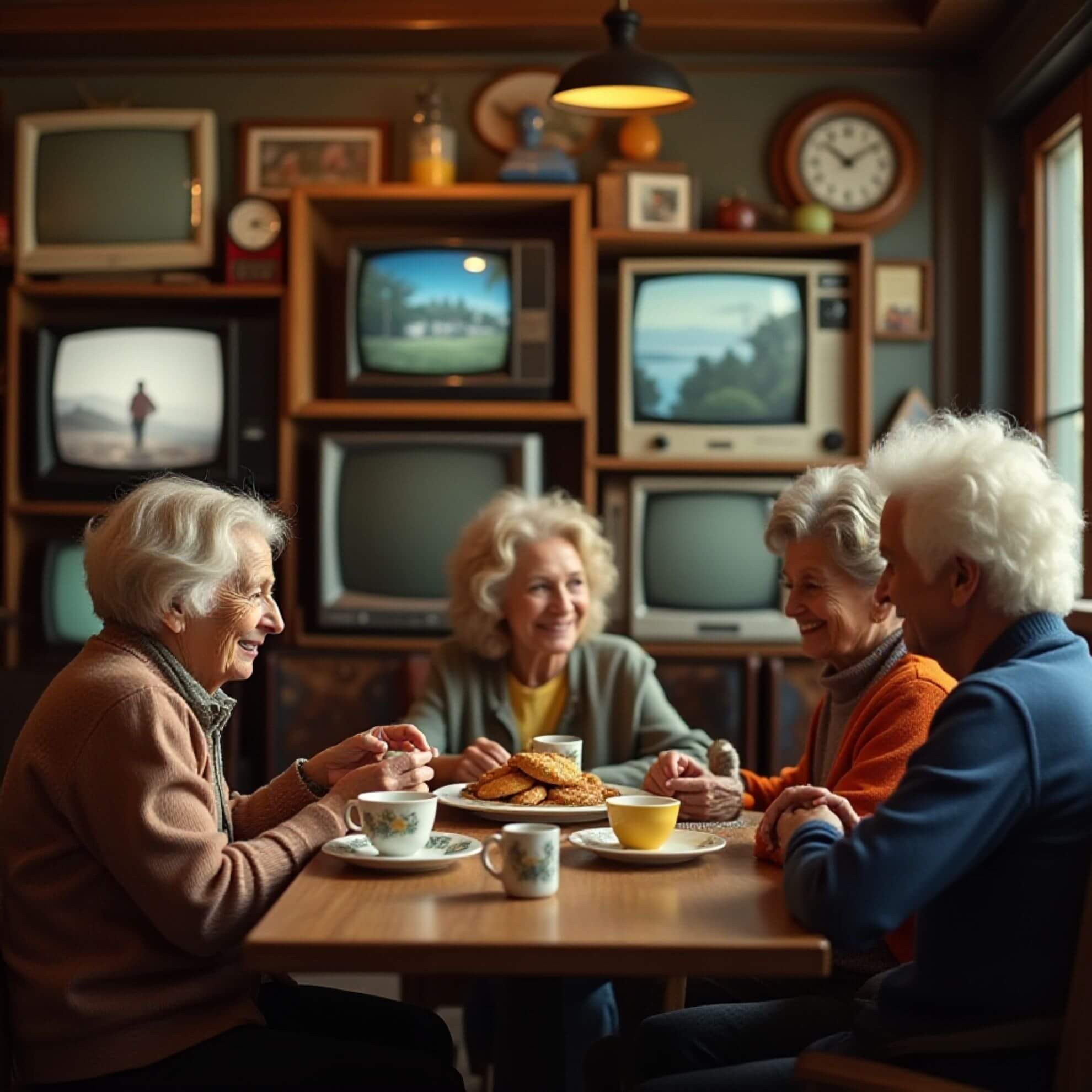} 
& 
\includegraphics[width=0.13\textwidth, height=0.13\textwidth]{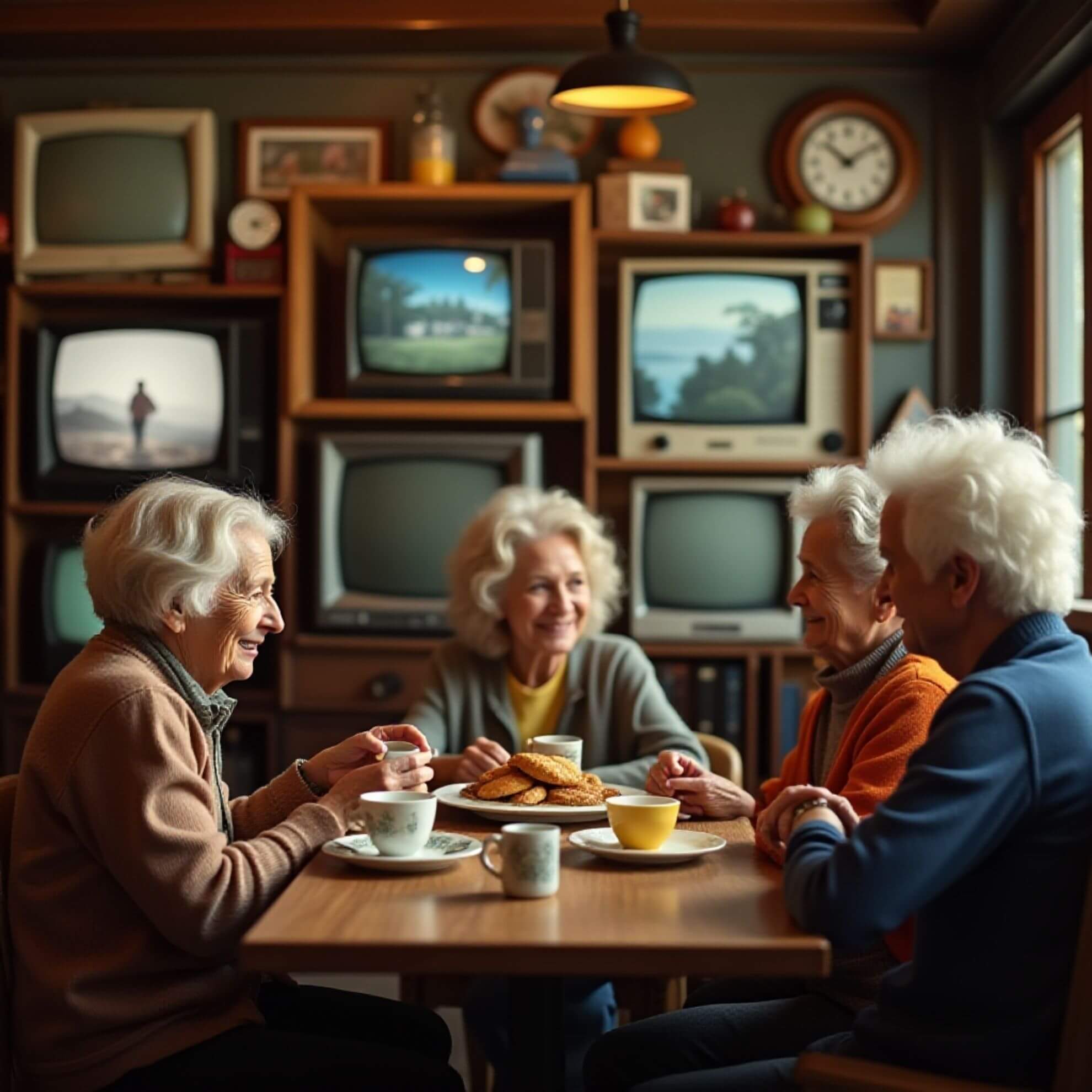} &
\includegraphics[width=0.13\textwidth, height=0.13\textwidth]{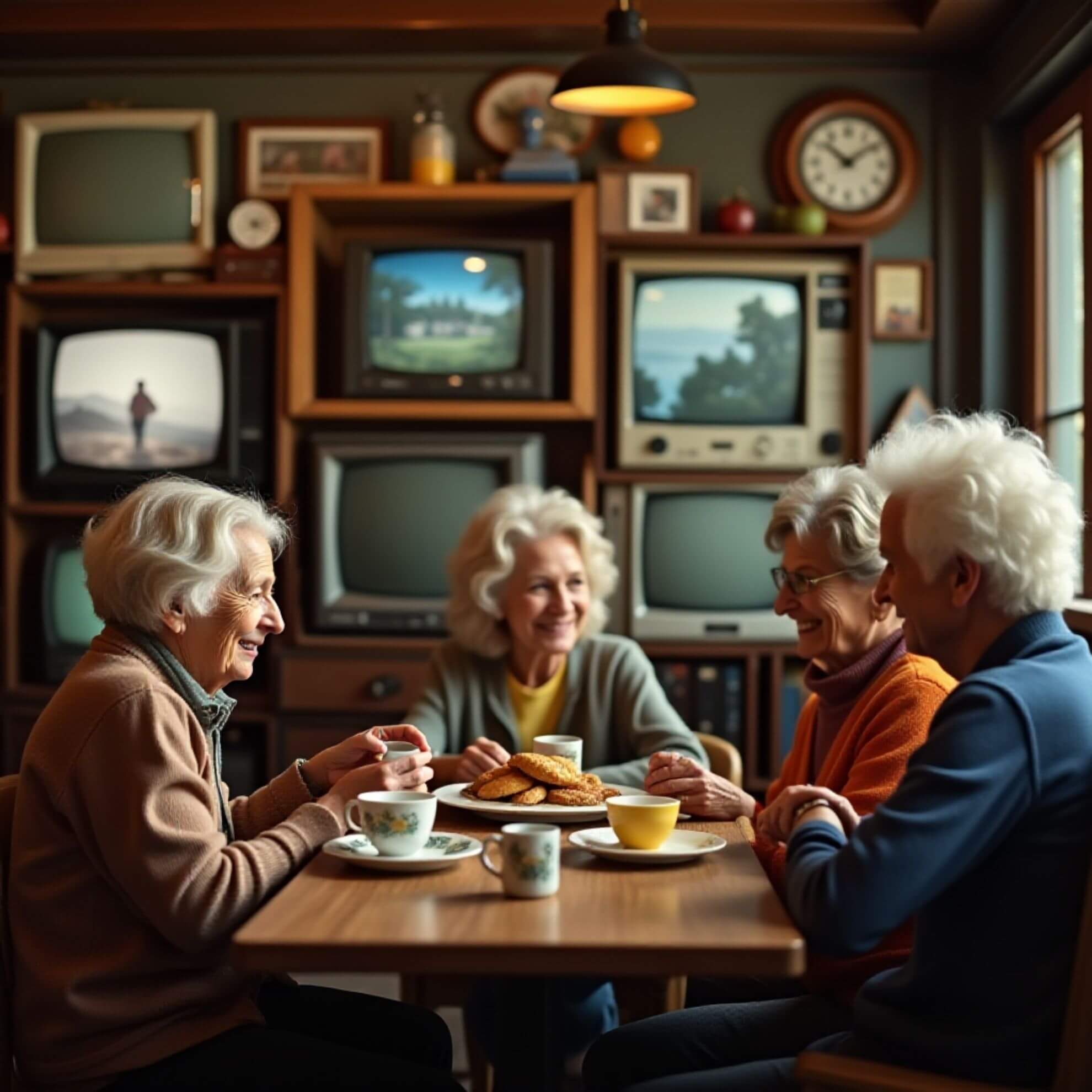} &
\includegraphics[width=0.13\textwidth, height=0.13\textwidth]{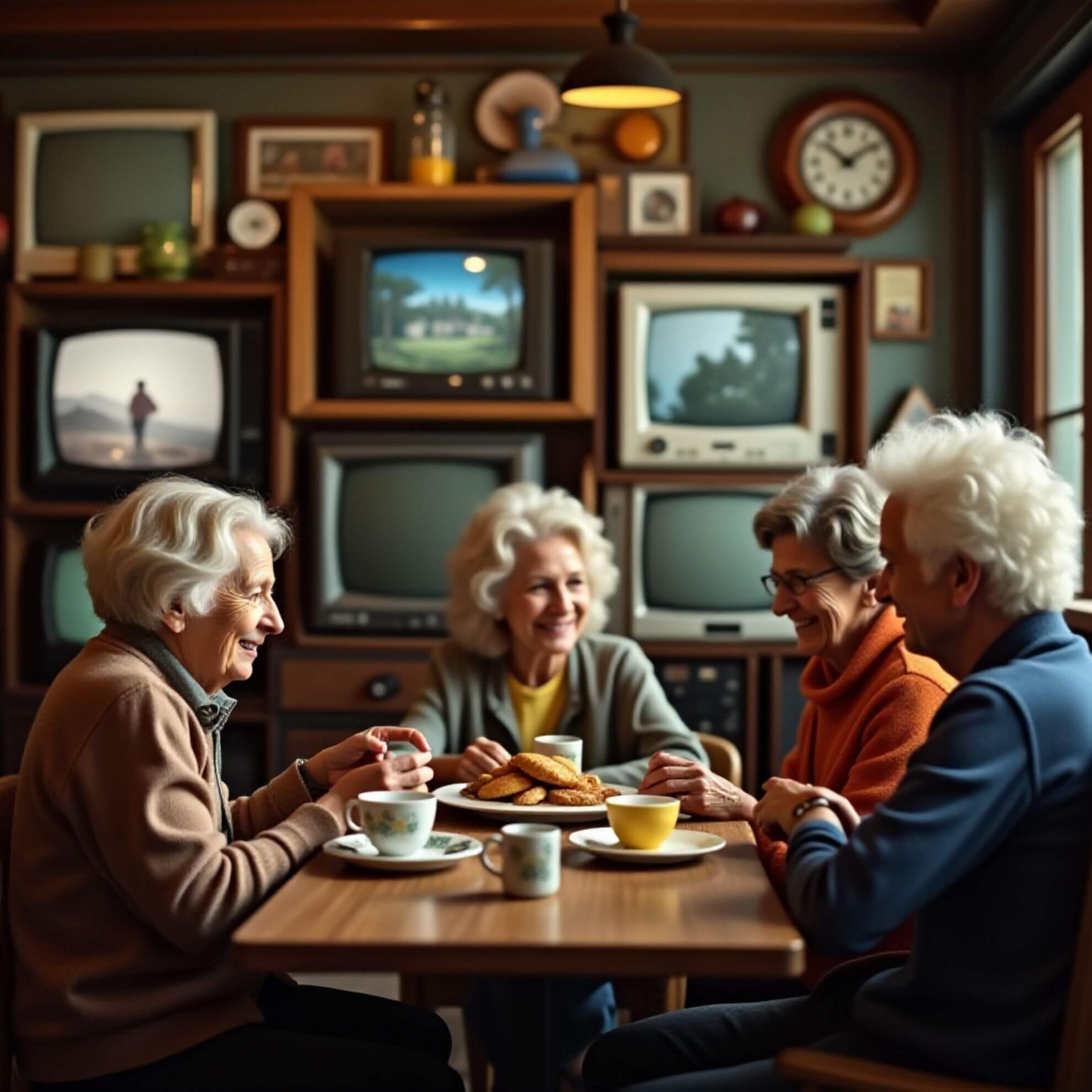} \\ 

\includegraphics[width=0.13\textwidth, height=0.13\textwidth]{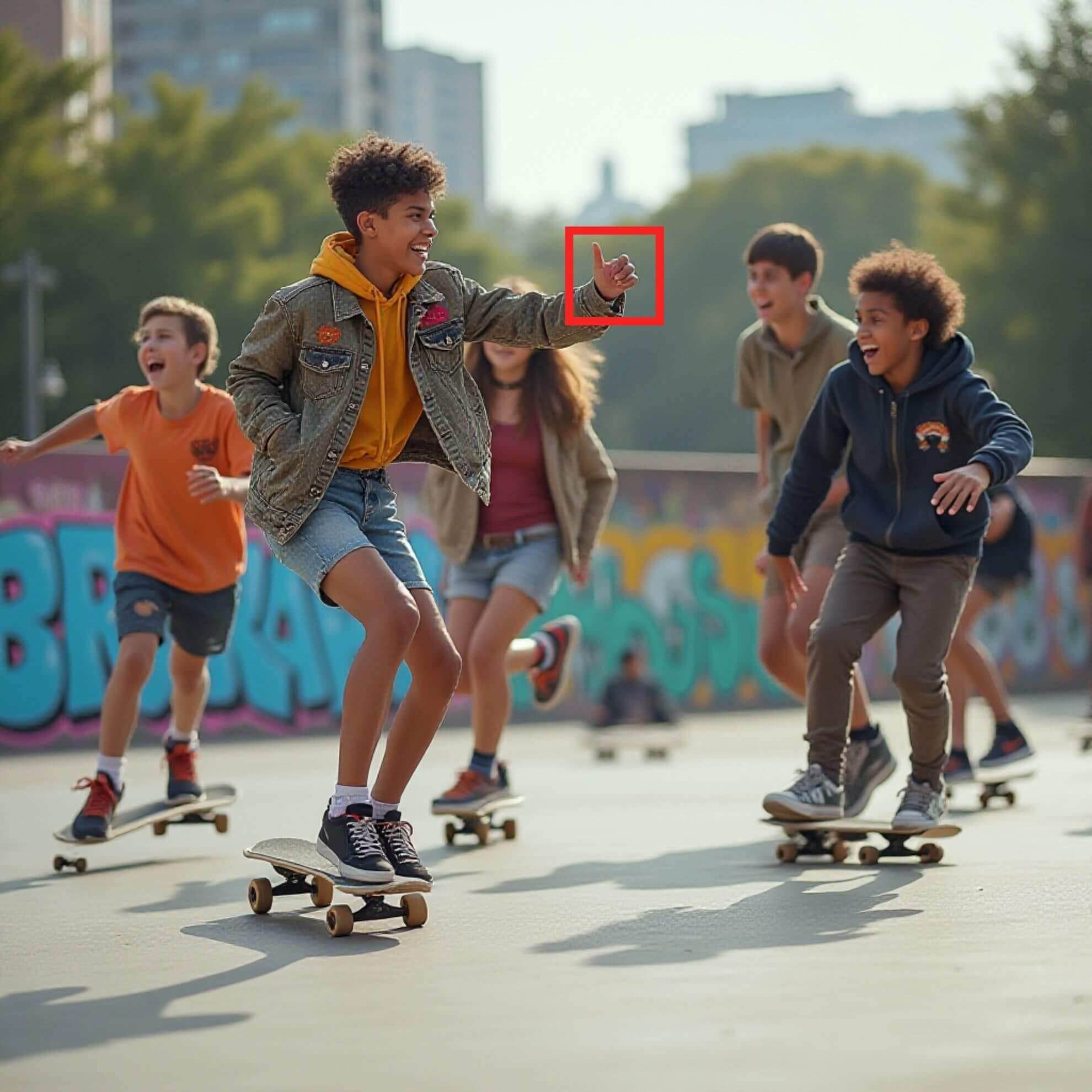} & 
\includegraphics[width=0.13\textwidth, height=0.13\textwidth]{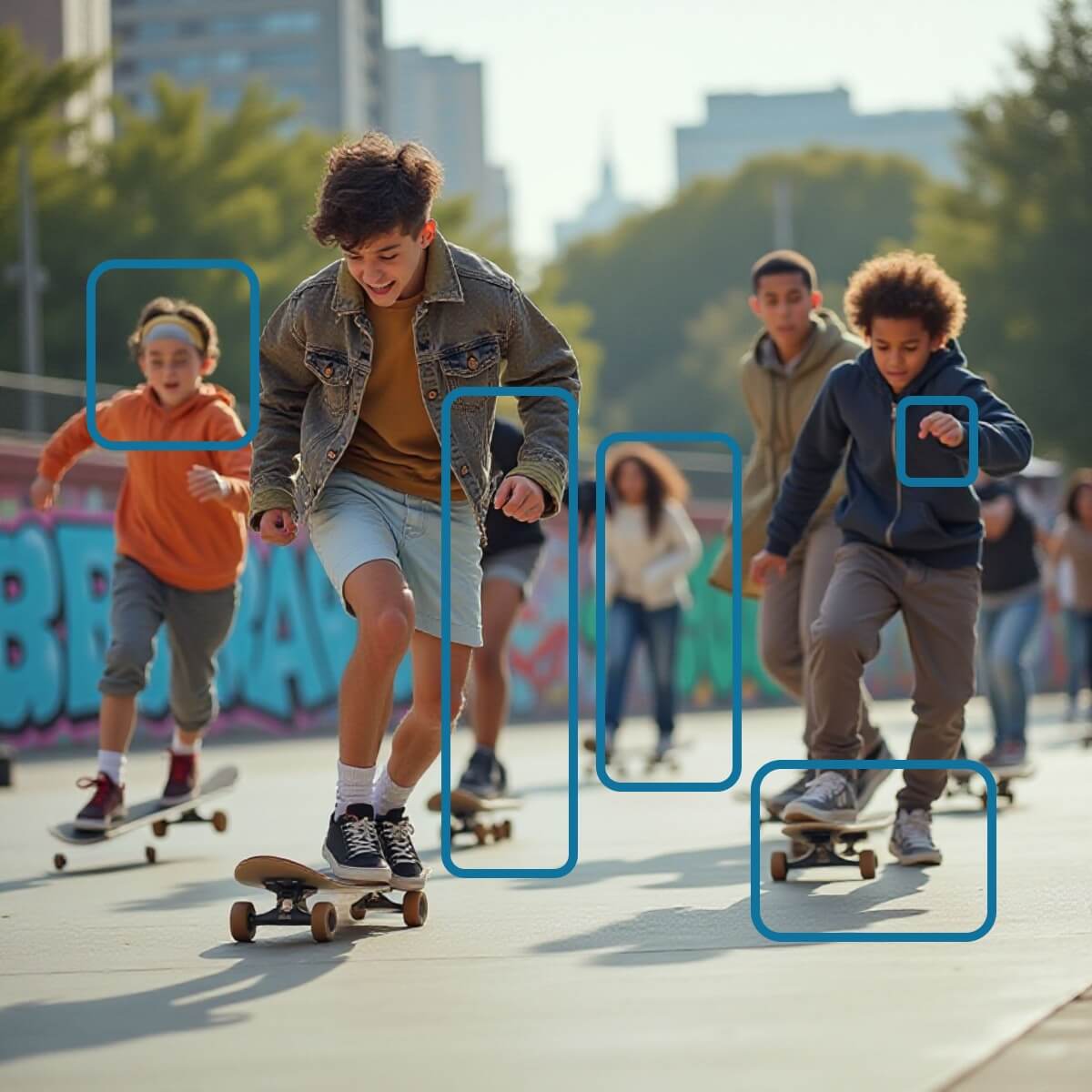} &
\includegraphics[width=0.13\textwidth, height=0.13\textwidth]{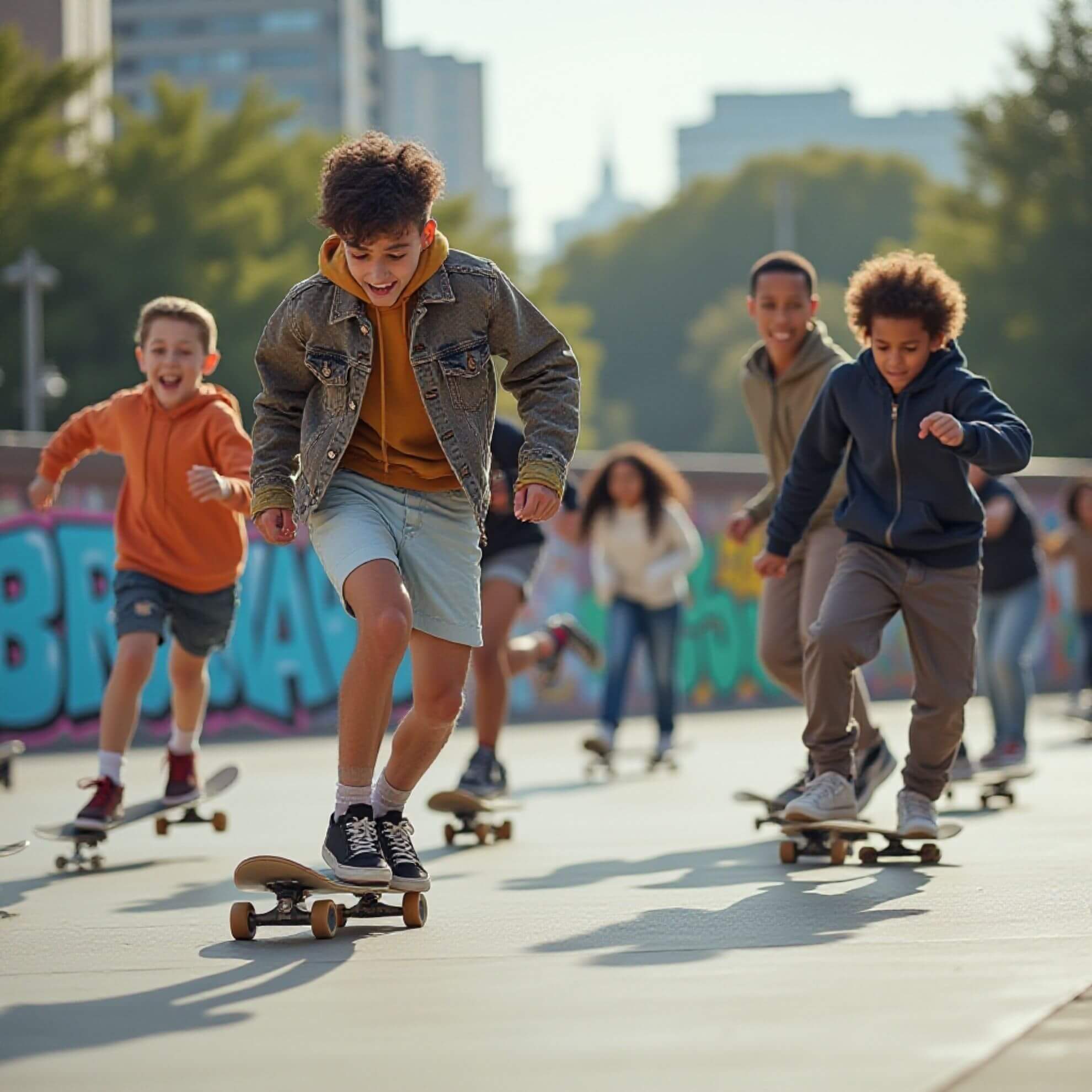} &
\includegraphics[width=0.13\textwidth, height=0.13\textwidth]{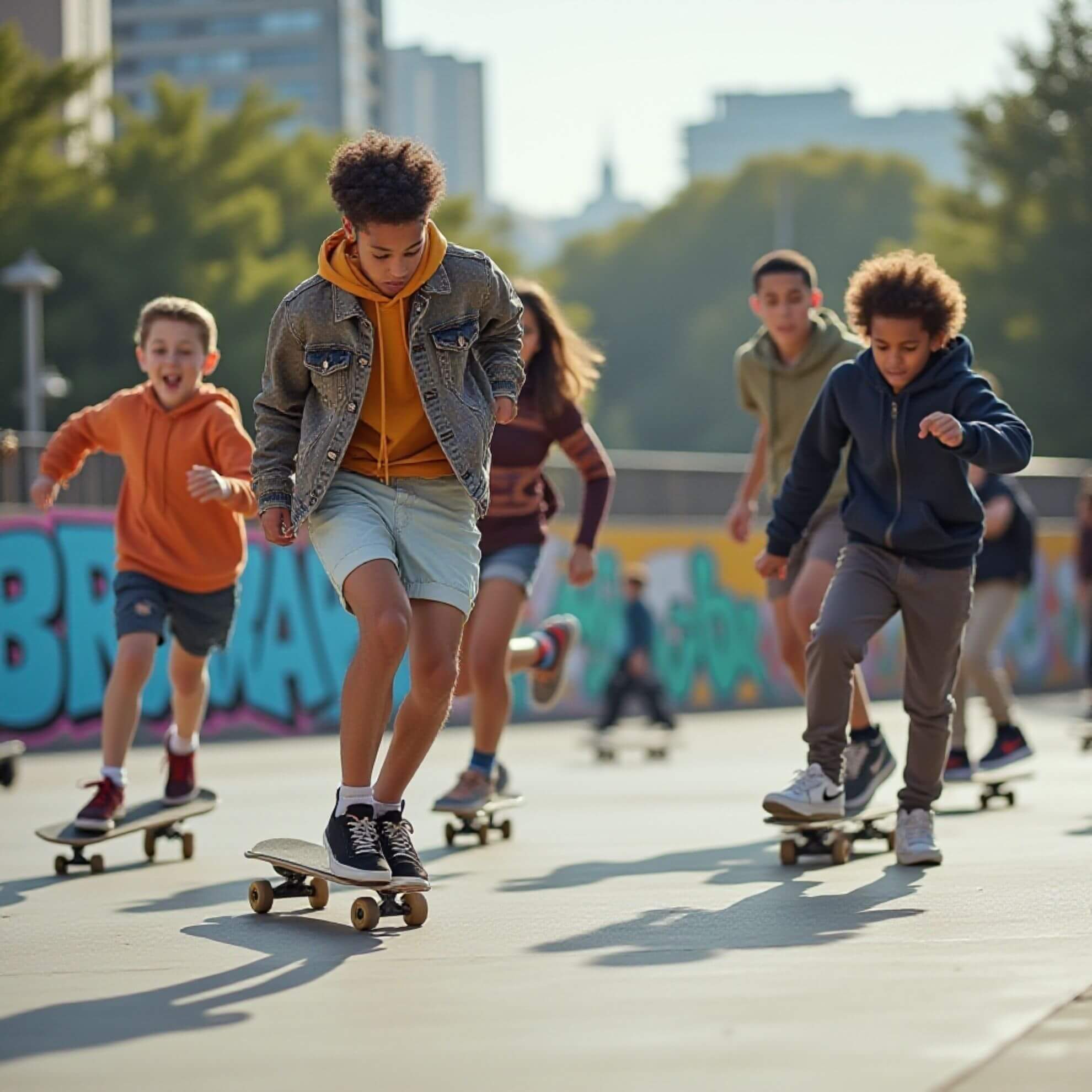} 
& 
\includegraphics[width=0.13\textwidth, height=0.13\textwidth]{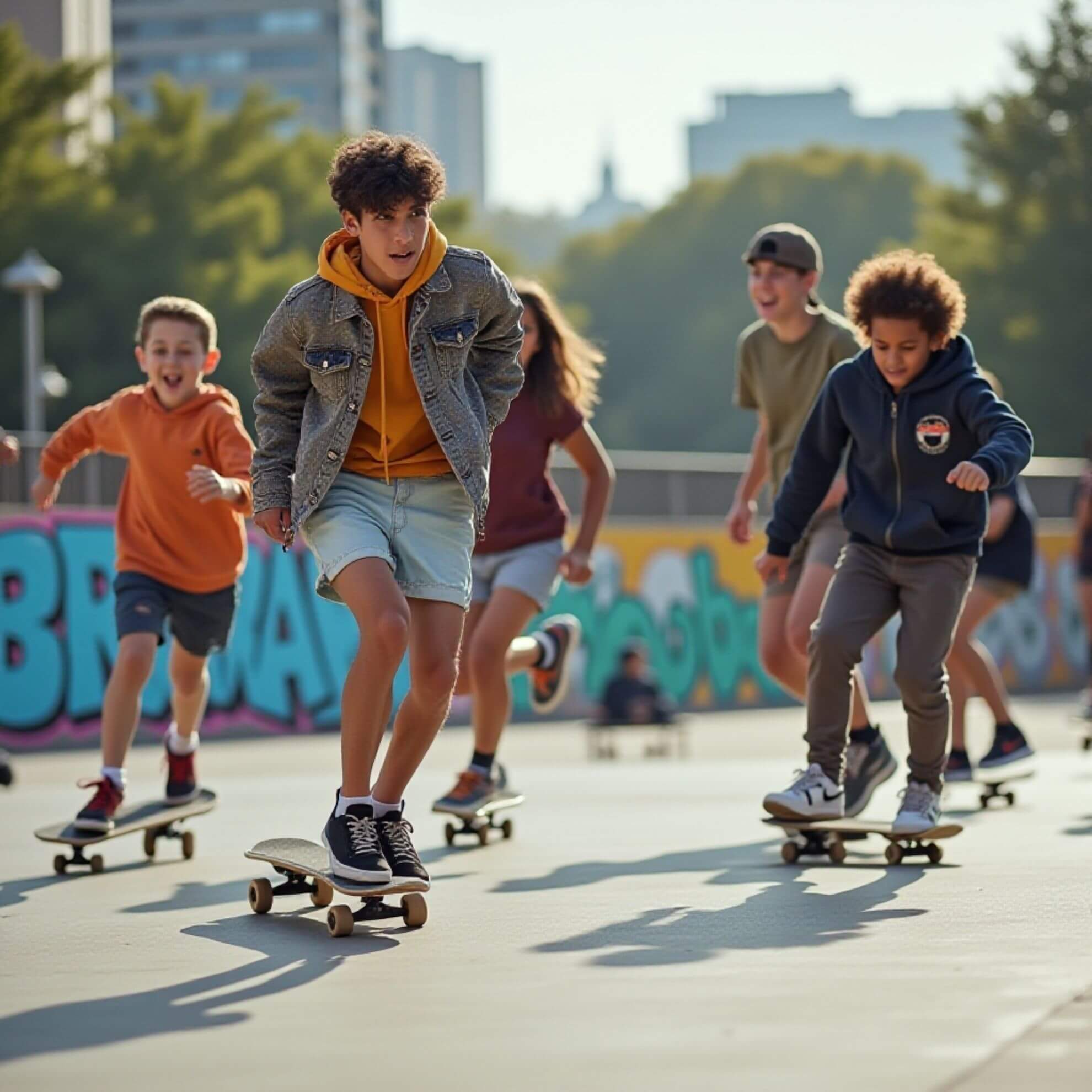} &
\includegraphics[width=0.13\textwidth, height=0.13\textwidth]{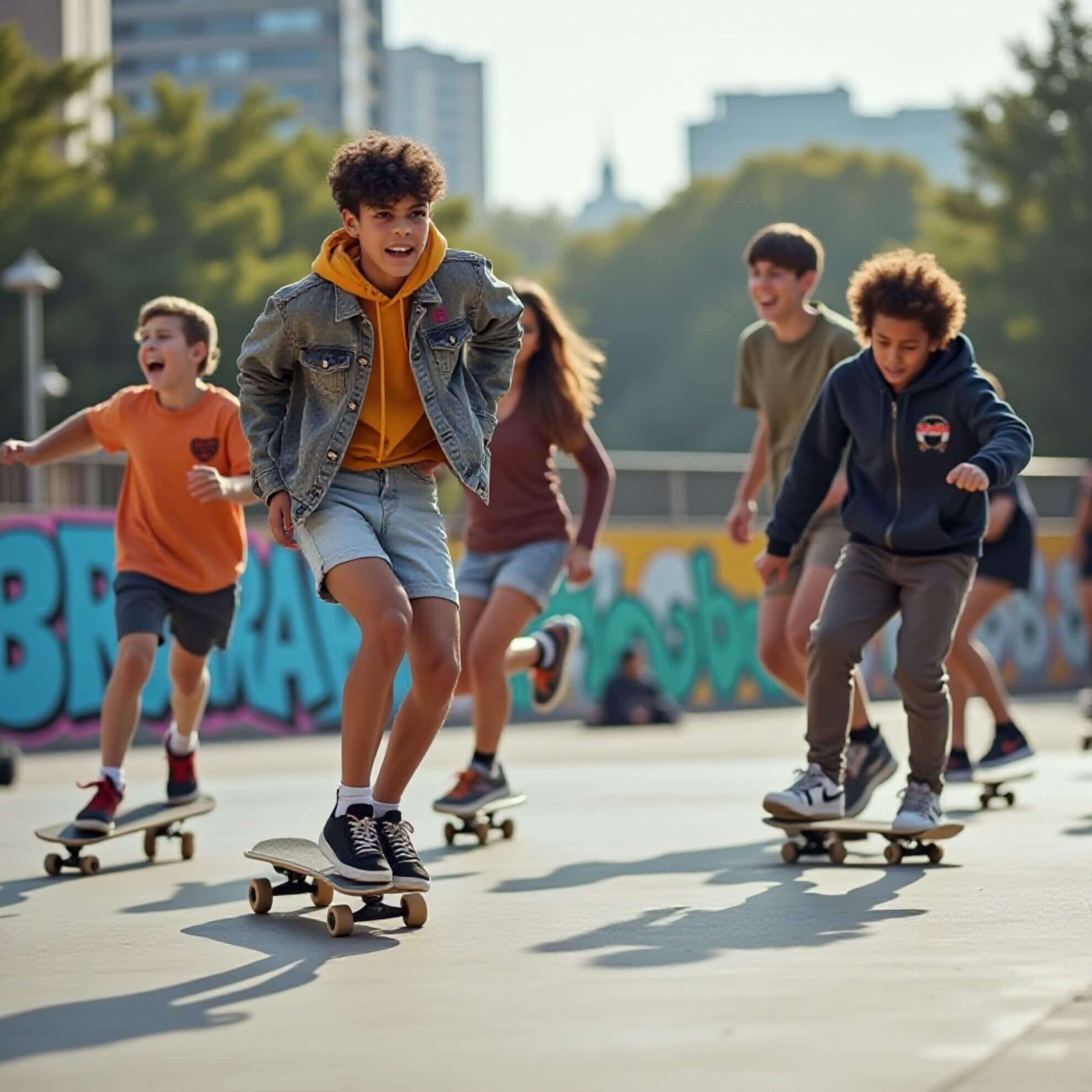} &
\includegraphics[width=0.13\textwidth, height=0.13\textwidth]{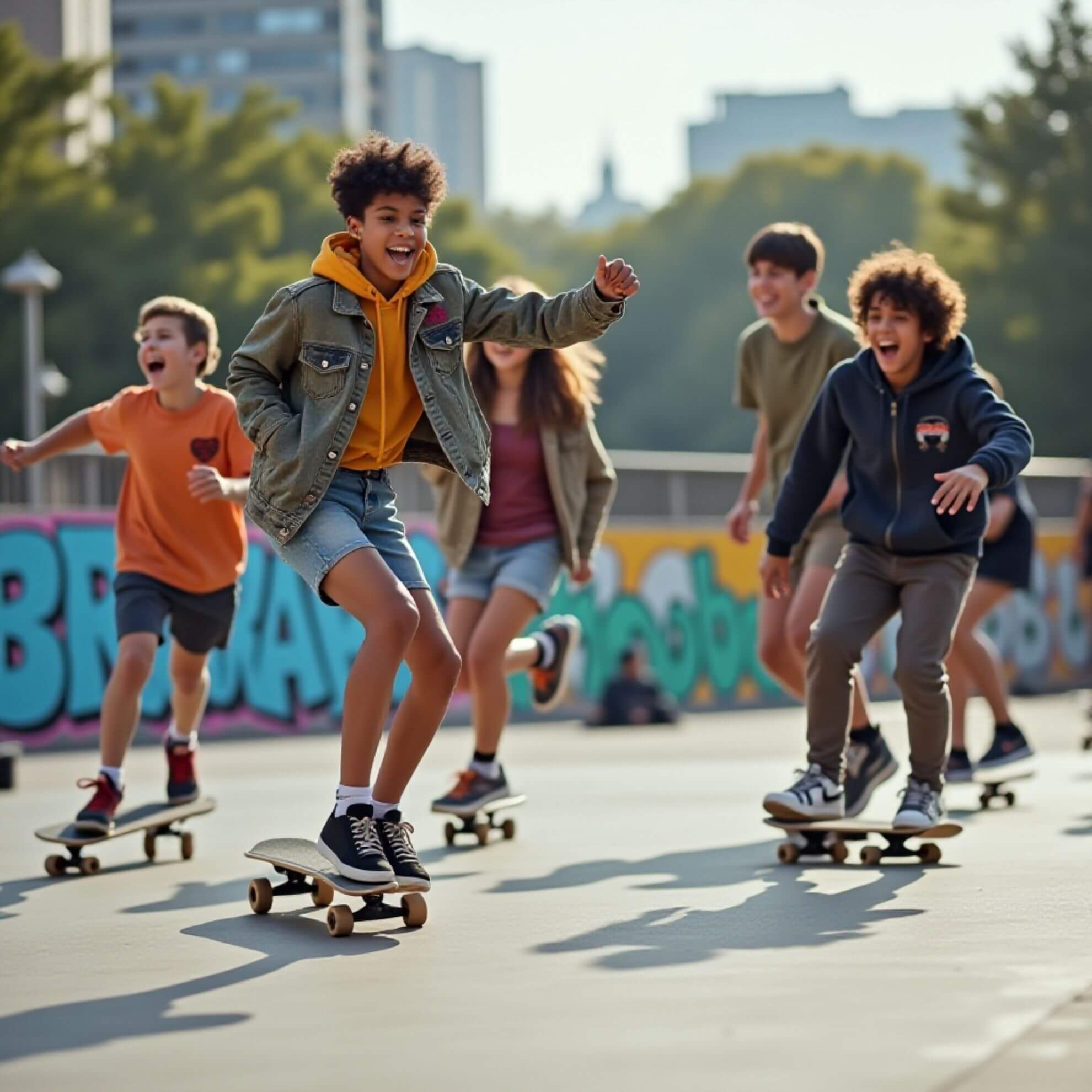} \\
\end{tabular}
\caption{\textbf{Visual comparison illustrating the effect of the $\alpha$ parameter.}
When $\alpha=0$, the method may in some cases modify regions that were not identified as artifacts; representative examples are highlighted in blue. Increasing $\alpha$ introduces regularization that brings the output closer to the original baseline (FLUX.1 [dev]) while preserving visual similarity. For example, in the first row, the woman seated at the center of the table and the correct number of people are better preserved, and in the second row with the boy’s raised arm is maintained. }
\label{fig:lambdamaeablation}
\vspace{-0.7cm}
\end{figure*}

\subsubsection{Base model identity preservation - extended results}
\label{subsubsec:mae}

An ablation study of the effect of adding the MAE coefficient is presented in Tab. \ref{tab:mae_people} to the loss function with value $\alpha$. The corresponding visual results are shown in Fig. \ref{fig:lambdamaeablation}. For low values of the $\alpha$ parameter, we observe a decrease in the MAE (NA) value, indicating a better match of pixels to the non-artifactual region, which should remain as little changed as possible from the original image. For high $\alpha$ ($\alpha>=0.7$), further correction leads to an increase in image contrast. This effect should not be considered a deterioration in the correction quality, but rather a consequence of the high sensitivity of the MAE metric to changes in pixel intensity, without taking into account the image's semantic context. We observe an increase in the CLIP-T metric, indicating better compliance of the generated images with the prompts, as the model is directed towards the target state from the beginning of the trajectory. Ultimately, the model better reflects the semantic content of the image compared to the baseline image, despite the larger difference in pixels. Fig.~\ref{fig:lambdamaeablation} shows that high $\alpha$ are characterized by higher contrast but allow for less modification of areas not marked by the artifact detector. The introduced approach allows for early targeting of the generation trajectory, but it should be emphasized that there is a trade-off between artifact reduction and faithful representation of the scene. Therefore, it is important to adjust the appropriate value of this parameter $\alpha$ to prevent artifacts from reappearing.

\label{subsec:text_art}
\begin{figure*}[!h]
\centering
\setlength{\tabcolsep}{1.2pt}
\renewcommand{\arraystretch}{0.9}
\begin{tabular}{ccccc}
FLUX.1 [dev] & +DiffDoctor & +HPSv2 & +\our{} \\
\includegraphics[width=0.19\textwidth]{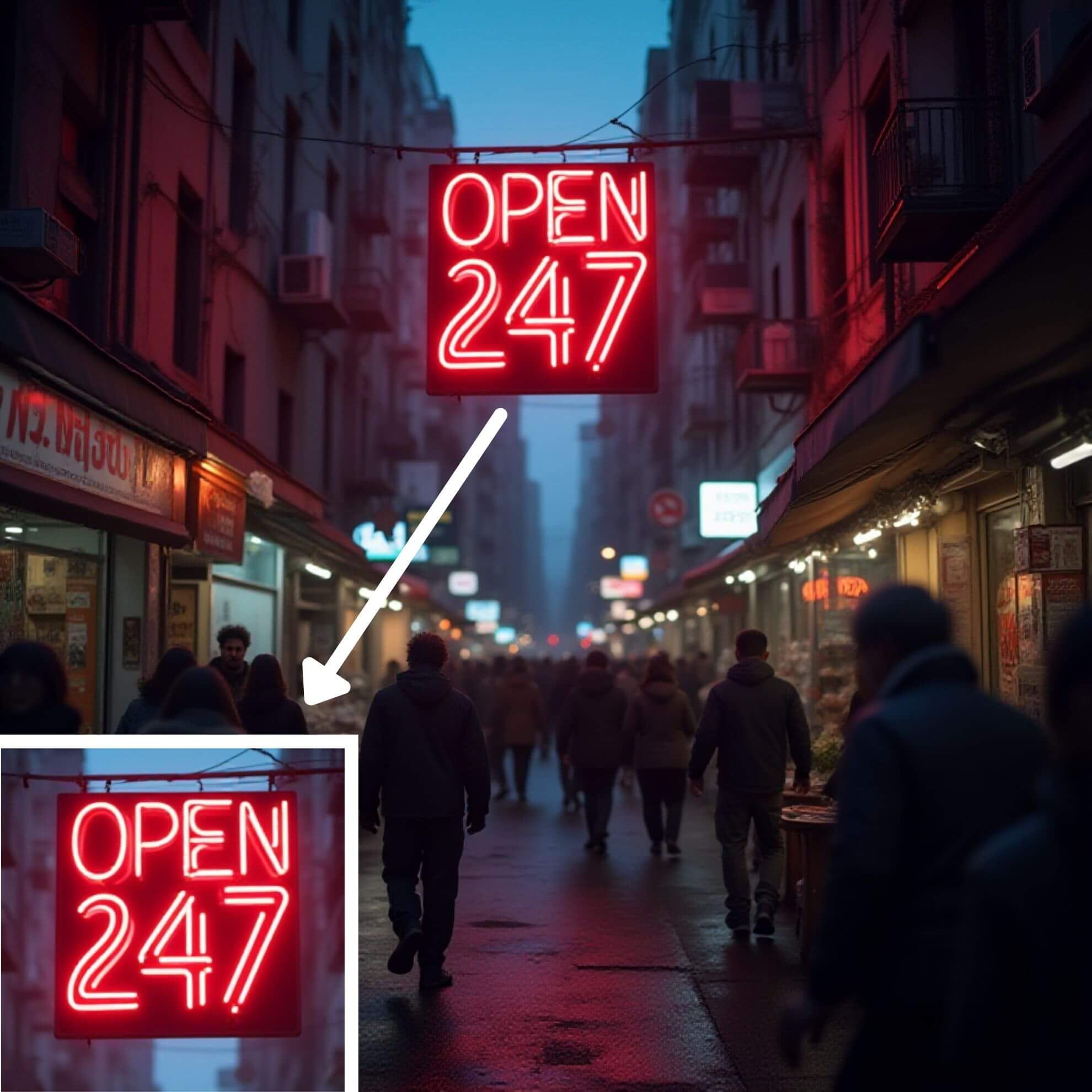} &
\includegraphics[width=0.19\textwidth]{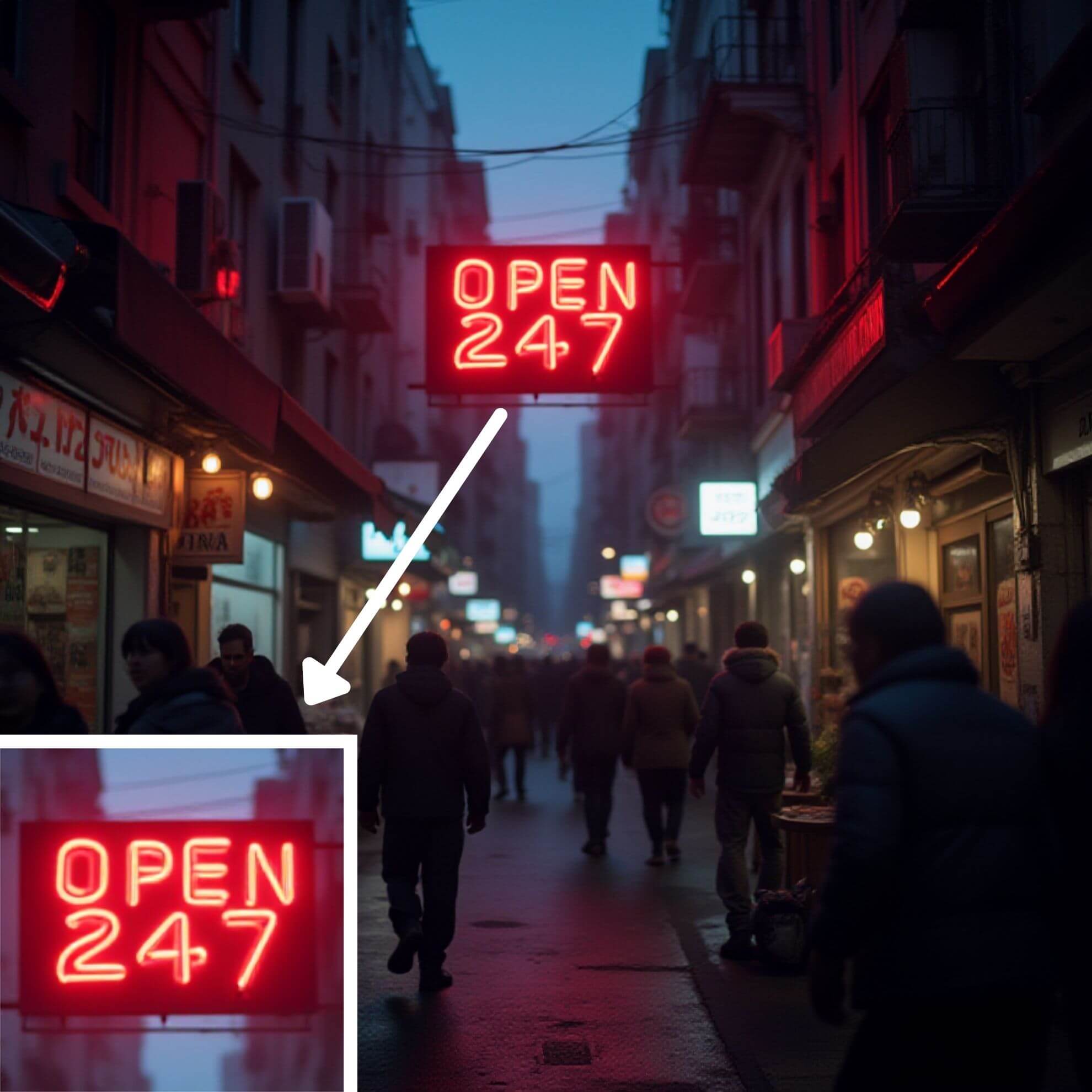} &
\includegraphics[width=0.19\textwidth]{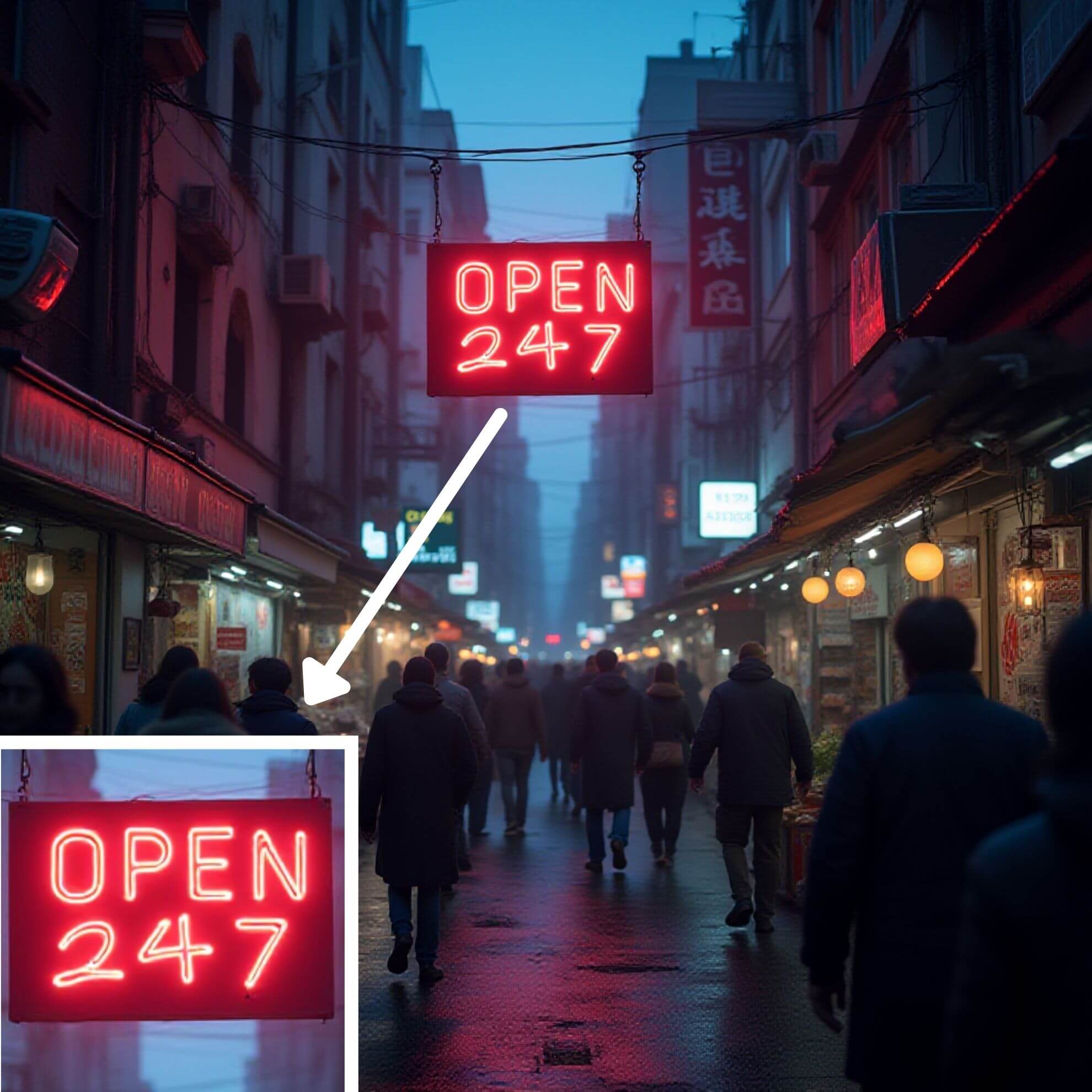} &
\includegraphics[width=0.19\textwidth]{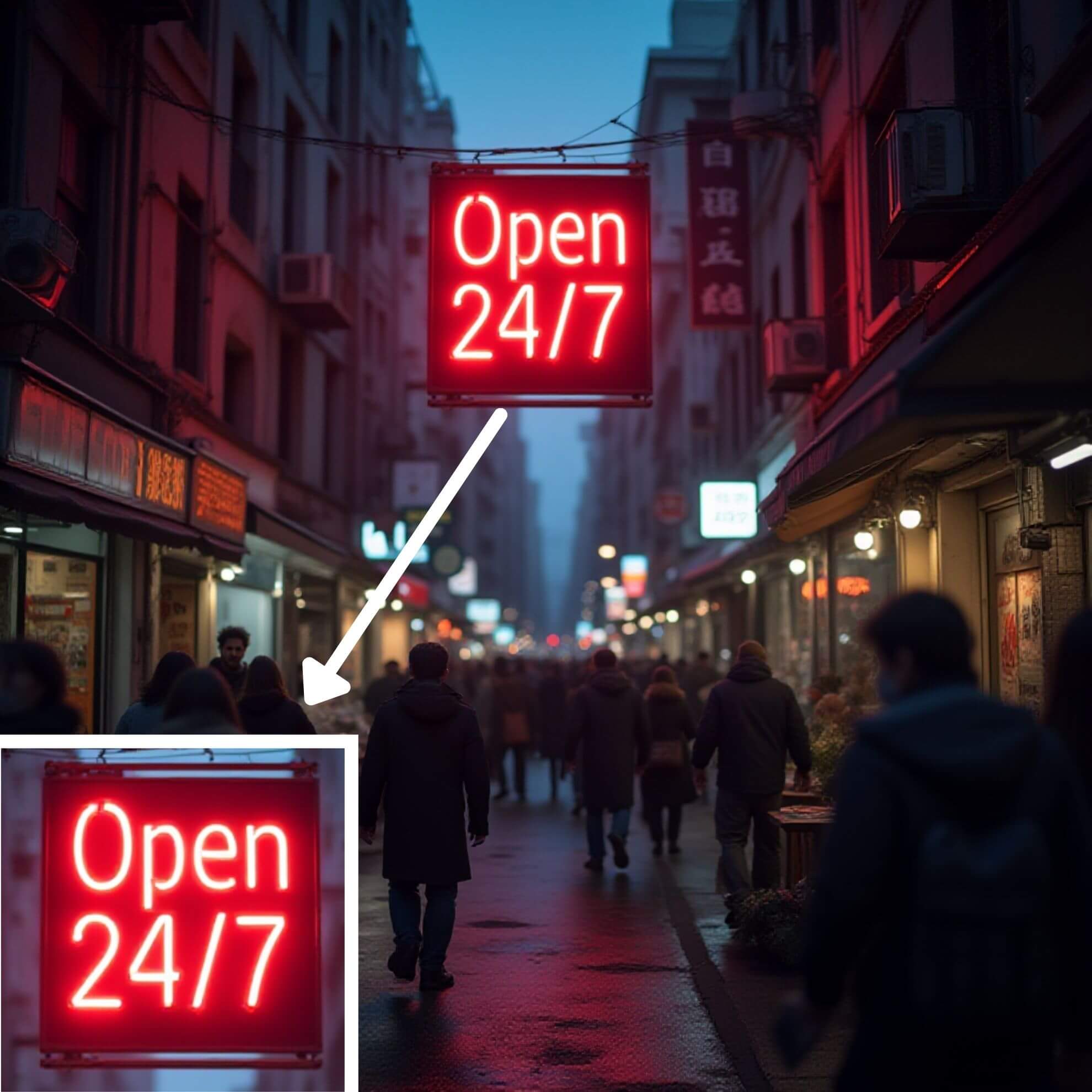} \\

\includegraphics[width=0.19\textwidth]{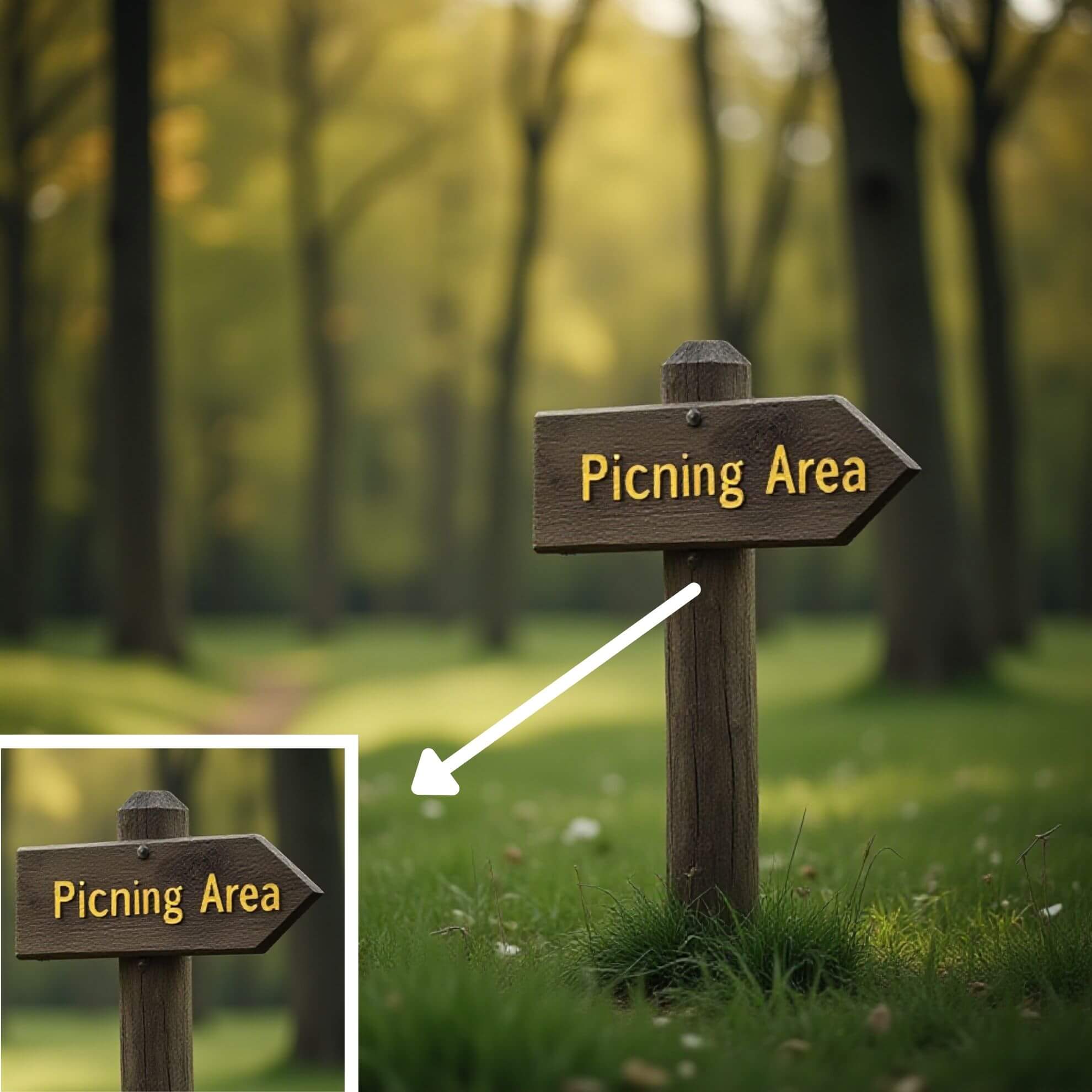} &
\includegraphics[width=0.19\textwidth]{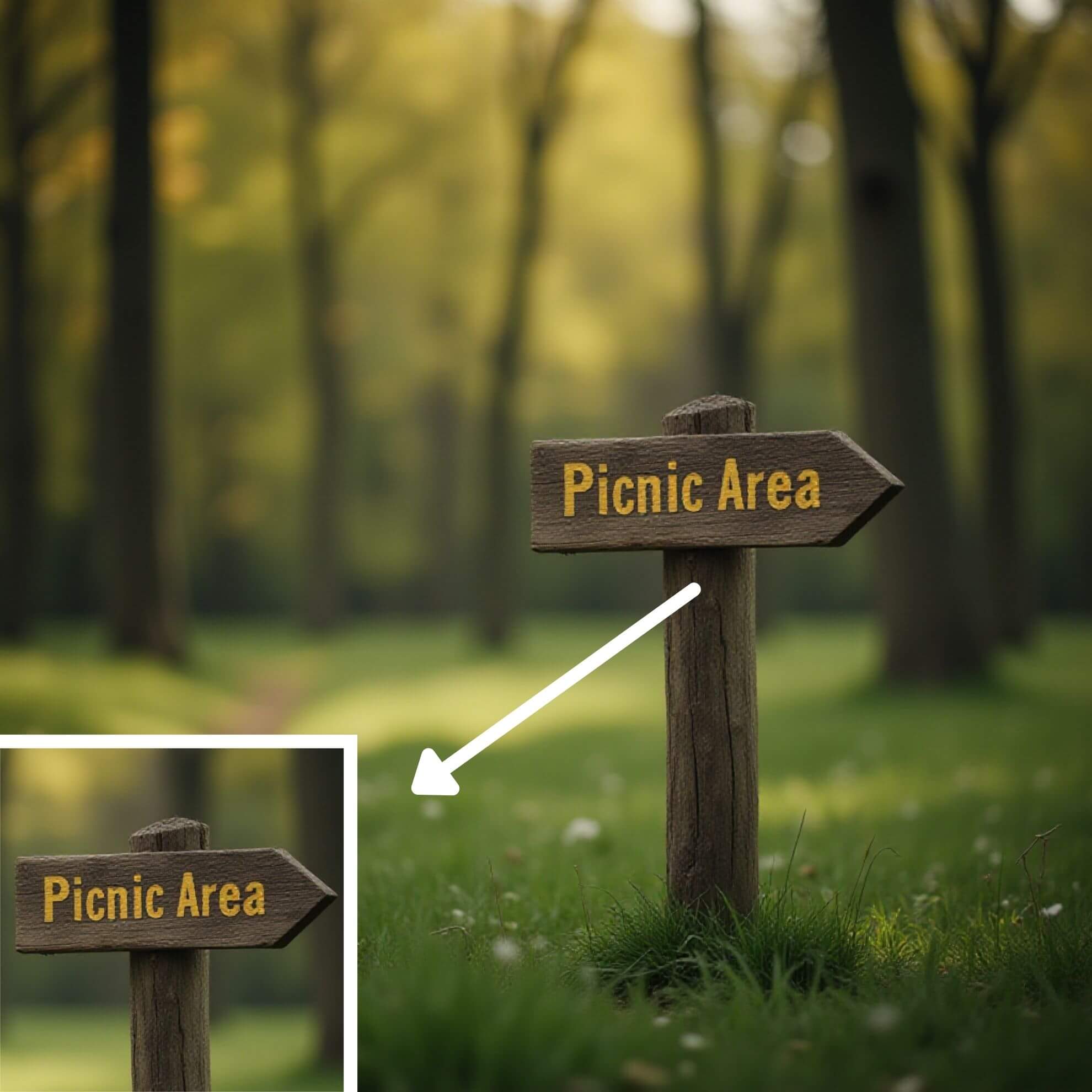} &
\includegraphics[width=0.19\textwidth]{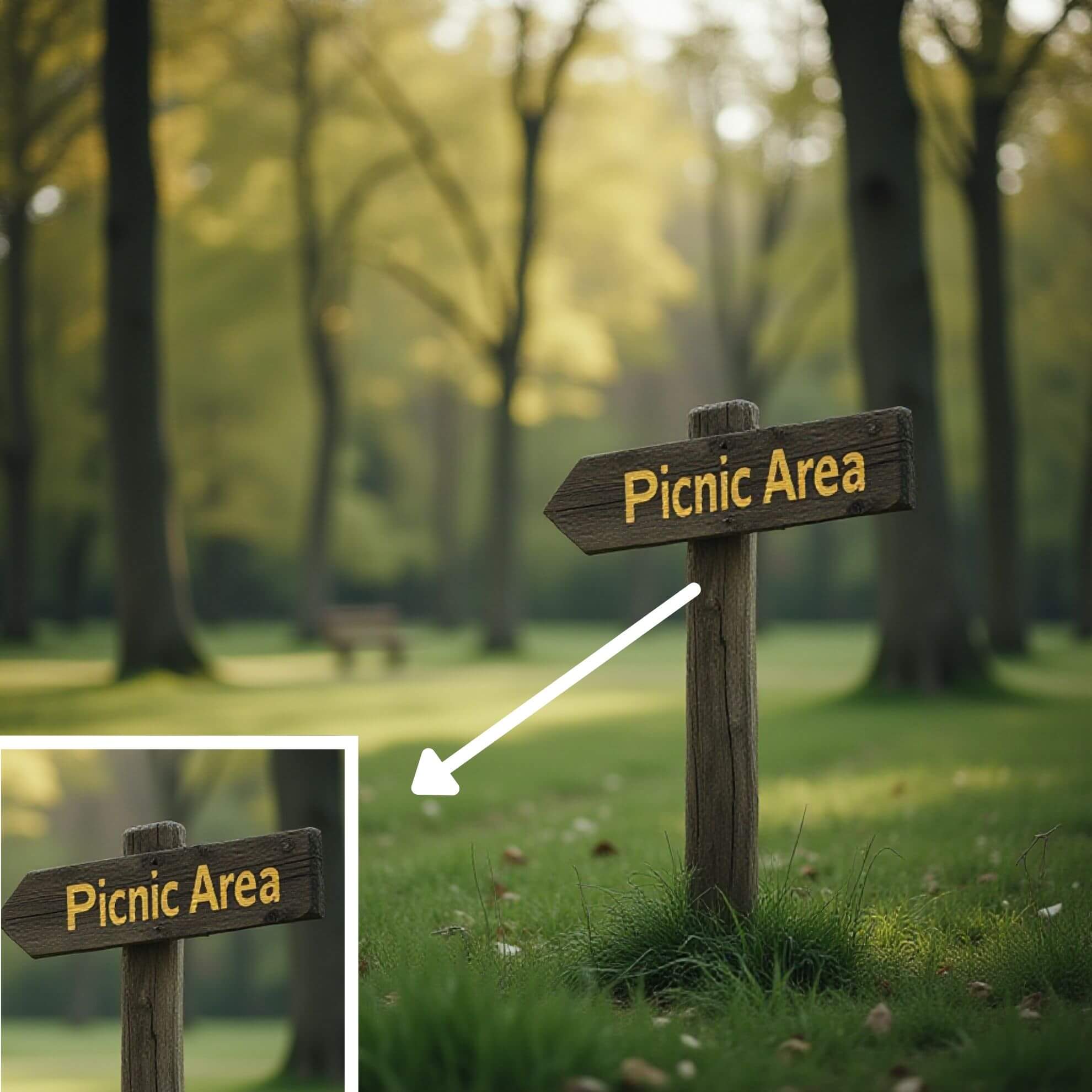} &
\includegraphics[width=0.19\textwidth]{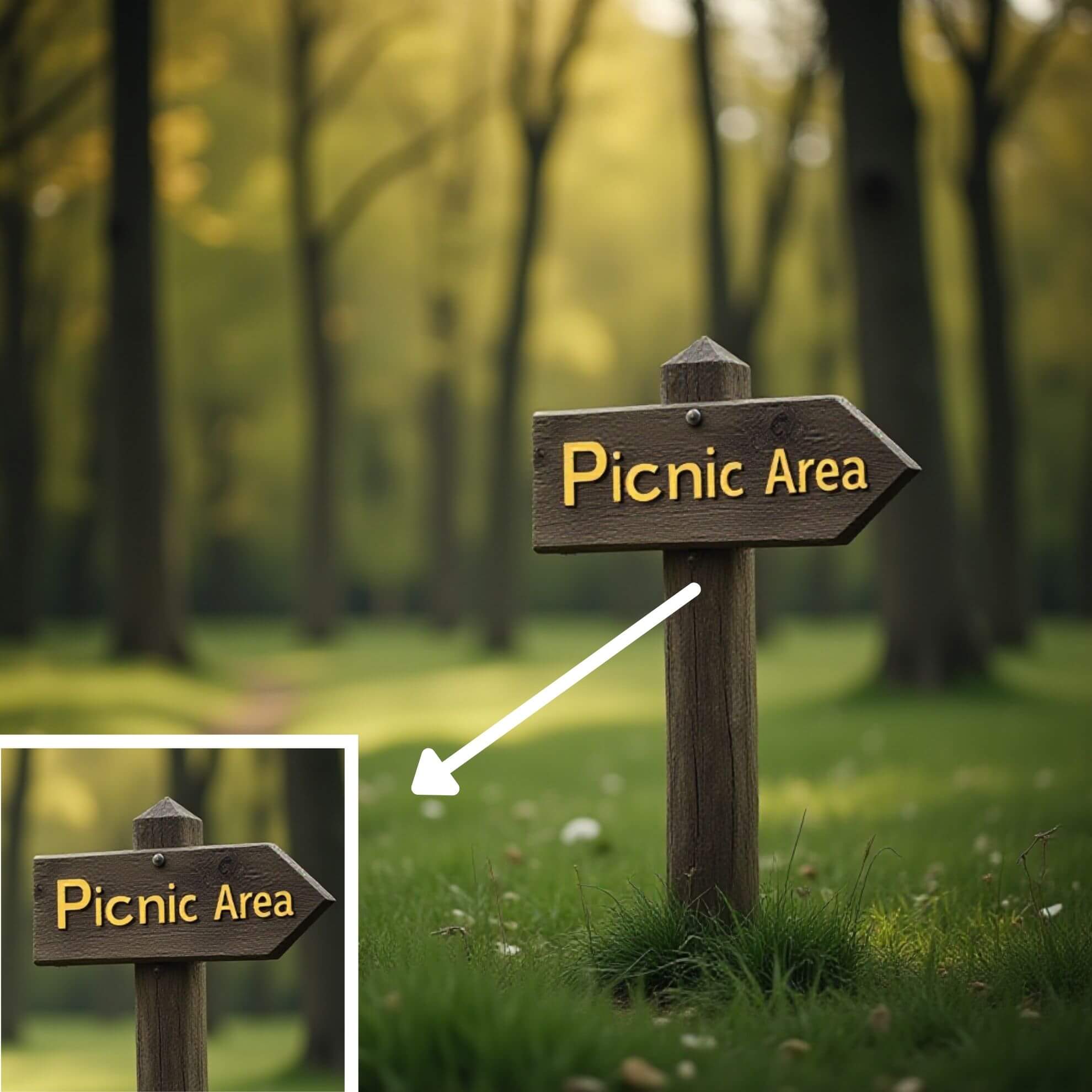} \\

\includegraphics[width=0.19\textwidth]{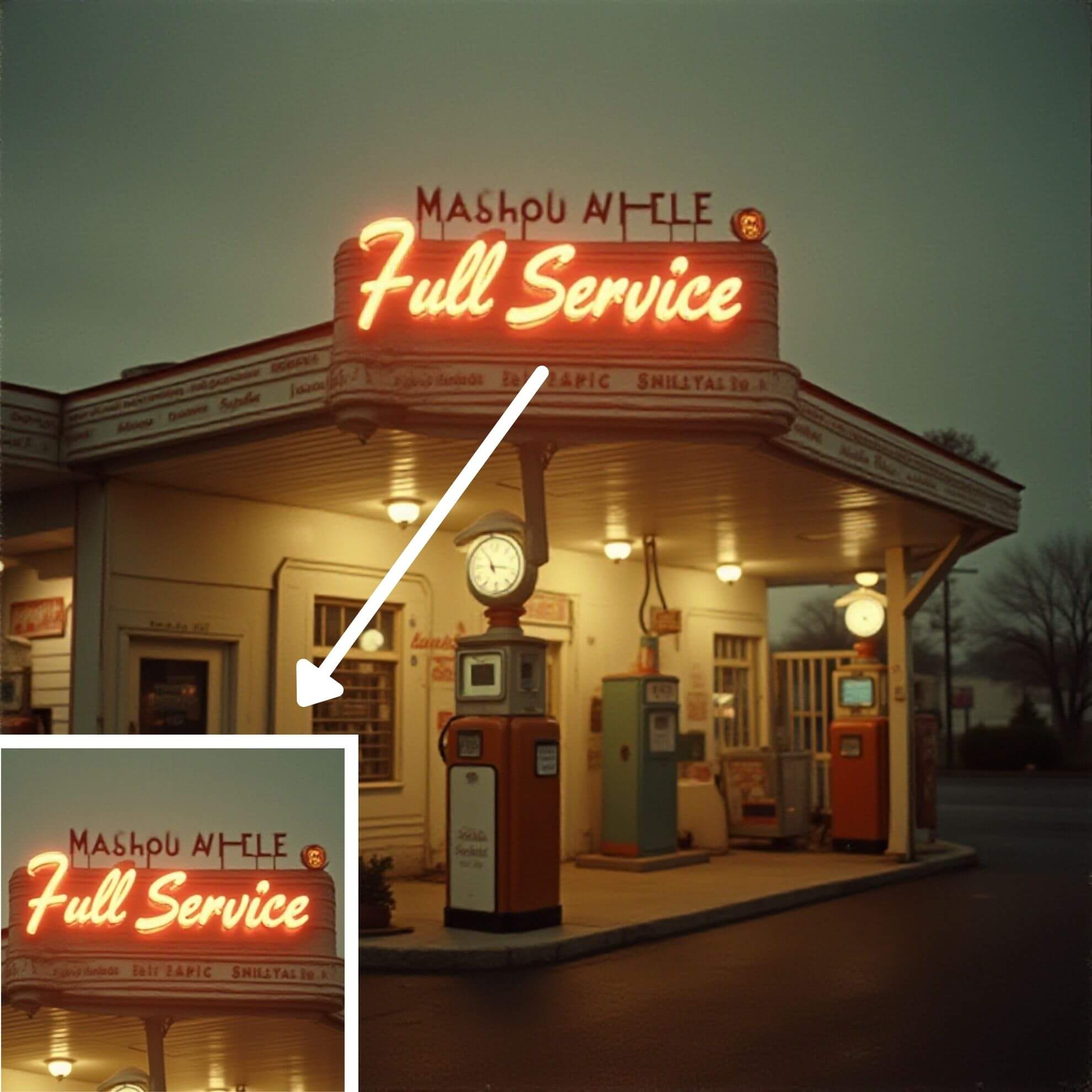} &
\includegraphics[width=0.19\textwidth]{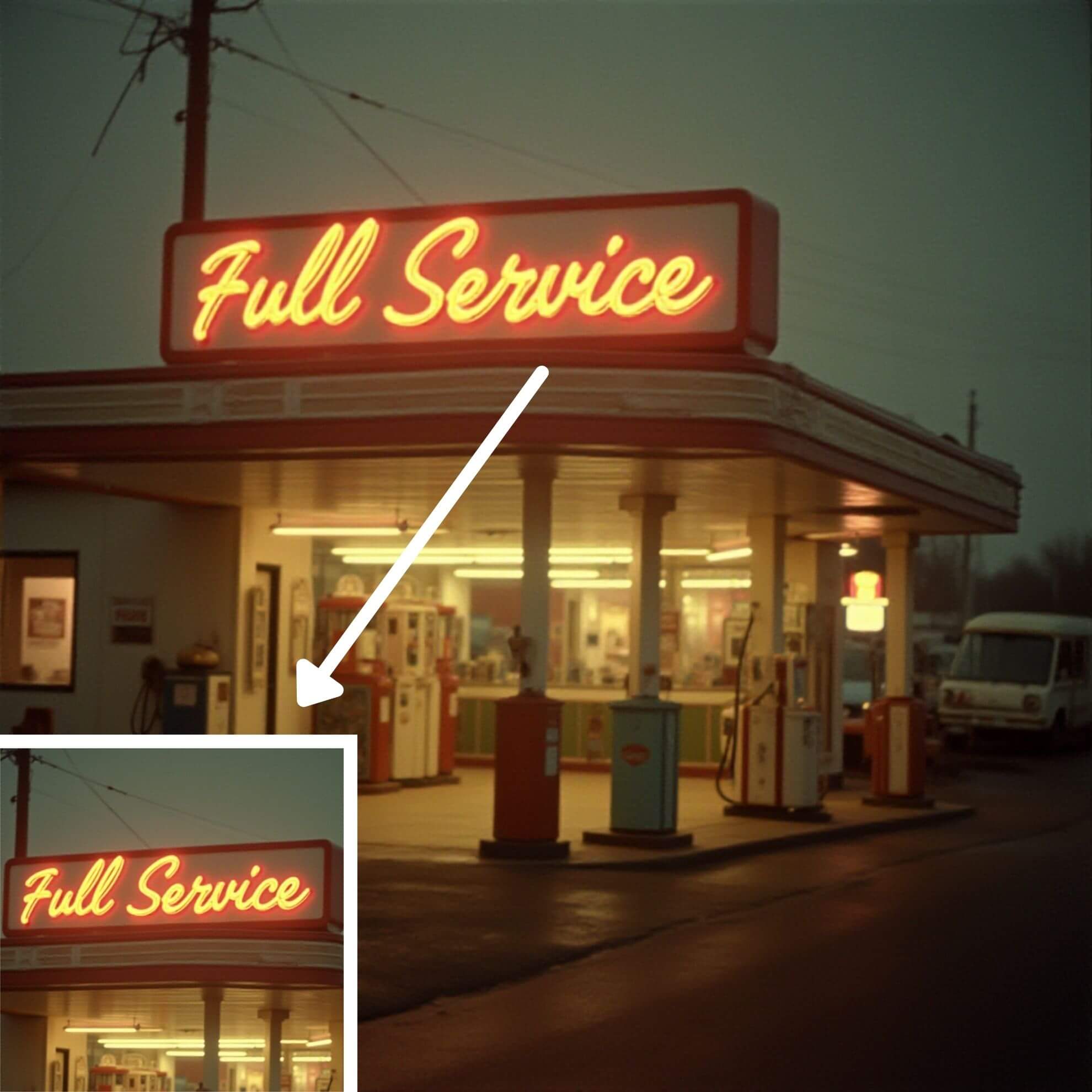} &
\includegraphics[width=0.19\textwidth]{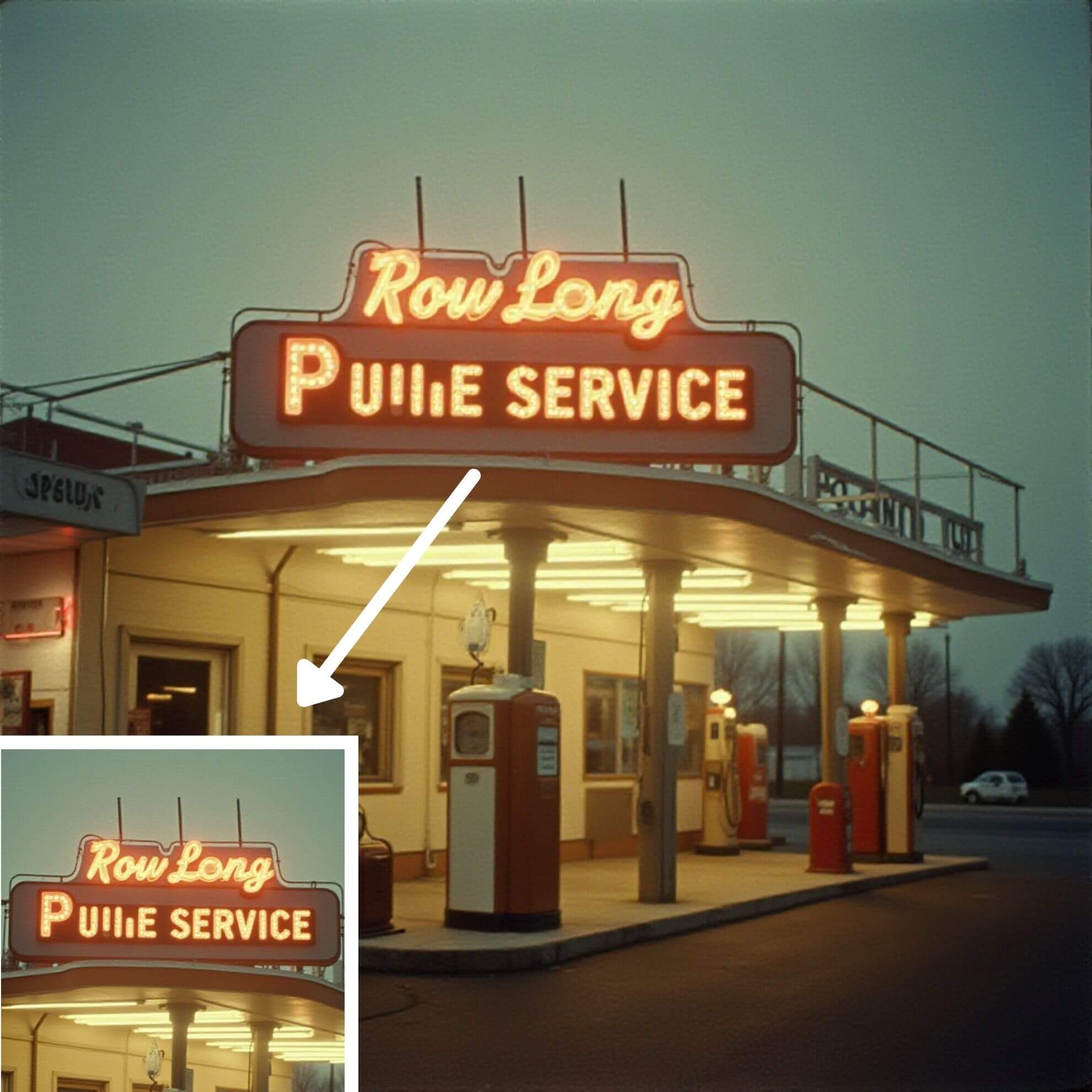} &
\includegraphics[width=0.19\textwidth]{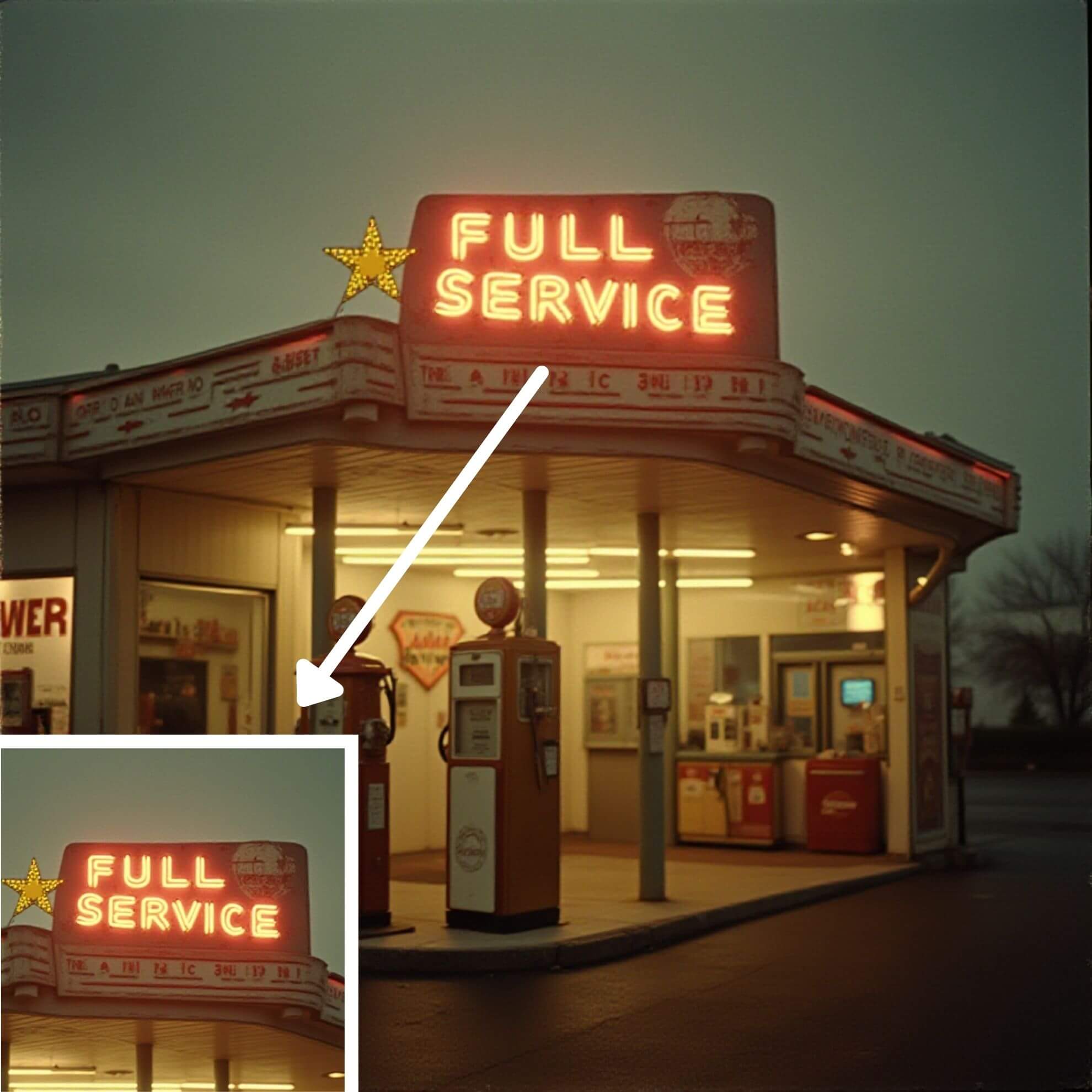} \\
 
\includegraphics[width=0.19\textwidth]{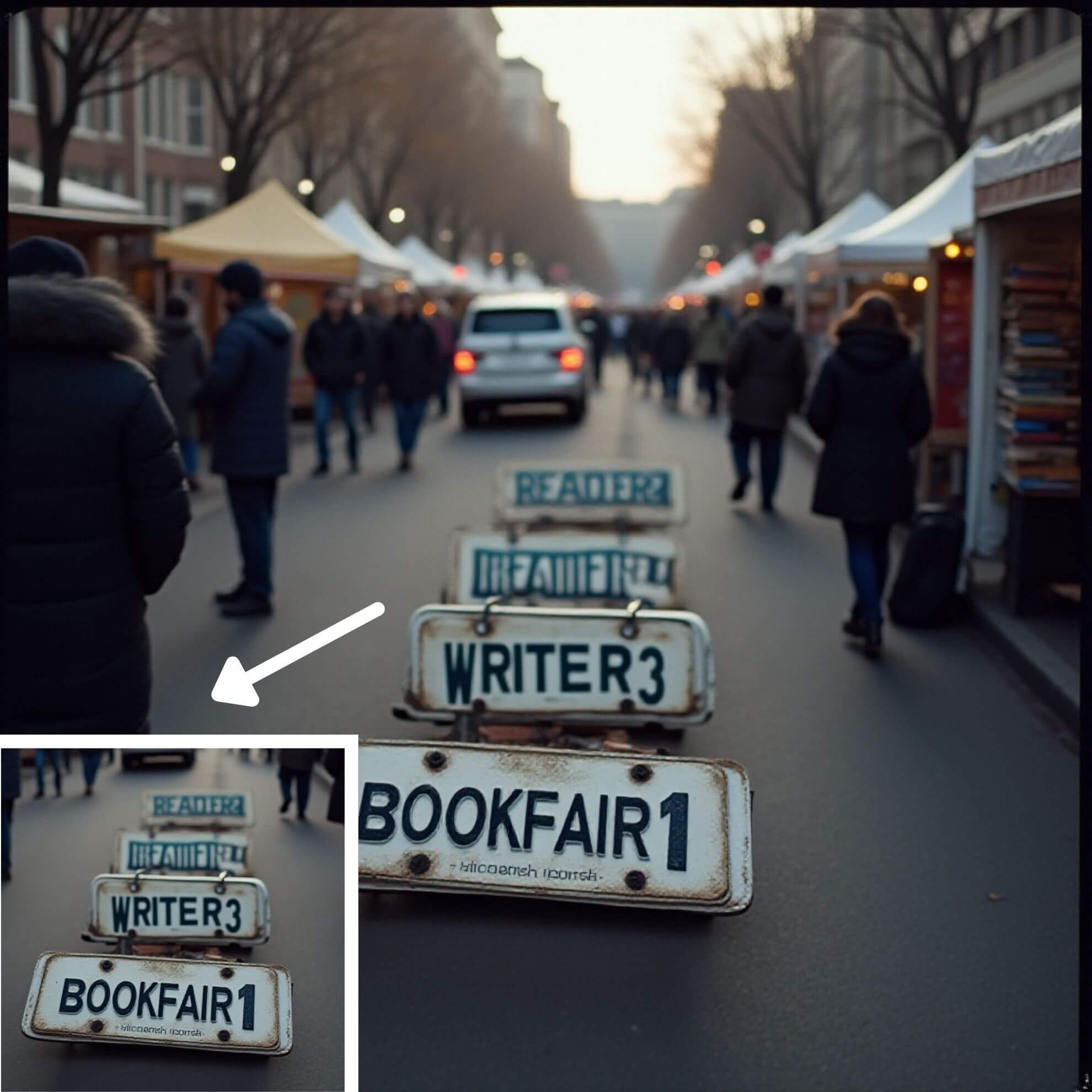} &
\includegraphics[width=0.19\textwidth]{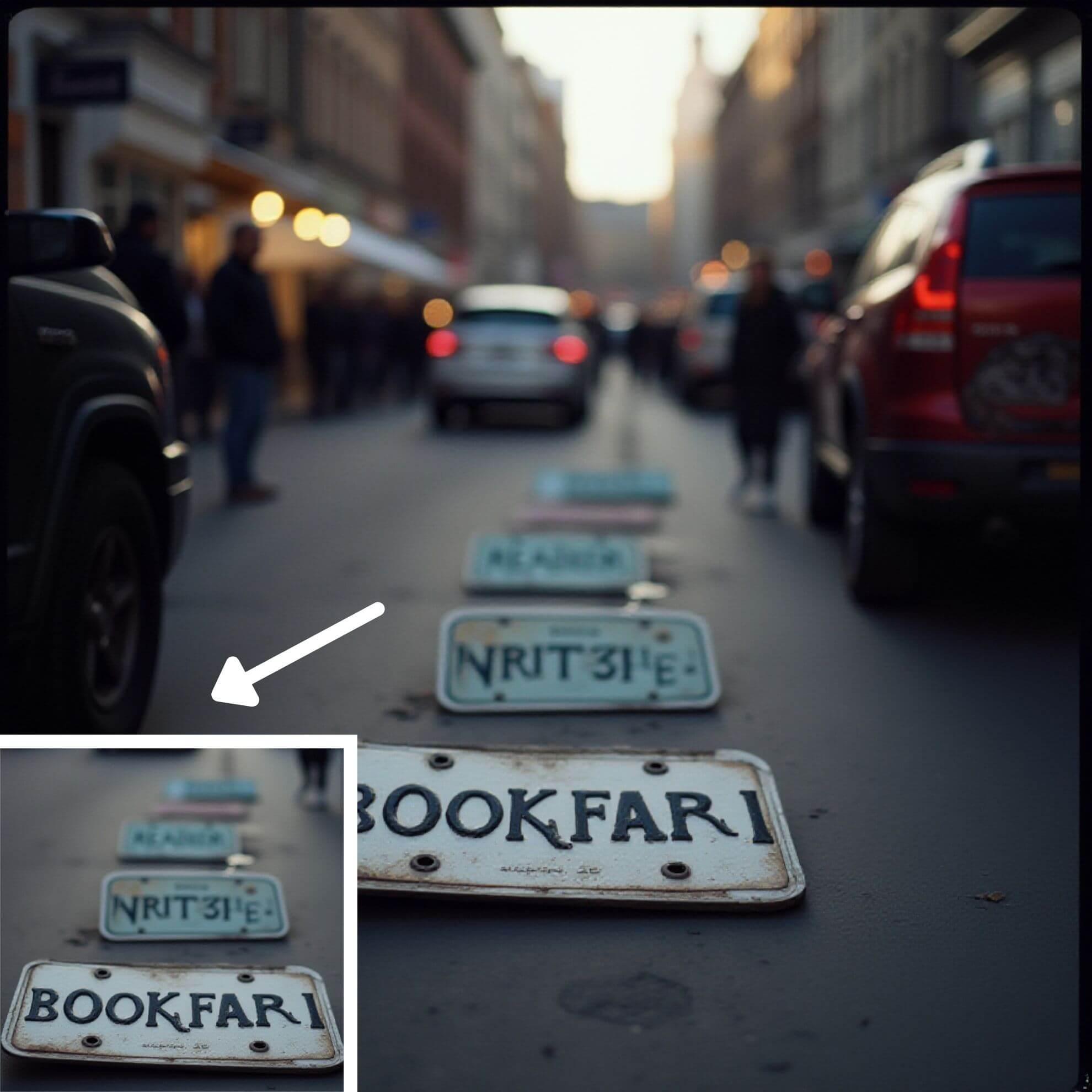} &
\includegraphics[width=0.19\textwidth]{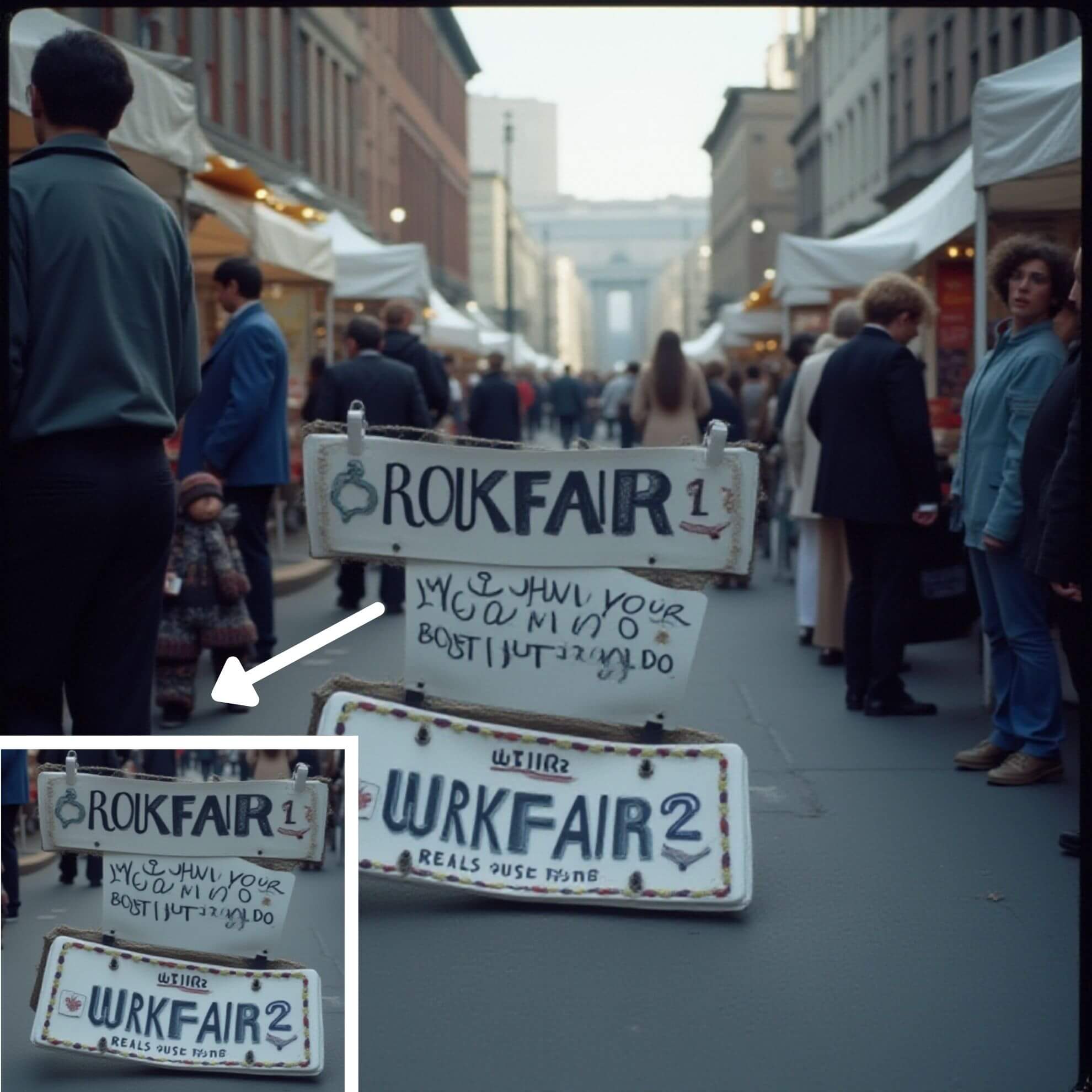} &
\includegraphics[width=0.19\textwidth]{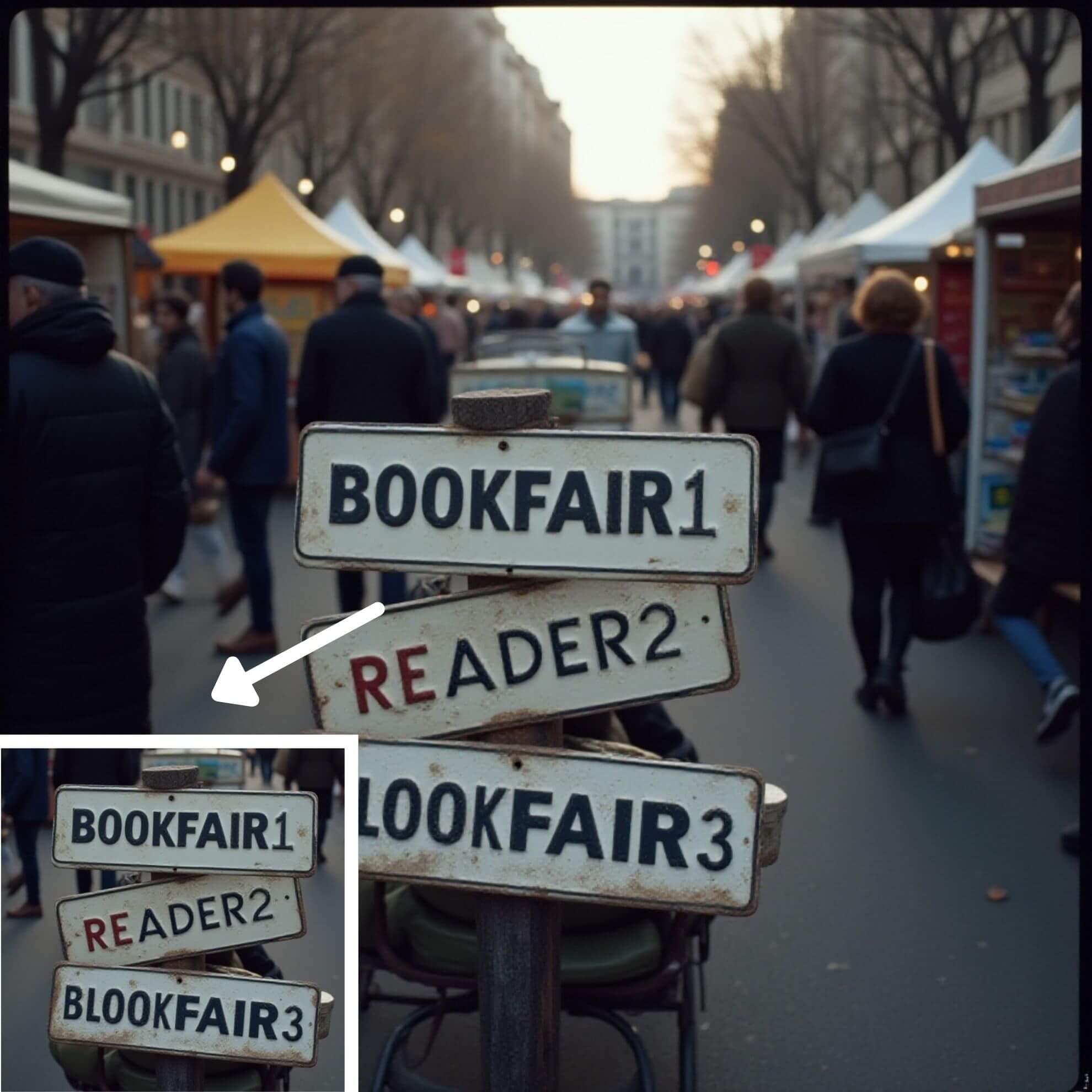} \\

\includegraphics[width=0.19\textwidth]{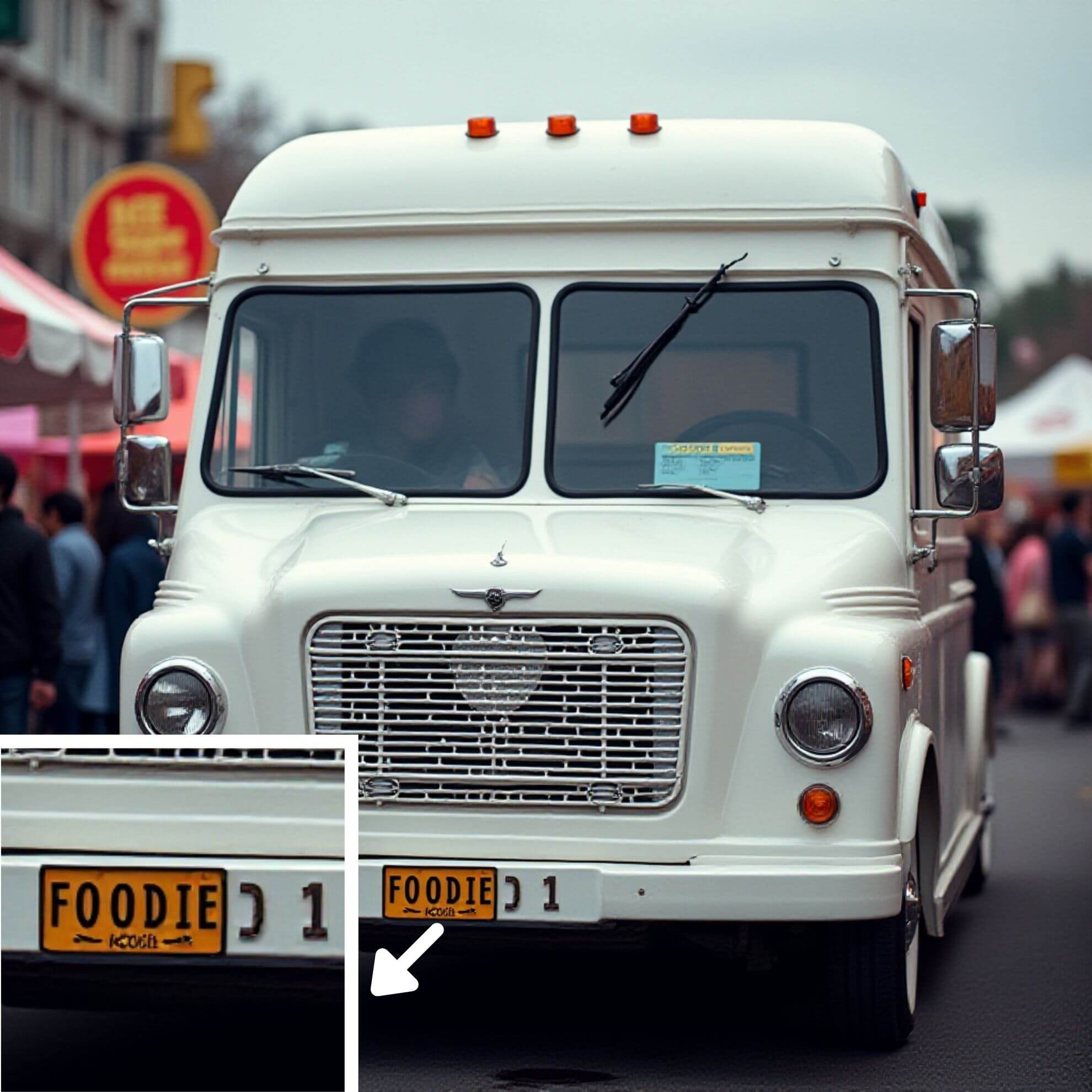} &
\includegraphics[width=0.19\textwidth]{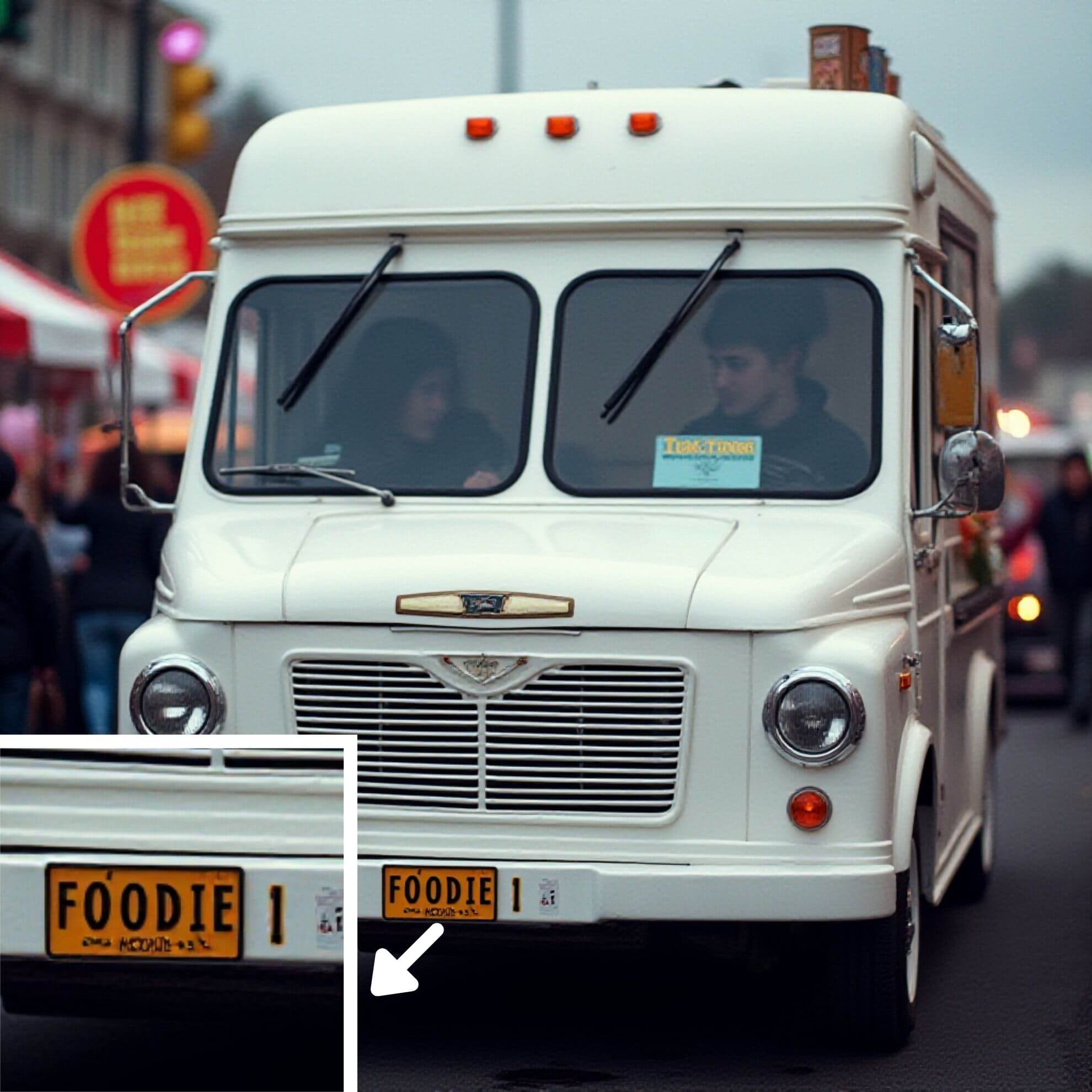} &
\includegraphics[width=0.19\textwidth]{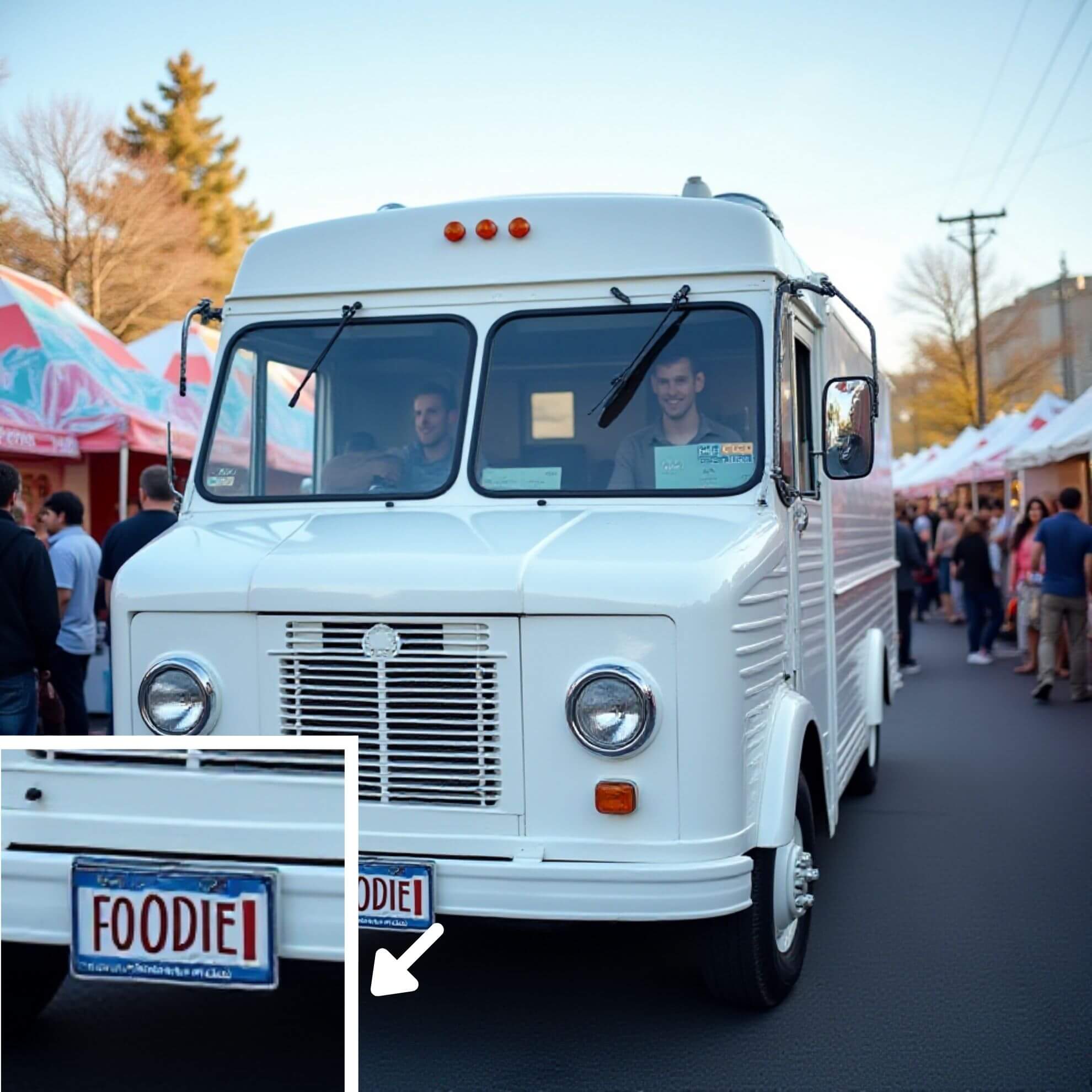} &
\includegraphics[width=0.19\textwidth]{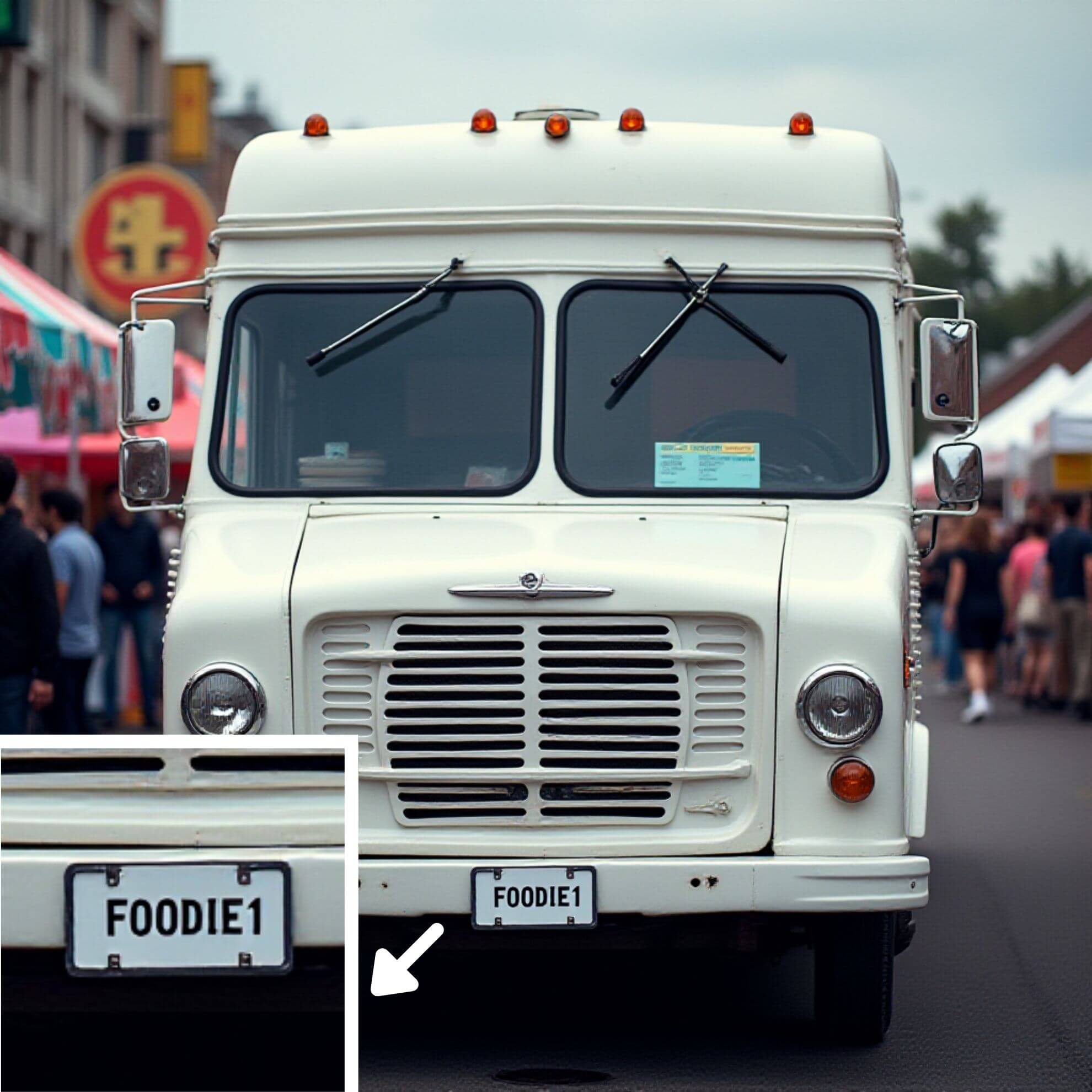} \\

\includegraphics[width=0.19\textwidth]{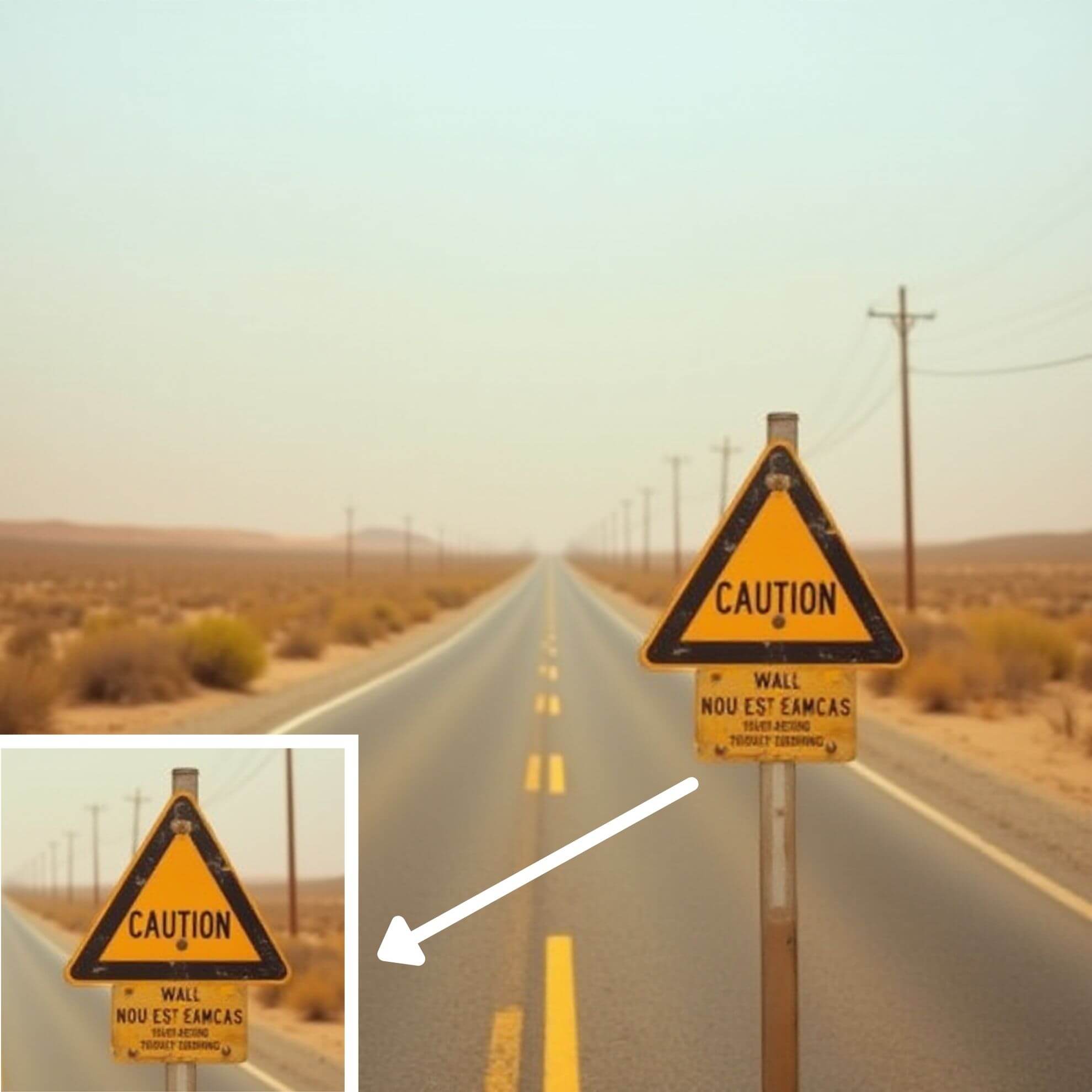} &
\includegraphics[width=0.19\textwidth]{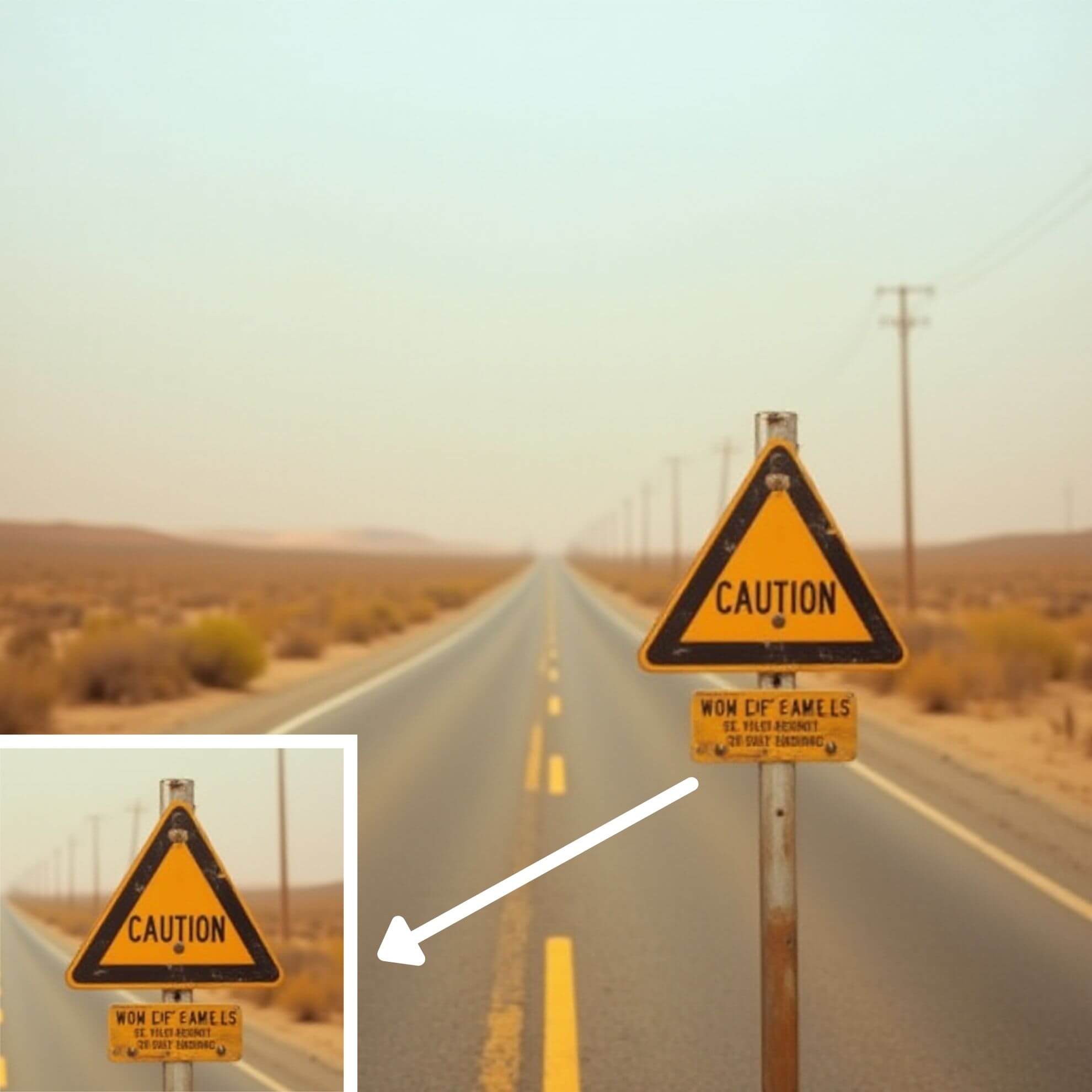} &
\includegraphics[width=0.19\textwidth]{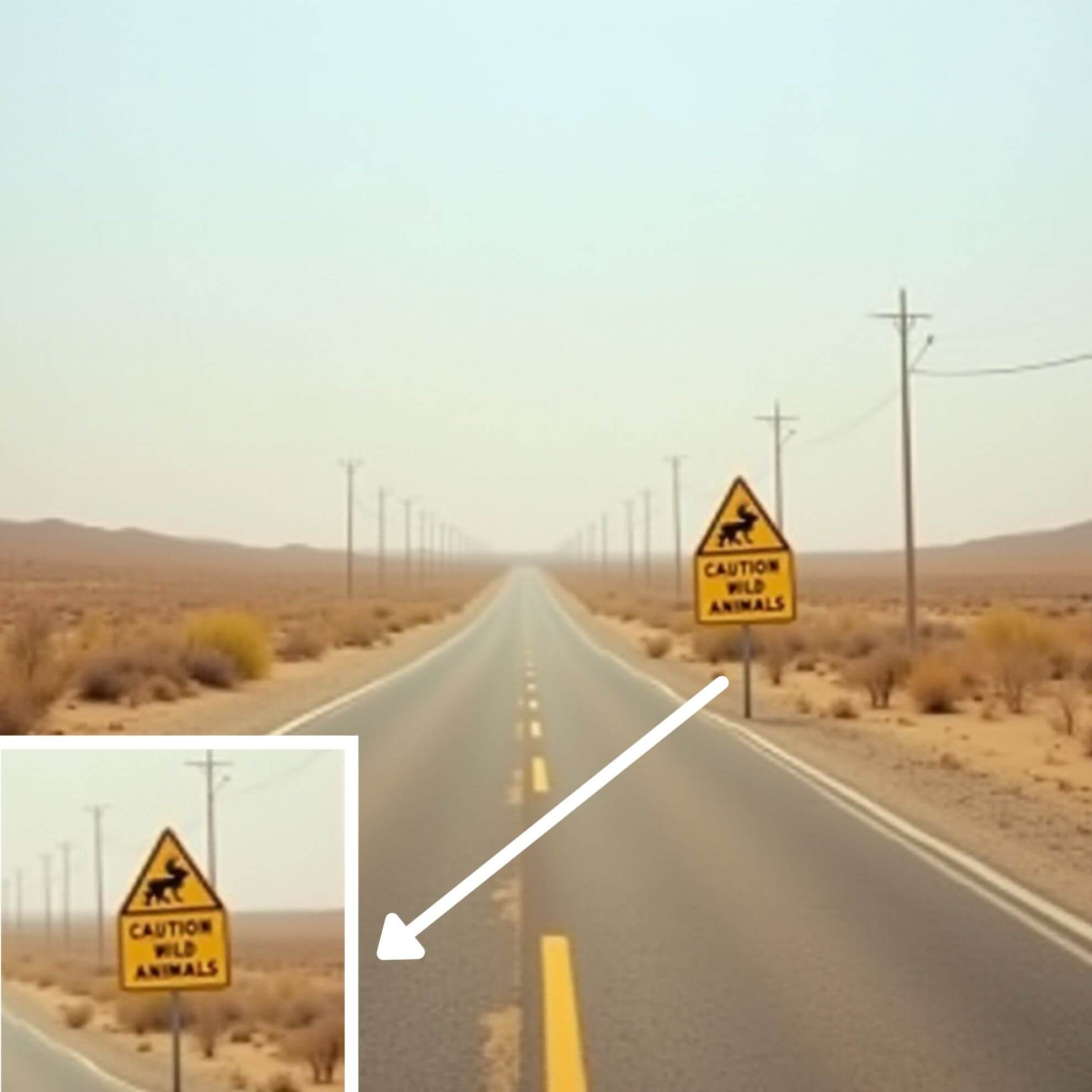} &
\includegraphics[width=0.19\textwidth]{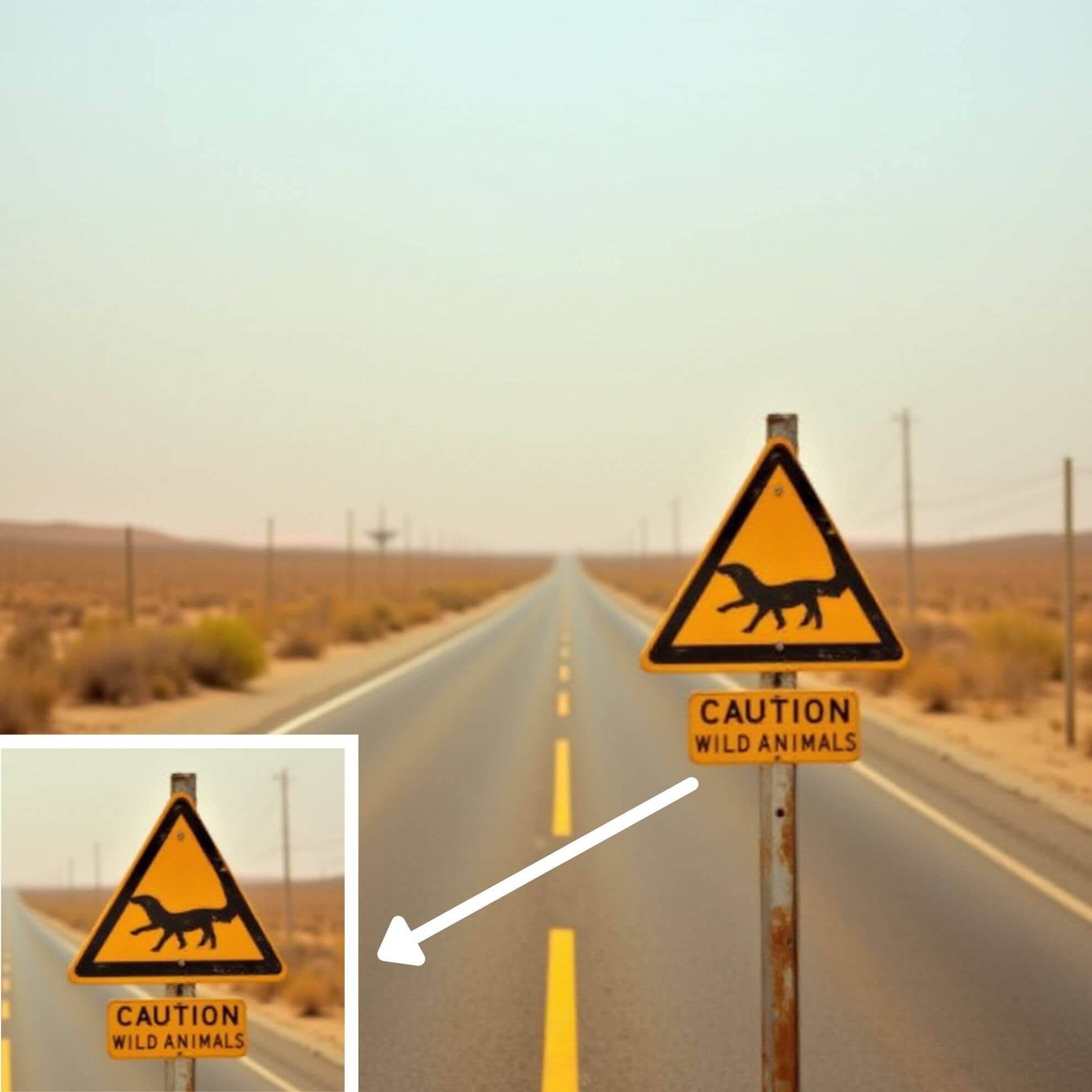}

\end{tabular}
\caption{
\textbf{Visual comparison with other methods. Images for the FLUX.1 [dev] on the \textit{words} dataset using \our{} for 10 inference steps.} \our{} can correct car registrations, road signs, plates with inscriptions, including availability information, such as opening hours (e.g., "24/7").}
\label{fig:dev_text2}
\end{figure*}

\begin{figure*}[!h]
\centering
\setlength{\tabcolsep}{1.5pt}
\renewcommand{\arraystretch}{0.9}
\begin{tabular}{cc}
FLUX.1 [dev]\hspace{80pt}+\our{} & FLUX.1 [dev]\hspace{80pt}+\our{} \\
\includegraphics[width=0.48\textwidth]{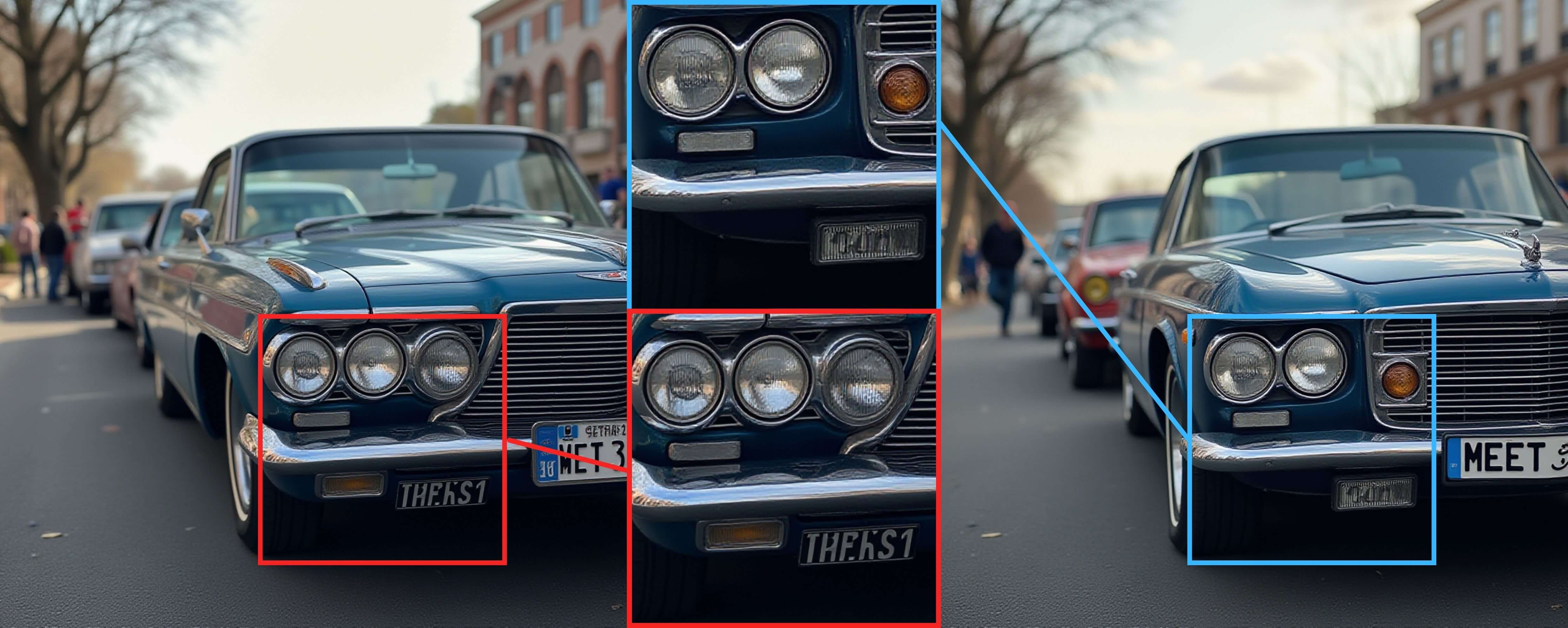} &
\includegraphics[width=0.48\textwidth]{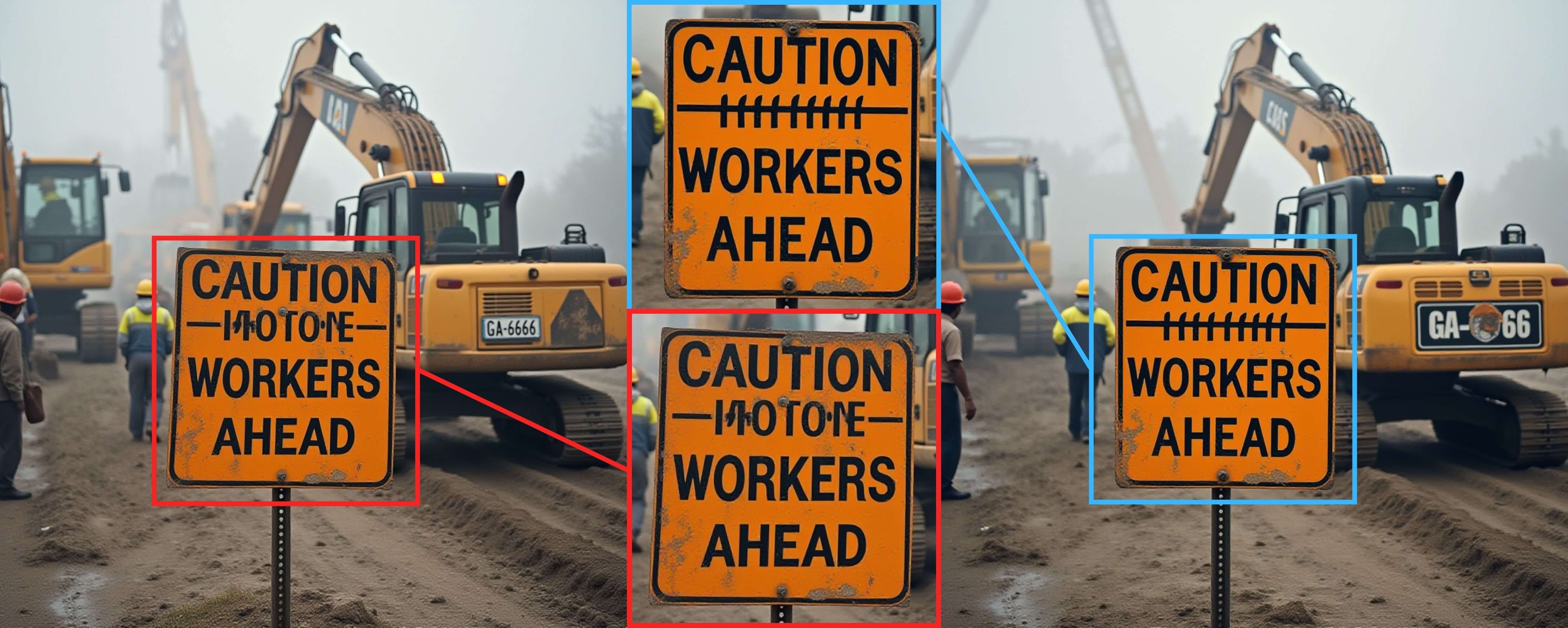} \\

\includegraphics[width=0.48\textwidth]{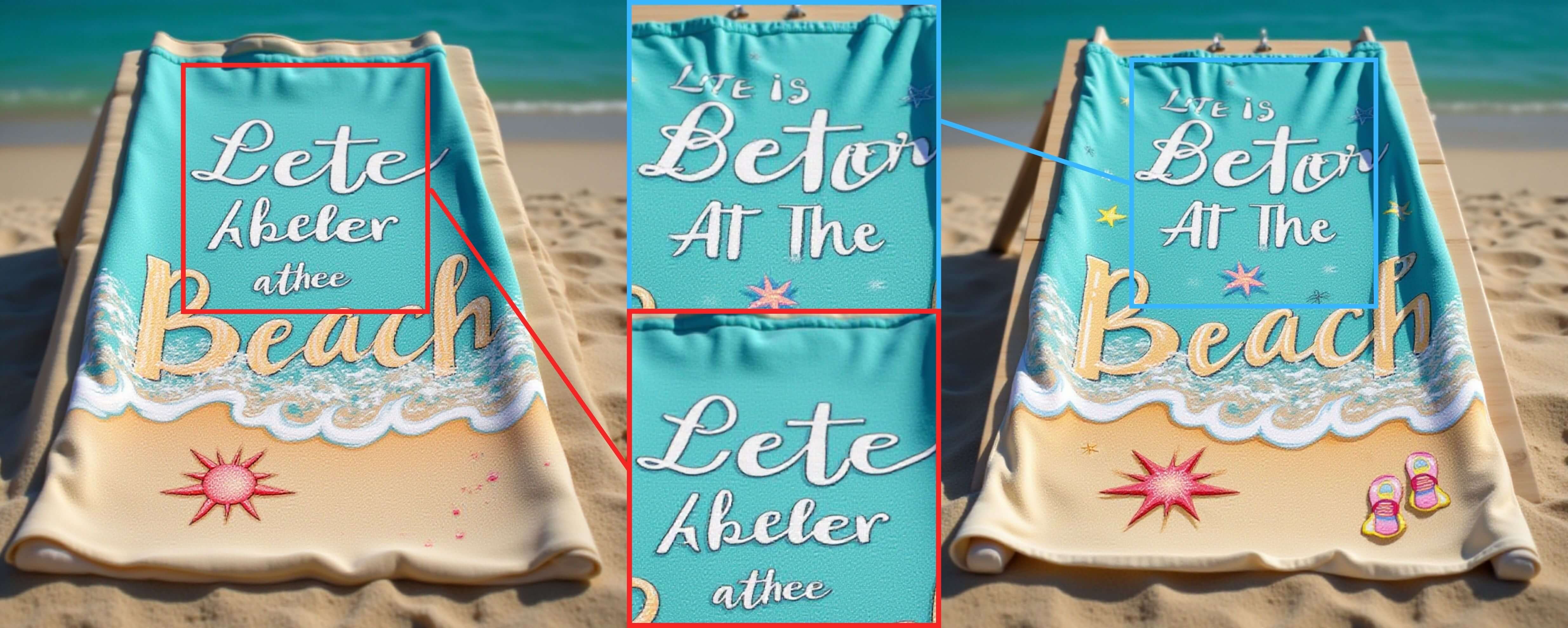} &
\includegraphics[width=0.48\textwidth]{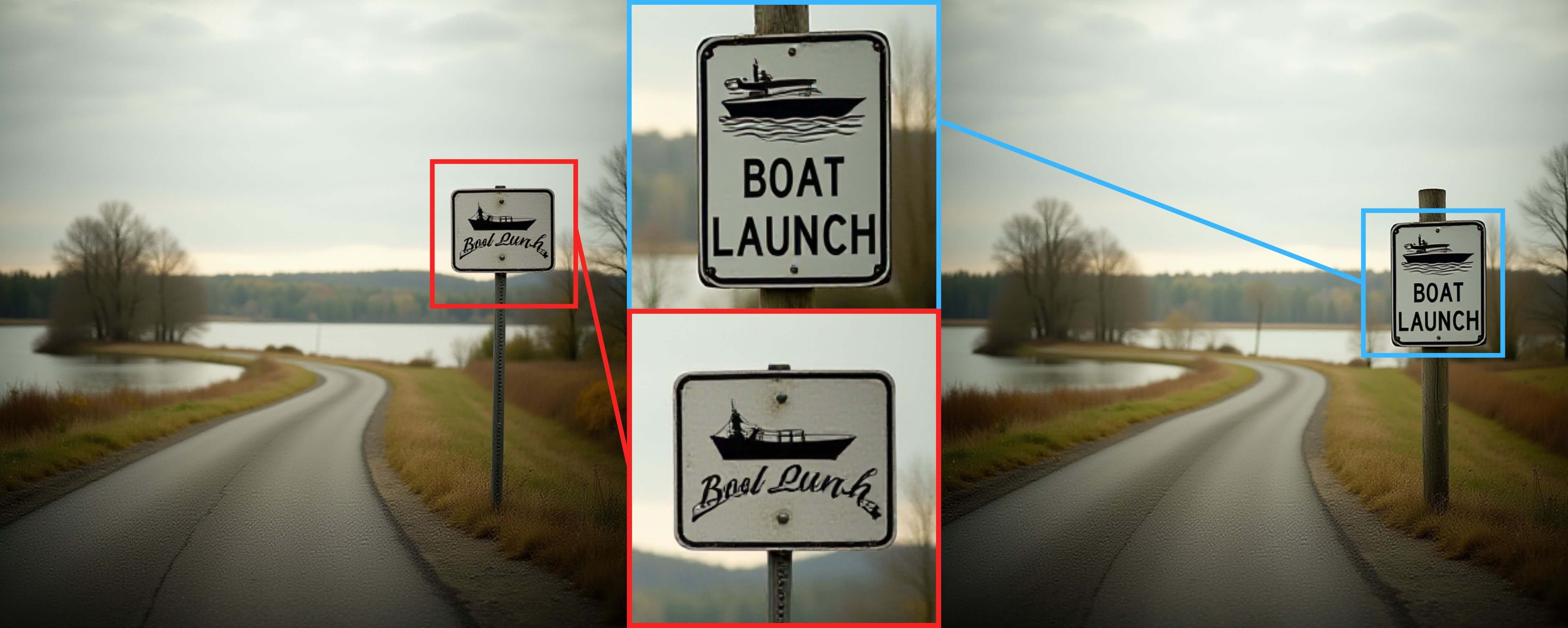} \\

\includegraphics[width=0.48\textwidth]{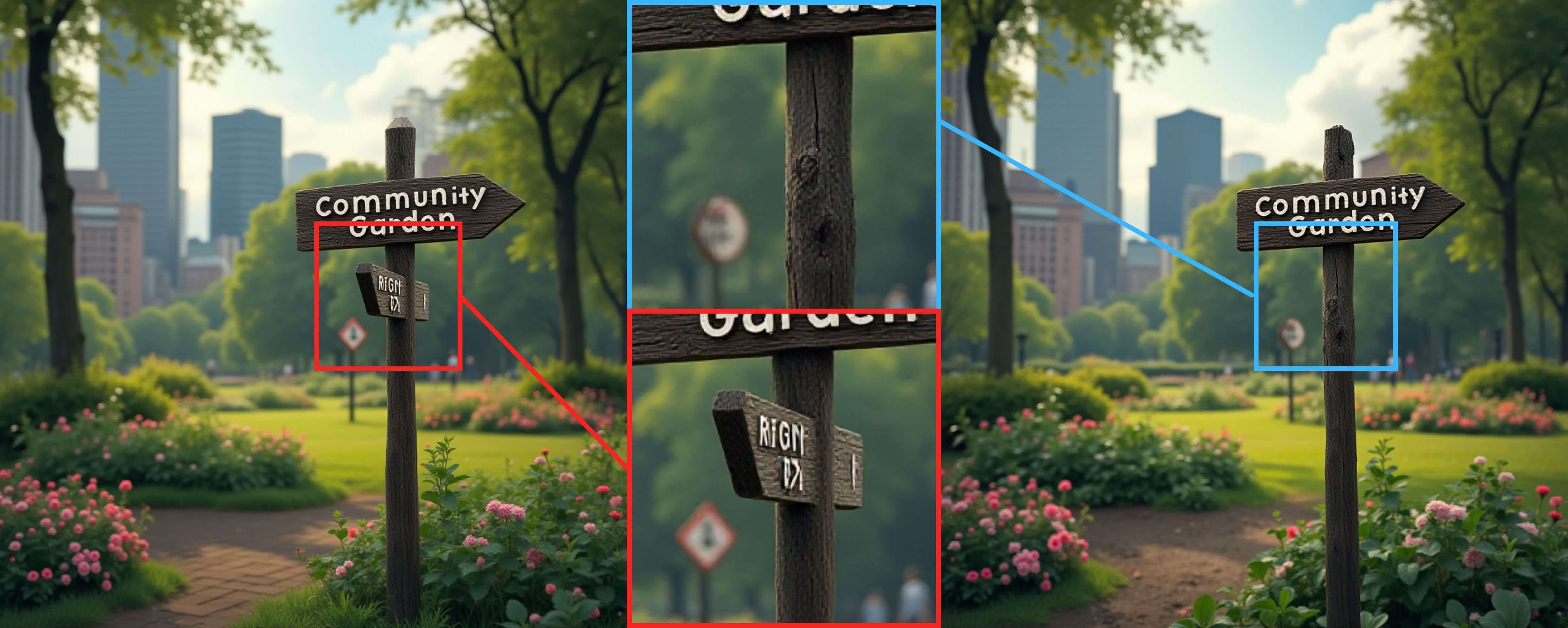} &
\includegraphics[width=0.48\textwidth]{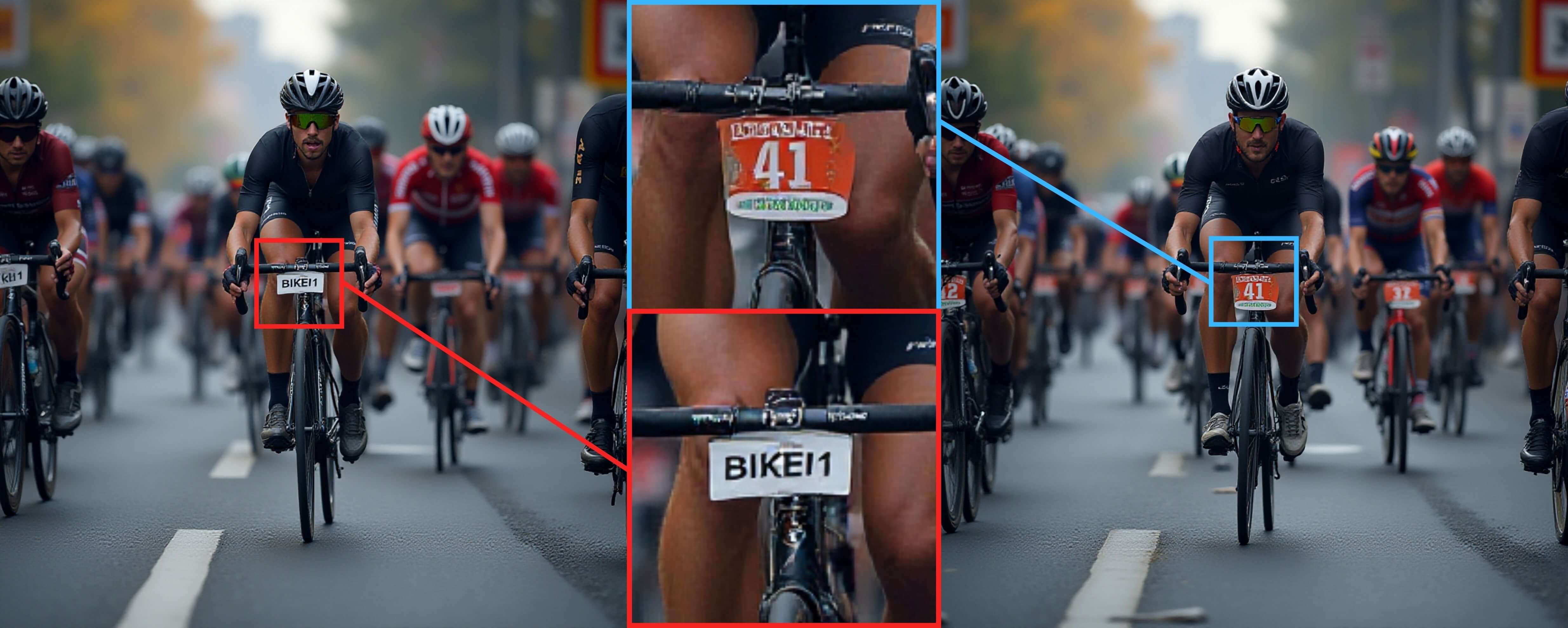} \\

\includegraphics[width=0.48\textwidth]{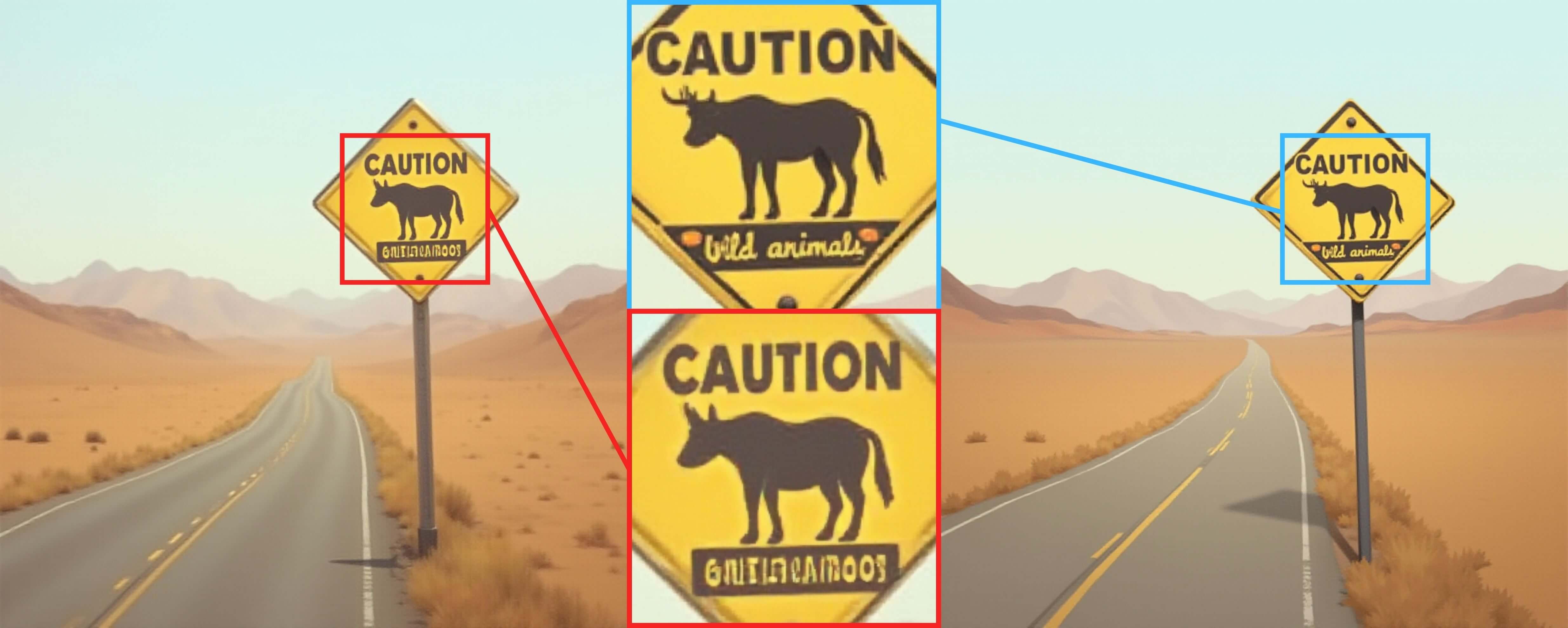} &
\includegraphics[width=0.48\textwidth]{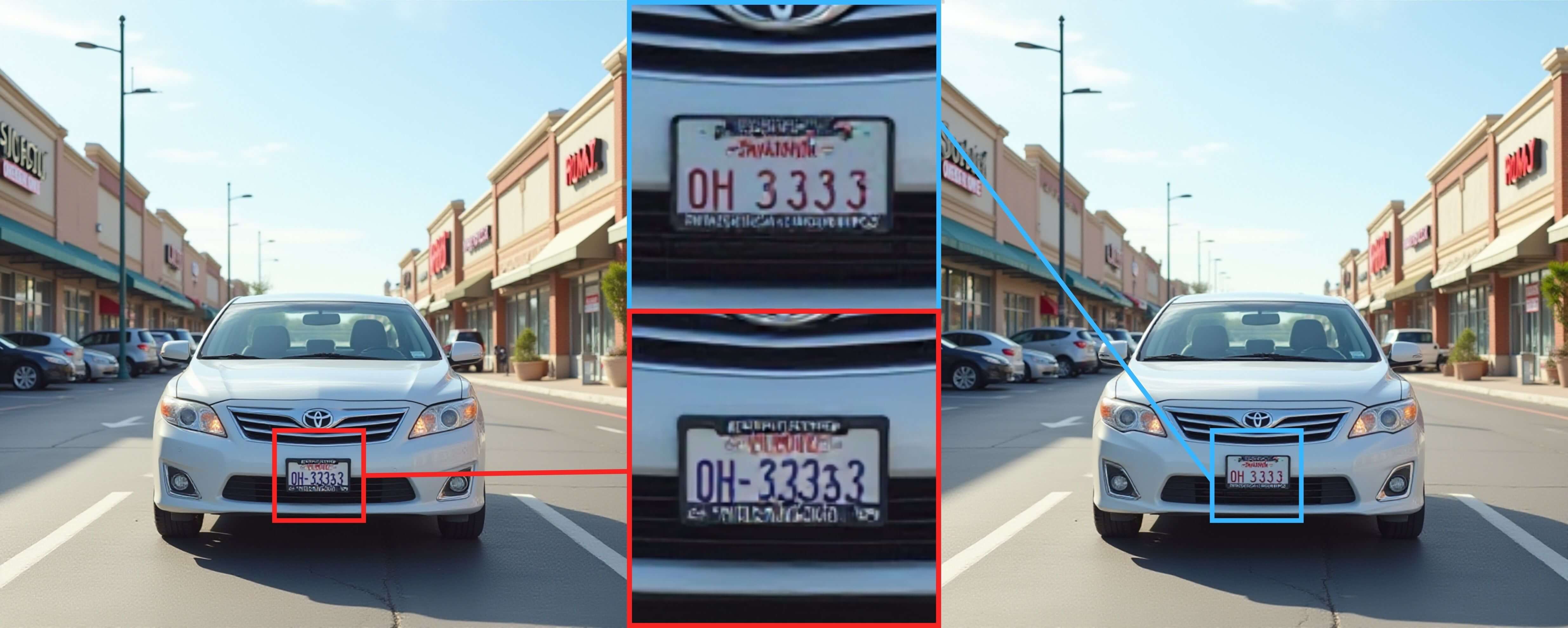} \\

\includegraphics[width=0.48\textwidth]{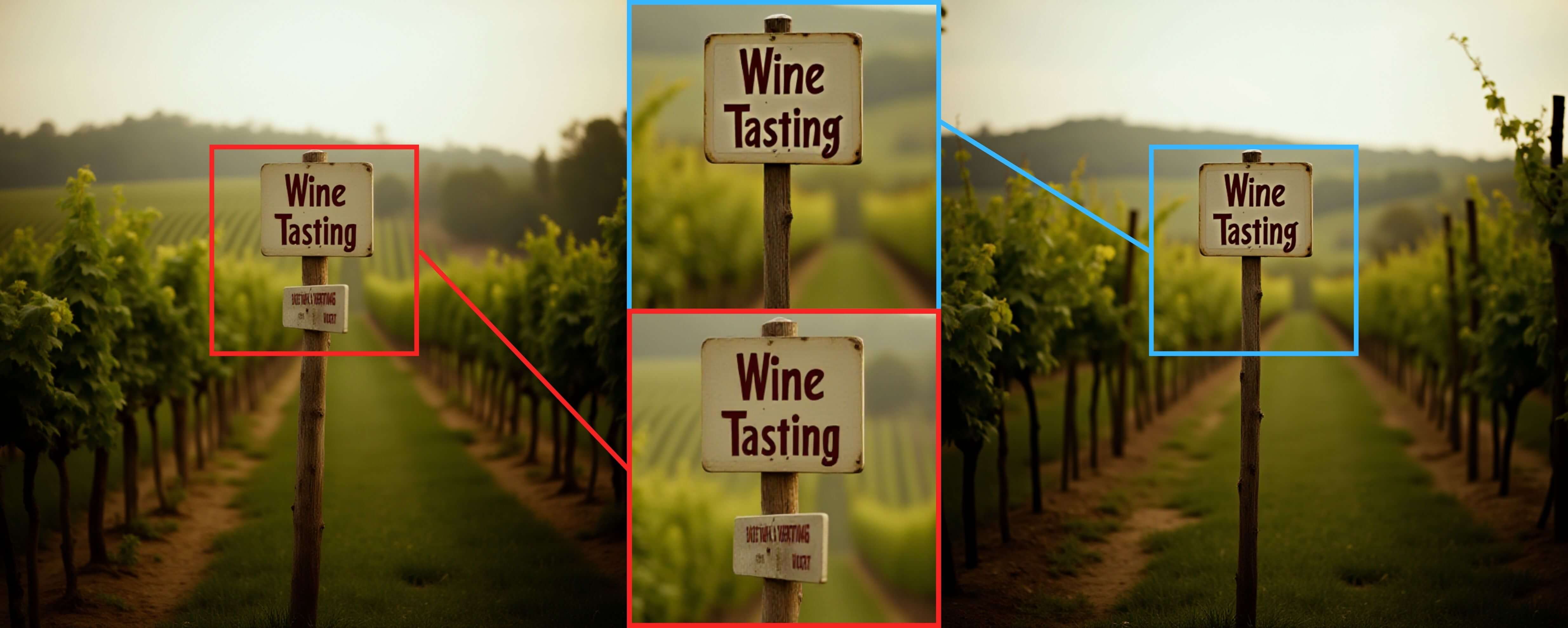} &
\includegraphics[width=0.48\textwidth]{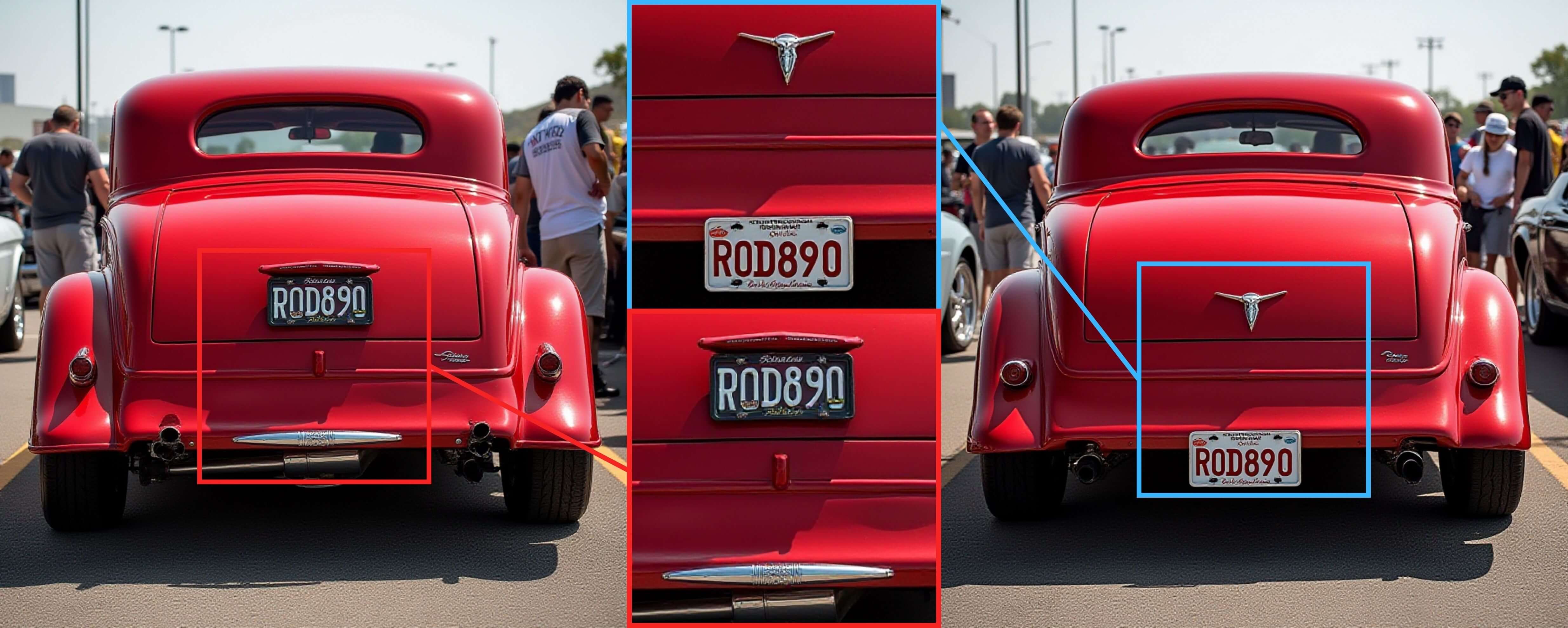} \\

\includegraphics[width=0.48\textwidth]{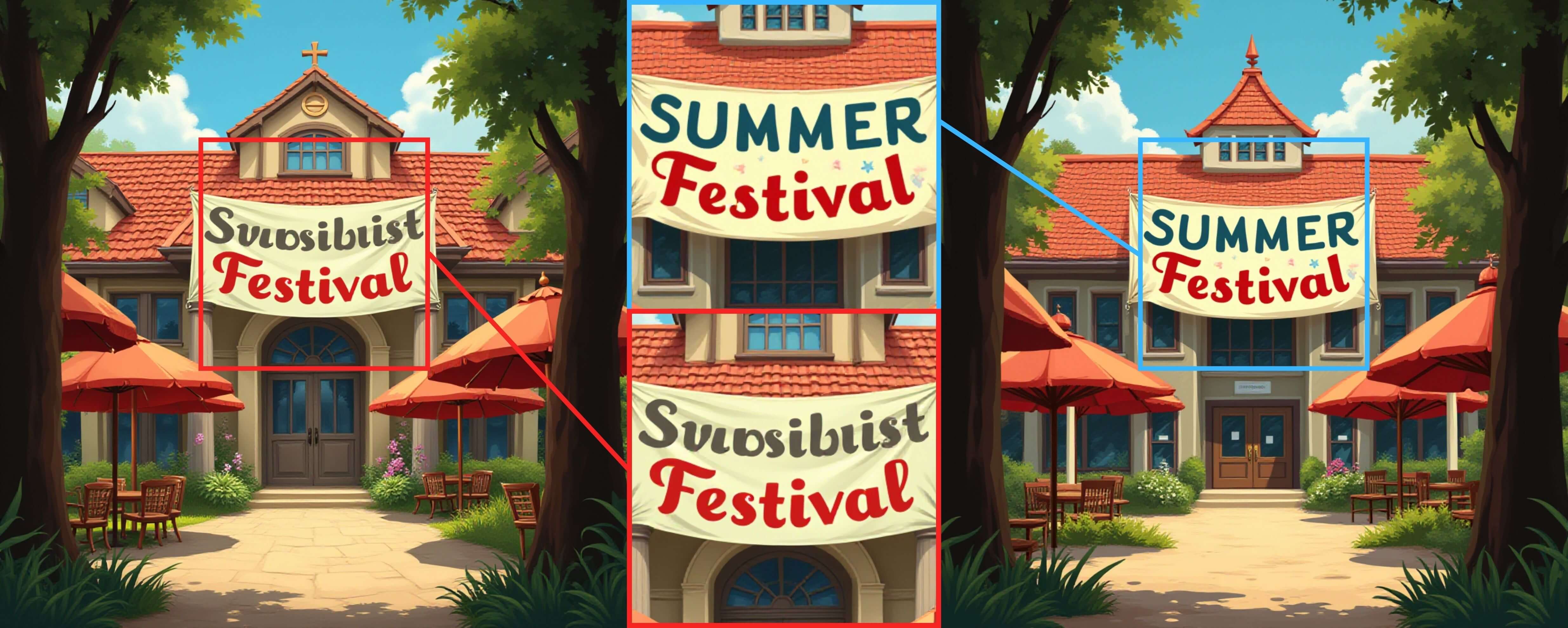} &
\includegraphics[width=0.48\textwidth]{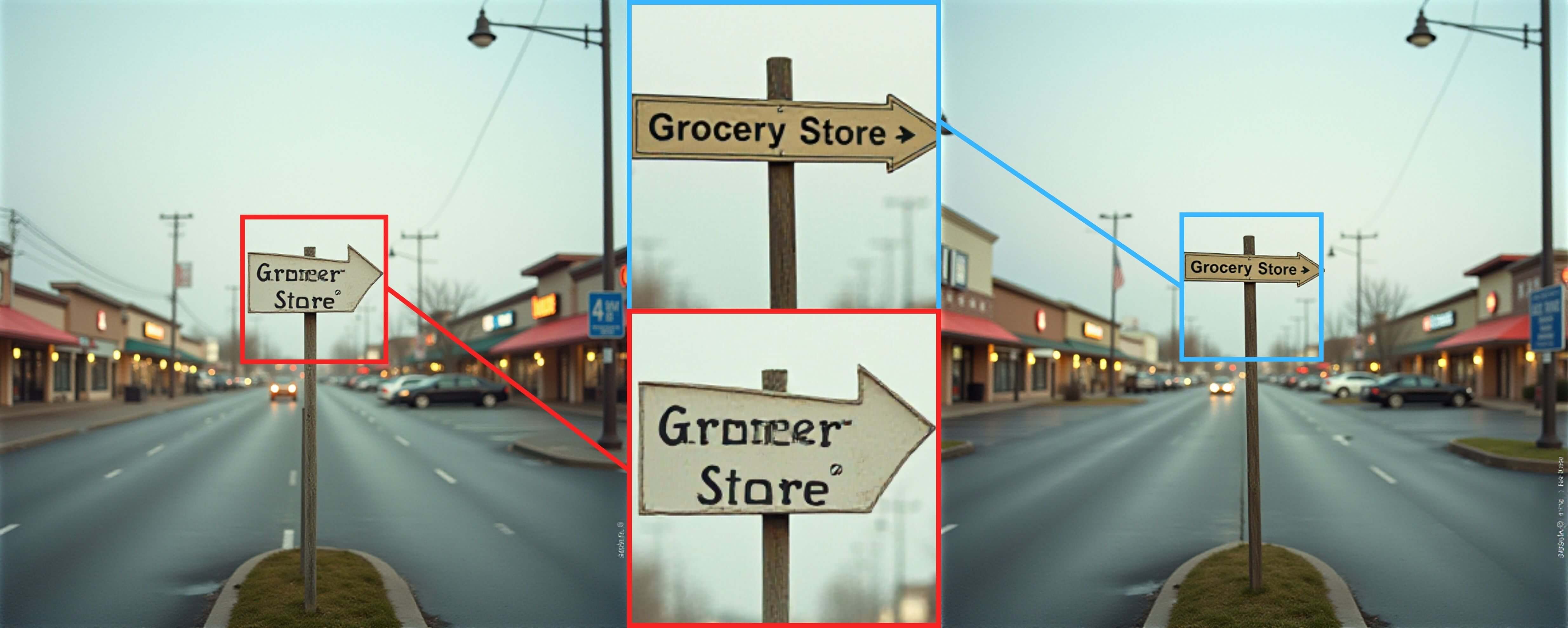}

\end{tabular}
\caption{\textbf{Images for the FLUX.1 [dev] on the \textit{words} dataset using \our{} for 10 inference steps.} In each example, we can see a significant improvement in image quality by removing text artifacts. It is worth noting that text accuracy exists even in the case of graphic text (slanted, e.g. on a banner) or artistic fonts.
}
\label{fig:dev_text}
\end{figure*}

\begin{figure*}[!h]
\centering
\setlength{\tabcolsep}{1.2pt}
\renewcommand{\arraystretch}{0.9}
\begin{tabular}{cccccccc}
 & $t_{9}$ & $t_{8}$ & $t_{7}$ & $t_{5}$ & $t_{3}$ & $t_{0}$ & Final\\
\rotatebox{90}{\hspace{+2mm}FLUX.1 [dev]} & \includegraphics[width=0.130\textwidth, height=0.130\textwidth]{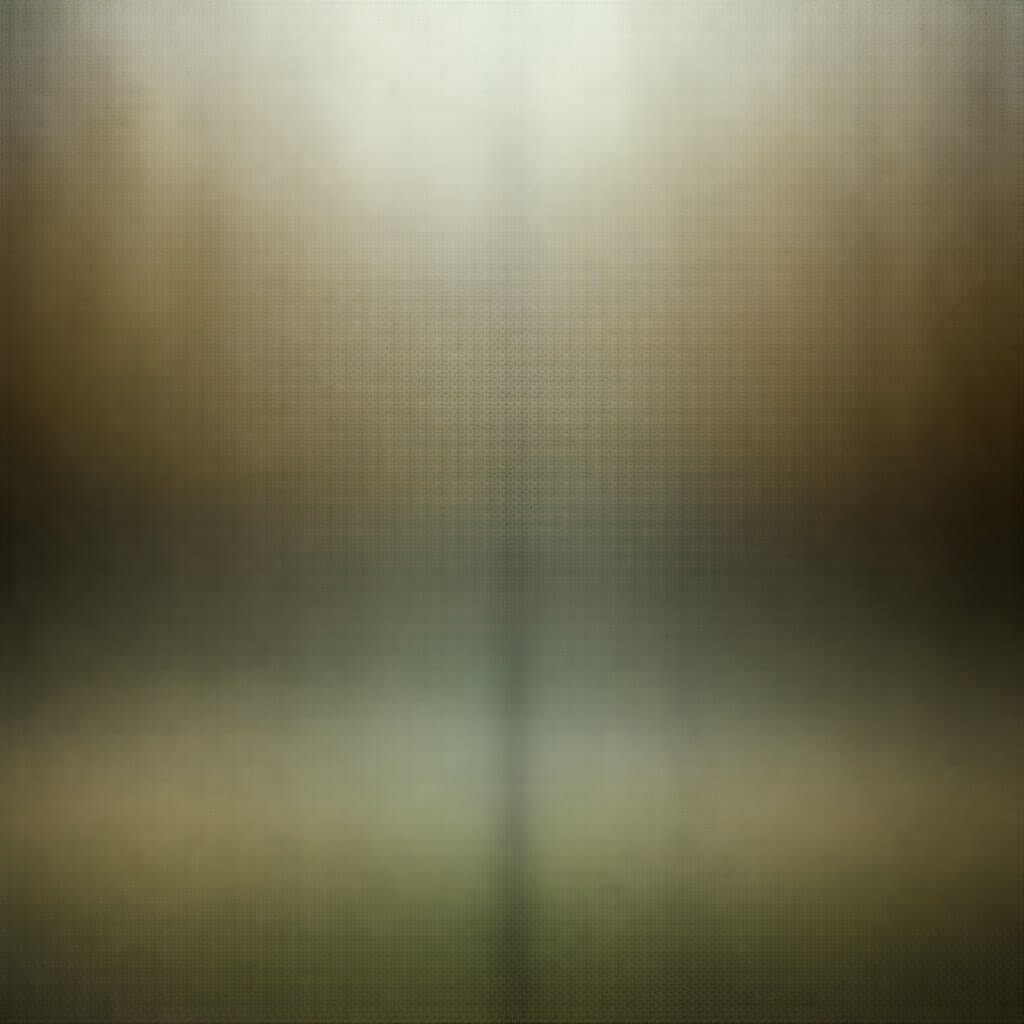} &
\includegraphics[width=0.130\textwidth, height=0.130\textwidth]{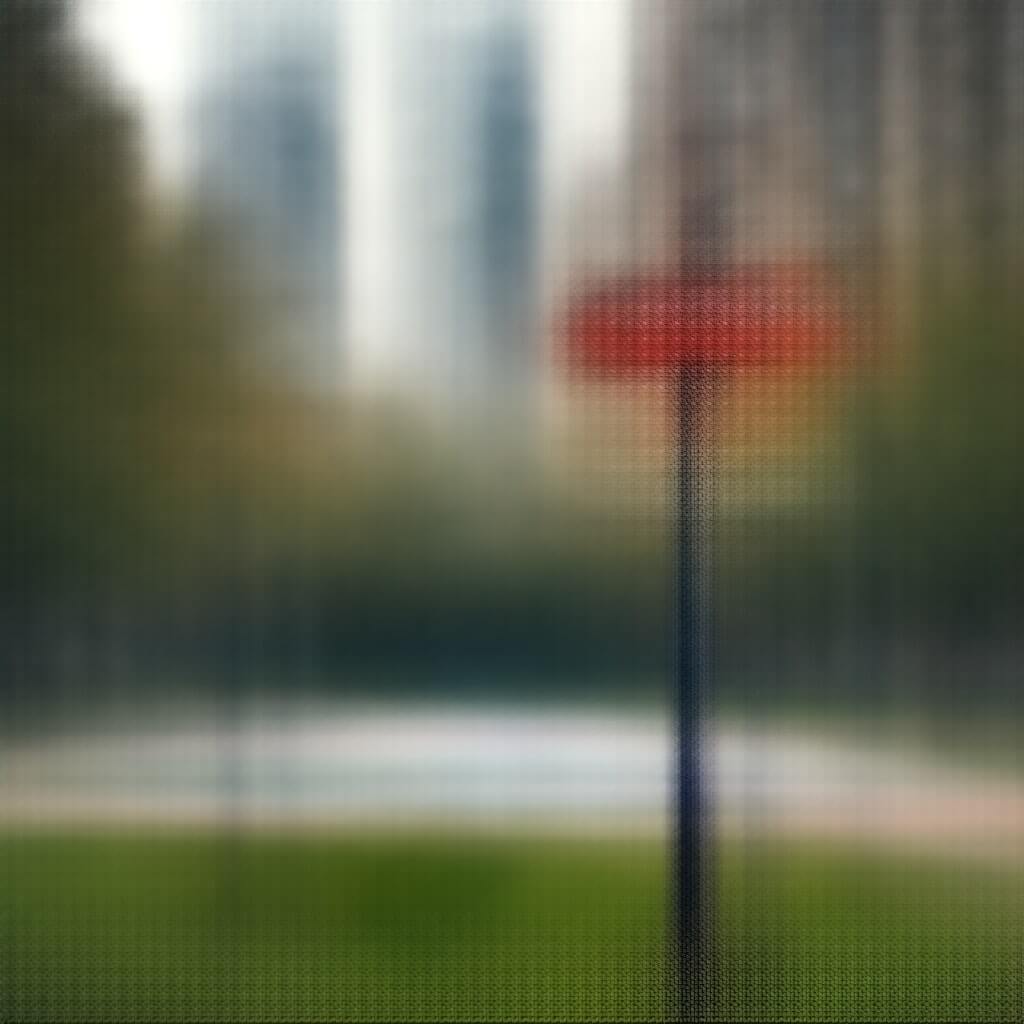} &
\includegraphics[width=0.130\textwidth, height=0.130\textwidth]{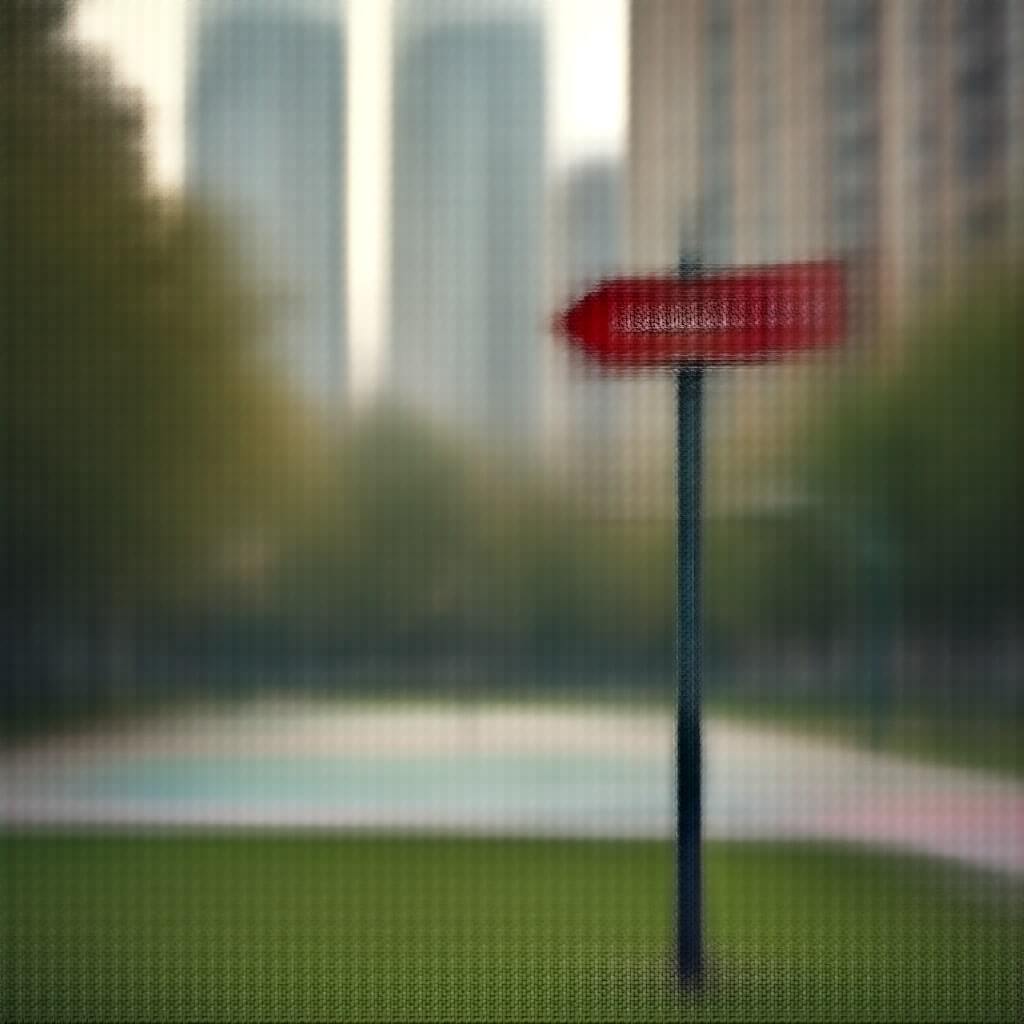} &
\includegraphics[width=0.130\textwidth, height=0.130\textwidth]{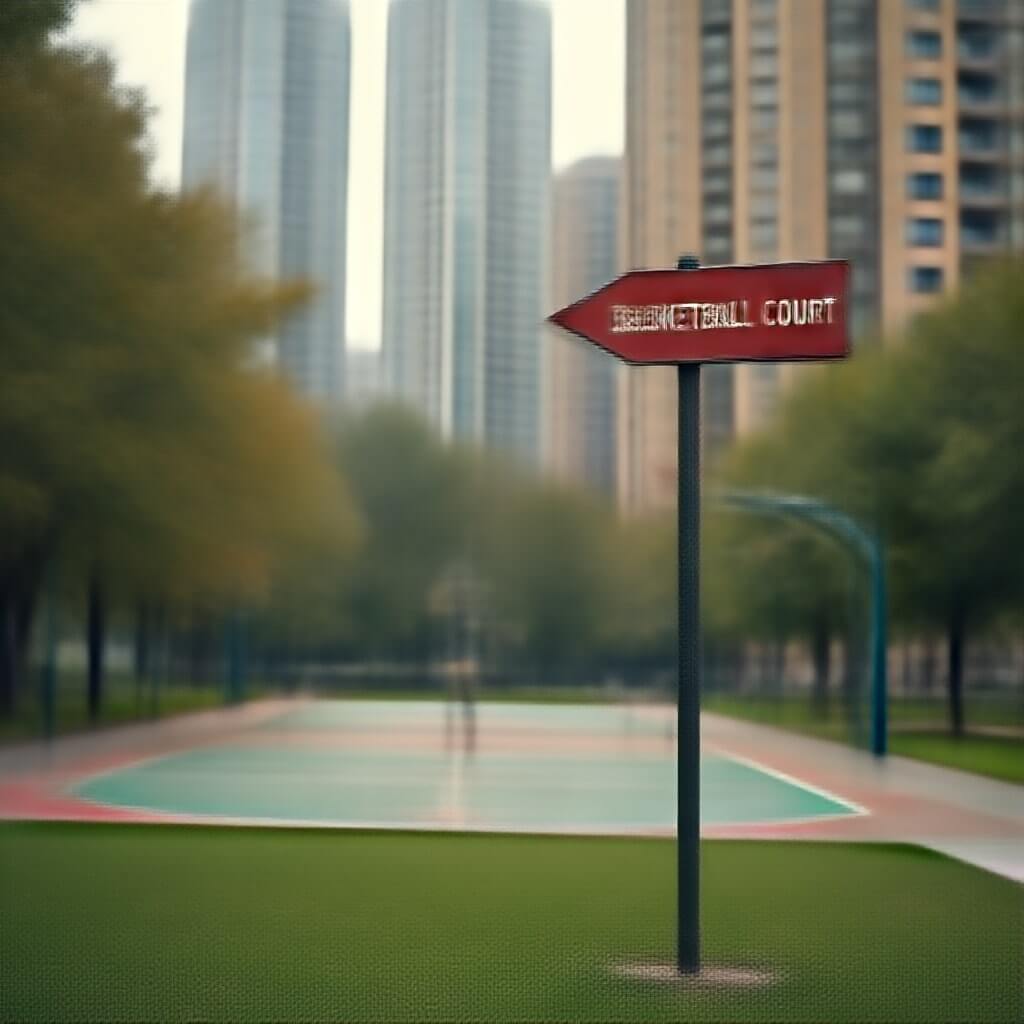} &
\includegraphics[width=0.130\textwidth, height=0.130\textwidth]{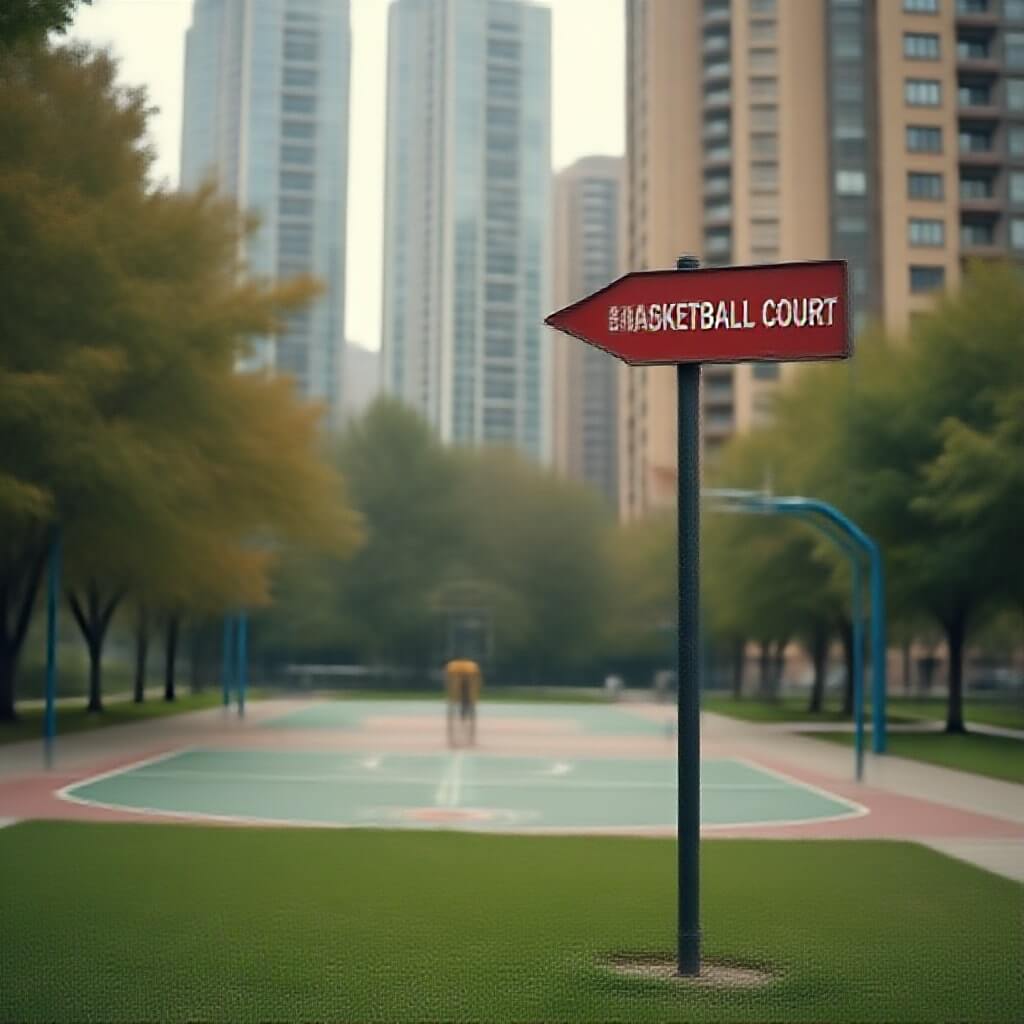}
& \includegraphics[width=0.130\textwidth, height=0.130\textwidth]{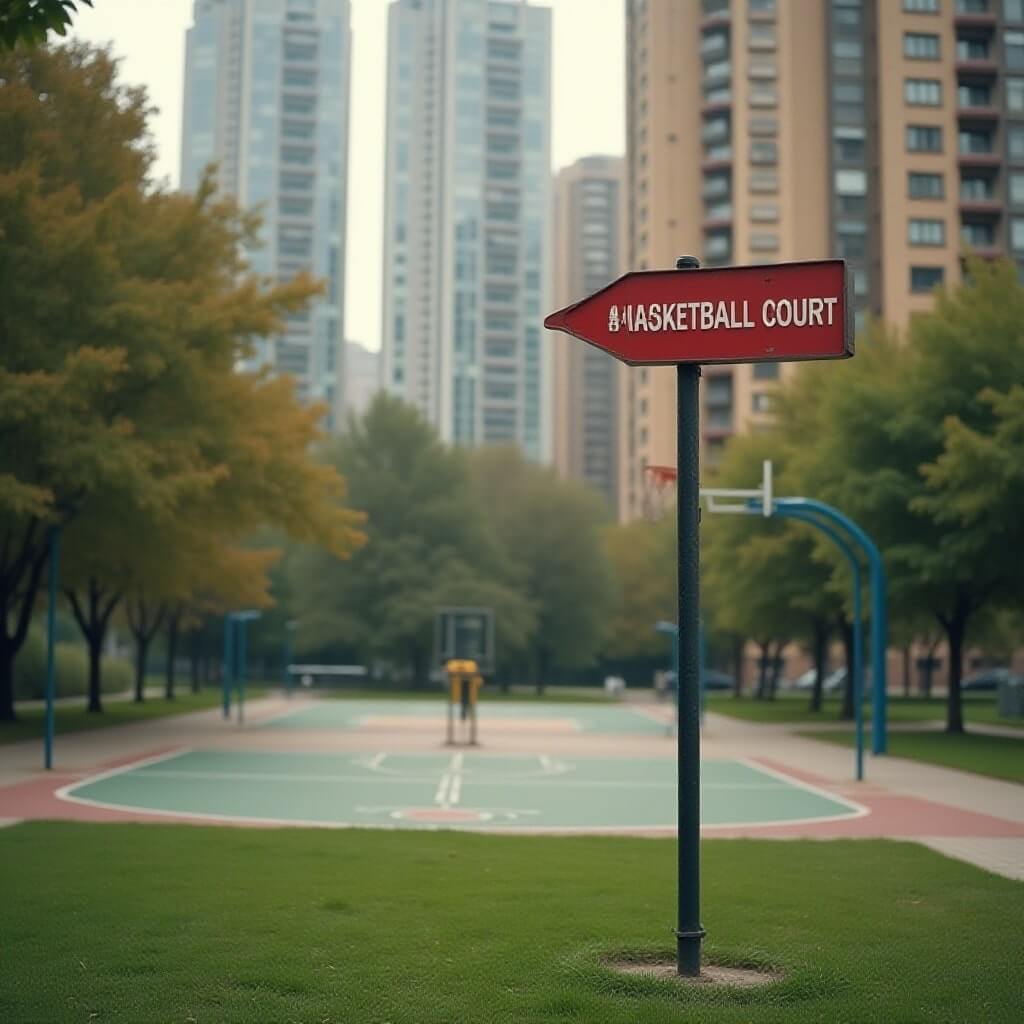} &
\includegraphics[width=0.130\textwidth, height=0.130\textwidth]{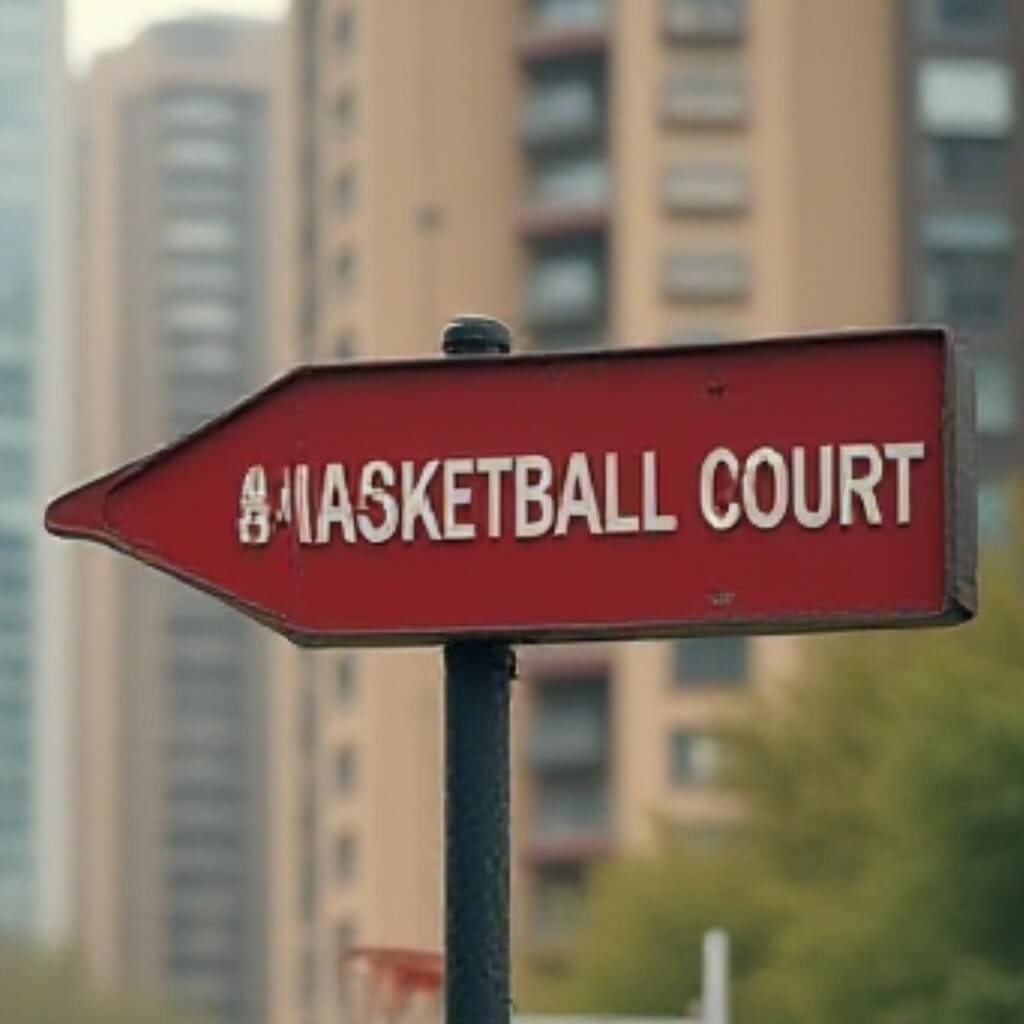} \\
\rotatebox{90}{{\quad \hspace{2mm}Artifact}} \rotatebox{90}{{\hspace{7mm}  Mask}} & \includegraphics[width=0.130\textwidth, height=0.130\textwidth]{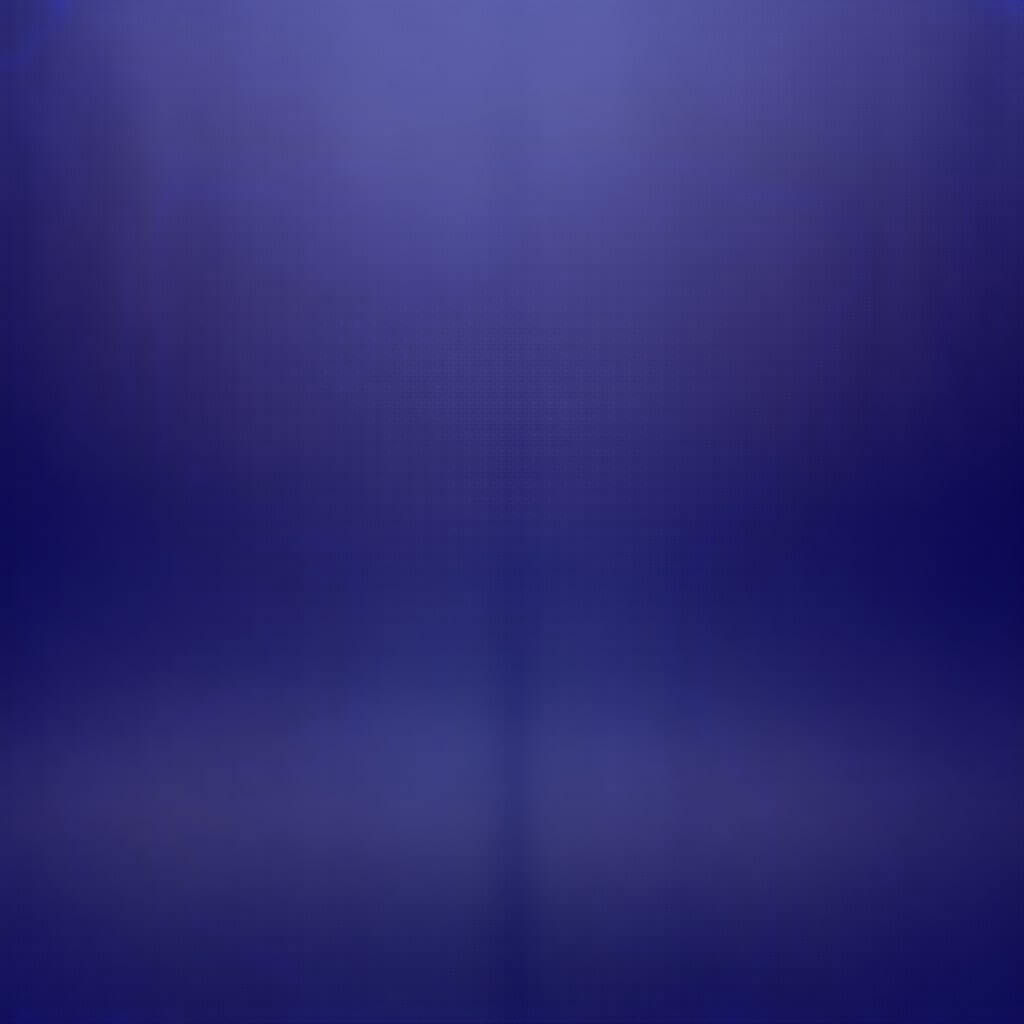} &
\includegraphics[width=0.130\textwidth, height=0.130\textwidth]{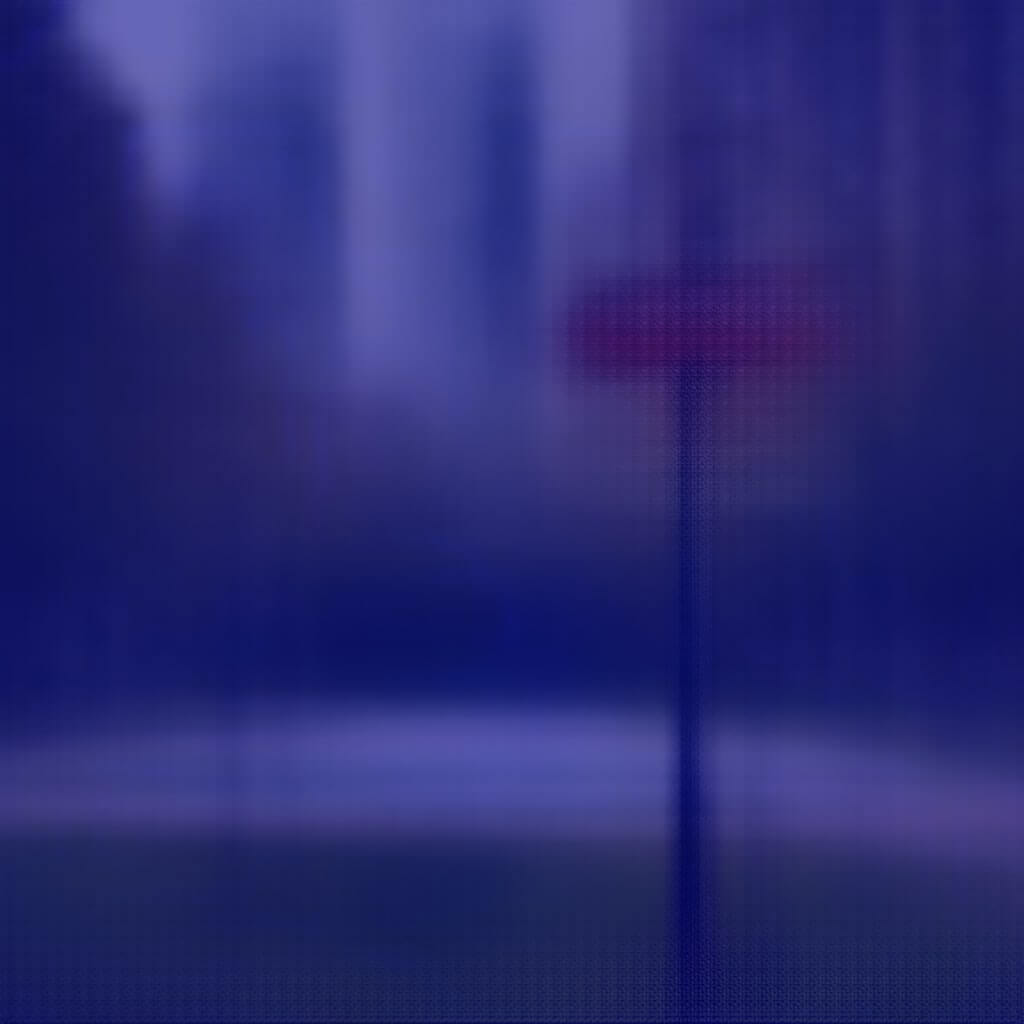} &
\includegraphics[width=0.130\textwidth, height=0.130\textwidth]{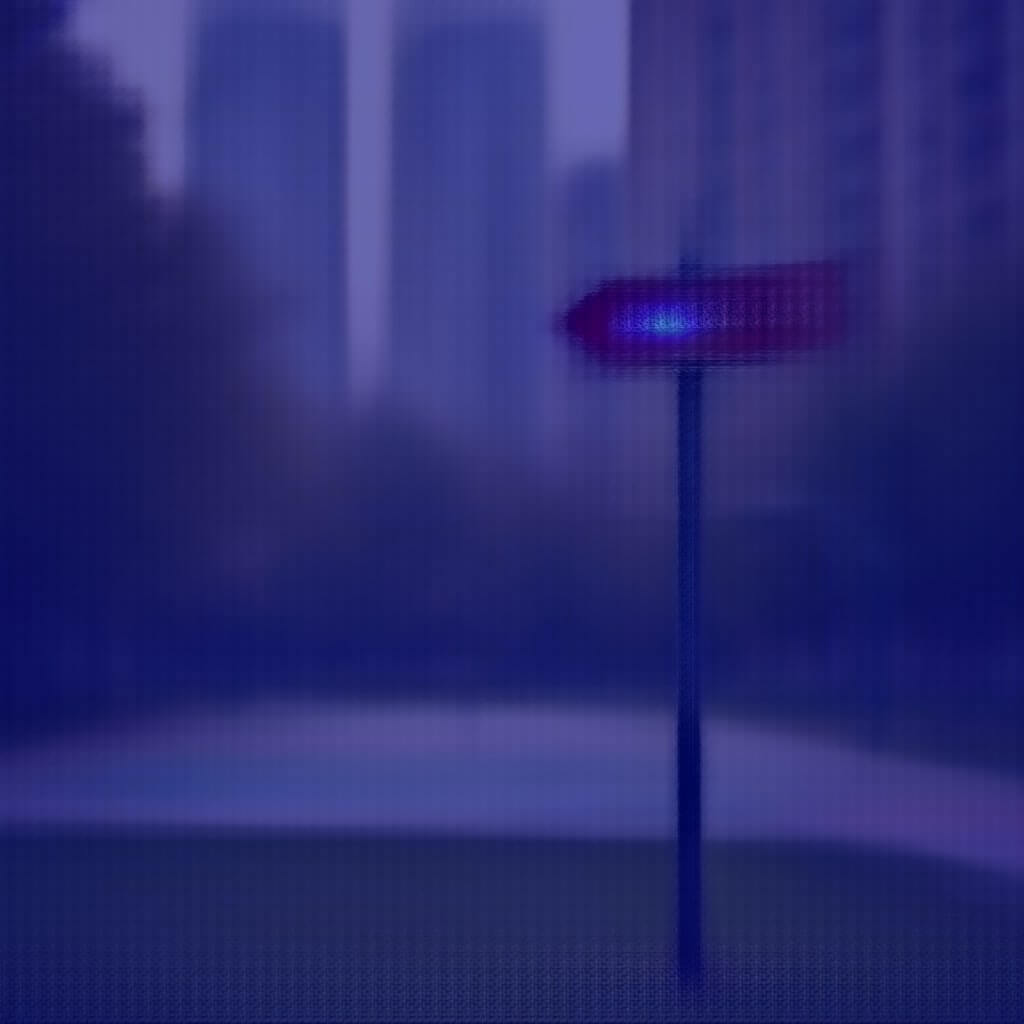} &
\includegraphics[width=0.130\textwidth, height=0.130\textwidth]{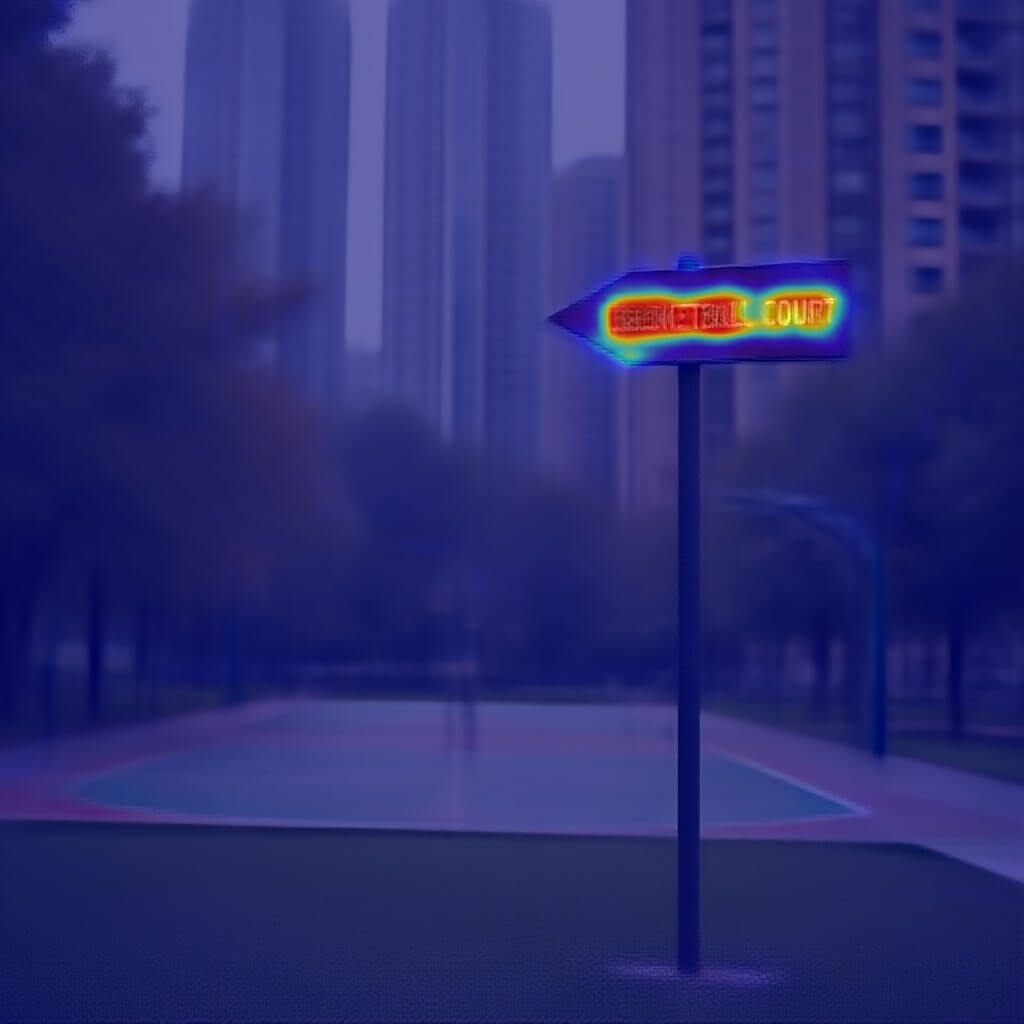} &
\includegraphics[width=0.130\textwidth, height=0.130\textwidth]{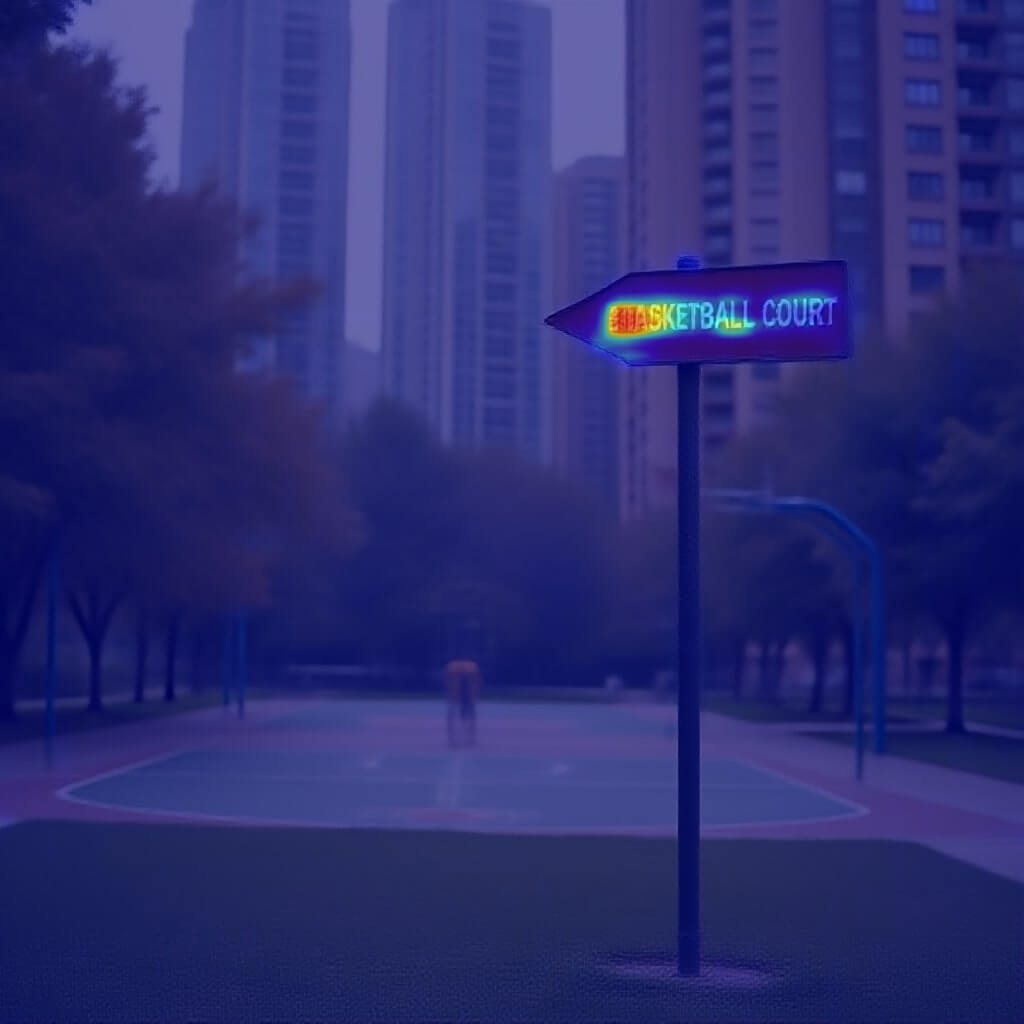}
& \includegraphics[width=0.130\textwidth, height=0.130\textwidth]{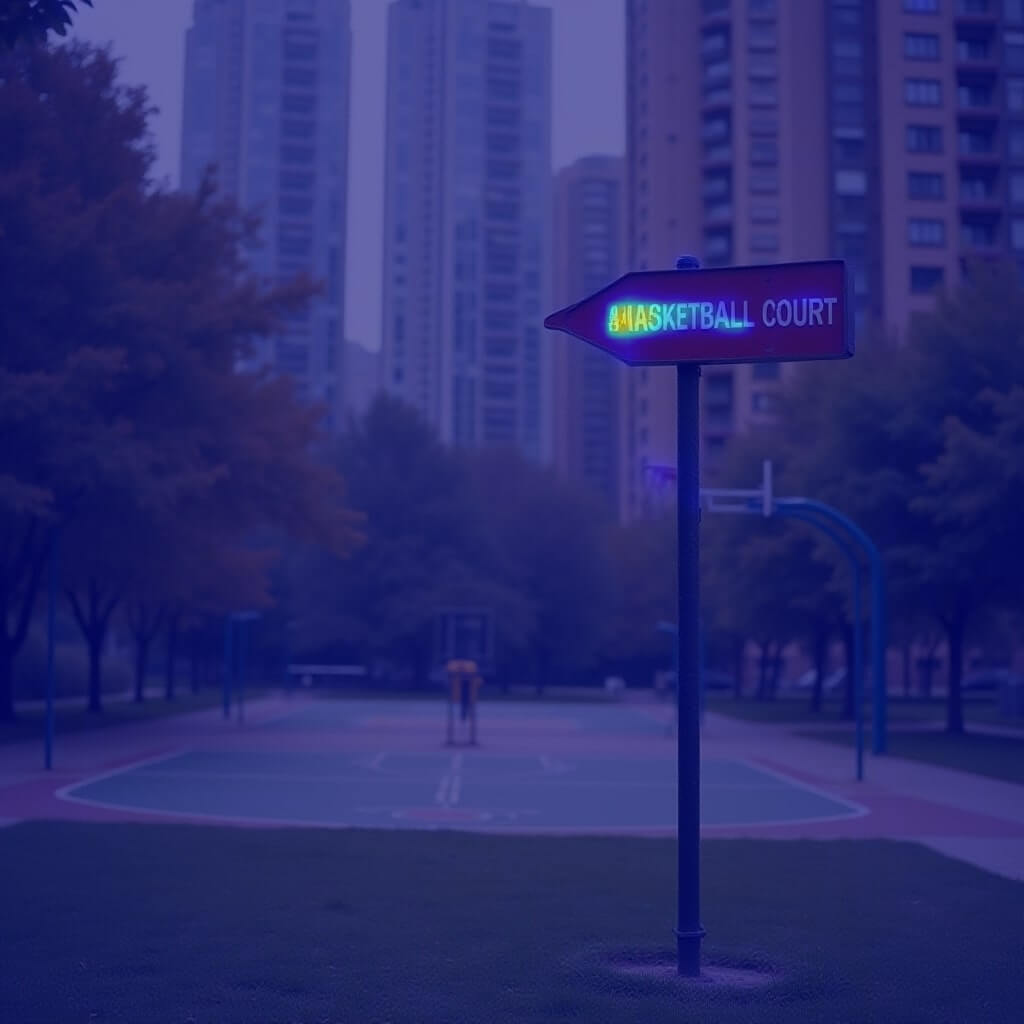} &
\includegraphics[width=0.130\textwidth, height=0.130\textwidth]{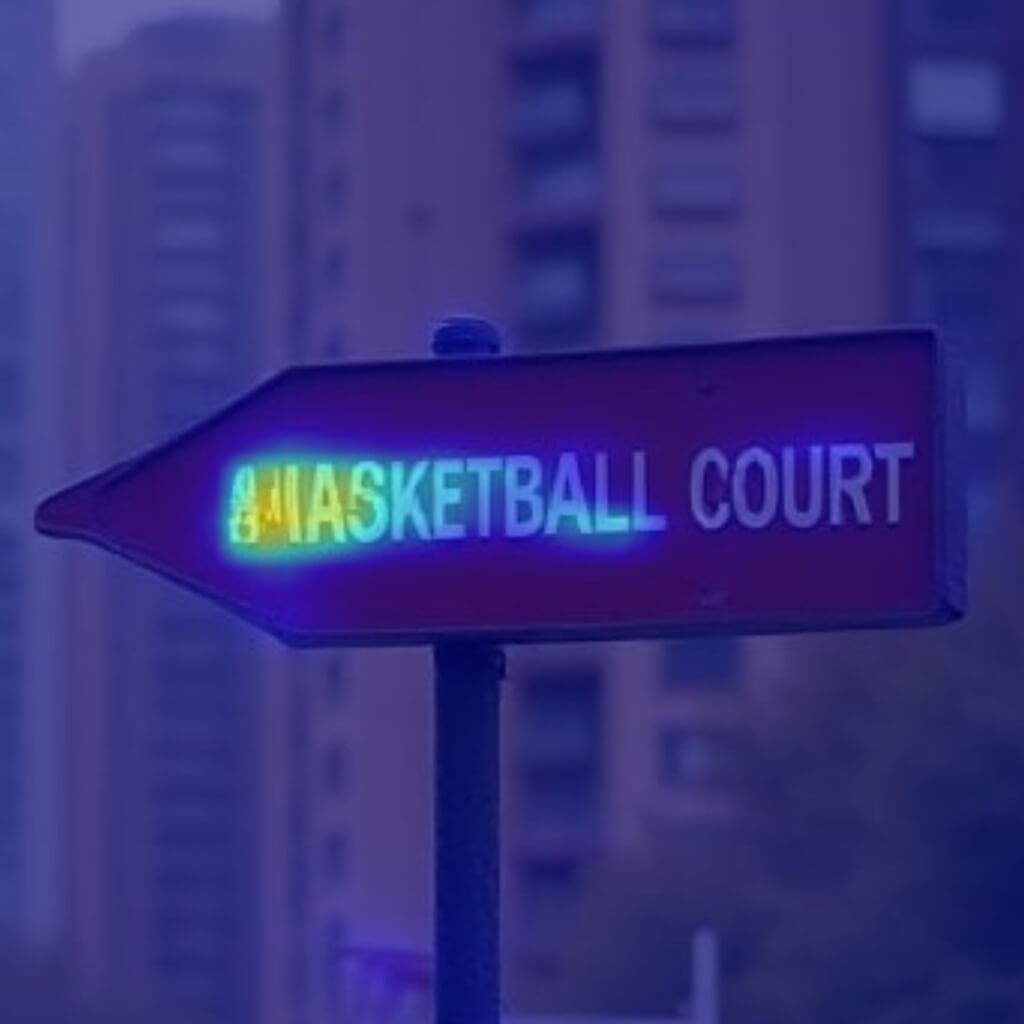} \\
\rotatebox{90}{\hspace{1mm}+\our{}} &  \includegraphics[width=0.130\textwidth, height=0.130\textwidth]{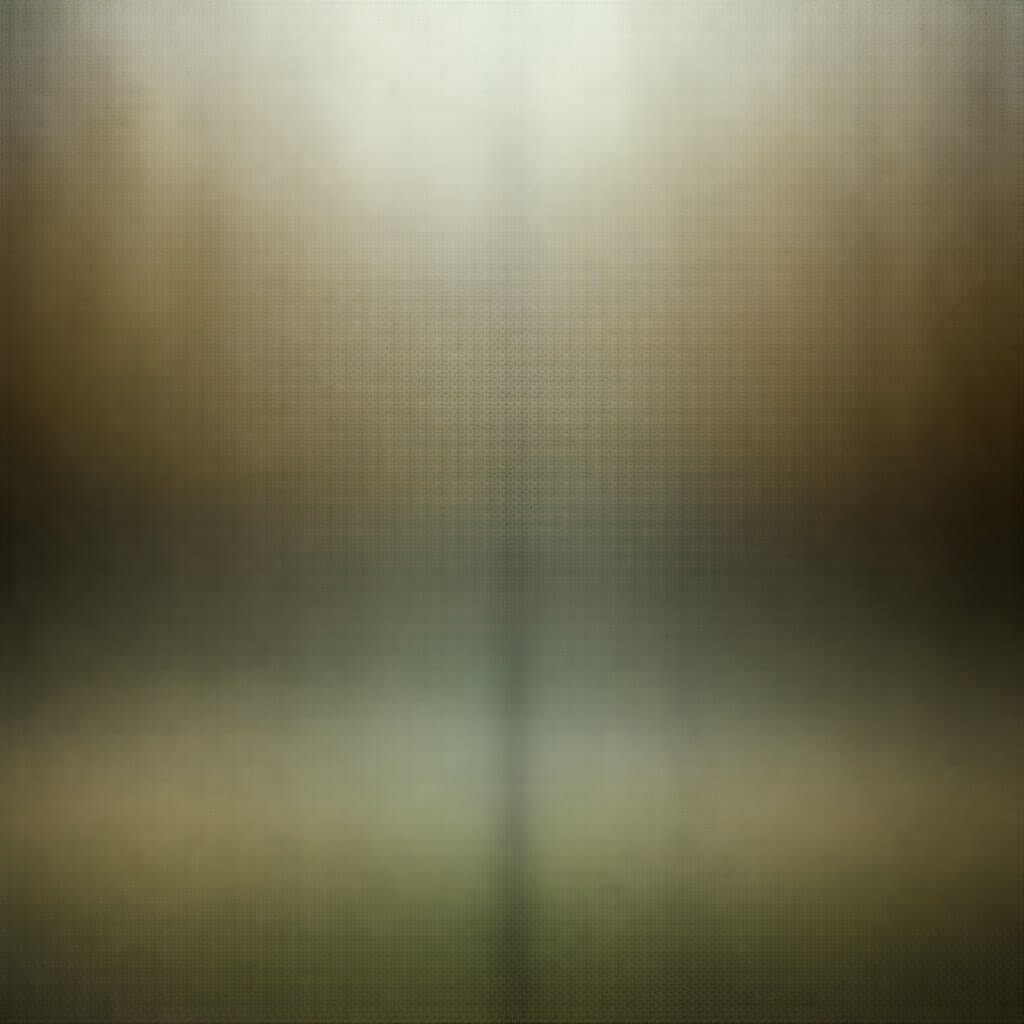} &
\includegraphics[width=0.130\textwidth, height=0.130\textwidth]{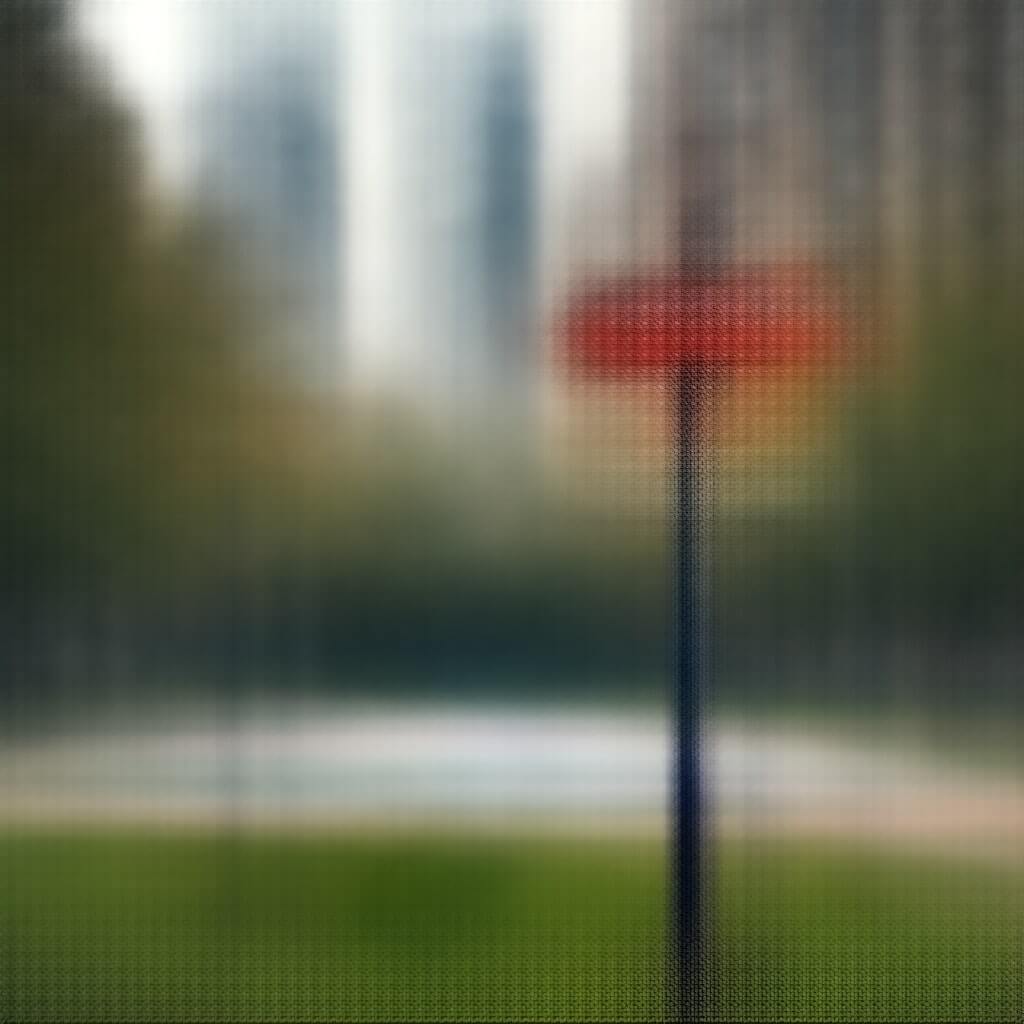} &
\includegraphics[width=0.130\textwidth, height=0.130\textwidth]{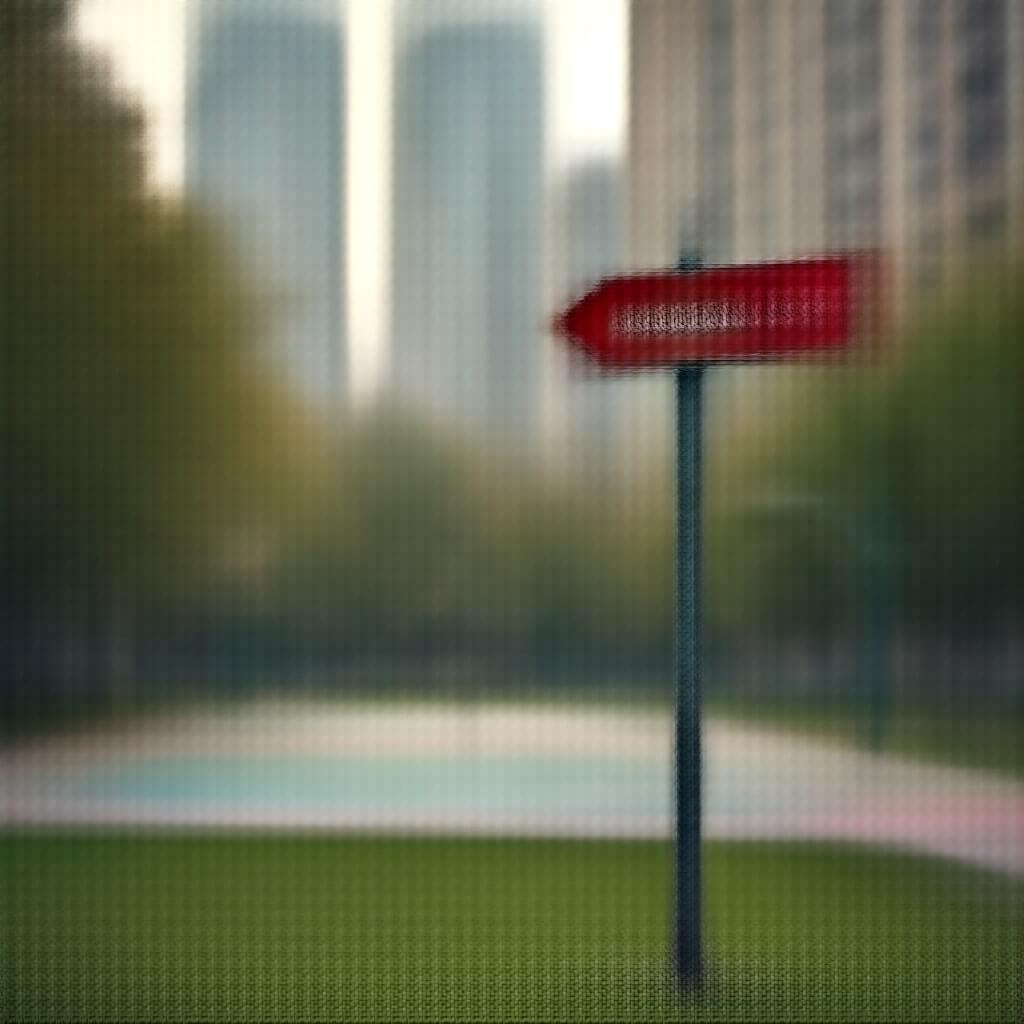} &
\includegraphics[width=0.130\textwidth, height=0.130\textwidth]{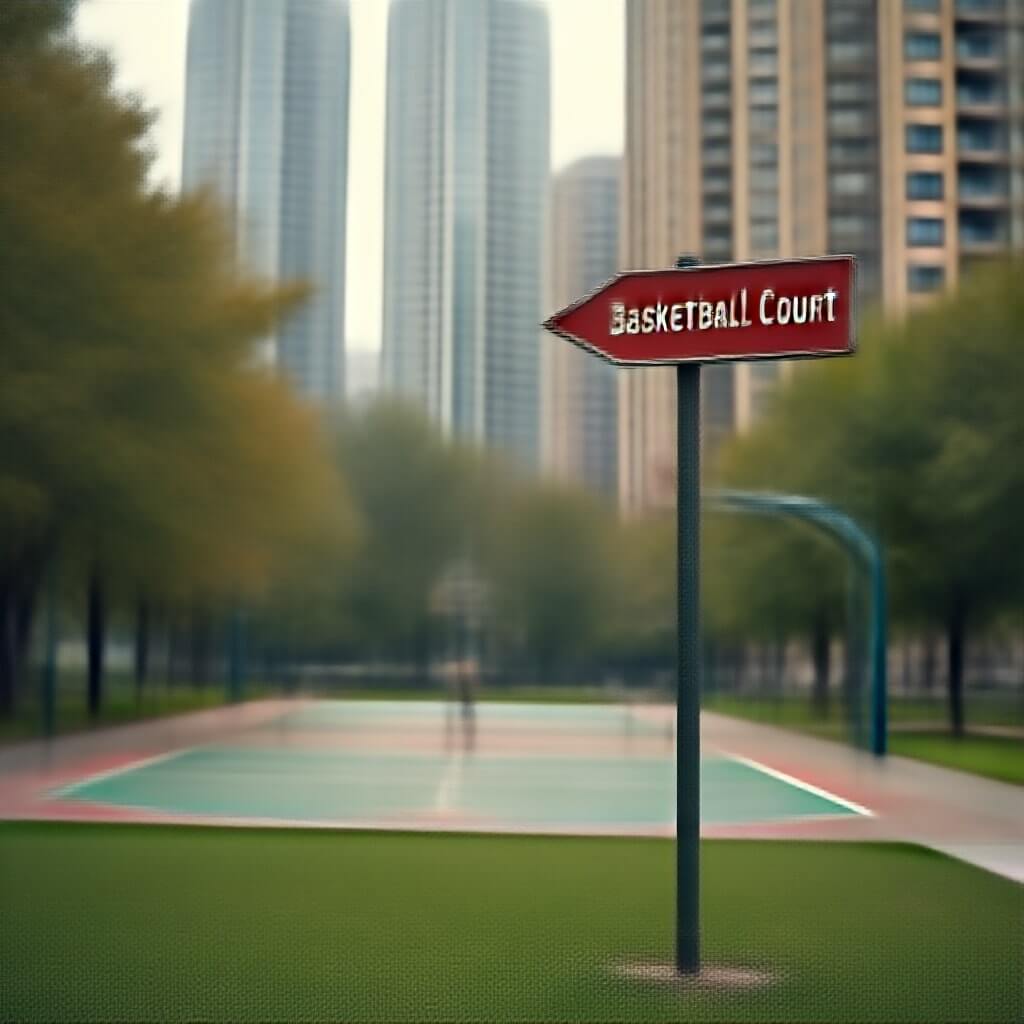} &
\includegraphics[width=0.130\textwidth, height=0.130\textwidth]{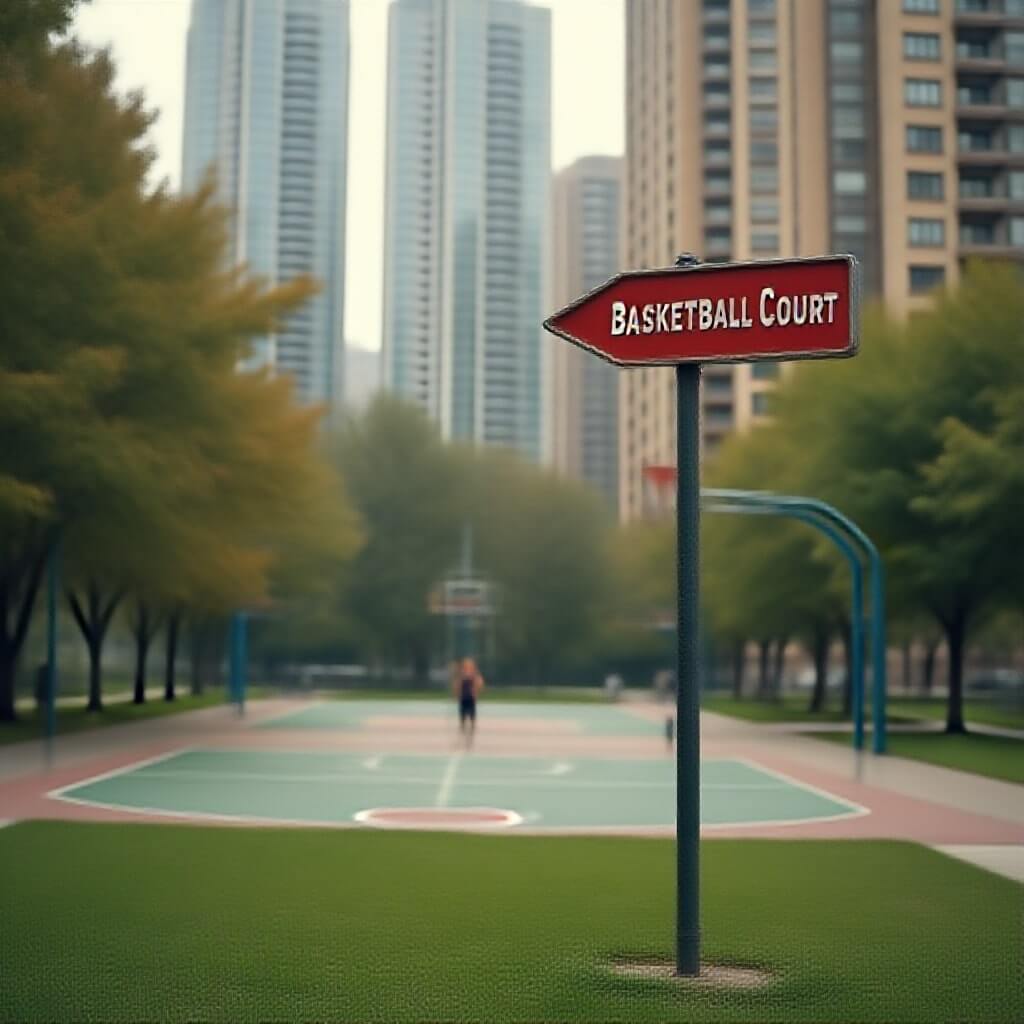}
& \includegraphics[width=0.130\textwidth, height=0.130\textwidth]{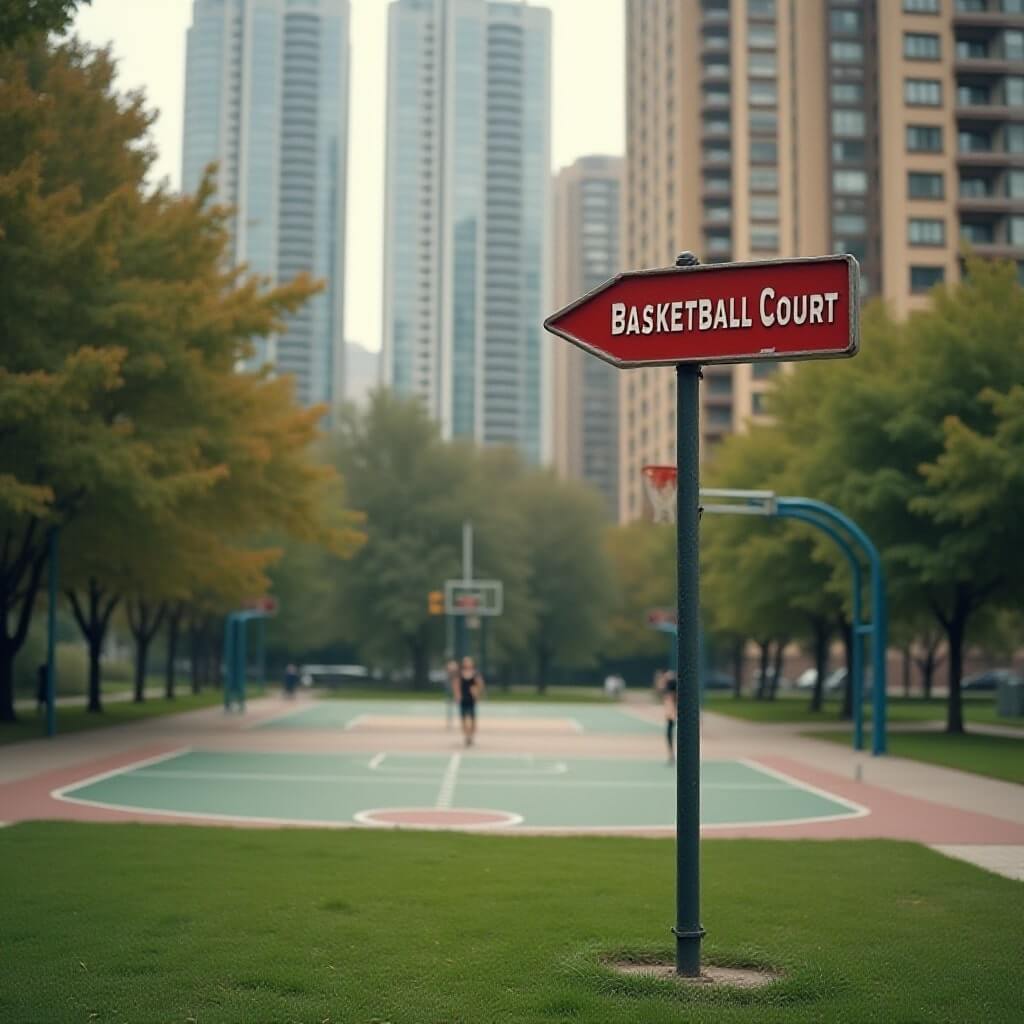} &
\includegraphics[width=0.130\textwidth, height=0.130\textwidth]{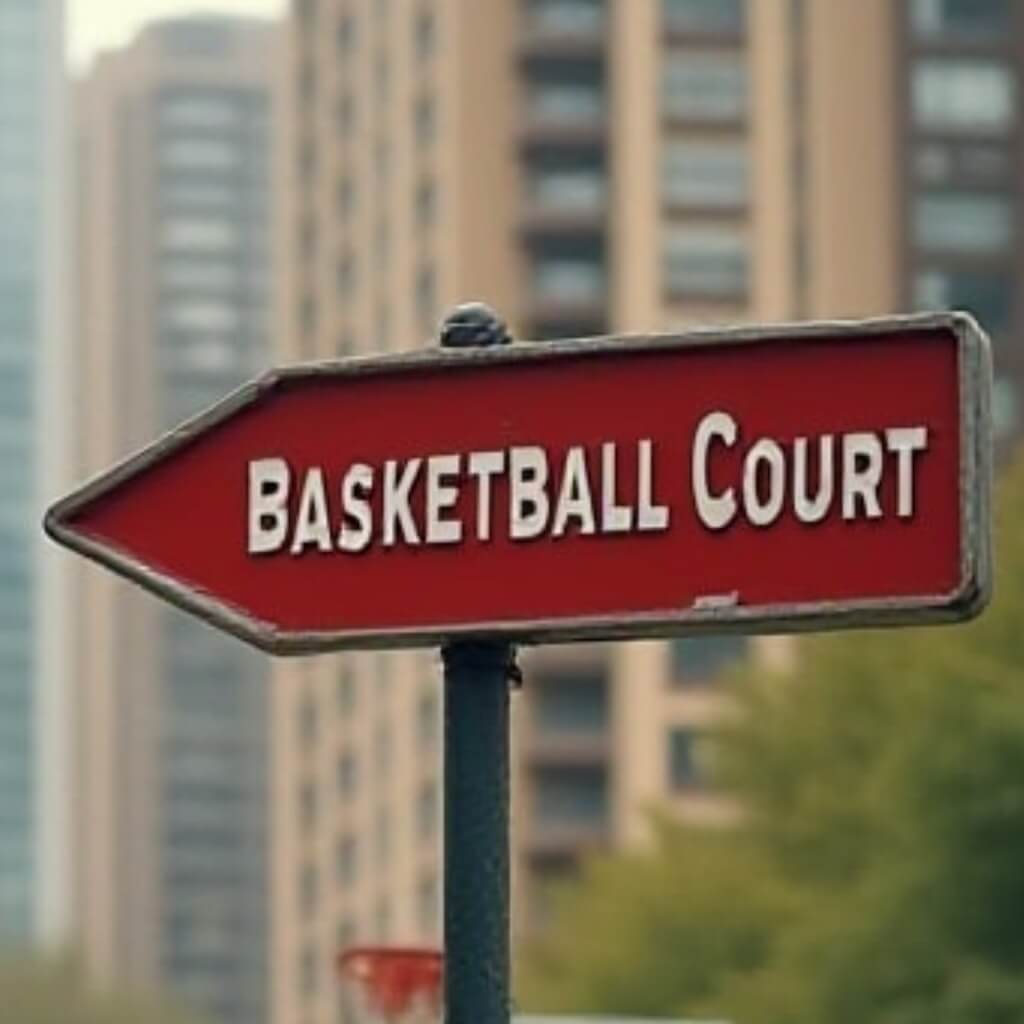} \\
\rotatebox{90}{{\quad \hspace{2mm} Artifact}} \rotatebox{90}{{\hspace{7mm}  Mask}} & \includegraphics[width=0.130\textwidth, height=0.130\textwidth]{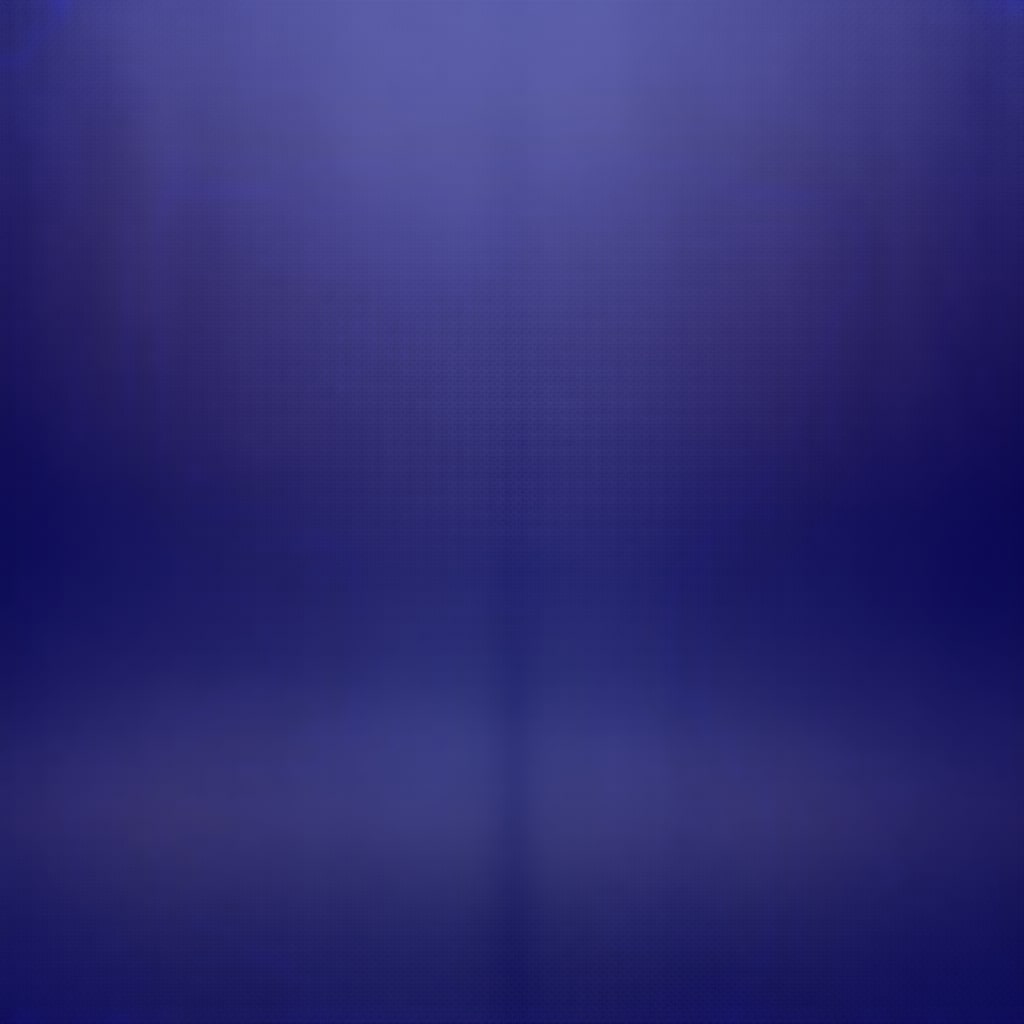} &
\includegraphics[width=0.130\textwidth, height=0.130\textwidth]{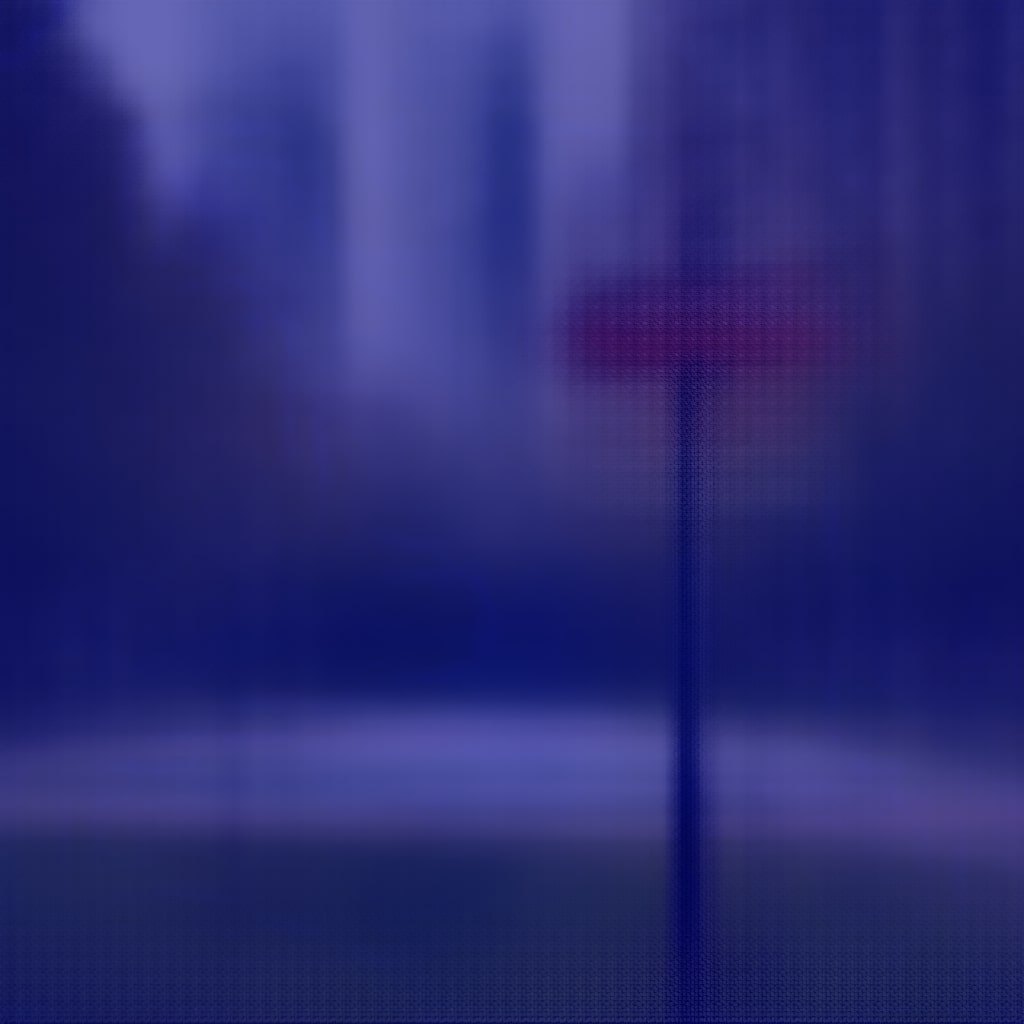} &
\includegraphics[width=0.130\textwidth, height=0.130\textwidth]{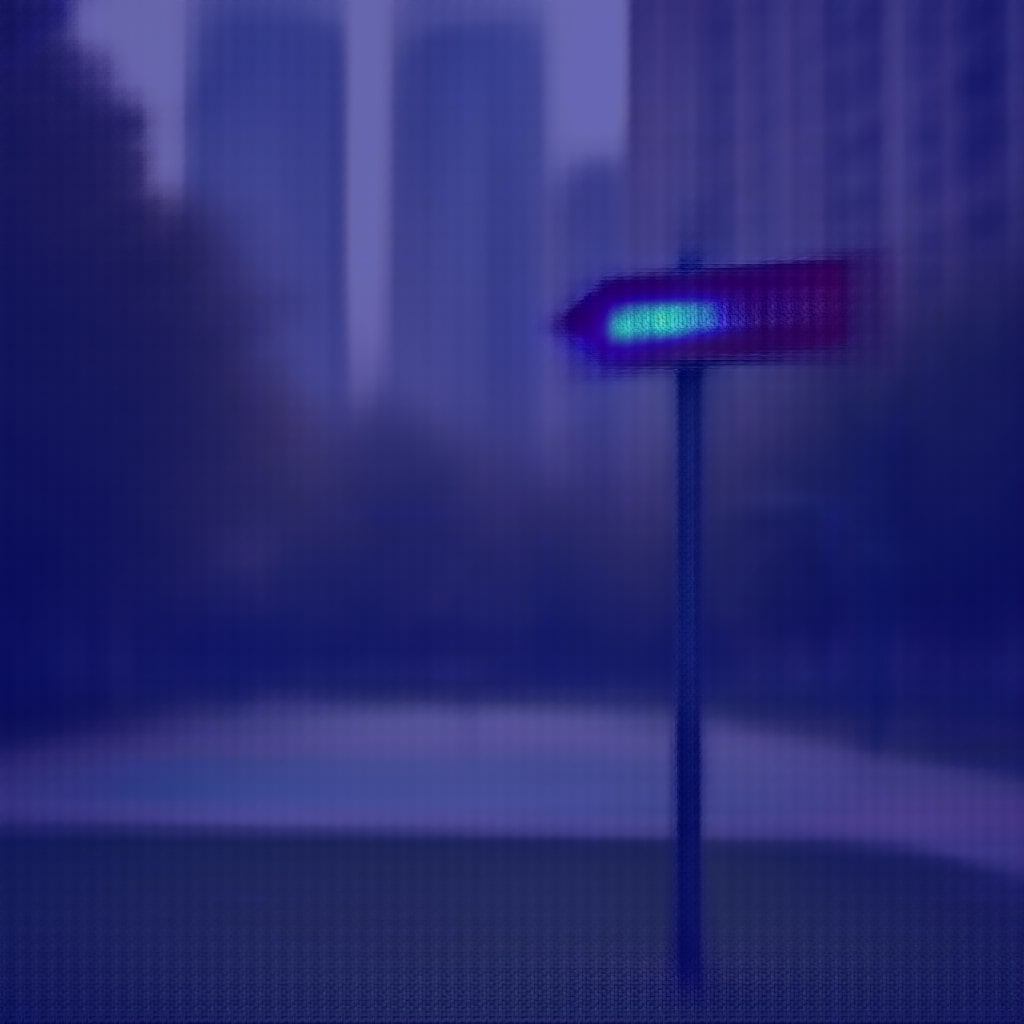} &
\includegraphics[width=0.130\textwidth, height=0.130\textwidth]{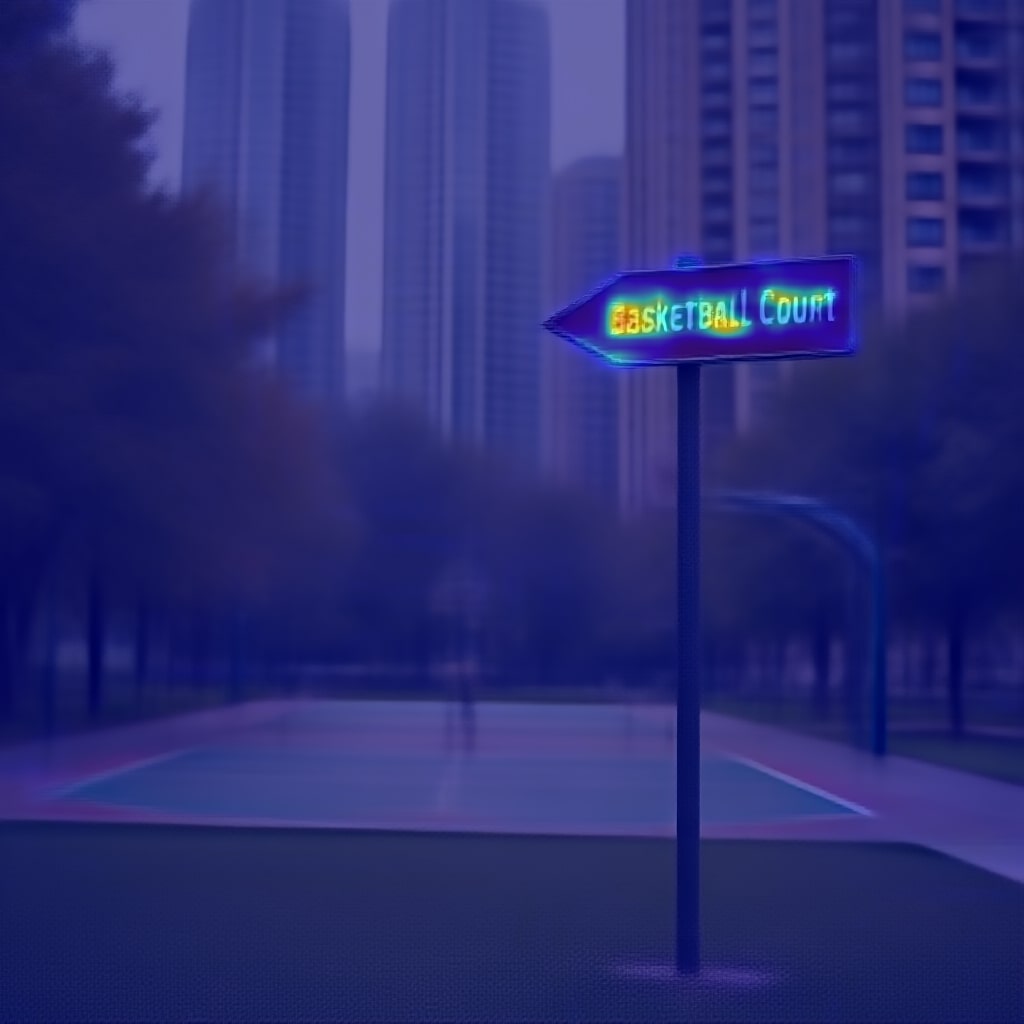} &
\includegraphics[width=0.130\textwidth, height=0.130\textwidth]{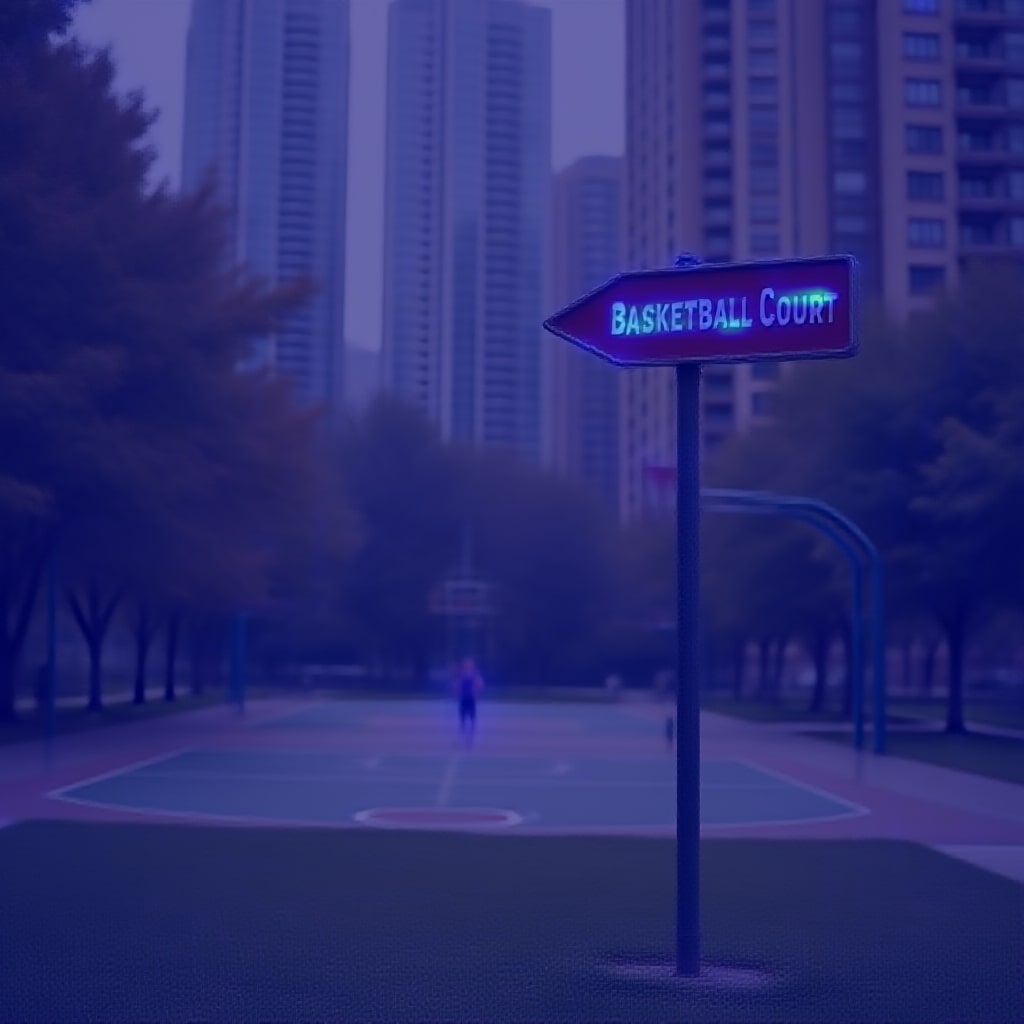}
& \includegraphics[width=0.130\textwidth, height=0.130\textwidth]{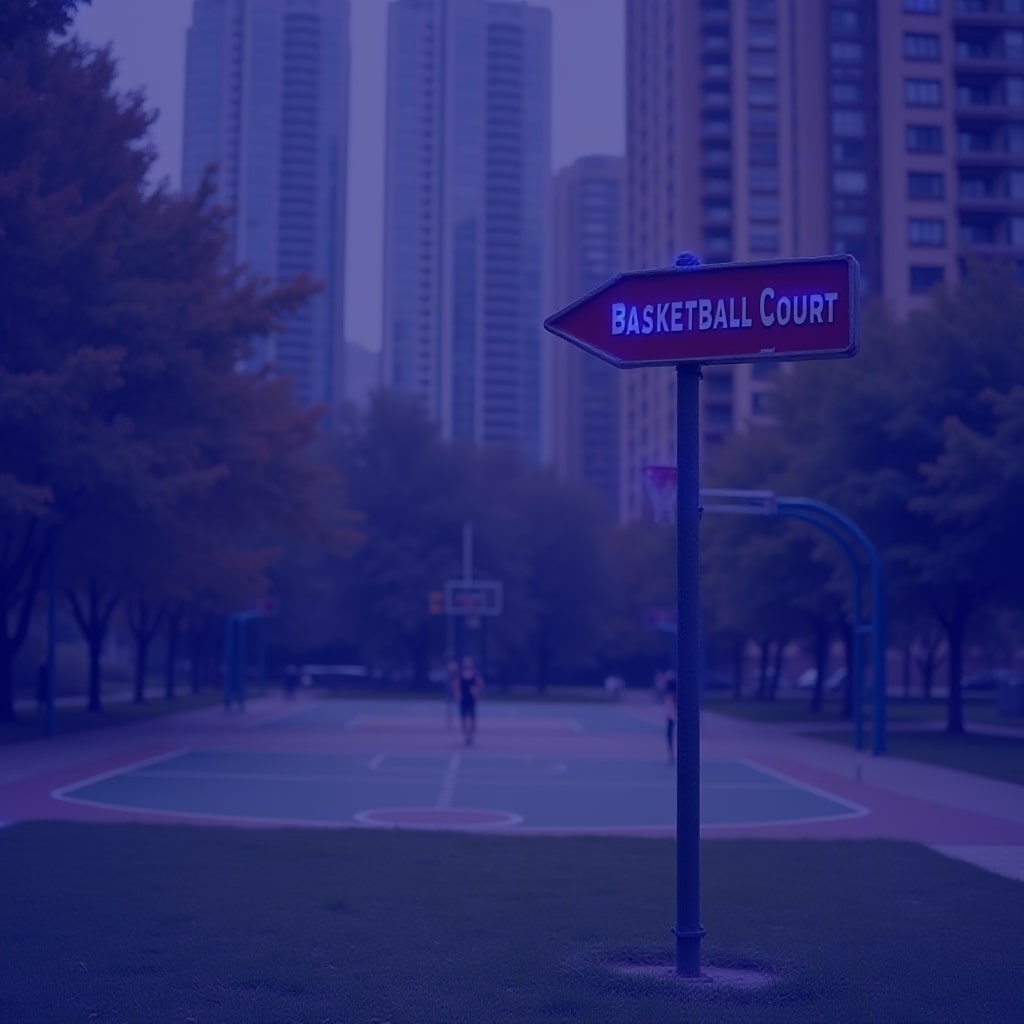} &
\includegraphics[width=0.130\textwidth, height=0.130\textwidth]{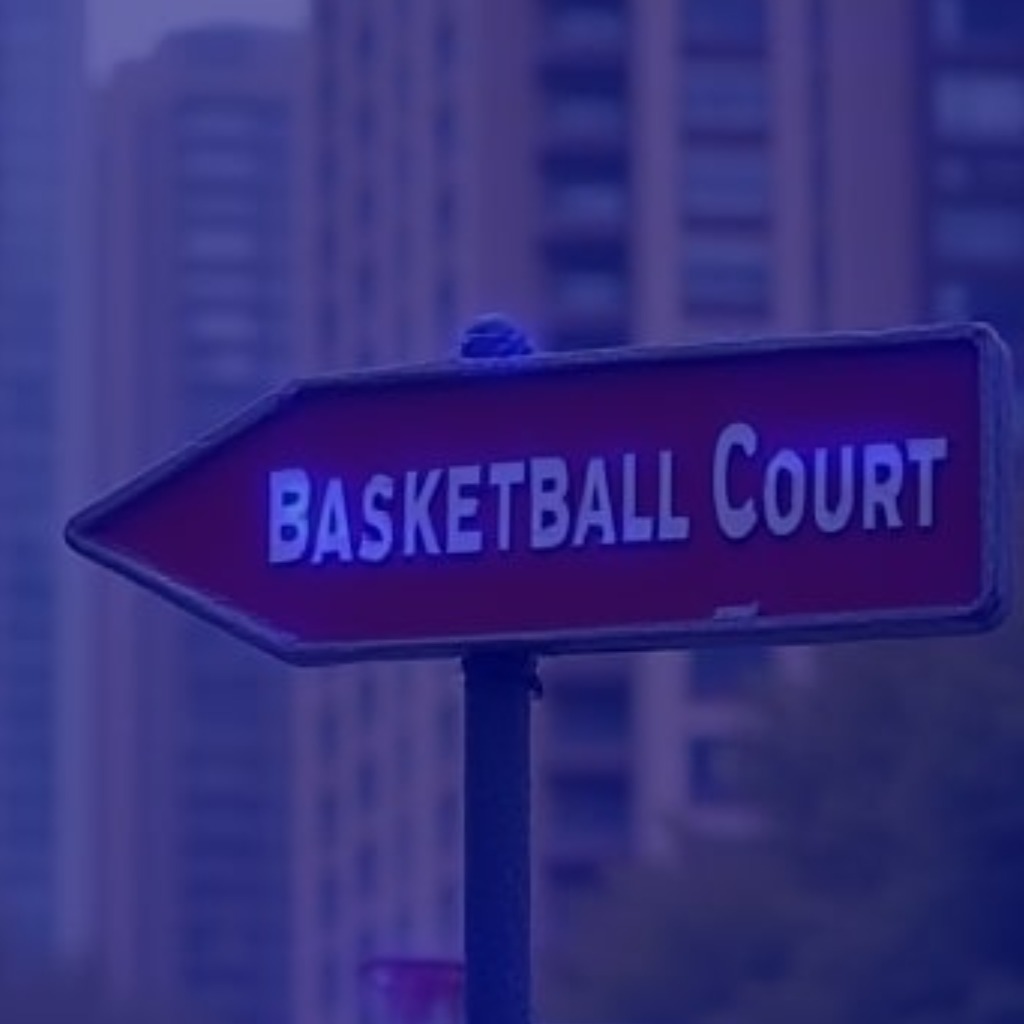}
\end{tabular}
\caption{\textbf{Baseline (top) and proposed (bottom) generation trajectories illustrating artifact reduction on FLUX.1 [dev]} During image generation, \our{} corrects the trajectories by improving the plate. Images for prompt: "A city park with a signpost pointing to \textit{"Basketball Court,"} urban, matte, sharp focus, vibrant colors."
}
\label{fig:maskintiem}
\end{figure*}

\newpage

\subsection{Text artifacts}

In FLUX.1 [dev], text in an image using 10 inferences can introduce small perturbations in the trajectory, leading to local character-level errors that are not ultimately corrected during generation. \our{} can correct these artifacts as shown in Figs. \ref{fig:dev_text2} and \ref{fig:dev_text}. We can distinguish many types of textual errors, such as letter substitutions, omitted or duplicated characters, ambiguous characters resembling other letters, such as 0 and O, local blurring of letters in other parts of the image, and inconsistencies between different versions of the same text. Additionally, although the text can be generated correctly, it is possible to have uneven letter widths, inconsistent spacing, or misalignment where one part of the text falls off. Unlike other methods, \our{} corrects these errors by stabilizing and improving the sampling trajectory, which allows for obtaining consistent text that was introduced in the prompt. Fig. \ref{fig:maskintiem} presents visualizations of the correction for improving the word "Basketball." The fading artifact mask during forward-pass is clearly visible, symbolizing the dynamic correction of intermediate states.

\begin{figure*}[!h]
\centering
\setlength{\tabcolsep}{1.2pt}
\renewcommand{\arraystretch}{0.9}
\begin{tabular}{cccc}
\rotatebox{90}{\hspace{+10mm}+\our{} \hspace{+60pt} FLUX.2 [dev]} &
\includegraphics[width=0.48\textwidth]{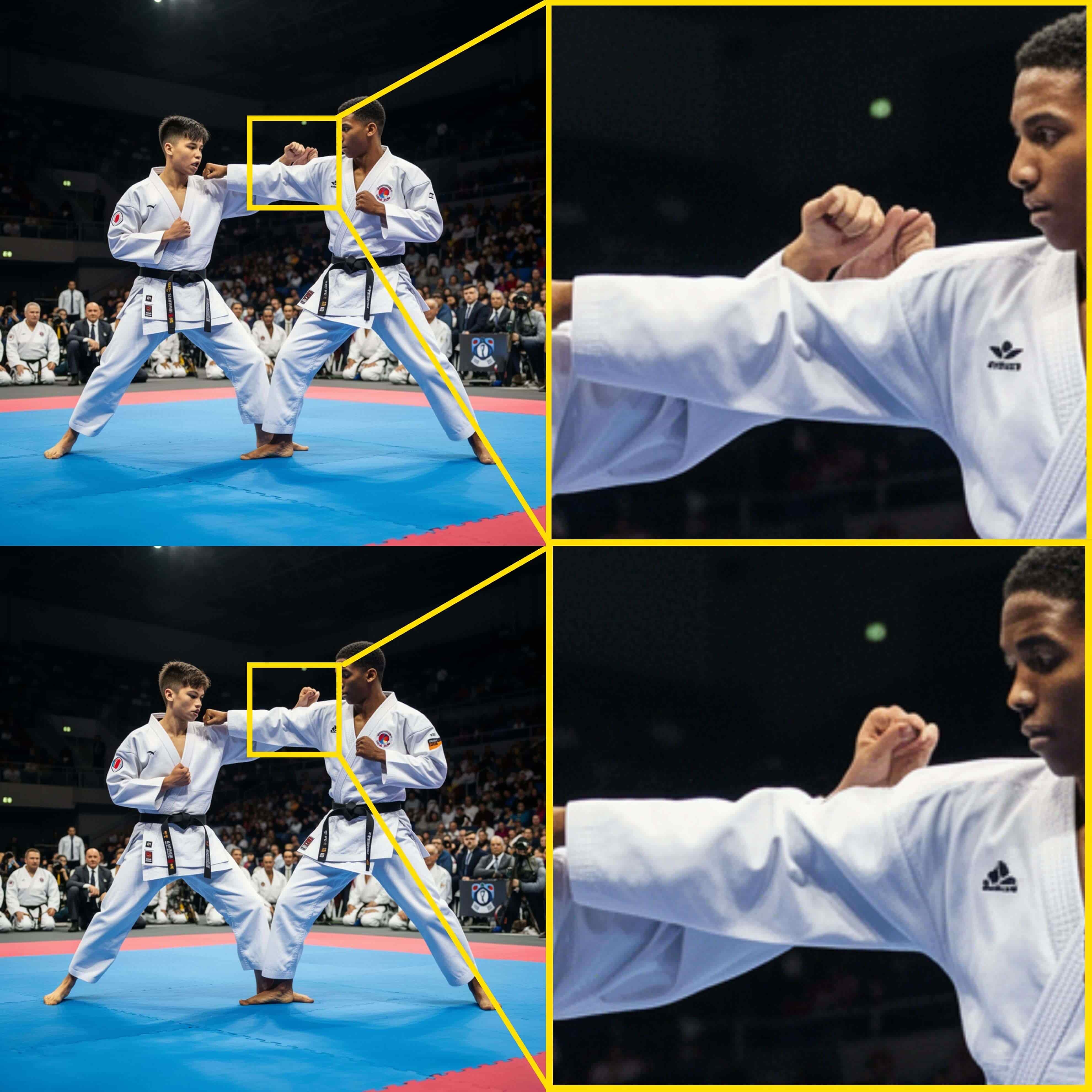} &
&
\includegraphics[width=0.48\textwidth]{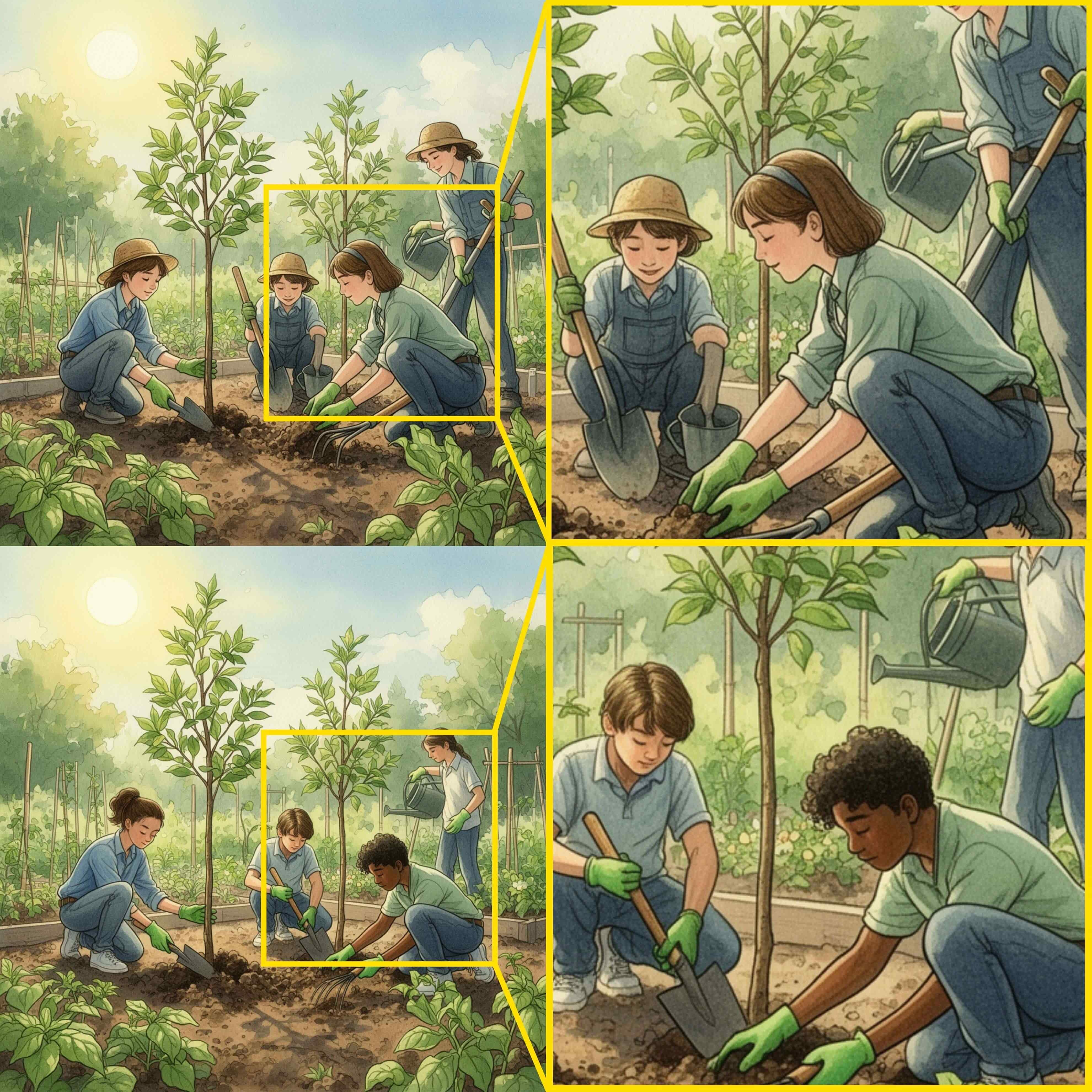} \\
\rotatebox{90}{\hspace{+10mm} +\our{} \hspace{+55pt} FLUX.2 [dev]}  &
\includegraphics[width=0.48\textwidth]{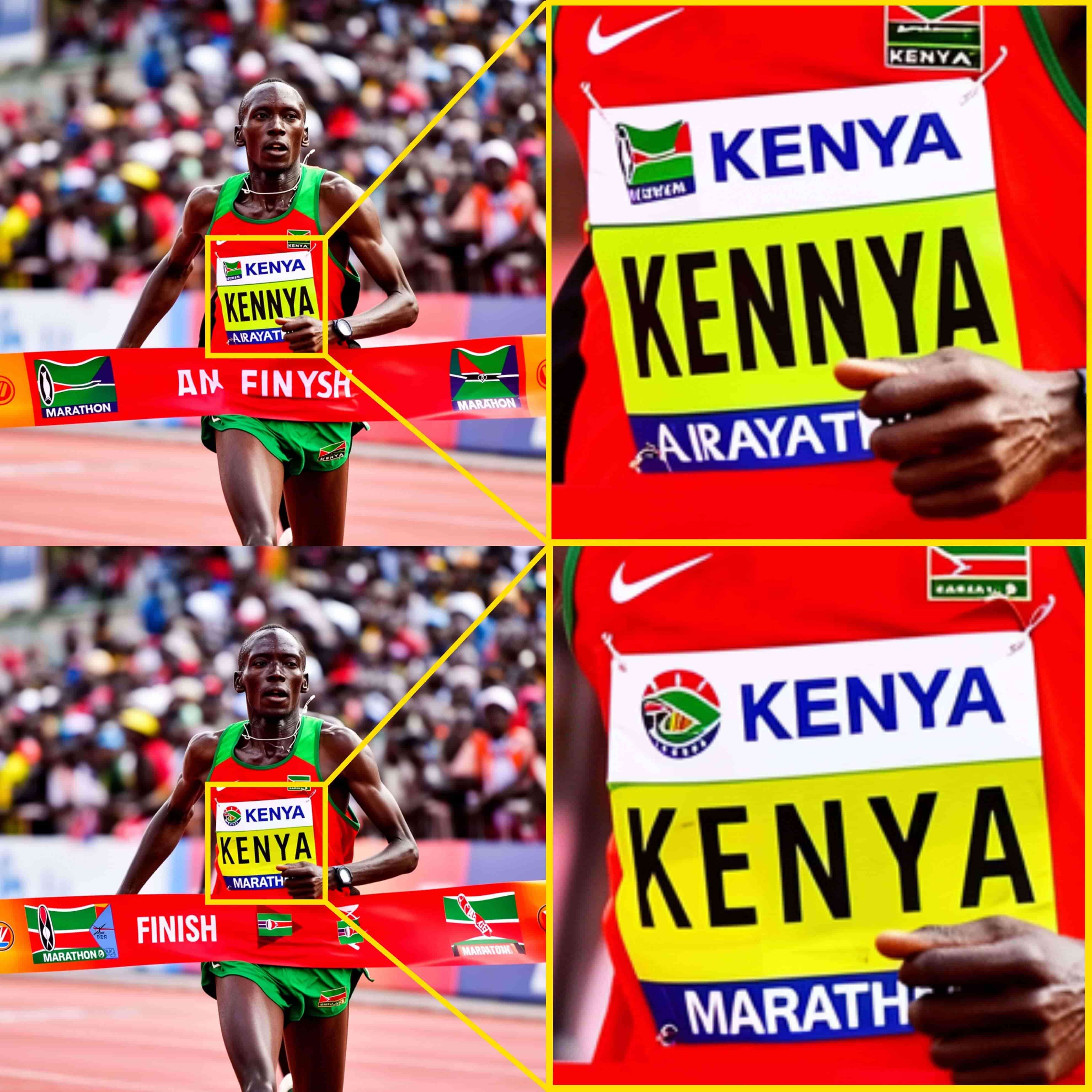} &
&
\includegraphics[width=0.48\textwidth]{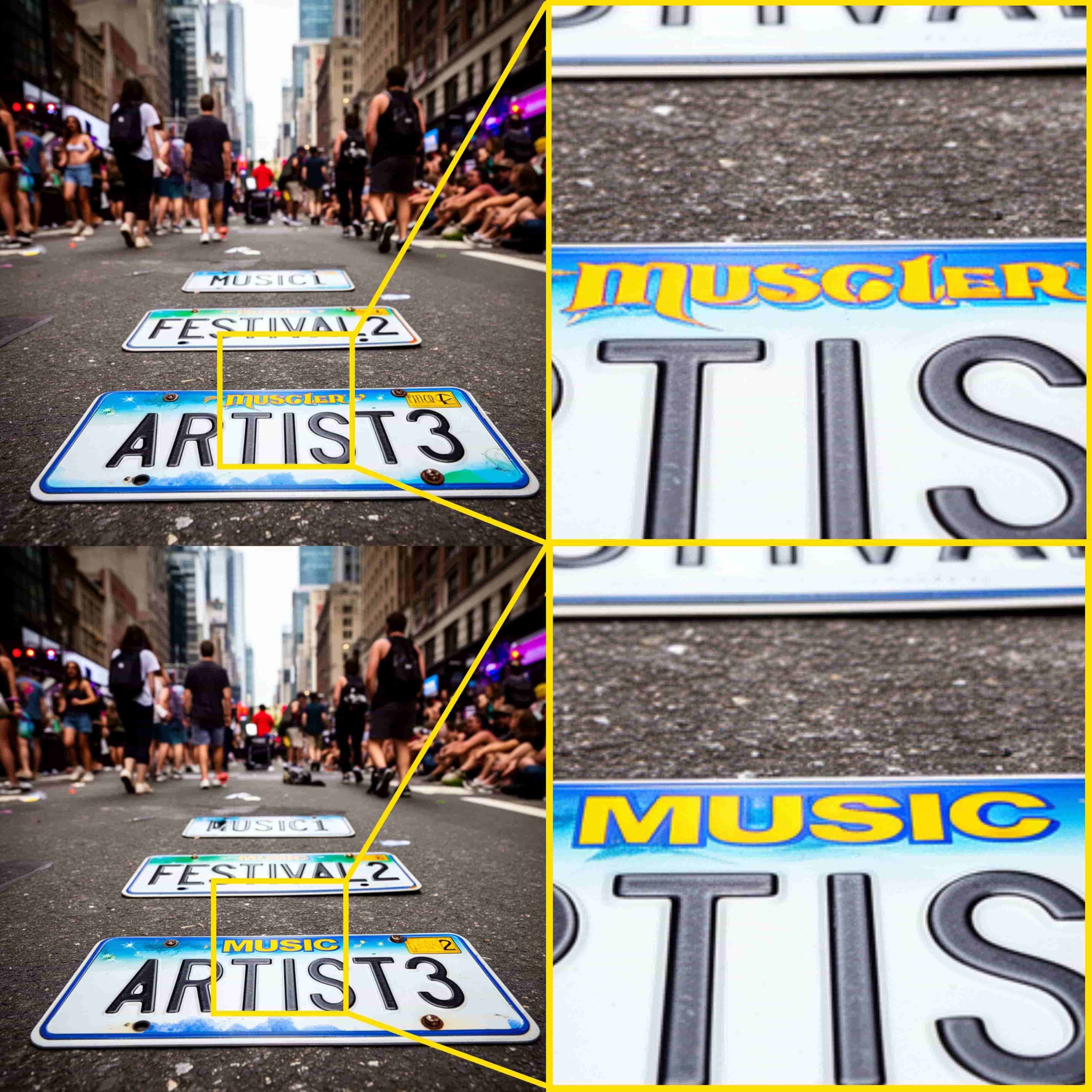}
\end{tabular}
\caption{\textbf{Images for the FLUX.2 [dev] using \our{} for 30 inference steps.} The FLUX.2 [dev] generates images with a high level of realism, with only a small fraction of them containing minor artifacts. The hyperparameters of the model used to generate these images are presented in Tab. \ref{tab:hyperparameters}. }
\label{fig:flux2_images}
\vspace{-0.3cm}
\end{figure*}

\newpage
\clearpage

\subsection{Animal artifacts}
\label{subsec:animals_art}

\our{} can correct artifacts in images of animals, as shown in Figs. \ref{fig:maskintiem_flux2}, \ref{fig:flux2_animal} and \ref{fig:dev_animal}, and in Tab. \ref{tab:animals_dev}. By focusing the generation process from the beginning, the model maintains semantic consistency and avoids unintended hybrids. For example, in prompts containing both an animal and a human, it does not accidentally generate a human-animal character (see example of cat and lawyer on Fig. \ref{fig:dev_animal}).

In the case of combined prompts, we observe that the model can adaptively shift the anatomy towards one of the semantic components, which can result in skin being replaced by fur. This is a result of the prompt's semantic ambiguity, see Fig. \ref{fig:flux2_animal}.

\begin{table*}[!h]
\centering
\caption{\textbf{Evaluation of FLUX.1 [dev] on the \textit{animals} dataset.} Results for our technique compared to other state-of-the-art models. }

\setlength{\tabcolsep}{4.8pt}
{\fontsize{7pt}{11pt}\selectfont
\begin{tabular}{lccccccc}
\hline
Model                              & CLIP-T $\uparrow$  & Mean Artifact Freq (\%) $\downarrow$ & ImageReward $\uparrow$ & Artfiact Pixel Ratio (\%) $\downarrow$ & MAE $\downarrow$  & MAE (A) $\downarrow$           & MAE (NA) $\downarrow$         \\ \hline
Flux.1 [dev]           &\textbf{ 38.888 $\pm$ 0.340} & 100.000 $\pm$ 0.000             & 1.228 $\pm$ 0.053      & 0.460 $\pm$ 0.078                 & - & - & - \\
+ DiffDoctor                       & 38.858 $\pm$ 0.155 & 58.653 $\pm$ 4.837              & \textbf{1.232 $\pm$ 0.063}      & 0.254 $\pm$ 0.041                 & 10.600 $\pm$ 1.130 & \textbf{25.015 $\pm$ 0.982}    & 10.524 $\pm$ 1.123                  \\
+ HPSv2                            & 38.231 $\pm$ 0.335 & 75.960 $\pm$ 1.920              & 1.122 $\pm$ 0.045      & 0.462 $\pm$ 0.140                 & 15.423 $\pm$ 1.293 & 32.098 $\pm$ 1.697  &  15.344 $\pm$ 1.296                 \\
\textbf{+ \our{}} & 38.797 $\pm$ 0.127 & \textbf{23.078 $\pm$ 10.880}             & 1.220 $\pm$ 0.066      & \textbf{0.062 $\pm$ 0.028}                 & \textbf{9.293 $\pm$ 0.922} & 26.292 $\pm$ 1.173 & \textbf{9.215 $\pm$ 0.915}                  \\ \hline
\end{tabular}
}
\label{tab:animals_dev}
\end{table*}

\begin{figure*}[!h]
\centering
\setlength{\tabcolsep}{1.2pt}
\renewcommand{\arraystretch}{0.9}
\begin{tabular}{ccccccccc}
 & $t_{29}$ & $t_{24}$ & $t_{19}$ & $t_{14}$ & $t_{9}$ & $t_{4}$ & $t_{0}$ & Final \\
\rotatebox{90}{FLUX.2 [dev]} &
\includegraphics[width=0.115\textwidth]{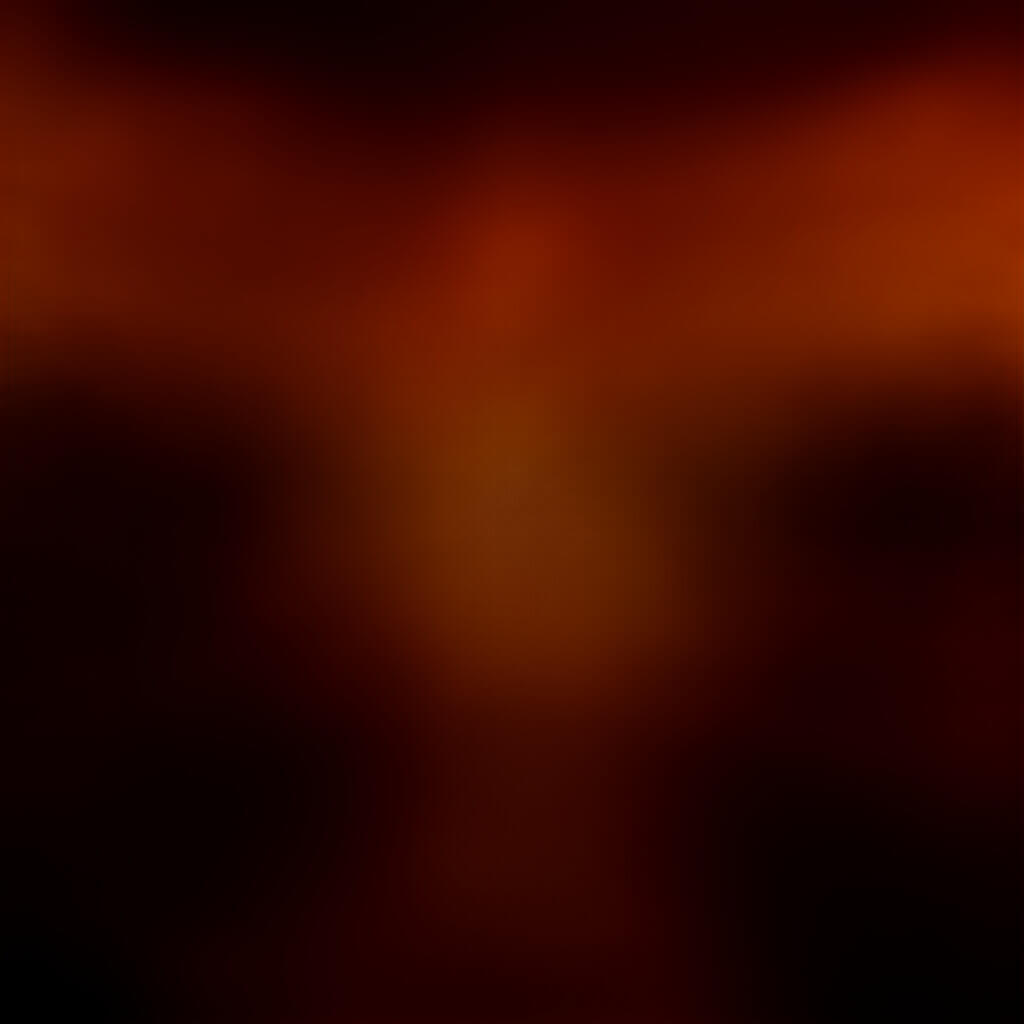} &
\includegraphics[width=0.115\textwidth]{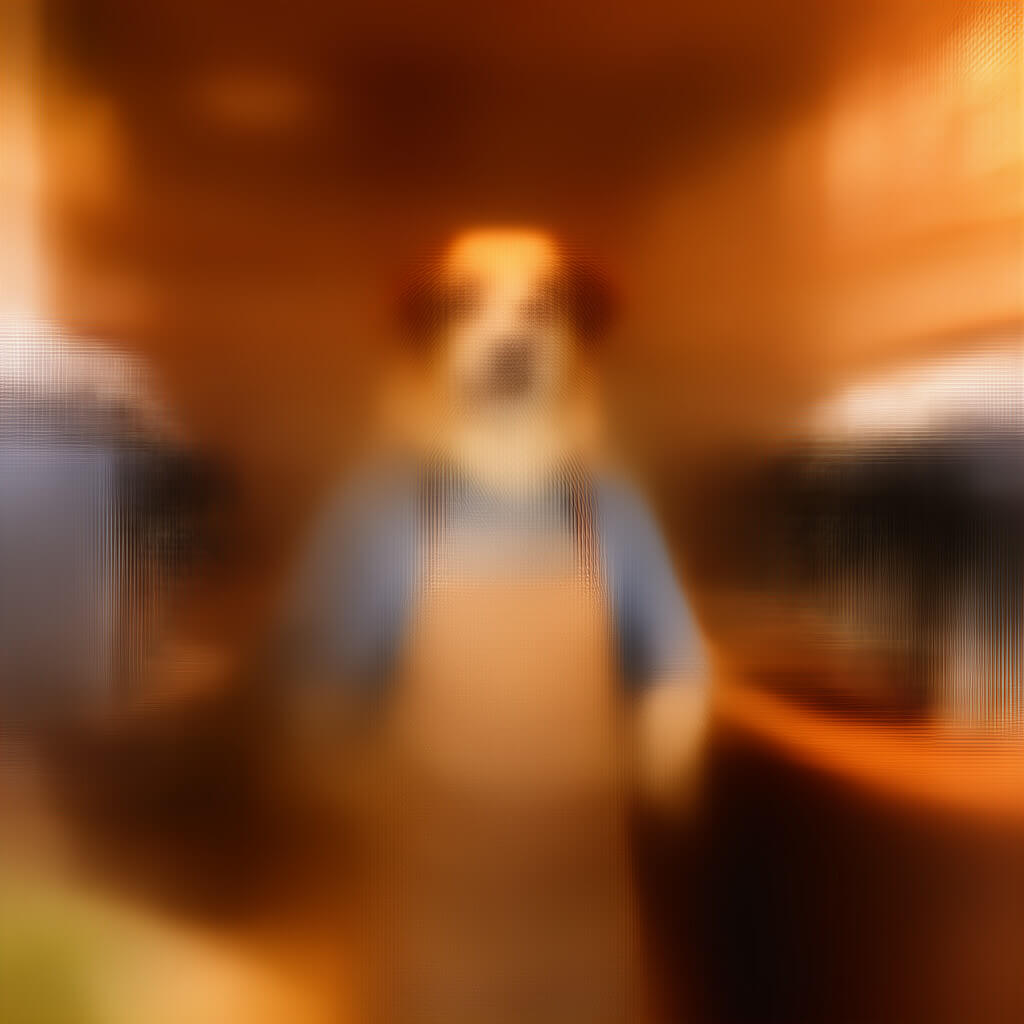} &
\includegraphics[width=0.115\textwidth]{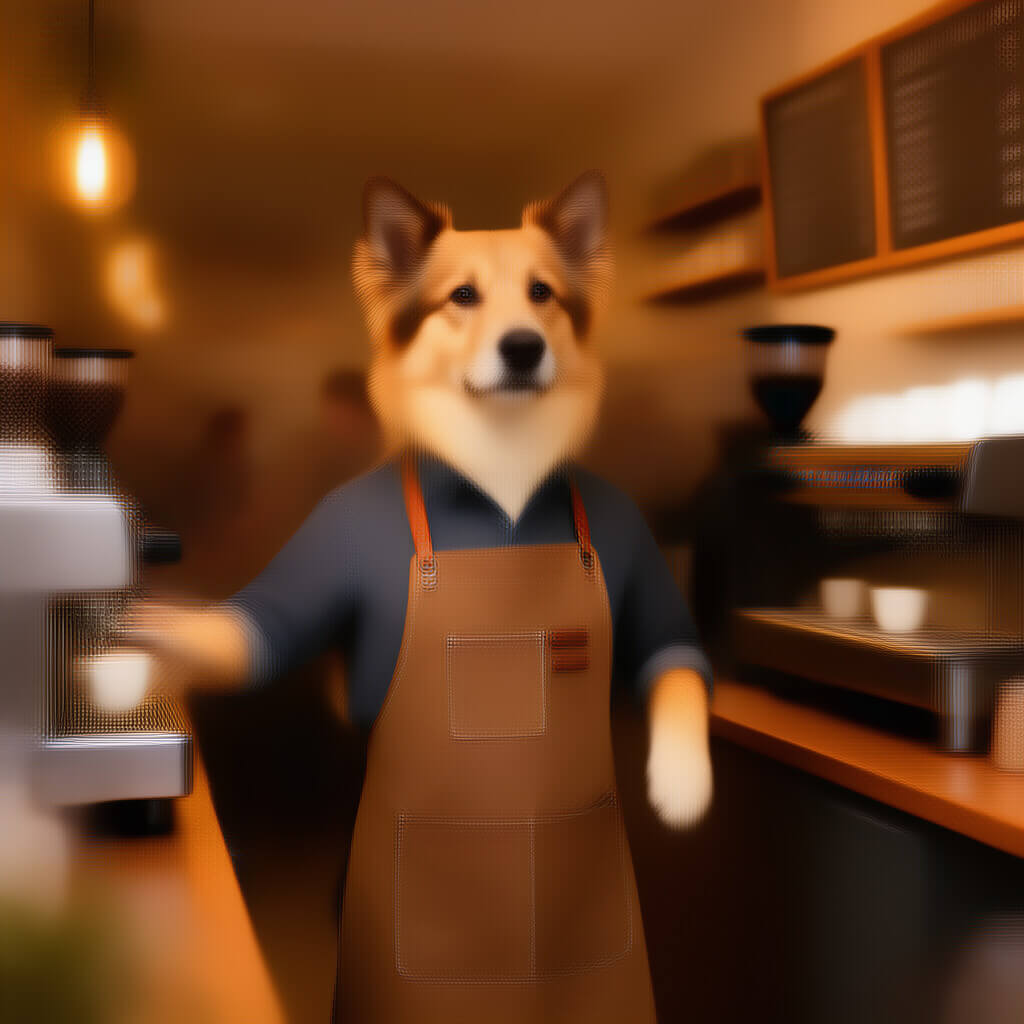} &
\includegraphics[width=0.115\textwidth]{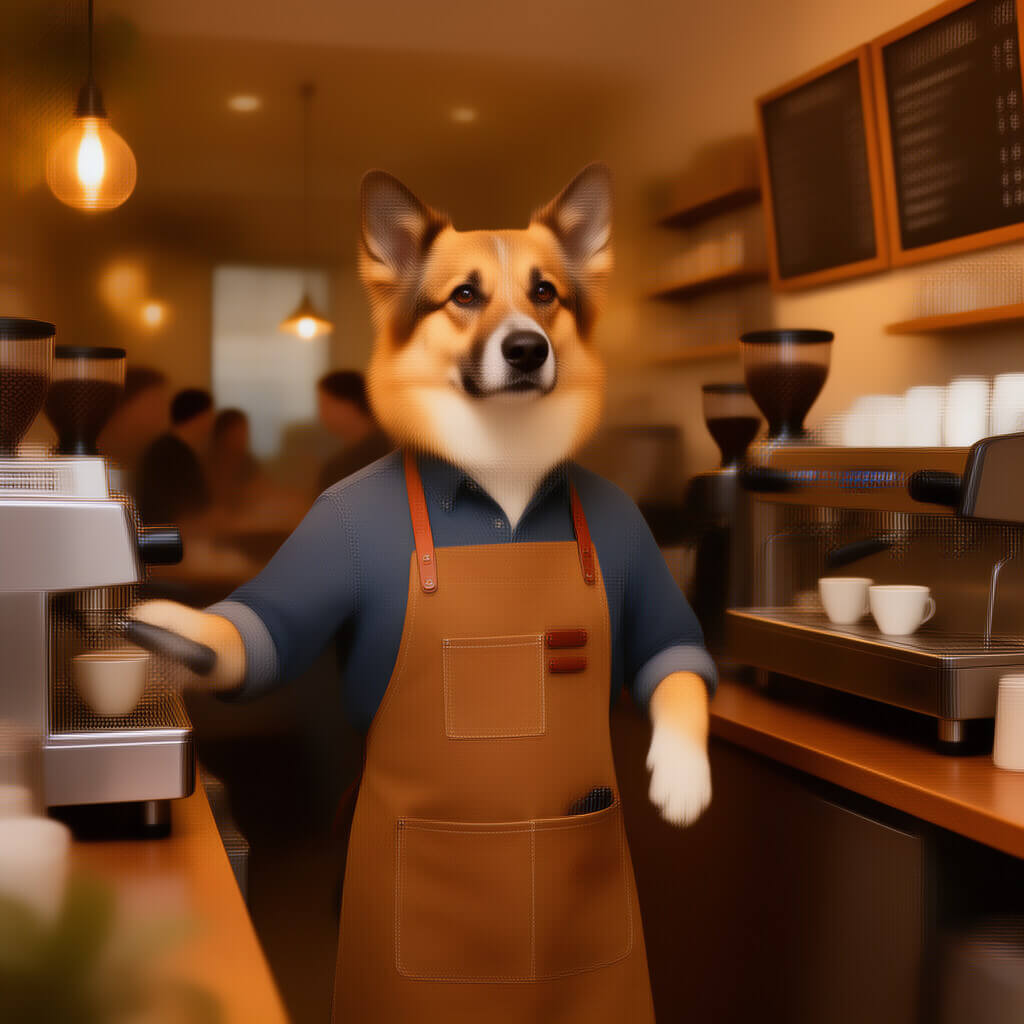} &
\includegraphics[width=0.115\textwidth]{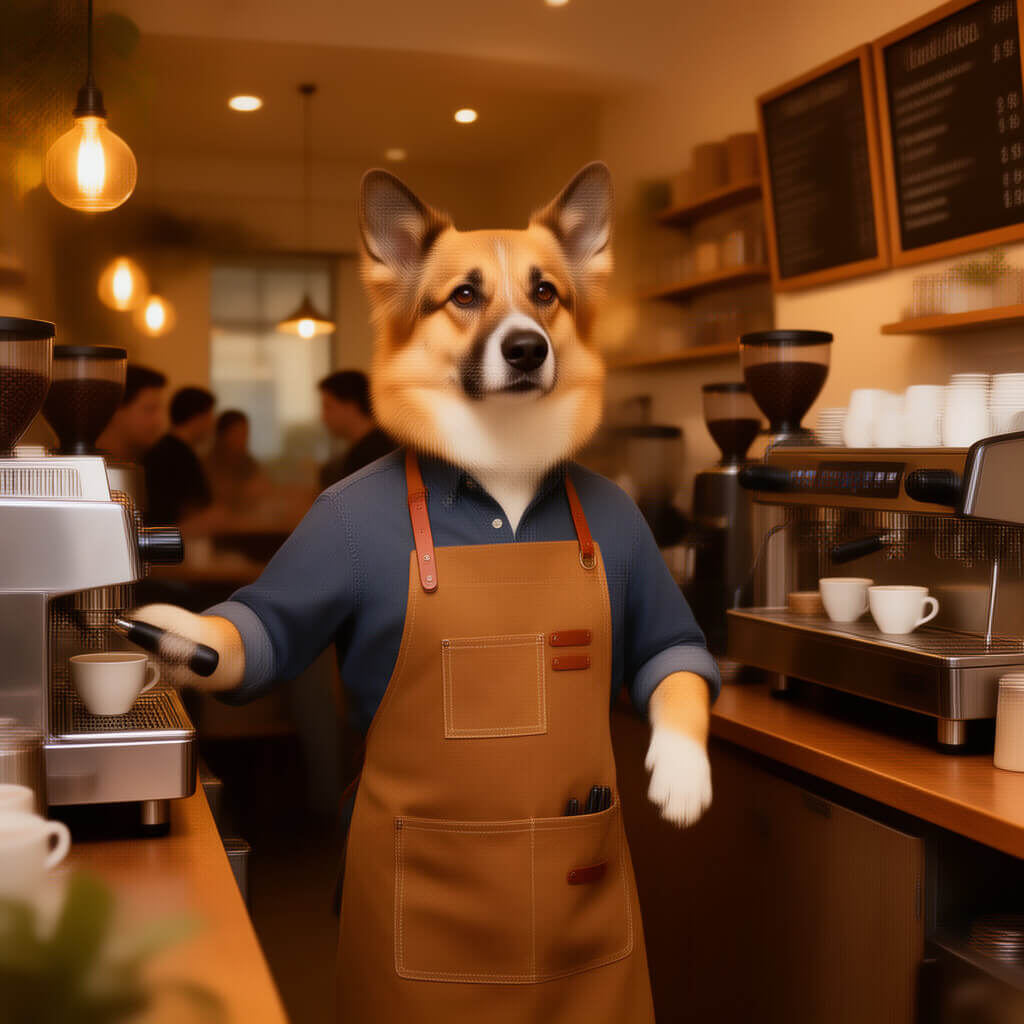} &
\includegraphics[width=0.115\textwidth]{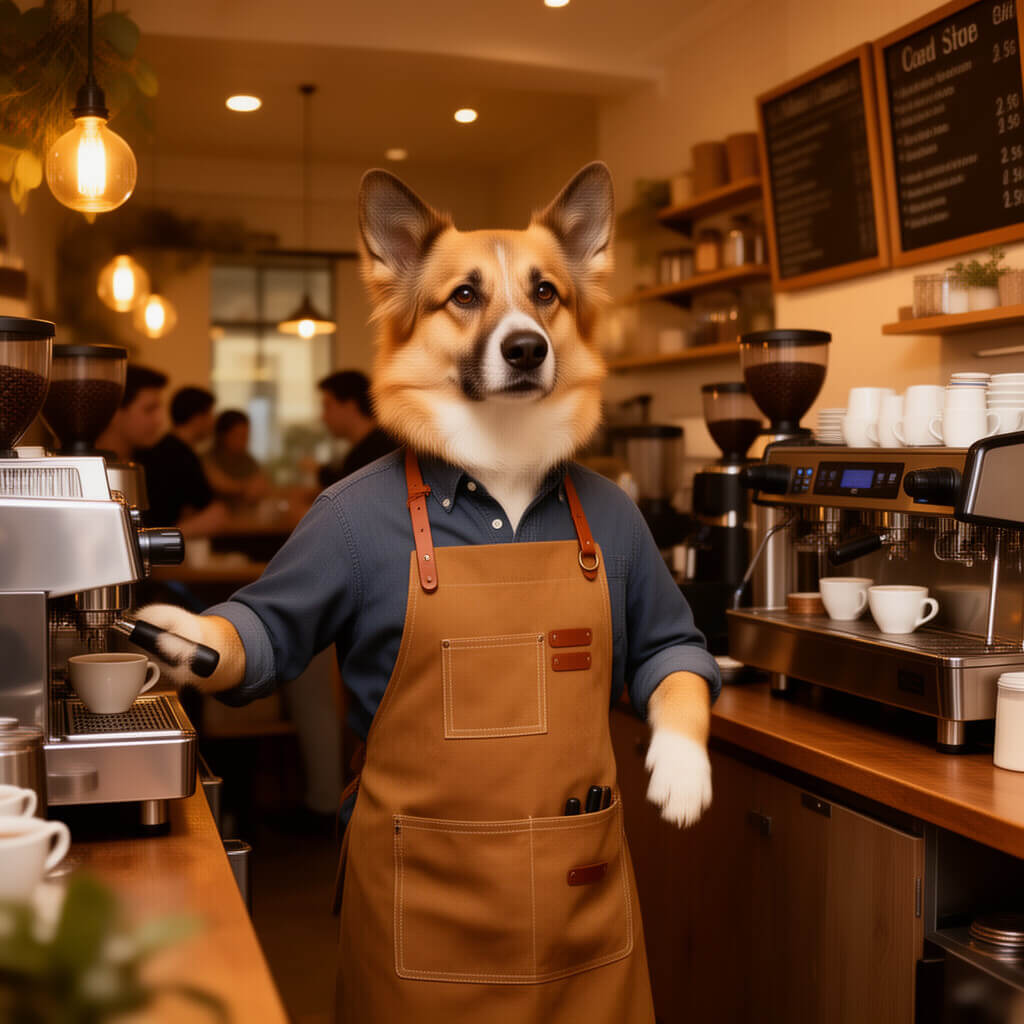} &
\includegraphics[width=0.115\textwidth]{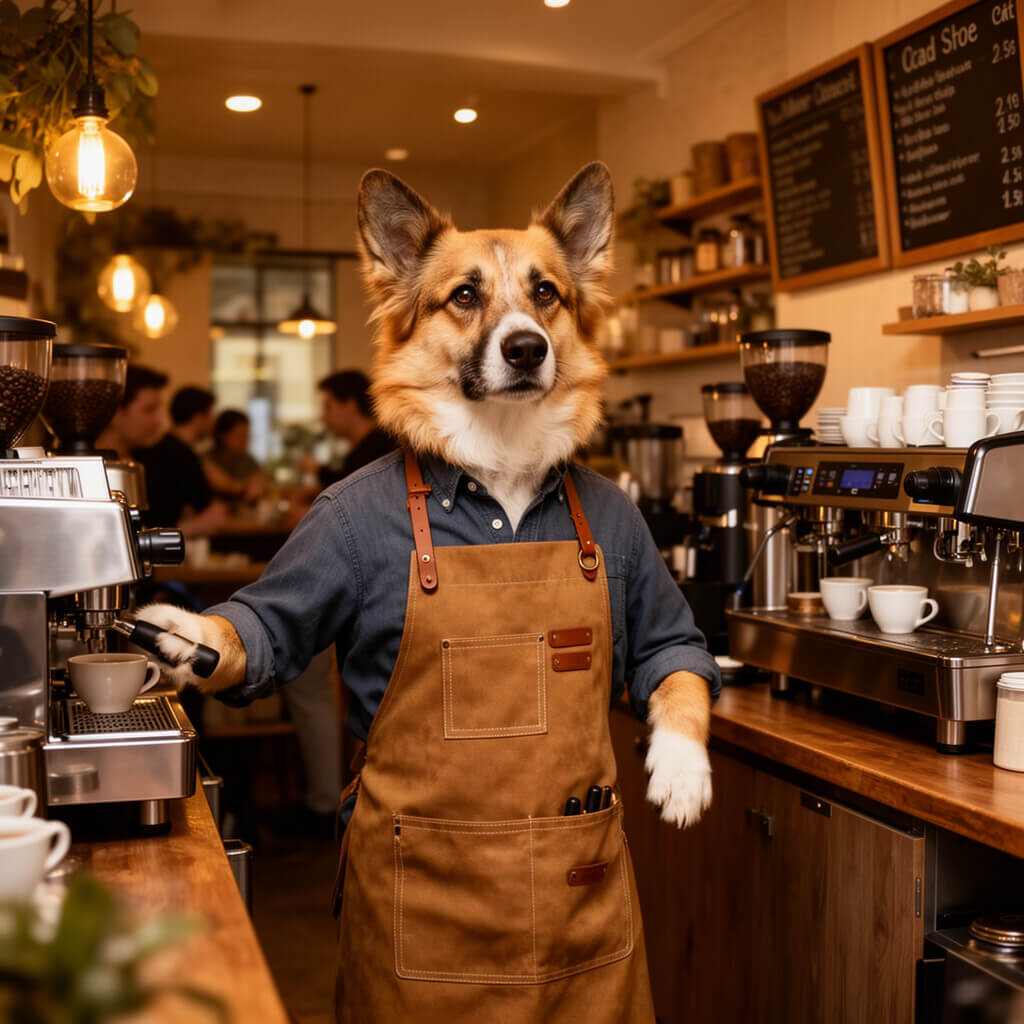} &
\includegraphics[width=0.115\textwidth]{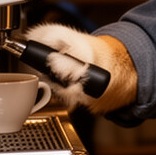} \\

\rotatebox{90}{{\quad \hspace{1mm}Artifact}} \rotatebox{90}{{\hspace{5mm}  Mask}}&
\includegraphics[width=0.115\textwidth]{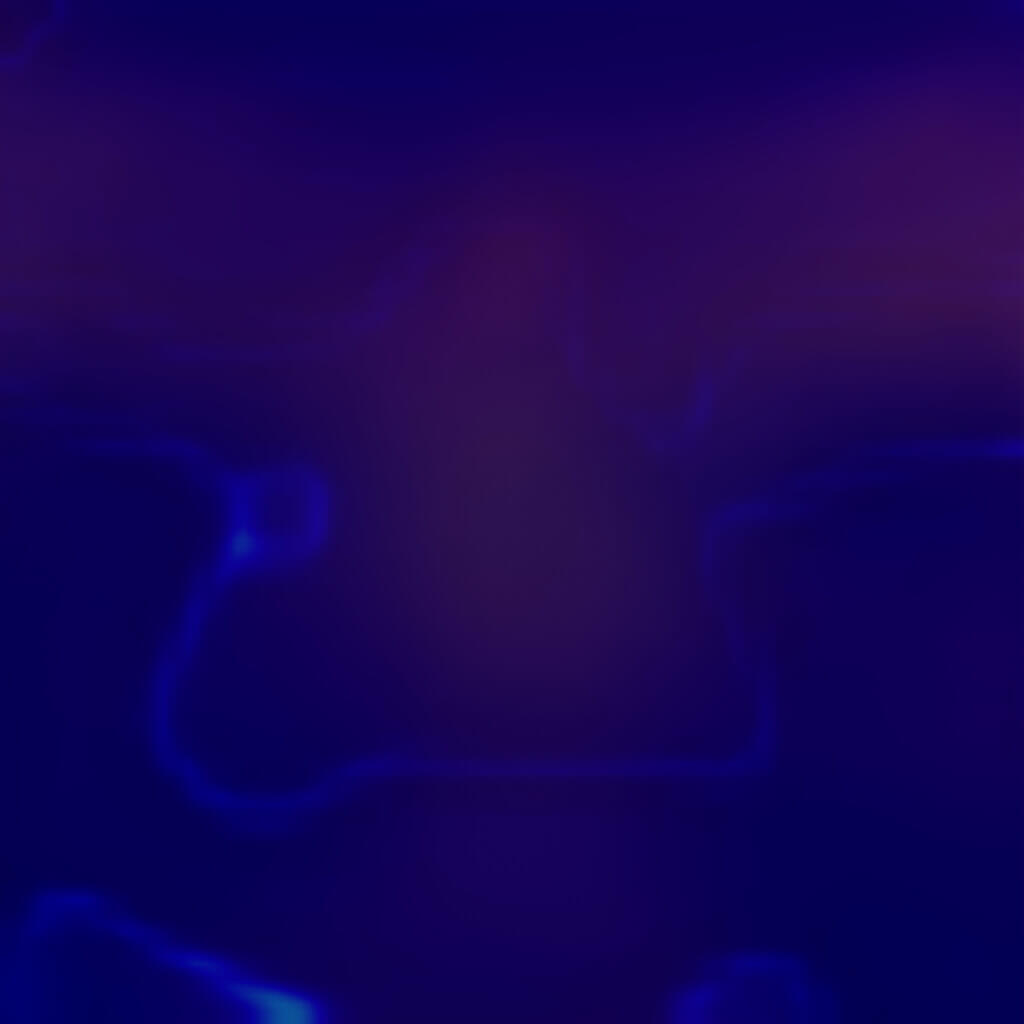} &
\includegraphics[width=0.115\textwidth]{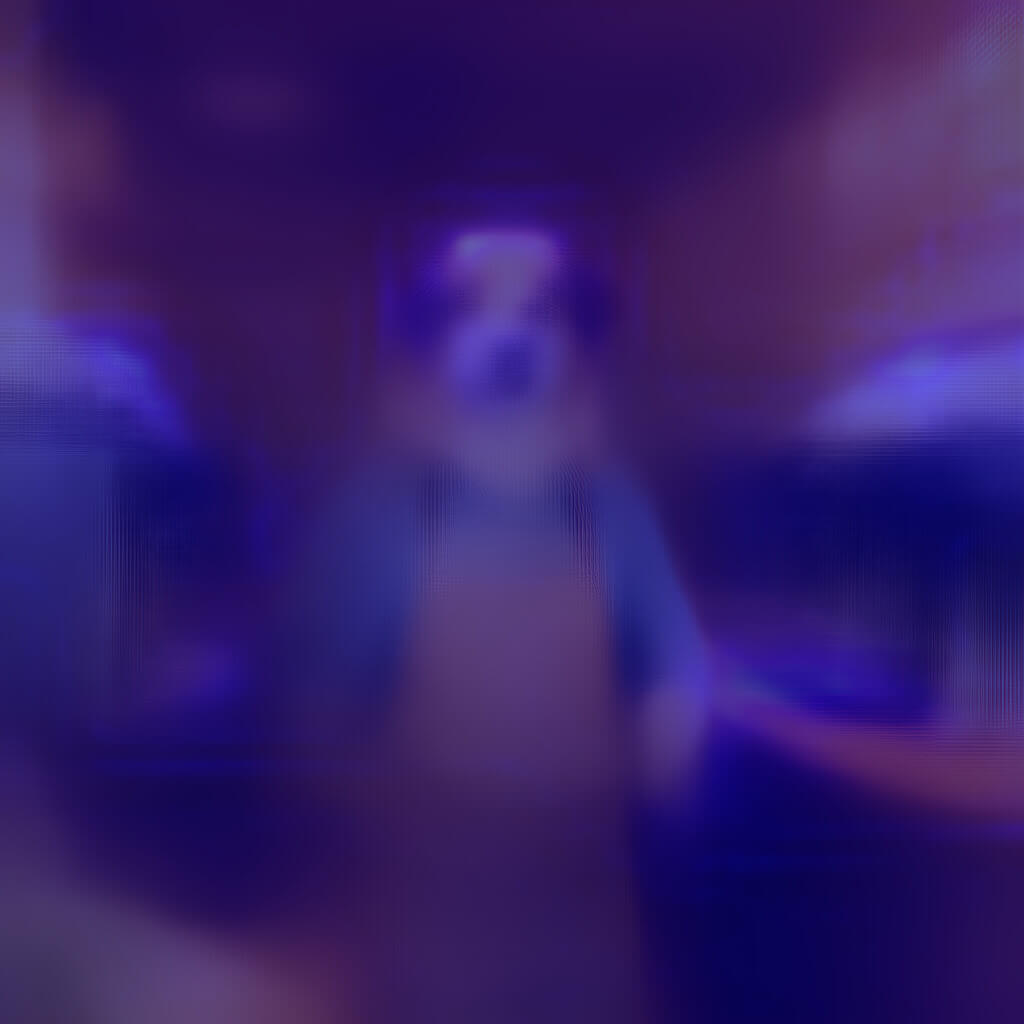} &
\includegraphics[width=0.115\textwidth]{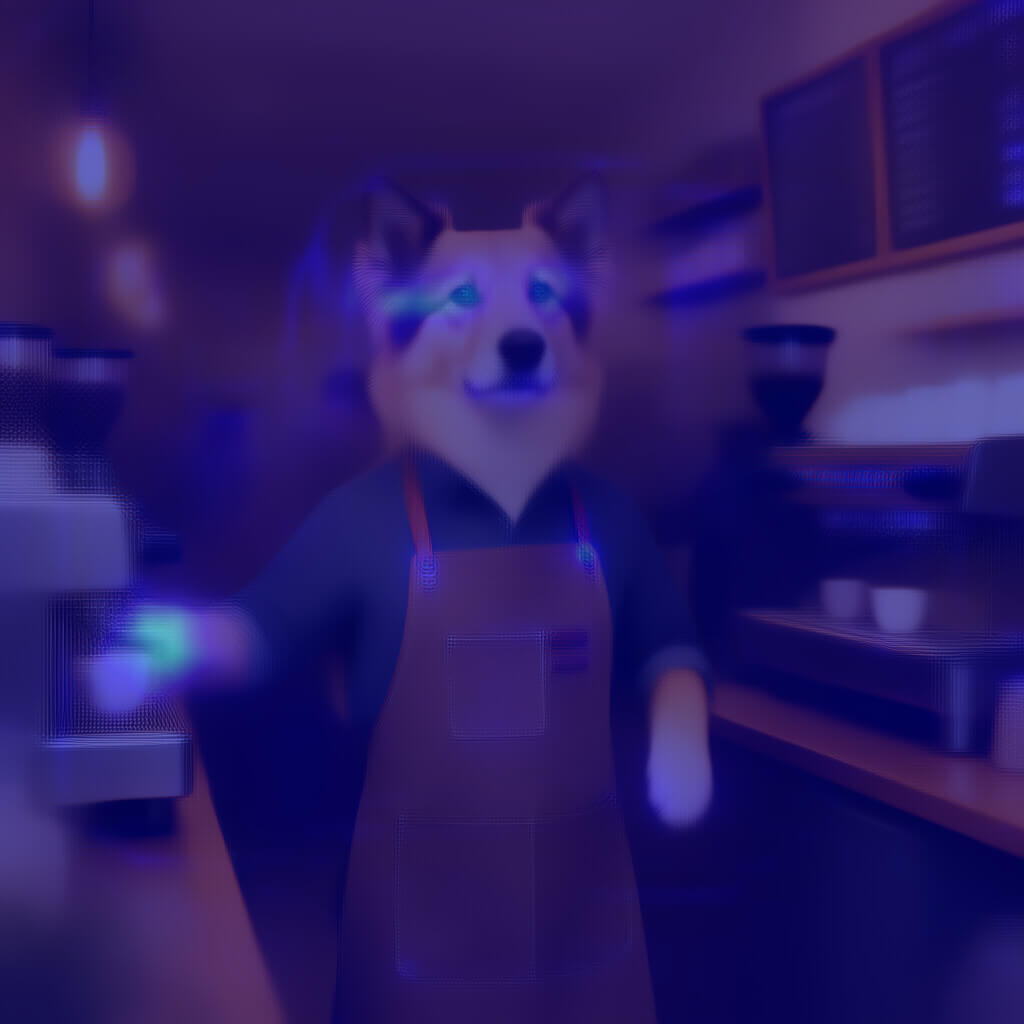} &
\includegraphics[width=0.115\textwidth]{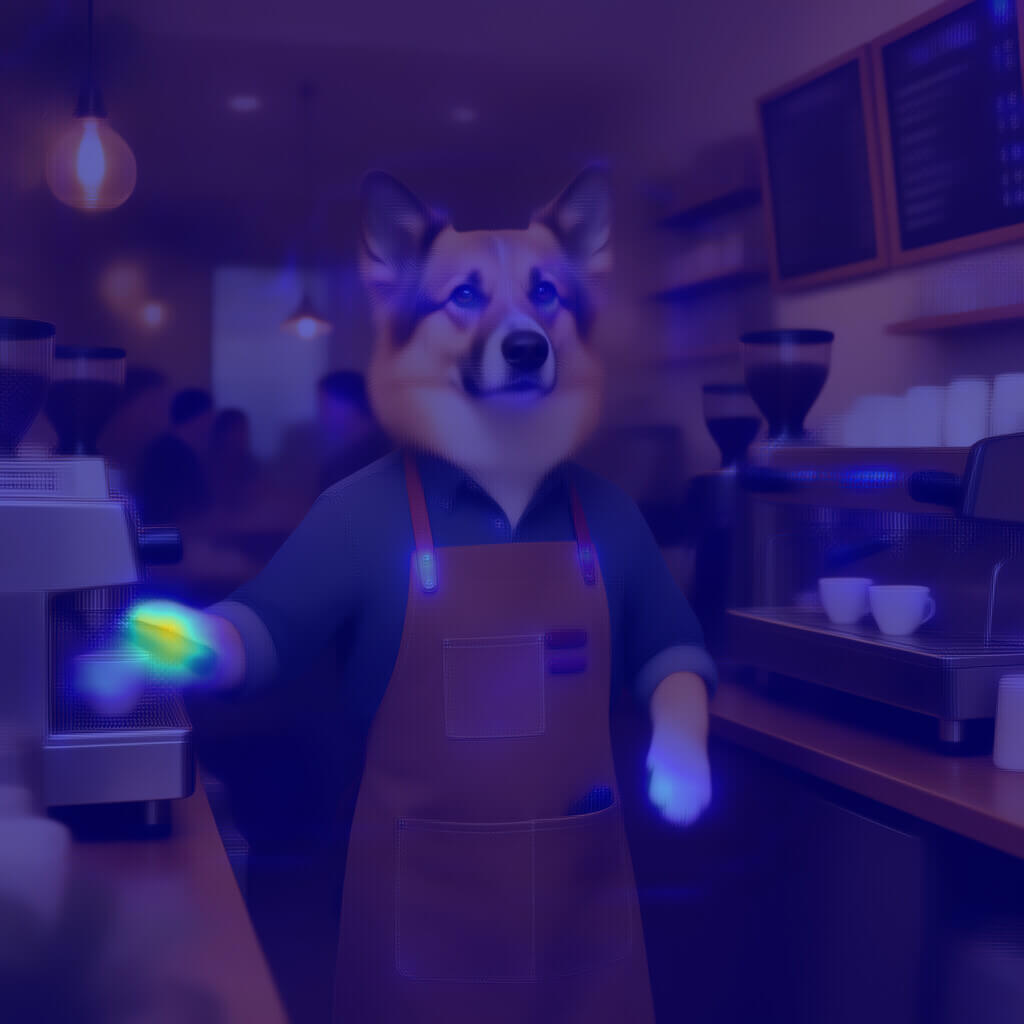} &
\includegraphics[width=0.115\textwidth]{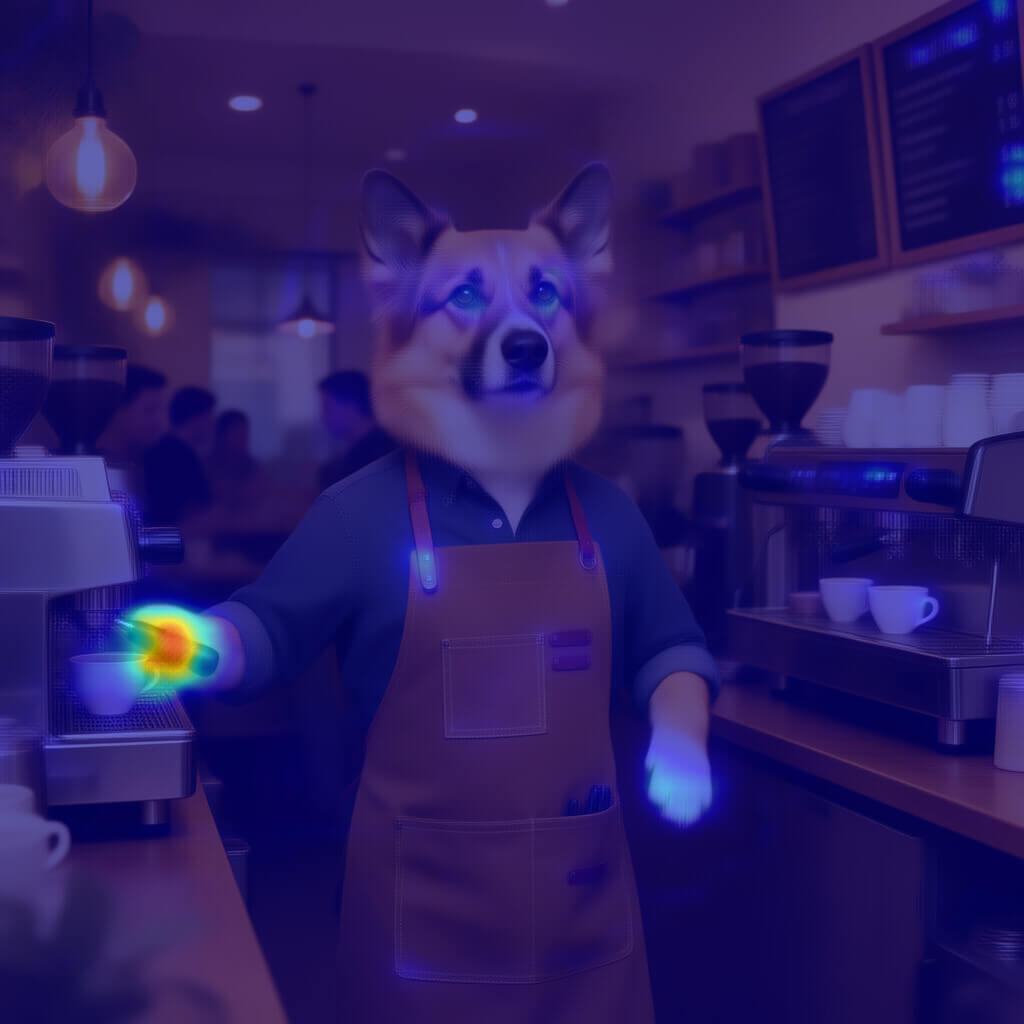} &
\includegraphics[width=0.115\textwidth]{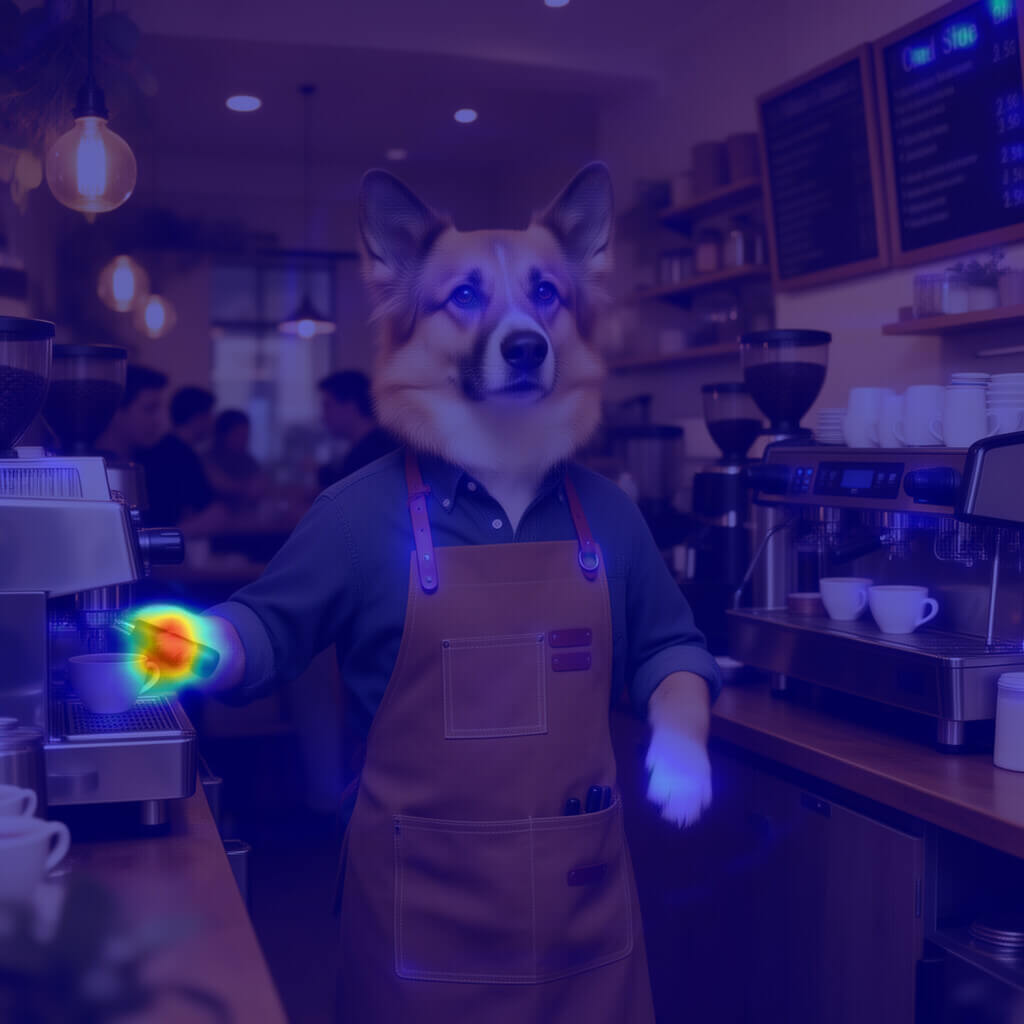} &
\includegraphics[width=0.115\textwidth]{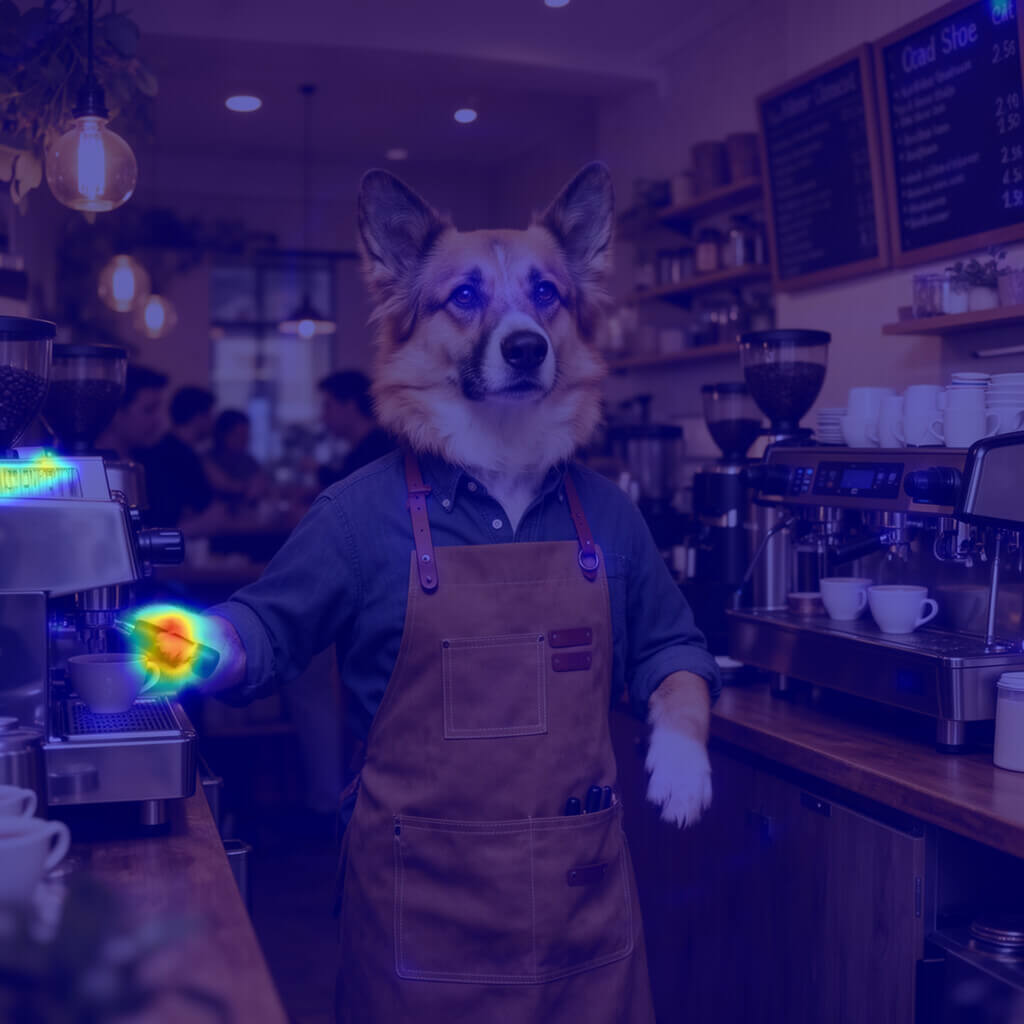} &
\includegraphics[width=0.115\textwidth]{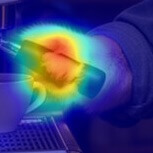} \\

\rotatebox{90}{+\our{}} &
\includegraphics[width=0.115\textwidth]{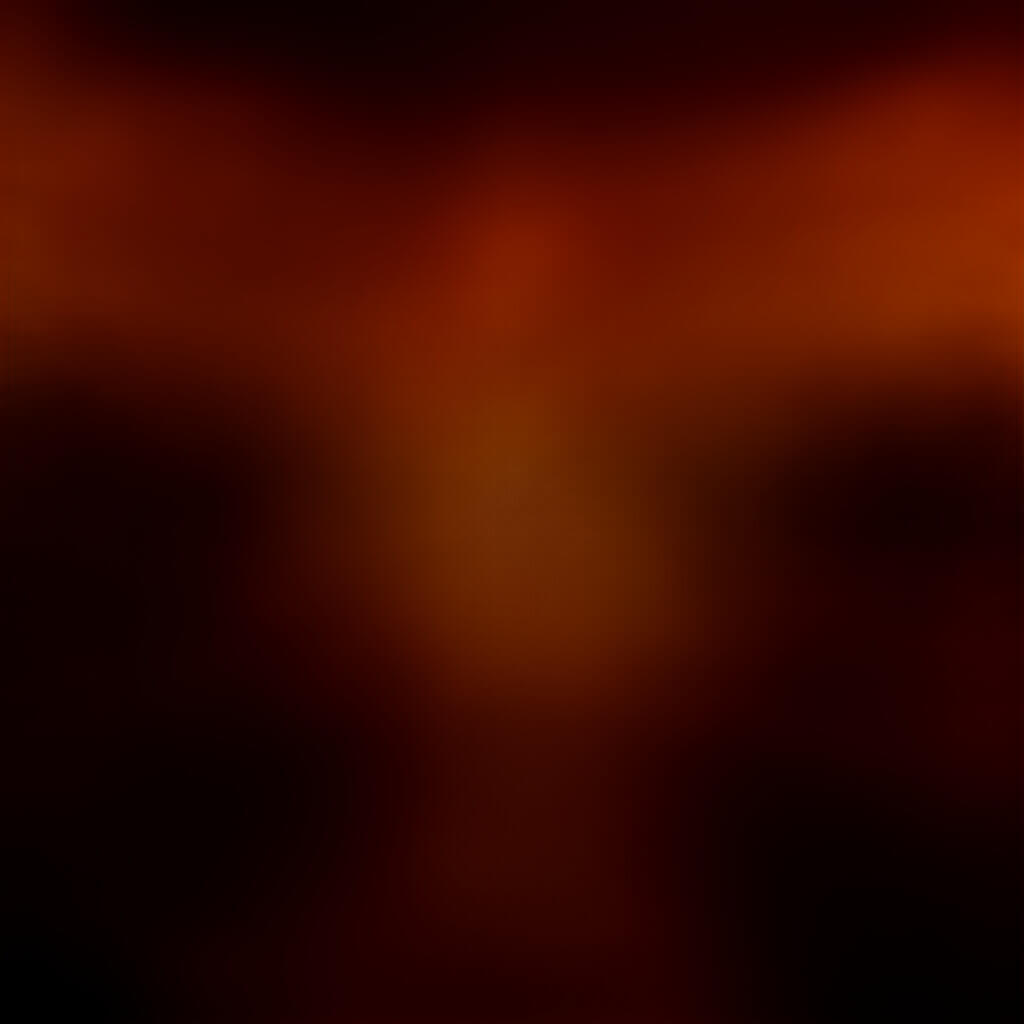} &
\includegraphics[width=0.115\textwidth]{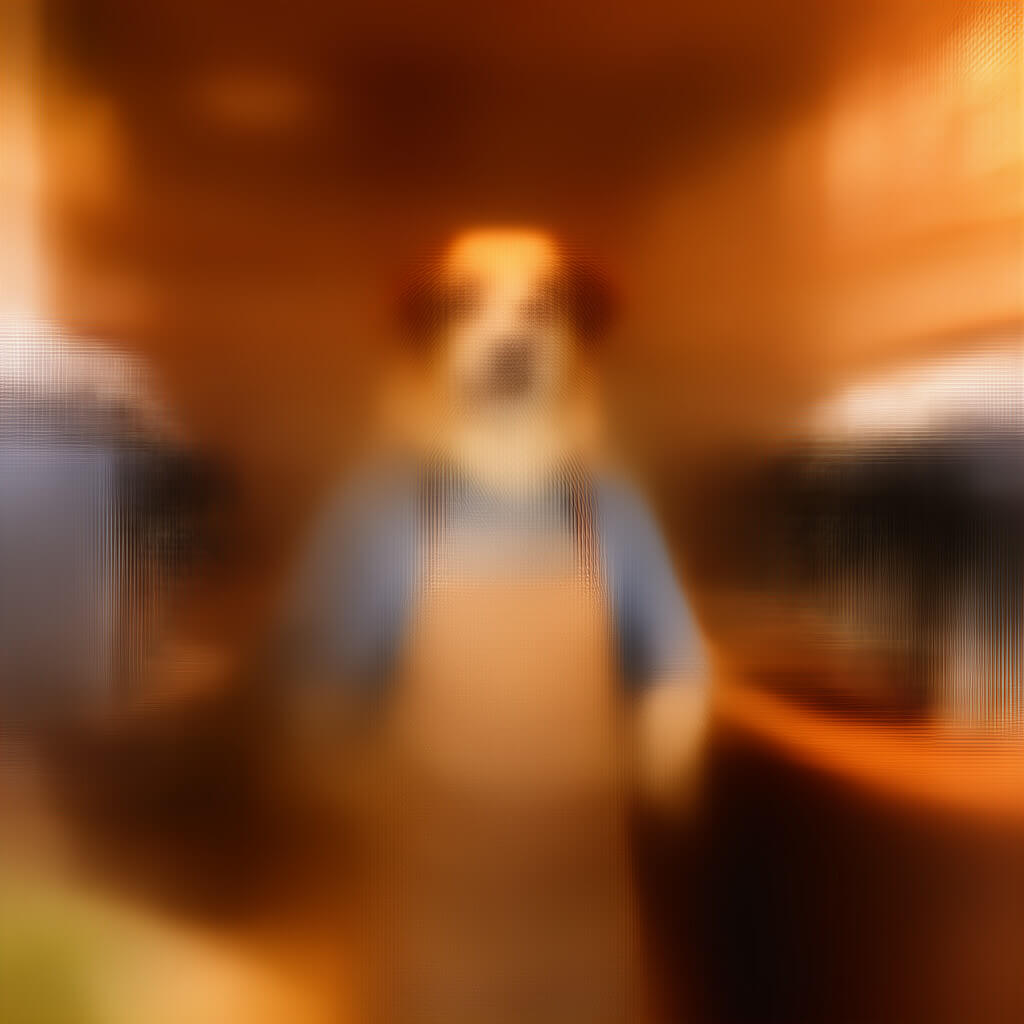} &
\includegraphics[width=0.115\textwidth]{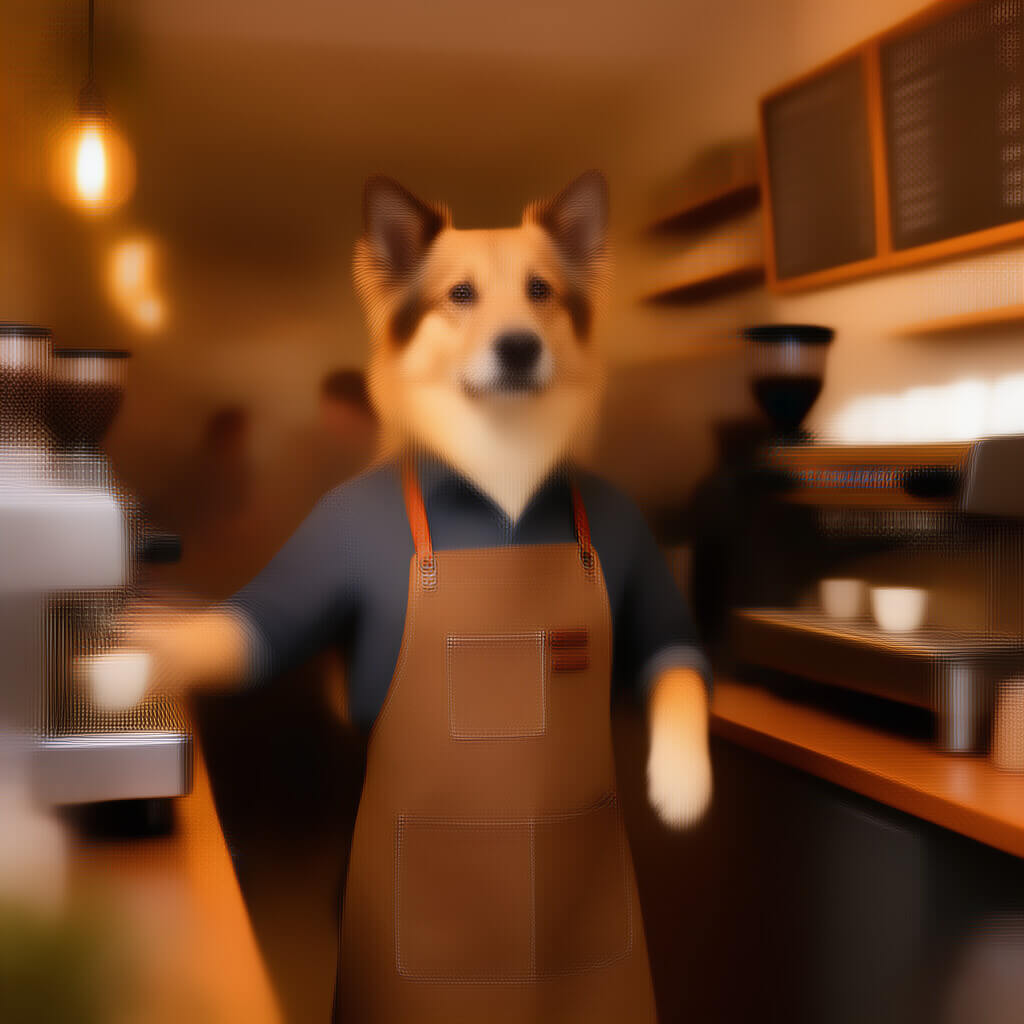} &
\includegraphics[width=0.115\textwidth]{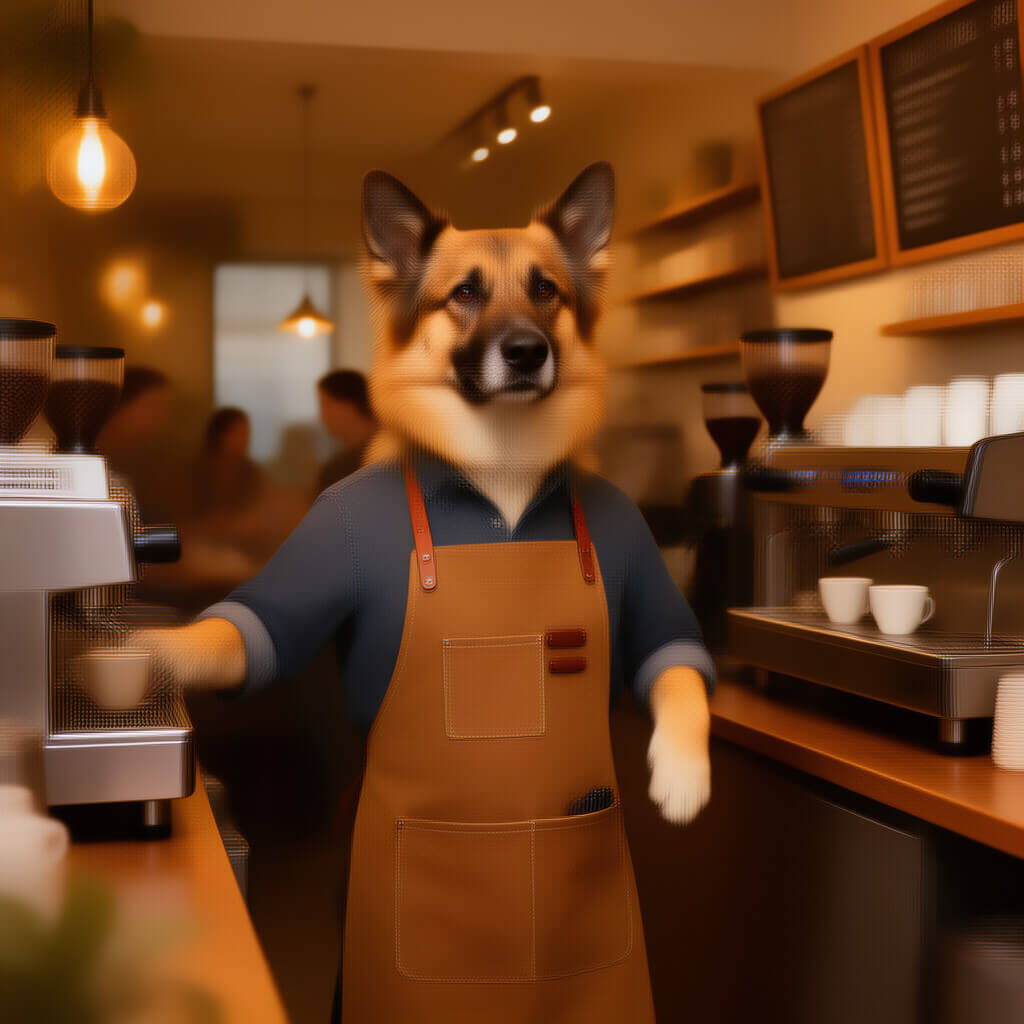} &
\includegraphics[width=0.115\textwidth]{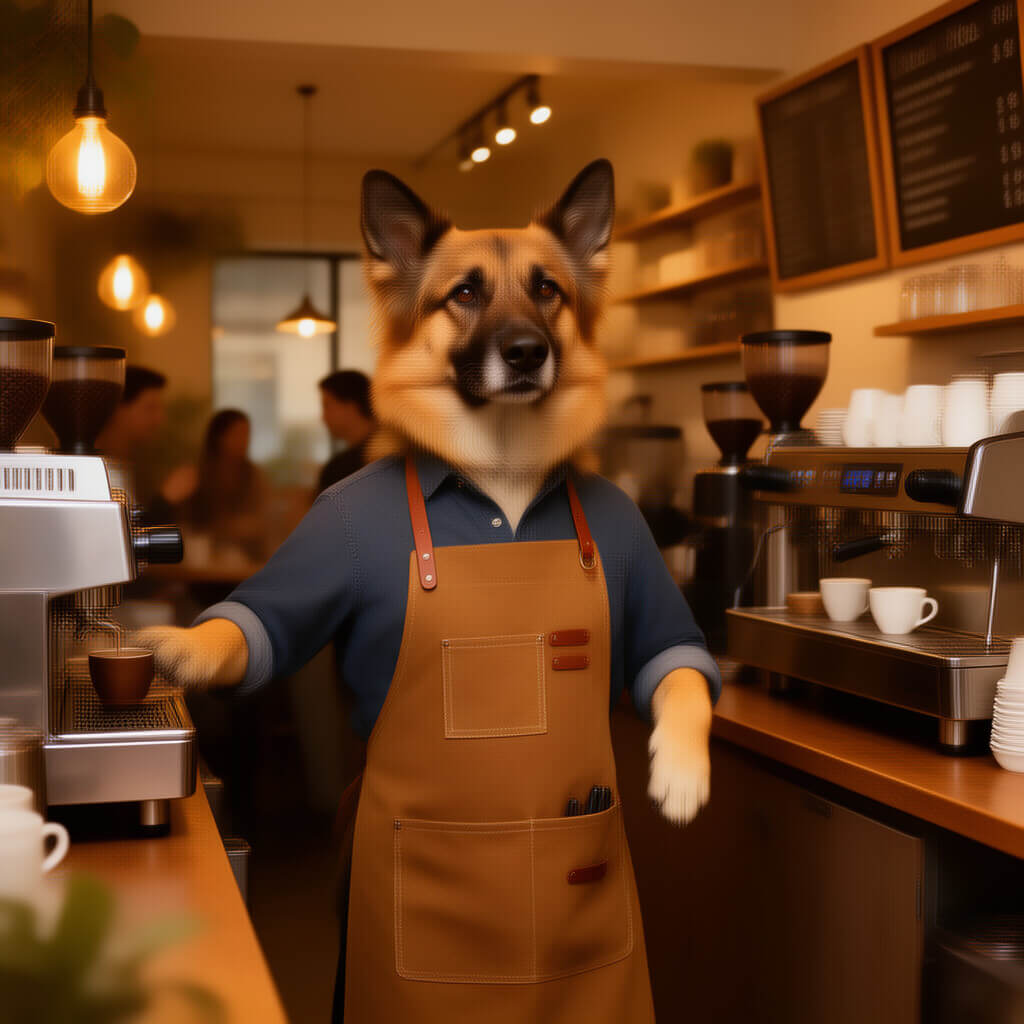} &
\includegraphics[width=0.115\textwidth]{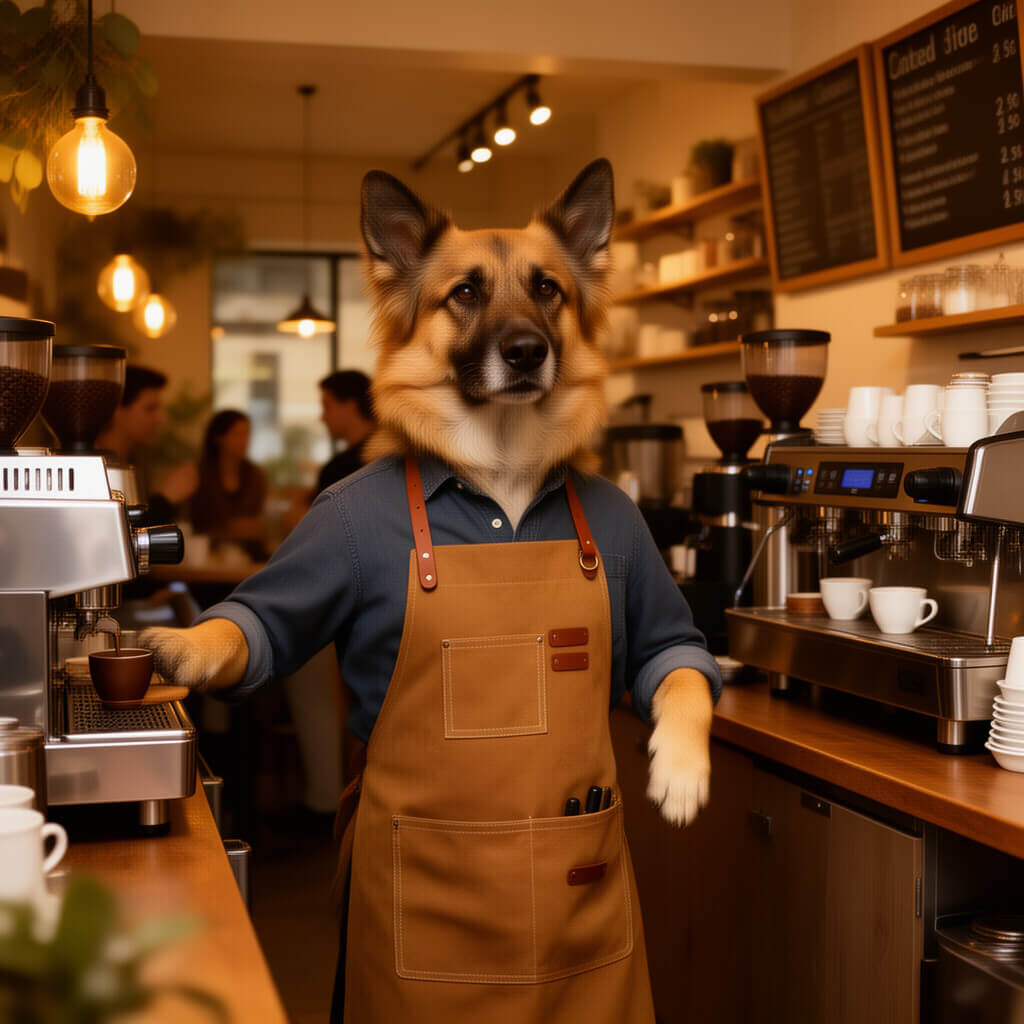} &
\includegraphics[width=0.115\textwidth]{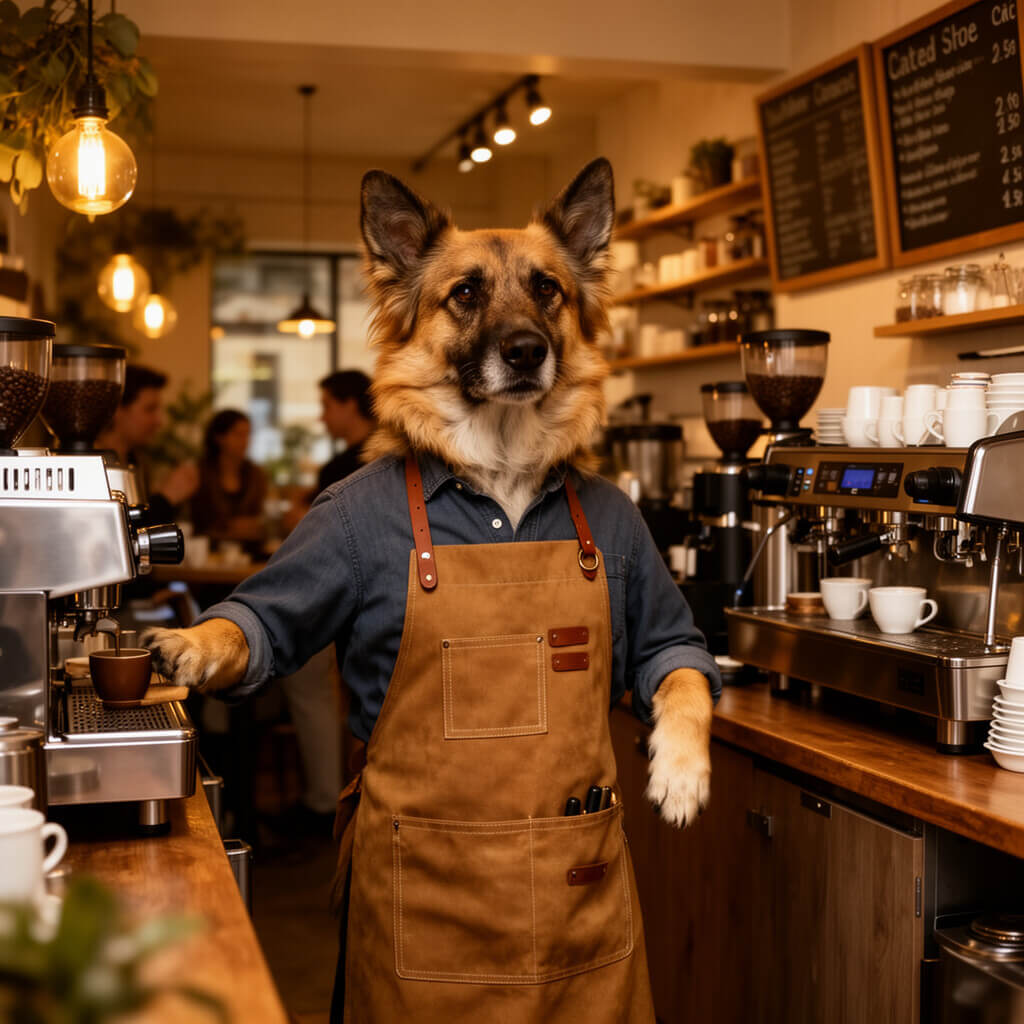} &
\includegraphics[width=0.115\textwidth]{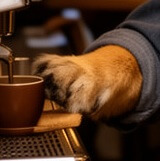} \\

\rotatebox{90}{{\quad \hspace{1mm}Artifact}} \rotatebox{90}{{\hspace{5mm}  Mask}} &
\includegraphics[width=0.115\textwidth]{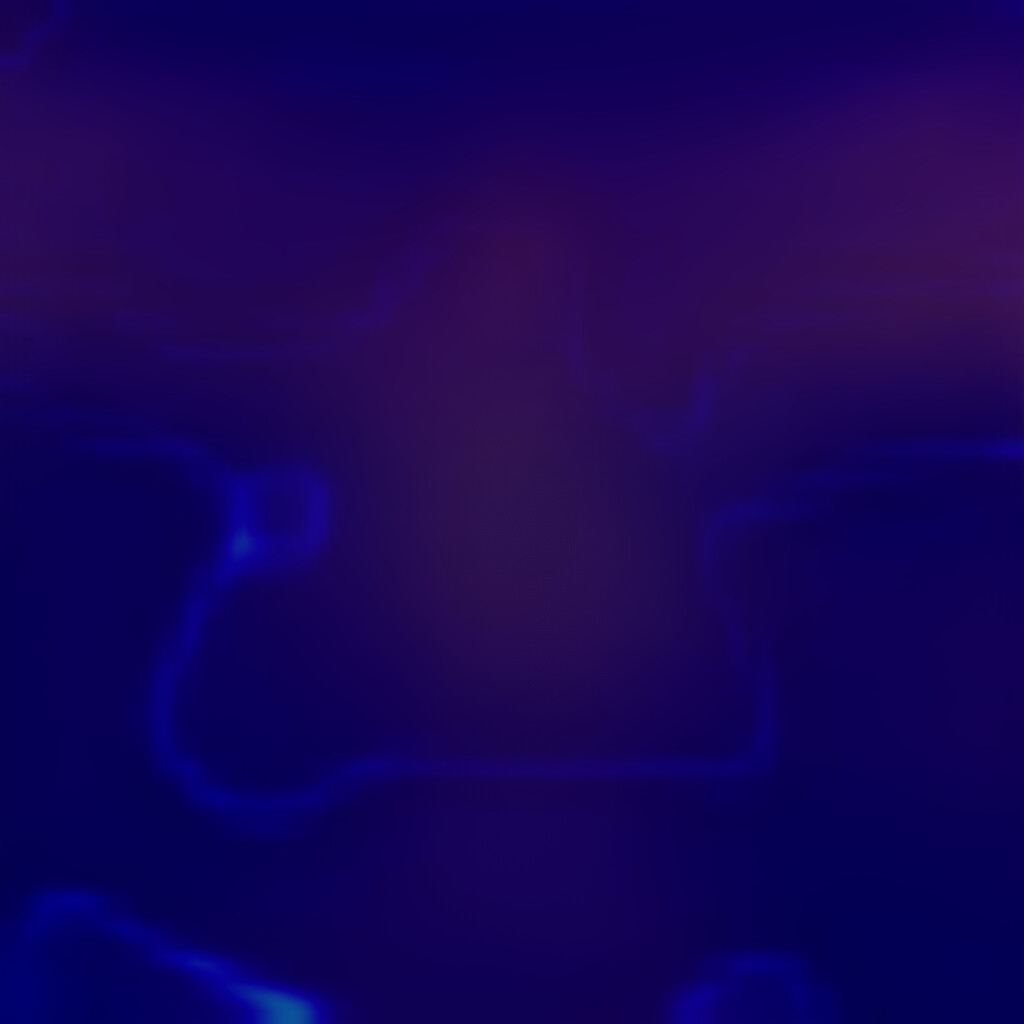} &
\includegraphics[width=0.115\textwidth]{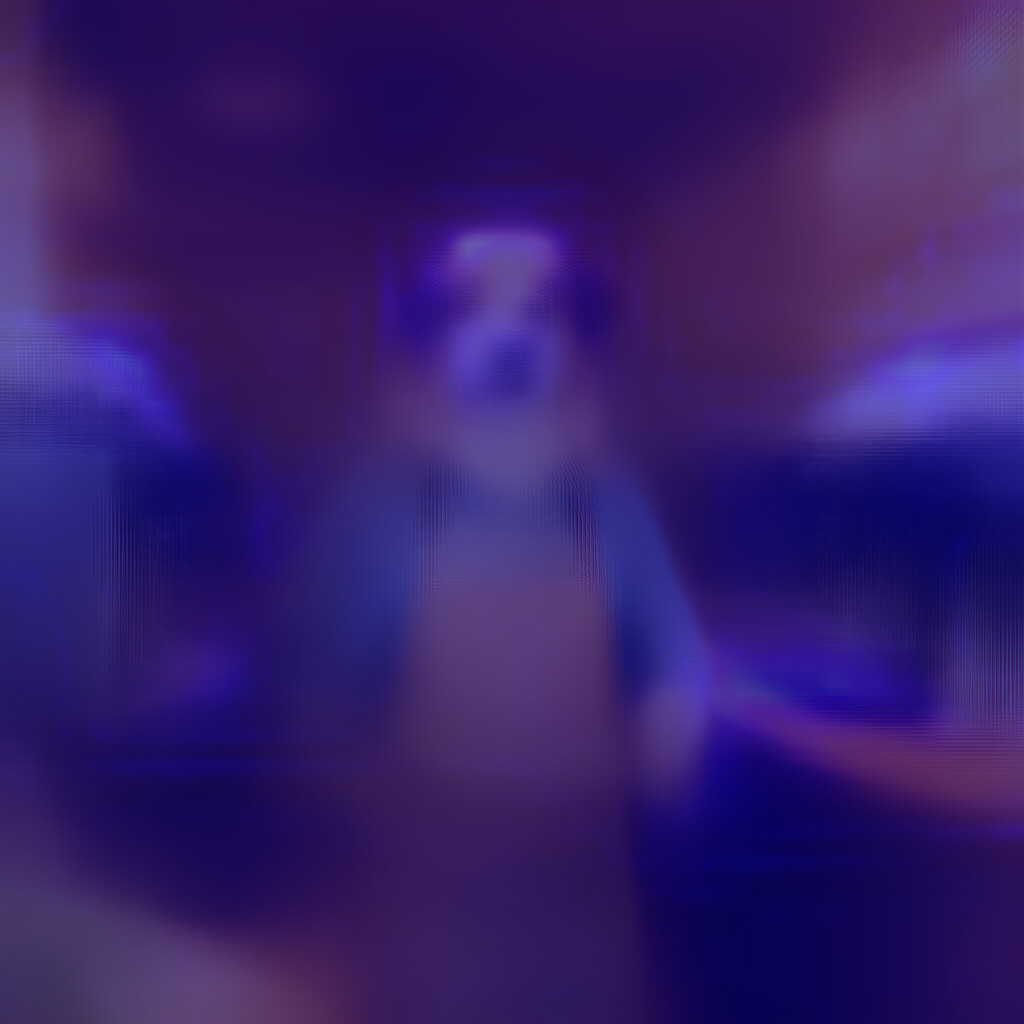} &
\includegraphics[width=0.115\textwidth]{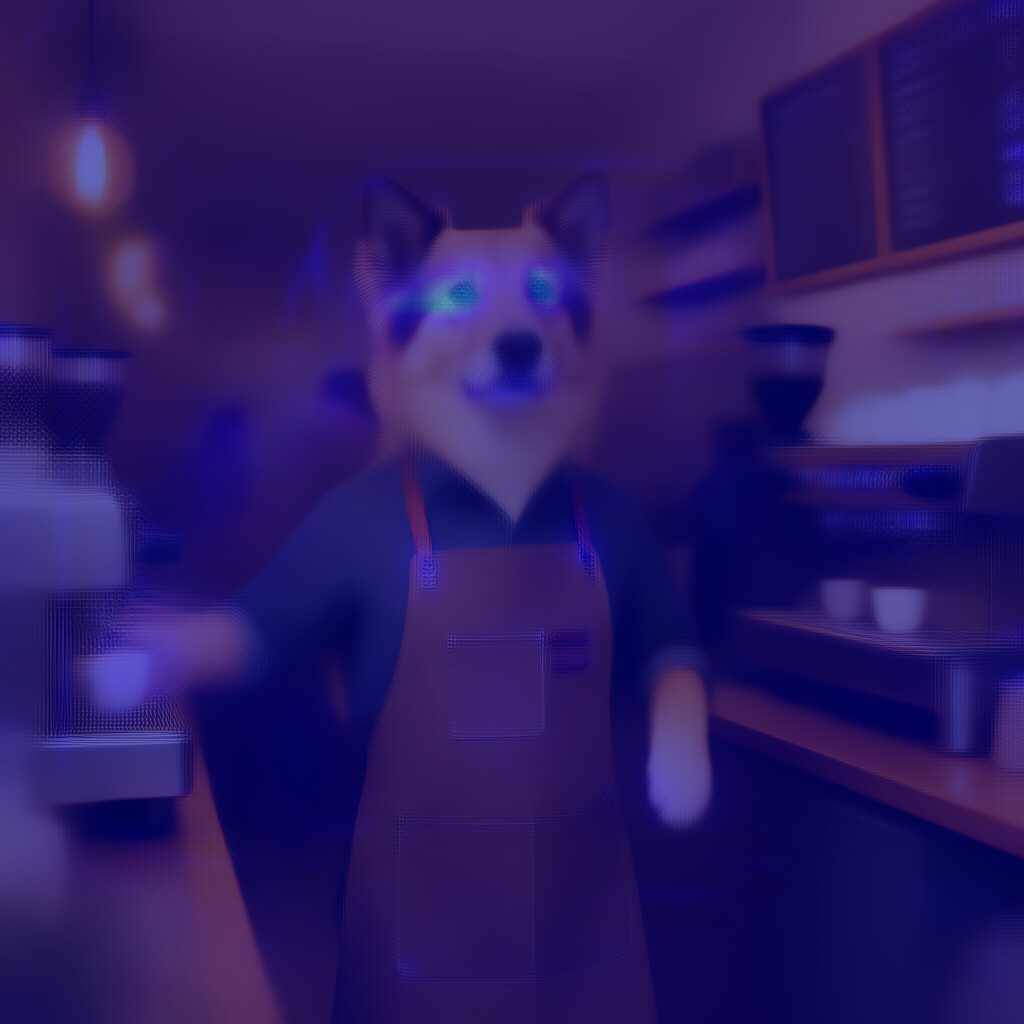} &
\includegraphics[width=0.115\textwidth]{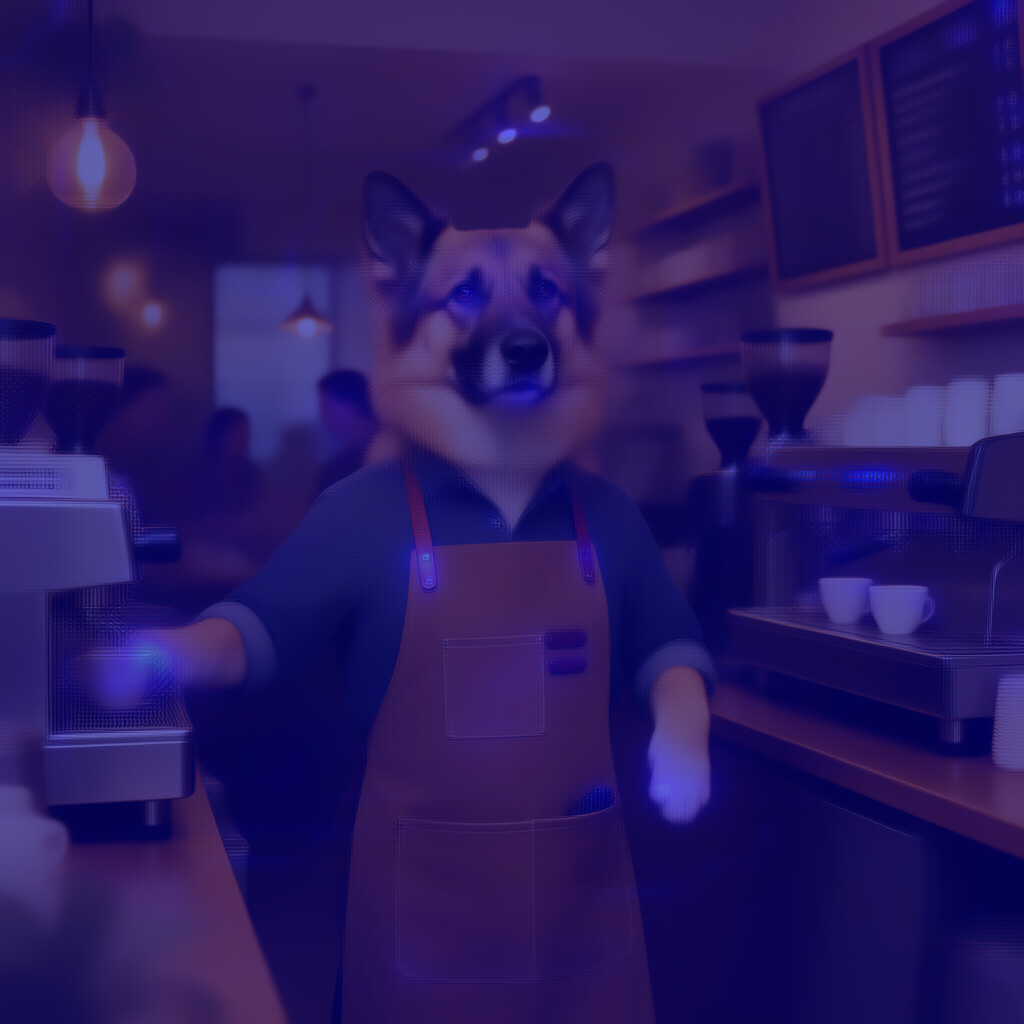} &
\includegraphics[width=0.115\textwidth]{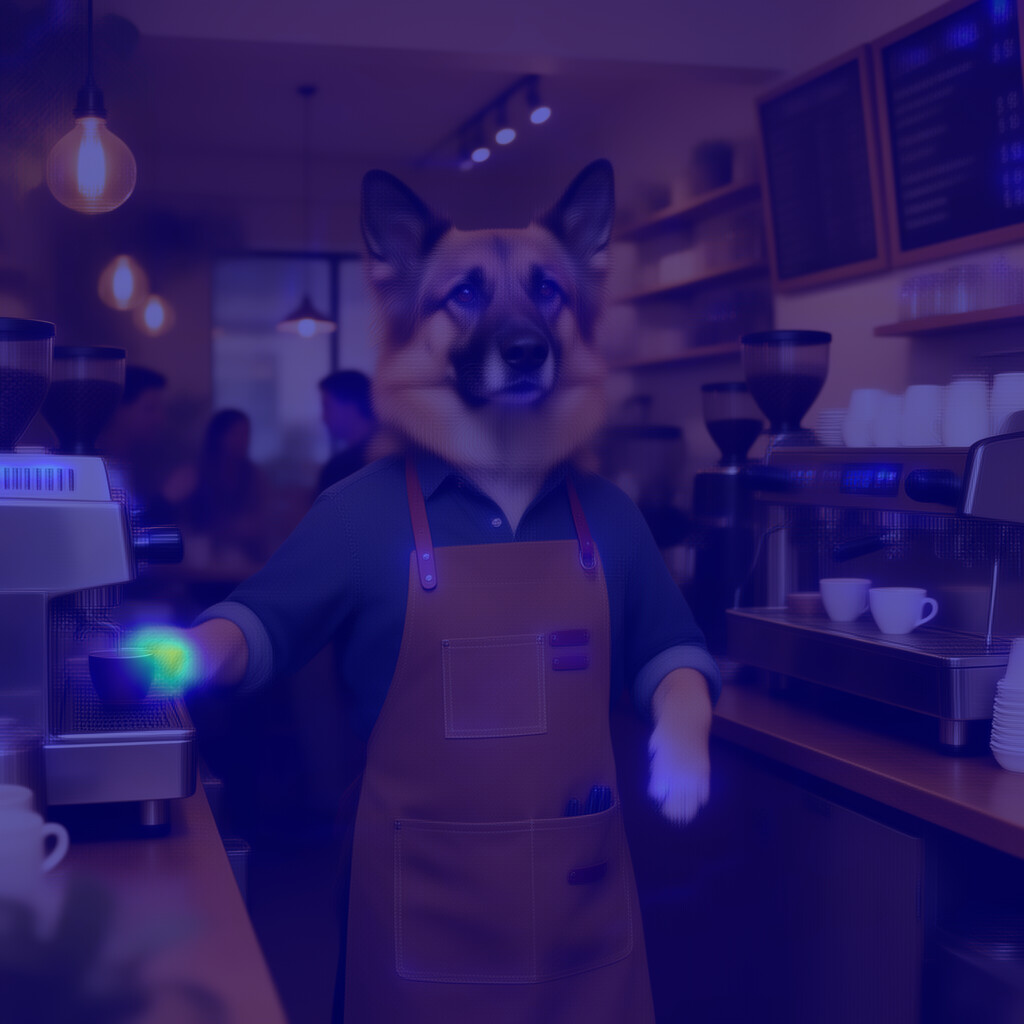} &
\includegraphics[width=0.115\textwidth]{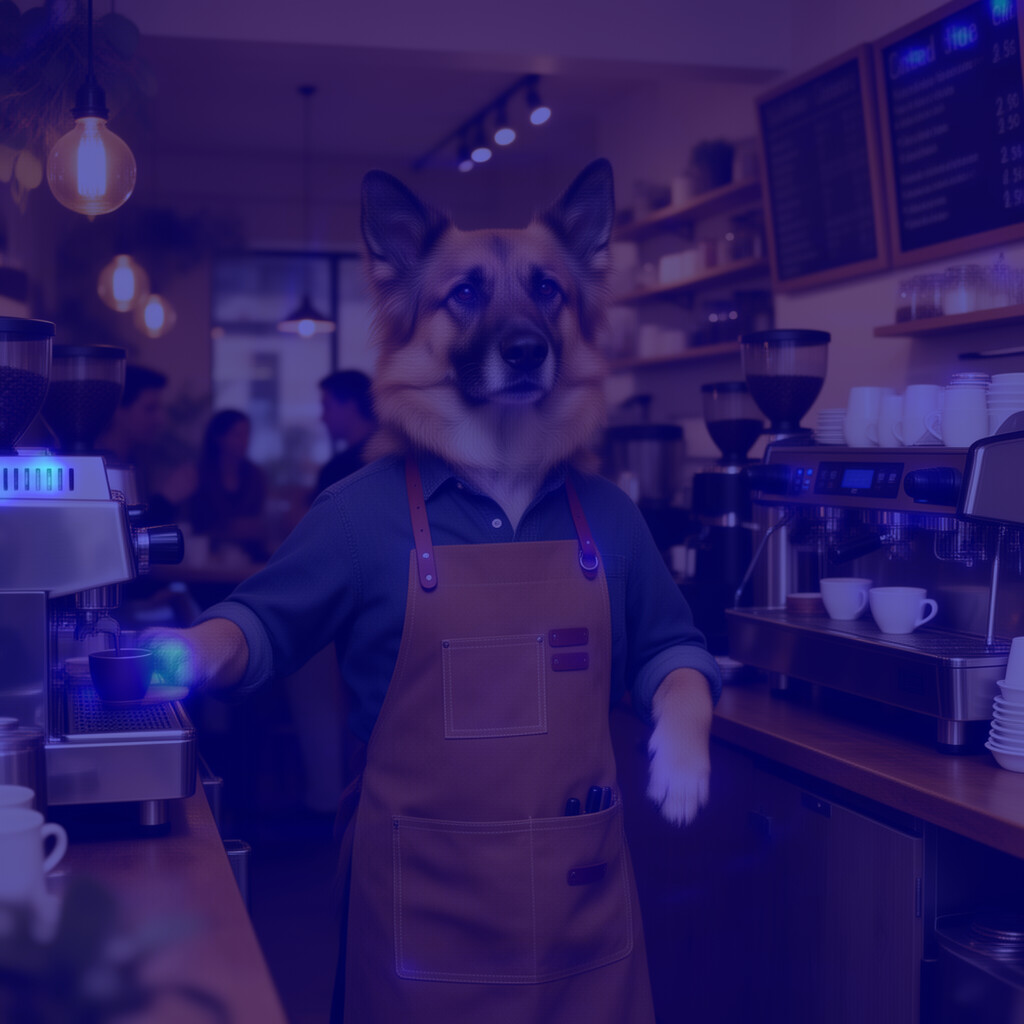} &
\includegraphics[width=0.115\textwidth]{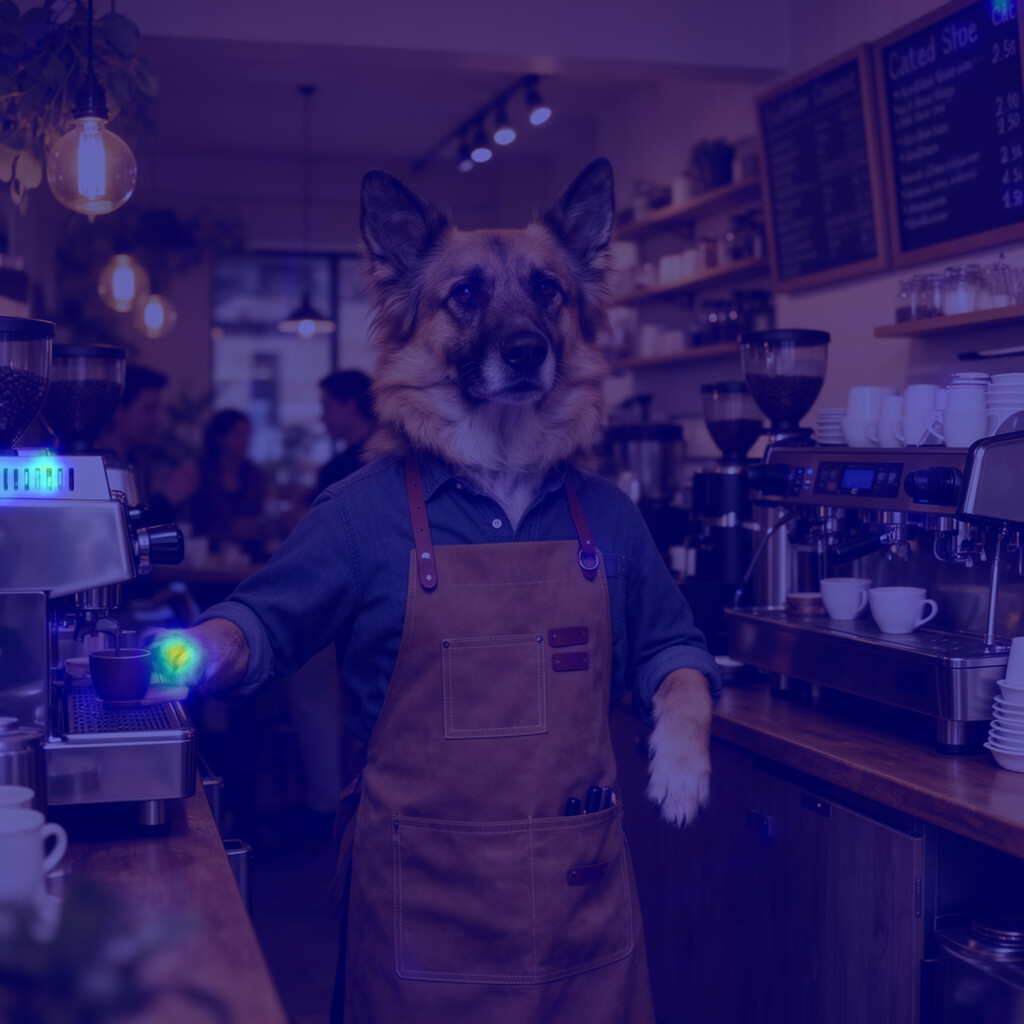} &
\includegraphics[width=0.115\textwidth]{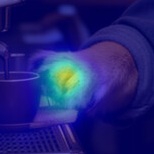} \\

\end{tabular}
\caption{\textbf{Baseline (top) and proposed (bottom) generation trajectories illustrating artifact reduction on FLUX.2 [dev].} When generating an image, \our{} modifies the trajectory, improving the dog's paw while reducing the warm spot effect visible on the overlay. Images for prompt: "A dog operating a coffee shop, professional, detailed apron, bustling cafĂ©, warm lighting, cozy."
}
\label{fig:maskintiem_flux2}
\end{figure*}

\begin{figure*}[!h]
\centering
\setlength{\tabcolsep}{1.2pt}
\renewcommand{\arraystretch}{0.9}
\begin{tabular}{cc}
FLUX.2 [dev]\hspace{80pt}+\our{} & FLUX.2 [dev]\hspace{80pt}+\our{}    \\
\includegraphics[width=0.49\textwidth]{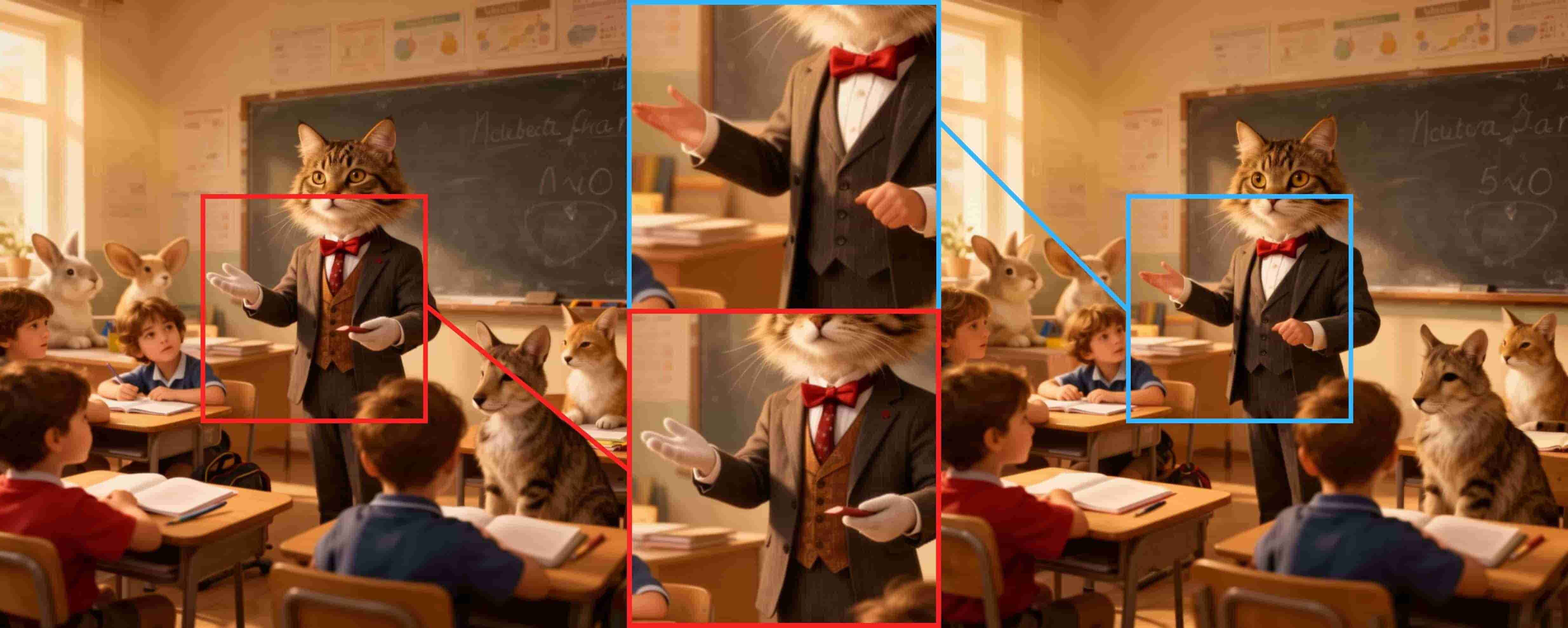} &
\includegraphics[width=0.49\textwidth]{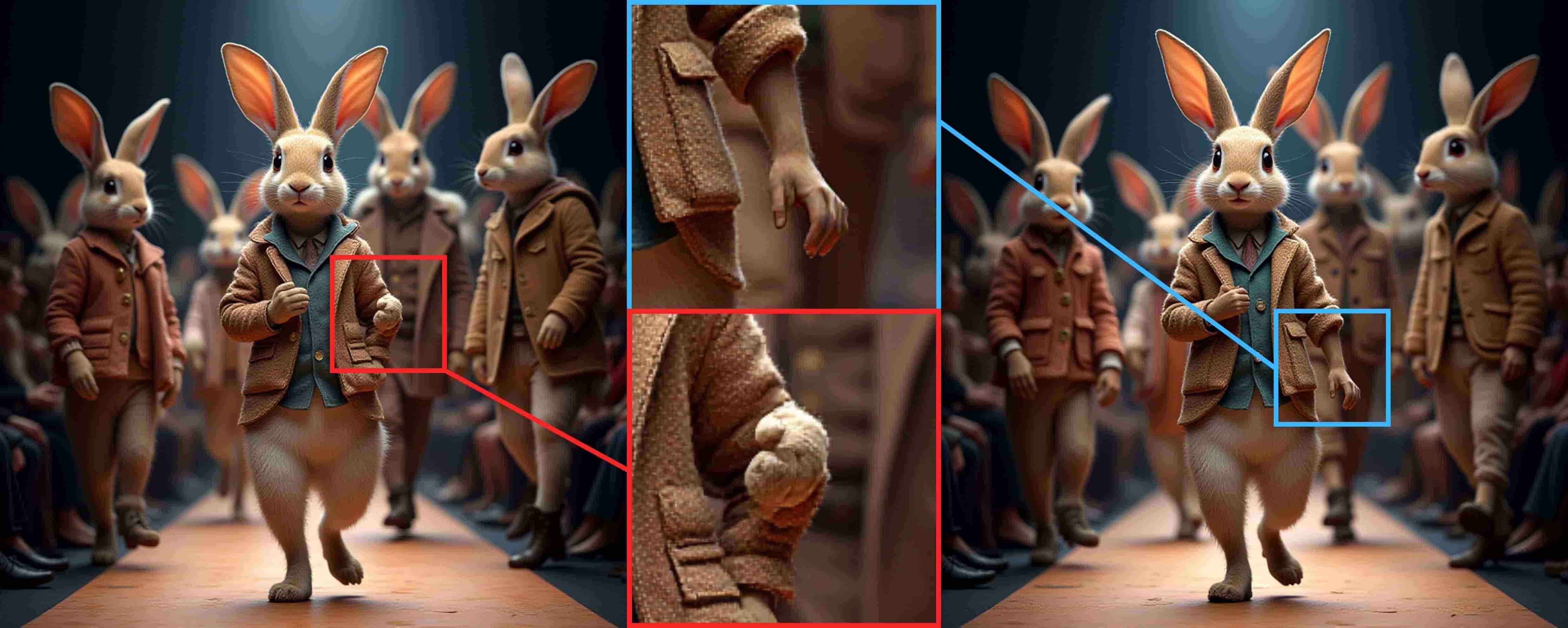} 
\end{tabular}
\caption{\textbf{Images for the FLUX.2 [dev] on the \textit{animals} dataset using \our{} for 30 inference steps.} For prompts that combine multiple semantic entities (human-animal), the model may implicitly bias the generated anatomy toward a single component.}
\label{fig:flux2_animal}
\end{figure*}

\begin{figure*}[!h]
\centering
\setlength{\tabcolsep}{1.2pt}
\renewcommand{\arraystretch}{0.9}
\begin{tabular}{ccccc}
FLUX.1 [dev] & +DiffDoctor & +HPSv2 & +\our{} \\

\includegraphics[width=0.22\textwidth]{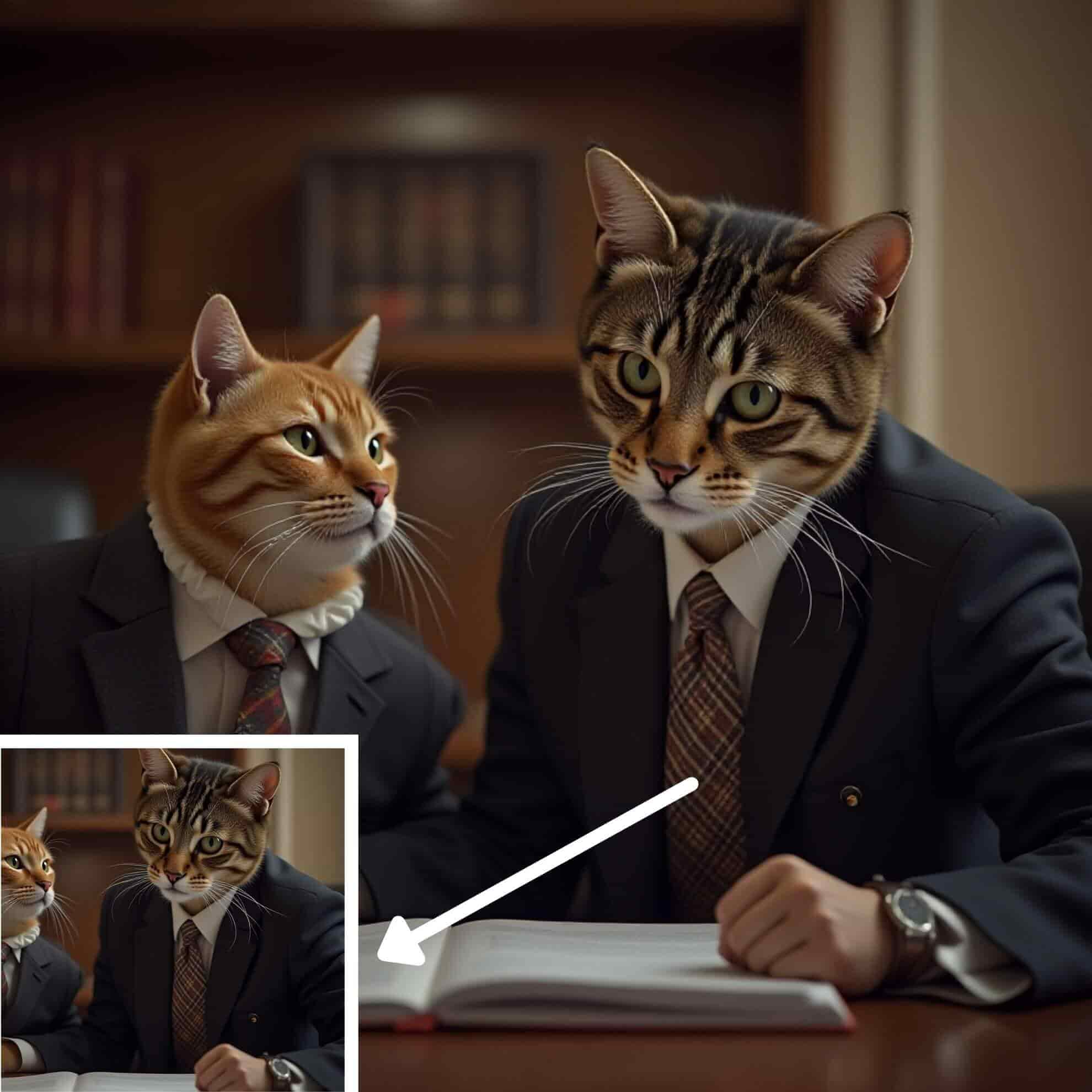} &
\includegraphics[width=0.22\textwidth]{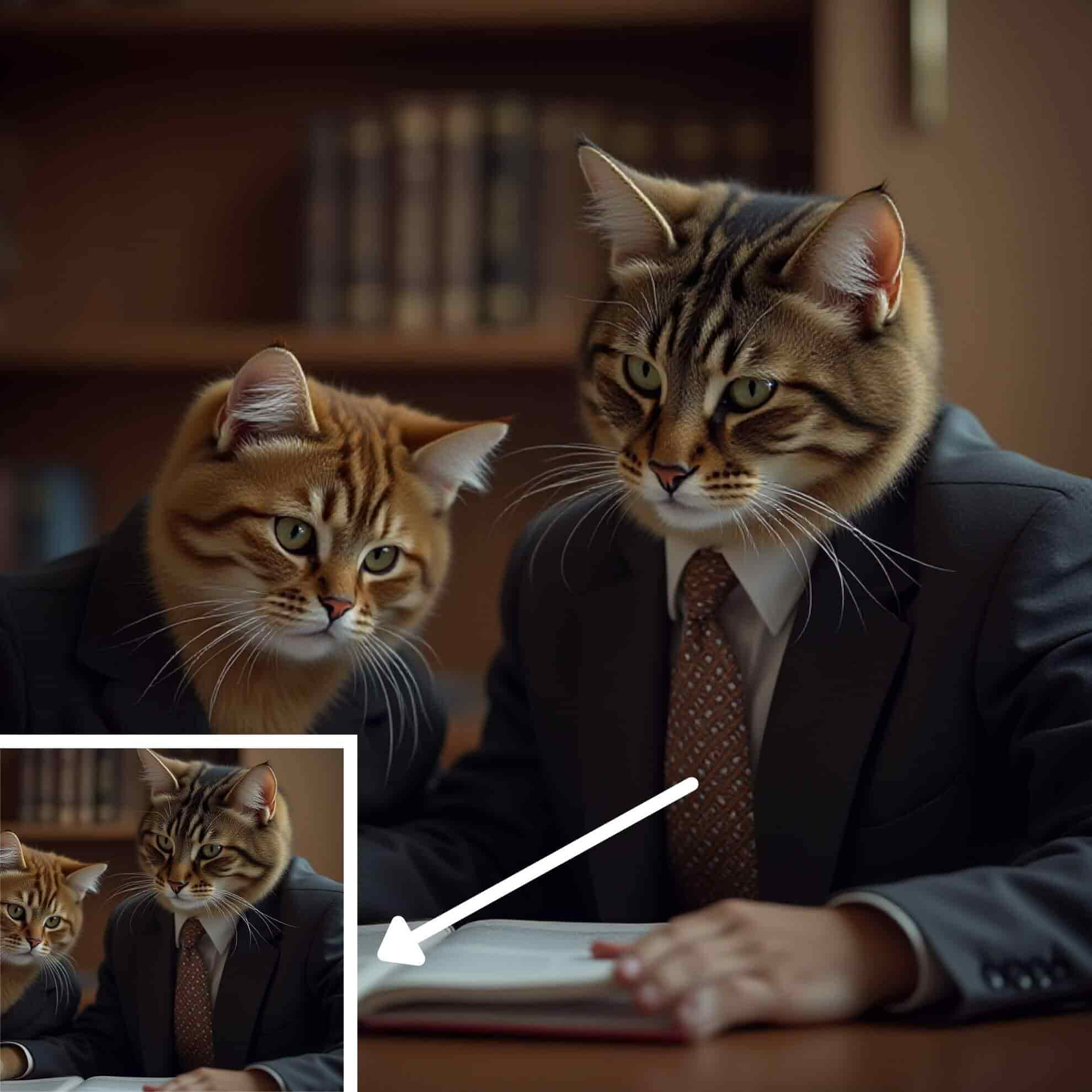} &
\includegraphics[width=0.22\textwidth]{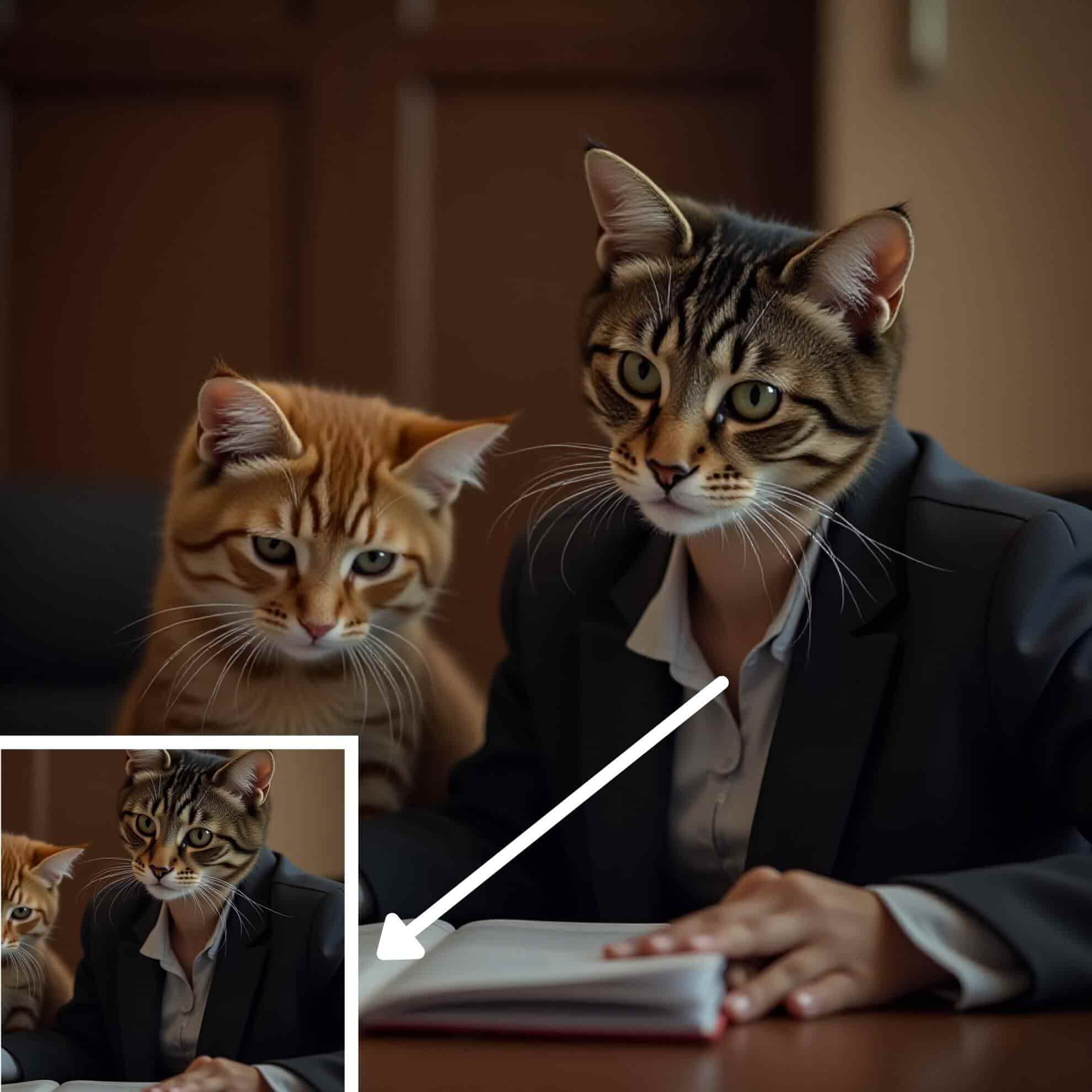} &
\includegraphics[width=0.22\textwidth]{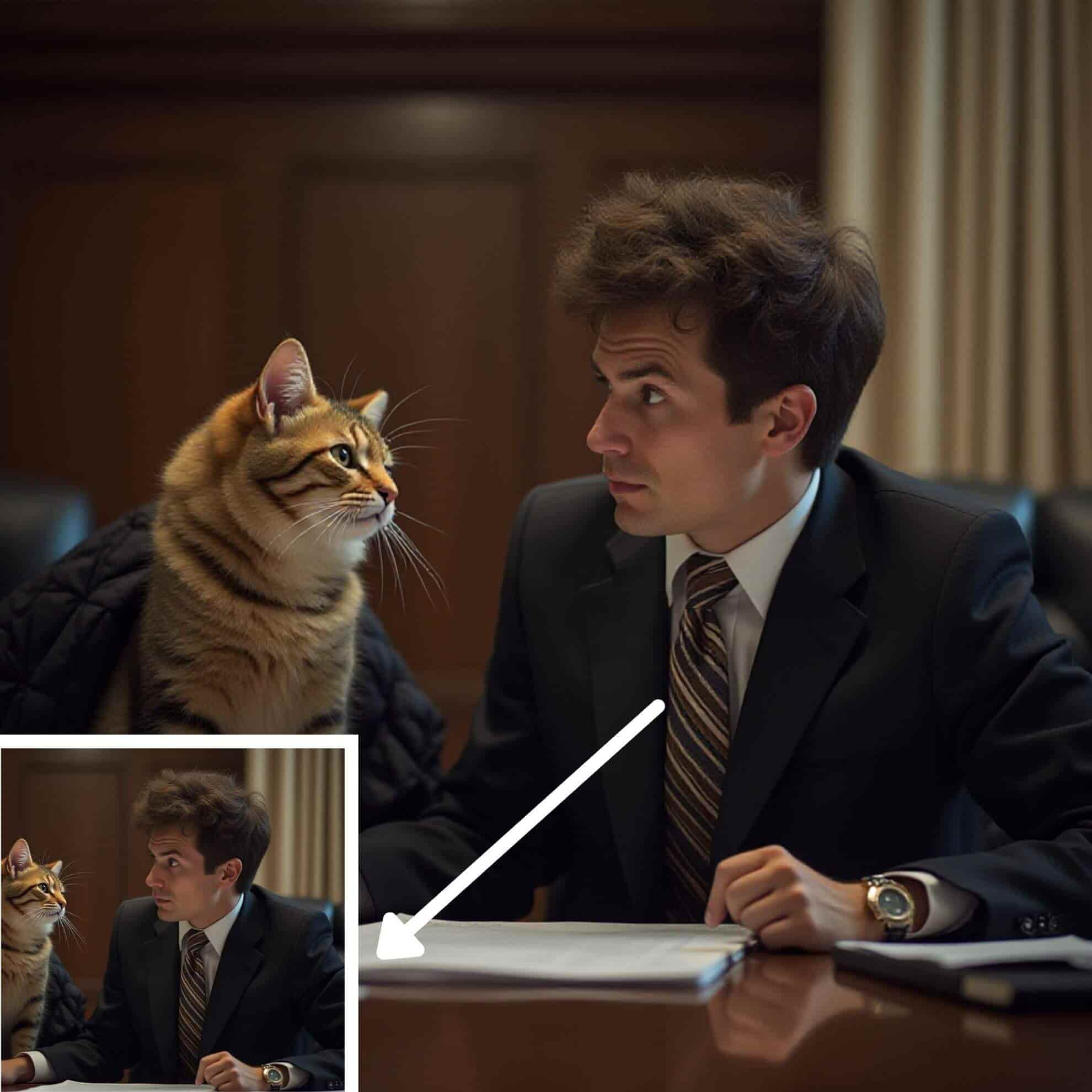} \\

\includegraphics[width=0.22\textwidth]{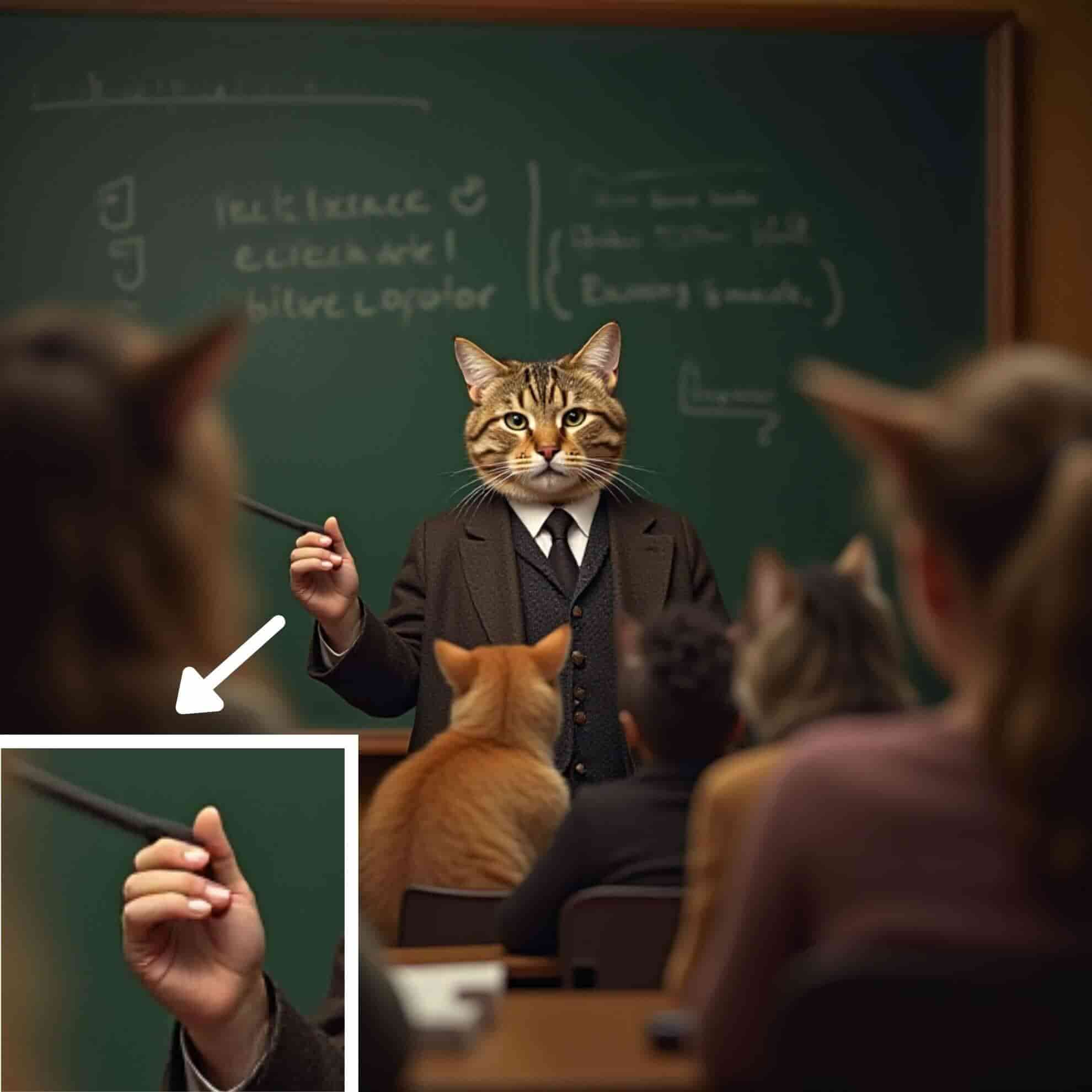} &
\includegraphics[width=0.22\textwidth]{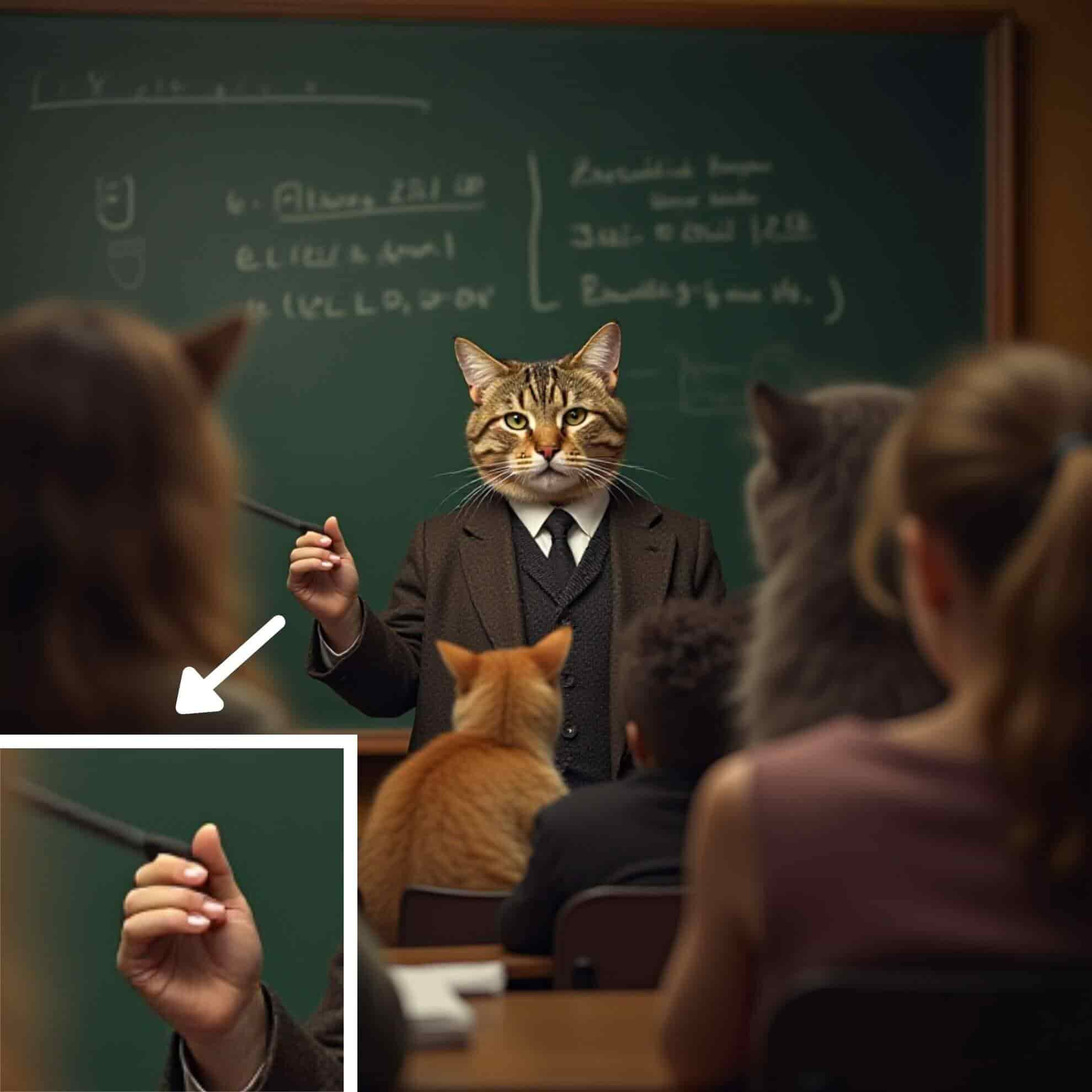} &
\includegraphics[width=0.22\textwidth]{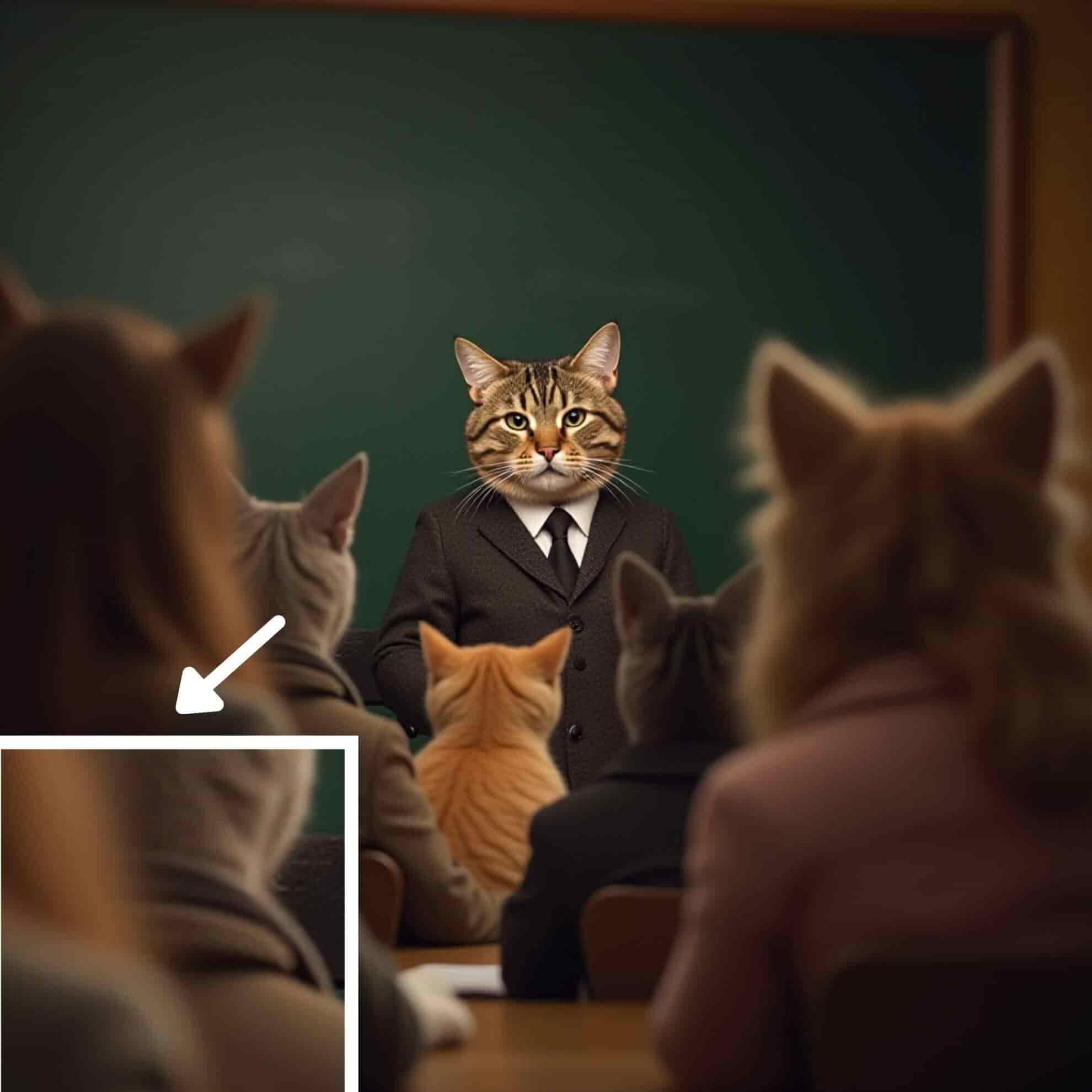} &
\includegraphics[width=0.22\textwidth]{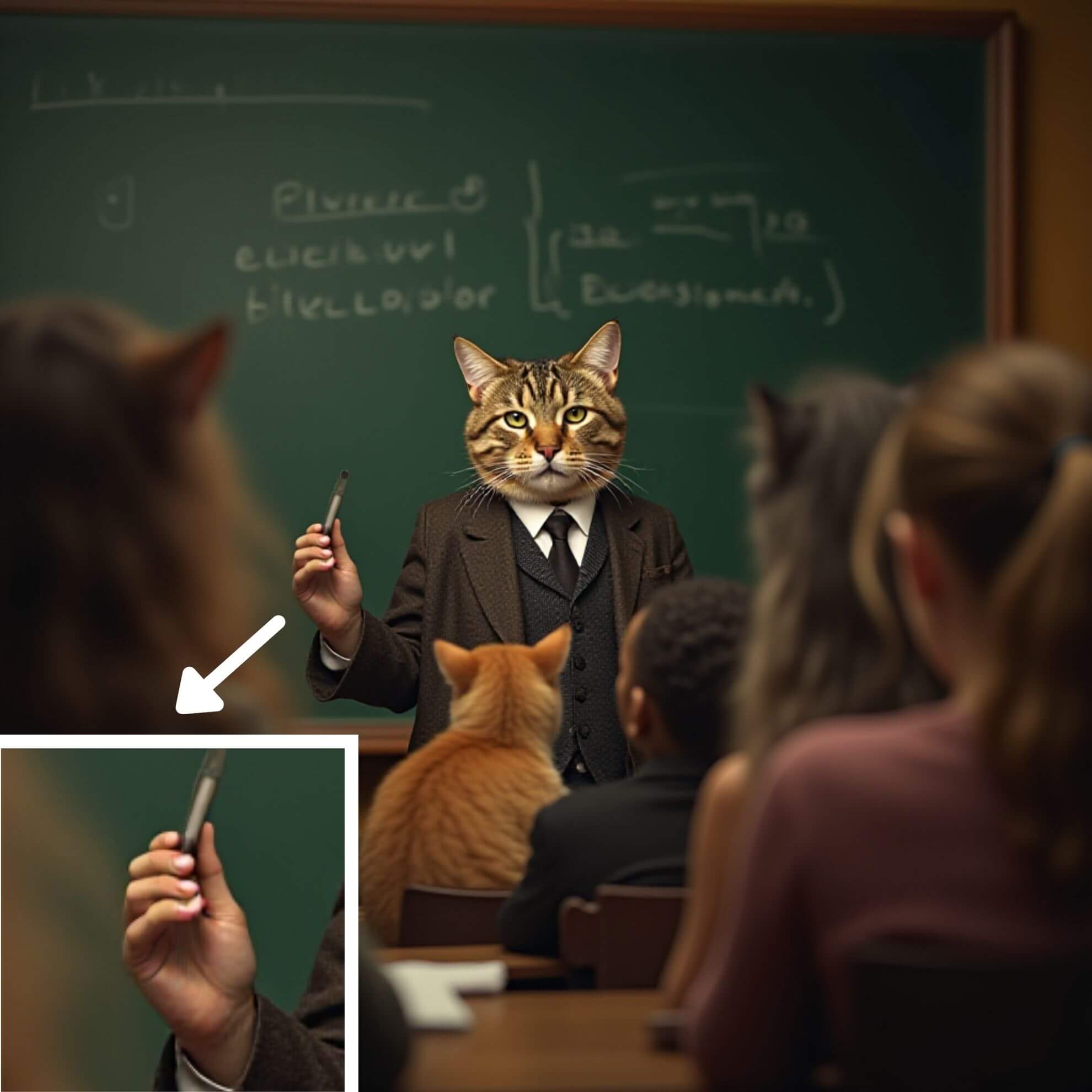} \\

\includegraphics[width=0.22\textwidth]{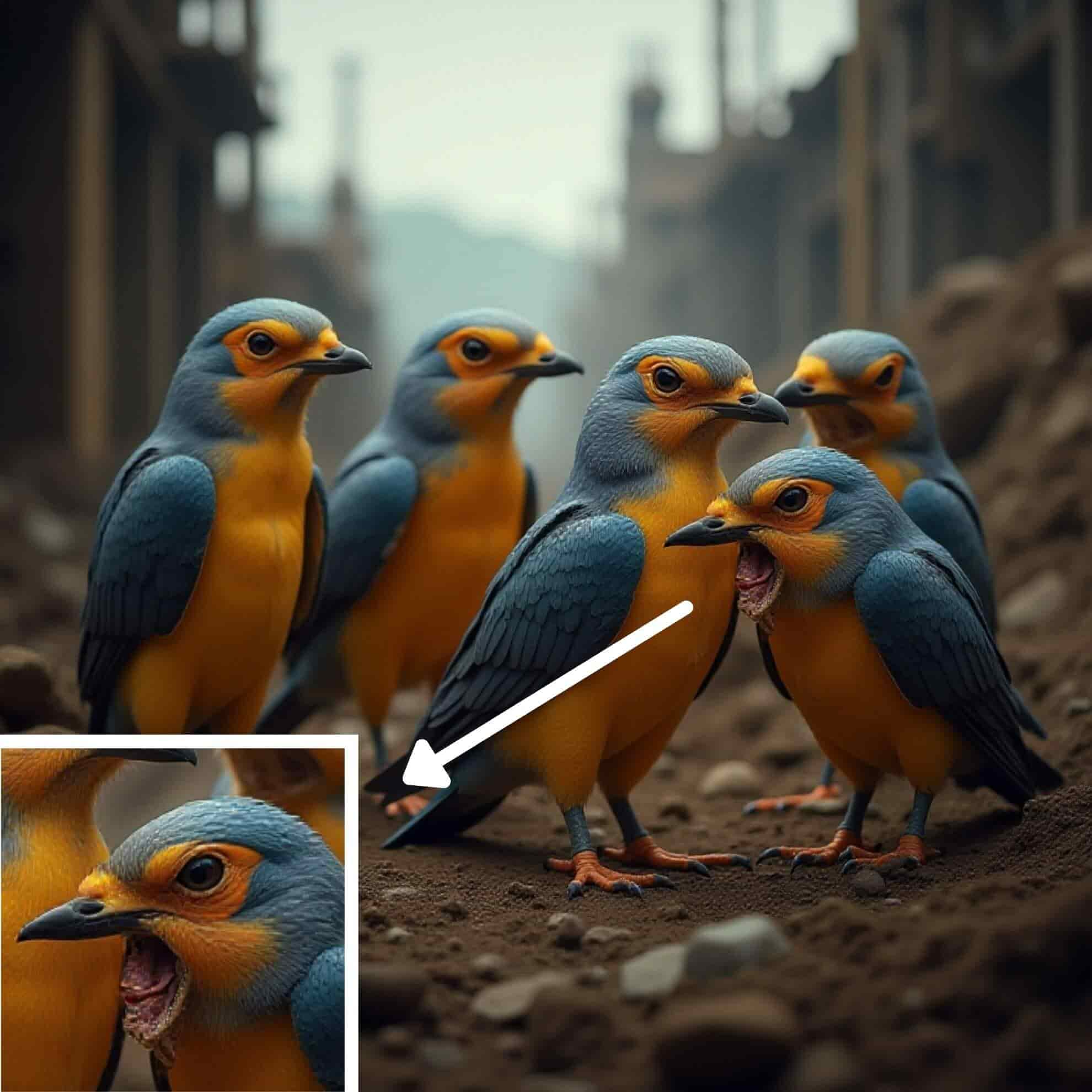} &
\includegraphics[width=0.22\textwidth]{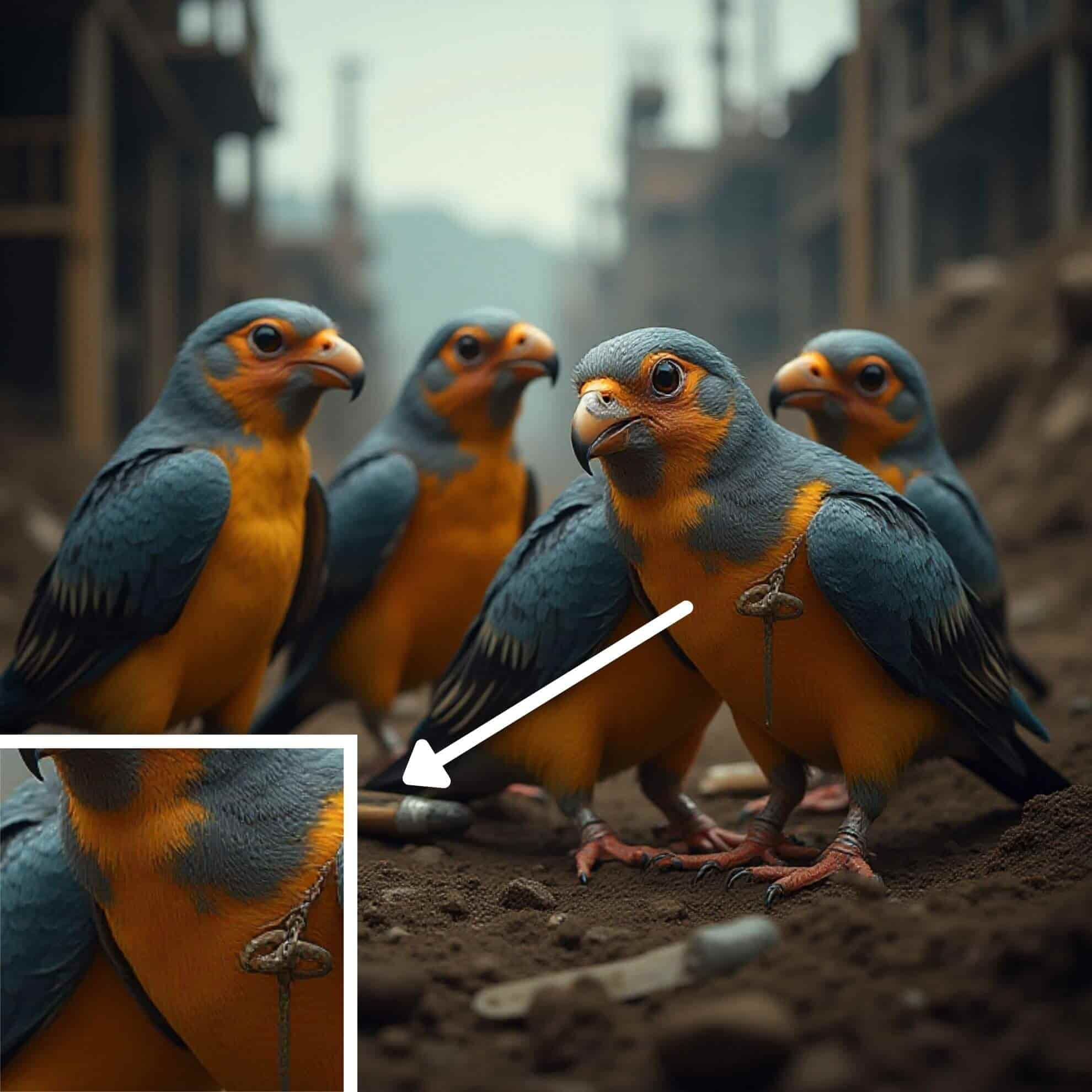} &
\includegraphics[width=0.22\textwidth]{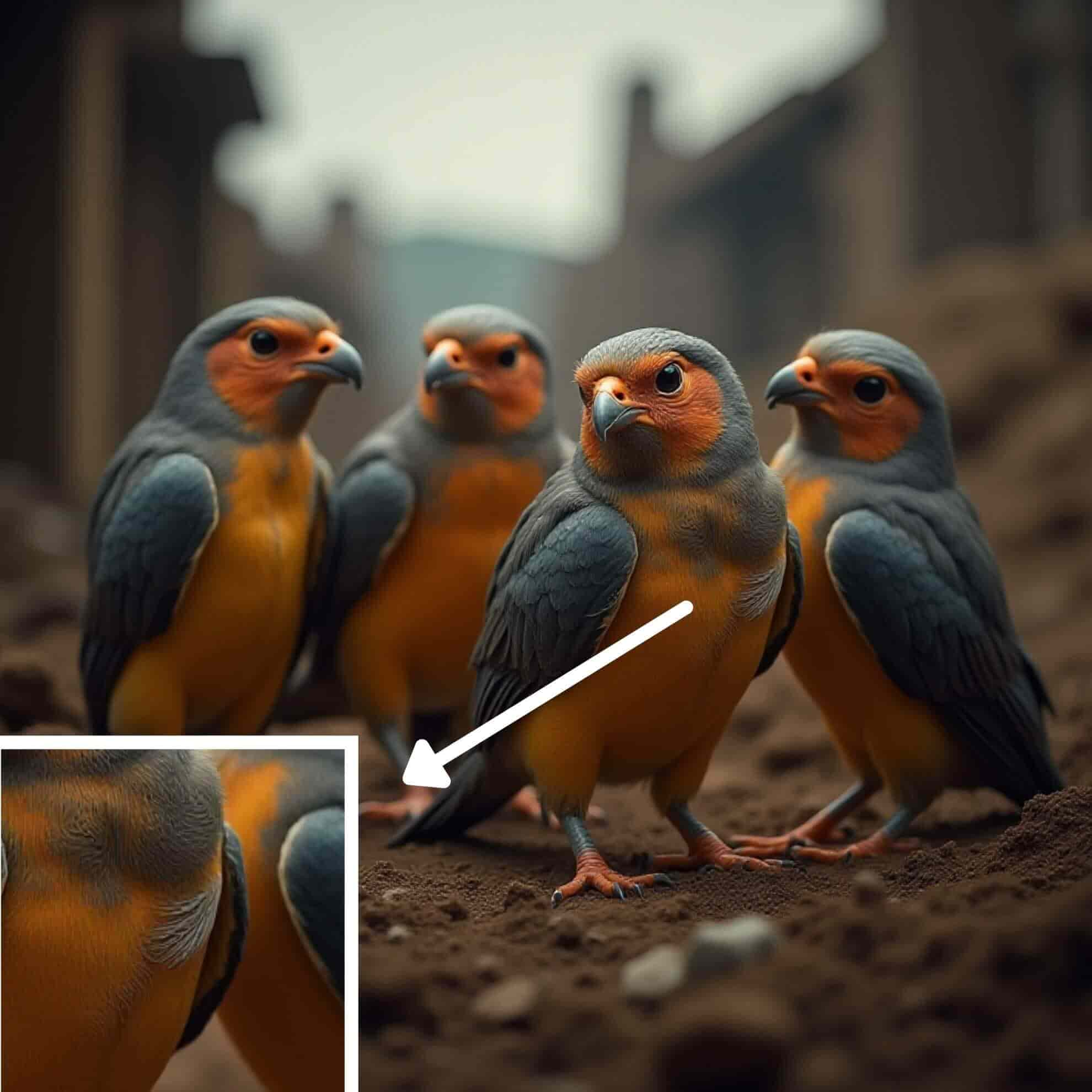} &
\includegraphics[width=0.22\textwidth]{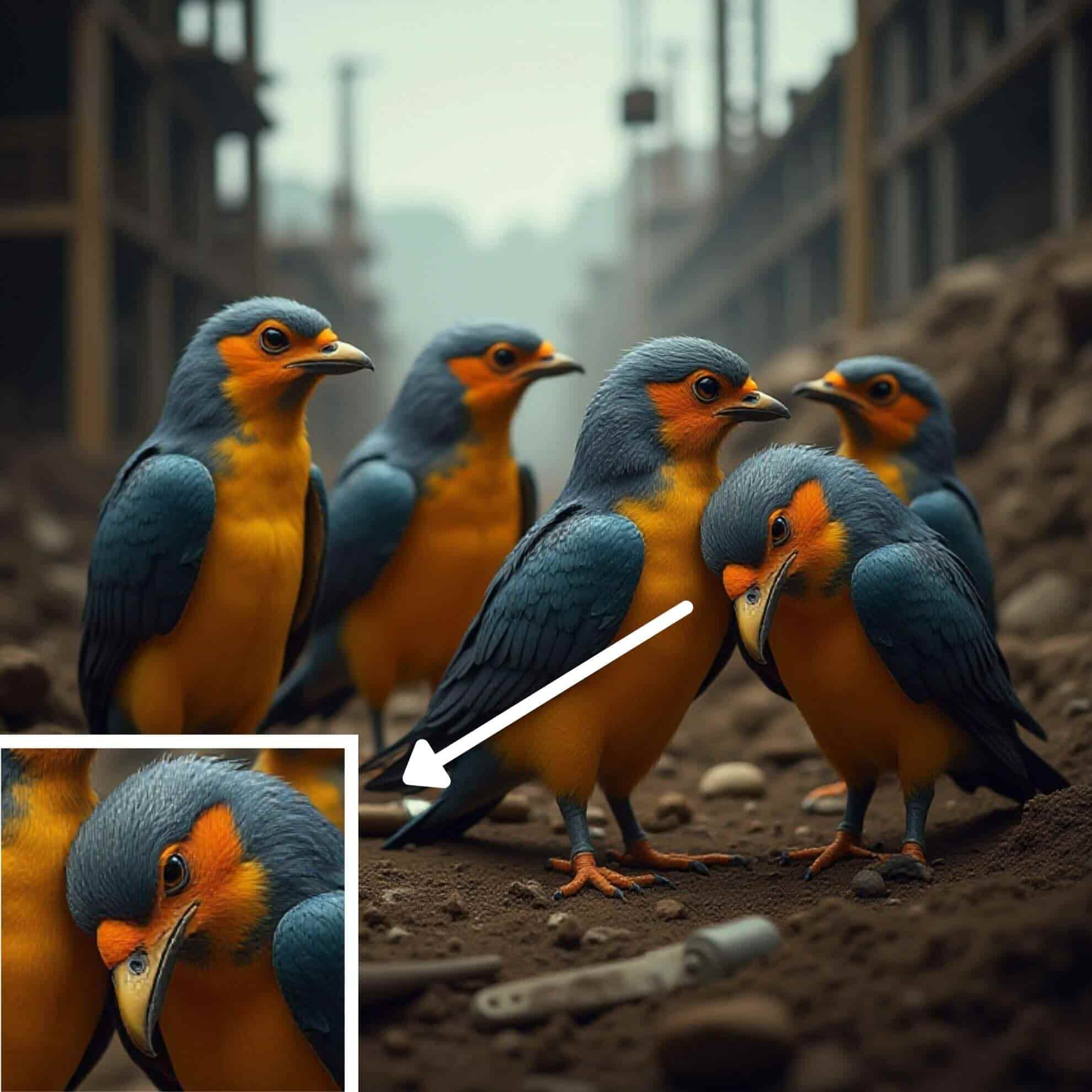} \\

\includegraphics[width=0.22\textwidth]{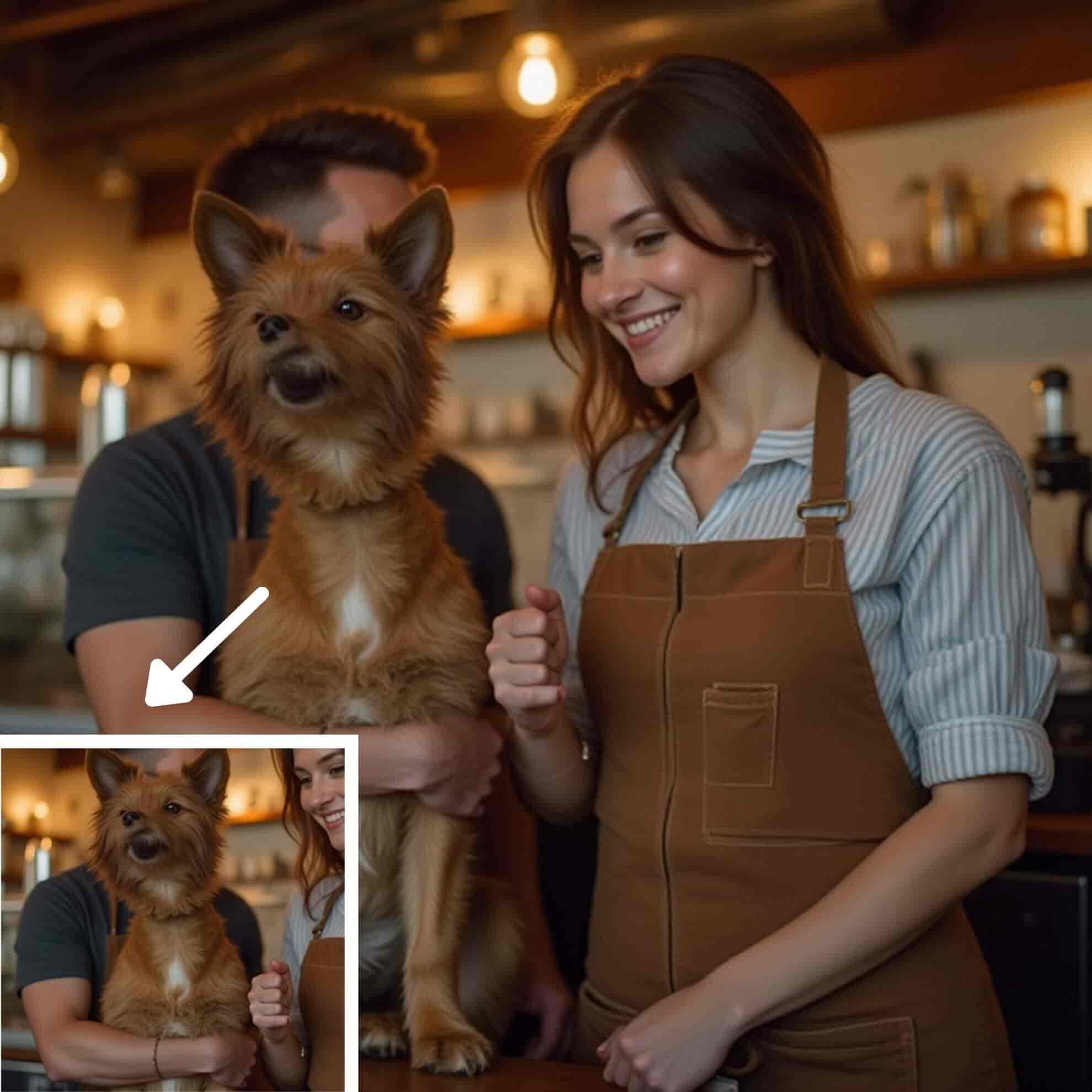} &
\includegraphics[width=0.22\textwidth]{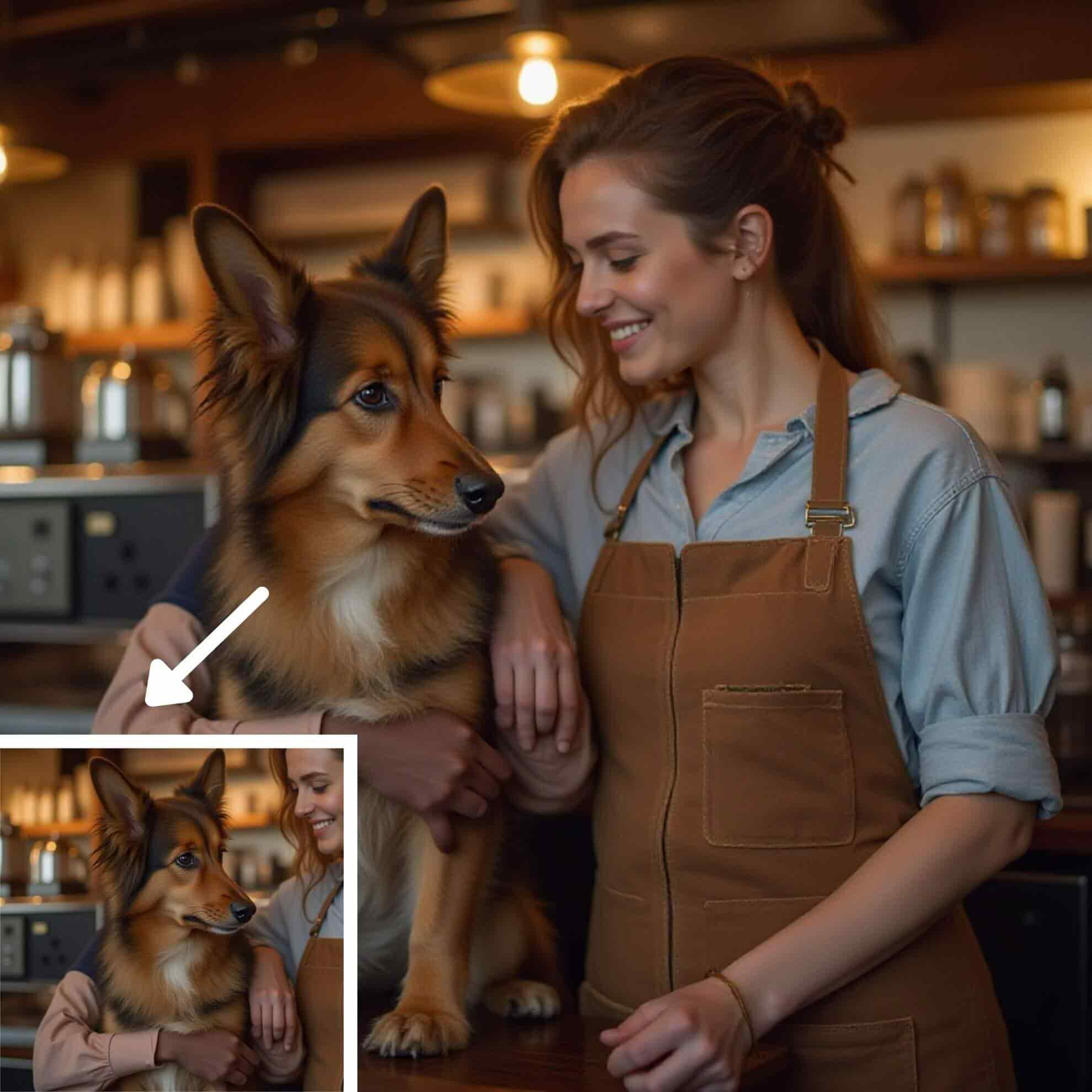} &
\includegraphics[width=0.22\textwidth]{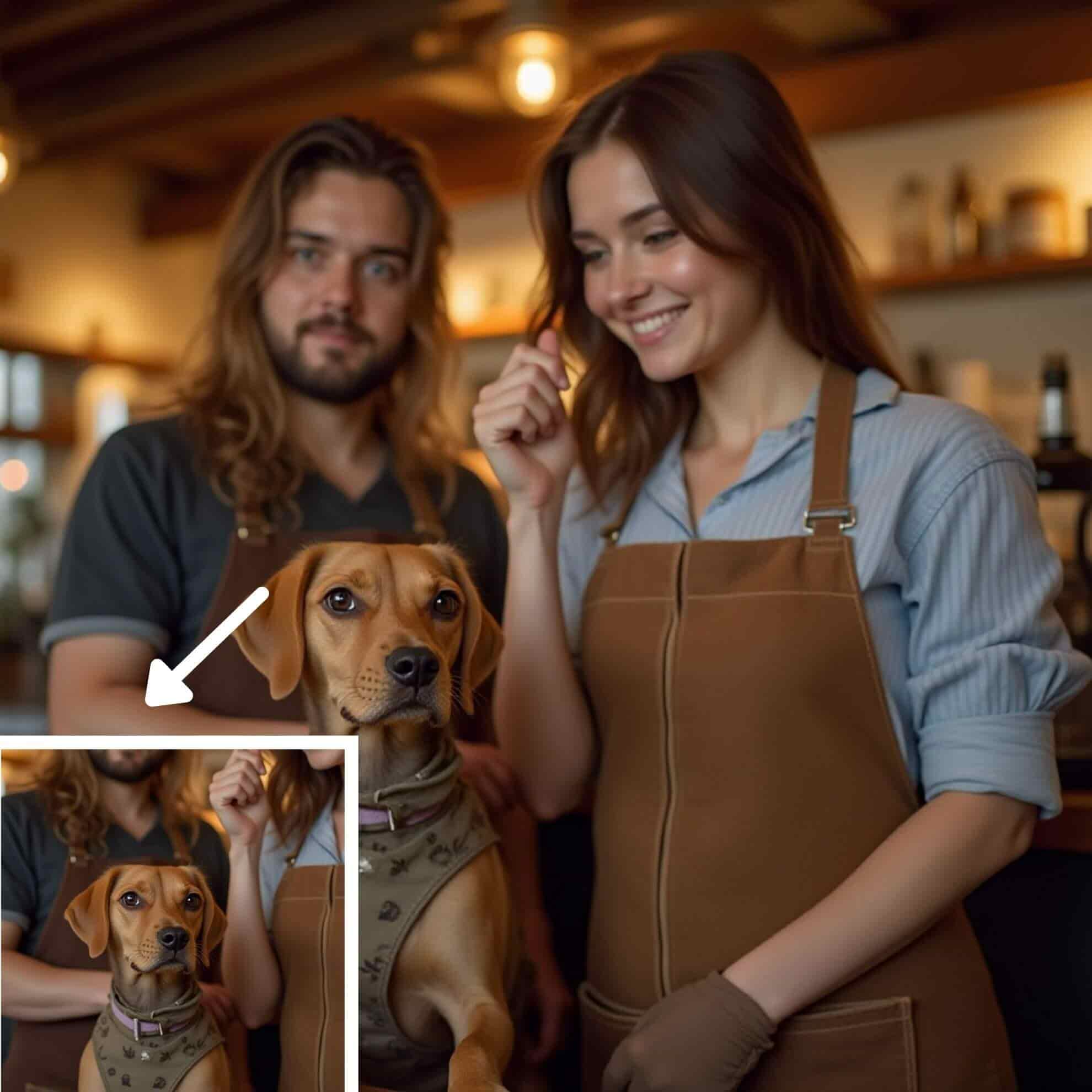} &
\includegraphics[width=0.22\textwidth]{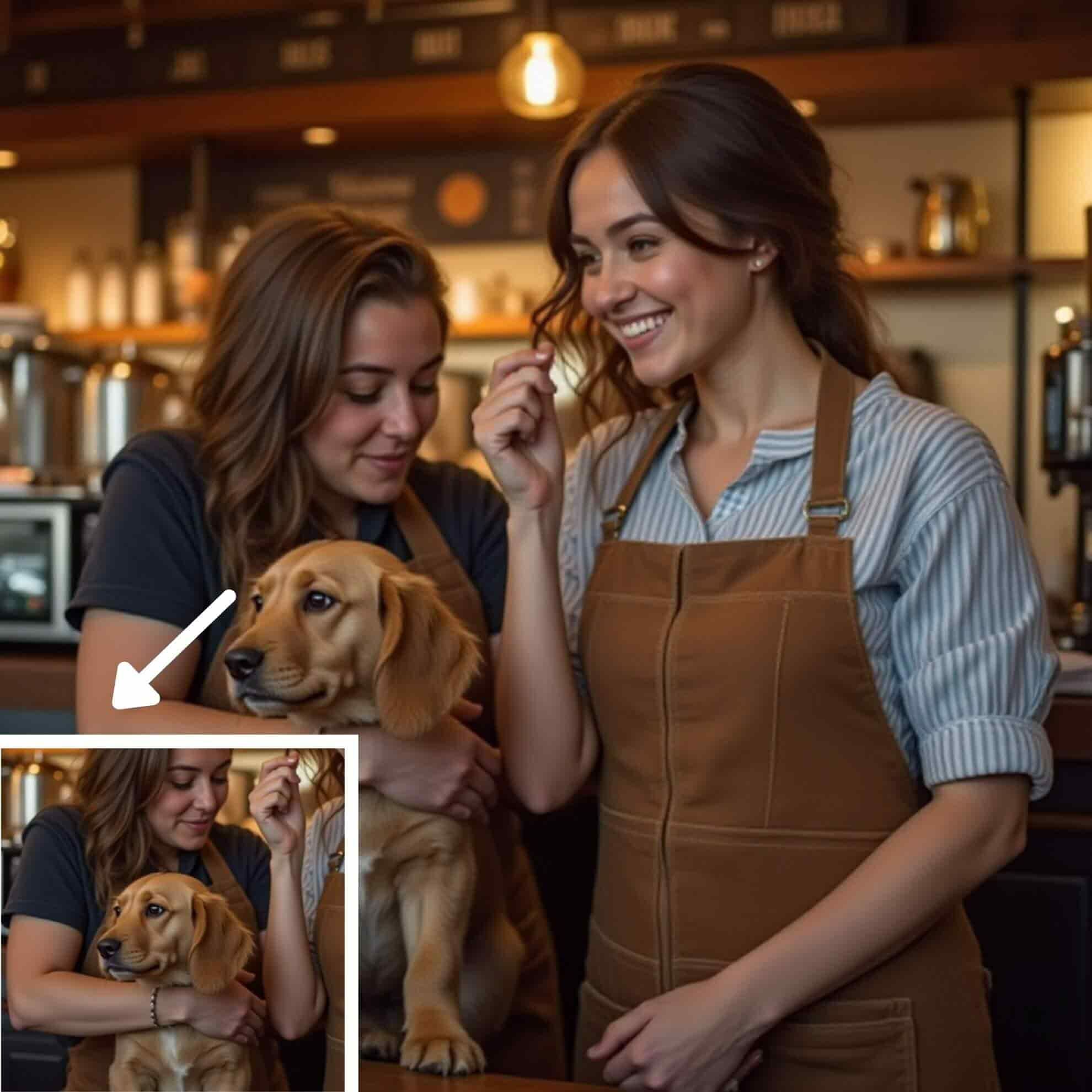}

\end{tabular}
\caption{
\textbf{comparison with other methods} Images for the FLUX.1 [dev] on the \textit{animals} dataset using \our{} for 10 inference steps.
}

\label{fig:dev_animal}
\end{figure*}

\newpage

\subsection{Effect of $\lambda_t$ on the Generation Trajectory}
\label{subsec:effect_lambda}
\begin{figure*}[!h]
\centering
\setlength{\tabcolsep}{1.2pt}
\renewcommand{\arraystretch}{0.9}
\begin{tabular}{cccc}
FLUX.1 [dev] & $\lambda_{start}=45,\lambda_{end}=45, p=2$ & $\lambda_{start}=1, \lambda_{end}=1, p=2$ & $\lambda_{start}=45,\lambda_{end}=1,p=2$ \\
\includegraphics[width=0.24\textwidth]{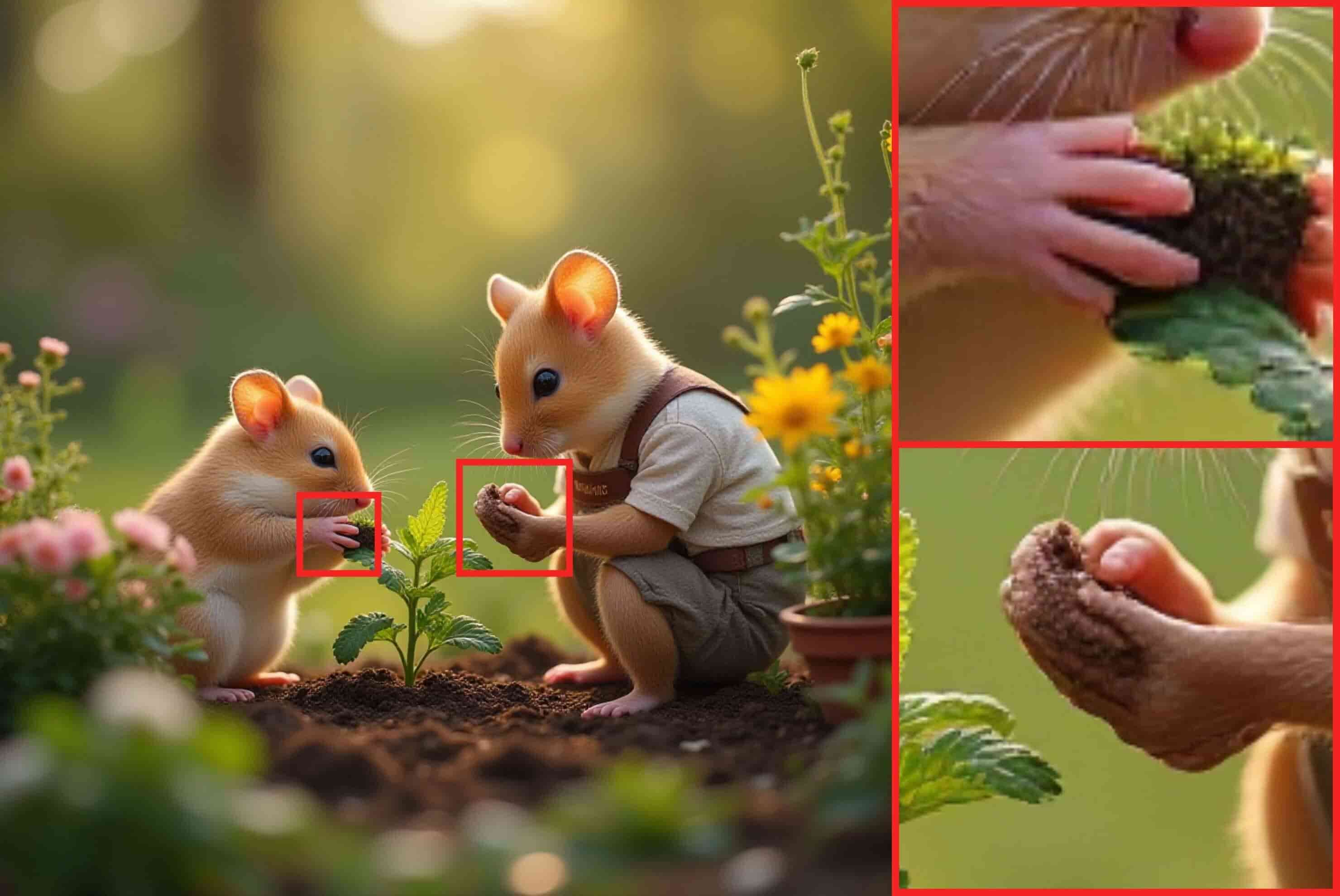} &
\includegraphics[width=0.24\textwidth]{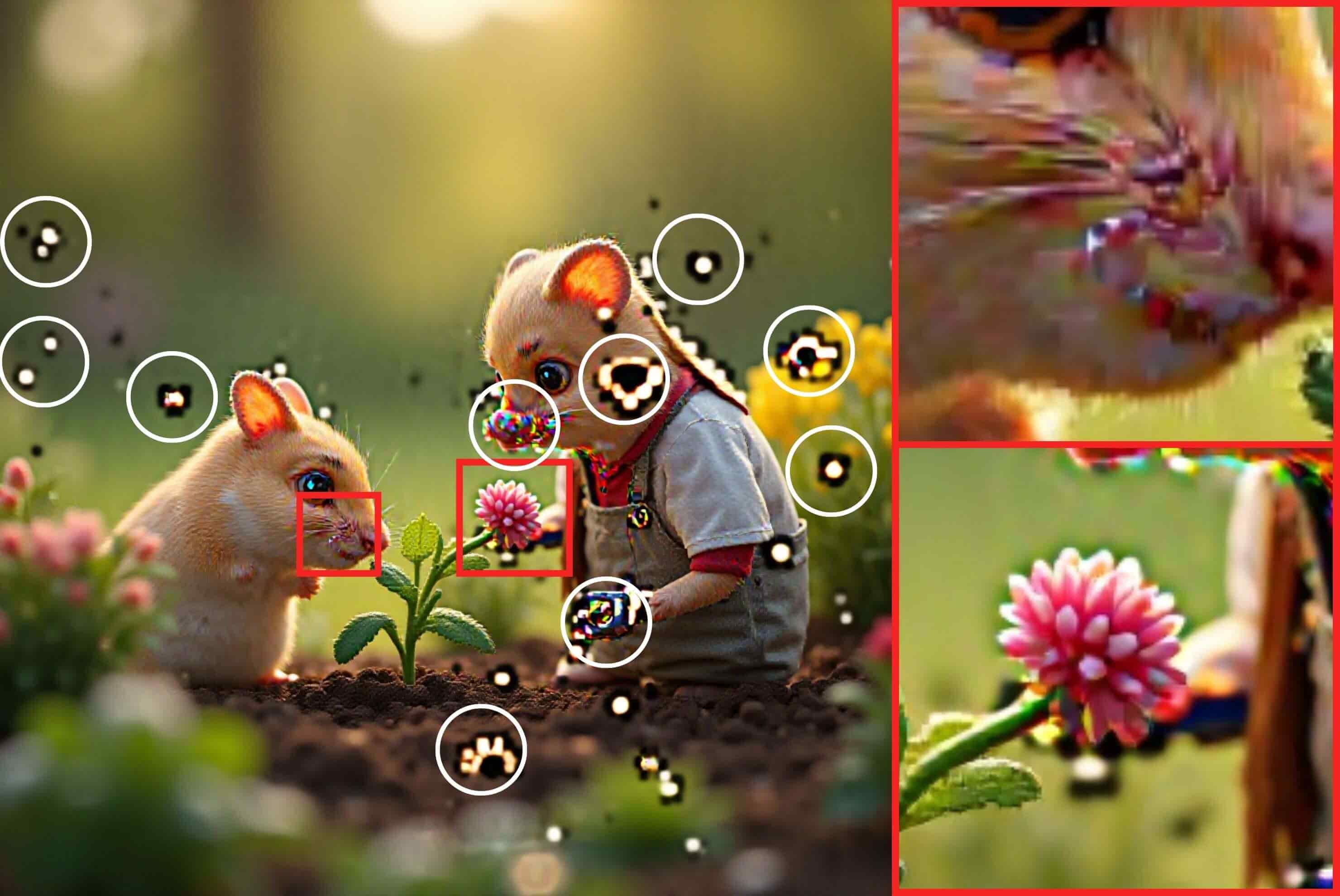} &
\includegraphics[width=0.24\textwidth]{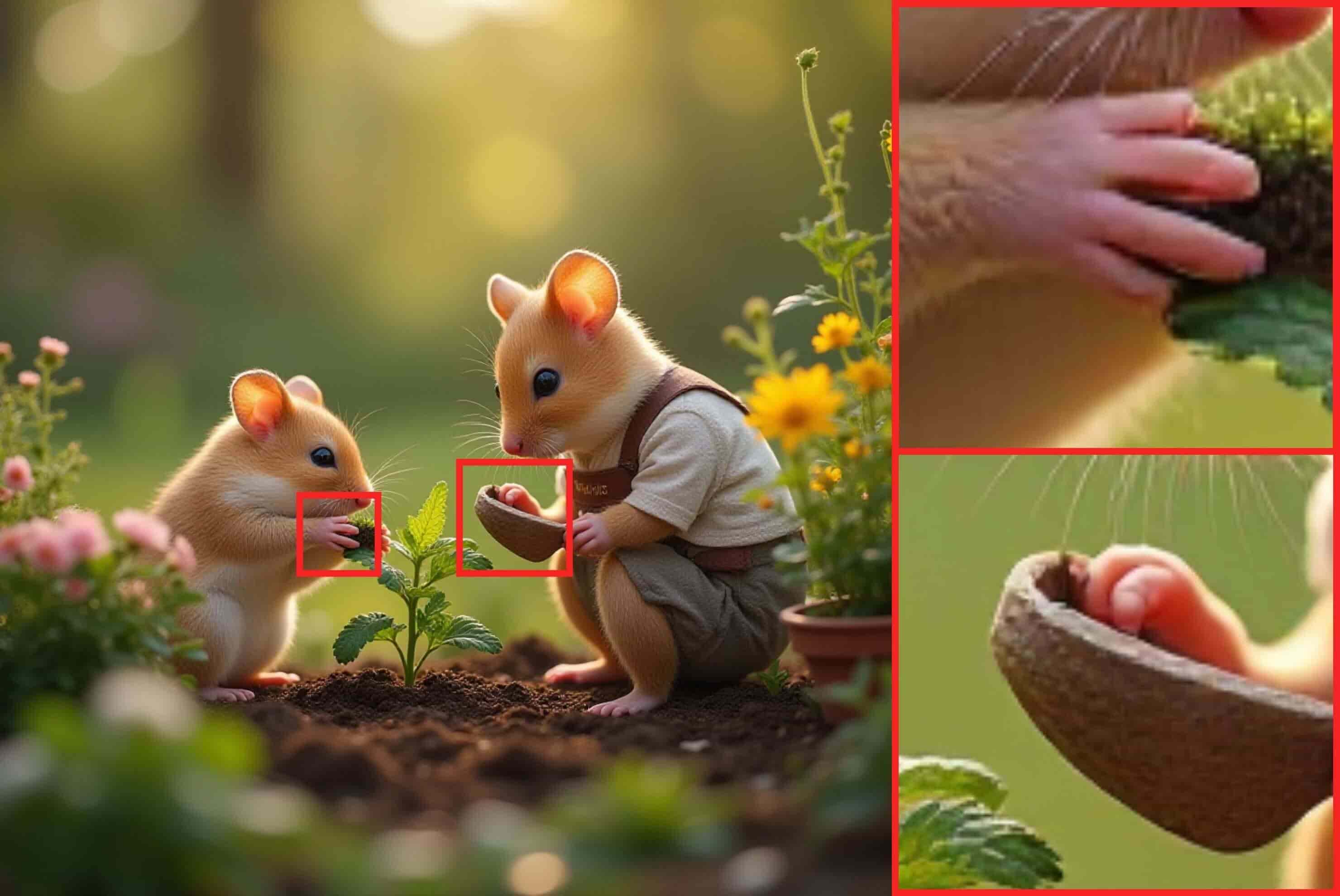} &
\includegraphics[width=0.24\textwidth]{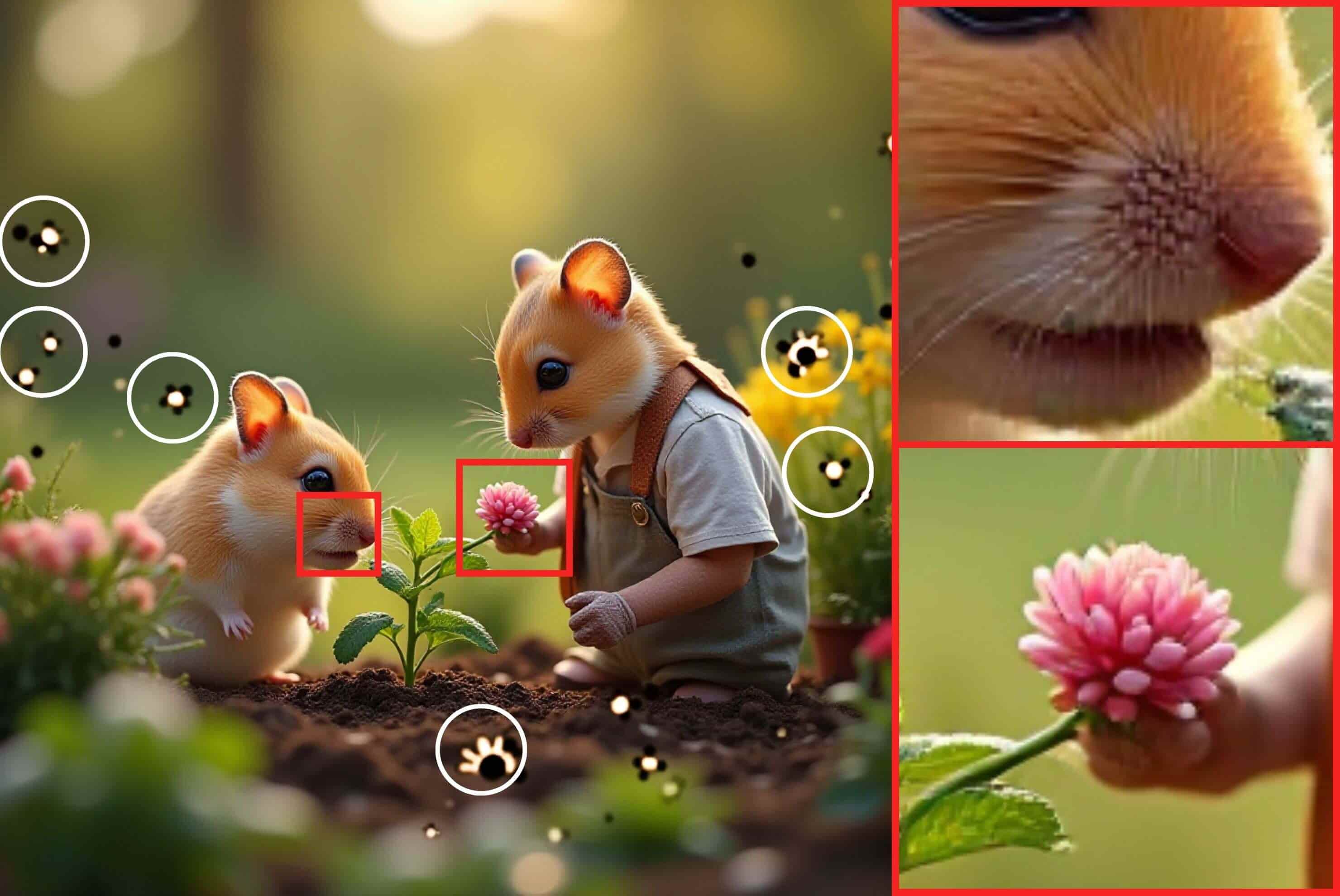} \\
\includegraphics[width=0.24\textwidth]{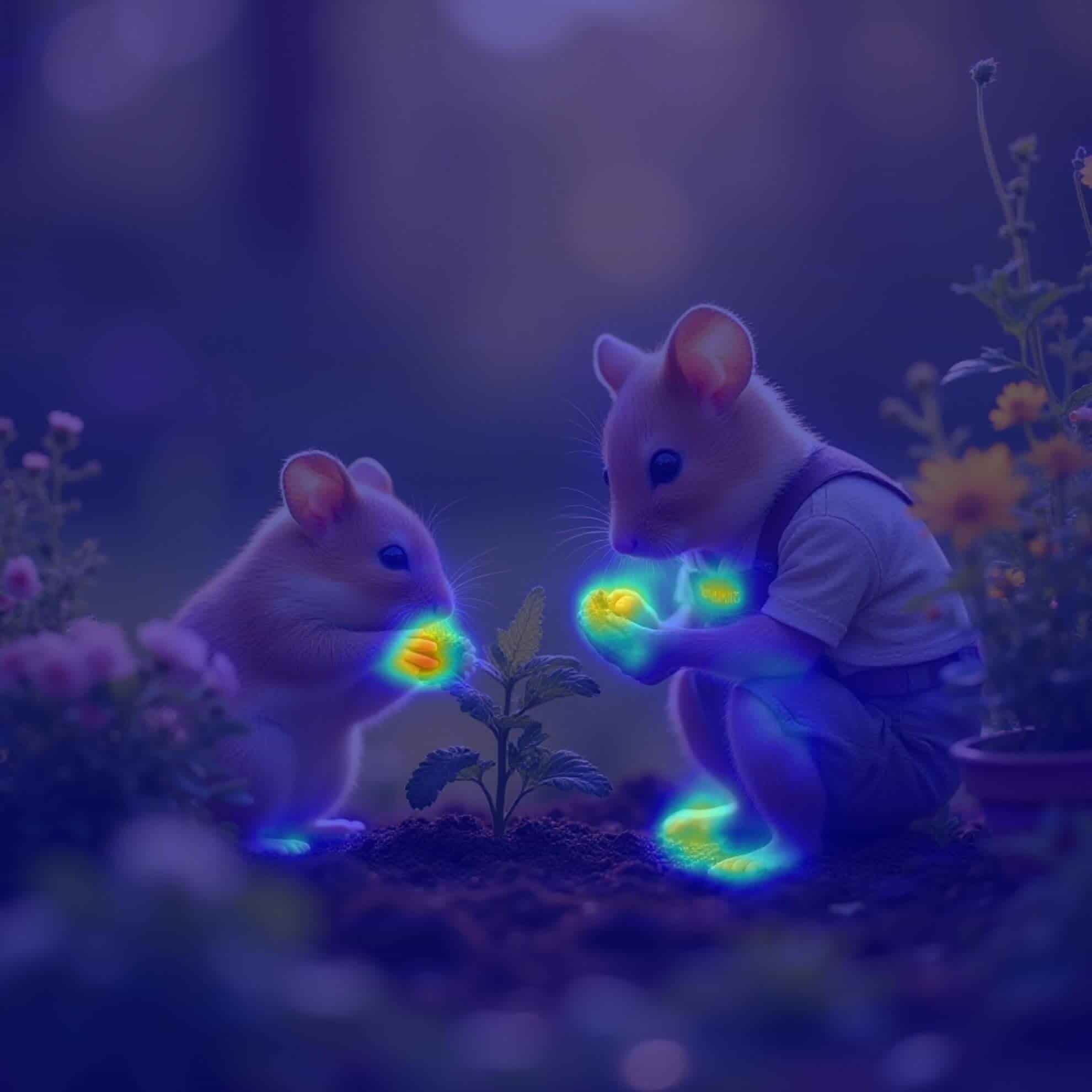} &
\includegraphics[width=0.24\textwidth]{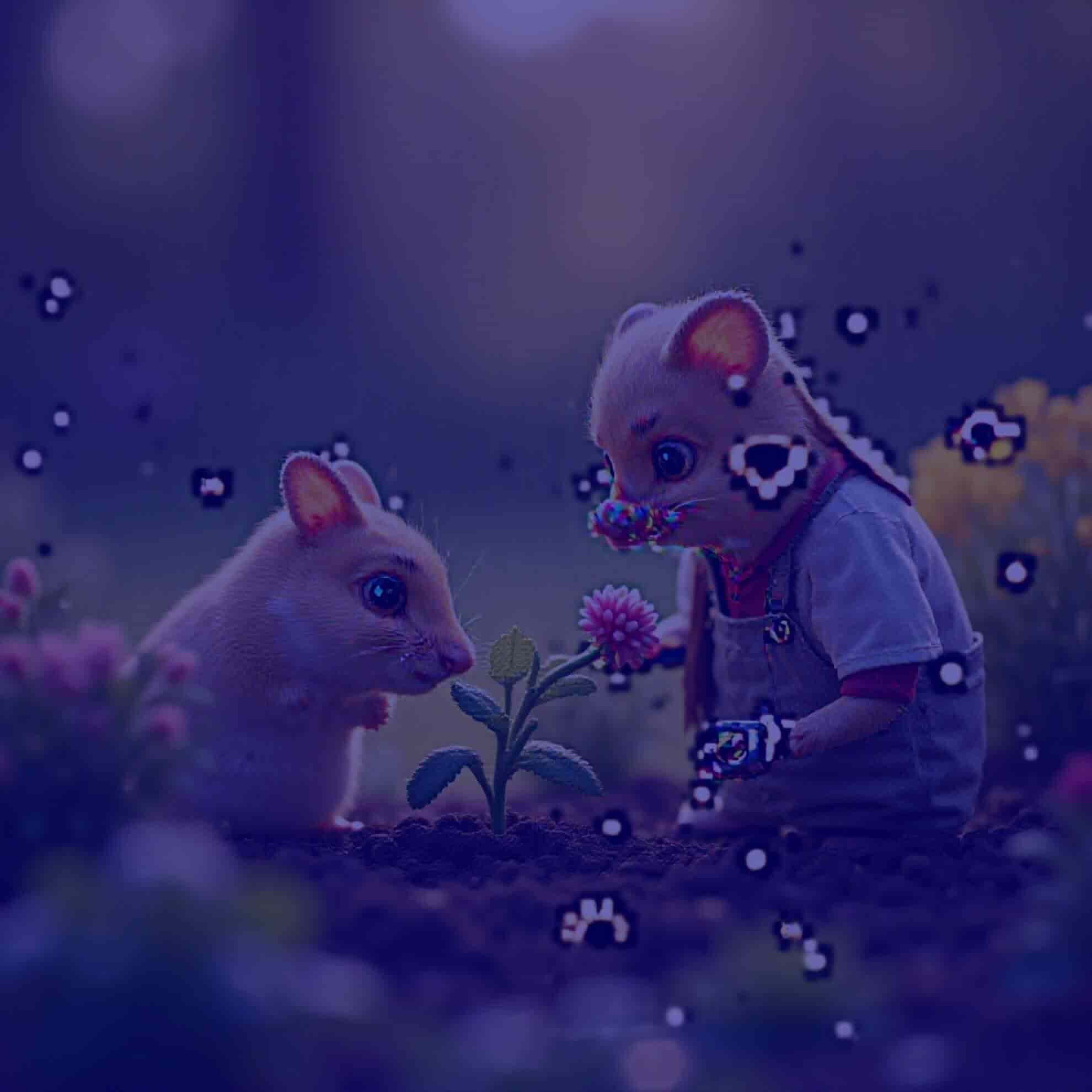} &
\includegraphics[width=0.24\textwidth]{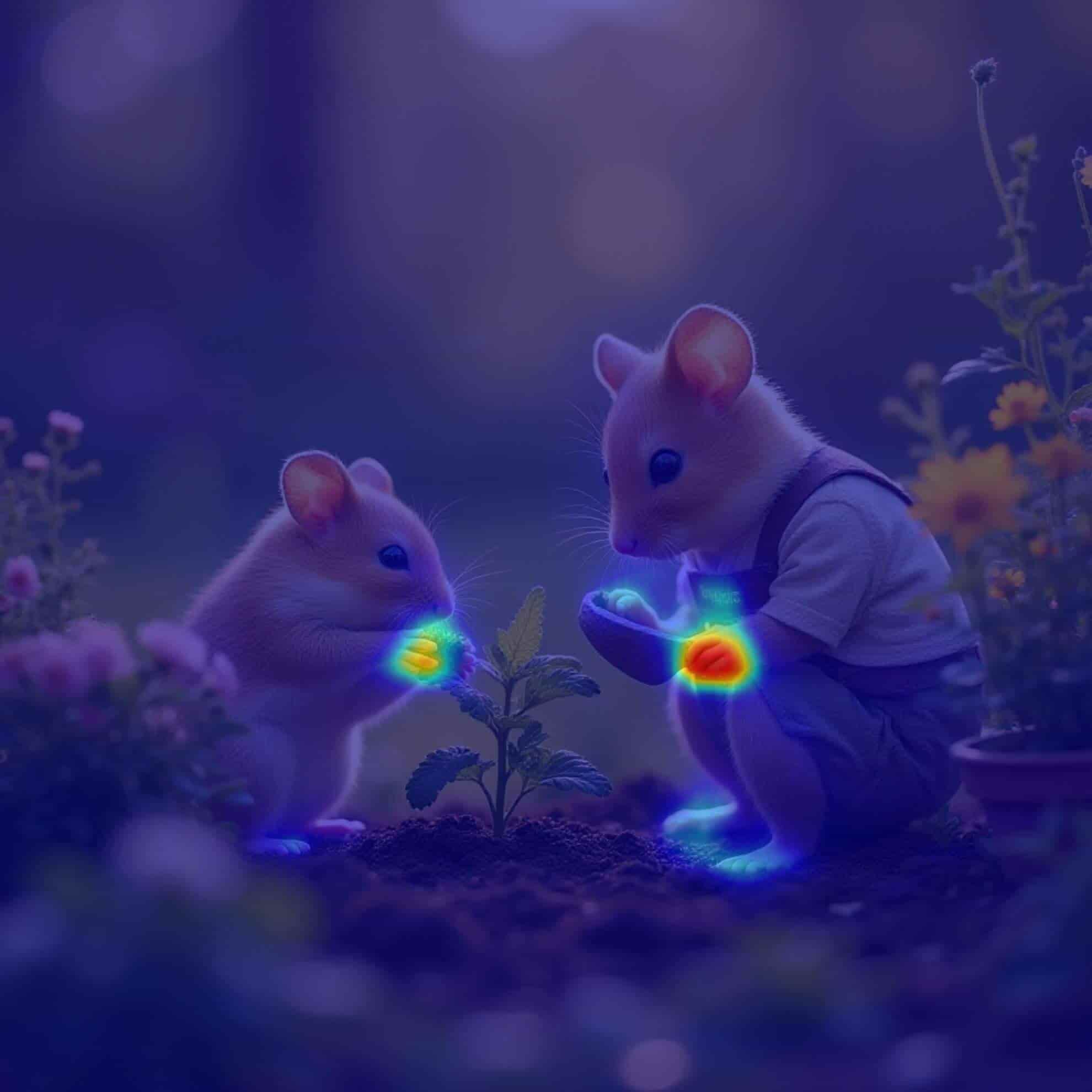} &
\includegraphics[width=0.24\textwidth]{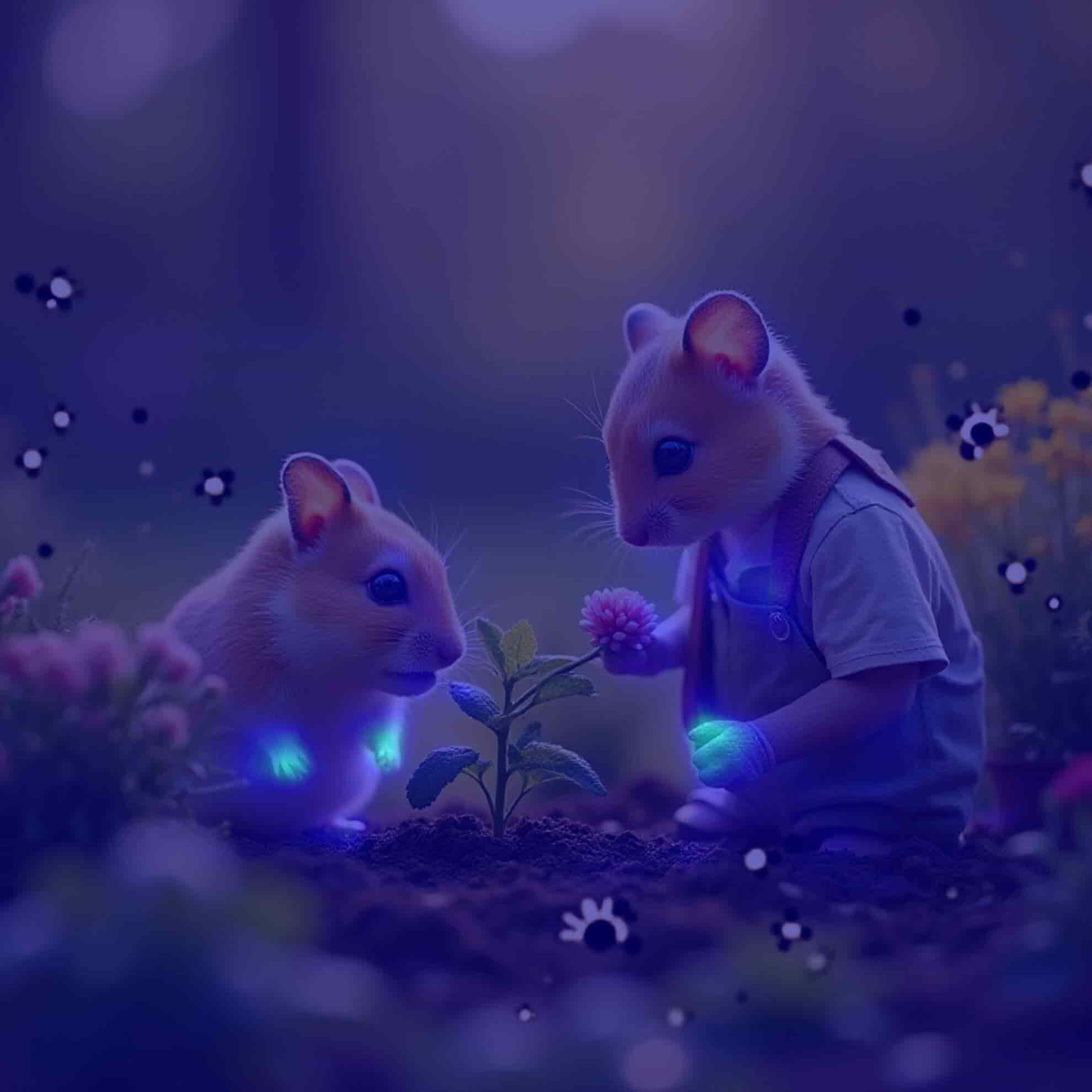} \\
$\lambda_{start}=15,\lambda_{end}=1,p=2$ & $\lambda_{start}=45,\lambda_{end}=1,p=3$ &  $\lambda_{start}=15,\lambda_{end}=1,p=3$ & $\lambda_{start}=25,\lambda_{end}=1,p=2$ \\
\includegraphics[width=0.24\textwidth]{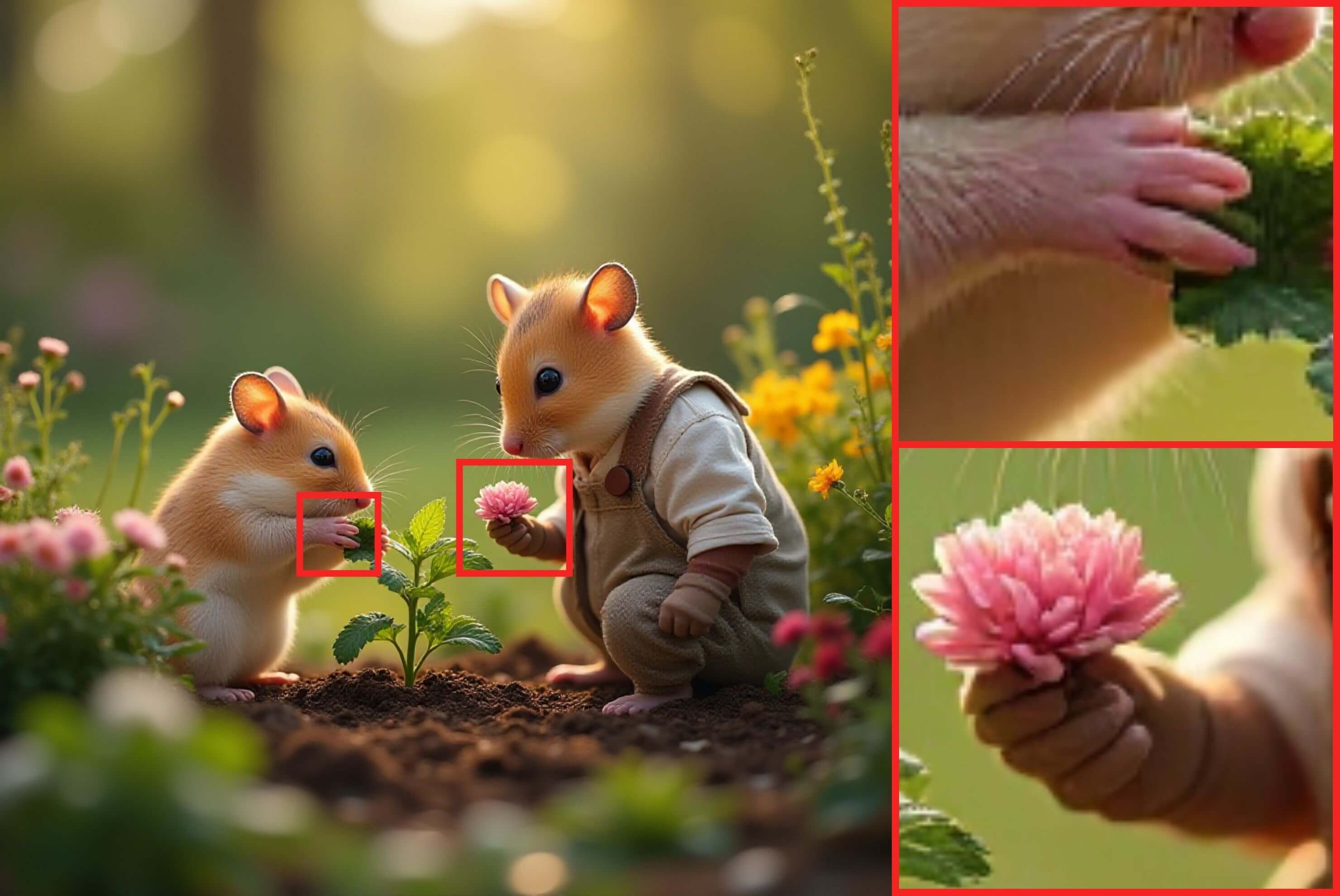} &
\includegraphics[width=0.24\textwidth]{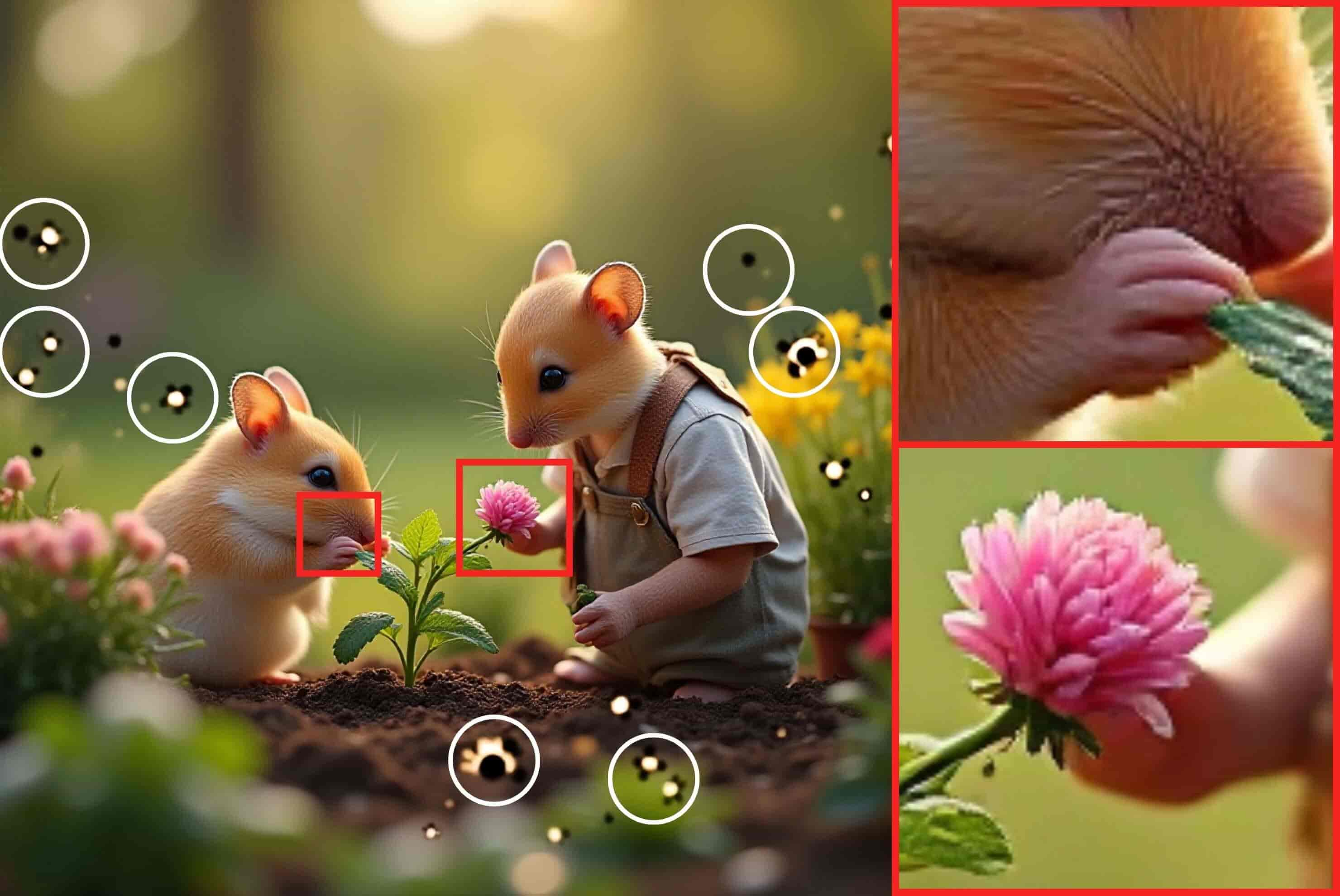} & 
\includegraphics[width=0.24\textwidth]{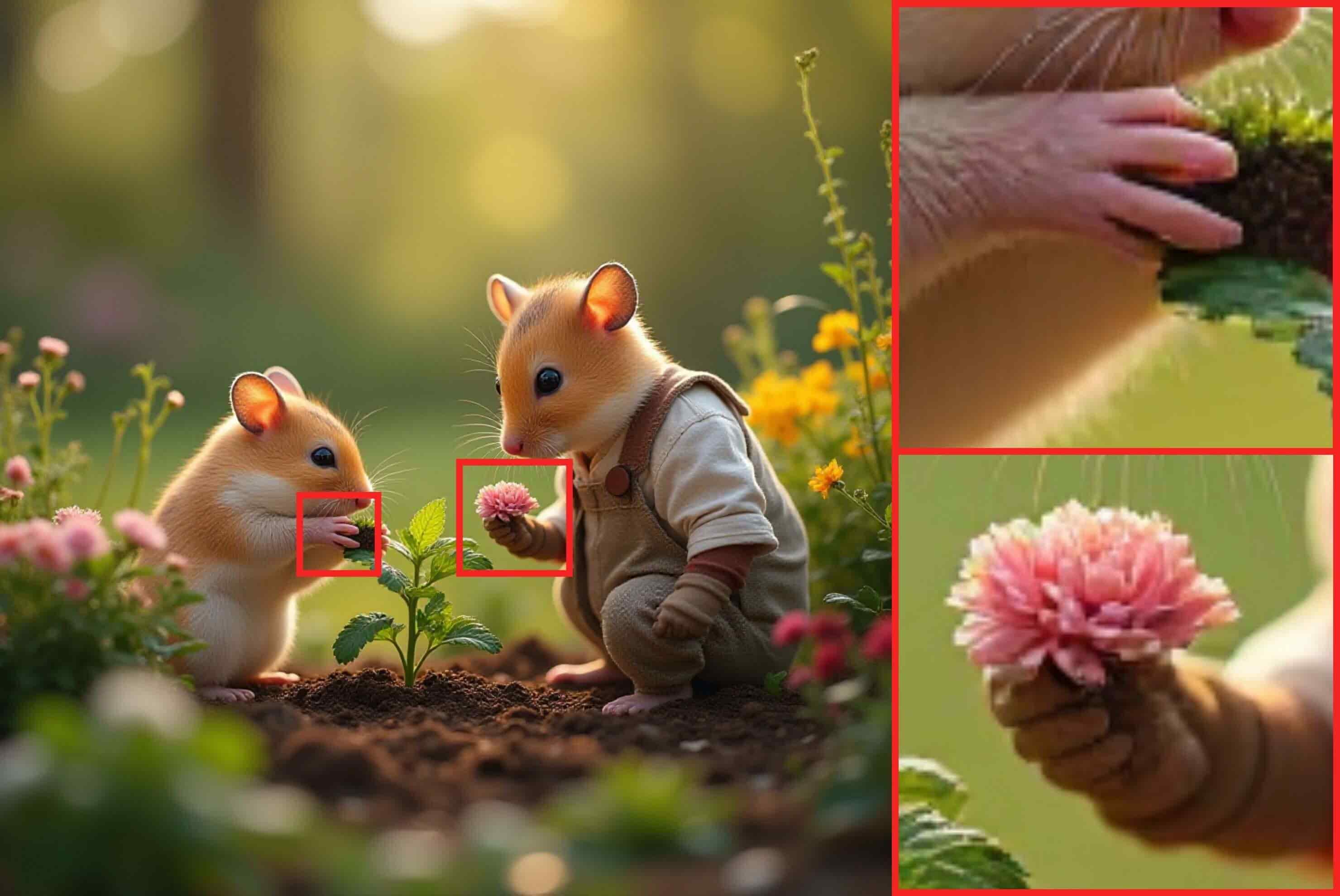} &
\includegraphics[width=0.24\textwidth]{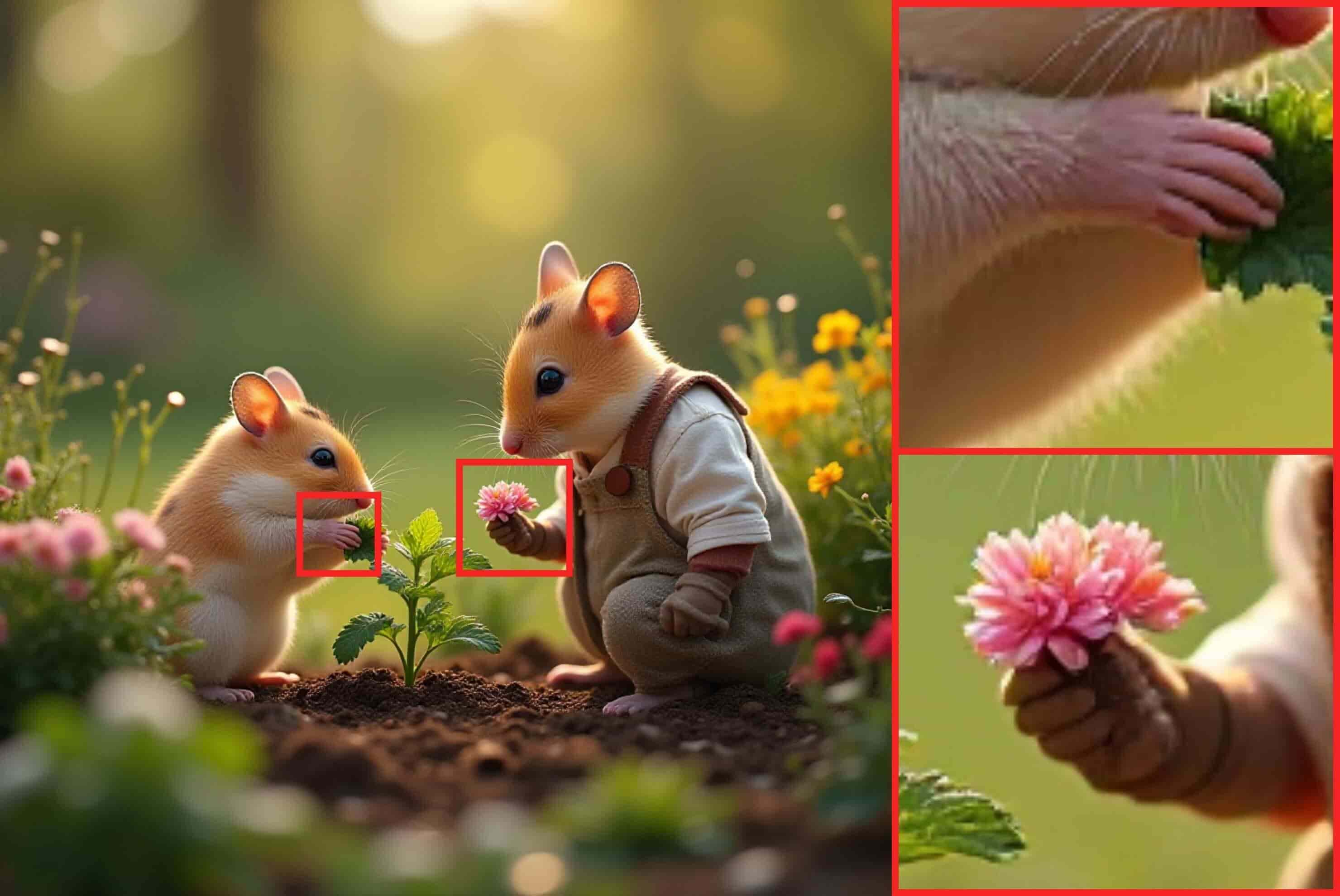} \\
\includegraphics[width=0.24\textwidth]{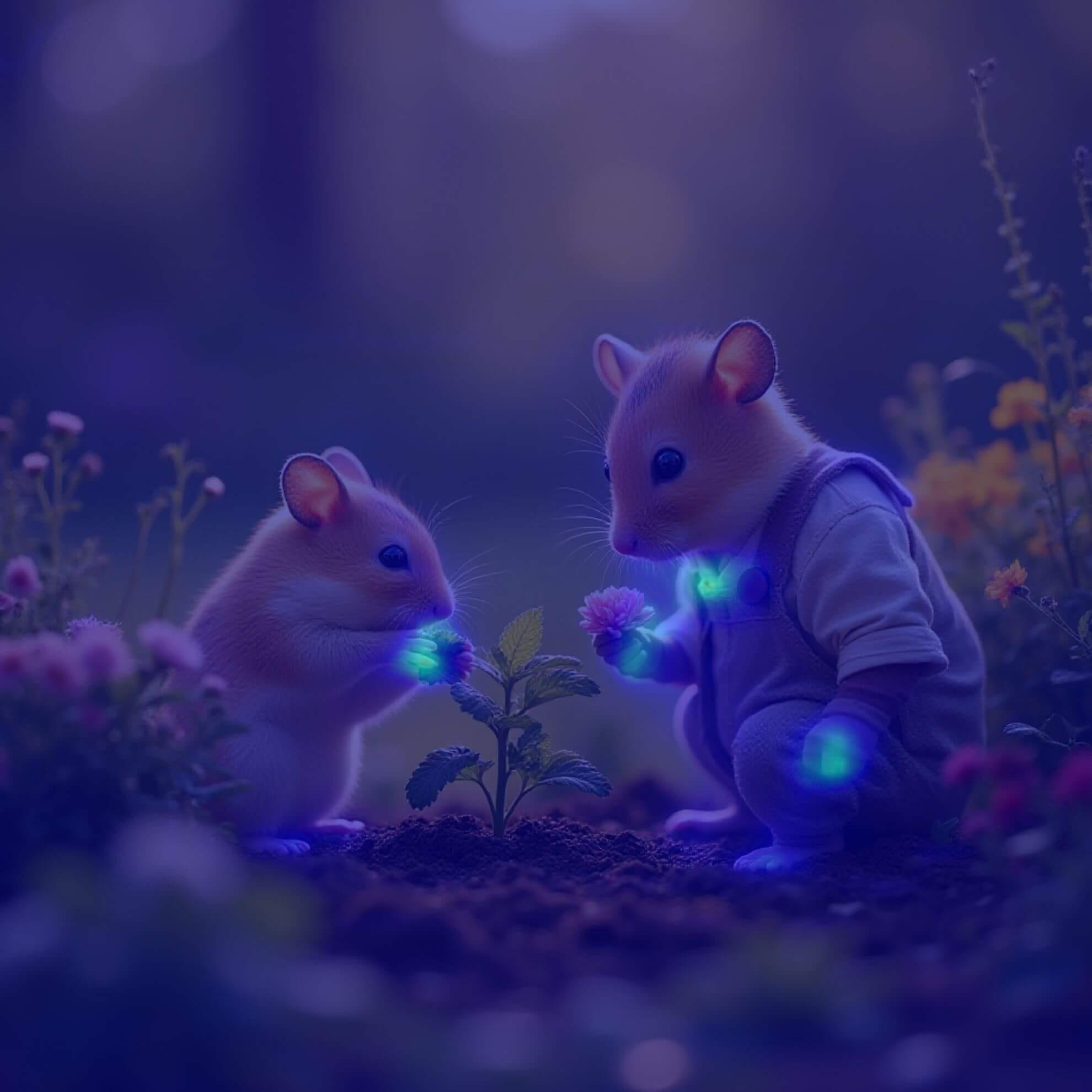} &
\includegraphics[width=0.24\textwidth]{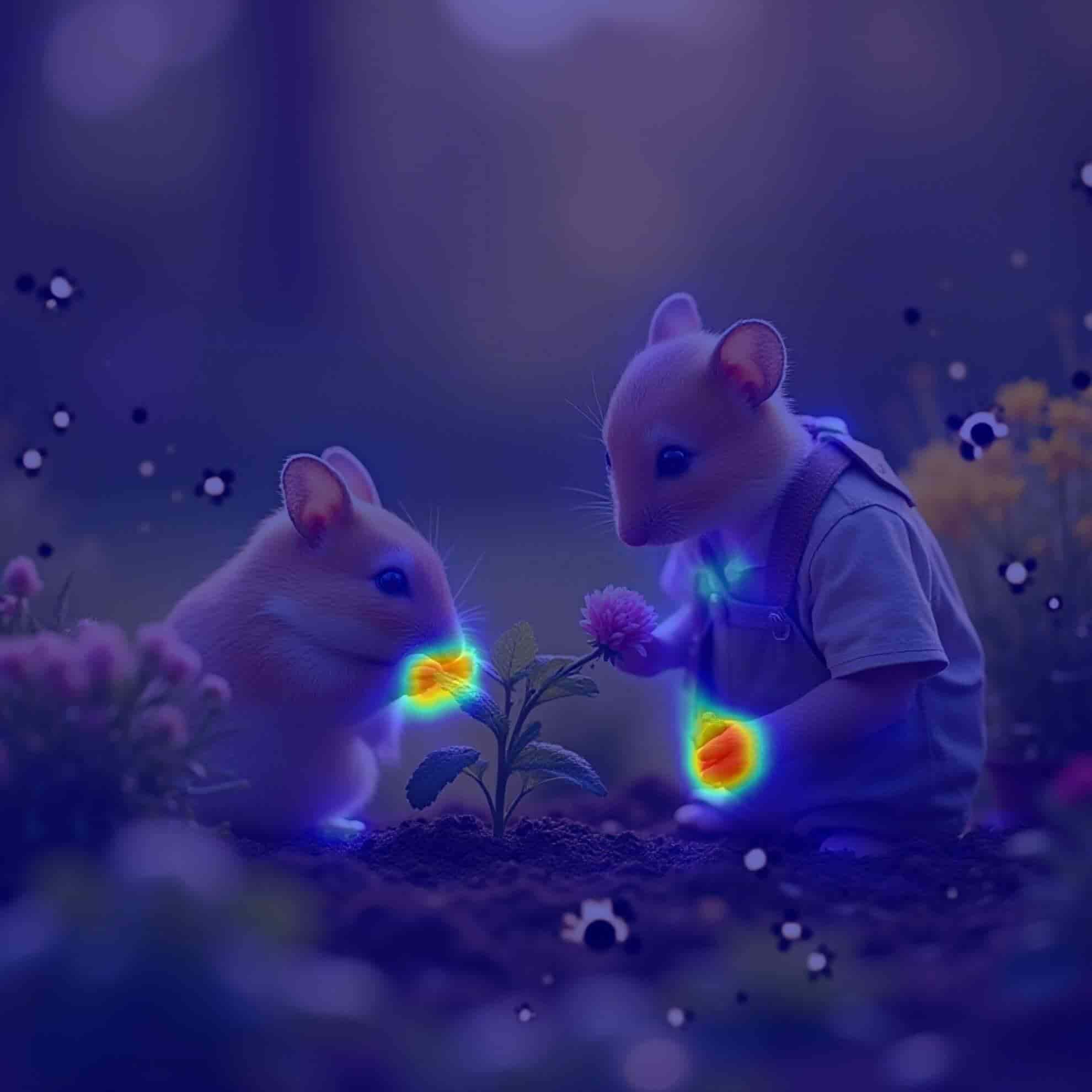} &
\includegraphics[width=0.24\textwidth]{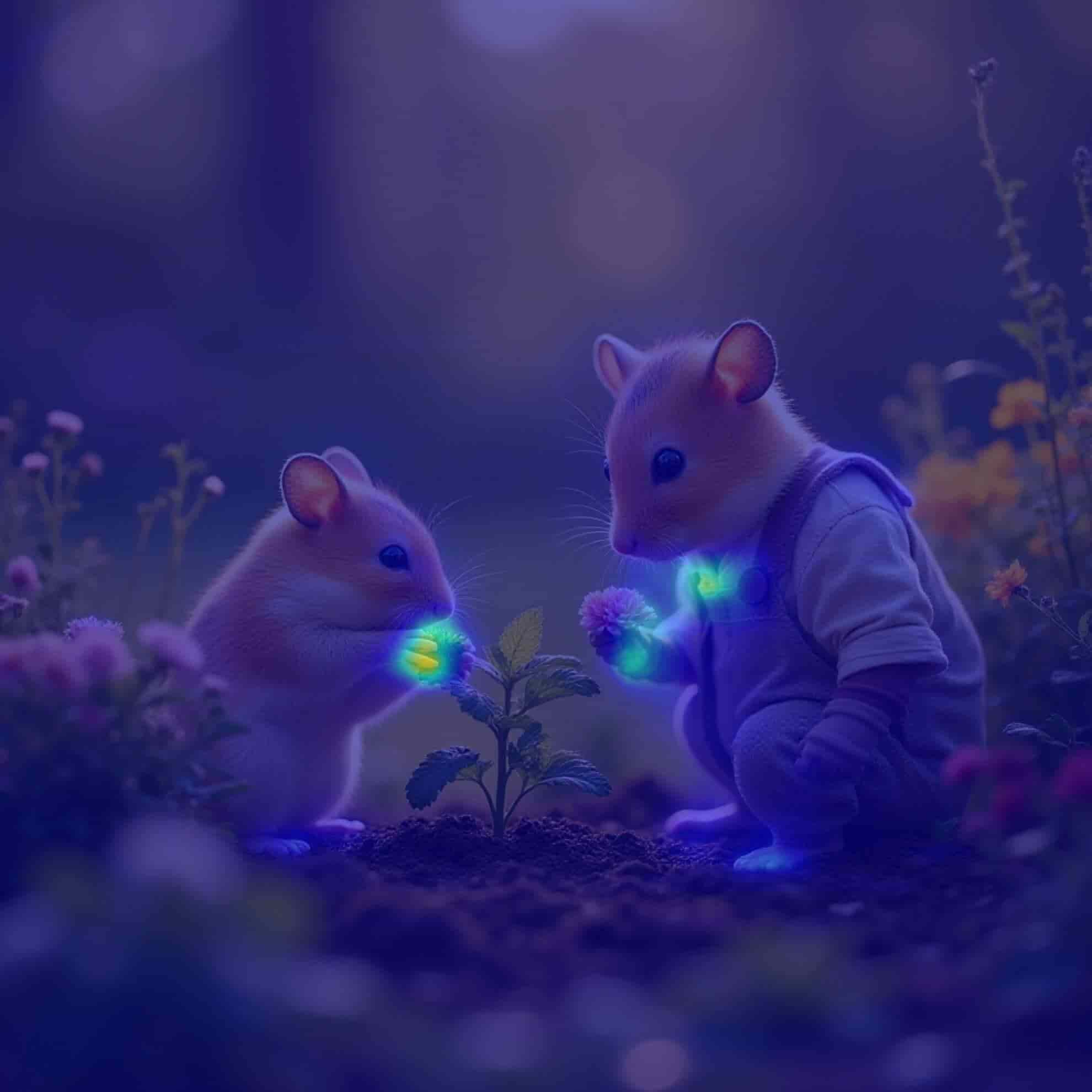} &
\includegraphics[width=0.24\textwidth]{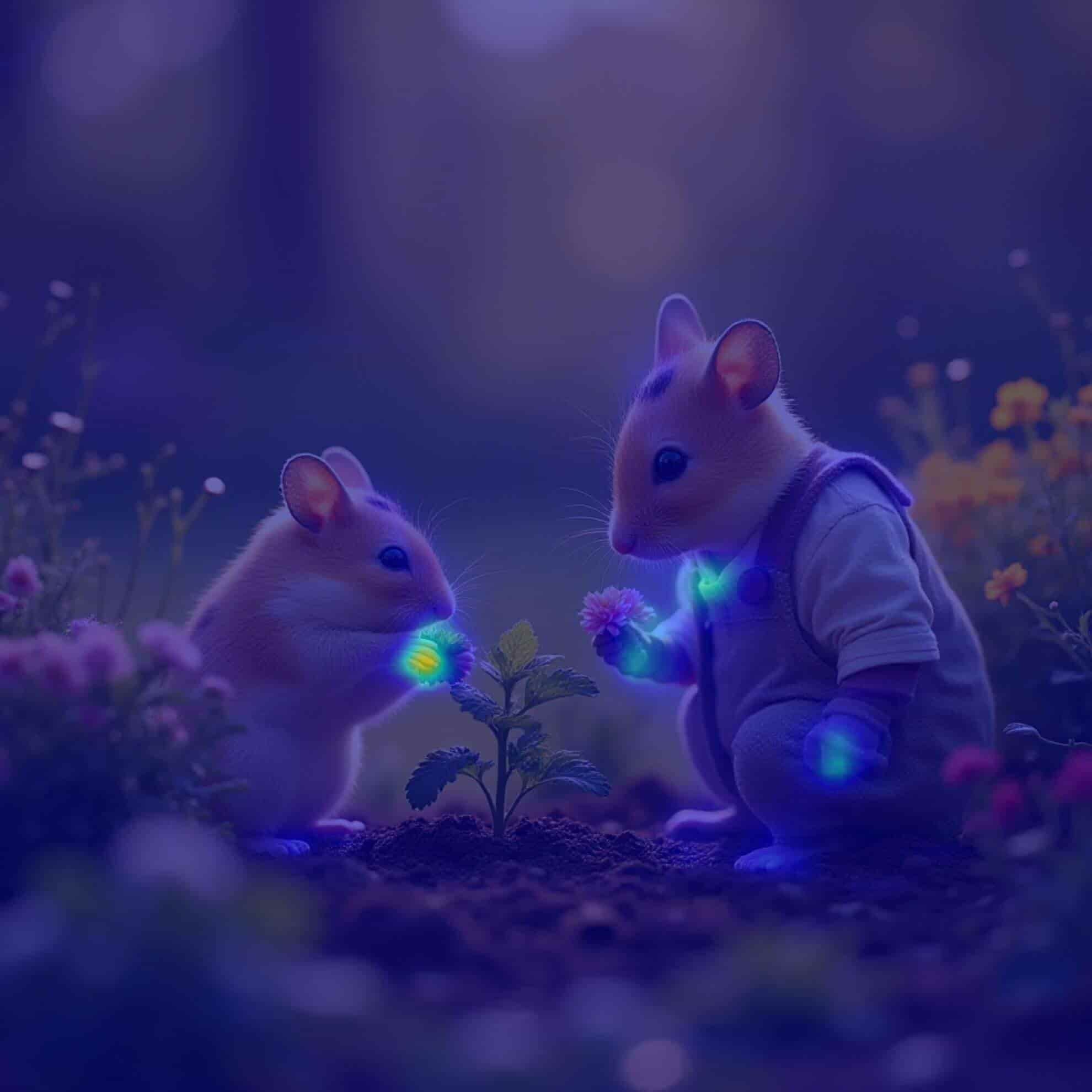} \\
\end{tabular}
\caption{\textbf{Visual Comparison of Different Exponential Schedule Configurations ($\lambda_{start}$, $\lambda_{end}$, $p$).} For large $\lambda_t$ the generated images exhibit glitches and pixel artifacts, whereas for smaller $\lambda_t$ the artifact removal effect is better and visually consistent.}
\label{fig:lambdatconstans}
\end{figure*}

Additional ablation experiments were conducted to evaluate the impact of the hyperparameters of this schedule:
\[
    \lambda_t = \lambda_{end} + (\lambda_{start} - \lambda_{end}) \left(1 - \frac{i}{N-1}\right)^p
\]
where $\lambda_{start}$ - start value, $\lambda_{end}$ - end value, $N$ - number of steps, $i$ - current step index and $p$ - power factor.
We performed an additional analysis of the \our{} on the trajectory performance and image quality using the FLUX.1 [dev] model on the \textit{people} dataset. Tab. \ref{tab:lambda} presents the numerical results for different $\lambda_t$ schedules. The results clearly indicate that maintaining too high a value of this parameter in the generation steps can lead to trajectory instability, causing image distortion and noise (low ImageReward value) (see Fig. \ref{fig:lambdatconstans}), for which the detector has trouble finding the artifact. It should be emphasized that these $\lambda_t$ values must be tuned to the specific model.

\begin{table*}[!h]
\centering
\caption{\textbf{Ablation Study of Exponential Scheduler Hyperparameters for \textit{people} dataset on FLUX.1 [dev].} All configurations are for $\tau_{start} = 0$ and $\tau_{end}= 0$. 
}
\setlength{\tabcolsep}{5.3pt}
{\fontsize{6.5pt}{11pt}\selectfont
\begin{tabular}{lllccccccc}
\hline
$\lambda_{start}$ & $\lambda_{end}$ &  $p$ &  CLIP-T $\uparrow$ & Mean Artifact Freq (\%) $\downarrow$ & ImageReward $\uparrow$ & Artfiact Pixel Ratio (\%) $\downarrow$ & MAE $\downarrow$ & MAE (A) $\downarrow$& MAE (NA) $\downarrow$\\ \hline
45 & 45 & 2 & 34.292 $\pm$ 0.231 & 3.500 $\pm$ 2.887  & 0.668 $\pm$ 0.093 & 0.019 $\pm$ 0.019 & 19.103 $\pm$ 1.003 & 49.885 $\pm$ 3.215  & 18.966 $\pm$ 1.007    \\
1  & 1  & 2 &  35.787 $\pm$ 0.202 & 26.500 $\pm$ 3.873 & 0.959 $\pm$ 0.027 & 0.122 $\pm$ 0.032 & 1.778 $\pm$ 0.081  & 8.318 $\pm$  0.356  & 1.753 $\pm$ 0.080  \\ 
\hline
45 & 1  & 2 &  35.501 $\pm$ 0.161 & 14.250 $\pm$ 0.957 & 0.891 $\pm$ 0.045 & 0.059 $\pm$ 0.022 & 14.345 $\pm$ 0.239 & 33.111 $\pm$ 0.989 & 14.261 $\pm$ 0.236 \\
25 & 1 & 2  &  35.762 $\pm$ 0.101  & 15.500 $\pm$ 2.380                & 0.968 $\pm$ 0.034        & 0.068 $\pm$ 0.022            & 9.617 $\pm$ 0.240  & 25.764 $\pm$ 0.604       & 9.545 $\pm$ 0.240   \\ 
15 & 1  & 2 & 35.710 $\pm$ 0.225 & 18.000 $\pm$ 3.559 & 0.963 $\pm$ 0.037 & 0.079 $\pm$ 0.023 & 7.122 $\pm$ 0.165  & 21.316 $\pm$ 0.944 & 7.060 $\pm$ 0.163  \\
\hline
45 & 1 & 3 & 35.500 $\pm$ 0.219  & 21.75 $\pm$ 5.560 & 0.882 $\pm$ 0.051 & 0.110 $\pm$ 0.026 & 13.719 $\pm$ 0.236 & 30.894 $\pm$ 0.869 & 13.642 $\pm$ 0.234 \\
15 & 1 & 3 & 35.772 $\pm$ 0.197  & 22.75 $\pm$ 3.500 & 0.961 $\pm$ 0.039 & 0.103 $\pm$ 0.027 & 6.868 $\pm$ 0.171  & 20.123 $\pm$ 1.237 & 6.811 $\pm$ 0.168  \\ \hline
\multicolumn{3}{l}{FLUX.1 [dev]}& 35.820 $\pm$ 0.175  & 100.000 $\pm$ 0.000               & 0.961 $\pm$ 0.035       & 0.525 $\pm$ 0.039            & -  & -       & -  \\
\hline 
\end{tabular}
}
\label{tab:lambda}
\end{table*}



\end{document}